\pdfoutput=1
\documentclass{book}

\usepackage{llm}
\newcommand{\mymacro}[1]{{#1}}

\newcommand{\defn}[1]{\mymacro{\textbf{#1}}\index{#1}}

\newcommand{\vbeta}{\mymacro{\boldsymbol{\beta}}}
\newcommand{\valpha}{\mymacro{\boldsymbol{\alpha}}}

\newcommand{\vgamma}{\mymacro{\boldsymbol{\gamma}}}

\newcommand{\inv}[1]{\mymacro{#1^{-1}}}

\newcommand{\biasVector}{\mymacro{\vb}}

\newcommand{\mean}[1]{\mymacro{\overline{#1}}}
\newcommand{\var}[1]{\mymacro{\sigma^2\left(#1\right)}}

\newcommand{\pdens}{\mymacro{p}}

\newcommand{\pdensTildeEOS}{\mymacro{\tilde{\pdens}_\eos}}

\newcommand{\loglikelihood}{\mymacro{\ell\ell}}
\newcommand{\samplespace}{\mymacro{\Omega}}
\newcommand{\eventspace}{\mymacro{\mathcal{F}}}
\newcommand{\eventspaceA}{\mymacro{\mathcal{A}}}
\newcommand{\probmeasure}{\mymacro{\mathbb{P}}}
\newcommand{\probfunction}{\mymacro{\mathbb{P}}}
\newcommand{\pushfwdMeasure}{\mymacro{\probfunction_*}}

\newcommand{\bomega}{\mymacro{\bm{\omega}}}
\newcommand{\overC}{\mymacro{\overline{C}}}
\newcommand{\overcalC}{\mymacro{\overline{\mathcal{C}}}}
\newcommand{\calC}{\mymacro{\mathcal{C}}}
\newcommand{\strCount}{\mymacro{C}}

\newcommand{\paramspace}{\mymacro{\boldsymbol{\Theta}}}
\newcommand{\param}{\mymacro{\theta}}
\newcommand{\modelparam}{\mymacro{\theta}}
\newcommand{\params}{\mymacro{\vtheta}}
\newcommand{\modelparams}{\mymacro{\vtheta}}
\newcommand{\modelparamstrue}{\mymacro{\modelparams^\star}}
\newcommand{\modelparamsopt}{\mymacro{\widehat\modelparams}}
\newcommand{\model}{\mymacro{p_{\scaleto{\modelparams}{4pt}}}}
\newcommand{\modelopt}{\mymacro{p_{\scaleto{\modelparamsopt}{4pt}}}}

\newcommand{\ind}[1]{\mathbbm{1} \left\{ #1 \right\}}
\newcommand{\floor}[1]{\lfloor #1 \rfloor}

\newcommand{\field}{\mymacro{\mathbb{F}}}
\newcommand{\vectorspace}{\mymacro{\mathbb{V}}}
\newcommand{\vspacedim}{\mymacro{D}}
\newcommand{\Z}{\mymacro{\mathbb{Z}}}
\newcommand{\Q}{\mymacro{\mathbb{Q}}}
\newcommand{\R}{\mymacro{\mathbb{R}}}
\newcommand{\Rex}{\mymacro{\overline{\R}}}
\newcommand{\C}{\mymacro{\mathbb{C}}}
\newcommand{\B}{\mymacro{\mathbb{B}}}

\newcommand{\BDD}{{\mymacro{\B^{D \times D}}}}

\newcommand{\Rplus}{\mymacro{\mathbb{R}_{+}}}
\newcommand{\Rnonnegative}{\mymacro{\mathbb{R}_{\geq 0}}}

\newcommand{\sigalgebra}{\mymacro{\sigma\text{-}\text{algebra}}}
\newcommand{\func}{\mymacro{f}}
\newcommand{\funcg}{\mymacro{g}}
\newcommand{\vfunc}{\mymacro{\mathbf{f}}}

\newcommand{\vfuncg}{\mymacro{\mathbf{g}}}

\newcommand{\norm}[1]{\mymacro{\left\lVert #1 \right\rVert}}

\newcommand{\normsubtwo}[1]{\mymacro{\left\lVert #1 \right\rVert_{2}}}

\newcommand{\eval}{\mymacro{\lambda}}

\newcommand{\evalmax}{\mymacro{\eval_{\text{max}}}}
\newcommand{\ordering}{\mymacro{n}}

\newcommand{\innerProd}[2]{\mymacro{\langle #1, #2 \rangle }}
\newcommand{\basisVect}{\mymacro{\ve}}
\newcommand{\perm}{{\mymacro{\pi}}}
\newcommand{\homomorphism}{{\mymacro{h}}}

\newcommand{\alphabet}{\mymacro{\Sigma}}

\newcommand{\stackalphabet}{\mymacro{\Gamma}}
\newcommand{\oalphabet}{\mymacro{\Delta}}
\newcommand{\outalphabet}{\oalphabet}
\newcommand{\eosalphabet}{\mymacro{\overline{\alphabet}}}
\newcommand{\lang}{\mymacro{L}}
\newcommand{\langFun}[1]{\lang\left(#1\right)}
\newcommand{\kleene}[1]{\mymacro{#1^*}}
\newcommand{\kleeneplus}[1]{\mymacro{#1^+}}

\newcommand{\compModel}{{\mymacro{\mathcal{C}}}}

\newcommand{\str}{\mymacro{\boldsymbol{y}}}
\newcommand{\strt}{\mymacro{\str_{\tstep}}}
\newcommand{\strlt}{\mymacro{\str_{<\tstep}}}
\newcommand{\strlet}{\mymacro{\str_{\leq\tstep}}}
\newcommand{\strltplus}{\mymacro{\str_{<\tstep+1}}}

\newcommand{\strPrime}{\mymacro{\str'}}
\newcommand{\eosstr}{\mymacro{\overline{\str}}}
\newcommand{\strlen}{\mymacro{T}}

\newcommand{\strx}{\mymacro{\boldsymbol{x}}}
\newcommand{\stry}{\mymacro{\boldsymbol{y}}}
\newcommand{\strz}{\mymacro{\boldsymbol{z}}}
\newcommand{\sym}{\mymacro{y}}
\newcommand{\eossym}{\mymacro{\overline{\sym}}}
\newcommand{\syma}{\mymacro{a}}
\newcommand{\symb}{\mymacro{b}}
\newcommand{\symc}{\mymacro{c}}
\newcommand{\symd}{\mymacro{d}}
\newcommand{\symx}{\mymacro{x}}
\newcommand{\stacksym}{\mymacro{\stacksymbol{\gamma}}}

\newcommand{\defeq}{\mathrel{\stackrel{\textnormal{\tiny def}}{=}}}

\newcommand{\NTo}[1]{{\mymacro{\left[ #1 \right]}}}
\newcommand{\Zmod}[1]{{\mymacro{\Z_{#1}}}}
\newcommand{\set}[1]{\mymacro{\left\{ #1 \right\}}}
\newcommand{\powerset}[1]{\mymacro{\mathcal{P}{\left(#1\right)}}}
\newcommand{\setcomplement}[1]{\mymacro{#1^{\textsf{c}}}}

\newcommand{\justification}[1]{\text{{\color{ETHGray}(#1)}}}

\newcommand{\textexample}[1]{\mymacro{\fontfamily{ppl}\selectfont \text{``#1''}}}

\newcommand{\bigo}{\mymacro{\mathcal{O}}}
\newcommand{\idx}{\mymacro{n}}
\newcommand{\idxn}{\mymacro{n}}
\newcommand{\idxd}{\mymacro{d}}
\newcommand{\idxi}{\mymacro{i}}
\newcommand{\idxj}{\mymacro{j}}
\newcommand{\idxk}{\mymacro{k}}
\newcommand{\idxl}{\mymacro{l}}
\newcommand{\idxm}{\mymacro{m}}

\newcommand{\nstates}{{\mymacro{|\states|}}}
\newcommand{\nsymbols}{{\mymacro{|\alphabet|}}}

\newcommand{\tstep}{\mymacro{t}}
\newcommand{\finaltstep}{\mymacro{T}}

\newcommand{\pM}{\mymacro{p_{\scaleto{\text{M}}{4pt}}}}
\newcommand{\pLM}{\mymacro{p_{\scaleto{\text{LM}}{4pt}}}}
\newcommand{\pLNSM}{\mymacro{p_{\scaleto{\text{SM}}{4pt}}}}
\newcommand{\pLNSMFun}[2]{{\mymacro{\pLNSM\left(#1\mid#2\right)}}}
\newcommand{\pPrefix}{\mymacro{\pi}}
\newcommand{\pGN}{\mymacro{p_{\scaleto{\text{GN}}{4pt}}}}
\newcommand{\pLN}{\mymacro{p_{\scaleto{\text{LN}}{4pt}}}}

\newcommand{\scoringfunc}{\mymacro{\widehat{p}}}
\newcommand{\unnormalizedpGN}{\mymacro{\scoringfunc_{\scaleto{\text{GN}}{4pt}}}}

\newcommand{\pSM}{\mymacro{p_{\scaleto{\text{SM}}{4pt}}}}

\newcommand{\bos}{\mymacro{\textsc{bos}}}
\newcommand{\eos}{\mymacro{\textsc{eos}}\xspace}
\newcommand{\ngr}{\mymacro{\textit{n}}}
\newcommand{\ngram}{\mymacro{\textit{n}-gram}}
\newcommand{\nStrictLocal}{\mymacro{\text{SL}_\ngr}}
\newcommand{\strictLocal}{\mymacro{\text{SL}}}
\newcommand{\normConstant}{\mymacro{Z}}

\newcommand{\embedDim}{\mymacro{R}}

\newcommand{\suffixOf}[2]{\mymacro{#1 \; \triangleleft \; #2}}

\newcommand{\dyck}[1]{{\mymacro{\mathrm{D}\!\left(#1\right)}}}

\newcommand{\nBracketTypes}{{\mymacro{k}}}

\newcommand{\openBr}[1]{{\mymacro{\langle_{#1}}}}
\newcommand{\closeBr}[1]{{\mymacro{\rangle_{#1}}}}

\newcommand{\hToQ}{{\mymacro{s}}}
\newcommand{\hToQFun}[1]{{\mymacro{\hToQ \!\left( #1 \right)}}}
\newcommand{\hToQinv}{{\mymacro{\inv{s}}}}
\newcommand{\hToQinvFun}[1]{{\mymacro{\hToQinv\!\left( #1 \right)}}}
\newcommand{\qToH}{{\mymacro{h}}}

\newcommand{\qToHinv}{{\mymacro{\inv{h}}}}

\newcommand{\onehot}[1]{\mymacro{\llbracket#1\rrbracket}}
\newcommand{\embedding}[2][]{\mymacro{\ve_{\textsf{#1}}\!\left(#2\right)}}
\newcommand{\inEmbedding}[1][]{\mymacro{\ve_{\textsf{#1}}'}}
\newcommand{\inEmbeddingFun}[2][]{\mymacro{\ve_{\textsf{#1}}'\!\left(#2\right)}}
\newcommand{\embedSym}{\mymacro{\embedding{\sym}}}

\newcommand{\inEmbedSymt}{\mymacro{\inEmbeddingFun{\sym_\tstep}}}

\newcommand{\embedMtx}{\mymacro{\mE}}

\newcommand{\embedEOS}{\mymacro{\embedding{\eos}}}

\newcommand{\embedSymt}{\mymacro{\embedding{{\sym_\tstep}}}}

\newcommand{\symt}{\mymacro{\sym_{\tstep}}}
\newcommand{\eossymt}{\mymacro{\eossym_{\tstep}}}
\newcommand{\eossymtplus}{\mymacro{\eossym_{\tstep+1}}}

\newcommand{\symtminus}{\mymacro{\sym_{\tstep-1}}}
\newcommand{\symtplus}{\mymacro{\sym_{\tstep+1}}}
\newcommand{\symzero}{\mymacro{\sym_{0}}}
\newcommand{\symone}{\mymacro{\sym_{1}}}
\newcommand{\symT}{\mymacro{\sym_{T}}}
\newcommand{\symTminus}{\mymacro{\sym_{T-1}}}

\newcommand{\bias}{\mymacro{\vb}}

\newcommand{\biasStr}{\mymacro{b_\str}}

\newcommand{\biasVech}{\mymacro{\vb_{h}}}

\newcommand{\vot}{\mymacro{\vo_{t}}}

\newcommand{\GNMAcronym}{\mymacro{GNM\xspace}}
\newcommand{\LNMAcronym}{\mymacro{LNM\xspace}}
\newcommand{\SMAcronym}{\mymacro{SM\xspace}}

\newcommand{\graph}{\mymacro{\mathcal{G}}}

\newcommand{\zero}{\mymacro{\mathbf{0}}}
\newcommand{\one}{\mymacro{\mathbf{1}}}

\newcommand{\automaton}{\mymacro{\mathcal{A}}}
\newcommand{\wfsa}{\mymacro{\automaton}}

\newcommand{\dpfsaAcr}{{\mymacro{dPFSA}}}

\newcommand{\stateq}{\mymacro{q}}
\newcommand{\statep}{\mymacro{p}}
\newcommand{\stater}{\mymacro{r}}

\newcommand{\states}{\mymacro{Q}}

\newcommand{\trans}{\mymacro{\delta}}

\newcommand{\prevq}{\mymacro{p }}
\newcommand{\nextq}{\mymacro{n }}

\newcommand{\weight}{\mymacro{\textnormal{w}}}
\newcommand{\innerweight}{\mymacro{\weight_{\text{I}}}}

\newcommand{\apath}{\mymacro{\boldsymbol \pi}}
\newcommand{\pathlen}{\mymacro{N}}
\newcommand{\paths}{\mymacro{\Pi}}

\newcommand{\initial}{\mymacro{I}}
\newcommand{\final}{\mymacro{F}}
\newcommand{\initf}{\mymacro{\lambda}}
\newcommand{\finalf}{\mymacro{\rho}}

\newcommand{\initfVect}{\mymacro{\overrightarrow{\initf}}}
\newcommand{\finalfVect}{\mymacro{\overrightarrow{\finalf}}}
\newcommand{\qinit}{\mymacro{q_{\iota}}}
\newcommand{\qfinal}{\mymacro{q_{\varphi}}}

\newcommand{\transMtx}{\mymacro{\mT}}
\newcommand{\allsum}{\mymacro{Z}}

\newcommand{\transitionWeight}{{\mymacro{\omega}}}
\newcommand{\transitionWeightFun}[1]{{\mymacro{\transitionWeight(#1)}}}

\newcommand{\fsatuple}{\mymacro{\left( \alphabet, \states, \initial, \final, \trans \right)}}
\newcommand{\wfsatuple}{\mymacro{\left( \alphabet, \states, \trans, \initf, \finalf \right)}}

\newcommand{\edgenoweight}[3]{#1 \xrightarrow{#2} #3}
\newcommand{\edge}[4]{\mymacro{#1 \xrightarrow{#2 / #3} #4}}
\newcommand{\uwedge}[3]{\mymacro{#1 \xrightarrow{#2} #3}}

\newcommand{\backward}[1][]{\mymacro{\boldsymbol{\beta}_{#1}}}

\newcommand{\backwardFun}[2][]{\mymacro{\backward{#1}\left(#2\right)}}

\newcommand{\yield}{\mymacro{\textbf{s}}}

\newcommand{\tm}{\mymacro{\mathcal{M}}}

\newcommand{\dyMap}{\mymacro{\vfunc}}
\newcommand{\generalrnntuple}{\mymacro{\left( \alphabet, \hiddDim, \dyMap, \hiddStateZero\right)}}
\newcommand{\rnntuple}{\mymacro{\left( \alphabet, \hiddDim, \dyMap, \outMtx, \hiddStateZero\right)}}
\newcommand{\elmanrnntuple}{\mymacro{\left( \alphabet, \hiddDim, \recMtx, \inMtx, \outMtx, \biasVech, \hiddStateZero\right)}}
\newcommand{\elmanUpdate}[2]{{\mymacro{\sigmoid\left(\recMtx #1 + \inMtx \onehot{#2} + \biasVech\right)}}}
\newcommand{\rnn}{\mymacro{\mathcal{R}}}

\newcommand{\hernnAcr}{{\mymacro{HRNN}}}
\newcommand{\recMtx}{\mymacro{\mU}}
\newcommand{\inMtx}{\mymacro{\mV}}
\newcommand{\outMtx}{\mymacro{\mE}}

\newcommand{\eRecMtx}{\mymacro{\emU}}
\newcommand{\eInMtx}{\mymacro{\emV}}
\newcommand{\hiddDim}{\mymacro{D}}
\newcommand{\gate}{\mymacro{\vg}}
\newcommand{\gatet}{\mymacro{\gate_\tstep}}

\newcommand{\hiddState}{\mymacro{\vh}}
\newcommand{\hiddStatet}{\mymacro{\vh_\tstep}}
\newcommand{\hiddStateT}{{\mymacro{\hiddState_\strlen}}}
\newcommand{\hiddStateTminus}{{\mymacro{\hiddState_{\strlen - 1}}}}
\newcommand{\hiddStatetplus}{{\mymacro{\hiddState_{\tstep + 1}}}}
\newcommand{\hiddStatetminus}{{\mymacro{\hiddState_{\tstep - 1}}}}

\newcommand{\vhzero}{\mymacro{\vh_0}}

\newcommand{\hiddStateZero}{\mymacro{\vhzero}}

\newcommand{\vht}{\mymacro{\vh_t}}

\newcommand{\vhtminus}{\mymacro{\vh_{t-1}}}

\newcommand{\symordering}{{\mymacro{m}}}
\newcommand{\eossymordering}{{\mymacro{\overline{\symordering}}}}
\newcommand{\stateordering}{{\mymacro{r}}}
\newcommand{\anOrdering}{{\mymacro{s}}}

\newcommand{\phiCoor}{{\mymacro{\vvarphi}}}
\newcommand{\phitwo}{{\mymacro{\phi_2}}}
\newcommand{\phifour}{{\mymacro{\phi_4}}}
\newcommand{\phitwoFun}[1]{{\mymacro{\phitwo\left(#1\right)}}}
\newcommand{\phifourFun}[1]{{\mymacro{\phifour\left(#1\right)}}}
\newcommand{\phitwoInv}{{\mymacro{\inv{\phi_2}}}}
\newcommand{\phifourInv}{{\mymacro{\inv{\phi_4}}}}
\newcommand{\phitwoInvFun}[1]{{\mymacro{\phitwoInv\left(#1\right)}}}
\newcommand{\phifourInvFun}[1]{{\mymacro{\phifourInv\left(#1\right)}}}

\newcommand{\srNstates}{{\mymacro{s}}}
\newcommand{\frNstates}{{\mymacro{r}}}
\newcommand{\BQQ}{{\mymacro{\B^{\nstates \times \nstates}}}}
\newcommand{\BtwoQ}{{\mymacro{\B^{2 \nstates}}}}
\newcommand{\Bss}{{\mymacro{\B^{\srNstates \times \srNstates}}}}
\newcommand{\Btwos}{{\mymacro{\B^{2 \srNstates}}}}
\newcommand{\qvectv}[1]{{\mymacro{\vv\left(#1\right)}}}

\newcommand{\stateSet}{{\mymacro{\mathcal{Q}}}}
\newcommand{\qPreds}[1]{{\mymacro{\texttt{Pred}\left(#1\right)}}}
\newcommand{\qt}{{\mymacro{\stateq_\tstep}}}
\newcommand{\qT}{{\mymacro{\stateq_\strlen}}}
\newcommand{\qTminus}{{\mymacro{\stateq_\strlen}}}

\newcommand{\qtplus}{{\mymacro{\stateq_{\tstep + 1}}}}
\newcommand{\balpha}{{\mymacro{\boldsymbol{\alpha}}}}
\newcommand{\nLines}[1]{{\mymacro{L\left(#1\right)}}}
\newcommand{\compMtx}[2]{{\mymacro{\Phi_{#1, #2}}}}

\newcommand{\grammar}{\mymacro{\mathcal{G}}}
\newcommand{\nonterm}{\mymacro{\mathcal{N}}}
\newcommand{\rules}{\mymacro{\mathcal{P}}}
\newcommand{\arule}{\mymacro{p}}
\newcommand{\cfgstr}{\mymacro{\valpha}} 
\newcommand{\start}{\mymacro{\NT{S}}}
\newcommand{\cfgtuple}{\mymacro{\left(\alphabet, \nonterm, \start, \rules \right)}}
\newcommand{\derivationset}[2]{\mymacro{\mathcal{D}_{#1}(#2)}}
\newcommand{\grammarDerivationset}[1]{\mymacro{\mathcal{D}_{#1}}}
\newcommand{\tree}{\mymacro{\boldsymbol{d}}}

\newcommand{\production}[2]{\mymacro{#1 \rightarrow #2}}
\newcommand{\wproduction}[3]{\mymacro{#1 \xrightarrow {#3} #2}}
\newcommand{\derives}{\mymacro{\overset{*}{\Rightarrow}}}
\newcommand{\derivesbase}{\mymacro{\Rightarrow}}
\newcommand{\wcfgtuple}{(\alphabet, \nonterm, \start, \rules, \productionWeight)}

\newcommand{\height}[1]{\textsf{height}(#1)}

\newcommand{\NT}[1]{\mymacro{\mathrm{#1}}}

\newcommand{\NTX}{\NT{X}}
\newcommand{\NTY}{\NT{Y}}

\newcommand{\productionWeight}{\mymacro{\mathcal{W}}}
\newcommand{\genFunc}{\mymacro{G}}

\newcommand{\enc}{\mymacro{{\displaystyle \mathrm{enc}}}}
\newcommand{\encfunc}[1]{\enc\left(#1\right)}
\newcommand{\encRNN}{\mymacro{{\enc_{\rnn}}}}
\newcommand{\encRNNfunc}[1]{\mymacro{\encRNN\!\left(#1\right)}}

\newcommand{\Rd}{\mymacro{\mathbb{R}^d}}

\newcommand{\Simplextminus}{\mymacro{\boldsymbol{\Delta}^{t-1}}}

\newcommand{\Simplexdminus}{\mymacro{\boldsymbol{\Delta}^{D-1}}}

\newcommand{\SimplexEosalphabetminus}{\mymacro{\boldsymbol{\Delta}^{|\eosalphabet|-1}}}
\newcommand{\ent}{\mymacro{\mathrm{H}}}

\newcommand{\mtxDim}{\mymacro{N}}
\newcommand{\spectRad}{\mymacro{\rho_{\text{s}}}}

\DeclareMathSymbol{\mlq}{\mathord}{operators}{``} 
\DeclareMathSymbol{\mrq}{\mathord}{operators}{`'} 

\newcommand{\negterm}[1]{\mymacro{{\raise.17ex\hbox{$\scriptstyle\sim$}} #1}}

\newcommand{\algoname}[1]{\textsf{\smaller\color{blue!50!black}#1}}
\newcommand{\ifcondition}{\textbf{if }}
\newcommand{\otherwisecondition}{\textbf{otherwise }}

\newcommand{\separationAlgoName}{{\algoname{Separate}}}

\newcommand{\pdatuple}{\left(\alphabet, \states,  \stackalphabet, \trans, \initconfig, \finalconfig\right)}

\newcommand{\twostackpdatuple}{\left(\states, \alphabet, \stackalphabet, \trans, \pdaconfig{\initstack,\initstacktwo}{\qinit}, \pdaconfig{\finalstack,\finalstacktwo}{\qfinal}\right)}

\newcommand{\pushdown}{\mymacro{\mathcal{P}}}
\newcommand{\pda}{\mymacro{\pushdown}}

\newcommand{\pdaEdge}[6]{\mymacro{\edge{#2}{#3, #5 \rightarrow #6}{#1}{#4}}}
\newcommand{\twoPdaEdge}[8]{\mymacro{\edge{#2}{#3, #5 \rightarrow #6, #7 \rightarrow #8}{#1}{#4}}}
\newcommand{\pdaEdgenoweight}[5]{\mymacro{\edgenoweight{#1}{#2, #4 \rightarrow #5}{#3}}}
\newcommand{\arun}{\mymacro{\apath}}
\newcommand{\runs}{\mymacro{\paths}}
\newcommand{\stackseq}{\mymacro{{\boldsymbol{\gamma}}}}
\newcommand{\stackseqtwo}{\mymacro{{\boldsymbol{\sigma}}}}
\newcommand{\emptystack}{\mymacro{{\emptyset}}}
\newcommand{\initstack}{\mymacro{{\stackseq_{\iota}}}}
\newcommand{\finalstack}{\mymacro{{\stackseq_{\varphi}}}}
\newcommand{\initstacktwo}{\mymacro{{\stackseqtwo_{\iota}}}}
\newcommand{\finalstacktwo}{\mymacro{{\stackseqtwo_{\varphi}}}}
\newcommand{\initconfig}{\mymacro{ \pdaconfig{\initstack}{\qinit}}}
\newcommand{\finalconfig}{\mymacro{ \pdaconfig{\finalstack}{\qfinal}}}
\newcommand{\pdaconfig}[2]{\mymacro{\left(#2, #1\right)}}
\newcommand{\twoPdaconfig}[3]{\mymacro{\left(#3, #1, #2\right)}}

\newcommand{\stacksymbol}[1]{\mymacro{#1 }}
\newcommand{\transweight}[5]{
{\mymacro{\tau(#1,\stacksymbol{#4}\xrightarrow{#2}#3,\stacksymbol{#5})}}}
\newcommand{\atrans}{\mymacro{\tau}}
\newcommand{\pdaAction}{\mymacro{a}}
\newcommand{\pushOp}{\mymacro{\texttt{PUSH}}}
\newcommand{\popOp}{\mymacro{\texttt{POP}}}
\newcommand{\noOp}{\mymacro{\texttt{NO-OP}}}

\newcommand{\stackCell}{\mymacro{\texttt{STACK}}}

\newcommand{\bufferCell}[1]{\mymacro{\texttt{BUFF}_{#1}}}
\newcommand{\acceptCell}{\mymacro{\texttt{ACCEPT}}}
\newcommand{\stackTopCell}[1]{\mymacro{\texttt{STACK}_{#1}}}
\newcommand{\stackTopCellEmpty}{\mymacro{\texttt{STACK}_{\eps}}}
\newcommand{\stackTopCellZero}{\mymacro{\texttt{STACK}_{0}}}
\newcommand{\stackTopCellOne}{\mymacro{\texttt{STACK}_{1}}}
\newcommand{\stackInputConfig}[2]{\mymacro{\texttt{CONF}_{#1, #2}}}
\newcommand{\stackComp}[2]{\mymacro{\texttt{OP}_{#1, #2}}}
\newcommand{\stackCompNoOp}{\mymacro{\texttt{OP}_{\noOp}}}
\newcommand{\stackToNumeric}[1]{\mymacro{\texttt{rep}\left(#1\right)}}
\newcommand{\stackSymToDigit}[1]{\mymacro{\texttt{digit}\left(#1\right)}}

\newcommand{\ignore}[1]{}
\newcommand{\expandLater}[1]{}

\newcommand{\tfslen}{\mymacro{\finaltstep}}
\newcommand{\tfheadnum}{\mymacro{H}}

\newcommand{\qTransf}{\mymacro{Q}}
\newcommand{\kTransf}{\mymacro{K}}
\newcommand{\vTransf}{\mymacro{V}}
\newcommand{\oTransf}{\mymacro{O}}
\newcommand{\fTransf}{\mymacro{F}}

\newcommand{\qMtx}{\mymacro{\mW_Q}}
\newcommand{\kMtx}{\mymacro{\mW_K}}
\newcommand{\vMtx}{\mymacro{\mW_V}}

\newcommand{\attn}{\mymacro{\texttt{Att}}}
\newcommand{\attnBlock}{\mymacro{\texttt{A}}}
\newcommand{\attnBlockMH}{\mymacro{\texttt{MH-A}}}
\newcommand{\attnfunc}{\mymacro{\vfunc_\texttt{Att}}}
\newcommand{\tfheadCombine}{\mymacro{\vfunc_{\textnormal{H}}}}

\newcommand{\transformernetwork}{\mymacro{\mathcal{T}}}
\newcommand{\tfencfun}{\mymacro{{\enc_{\transformernetwork}}}}
\newcommand{\transformertuple}{\mymacro{(\alphabet, \hiddDim, \tfencfun)}}
\newcommand{\tfseqmodel}{\mymacro{(\alphabet, \hiddDim, \tfencfun, \embedMtx)}}

\newcommand{\tfscorefun}{\mymacro{\func}}
\newcommand{\tfembfun}{\mymacro{\vr}}
\newcommand{\hardmax}{\mymacro{\mathrm{hardmax}}}
\newcommand{\hardmaxAvg}{\mymacro{\hardmax_\textnormal{avg}}}
\newcommand{\hardmaxUni}{\mymacro{\hardmax_\textnormal{uni}}}

\newcommand{\tflayer}{\mymacro{\mathrm{\textsf{T}}}}
\newcommand{\tflayerinputmat}{\mymacro{\mX}}
\newcommand{\tflayerinputsy}{\mymacro{\vx}}
\newcommand{\tflayerinputseq}{\mymacro{(\tflayerinputsy_1^{\top}, \tflayerinputsy_2^{\top}, \dots, \tflayerinputsy_\tfslen^{\top})}}
\newcommand{\tflayeroutputmat}{\mymacro{\mZ}}
\newcommand{\tflayeroutputsy}{\mymacro{\vz}}
\newcommand{\tflayeroutputseq}{\mymacro{(\tflayeroutputsy_1^{\top}, \tflayeroutputsy_2^{\top}, \dots, \tflayeroutputsy_\tfslen^{\top})}}
\newcommand{\tflayeridx}{\mymacro{\ell}}
\newcommand{\tfnumlayer}{\mymacro{L}}

\newcommand{\posEnc}{\mymacro{\vfunc_\texttt{pos}}}
\newcommand{\posEncFun}[1]{\mymacro{\posEnc\left(#1\right)}}
\newcommand{\posInEmbedding}{\mymacro{\inEmbedding_\texttt{pos}}}
\newcommand{\posInEmbeddingFun}[1]{\mymacro{\inEmbedding_\texttt{pos}\left(#1\right)}}

\newcommand{\layerNorm}{\mymacro{\texttt{LN}}}
\newcommand{\layerNormFun}[1]{\mymacro{\layerNorm\left(#1\right)}}

\newcommand{\extAlphabet}{\mymacro{\Delta}}
\newcommand{\actionToNum}[1]{\mymacro{\texttt{d}\left(#1\right)}}

\def\ceil#1{\lceil #1 \rceil}
\def\1{\mathbf{1}}
\newcommand{\dataset}{\mymacro{\mathcal{D}}}

\def\eps{\mymacro{\varepsilon}}

\def\rr{{\mymacro{\textnormal{r}}}}

\def\rv{{\mymacro{\textnormal{v}}}}

\def\rx{{\mymacro{\textnormal{x}}}}

\def\rvd{{\mymacro{\mathbf{d}}}}

\def\rvp{{\mymacro{\mathbf{p}}}}

\def\rvx{{\mymacro{\mathbf{x}}}}

\def\ervp{{\mymacro{\textnormal{p}}}}

\def\vtheta{{\mymacro{\boldsymbol{\theta}}}}

\def\vvarphi{{{\mymacro{\boldsymbol{\varphi}}}}}
\def\va{{\mymacro{\mathbf{a}}}}
\def\vb{{\mymacro{\mathbf{b}}}}

\def\vd{{\mymacro{\mathbf{d}}}}
\def\ve{{\mymacro{\mathbf{e}}}}

\def\vg{{\mymacro{\mathbf{g}}}}
\def\vh{{\mymacro{\mathbf{h}}}}

\def\vk{{\mymacro{\mathbf{k}}}}

\def\vo{{\mymacro{\mathbf{o}}}}
\def\vp{{\mymacro{\mathbf{p}}}}
\def\vq{{\mymacro{\mathbf{q}}}}
\def\vr{{\mymacro{\mathbf{r}}}}
\def\vs{{\mymacro{\mathbf{s}}}}

\def\vu{{\mymacro{\mathbf{u}}}}
\def\vv{{\mymacro{\mathbf{v}}}}
\def\vw{{\mymacro{\mathbf{w}}}}
\def\vx{{\mymacro{\mathbf{x}}}}

\def\vz{{\mymacro{\mathbf{z}}}}

\def\eveta{{\mymacro{\eta}}}
\def\evlambda{{\mymacro{\lambda}}}

\def\evlambda{{\mymacro{\lambda}}}

\def\evvarphi{{{\mymacro{\varphi}}}}

\def\eva{{\mymacro{a}}}
\def\evb{{\mymacro{b}}}

\def\evg{{\mymacro{g}}}

\def\evs{{\mymacro{s}}}

\def\evu{{\mymacro{u}}}
\def\evv{{\mymacro{v}}}
\def\evw{{\mymacro{w}}}
\def\evx{{\mymacro{x}}}

\def\mA{{\mymacro{\mathbf{A}}}}
\def\mB{{\mymacro{\mathbf{B}}}}

\def\mE{{\mymacro{\mathbf{E}}}}

\def\mH{{\mymacro{\mathbf{H}}}}
\def\mI{{\mymacro{\mathbf{I}}}}

\def\mK{{\mymacro{\mathbf{K}}}}
\def\mL{{\mymacro{\mathbf{L}}}}
\def\mM{{\mymacro{\mathbf{M}}}}

\def\mP{{\mymacro{\mathbf{P}}}}
\def\mQ{{\mymacro{\mathbf{Q}}}}

\def\mS{{\mymacro{\mathbf{S}}}}
\def\mT{{\mymacro{\mathbf{T}}}}
\def\mU{{\mymacro{\mathbf{U}}}}
\def\mV{{\mymacro{\mathbf{V}}}}
\def\mW{{\mymacro{\mathbf{W}}}}
\def\mX{{\mymacro{\mathbf{X}}}}

\def\mZ{{\mymacro{\mathbf{Z}}}}

\def\gL{{\mymacro{\mathcal{L}}}}

\def\gY{{\mymacro{\mathcal{Y}}}}

\def\sA{{\mymacro{\mathcal{A}}}}
\def\sB{{\mymacro{\mathcal{B}}}}
\def\sC{{\mymacro{\mathcal{C}}}}

\def\sE{{\mymacro{\mathcal{E}}}}

\def\sG{{\mymacro{\mathcal{G}}}}
\def\sH{{\mymacro{\mathcal{H}}}}

\def\sK{{\mymacro{\mathcal{K}}}}
\def\sL{{\mymacro{\mathcal{L}}}}
\def\sM{{\mymacro{\mathcal{M}}}}

\def\sS{{\mymacro{\mathcal{S}}}}
\def\sT{{\mymacro{\mathcal{T}}}}
\def\sU{{\mymacro{\mathcal{U}}}}

\def\sX{{\mymacro{\mathcal{X}}}}
\def\sY{{\mymacro{\mathcal{Y}}}}

\def\emB{\mymacro{B}}

\def\emE{\mymacro{E}}

\def\emM{\mymacro{M}}

\def\emP{\mymacro{P}}

\def\emU{\mymacro{U}}
\def\emV{\mymacro{V}}

\newcommand{\N}{\mymacro{\mathbb{N}}}
\newcommand{\Nzero}{\mymacro{\mathbb{N}_{\geq 0}}}

\newcommand{\projfunc}{\mymacro{\vfunc_{\Simplexdminus}}}
\newcommand{\projfuncEosalphabetminus}{\mymacro{\vfunc_{\SimplexEosalphabetminus}}}
\newcommand{\projfuncEosalphabetminusFunc}[1]{\mymacro{\vfunc_{\SimplexEosalphabetminus}\left(#1\right)}}
\newcommand{\tempParam}{\mymacro{\tau}}
\newcommand{\KL}{\mymacro{D_{\mathrm{KL}}}}

\newcommand{\loss}{\mymacro{\ell}}
\newcommand{\softmax}{\mymacro{\mathrm{softmax}}}
\newcommand{\sparsemax}{\mymacro{\mathrm{sparsemax}}}
\newcommand{\ReLU}{\mymacro{\mathrm{ReLU}}}
\newcommand{\softmaxfunc}[2]{\mymacro{\mathrm{softmax}\!\left(#1\right)_{#2}}} 
\newcommand{\sparsemaxfunc}[2]{\mymacro{\mathrm{sparsemax}\!\left(#1\right)_{#2}}} 

\newcommand{\heaviside}{\mymacro{H}}
\newcommand{\heavisideFun}[1]{{\mymacro{\heaviside\left(#1\right)}}}
\newcommand{\sigmoid}{\mymacro{\sigma}}
\newcommand{\sigmoidFun}[1]{{\mymacro{\sigmoid\left(#1\right)}}}

\newcommand{\constbound}{\mymacro{M}}

\DeclareMathOperator*{\argmax}{\mymacro{argmax}}
\DeclareMathOperator*{\argmin}{\mymacro{argmin}}

\newcommand{\bigO}[1]{\mymacro{\mathcal{O}\left(#1\right)}}

\toggletrue{publish-notes}

\title{
    Formal Aspects \\
    of \\ 
    Language Modeling
}
\author{Ryan Cotterell, Anej Svete, Clara Meister, \\Tianyu Liu, and Li Du}
\date{\today}

\begin{document}

\includepdf{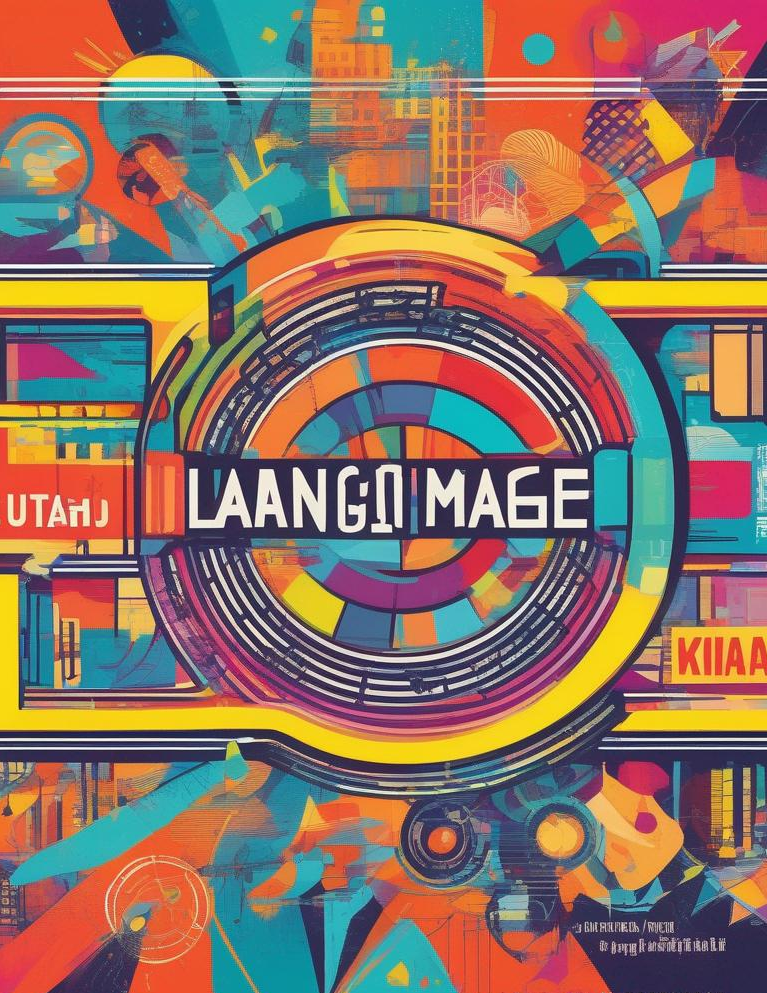}

\thispagestyle{empty}

\maketitle

\thispagestyle{empty}

\tableofcontents


\chapter{Introduction}

\section{Introduction}
Welcome to the class notes for the first third of Large Language Models (263-5354-00L).
The course comprises an omnibus introduction to language modeling.
The first third of the lectures focuses on a formal treatment of the subject.
The second part focuses on the practical aspects of implementing a language model and its applications.
Many universities are offering similar courses at the moment, e.g., CS324 at Stanford University (\url{https://stanford-cs324.github.io/winter2022/}) and CS 600.471 (\url{https://self-supervised.cs.jhu.edu/sp2023/}) at Johns Hopkins University.
Their syllabi may serve as useful references.

\paragraph{Disclaimer.}
This is the third time the course is being taught and we are improving the notes as we go.
We will try to be as careful as possible to make them typo- and error-free.
However, there will undoubtedly be mistakes scattered throughout.
We will be very grateful if you report any mistakes you spot, or anything you find unclear and confusing in general---this will benefit the students as well as the teaching staff by helping us organize a better course!

\chapter{Probabilistic Foundations} \label{chapter:prob-founds}

\section{An Invitation to Language Modeling}
The first module of the course focuses on \emph{defining} a language model mathematically.
To see why such a definition is nuanced, we are going to give an informal definition of a language model and demonstrate two ways in which that definition breaks and fails to meet our desired criteria.
\begin{definition}{Language Model (Informal)}{language-model-informal}
    Given an alphabet\footnote{An alphabet is a finite, non-empty set. It is also often referred to as a vocabulary.} $\alphabet$ and a distinguished \underline{e}nd-\underline{o}f-\underline{s}equence symbol $\eos \not\in \alphabet$, a language model is a collection of conditional probability distributions $\pdens(\sym \mid \str)$ for $\sym \in \alphabet \cup \{\eos\}$ and $\str \in \kleene{\alphabet}$, where $\kleene{\alphabet}$ is the set of all strings over the alphabet $\alphabet$.
    The term $\pdens(\sym \mid \str)$ represents the probability of the symbol $\sym$ occurring as the next symbol after the string $\str$.
\end{definition}
\cref{def:language-model-informal} is the definition of a language model that is implicitly assumed in most papers on language modeling.
We say implicitly since most technical papers on language modeling simply write down the following autoregressive factorization
\begin{equation}\label{def:autoregressive-lm}
    \pdens(\str) = \pdens(\sym_1\cdots \sym_\strlen) = \pdens(\eos \mid \str) \prod_{\tstep=1}^\strlen \pdens(\sym_\tstep \mid \str_{<\tstep})
\end{equation}
as the probability of a string according to the distribution $\pdens$.\footnote{Many authors (erroneously) avoid writing $\eos$ for concision.}
The part that is left implicit in \cref{def:autoregressive-lm} is whether or not $\pdens$ is indeed a probability distribution and, if it is, over what space.
The natural assumption in \cref{def:language-model-informal} is that $\pdens$ is a distribution over $\kleene{\alphabet}$, i.e., the set of all \emph{finite} strings\footnote{Some authors assert that strings are by definition finite.} over an alphabet $\alphabet$.
However, in general, it is not true that all such collections of conditionals will yield a valid probability distribution over $\kleene{\alphabet}$; some may ``leak'' probability mass to infinite sequences.\footnote{However, the converse \emph{is} true: All valid distributions over $\kleene{\alphabet}$ may be factorized as the above.}
More subtly, we additionally have to be very careful when dealing with uncountably infinite spaces lest we run into a classic paradox.
We highlight these two issues with two very simple examples.
The first example is a well-known paradox in probability theory.
\begin{example}{Infinite Coin Toss}{inf-coin-toss}
    Consider the infinite independent fair coin toss model, where we aim to place a distribution over $\{\texttt{H},\texttt{T}\}^\infty$, the (uncountable) set of infinite sequences of $\{\texttt{H},\texttt{T}\}$ (\texttt{H} represents the event of throwing heads and \texttt{T} the event of throwing tails).
    Intuitively, such a distribution corresponds to a ``language model'' as defined above in which for all $\str_{<\tstep}$, $\pdens(\texttt{H} \mid \str_{<\tstep}) = \pdens(\texttt{T} \mid \str_{<\tstep}) = \frac{1}{2}$ and $\pdens(\eos \mid \str_{<\tstep})=0$.
    However, each individual infinite sequence over $\{\texttt{H},\texttt{T}\}$ should also be assigned probability $(\frac{1}{2})^\infty = 0$.
    Without a formal foundation, one arrives at the following paradox:\looseness=-1
    \begin{align*}
        1 & =\pdens\left(\{\texttt{H},\texttt{T}\}^\infty\right)                                 \\
          & =\pdens\left(\bigcup_{\bomega\in\{\texttt{H},\texttt{T}\}^\infty} \{\bomega\}\right) \\
          & =\sum_{\bomega\in\{\texttt{H},\texttt{T}\}^\infty}\pdens(\{\bomega\})                \\
          & = \sum_{\bomega\in\{\texttt{H},\texttt{T}\}^\infty} 0 \stackrel{?}{=} 0. \nonumber
    \end{align*}
\end{example}
The second example is more specific to language modeling.
As we stated above, an implicit assumption made by most language modeling papers is that a language model constitutes a distribution over $\kleene{\alphabet}$.
However, in our next example, we show that a collection of conditions that satisfy \cref{def:language-model-informal} may not sum to 1 if the sum is restricted to elements of $\kleene{\alphabet}$.
This means that it is not a priori clear what space our probability distribution is defined over.\footnote{This also holds for the first example.}
\begin{figure}[h]
    \centering
    \begin{tikzpicture}[minimum size=7mm]
        \node[state, initial] (q0) {$0/1$};
        \node[state,above right=0.5cm and 0.8cm of q0] (q1) {$1$};
        \node[state,below right=0.5cm and 0.8cm of q0,accepting] (q2) {$2/\frac{1}{2}$};

        \draw[transition] (q0) edge [bend left] node [above, sloped] {$\texttt{H}/\frac{1}{2}$} (q1);
        \draw[transition] (q0) edge [bend right] node [below, sloped] {$\texttt{T}/\frac{1}{2}$} (q2);
        \draw[transition] (q1) edge [loop right] node [right] {$\texttt{H}/1$} (q1);
        \draw[transition] (q2) edge [loop right] node [right] {$\texttt{T}/\frac{1}{2}$} (q2);
    \end{tikzpicture}
    \caption{Graphical depiction of the possibly finite coin toss model. The final weight $\frac{1}{2}$ of the state $2$ corresponds to the probability $\pdens\left(\eos \mid \sym_{\tstep - 1} = \texttt{T} \right) = \frac{1}{2}$.}
    \label{fig:possibly-finite-coin-toss}
\end{figure}
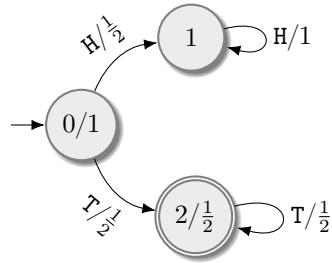
\begin{example}{Possibly Finite Coin Toss}{possibly-inf-coin-toss}
    Consider now the possibly finite ``coin toss'' model with a rather peculiar coin: when tossing the coin for the first time, both \texttt{H} and \texttt{T} are equally likely.
    After the first toss, however, the coin gets stuck: If $\sym_1 = \texttt{H}$, we can only ever toss another \texttt{H} again, whereas if $\sym_1 = \texttt{T}$, the next toss can result in another \texttt{T} or ``end'' the sequence of throws ($\eos$) with equal probability.
    We, therefore, model a probability distribution over $\kleene{\left\{\texttt{H}, \texttt{T}\right\}} \cup \left\{\texttt{H}, \texttt{T}\right\}^\infty$, the set of finite and infinite sequences of tosses.
    Formally:\footnote{Note that $\pdens(\texttt{H} \mid \str_{<1}) = \pdens(\texttt{H} \mid \varepsilon)$ and $ \pdens(\texttt{T} \mid \str_{<1}) = \pdens(\texttt{T} \mid \varepsilon)$.}
    \begin{align*}
        \pdens(\texttt{H} \mid \str_{<1}) = \pdens(\texttt{T} \mid \str_{<1}) & = \frac{1}{2} \\
        \pdens(\texttt{H} \mid \str_{<\tstep})                                & =
        \begin{cases}
            1 & \text{ if } \tstep > 1 \text{ and } \sym_{\tstep - 1} = \texttt{H} \\
            0 & \text{ if } \tstep > 1 \text{ and } \sym_{\tstep - 1} = \texttt{T}
        \end{cases}                \\
        \pdens(\texttt{T} \mid \str_{<\tstep})                                & =
        \begin{cases}
            \frac{1}{2} & \text{ if } \tstep > 1 \text{ and } \sym_{\tstep - 1} = \texttt{T} \\
            0           & \text{ if } \tstep > 1 \text{ and } \sym_{\tstep - 1} = \texttt{H}
        \end{cases}      \\
        \pdens(\eos \mid \str_{<\tstep})                                      & =
        \begin{cases}
            \frac{1}{2} & \text{ if } \tstep > 1 \text{ and } \sym_{\tstep - 1} = \texttt{T} \\
            0           & \text{ otherwise.}
        \end{cases}
    \end{align*}
    If you are familiar with (probabilistic) finite-state automata,\footnote{They will be formally introduced in \cref{sec:n-gram-models}} you can imagine the model as depicted in \cref{fig:possibly-finite-coin-toss}.
    It is easy to see that this model only places the probability of $\frac{1}{2}$ on \emph{finite} sequences of tosses.
    If we were only interested in those (analogously to how we are only interested in finite strings when modeling language), yet still allowed the model to specify the probabilities as in this example, the resulting probability distribution would not model what we require.
\end{example}

It takes some mathematical heft to define a language model in a manner that avoids such paradoxes.
The tool of choice for mathematicians is measure theory, as it allows us to define
probability over uncountable sets\footnote{As stated earlier, $\{\texttt{H},\texttt{T}\}^\infty$ is uncountable. It's easy to see there exists a surjection from $\{\texttt{H},\texttt{T}\}^\infty$ to the binary expansion of the real interval $(0, 1]$. Readers who are interested in more details and mathematical implications can refer to \S1 in \cite{Billingsley1986}.} in a principled way.
Thus, we begin our formal treatment of language modeling with a primer of measure theory in \cref{sec:measure-theory}.
Then, we will use concepts discussed in the primer to work up to a formal definition of a language model.

\newpage{}

\section{A Measure-theoretic Foundation}
\label{sec:measure-theory}

At their core, (large) language models are an attempt to place a probabilistic distribution over natural language utterances.
However, our toy examples in \cref{ex:inf-coin-toss,ex:possibly-inf-coin-toss} in the previous section reveal that it can be relatively tricky to get a satisfying definition of a language model.
Thus, our first step forward is to review the basics of rigorous probability theory,\footnote{By rigorous probability theory we mean a measure-theoretic treatment of probability theory.} the tools we need to come to a satisfying definition.
Our course will assume that you have had some exposure to rigorous probability theory before, and just review the basics.
However, it is also possible to learn the basics of rigorous probability on the fly during the course if it is new to you.
Specifically, we will cover \emph{measure-theoretic} foundations of probability theory.
This might come as a bit of a surprise since we are mostly going to be talking about \emph{language}, which is made up of discrete objects---strings.
However, as we will see in \cref{sec:lm-tightness} soon, formal treatment of language modeling indeed requires some mathematical rigor from measure theory.

The goal of measure-theoretic probability is to assign probabilities to \emph{subsets}
of an \defn{outcome space}\index{outcome space} $\samplespace$.
However, in the course of the study of measure theory, it has become clear that for many common $\samplespace$, it is impossible to assign probabilities in a way that satisfies a set of reasonable desiderata.\footnote{Measure theory texts commonly discuss such desiderata and the dilemma that comes with it. See, e.g., Chapter 7 in \citet{Tao2016}, Chapter 3 in \citet{Royden1988} or Chapter 3 in \citet{Billingsley1986}.
    We also give an example later.}
Consequently, the standard approach to probability theory resorts to only assigning probability to certain ``nice'' (but not necessarily all) subsets of $\samplespace$, which are referred to as \defn{events}\index{event} or \defn{measurable subsets}\index{measurable subset}, as in the theory of integration or functional analysis.
The set of measurable subsets is commonly denoted as $\eventspace$ (\cref{def:sigma-algebra}) and a probability measure $\probfunction:\eventspace\to[0,1]$ is the function that assigns a probability to each measurable subset.
The triple $(\samplespace,\eventspace,\probfunction)$ is collectively known as a probability space (\cref{def:probability-measure}).
As it turns out, the following simple and reasonable requirements imposed on $\eventspace$ and $\probfunction$ are enough to rigorously discuss probability.

\begin{definition}{$\sigalgebra$}{sigma-algebra}
    Let $\powerset{\samplespace}$ be the power set of $\samplespace$.
    Then $\eventspace \subseteq \powerset\samplespace$ is called a \defn{$\sigalgebra$}\index{$\sigalgebra$} (or $\sigma$-field) over $\samplespace$ if the following conditions hold:
    \begin{enumerate}[label=\arabic*)]
        \item $\samplespace\in\eventspace$,
        \item if $\sE\in\eventspace$, then $\setcomplement{\sE} \in \eventspace$,
        \item if $\sE_1,\sE_2,\dots$ is a finite or infinite sequence of sets in $\eventspace$, then $\bigcup_n \sE_n \in\eventspace$.
    \end{enumerate}
    If $\eventspace$ is a $\sigma$-algebra over $\samplespace$, we call the tuple $(\samplespace,\eventspace)$ a \defn{measurable space}\index{measurable space}.
\end{definition}

\begin{example}{$\sigalgebra$s}{sigma-algebras}
    Let $\samplespace$ be any set.
    Importantly, there is more than one way to construct a $\sigalgebra$ over $\samplespace$:
    \begin{enumerate}
        \item The family consisting of only the empty set $\emptyset$ and the set $\samplespace$, i.e., $\eventspace\defeq \left\{\emptyset, \samplespace\right\}$, is called the \emph{minimal} or \emph{trivial} $\sigalgebra$.
        \item The full power set $\eventspace \defeq \powerset{\samplespace}$ is called the \emph{discrete $\sigalgebra$}.
        \item Given $\sA \subseteq \samplespace$, the family $\eventspace \defeq \left\{\emptyset, \sA, \samplespace \setminus \sA, \samplespace \right\}$ is a $\sigalgebra$ induced by $\sA$.
        \item Suppose we are rolling a six-sided die.
              There are six events that can happen: We can roll any of the numbers $1\text{--}6$.
              In this case, we will then define the set of outcomes $\samplespace$ as $\samplespace \defeq \left\{\text{The number observed is } n \mid n = 1, \ldots, 6\right\}$.
              There are of course multiple ways to define an event space $\eventspace$ and with it a $\sigalgebra$ over this outcome space.
              By definition, $\emptyset \in \eventspace$ and $\samplespace \in \eventspace$.
              One way to intuitively construct a $\sigalgebra$ is to consider that all individual events (observing any number) are possible, meaning that we would like to later assign probabilities to them (see \cref{def:probability-measure}).
              This means that we should include individual singleton events in the event space: $\left\{\text{The number observed is } n \right\} \in \eventspace$ for $n =1, \ldots, 6$.
              It is easy to see that in this case, to satisfy the axioms in \cref{def:sigma-algebra}, the resulting event space should be $\eventspace = \powerset{\samplespace}$.
    \end{enumerate}
    You might want to confirm these are indeed $\sigalgebra$s by checking them against the axioms in \cref{def:sigma-algebra}.
\end{example}
A measurable space guarantees that operations on countably many sets are always valid, and hence permits the following definition.
\begin{definition}{Probability measure}{probability-measure}
    A \defn{probability measure}\index{probability measure} $\probfunction$ over a measurable space $(\samplespace,\eventspace)$ is a function $\probfunction:\eventspace\to[0,1]$ such that
    \begin{enumerate}[label=\arabic*)]
        \item $\probfunction(\samplespace)=1$,
        \item if $\sE_1,\sE_2,\dots$ is a countable sequence of disjoint sets in $\eventspace$, then $\probfunction(\bigcup_n \sE_n)=\sum_n \probfunction(\sE_n)$.
    \end{enumerate}
    In this case we call $(\samplespace,\eventspace, \probfunction)$ a \defn{probability space}\index{probability space}.\looseness=-1
\end{definition}

As mentioned, measure-theoretic probability only assigns probabilities to ``nice'' subsets of $\samplespace$.
In fact, it is often impossible to assign a probability measure to every single subset of $\samplespace$ and we must restrict our probability space to a strict subset of $\powerset{\samplespace}$.
More precisely, the sets $\sB \subseteq \samplespace$\clara{Just checking, does it make sense that this is $\subseteq$ rather than $\subset$?} for which a probability (or more generally, a \emph{volume}) can not be defined are called \emph{non-measurable sets}.
An example of such sets is the Vitali set.\footnote{See \url{https://en.wikipedia.org/wiki/Non-measurable_set} and \url{https://en.wikipedia.org/wiki/Vitali_set}.}
See also Appendix A.2 in \citet{Durrett2019}.

Later, we will be interested in modeling probability spaces over sets of (infinite) sequences.
By virtue of a theorem due to Carath\'{e}odory, there is a natural way to construct such a probability space for sequences (and many other spaces) that behaves in accordance with our intuition, as we will clarify later.
Here, we shall lay out a few other necessary definitions.
\begin{definition}{Algebra}{algebra}
    $\eventspaceA \subseteq \powerset{\samplespace}$ is called an \defn{algebra}\index{algebra} (or field) over $\samplespace$ if
    \begin{enumerate}[label=\arabic*)]
        \item $\samplespace\in\eventspaceA$,
        \item if $\sE\in\eventspaceA$, then $\setcomplement{\sE} \in \eventspaceA$,
        \item if $\sE_1,\sE_2 \in \eventspaceA$, then $\sE_1 \cup \sE_2 \in \eventspaceA$.
    \end{enumerate}
\end{definition}

\begin{definition}{Probability pre-measure}{pre-measure}
    Let $\eventspaceA$ be an algebra over some set $\samplespace$. A \defn{probability pre-measure}\index{probability pre-measure} over $(\samplespace,\eventspaceA)$ is a function $\probfunction_0:\eventspaceA \to [0,1]$ such that
    \begin{enumerate}[label=\arabic*)]
        \item $\probfunction_0(\samplespace)=1$,
        \item if $\sE_1,\sE_2,\dots$ is a (countable) sequence of disjoint sets in $\eventspaceA$ \emph{whose (countable) union is also in} $\eventspaceA$, then $\probfunction_0(\cup_{\idx=1}^\infty \sE_\idx)=\sum_{\idx=1}^{\infty} \probfunction_0(\sE_\idx)$.
    \end{enumerate}
\end{definition}

Note that the only difference between a $\sigma$-algebra (\cref{def:sigma-algebra}) and an algebra is that condition 3 is weakened from countable to finite, and the only difference between a probability measure (\cref{def:probability-measure}) and a pre-measure is that the latter is defined with respect to an algebra instead of a $\sigma$-algebra.

The idea behind Carath\'eodory's extension theorem is that there is often a simple construction of an algebra $\eventspaceA$ over $\samplespace$ such that there is a natural way to define a probability pre-measure.
One can then \emph{extend} this probability pre-measure to a probability measure that is both minimal and unique in a precise sense.
For example, the standard Lebesgue measure over the real line can be constructed this way.

Finally, we define random variables.
\begin{definition}{Random}{rv}
    A mapping $\rx:\samplespace\to \sS$ between two measurable spaces $(\samplespace, \eventspace)$ and $(\sS,\sT)$ is an $(\sS,\sT)$-valued \defn{random variable}\index{random variable}, or a measurable mapping, if, for all $\sB \in \sT$,
    \begin{equation}
        \inv{\rx}(\sB)\defeq \{\omega\in\samplespace:\rx(\omega)\in \sB\} \in \eventspace.
    \end{equation}
\end{definition}

Any measurable function (random variable) induces a new probability measure on the \emph{output} $\sigma$-algebra based on the one defined on the original $\sigma$-algebra.
This is called the \defn{pushforward measure}\index{pushforward measure} (cf. \S2.4 in \citealp{Tao2011}), which we will denote by $\pushfwdMeasure$, given by
\begin{align}
    \pushfwdMeasure\left(\rx\in \sE\right)\defeq\probfunction\left(\inv{\rx}\left(\sE\right)\right),
\end{align}
that is, the probability of the result of $\rx$ being in some event $\sE$ is determined by the probability of the event of all the elements which $\rx$ maps into $\sE$, i.e., the pre-image of $\sE$ given by $\rx$.

\begin{example}{Random Variables}{random-variables}
    We give some simple examples of random variables.
    \begin{enumerate}
        \item Let $\samplespace$ be the set of possible outcomes of throwing a fair coin, i.e., $\samplespace \defeq \left\{\texttt{T}, \texttt{H}\right\}$. Define $\eventspace \defeq \powerset{\samplespace}$, $\sS \defeq \left\{0, 1\right\}$, and $\sT \defeq \powerset{\sS}$.
              Then, the random variable
              \begin{equation*}
                  \rvx: \begin{cases} \texttt{T} \mapsto 0 \\ \texttt{H} \mapsto 1 \end{cases}
              \end{equation*}
              assigns tails (\texttt{T}) the value $0$ and heads (\texttt{H}) the value $1$.
        \item
              Consider the probability space of throwing two dice (similar to \cref{ex:sigma-algebras}) where $\samplespace=\left\{(i, j): i, j=1, \ldots, 6\right\}$ where the element $(i, j)$ refers to rolling $i$ on the first and $j$ on the second die and $\eventspace=\powerset{\samplespace}$. Define $\sS \defeq \mathbb{Z}$ and $\sT \defeq \powerset{\sS}$.
              Then, the random variable
              \begin{equation*}
                  \rvx: (i,j) \mapsto i+j
              \end{equation*}
              is an $\left(\sS,\sT\right)$-valued random variable which represents the sum of two dice.
    \end{enumerate}
\end{example}

\newpage{}

\section{Language Models: Distributions over Strings}
Language models are defined as probability distributions over sequences of words, referred to as utterances. This chapter delves into the formalization of the term ``utterance'' and introduces fundamental concepts such as the alphabet, string, and language. Utilizing these concepts, a formal definition of a language model is presented, along with a discussion on the intricacies of defining distributions over infinite sets.

\subsection{Sets of Strings}
We begin by defining the very basic notions of alphabets and strings, where we take inspiration from \defn{formal language theory}\index{formal language theory}.
First and foremost, formal language theory concerns itself with \emph{sets of structures}.
The simplest structure it considers is a \defn{string}\index{string}.
So what is a string?
We start with the notion of an alphabet.
\begin{definition}{Alphabet}{alphabet}
    An \defn{alphabet}\index{alphabet} is a finite, non-empty set.
    In this course, we will denote an alphabet using Greek capital letters, e.g., $\alphabet$ and $\outalphabet$.
    We refer to the elements of an alphabet as \defn{symbols}\index{symbols} or letters and will denote them with lowercase letters: $\syma$, $\symb$, $\symc$.
\end{definition}

\begin{definition}{String}{string}
    A \defn{string}\footnote{A string is also referred to as a \defn{word}\index{word}, which continues with the linguistic terminology.} over an alphabet is any \emph{finite} sequence of letters.
    Strings made up of symbols from $\alphabet$ will denoted by bolded Latin letters, e.g., $\stry = \sym_1 \cdots \sym_\strlen$ where each $\sym_n \in \alphabet$.
\end{definition}

The length of a string, written as $|\str|$, is the number of letters it contains.
Usually, we will use $\strlen$ to denote $|\str|$ more concisely whenever the usage is clear from the context.
There is only one string of length zero, which we denote with the distinguished symbol $\eps$ and refer to as the \emph{empty string}\index{empty string}.
By convention, $\eps$ is \emph{not} an element of the original alphabet.

New strings are formed from other strings and symbols with \defn{concatenation}\index{concatenation}.
Concatenation, denoted with $\strx \circ \stry$ or just $\strx\stry$, is an associative operation on strings.
Formally, the concatenation of two words $\str$ and $\strx$ is the word $\str \circ \strx = \str \strx$, which is obtained by writing the second argument after the first one.
The result of concatenating with $\eps$ from either side results in the original string, which means that $\eps$ is the \defn{unit} of concatenation and the set of all words over an alphabet with the operation of concatenation forms a \defn{monoid}\index{monoid}.

We have so far only defined strings as individual sequences of symbols.
To give our strings made up of symbols in $\alphabet$ a set to live in, we now define Kleene closure of an alphabet $\alphabet$.
\begin{definition}{Kleene Star}{kleene-star}
    Let $\alphabet$ be an alphabet.
    The \defn{Kleene star}\index{Kleene star} $\kleene{\alphabet}$ is defined as
    \begin{equation}
        \kleene{\alphabet} = \bigcup_{\textcolor{ETHRed}{n = 0}}^\infty \alphabet^n
    \end{equation}
    where
    \begin{equation}
        \alphabet^n \defeq \underbrace{\alphabet \times \cdots \times \alphabet}_{n \text{ times}}
    \end{equation}
    Note that we define $\alphabet^0 \defeq \{\eps\}$.
    We call the $\kleene{\alphabet}$ the \defn{Kleene closure}\index{Kleene closure} of the alphabet $\alphabet$.
    We also define
    \begin{equation}
        \kleeneplus{\alphabet} \defeq \bigcup_{\textcolor{ETHRed}{n = 1}}^\infty \alphabet^n = \alphabet \kleene{\alphabet}.
    \end{equation}
\end{definition}

Finally, we also define the set of all infinite sequences of symbols from some alphabet $\alphabet$ as $\alphabet^\infty$.
\begin{definition}{Infinite sequences}{infinite-sequences}
    Let $\alphabet$ be an alphabet.
    The set of all \defn{infinite sequences}\index{infinite sequence} over $\alphabet$ is defined as:
    \begin{equation}
        \alphabet^\infty \defeq \underbrace{\alphabet \times \cdots \times \alphabet}_{\infty \text{-times}},
    \end{equation}
\end{definition}
Since strings are canonically \emph{finite} in computer science, we will explicitly use the terms infinite sequence or infinite string to refer to elements of $\alphabet^\infty$.

More informally, we can think of $\kleene{\alphabet}$ as the set which contains $\eps$ and all (finite-length) strings which can be constructed by concatenating arbitrary symbols from $\alphabet$.
$\kleeneplus{\alphabet}$, on the other hand, does \emph{not} contain $\varepsilon$, but contains all other strings of symbols from $\alphabet$.
The Kleene closure of an alphabet is a \emph{countably infinite} set (this will come into play later!).
In contrast, the set $\alphabet^\infty$ is \emph{uncountably infinite} for any $\alphabet$ such that $|\alphabet| \geq 2$.

The notion of the Kleene closure leads us very naturally to our next definition.
\begin{definition}{Formal language}{formal-language}
    Let $\alphabet$ be an alphabet.
    A \defn{language}\index{language} $\lang$ is a subset of $\kleene{\alphabet}$.
\end{definition}
That is, a language is just a specified subset of all possible strings made up of the symbols in the alphabet.
This subset can be specified by simply enumerating a finite set of strings, or by a \emph{formal model}.
We will see examples of those later.
Importantly, these strings are \emph{finite}.
If not specified explicitly, we will often assume that $\lang = \kleene{\alphabet}$.

\paragraph{A note on terminology.}
As we mentioned, these definitions are inspired by formal language theory.
We defined strings as our main structures of interest and symbols as their building blocks.
When we talk about natural language, the terminology is often slightly different: we may refer to the basic building blocks (symbols) as \defn{tokens}\index{token} or \defn{words}\index{word} (which might be composed of one or more \emph{characters} and form some form of ``words'') and their compositions (strings) as \defn{sequences}\index{sequence} or \defn{sentences}\index{sentence}.
Furthermore, what we refer to here as an alphabet may be called a \defn{vocabulary}\index{vocabulary} (of words or tokens) in the context of natural language.
Sentences are therefore concatenations of words from a vocabulary in the same way that strings are concatenations of symbols from an alphabet.

\begin{example}{Kleene Closure}{}
    Let $\alphabet = \left\{ \syma, \symb, \symc \right\}$. Then $$\kleene{\alphabet} = \{ \eps, \syma, \symb, \symc, \syma\syma, \syma \symb, \syma \symc, \symb\syma , \symb\symb, \symb\symc, \symc\syma , \symc\symb, \symc\symc, \syma \syma \syma , \syma \syma \symb, \syma \syma \symc, \dots \}.$$
    Examples of a languages over this alphabet include $\lang_1 \defeq \set{\syma, \symb, \syma\symb, \symb\syma}$, $\lang_2 \defeq \set{\str \in \kleene{\alphabet} \mid \sym_1 = \syma}$, and $\lang_3 \defeq \set{\str \in \kleene{\alphabet} \mid |\str| \text{ is even}}$.
\end{example}

Next, we introduce two notions of subelements of strings.
\begin{definition}{String Subelements}{str-subelements}
    A \defn{subsequence}\index{subsequence} of a string $\str$ is defined as a sequence that can be formed from $\str$ by deleting some or no symbols, leaving the order untouched.
    A \defn{substring}\index{substring} is a contiguous subsequence.
    For instance, $\syma\symb$ and $\symb\symc$ are substrings and subsequences of $\str=\syma\symb\symc$, while $\syma\symc$ is a subsequence but not a substring.
    \defn{Prefixes}\index{prefix} and \defn{suffixes}\index{suffix} are special cases of substrings.
    A prefix is a substring of $\str$ that shares the same first letter as $\str$ and a suffix is a substring of $\str$ that shares the same last letter as $\str$.
    We will also denote a prefix $\sym_1\ldots\sym_{\idx - 1}$ of the string $\str = \sym_1\ldots\sym_\strlen$ as $\str_{< \idx}$. We will also use the notation $\suffixOf{\str}{\str'}$ to denote that $\str$ is a suffix of $\str'$.
\end{definition}

\subsection{Defining a Language Model}

We are now ready to introduce the main interest of the entire lecture series: language models.

\begin{definition}{Language model}{language-model}
    Let $\alphabet$ be an alphabet.
    A \defn{language model}\index{language model} is a (discrete) distribution $\pLM$ over $\kleene{\alphabet}$.
\end{definition}

\begin{example}{A very simple language model}{simple-lm}
    Let $\alphabet \defeq \set{\syma}$.
    For $n \in \Nzero$, define
    \begin{equation*}
        \pLM\left(\syma^n\right) \defeq 2^{-\left(n + 1\right)},
    \end{equation*}
    where $\syma^0 \defeq \eps$ and $\syma^n \defeq \underbrace{\syma\ldots\syma}_{n \text{ times}}$.

    We claim that $\pLM$ is a language model.
    To see that, we verify that it is a valid probability distribution over $\kleene{\alphabet}$.
    It is easy to see that $\pLM\left(\syma^\idx\right) \geq 0$ for any $\idx$.
    Additionally, we see that the probabilities of finite sequences indeed sum to $1$:
    \begin{equation*}
        \sum_{\str \in \kleene{\alphabet}} \pLM\left(\str\right) = \sum_{\idx = 0}^{\infty} \pLM\left(\syma^\idx\right)
        = \sum_{\idx = 0}^\infty 2^{-\left(\idx + 1\right)}
        = \frac{1}{2}\sum_{\idx = 0}^\infty 2^{-\idx}
        = \frac{1}{2} \frac{1}{1 - \frac{1}{2}} = 1.
    \end{equation*}
\end{example}

In our formal analysis of language models, we will also often refer to the \emph{language} defined by a language model.
\begin{definition}{Weighted language}{weighted-language}
    Let $\pLM$ be a language model.
    The \defn{weighted language}\index{language model!weighted language} of $\pLM$ is defined as
    \begin{equation}
        \lang\left(\pLM\right) \defeq \left\{\left(\str, \pLM\left(\str\right) \right) \mid \str \in \kleene{\alphabet}\right\}
    \end{equation}
\end{definition}

\begin{example}{Languge of a langauge model}{}
    The language of the language model from \cref{ex:simple-lm} is
    \begin{equation}
        \lang\left(\pLM\right) \defeq \left\{\left(\syma^n, 2^{-\left(n + 1\right)} \right) \mid n \in \Nzero\right\}
    \end{equation}
\end{example}

A language model is itself a very simple concept---it is simply a distribution that weights strings (natural utterances) by their probabilities to occur in a particular language.
Note that we have not said anything about how we can represent or model this distribution yet.
Besides, for any (natural) language, the ground-truth language model $\pLM$ is of course \emph{unknown} and complex.
The next chapter, therefore, discusses in depth the computational models which we can use to try to tractably represent distributions over strings and ways of \emph{approximating} (learning) the ground-truth distribution based on finite datasets using such models.

\newpage{}

\section{Global and Local Normalization} \label{sec:global-local-normalization}
The previous chapter introduced a formal definition of a language as a set of strings and the definition of a language model as a distribution over strings.
We now delve into a potpourri of technical questions to complete the theoretical minimum for discussing language models.
While doing so, we will introduce (and begin to answer) three fundamental questions in the first part of the course.
We will introduce them later in the section.

\paragraph{A note on terminology.}
Unfortunately, we will encounter some ambiguous terminology.
In \cref{sec:lm-tightness}, we explicitly define a language model as a valid probability distribution over $\kleene{\alphabet}$, the Kleene closure of some alphabet $\alphabet$, which means that $\sum_{\str \in \kleene{\alphabet}} \pLM\left(\str\right) = 1$.
As we will see later, this means that the model is \emph{tight}, whereas it is \emph{non-tight} if $\sum_{\str \in \kleene{\alphabet}} \pLM\left(\str\right) < 1$.
Definitionally, then, all language models are tight.
However, it is standard in the literature to refer to many non-tight language models as language models as well.
We pardon in advance the ambiguity that this introduces.
Over the course of the notes, we attempt to stick to the convention that the term ``language model'' without qualification only refers to a tight language model whereas a ``non-tight language model'' is used to refer to a language model in the more colloquial sense.
Linguistically, tight is acting as a non-intersective adjective.
Just as in English, where a fake gun is not a gun, so too in our course notes a non-tight language model is not a language model.
This distinction does in fact matter.
On one hand, we can prove that many language models whose parameters are estimated from data (e.g., a finite-state language model estimated by means of maximum-likelihood estimation) are, in fact, tight.
On the other hand, we can show that this is \emph{not} true in general, i.e., \emph{not} all language models estimated from data will be tight.
For instance, a recurrent neural network language model estimated through gradient descent may not be tight \citep{Chen2018}.

When specifying $\pLM$, we have two fundamental options.
Depending on whether we model $\pLM\left(\str\right)$ for each string $\str$ \emph{directly} or we model \emph{individual} conditional probabilities $\pLM\left(\sym_\idx \mid \str_{<\tstep}\right)$ we distinguish \emph{globally} and \emph{locally} normalized models.
The names naturally come from the way the distributions in the two families are normalized: whereas globally normalized models are normalized by summing over the entire (infinite) space of strings, locally normalized models define a sequence of \emph{conditional distributions} and make use of the chain rule of probability to define the joint probability of a whole string.

\paragraph{The beginning of sequence string symbol.}
Conventionally, we will include a special symbol over which globally or locally normalized models operate: the \defn{\underline{b}eginning \underline{o}f \underline{s}equence} ($\bos$) symbol\index{$\bos$ symbol}, which, as the name suggests, denotes the beginning of a string or a sequence.
For a string $\str = \sym_1 \cdots \sym_\strlen$, we will suggestively denote $\sym_{0} \defeq \bos$.

\subsection{Globally Normalized Language Models}

We start with globally normalized models.
Such models are also called \defn{energy-based} language models\index{language model!energy-based} in the literature \citep{Bakhtin2021}.
To define a globally normalized language model, we start with the definition of an energy function.
\begin{definition}{Energy function}{energy-function}
    An \defn{energy function}\index{energy function} is a function $\scoringfunc: \kleene{\alphabet} \to \R$.
\end{definition}
\clara{Hmmm this is a very specific energy function. Perhaps we should call it like a "language energy function" if we're going to specifically say $\scoringfunc: \kleene{\alphabet} \to \R$}
\ryan{I think we should write a bit more about the negative exp bit that comes up.}
Inspired by concepts from statistical mechanics, an energy function can be used to define a very general class of probability distributions by normalizing its exponentiated negative values.

Now, we can define a globally normalized language model in terms of an energy function over $\kleene{\alphabet}$.
\begin{definition}{Globally normalized models}{globally-normalized-model}
    Let $\unnormalizedpGN\left(\str\right) : \kleene{\alphabet} \rightarrow \R$ be an energy function.
    A \defn{globally normalized model}\index{language model!globally normalized}  (\GNMAcronym{}) is
    defined as
    \begin{equation} \label{eq:globally-normalized-model}
        \pLM\left(\str\right) \defeq \frac{\exp \left[- \unnormalizedpGN\left(\str\right)\right]}{\sum_{\str' \in \kleene{\alphabet}}\exp\left[-\unnormalizedpGN\left(\str'\right)\right]} \defeq \frac{1}{\normConstant_G} \exp\left[-\unnormalizedpGN\left(\str\right)\right],
    \end{equation}
    where $\normConstant_G \defeq \sum_{\str' \in \kleene{\alphabet}}\exp\left[-\unnormalizedpGN\left(\str'\right)\right]$.\footnote{We will later return to this sort of normalization when we define the $\softmax$ function in \cref{sec:general-framework}.}
    We call $\normConstant_G$ the \defn{normalization constant}\index{normalization constant}.

\end{definition}
Globally normalized models are attractive because one only needs to define an (unnormalized) energy function $\unnormalizedpGN$, which scores entire sequences at once.
This is often easier than specifying a probability distribution.
Furthermore, they define a probability distribution over strings $\str \in \kleene{\alphabet}$ \emph{directly}.
As we will see in \cref{sec:locally-normalized-models}, this stands in contrast to locally normalized language models which require care with the space over which they operate.
However, the downside is that it may be difficult to compute the normalizer $\normConstant_G$.

\subsubsection{Normalizability}
In defining the normalizer $\normConstant_G \defeq \sum_{\str' \in \kleene{\alphabet}} \exp\left[- \unnormalizedpGN\left(\str'\right)\right]$, we notationally cover up a certain subtlety.
The set $\kleene{\alphabet}$ is countably infinite, so $\normConstant_G$ may diverge to $\infty$.
In this case, \cref{eq:globally-normalized-model} is not well-defined.
This motivates the following definition.
\begin{definition}{Normalizable energy function}{norm-energy-func}
    We say that an energy function is \defn{normalizable}\index{normalizable energy function} if the quantity $\normConstant_G$ in \cref{eq:globally-normalized-model} is finite, i.e., if $\normConstant_G < \infty$.
\end{definition}
With this definition, we can state a relatively trivial result that characterizes when an energy function can be turned into a globally normalized language model.
\begin{theorem}{Normalizable energy functions induce language models}{}
    Any normalizable energy function $\pGN$ induces a language model, i.e., a distribution over $\kleene{\alphabet}$.
\end{theorem}
\begin{proof}
    Given an energy function $\unnormalizedpGN$, we have $\exp{\left[-\unnormalizedpGN\left(\str\right)\right]} \geq 0$ and
    \begin{align} \label{eq:globally-normalized-formulation}
        \sum_{\str \in \kleene{\alphabet}} \pGN\left(\str\right) & = \sum_{\str \in \kleene{\alphabet}} \frac{\exp{\left[-\unnormalizedpGN\left(\str\right)\right]}}{\sum_{\str' \in \kleene{\alphabet}} \exp{\left[-\unnormalizedpGN\left(\str'\right)\right]}}  \\
                                                                 & = \frac{1}{\sum_{\str' \in \kleene{\alphabet}} \exp{\left[-\unnormalizedpGN\left(\str'\right)\right]}}\sum_{\str \in \kleene{\alphabet}} \exp{\left[-\unnormalizedpGN\left(\str\right)\right]} \\
                                                                 & = 1,
    \end{align}
    which means that $\pGN$ is a valid probability distribution over $\kleene{\alphabet}$.
\end{proof}

While the fact that normalizable energy functions always form a language model is a big advantage, we will see later that \emph{ensuring} that they are normalizable can be difficult and restrictive.
This brings us to the first fundamental question of the section:
\begin{aquestion}{Normalizing an energy function}{}
    When is an energy function normalizable?
    More precisely, for which energy functions $\unnormalizedpGN$ is $\normConstant_G < \infty$?
\end{aquestion}
We will not discuss any specific results here, as there are no general necessary or sufficient conditions---the answer to this of course depends on the precise definition of $\unnormalizedpGN$.
Later in the course notes,\ryan{maybe give an explicit citation here.} we will present two formalisms where we can exactly characterize when an energy function is normalizable.
First, when it is weighted finite-state automaton (cf. \cref{sec:finite-state}), and, second, when it is defined through weighted context-free grammars (\cref{sec:context-free}) and discuss the specific sufficient and necessary conditions there.
However, under certain assumptions, determining whether an energy function is normalizable in the general case is undecidable.\ryan{Add forward reference.}

Moreover, even if it is known that an energy function is normalizable, we still need an efficient algorithm to compute it.
But, efficiently computing $\normConstant_G$ can be challenging: the fact that $\kleene{\alphabet}$ is \emph{infinite} means that we cannot always compute $\normConstant_G$ in a \emph{tractable} way.
In fact, there are no general-purpose algorithms for this.
Moreover, sampling from the model is similarly intractable, as entire sequences have to be drawn at a time from the large space $\kleene{\alphabet}$.

\subsection{Locally Normalized Language Models} \label{sec:locally-normalized-models}

The inherent difficulty in computing the normalizer, an  infinite summation over $\kleene{\alphabet}$, motivates the definition of locally normalized language models, which we will denote with $\pLN$.
Rather than defining a probability distribution over $\kleene{\alphabet}$ directly, they decompose the problem into the problem of modeling a series of conditional distributions over the next possible symbol in the string given the context so far, i.e., $\pLN\left(\sym\mid \str\right)$, which could be na{\"i}vely combined into the full probability of the string by multiplying the conditional probabilities.\footnote{We will soon see why this would not work and why we have to be a bit more careful.}
Intuitively, this reduces the problem of having to normalize the distribution over an infinite set $\kleene{\alphabet}$ to the problem of modeling the distribution of the \emph{next possible symbol} $\sym_\idx$ given the symbols seen so far $\str_{<\idx}$.
This means that normalization would only ever require summation over $|\alphabet|$ symbols at a time, solving the tractability issues encountered by globally normalized models.

However, we immediately encounter another problem:
In order to be a language model, $\pLN\left(\sym\mid \str\right)$ must
constitute a probability distribution over $\kleene{\alphabet}$.
However, as we will discuss in the next section, this may not be the case because locally normalized models can place positive probability mass on \emph{infinitely long} sequences (cf. \cref{ex:non-tight-2-gram} in \cref{sec:tightness-intro}).
Additionally, we also have to introduce a new symbol that tells us to ``stop'' generating a string,
which we call the \defn{\underline{e}nd \underline{o}f \underline{s}equence} symbol, $\eos$\index{$\eos$ symbol}.
Throughout the notes, we will assume $\eos \notin \alphabet$ and we define
\begin{equation}
    \eosalphabet \defeq \alphabet \cup \left\{\eos\right\}.
\end{equation}
Moreover, we will explicitly denote elements of $\kleene{\eosalphabet}$ as $\eosstr$ and symbols in $\eosalphabet$ as $\eossym$.
Given a sequence of symbols and the $\eos$ symbol, we take the string to be the sequence of symbols encountered \emph{before} the \emph{first} $\eos$ symbol.
Informally, you can think of the $\bos$ symbol as marking the beginning of the string, and the $\eos$ symbol as denoting the end of the string or even as a language model terminating its generation, as we will see later.

Due to the issues with defining valid probability distributions over $\kleene{\alphabet}$, we will use the term sequence model to refer to any model that may place positive probability on infinitely long sequences.
Thus, sequence models are strictly more general than language models, which, by definition, only place positive probability mass on strings, i.e., finite sequences.
\begin{definition}{Sequence model}{sequence-model-1}
    Let $\alphabet$ be an alphabet.
    A \defn{sequence model}\index{sequence model} (\SMAcronym{}) over $\alphabet$ is defined as a set of conditional probability distributions
    \begin{equation} \label{eq:locally-normalized-model}
        \pLNSM\left(\sym\mid\str\right)
    \end{equation}
    for $\sym \in \alphabet$ and $\str \in \kleene{\alphabet}$.
    We will refer to the string $\str$ in $\pLNSM\left(\sym \mid \str\right)$ as the \defn{history}\index{history} or the \defn{context}\index{context}.
\end{definition}
Note that we will mostly consider \SMAcronym{}s over the set $\eosalphabet$.
To reiterate, we have just formally defined locally normalized \emph{sequence} models rather than locally normalized \emph{language} models.
That has to do with the fact that, in contrast to a globally normalized model with a normalizable energy function, a \SMAcronym{} might not correspond to a \emph{language} model, as alluded to at the beginning of this section and as we discuss in more detail shortly.

We will now work up to a locally normalized \emph{language} model.
\begin{definition}{Locally normalized language model}{locally-normalized-model}
    Let $\alphabet$ be an alphabet.
    Next, let $\pLNSM$ be a sequence model over $\eosalphabet$.
    A \defn{locally normalized language model}\index{language model!locally normalized} (\LNMAcronym{}) over $\alphabet$ is defined as
    \begin{equation}
        \pLN(\str) \defeq \pLNSM(\eos \mid \str) \prod_{\tstep=1}^{\finaltstep} \pLNSM(\sym_\tstep \mid \str_{<\tstep})
    \end{equation}
    for $\str \in \kleene{\alphabet}$ with $|\str| = \strlen$.
    We say a locally normalized language model is \defn{tight} if
    \begin{equation}
        \sum_{\str\in \kleene{\alphabet}} \pLN(\str) = 1.
    \end{equation}
    Tightness is a nuanced concept that will be discussed in great detail in \cref{sec:lm-tightness}.
\end{definition}

\begin{figure}
    \centering
    \begin{subfigure}{\textwidth}
        \begin{tikzpicture}[level distance=1.75cm,sibling distance=0.3cm,
                edge from parent path={(\tikzparentnode) -- (\tikzchildnode)}]
            \Tree
            [.$\bos$
            \edge node[auto, sloped] {{\scriptsize $0.04$}};
            [.The
            \edge node[auto, sloped] {{\scriptsize $0.08$}};
            [.quick
            \edge node[auto, sloped] {{\scriptsize $0.12$}};
            [.brown
            \edge node[auto, sloped] {{}};
            [.$\cdots$ ]
            ]
            \edge node[auto, sloped] {{\scriptsize $0.01$}};
            [.and
            \edge node[auto, sloped] {{}};
            [.$\cdots$ ]
            ]
            ]
            \edge node[auto, sloped] {{\scriptsize $0.13$}};
            [.best
            \edge node[auto, sloped] {{\scriptsize $0.22$}};
            [.$\eos$ ]
            \edge node[auto, sloped] {{\scriptsize $0.07$}};
            [.!
            \edge node[auto] {{\scriptsize $1$}};
            [.$\eos$ ]
            ]]
            ]
            \edge node[auto, sloped] {{\scriptsize $0.01$}};
            [.Please
            \edge node[auto, sloped] {{\scriptsize $0.09$}};
            [.don't
            \edge node[auto, sloped] {{}};
            [.$\cdots$ ]]
            \edge node[auto, sloped] {{\scriptsize $0.02$}};
            [.consider
            \edge node[auto, sloped] {{}};
            [.$\cdots$ ]]
            ]
            \edge node[auto, sloped] {{\scriptsize }};
            [.$\cdots$ ]
            \edge node[auto, sloped] {{\scriptsize $0.03$}};
            [.Hello
            \edge node[auto, sloped] {{\scriptsize $0.21$}};
            [.world
            \edge node[auto, sloped] {{}};
            [.$\cdots$ ]]
            \edge node[auto, sloped] {{\scriptsize $0.06$}};
            [.there
            \edge node[auto, sloped] {{}};
            [.$\cdots$ ]]
            ]
            ]
        \end{tikzpicture}
        \caption{An example of a locally normalized language model. The values of the edges represent the conditional probability of observing the new word given the observed words (higher up on the path from the root node $\bos$).
            Note that the probabilities stemming from any inner node should sum to $1$---however, to avoid clutter, only a subset of the possible arcs is drawn.
        }
        \label{fig:locally-normalized-lm}
    \end{subfigure}
    \begin{subfigure}{\textwidth}
        \centering
        \begin{tikzpicture}
            \node[rectangle, dashed, ETHPetrol!50, minimum width=8cm, minimum height=5cm, draw, fill=ETHPetrol!15] (oval) {};
            \node[anchor=east] (p) at ([shift={(-0.5,0)}]oval.west) {$\str \sim$};
            \node[rounded corners, fill=ETHPetrol!30, align=center] at  ([shift={(0.75,1.5cm)}]oval.center) {$\unnormalizedpGN\left(\text{ The best }\right)$};
            \node[rounded corners, fill=ETHPetrol!30, align=center] at  ([shift={(-2,0.5cm)}]oval.center) {$\unnormalizedpGN\left(\text{ The best! }\right)$};
            \node[rounded corners, fill=ETHPetrol!30, align=center] at  ([shift={(-0.75,-0.5cm)}]oval.center) {$\unnormalizedpGN\left(\text{ The quick fox. }\right)$ };
            \node[rounded corners, fill=ETHPetrol!30, align=center] at  ([shift={(0.5,-1.5cm)}]oval.center) {$\unnormalizedpGN\left(\text{ Hello World! }\right)$ };
        \end{tikzpicture}
        \caption{An example of a globally normalized model which can for example generate sentences based on the probabilities determined by normalizing the assigned scores $\unnormalizedpGN$.}
        \label{fig:globally-normalized-lm}
    \end{subfigure}
    \caption{``Examples'' of a locally and a globally normalized language model.}
\end{figure}
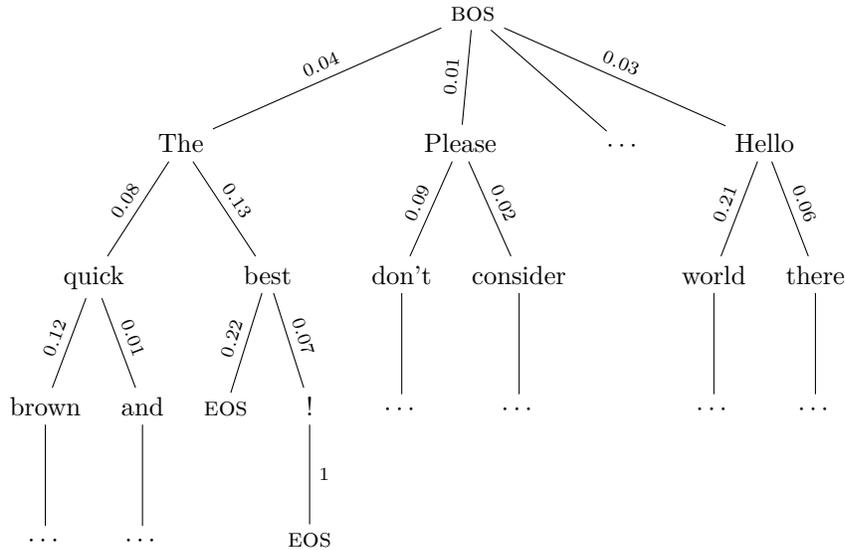
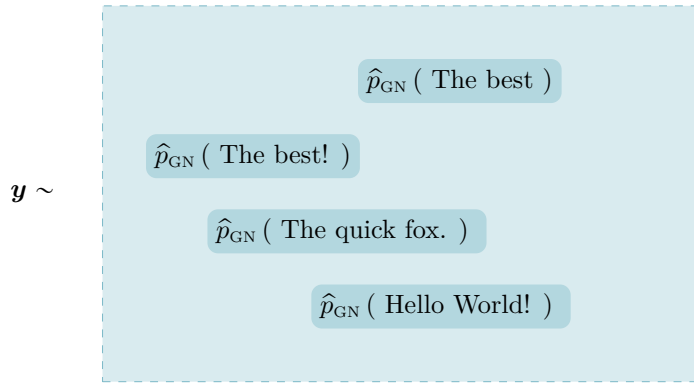

We now contrast globally and locally normalized models pictorially in the following example.
\begin{example}{Locally and globally normalized language models}{}
    \cref{fig:locally-normalized-lm} shows a simple instance of what a locally normalized language model would look like.
    We can compute the probabilities of various strings by starting at the root node $\bos$ and choosing one of the paths to a leaf node, which will always be $\eos$.
    The values on the edges represent the conditional probabilities of observing the new word given at the target of the edge given the context seen on the path so far, i.e., $\pLN\left(\eossym_\tstep\mid \eosstr_{<\tstep}\right)$ at the level $\tstep$ of the tree.
    For example, the probability of the string $\bos \text{ ``The best'' } \eos$ under this language model is $0.04 \cdot 0.13 \cdot 0.22 = 0.001144$.
    On the other hand, a globally normalized model would simply score all possible sentences using the score function $\unnormalizedpGN\left(\str\right)$, as is hinted at in \cref{fig:globally-normalized-lm}.

\end{example}

\subsubsection{Locally Normalizing a Language Model} \label{sec:locally-normalizing-lms}

The second fundamental question of this section concerns the relationship between language models and local normalization.
\begin{aquestion}{Locally normalizing a language model}{}
    When can a language model be locally normalized?
\end{aquestion}
The answer to that is simple: \emph{every} language model can be locally normalized!
While the intuition behind this is very simple, the precise formulation is not.
Before we discuss the details, we have to introduce the concept of prefix probabilities, which denote the sum of the probabilities of all strings beginning with a certain prefix.
\begin{definition}{Prefix probability}{prefix-probability}
    Let $\pLM$ be a language model.
    We define a $\pLM$'s \defn{prefix probability}\index{prefix probability} $\pPrefix$ as
    \begin{equation}
        \pPrefix\left(\str\right) \defeq \sum_{\str' \in \kleene{\alphabet}} \pLM\left(\str \str'\right),
    \end{equation}
    that is, the probability that $\str$ is a prefix of any string $\str \str'$ in the language, or, equivalently, the cumulative probability of all strings beginning with $\str$.
\end{definition}
Note that, naturally, $\pPrefix\left(\eps\right) = 1$.

\begin{theorem}{Any language model can be locally normalized}{locally-normalizing-lms}
    Let $\pLM$ be a language model.
    Then, there exists a locally normalized language model $\pLN$ such that, for all $\str \in \kleene{\alphabet}$ with $|\str| = \strlen$,
    \begin{equation}
        \pLM\left(\str\right) = \pLN\left(\str\right) = \pLNSM(\eos \mid \str) \prod_{\tstep=1}^{\strlen} \pLNSM(\sym_\tstep \mid \str_{<\tstep}).
    \end{equation}
\end{theorem}
\begin{proof}
    We define the individual conditional probability distributions over the next symbol of the \SMAcronym{} $\pLNSM$ using the chain rule of probability.
    If $\pPrefix(\str) > 0$, then define
    \begin{equation} \label{eq:lm-lnsm-1}
        \pLNSM\left(\sym\mid\str\right) \defeq \frac{ \pPrefix\left(\str\sym\right)}{ \pPrefix\left(\str\right)}
    \end{equation}
    for $\sym \in \alphabet$ and $\str \in \kleene{\alphabet}$ such that $\pdens\left(\str\right) > 0$.
    We still have to define the probabilities of \emph{ending} the sequence using $\pLNSM$ by defining the $\eos$ probabilities.
    We define, for any $\str \in \kleene{\alphabet}$ such that $\pPrefix(\str) > 0$,
    \begin{equation} \label{eq:lm-lnsm-2}
        \pLNSM\left(\eos\mid\str\right) \defeq \frac{\pLM(\str)}{\pPrefix(\str)}
    \end{equation}
    that is, the probability that the globally normalized model will generate \emph{exactly} the string $\str$ and not any continuation of it $\str \str'$, given that $\str$ has already been generated.
    Each of the conditional distributions of this model (\cref{eq:lm-lnsm-1,eq:lm-lnsm-2}) is clearly defined over $\eosalphabet$.
    This, therefore, defines a valid \SMAcronym{}.
    To see that $\pLN$ constitutes the same distribution as $\pLM$, consider two cases.
    \paragraph{Case 1:}
    Assume $\pPrefix(\str) > 0$.
    Then, we have\anej{add a note on why conditionals defined like this form a prob. dist.}
    \begin{align}
        \pLN\left(\str\right) & = \left[\prod_{\tstep = 1}^{\strlen} \pLNSM\left(\sym_{\tstep} \mid \str_{<\tstep}\right)\right] \pLNSM\left(\eos\mid \str\right)                                                                                                                                                                                                                                                                       \\
                              & = \frac{\cancel{\pPrefix\left(\sym_1\right)}}{\pPrefix\left(\eps\right)} \frac{\cancel{\pPrefix\left(\sym_1\sym_2\right)}}{\cancel{\pPrefix\left(\sym_1\right)}} \cdots \frac{\cancel{\pPrefix\left(\str_{<\strlen}\right)}}{\cancel{\pPrefix\left(\str_{<\strlen-1}\right)}} \frac{\pPrefix\left(\str\right)}{\cancel{\pPrefix\left(\str_{<\strlen}\right)}}\pLNSM\left(\eos\mid \str\right) \nonumber \\
                              & = \frac{\cancel{\pPrefix\left(\str\right)}}{\pPrefix\left(\eps\right)}\frac{ \pLM\left(\str\right)}{ \cancel{\pPrefix\left(\str\right)}}                                                                                                                                                                                                                                                                \\
                              & = \pLM\left(\str\right)
    \end{align}
    where $\pPrefix\left(\eps\right) = 1$.
    \paragraph{Case 2: }
    Assume $\pPrefix(\str) = 0$.
    Let $\str = \sym_1 \cdots \sym_\strlen$.
    Then, there must exist a $1 \leq \tstep' \leq \strlen$ such that $\pPrefix(\str_{<\tstep'}) = 0$.
    Note that
    \begin{equation}
        \pLN(\str) = \prod_{\tstep=1}^{\tstep'} \pLNSM(\sym_\tstep \mid \str_{<\tstep}) = 0
    \end{equation}
    whereas the conditional probabilities after $\tstep'$ can be arbitrarily defined since they do not affect the string having $0$ probability.
\end{proof}

\subsubsection{When Is a Locally Normalized Language Model a Language Model?}

\LNMAcronym{}s which specify distributions over strings $\pLN\left(\sym_1 \ldots \sym_\strlen\right)$ in terms of their conditional probabilities $\pLNSM(\sym_\tstep \mid \str_{<\tstep})$ for $\tstep = 1, \ldots, \finaltstep$ and $\pLNSM\left(\eos \mid \str\right)$ have become the standard in NLP literature.
However, \LNMAcronym{}s come with their own set of problems.
An advantage of normalizable globally normalized models is that they, by definition, always define a \emph{valid} probability space over $\eosalphabet$.
Although this might be counterintuitive at first, the same cannot be said for \LNMAcronym{}s---in this sense, locally normalized ``language models'' might not even be language models!
One might expect that in a \LNMAcronym{} $\pLN$, it would hold that $\sum_{\str \in \kleene{\alphabet}}\pLN\left(\str\right) = 1$.
However, this might not be the case!\ryan{maybe we could mention that it would be the case if we assumed that the LN assigned a positive probability to only a finite subset of $\Sigma^*$ or something.}
This is the issue with the terminology we brought up earlier and it brings us to the last fundamental question of this section.
\begin{aquestion}{Locally normalized language models}{locally-normalized-lm}
    When does an \LNMAcronym{} encode a language model?
\end{aquestion}
As the conditions are a bit more nuanced, it requires a longer treatment.
We explore this issue in much more detail in the next section.
\newpage{}

\section{Tight Language Models}\label{sec:lm-tightness}
We saw in the last section that any language model $\pLM$ can be converted into a locally normalized sequence model (cf. \cref{sec:locally-normalizing-lms}).
The converse, however, is \emph{not} true.
As alluded to in the previous section and as we detail in this section, there exist sets of conditional distributions $\pLN\left(\eossym \mid \eosstr\right)$ over $\kleene{\eosalphabet}$ such that $\pLN\left(\eosstr\right)$ as defined in \cref{eq:locally-normalized-model} does not represent a valid probability measure over $\kleene{\alphabet}$ (after taking into account the semantics of $\eos$), i.e., over the set of \emph{finite} strings.
Indeed, we will later show that some popular classes of locally normalized sequence models used in practice have parameter settings in which the generative process terminates with probability $< 1$.
This means that $\pLN$ ``leaks'' some of its probability mass to \emph{infinite} sequences.
This section investigates this behavior in a lot of detail.
It is based on the recent work from \citet{Du2023}.

\subsection{Tightness} \label{sec:tightness-intro}
Models whose generative process may fail to terminate are called \defn{non-tight} \citep{Chi1999}.\footnote{Tight models are also called \defn{consistent} \citep{Booth1973, Chen2018} and \defn{proper} \citep{Chi1999} in the literature.}
\begin{definition}{Tightness}{tightness}
    A locally normalized language model $\pLN$ derived from a sequence model $\pLNSM$ is called \defn{tight} if it defines a valid probability distribution over $\kleene{\alphabet}$:
    \begin{equation}
        \sum_{\str \in \kleene{\alphabet}} \pLN\left(\str\right) =
        \sum_{\str \in \kleene{\alphabet}} \left[\pLNSM\left(\eos\mid\str\right) \prod_{\tstep = 1}^{\strlen}\pLNSM\left(\sym_\tstep\mid\str_{<\tstep}\right)\right] = 1.
    \end{equation}
\end{definition}

Note that the individual conditional distributions $\pLNSM\left(\sym \mid \str\right)$ in a non-tight \LNMAcronym{} still are valid conditional distributions (i.e., they sum to one).
However, the distribution over all possible strings that they induce may not sum to $1$.
To be able to investigate this phenomenon more closely, let us first examine what the conditional probabilities of an \LNMAcronym{} actually define and how they can result in non-tightness.
We now ask ourselves: given a sequence model $\pLNSM$, what is $\pLN$?
Is $\pLN$ a language model, i.e., a distribution over $\kleene{\alphabet}$ (after taking into account the semantics of $\eos$)?
Certainly, the answer is yes if the \LNMAcronym{}'s conditional probabilities match the conditional probabilities of some known language model $\pLM$ as defined in \cref{sec:locally-normalizing-lms},\footnote{That is, $\pLM(\sym_\tstep \mid \str_{<\tstep}) = \pLN(\sym_\tstep\mid\str_{<\tstep})$ whenever the former conditional probability is well-defined under the language model $\pLM$, i.e., whenever $\sym_\tstep \in \eosalphabet$ and $\str_{<\tstep}\in\kleene{\alphabet}$ with $\pLM(\str_{<\tstep}) > 0$.} in which case $\pLN$ is specifically the language model $\pLM$ itself.
In this case clearly $\pLN\left(\kleene{\alphabet}\right) \defeq \sum_{\str\in\kleene{\alphabet}}\pLN\left(\str\right) = \sum_{\str\in\kleene{\alphabet}}\pLM\left(\str\right) = 1$.
If instead $\pLN\left(\kleene{\alphabet}\right) < 1$, the \LNMAcronym{}'s conditional probabilities do \emph{not} match the conditional probabilities of any language model $\pLM$.

To see how this can happen, we now exhibit such an \LNMAcronym{} in the following example.
\begin{example}{A non-tight $2$-gram model}{non-tight-2-gram}
    Consider the bigram model defined in \cref{fig:non-tight-2-gram-example} over the alphabet $\alphabet = \{\syma, \symb\}$.\footnote{The graphical representation of the \LNMAcronym{} depicts a so-called weighted finite-state automaton, a framework of language models we will introduce shortly. For now, it is not crucial that you understand the graphical representation and you can simply focus on the conditional probabilities specified in the figure.}
    Although the conditional probability distributions $\pLN(\cdot \mid \str_{<\idx})$ each sum to 1 over $\eosalphabet$, they fail to combine into a model $\pLN$ that sums to 1 over $\kleene{\alphabet}$ (i.e., a language model): under this model, any finite string that contains the symbol \symb{} will have probability 0, since $\pLN(\eos\mid\symb)=\pLN(\syma\mid\symb)=0$. This implies $\pLN\left(\kleene{\alphabet}\right) = \sum_{\idx=0}^\infty \pLN(\syma^\idx) = \sum_{\idx=0}^\infty (0.7)^\idx \cdot 0.1 = \frac{0.1}{1 - 0.7} = \frac{1}{3} < 1$.
\end{example}
\begin{example}{A tight $2$-gram model}{tight-2-gram}
    On the other hand, in the bigram model in \cref{fig:tight-2-gram-example}, obtained from \cref{ex:non-tight-2-gram} by changing the arcs from the \symb{} state, $\pLN\left(\kleene{\alphabet}\right)=1$.
    We can see that by calculating:
    \begin{align*}
        \probfunction(\kleene{\alphabet})
         & =\sum_{n=1}^\infty\sum_{m=0}^\infty \probfunction(\syma^n\symb^m)                                                          \\
         & = \sum_{n=1}^\infty \left(\probfunction(\syma^n)+\sum_{m=1}^\infty \probfunction(\syma^n\symb^m) \right)                   \\
         & = \sum_{n=1}^\infty \left(0.1\cdot(0.7)^{n-1} + \sum_{m=1}^\infty (0.7)^{n-1}\cdot 0.2 \cdot (0.9)^{m-1} \cdot 0.1 \right) \\
         & = \sum_{n=1}^\infty \left(0.1\cdot(0.7)^{n-1} + (0.7)^{n-1}\cdot 0.2 \cdot \frac{1}{1-0.9} \cdot 0.1 \right)               \\
         & = \sum_{n=1}^\infty \left(0.1\cdot(0.7)^{n-1} + 0.2 \cdot (0.7)^{n-1}\right)                                               \\
         & = \sum_{n=1}^\infty 0.3\cdot (0.7)^{n-1}=\frac{0.3}{1-0.7} = 1.
    \end{align*}
\end{example}
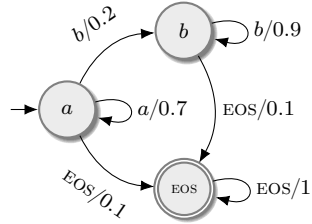
\begin{figure}[t]
    \centering
    \begin{subfigure}{\textwidth}
        \centering
        \raisebox{-0.65\height}{
            \footnotesize
            \begin{tabular}{ll}
                $\pLN(\syma\mid \bos)$  & 1   \\
                $\pLN(\syma\mid \syma)$ & 0.7 \\
                $\pLN(\symb\mid \syma)$ & 0.2 \\
                $\pLN(\eos\mid \syma)$  & 0.1 \\
                $\pLN(\symb\mid \symb)$ & 1   \\
                $\pLN(\eos\mid\eos)$    & 1
            \end{tabular}}
        \raisebox{-0.8\height}{
            \footnotesize
            \begin{tikzpicture}[minimum size=7mm]
                \node[state, initial] (a) {$\syma$};
                \node[state,above right=0.5cm and 1cm of a] (b) {$\symb$};
                \node[state,below right=0.5cm and 1cm of a,accepting] (eos) {{$\scriptstyle \eos$}};

                \draw[transition] (a) edge [loop right] node [right] {$\syma/0.7$} (a);
                \draw[transition] (b) edge [loop right] node {$\symb/1$} (b);
                \draw[transition] (eos) edge [loop right] node {$\eos/1$} (eos);
                \draw[transition] (a) edge [bend left] node [above, sloped] {$\symb/0.2$} (b);
                \draw[transition] (a) edge [bend right] node [below, sloped] {$\eos/0.1$} (eos);
            \end{tikzpicture}
        }
        \caption{A non-tight $2$-gram model.}
        \label{fig:non-tight-2-gram-example}
    \end{subfigure}
    \vspace{5mm}
    \centering
    \begin{subfigure}{\textwidth}
        \centering
        \raisebox{-0.65\height}{
            \footnotesize
            \begin{tabular}{ll}
                $\pLN(\syma\mid \bos)$  & 1   \\
                $\pLN(\syma\mid \syma)$ & 0.7 \\
                $\pLN(\symb\mid \syma)$ & 0.2 \\
                $\pLN(\eos\mid \syma)$  & 0.1 \\
                $\pLN(\symb\mid \symb)$ & 0.9 \\
                $\pLN(\eos\mid \symb)$  & 0.1 \\
                $\pLN(\eos\mid\eos)$    & 1
            \end{tabular}}
        \raisebox{-0.8\height}{
            \footnotesize
            \begin{tikzpicture}[minimum size=7mm]
                \node[state, initial] (a) {$\syma$};
                \node[state,above right=0.5cm and 1cm of a] (b) {$\symb$};
                \node[state,below right=0.5cm and 1cm of a,accepting] (eos) {{$\scriptstyle \eos$}};

                \draw[transition] (a) edge [loop right] node [right] {$\syma/0.7$} (a);
                \draw[transition] (b) edge [loop right] node {$\symb/0.9$} (b);
                \draw[transition] (b) edge [bend left] node [right] {$\eos/0.1$} (eos);
                \draw[transition] (eos) edge [loop right] node {$\eos/1$} (eos);
                \draw[transition] (a) edge [bend left] node [above, sloped] {$\symb/0.2$} (b);
                \draw[transition] (a) edge [bend right] node [below, sloped] {$\eos/0.1$} (eos);
            \end{tikzpicture}
        }
        \caption{A tight $2$-gram model.}
        \label{fig:tight-2-gram-example}
    \end{subfigure}
    \caption{Tight and non-tight bigram models, expressed as Mealy machines.
        Symbols with conditional probability of 0 are omitted.}
\end{figure}

\cref{ex:non-tight-2-gram} confirms that the local normalization does not necessarily yield $\pLN$ that is a valid distribution over $\kleene{\alphabet}$.
But if $\pLN$ is not a language model, \emph{what} is it?
It is intuitive to suspect that, in a model with $\pLN\left(\kleene{\alphabet}\right)<1$, the remainder of the probability mass ``leaks'' to infinite sequences, i.e., the generative process may continue forever with probability $>0$.
This means that, to be able to characterize $\pLN$, we will have to be able to somehow take into account infinite sequences.
We will make this intuition formal below.

Delving a bit deeper, the non-tightness of \cref{ex:non-tight-2-gram} is related to the fact that the conditional probability of $\eos$ is $0$ at some states, in contrast to \cref{ex:tight-2-gram}.
However, requiring $\pLN(\sym_\idx=\eos \mid \str_{<\idx}) > 0$ for all prefixes $\str_{<\idx}$ is neither \emph{necessary} nor \emph{sufficient} to ensure tightness.
It is not necessary because one can, for example, construct an \LNMAcronym{}
in which $\pLN(\sym_\idx=\eos \mid \str_{<\idx}) = 0.1$ when $\idx$ is even but $= 0$ otherwise.  Such a model generates only odd-length strings but is tight.
We will postpone non-sufficienty for later, where we will present specific \LNMAcronym{}s under which the conditional probability of $\eos$ is always $>0$, yet are non-tight.\anej{Add this part to RNNs and put a forward-reference here.}

\subsection{Defining the probability measure of an \texorpdfstring{\LNMAcronym{}}{LNM}}
We now rigorously characterize the kind of distribution induced by an \LNMAcronym{}, i.e., we investigate what $\pLN$ is.
As mentioned earlier, an \LNMAcronym{} can lose probability mass to the set of infinite sequences, $\alphabet^\infty$.
However, $\alphabet^\infty$, unlike $\kleene{\alphabet}$, is \emph{uncountable}, and it is due to this fact that we need to work explicitly with the \emph{measure-theoretic} formulation of probability which we introduced in \cref{sec:measure-theory}.
We already saw the peril of not treating distributions over uncountable sets carefully is necessary in \cref{ex:inf-coin-toss}---the set of all infinite sequences of coin tosses is indeed uncountable.

\paragraph{Including infinite strings and the end of string symbol.}
As we saw in \cref{ex:inf-coin-toss}, sampling successive symbols from a non-tight \LNMAcronym{} has probability $> 0$ of continuing forever, i.e., generating infinite strings.
Motivated by that, we hope to regard the \LNMAcronym{} as defining a valid probability space over $\samplespace = \kleene{\alphabet} \cup \alphabet^\infty$, i.e., both finite as well as infinite strings, and then ``relate'' it to our definition of true language models.
Notice, however, that we also have to account for the difference in the alphabets: while we would like to characterize language models in terms of strings over the alphabet $\alphabet$, \LNMAcronym{}s work over symbols in $\eosalphabet$.

\begin{figure}[t]
    \centering
    \begin{tikzpicture}[auto,
            start chain = going right,
            box/.style = {draw, rounded corners, blur shadow,fill=white, on chain, align=center, minimum height=1cm, minimum width=1.5cm}]
        \node[box,fill=ETHBlue!20] (b1)    {$\eosalphabet^\infty$};
        \node[box,fill=ETHBlue!20] (b2)   [right = 2cm of b1] {\begin{tabular}{c} Algebra \\ $\left(\eosalphabet^\infty, \overcalC\right)$\end{tabular}};
        \node[box,fill=ETHBlue!20] (b3)   [right = 4cm of b2] {Pre-measure \\ $\left(\eosalphabet^\infty, \overcalC, \probmeasure_0\right)$};
        \node[box,fill=ETHBlue!20] (b4)   [below = 4cm of b1, xshift = 1cm] {Measure \\ $\left(\eosalphabet^\infty, \sigma\left(\overcalC\right), \probmeasure\right)$};
        \node[box,fill=ETHBlue!20] (b5)   [right = 4cm of b4] {Measure (sequence model) \\ $\left(\alphabet^\infty\cup\kleene{\alphabet}, \sigma\left(\calC\right), \probmeasure'\right)$};
        \begin{scope}[rounded corners,-{Latex[length=1.8mm, width=1.4mm]}]
            \path       (b1) edge node[above, align=center, text width = 2cm] {Cylinder sets} (b2);
            \path       (b2) edge node[above, align=center, text width = 4cm] {Conditional probabilities from $\pLN$} (b3);
            \path       (b3) edge node[above, sloped, align=center, text width = 4cm] {Carath\'eodory's Extension} (b4);
            \path       (b4) edge node[above, align=center, text width = 4cm] {Random variable construction} (b5);
        \end{scope}
    \end{tikzpicture}
    \caption{The outline of our measure-theoretic treatment of \LNMAcronym{}s in this section to arrive at a precise characterization of $\pLN$. The final box corresponds to the sequence model (probability measure over $\kleene{\alphabet} \cup \alphabet^\infty$) constructed for $\pLN$.}
    \label{fig:asm-prob-measure-pipeline}
\end{figure}
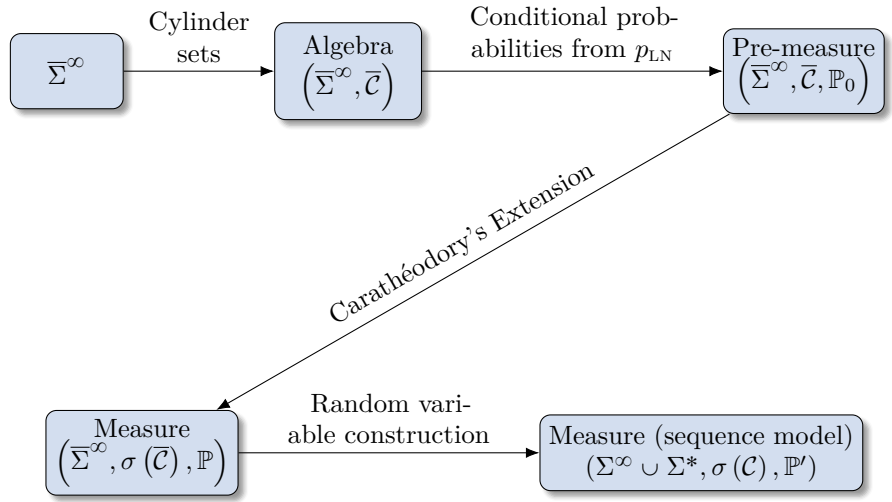

With this in mind, we now embark on our journey of discovering what $\pLN$ represents.
Given an \LNMAcronym{}, we will first need to turn its $\pLN$ into a measurable space by defining an appropriate $\sigalgebra$.
This type of distribution is more general than a language model as it works over both finite as well as infinite sequences.
To distinguish the two, we will expand our vocabulary and explicitly \emph{differentiate} between true language models and non-tight \LNMAcronym{}s.
We will refer to a distribution over $\kleene{\alphabet} \cup \alphabet^\infty$ as a sequence model.
As noted in our definition of a sequence model (cf. \cref{def:sequence-model-1}), an \LNMAcronym{} defines a probabilty measure over $\kleene{\alphabet} \cup \alphabet^\infty$.
Thus, an equivalent distribution, which will be useful for this section, would be the following.
\begin{definition}{Sequence model}{sequence-model}
    A \defn{sequence model} is a probability space over the set $\kleene{\alphabet} \cup \alphabet^\infty$.
\end{definition}
Intuitively, and we will make this precise later, the set $\alphabet^\infty\subset\kleene{\alphabet} \cup \alphabet^\infty$ in \cref{def:sequence-model} represents the \emph{event} where the sequence model is \defn{non-terminating}, i.e., it attempts to generate an infinitely long sequence.
We can then understand language models in a new sense.
\begin{definition}{Re-definition of a Language model}{language-model-eqivalent}
    A \defn{language model} is a probability space over $\kleene{\alphabet}$.
    Equivalently, a language model is a sequence model such that $\probfunction(\alphabet^\infty)=0$.
\end{definition}
Now buckle up!
Our goal through the rest of this section is to rigorously construct a probability space of a sequence model as in \cref{def:probability-measure} and \cref{def:sequence-model} which encodes the probabilities assigned by an \LNMAcronym{}.
Then, we will use this characterization to formally investigate tightness.
An outline of what this is going to look like is shown in \cref{fig:asm-prob-measure-pipeline}.

\subsubsection{Defining an Algebra over $\eosalphabet^\infty$ (Step 1)}
Since an \LNMAcronym{} produces conditional distributions over the augmented alphabet $\eosalphabet$ (first box in \cref{fig:asm-prob-measure-pipeline}) and results in possibly infinite strings, we will first construct a probability space over $\eosalphabet^\infty$, which will naturally induce a sequence model.
We will do that by first constructing an \emph{algebra} (cf. \cref{def:algebra}) over $\samplespace=\eosalphabet^\infty$ for some alphabet $\alphabet$ (second box in \cref{fig:asm-prob-measure-pipeline}).
Then, assuming we are given an \LNMAcronym{} $\pLN$ over $\eosalphabet$, we will associate the constructed algebra with a pre-measure (cf. \cref{def:pre-measure}) that is ``consistent'' with $\pLN$ (third box in \cref{fig:asm-prob-measure-pipeline}).

We will make use of the following definition to construct the algebra:
\begin{definition}{Cylinder set}{cylinder-set}
    Given any set $\sH\subseteq\eosalphabet^\idxk$, i.e., a set of sequences of symbols from $\eosalphabet$ of length $\idxk$, define its \defn{cylinder set} (of rank $\idxk$) to be
    \begin{equation} \label{eq:cylinder}
        \overC(\sH)\defeq\left\{\str\bomega~:~\str\in \sH, \bomega\in\eosalphabet^\infty\right\}
    \end{equation}
\end{definition}
In essence, a cylinder set of rank $\idxk$ is the set of infinite strings that share their $\idxk$-prefix with some string $\eosstr \in \sH\subseteq\eosalphabet^\idxk$.
In particular, for a length-$\idxk$ string $\eosstr=\eossym_1\dotsm \eossym_\idxk$, the cylinder set $\overC(\eosstr)\defeq\overC(\{\eosstr\})$ is the set of all infinite strings prefixed by $\eosstr$.\footnote{This type of cylinder set, i.e., one that is generated by a singleton, is also called a \defn{thin cylinder}.}

We denote the collection of all rank-$\idxk$ cylinder sets by
\begin{equation}
    \overcalC_\idxk\defeq\left\{\overcalC(\sH)~:~\sH\in\powerset{\eosalphabet^\idxk}\right\}
\end{equation}
and define\anej{should this start at k = 0?}
\begin{equation}
    \overcalC \defeq \bigcup_{\idxk=1}^\infty\overcalC_\idxk
\end{equation}
to be the collection of all cylinder sets over $\samplespace$.\footnote{We invite the reader to verify that $\overcalC_1\subset\overcalC_2\subset\overcalC_3\subset\dotsm$.}

The following lemma asserts $\overcalC\subseteq\powerset\samplespace$ is what we want in the second block of \cref{fig:asm-prob-measure-pipeline}.
\begin{lemma}{}{cylinder-algebra}
    $\overcalC\subseteq\powerset\samplespace$ is an algebra over $\samplespace=\eosalphabet^\infty$.
\end{lemma}
\begin{proof}
    First, $\eosalphabet^\infty=\overC(\eosalphabet^\idxk)$ for any $\idxk$, and in particular is a cylinder set of any rank.
    Secondly, given a cylinder set $\overC(\sH)$ of rank $\idxk$, i.e., $\sH\subseteq \eosalphabet^\idxk$, $\setcomplement{\big(\overC(\sH)\big)}=\overC\left(\eosalphabet^\idxk\setminus \sH\right)$. Hence, $\overcalC$ is closed under complements.
    Finally, notice that the intersection of two cylinder sets of ranks $k_1 \leq k_2$ is another cylinder set of rank $k_2$.
    Hence, $\overcalC$ is an algebra over $\Omega$.
\end{proof}
With this, the first step of \cref{fig:asm-prob-measure-pipeline} is done!

\subsubsection{Defining a Pre-measure over $\overcalC$ (Step 2)}
We are now ready to define the pre-measure $\probfunction_0$ for the cylinder algebra $\overcalC$.
Given an \LNMAcronym{} $\pLN$ and any set $\overC(\sH)\in\overcalC$, let\anej{Ryan says we should be using $\pPrefix$ here.}
\begin{align}
    \probfunction_0(\overC(\sH))\defeq\sum_{\eosstr\in \sH} \pLN(\eosstr) \label{eq:pre-measure}
\end{align}
where we have defined\looseness=-1
\begin{align}
    \pLN(\eosstr) \defeq \prod_{t=1}^\strlen \pLN(\eossym_\tstep\mid \eosstr_{<t}).
\end{align}
Note that there is a caveat here since the same cylinder set may admit different $\sH$.\footnote{For example, in the infinite coin toss model, $C(\texttt{H})=C(\{\texttt{HH},\texttt{HT}\})$.}
Before showing that $\probfunction_0$ defines a valid pre-measure, we address this and show that $\probfunction_0$ is indeed well defined.
\begin{proposition}{}{pre-measure-well-defined}
    $\probfunction_0$ as defined in \cref{eq:pre-measure} is a well-defined function.
\end{proposition}
\begin{proof}
    Suppose a cylinder set can be described by two different prefix sets: $H_1\subseteq\eosalphabet^{k_1}$ and $H_2\subseteq\eosalphabet^{k_2}$. In other words, $\overC(H_1)=\overC(H_2)$.
    Without loss of generality, assume that $k_1\leq k_2$.
    Then,
    \begin{subequations}
        \begin{align}
            \overC(H_2) & =\overC(H_1)                                                                              \\
                        & = \bigcup_{\str\in H_1} \overC(\str)                                                      \\
                        & = \bigcup_{\str\in H_1} \bigcup_{\eosstr \in \eosalphabet^{k_2-k_1}} \overC(\str\eosstr).
        \end{align}
    \end{subequations}
    All the unions above are disjoint, and hence $H_2=\bigcup_{\eosstr\in\eosalphabet^{k_2-k_1}} \{\str\eosstr: \str\in H_1\}$.
    Then, by the locally-normalizing property of $\pLN$, we have that
    \begin{align}
        \probfunction_0(\overC(H_1))=\probfunction_0(\overC(H_2)).
    \end{align}
\end{proof}

With this, we are able to state and prove the lemma which shows that $\probfunction_0$ is a pre-measure, which is what we need in the third block of \cref{fig:asm-prob-measure-pipeline}.
\begin{lemma}{}{cylinder-premeasure}
    $\probfunction_0$ is a pre-measure over $\overcalC$.
\end{lemma}

For the proof of \cref{lem:cylinder-premeasure}, we will mostly follow the proof of Theorem 2.3 in \citet{Billingsley1986}, with the exception of invoking the Tychonoff theorem directly.
This proof depends on the following lemma, which is Example 2.10 in \citet{Billingsley1986}.
We repeat the statement and proof here for the reader's convenience.
\begin{lemma}{}{pre-measure-continuity}
    Let $\probfunction_0$ be a finitely additive probability pre-measure over $\overcalC$ such that, given a decreasing sequence of sets $A_1\supset A_2\supset \dotsm$ in $\overcalC$ where $\bigcap_{\idx=1}^\infty A_\idx=\emptyset$, $\lim_{\idx\to\infty} \probfunction_0(A_\idx)=0$. Then, $\probfunction_0$ is also countably additive over $\overcalC$.
\end{lemma}
\begin{proof}
    \sloppy Let $\{A_n\}$ be a sequence of disjoint sets in $\overcalC$ such that $A=\bigcup_n A_n \in \overcalC$. Then, defining $B_n=\bigcup_{m>n} A_m$, we see that $B_1\supset B_2\supset \dotsm$ and $\bigcap_n B_n=\emptyset$. Notice that
    \begin{equation}
        A=A_1\cup B_1=A_1\cup A_2\cup B_2 = \dotsm =A_1\cup\dotsm\cup A_n \cup B_n
    \end{equation}
    for any $n$ and hence by finite additivity of $\probfunction_0$
    \begin{equation}
        \probfunction_0(A)=\probfunction_0(A_1)+\dotsm+\probfunction_0(A_n)+\probfunction_0(B_n)
    \end{equation}
    or equivalently
    \begin{equation}\label{eq:finite-countable-add}
        \probfunction_0(A_1)+\dotsm+\probfunction_0(A_n) = \probfunction_0(A)-\probfunction_0(B_n).
    \end{equation}
    Since, $B_n\downarrow\emptyset$ implies that $\probfunction_0(B_n)\downarrow0$ by assumption, taking the limits on both sides of \cref{eq:finite-countable-add} yields
    \begin{equation}
        \sum_{n} \probfunction_0(A_n)=\lim_{n\to\infty} \sum_{i\leq n}\probfunction_0(A_i)
        =\probfunction_0(A)-\lim_{n\to\infty}\probfunction_0(B_n) =\probfunction_0(A)
    \end{equation}
    which shows countable additivity. 
\end{proof}
We also recall the Tychonoff theorem.\footnote{See \S37 in \citet{Munkres2000} for a detailed and well-written treatise.}
\begin{theorem}{Tychonoff}{tychonoff}
    Let $\{\sX_\alpha\}_{\alpha\in J}$ be an indexed family of compact topologies.
    Then, their product topology $\prod_{\alpha\in J} \sX_\alpha$ is also compact.
\end{theorem}

We can now give the proof for \cref{lem:cylinder-premeasure}.
\begin{proof}[Proof of \cref{lem:cylinder-premeasure}]
    We first show that $\probfunction_0$ is finitely additive over $\overcalC$.
    Let $\sC(\sH_1)$ and $\sC(\sH_2)$ be two disjoint cylinder sets.
    By \cref{prop:pre-measure-well-defined}, we can assume they are of the same rank without loss of generality.
    Then,
    \begin{subequations}
        \begin{align}
            \sC(\sH_1)\cup \sC(\sH_2)
             & = \bigcup_{\str\in \sH_1} \{\str\bomega:\bomega\in \eosalphabet^\infty\}
            \cup \bigcup_{\str\in \sH_2} \{\str\bomega:\bomega\in \eosalphabet^\infty\}                                                                               \\
             & = \bigcup_{\str\in \sH_1 \cup \sH_2} \{\str\bomega:\bomega\in \eosalphabet^\infty\} \qquad \justification{$\sH_1$ and $\sH_2$ equal rank and disjoint} \\
             & =\sC(\sH_1\cup \sH_2)
        \end{align}
    \end{subequations}
    which leads to
    \begin{subequations}
        \begin{align}
            \probfunction_0(\sC(\sH_1)\cup \sC(\sH_2)) & = \probfunction_0(\sC(\sH_1\cup \sH_2))                    \\
                                                       & = \sum_{\str\in \sH_1\cup \sH_2} \pLN(\str)                \\
                                                       & = \probfunction_0(\sC(\sH_1))+\probfunction_0(\sC(\sH_2)).
        \end{align}
    \end{subequations}
    Hence, $\probfunction_0$ is finitely additive.

    Now, equip $\eosalphabet$ with the discrete topology.
    Since $\eosalphabet$ is finite, it is compact under the discrete topology and so is $\eosalphabet^\infty$ by \cref{thm:tychonoff}.
    Then, by properties of the product topology over discrete finite spaces, all cylinder sets in $\eosalphabet^\infty$ are compact.
    To apply \cref{lem:pre-measure-continuity}, let $\sC_1\supset \sC_2\supset \dotsm$ be a decreasing sequence of cylinder sets with empty intersection.
    Suppose to the contrary that $\probfunction_0\left(\bigcap_n \sC_n\right)>0$.
    This would imply that all $\sC_n$ are nonempty (any of these being empty would result in a measure 0).
    However, by Cantor's intersection theorem\footnote{Cantor's intersection theorem states that a decreasing sequence of nonempty compact sets have a nonempty intersection.
        A version of this result in introductory real analysis is the Nested Interval Theorem.}, $\bigcap_n \sC_n$ is nonempty, contradicting the assumption.
    Hence, $\probfunction_0\left(\bigcap_n \sC_n\right)=0$, and by \cref{lem:pre-measure-continuity}, $\probfunction_0$ is countably additive.

    With this, we have proved that $\probfunction_0$ is countably additive.
    To show that $\probfunction_0$ defines a pre-measure, we still have to show that $\probfunction_0\left(\Omega\right) = 1$.
    Recall from the proof of \cref{lem:cylinder-algebra} that $\eosalphabet^\infty=\overC(\eosalphabet^\idxk)$ for any $\idxk > 0$.
    In particular, $\eosalphabet^\infty=\overC(\eosalphabet^1) = \overC(\eosalphabet)$.
    This means that
    \begin{align}
        \probfunction_0\left(\Omega\right) & = \probfunction_0\left(\overC\left(\eosalphabet\right)\right)             \\
                                           & = \sum_{\eosstr\in \eosalphabet} \pLN\left(\eosstr\right)                 \\
                                           & = \sum_{\eossym\in \eosalphabet} \pLN\left(\eossym \mid \bos \right) = 1.
    \end{align}
    The last equality follows from local normalization of the sequence model.
\end{proof}

With this, we have successfully completed the first two steps of \cref{fig:asm-prob-measure-pipeline}!
However, we have only defined a \emph{pre}-measure over the set of \emph{infinite} $\eos$-containing sequences $\eosalphabet^\infty$.
This does not yet satisfy all the properties we would like from a probability space.
Because of that, we next \emph{extend} the constructed probability pre-measure $\probfunction_0$ into a valid probability measure $\probfunction$ to arrive to a valid probability space.

\subsubsection{Extending the Pre-measure $\probfunction_0$ into a Measure $\probfunction$ (Step 3)}

To extend $\probfunction_0$ into a measure, we will use Carath\'eodory's theorem:
\begin{theorem}{Carath\'eodory's Extension Theorem}{caratheodory}
    Given an algebra $\eventspaceA$ over some set $\samplespace$ and a probability pre-measure $\probfunction_0:\eventspaceA\to[0,1]$, there exists a probability space $(\samplespace,\eventspace,\probfunction)$ such that $\mathcal{A} \subset \eventspace$ and $\probfunction|_\mathcal{A}=\probfunction_0$.
    Furthermore, the $\sigalgebra$ $\eventspace$
    depends only on $\eventspaceA$ and is minimal and unique, which we will also denote by $\sigma(\eventspaceA)$, and the probability measure $\probfunction$ is unique.
\end{theorem}
\begin{proof}[Proof Sketch]
    First, construct an outer measure by approximation with countable coverings.
    Then, show that the collection of sets that is measurable with respect to this outer measure is a $\sigalgebra$ $\eventspace$ that contains $\eventspaceA$.
    Finally, restricting the outer measure to this $\sigalgebra$, one is then left with a probability space.
    To show minimality, one can show that $\eventspace$ is contained in any $\sigalgebra$ that contains $\eventspaceA$.
    Uniqueness is given by applying Dynkin's $\pi$-$\lambda$ theorem (Theorem 3.2 in \citealp{Billingsley1986}).

    Great care must be taken in each step involved in the outline above.
    To address these is well beyond the scope of this treatment and we refer the reader to the many excellent texts with a proof of this theorem, such as Chapter 12 in \citet{Royden1988} and Chapter 11 in \citet{Billingsley1986}.
\end{proof}

Applying Carath\'eodory's extension theorem to our cylinder algebra $\overcalC$ and pre-measure $\probfunction_0$, we see that there exists a probability space $(\eosalphabet^\infty,\sigma(\overcalC),\probfunction)$ over $\eosalphabet^\infty$ that agrees with the \LNMAcronym{} $\pLN$'s probabilities.

Phew!
This now gets us to the fourth box in \cref{fig:asm-prob-measure-pipeline} and we only have one step remaining.

\subsubsection{Defining a Sequence Model (Step 4)}
We now have to make sure that the \emph{outcome space} of the defined probability space fits the definition of a sequence model.
That is, we have to find a way to convert (map) the infinite $\eos$-containing sequences from $\eosalphabet^\infty$ into $\eos$-free finite or possibly infinite strings processed by a sequence model as required by \cref{def:sequence-model}.
We will achieve this through the use of a \emph{random variable}.

Recall from \cref{def:rv} that a random variable is a mapping between \emph{two} $\sigalgebra$s. 
Since we want our final measure space to work with the outcome space $\alphabet^\infty \cup \kleene{\alphabet}$, we, therefore, want to construct a $\sigalgebra$ over $\kleene{\alphabet}\cup\alphabet^\infty$ and then map elements from $\eosalphabet^\infty$ to $\kleene{\alphabet}\cup\alphabet^\infty$ to have the appropriate objects.
We will do so in a similar fashion as we constructed $(\eosalphabet^\infty,\overcalC)$.
Given $\sH\subseteq\alphabet^\idxk$, define a rank-$\idxk$ cylinder set in $\kleene{\alphabet}\cup\alphabet^\infty$ to be
\begin{align}
    \sC(\sH)\defeq\{\str\bomega~:~\str\in \sH, \bomega \in \kleene{\alphabet}\cup\alphabet^\infty\}. \label{eq:rv-cylinder}
\end{align}
Notice the major change from \cref{eq:cylinder}: the suffixes $\bomega$ of the elements in $C\left(\sH\right)$ now come from $\kleene{\alphabet}\cup\alphabet^\infty$ rather than $\eosalphabet^\infty$.
This means \textit{(i)} that they do not contain $\eos$ and \textit{(ii)} that they (and thus, elements of $C\left(\sH\right)$) can also be finite.
Let $\calC_\idxk$ be the set of all rank-$\idxk$ cylinder sets.
Define $\calC \defeq \bigcup_{\idxk=1}^\infty \calC_\idxk$.
Then, $\sigma\left(\calC\right)$ is a $\sigalgebra$ by the same reasoning as in \cref{lem:cylinder-algebra} and \cref{thm:caratheodory}.
We can now define the following random variable
\begin{align}\label{eq:rv}
    \rx(\bomega) = \begin{cases}
                       \bomega_{<\idxk} & \textbf{if } \text{$k$ is the first $\eos$ in } \bomega,                 \\
                       \bomega          & \textbf{otherwise} \text{     (}\textbf{if } \eos \notin \bomega\text{)}
                   \end{cases}
\end{align}
given any $\bomega\in \eosalphabet^\infty$.
The proposition below shows that $\rx$ is well-defined.
\begin{proposition}{}{}
    The function $\rx:(\eosalphabet^\infty,\sigma(\overcalC))\to(\kleene{\alphabet}\cup\alphabet^\infty, \sigma(\calC))$
    defined in \cref{eq:rv} is a measurable mapping.
\end{proposition}
\begin{proof}
    To show that $\rx$ is measurable, it suffices to show the measurability of preimage of a generating set of the $\sigalgebra$.
    Note that the set of thin cylinder sets is a generating set. Let $\sC(\str)$ be a thin cylinder set,\looseness=-1
    \begin{subequations}
        \begin{align}
            \rx^{-1}(\sC(\str))
            = & \rx^{-1}(\{\str\bomega: \bomega\in\kleene{\alphabet}\cup \alphabet^\infty\})                                                                                         \\
            = & \rx^{-1}(\{\str \bomega: \bomega\in\kleene{\alphabet} \})  \cup \rx^{-1}(\{\str \bomega: \bomega\in\alphabet^\infty \})                                              \\
            = & \left(\bigcup_{\bomega\in \kleene{\alphabet}} \overC(\str\bomega\eos)\right)  \cup \left(\overC(\str)\cap \bigcap_{\idxk=1}^\infty \setcomplement{\sA_\idxk} \right)
        \end{align}
    \end{subequations}
    Note that the sets $\sA_k$ above are defined in \cref{eq:term-at-k} which are cylinder sets representing the event of terminating at step $k$.
    Then, from the derivation above, we can see that $\rx^{-1}(\sC(\str))$ is formed by countable operations over measurable sets (cylinder sets) in $\eosalphabet^\infty$, and is hence measurable.
    So $\rx$ is a measurable function.
\end{proof}

$\rx$ intuitively ``cuts out'' the first stretch of $\bomega$ before the first $\eos$ symbol (where an \LNMAcronym{} would stop generating) or leaves the sequence intact if there is no termination symbol $\eos$.
One can check that $\pushfwdMeasure$, defined using $\probfunction$, is indeed a probability measure on $(\kleene{\alphabet}\cup\alphabet^\infty, \sigma(\calC))$ and hence $(\kleene{\alphabet}\cup\alphabet^\infty, \sigma(\calC), \pushfwdMeasure)$ is a probability space.
We have therefore arrived at the final box of \cref{fig:asm-prob-measure-pipeline} and shown that, given any \LNMAcronym{}, we can construct an associated sequence model as defined in \cref{def:sequence-model}!
In other words, given an \LNMAcronym{} $\pLN$, we have constructed a sequence model $\pSM$ (a probability space over $\alphabet^\infty \cup \kleene{\alphabet}$ where the probabilities assigned to (infinite) strings by $\pSM$ agree with $\pLN$.

\subsection{Interpreting the Constructed Probability Space}
Under the formulation of a probability space together with a random variable, useful probability quantities arise naturally and intuitively.

Consider, for example, the probability of a single finite string $\str \in \kleene{\alphabet}$, $\pushfwdMeasure\left(\str \right)$.
By definition of $\rx$, this equals
\begin{align}
    \pushfwdMeasure\left(\str \right) & = \pushfwdMeasure\left(\rx = \str \right)                                                                  \\
                                      & = \probfunction\left(\inv{\rx}\left(\str\right)\right)                                                     \\
                                      & = \probfunction\left(\text{All the sequences $\bomega\in \eosalphabet^\infty$ which map to $\str$.}\right)
\end{align}
All the sequences $\bomega\in \eosalphabet^\infty$ which map to $\str$ are sequences of the form $\bomega = \str \eos \bomega'$ for $\bomega' \in \eosalphabet^\infty$---this is exactly the cylinder $\overC\left(\str\eos\right)$!
By the definition of the probability space $\left(\eosalphabet, \sigma\left(\overC\right), \probfunction\right)$, this is
\begin{equation}
    \probfunction\left(\overC\left(\str\eos\right)\right) = \sum_{\strPrime \in \left\{\str\eos\right\}} \pLN\left(\strPrime\right) = \pLN\left(\str \eos\right)
\end{equation}
and as before $\pLN\left(\str \eos\right) = \prod_{\tstep = 1}^{\strlen}\pLN\left(\sym_\tstep\mid\str_{<\tstep}\right)\pLN\left(\eos\mid \str\right)$.

Altogether, this means that, given a finite string $\str\in\kleene{\alphabet}$, we intuitively have
\begin{equation}
    \pushfwdMeasure(\rx=\str) =\pLN(\eos \mid \str)\pLN(\str).\label{eq:prob-finite-string}
\end{equation}
Additionally, as we will show in the next section, the probability of the set of infinite strings $\pushfwdMeasure(\rx \in \alphabet^{\infty})$ is the probability of generating an infinite string.

An important technical detail left out in this discussion so far is that both the singleton set $\{\str\}$ and $\alphabet^\infty$ need to be measurable in $(\kleene{\alphabet}\cup\alphabet^\infty, \sigma(\calC))$ for the above to make sense.
This is addressed by \cref{prop:rv-singleton-measurable} and \cref{prop:rv-infinite-measurable}.

\begin{proposition}{}{rv-singleton-measurable}
    In measure space $(\kleene{\alphabet}\cup\alphabet^\infty, \sigma(\calC))$, $\{\str\}$ is measurable for all $\str\in\kleene{\alphabet}$.
\end{proposition}
\begin{proof}
    By definition in \cref{eq:rv-cylinder}, for any $\str\in\kleene{\alphabet}$,
    \begin{subequations}
        \begin{align}
            \sC(\str) & =\{\str\bomega: \bomega\in \kleene{\alphabet}\cup\alphabet^\infty \}                                \\
                      & =\{\str\bomega: \bomega\in \kleene{\alphabet} \} \cup \{\str\bomega: \bomega\in \alphabet^\infty \}
        \end{align}
    \end{subequations}
    where
    \begin{subequations}
        \begin{align}
            \{\str\bomega: \bomega\in \kleene{\alphabet} \} = \{\str\} \cup \bigcup_{a\in\alphabet} \{\str a\bomega: \bomega\in \kleene{\alphabet} \}
        \end{align}
    \end{subequations}
    and
    \begin{align}
        \{\str\bomega: \bomega\in \alphabet^\infty \} =
        \bigcup_{a\in\alphabet} \{\str a\bomega: \bomega\in \alphabet^\infty \}.
    \end{align}
    So,
    \begin{subequations}
        \begin{align}
            \sC(\str) & = \{\str\} \cup \bigcup_{a\in\alphabet} \bigg(
            \{\str a\bomega: \bomega\in \kleene{\alphabet} \}
            \cup\{\str a\bomega: \bomega\in \alphabet^\infty \}\bigg)       \\
                      & = \{\str\} \cup \bigcup_{a\in\alphabet} \sC(\str a)
        \end{align}
    \end{subequations}
    which implies that $\{\str\}=\sC(\str)\setminus \bigcup_{a\in\alphabet} \sC(\str a)$ and hence measurable.
\end{proof}

\begin{proposition}{}{rv-infinite-measurable}
    In the measure space $(\kleene{\alphabet}\cup\alphabet^\infty, \sigma(\calC))$, $\alphabet^\infty$ is measurable.
\end{proposition}
\begin{proof}
    First, the outcome space $\kleene{\alphabet}\cup\alphabet^\infty$ is measurable by definition of $\sigalgebra$. Notice that
    \begin{align}
        \alphabet^\infty = (\kleene{\alphabet}\cup\alphabet^\infty) \setminus
        \bigcup_{\str\in\kleene{\alphabet}} \{\str\}.
    \end{align}
    Since each $\{\str\}$ in the above is measurable by \cref{prop:rv-singleton-measurable} and $\kleene{\alphabet}$ is a countable set, $\alphabet^\infty$ is then measurable.
\end{proof}

Since both $\{\str\}$ and $\alphabet^\infty$ are measurable in $(\kleene{\alphabet}\cup\alphabet^\infty, \sigma(\calC))$ by \cref{prop:rv-singleton-measurable,prop:rv-infinite-measurable}, we have the following.

\begin{proposition}{}{seq-tight}
    A sequence model $(\kleene{\alphabet}\cup\alphabet^\infty, \sigma(\calC), \probfunction)$ is tight if and only if $\sum_{\str\in\kleene{\alphabet}}\probfunction(\{\str\})=1$.
\end{proposition}
\begin{proof}
    By definition, a sequence model is tight if and only if $\probfunction(\alphabet^\infty)=0$.
    By \cref{prop:rv-singleton-measurable,prop:rv-infinite-measurable}, we can write
    \begin{subequations}
        \begin{align}
            \probfunction(\kleene{\alphabet}\cup\alphabet^\infty)
             & =\probfunction(\alphabet^\infty)+\probfunction(\kleene{\alphabet})                         & \justification{countable additivity} \\
             & =\probfunction(\alphabet^\infty)+\sum_{\str\in\kleene{\alphabet}} \probfunction(\{\str\}). & \justification{countable additivity}
        \end{align}
    \end{subequations}
    Hence, a sequence model is tight if and only if $\sum_{\str\in\kleene{\alphabet}}\probfunction(\{\str\})=1$.
\end{proof}

\subsubsection{Deriving $\eos$}
As an aside, the preceding section allows us to motivate the $\eos$ token in \LNMAcronym{} as a construct that emerges naturally.
Specifically, for any $\str \in \kleene{\alphabet}$, rearranging \cref{eq:prob-finite-string}:
\begin{subequations}
    \begin{align}
        \pLN(\eos \mid \str) & = \frac{\pushfwdMeasure(\rx = \str)}{\pLN(\str)}                           \\
                             & = \frac{\pushfwdMeasure(\rx = \str)}{\pushfwdMeasure(\rx \in \sC({\str}))} \\
                             & = \pushfwdMeasure(\rx = \str \mid \rx \in \sC({\str}))
    \end{align}
\end{subequations}
where we have used $\pLN(\str) = \probfunction(\overC(\str)) = \probfunction(\rx^{-1}(\sC(\str))) = \pushfwdMeasure(\rx\in \sC(\str))$.
This means that the $\eos$ probability in an \LNMAcronym{} emerges as the conditional probability that, given that we must generate a string with a prefix $\str \in \kleene{\alphabet}$, the string is \emph{exactly} $\str$, i.e., that generation ends there.

\subsection{Characterizing Tightness} \label{sec:tightness}
Now that we have derived a measure-theoretic formalization of the probability space induced by locally-normalized models, we can use it to provide an exact characterization of tightness in \LNMAcronym{}s.
First, we consider the event
\begin{align}
    \sA_\idxk\defeq\{\bomega\in\eosalphabet^\infty:\omega_\idxk=\eos\} \label{eq:term-at-k}
\end{align}
in the probability space $(\eosalphabet^\infty,\sigma(\overcalC),\probfunction)$.
Intuitively, $\sA_\idxk$ is the event that an $\eos$ symbol appears at position $k$ in the string.
Note that under this definition the $\sA_\idxk$ are not disjoint.
For example, the string $\bomega = \syma\symb\,\eos\,\symc\,\eos\, \symd\symd\symd\symd\dotsm$ lives in the intersection of $\sA_3$ and $\sA_5$ since $\eos$ appears at both position 3 and position 5.
Using \cref{eq:term-at-k}, we can express the event consisting of all finite strings as
\begin{equation}
    \bigcup_{\idxk=1}^\infty \sA_\idxk.
\end{equation}
It follows that we can express the event of an infinite string as
\begin{equation}
    \setcomplement{\left(\bigcup_{\idxk=1}^\infty \sA_\idxk\right)}=\bigcap_{\idxk=1}^\infty \setcomplement{\sA_\idxk}.
\end{equation}
Thus, using the random variable $\rx$, we can express the probability of generating an infinite string as
\begin{subequations}
    \begin{align}
        \pushfwdMeasure(\rx\in \alphabet^\infty) & =\probfunction(\rx^{-1} (\alphabet^\infty))                                    \\
                                                 & =\probfunction\left(\bigcap_{\idxk=1}^\infty \setcomplement{\sA_\idxk}\right).
    \end{align}
\end{subequations}
Hence, we can now restate and formalize the notion of tightness.
\begin{definition}{Tight sequence model}{tight-seq-model}
    A sequence model is said to be \defn{tight} if $\pushfwdMeasure(\rx\in\alphabet^\infty) = 0$, in which case it is also a language model.
    Otherwise, we say that it is \defn{non-tight}.
\end{definition}

Note that the definition of $\sA_\idxk$ only uses a string's $\idxk$-prefix, and hence is a cylinder set of rank $\idxk$.
Recalling that the cylinder sets are measurable and so are the sets countably generated by them, we see that both the event consisting of all finite strings and the event consisting of all infinite strings are measurable.
Thus, $\probfunction\left(\bigcup_{\idxk=1}^\infty \sA_\idxk\right)$ and $\probfunction\left(\bigcap_{\idxk=1}^{\infty} \setcomplement{\sA_\idxk}\right)$ are well defined.

\subsubsection{A Lower Bound Result}
We have characterized tightness in terms of the probability of a specific event $\probfunction\left(\bigcap_{k=1}^{\infty} \setcomplement{\sA_\idxk}\right)$, a quantity we now seek to determine.
\begin{lemma}{}{durrett-435}
    If $\sum_{\idx=2}^\infty \probfunction\left(\sA_\idx\mid \bigcap_{m=1}^{\idx-1}\setcomplement{\sA_\idxm}\right)=\infty$, then
    $\probfunction\left(\bigcap_{\idxm=1}^\infty \setcomplement{\sA_\idxm}\right)=0$.
\end{lemma}
\begin{proof}
    First, recall an elementary inequality that for $x>0$,
    \begin{align}
        x-1\geq \log x \quad
        \Leftrightarrow \quad 1-x \leq \log\frac{1}{x}. \label{ineq:durrett-435}
    \end{align}
    Note that $\probfunction(\bigcap_{m=1}^n\setcomplement{A_m})>0$ for any $n$, for otherwise the conditional probabilities would be undefined.
    Let $p_n \defeq \probfunction(\bigcap_{m=1}^n\setcomplement{A_m})$.
    Then we have that $p_n>0$ for all $n$, and
    \begin{subequations}
        \begin{align}
            \infty
             & =\sum_{n=2}^\infty \probfunction(\sA_n\mid\bigcap_{m=1}^{n-1}\setcomplement{\sA_m})                                                                                                                    \\
             & =\sum_{n=2}^\infty 1-\probfunction(\setcomplement{\sA_n}\mid\bigcap_{m=1}^{n-1}\setcomplement{\sA_m})                                                                                                  \\
             & =\lim_{N\to\infty}\sum_{n=2}^N 1-\probfunction(\setcomplement{\sA_n}\mid\bigcap_{m=1}^{n-1}\setcomplement{\sA_m})                                                                                      \\
             & \leq\lim_{N\to\infty}\sum_{n=2}^N \log 1/\probfunction(\setcomplement{\sA_n}\mid\bigcap_{m=1}^{n-1}\setcomplement{\sA_m})                                 & \justification{by \cref{ineq:durrett-435}} \\
             & =\lim_{N\to\infty}\sum_{n=2}^N \log \frac{\probfunction(\bigcap_{m=1}^{n-1}\setcomplement{\sA_m})}{\probfunction(\bigcap_{m=1}^{n}\setcomplement{\sA_m})}                                              \\
             & =\lim_{N\to\infty}\sum_{n=2}^N \log \frac{p_{n-1}}{p_n}                                                                                                                                                \\
             & =\lim_{N\to\infty}\sum_{n=2}^N (\log p_{n-1}- \log p_n)                                                                                                                                                \\
             & =\lim_{N\to\infty} (\log p_1 - \log p_N)                                                                                                                                                               \\
             & =\log p_1 - \lim_{N\to\infty} \log p_N
        \end{align}
    \end{subequations}
    which implies that
    \begin{subequations}
        \begin{align}
                                 & \lim_{N\to\infty} \log p_N = -\infty                                                                                  \\
            \Leftrightarrow\quad & \lim_{N\to\infty} p_N = 0                                                                                             \\
            \Leftrightarrow\quad & \lim_{N\to\infty} \probfunction(\bigcap_{m=1}^N \setcomplement{\sA_m}) = 0                                            \\
            \Leftrightarrow\quad & \probfunction(\bigcap_{m=1}^\infty \setcomplement{\sA_m}) = 0.             & \justification{by continuity of measure}
        \end{align}
    \end{subequations}
\end{proof}

Using \cref{lem:durrett-435}, we can derive the following useful condition of tightness of a language model. Specifically, it applies when the probability of $\eos$ is lower bounded by a function that depends only on the \emph{length} and not the \emph{content} of the prefix.
\begin{proposition}{}{div-implies-tight}
    If $\pLN(\eos\mid\str)\geq \func(\tstep)$ for all $\str\in\alphabet^\tstep$ and for all $\tstep$ and $\sum_{\tstep=1}^\infty \func(\tstep)=\infty$, then $\probfunction(\bigcap_{\idxk=1}^\infty \setcomplement{\sA_\idxk})=0$. In other words, $\pLN$ is tight.
\end{proposition}
\begin{proof}
    Suppose $\pLN(\eos\mid\str)\geq \func(\tstep)$ for all $\str\in\alphabet^\tstep$. To apply \cref{lem:durrett-435}, we observe that
    \begin{subequations}
        \begin{align}
            \sA_n \cap (\setcomplement{\sA_1} \cap \dotsm \cap \setcomplement{\sA_{n-1}}) = & \{\bomega \in \eosalphabet^\infty: \omega_n=\eos\}~\cap \left(
            \bigcap_{i=1}^{n-1} \{\bomega \in \eosalphabet^\infty: \omega_i\not=\eos\}
            \right)                                                                                                                                                                 \\
            =                                                                               & \{\bomega \in \eosalphabet^\infty: \bomega=\eos, \forall~i<n, \bomega\not=\eos \}     \\
            =                                                                               & \{\bomega \in \eosalphabet^\infty: \text{$\bomega$'s first \eos is at position $n$}\}
        \end{align}
    \end{subequations}
    and similarly
    \begin{equation}
        \setcomplement{\sA_1} \cap \dotsm \cap \setcomplement{\sA_{n-1}} = \{ \bomega \in \eosalphabet^\infty: \text{There is no \eos in $\bomega$'s first $n-1$ positions}\}
    \end{equation}
    Setting $\sG\defeq\{\bomega\eos:\bomega\in \alphabet^{n-1} \}\subset\eosalphabet^n$, we get
    \begin{subequations}
        \begin{align}
            \probfunction(\sA_n \mid \setcomplement{\sA_1} \cap \dotsm \cap \setcomplement{\sA_{n-1}}) \label{eq:weighted-eos} & = \frac{
                \probfunction(\sA_n \cap (\setcomplement{\sA_1} \cap \dotsm \cap \setcomplement{\sA_{n-1}}))
            }{
                \probfunction(\setcomplement{\sA_1} \cap \dotsm \cap \setcomplement{\sA_{n-1}})
            }                                                                                                                                                                                      \\
                                                                                                                               & = \frac{\probfunction(\overC(\sG))
            }{
            \probfunction(\overC(\alphabet^{n-1}))}                                                                            & \justification{definition of $\sG$}                               \\
                                                                                                                               & =\frac{
                \sum_{\bomega\in \alphabet^{n-1}} p(\eos\mid\bomega)p(\bomega)}
            {\sum_{\bomega\in \alphabet^{n-1}} p(\bomega)}                                                                     & \justification{by \cref{eq:pre-measure}}                          \\
                                                                                                                               & \geq \frac{\sum_{\bomega\in \alphabet^{n-1}} \func(n-1)p(\bomega)
            }{
                \sum_{\bomega\in \alphabet^{n-1}} p(\bomega)
            }                                                                                                                  & \justification{definition of $\func(\tstep)$}                     \\
                                                                                                                               & = \func(n-1) \frac{\sum_{\bomega\in \alphabet^{n-1}} p(\bomega)
            }{
                \sum_{\bomega\in \alphabet^{n-1}} p(\bomega)
            }                                                                                                                                                                                      \\
                                                                                                                               & = \func(n-1).
        \end{align}
    \end{subequations}
    Since $\sum_{\tstep=0}^\infty \func(\tstep)=\infty$, \cref{lem:durrett-435} shows that the event of a string never terminating, i.e., $\bigcap_{k=1}^\infty \setcomplement{\sA_k}$ has probability measure $\probfunction(\bigcap_{k=1}^\infty \setcomplement{\sA_k})=0$.
    In other words, if the $\eos$ probability of a language model is lower bounded by a divergent sequence at every step, then the event that this language model terminates has probability 1.
\end{proof}

\subsubsection{The Borel--Cantelli Lemmata}
It turns out that \cref{prop:div-implies-tight} admits a converse statement in which we can prove a similar property of $\pLN$ by assuming that the model is tight.
To show this result, we will use a fundamental inequality from probability theory---the Borel--Cantelli lemmata.
The Borel--Cantelli lemmata are useful for our purposes because they relate the probability measure of sets of the form $\bigcap_{\idx=0}^\infty \sA_\idx$ or $\bigcup_{\idx=0}^\infty \sA_\idx$ to a series $\sum_{\idx=0}^\infty p_\idx$.
We will only state the lemmata here without supplying their proofs;\footnote{See \S2.3 in \citet{Durrett2019} or \S4 in \citet{Billingsley1986} instead.} however, we point out that \cref{lem:durrett-435} can be viewed as a parallel statement to the Borel--Cantelli lemmata and one can prove the lemmata using a very similar proof (cf. proof of Theorem 2.3.7 in \citealp{Durrett2019}).

Concretely, given a sequence of events $\{\sA_\idx\}_{\idx=1}^\infty$ in some probability space, the Borel--Cantelli lemmata are statements about the event
\begin{equation} \label{eq:defn-io}
    \{\sA_\idx \text{ i.o.}\}\defeq\bigcap_{\idxm=1}^\infty\bigcup_{\idx=\idxm}^\infty \sA_\idx
\end{equation}
where i.o. stands for ``infinitely often.''
Intuitively, $\{\sA_\idx \text{ i.o.}\}$ is the set of outcomes that appear in infinitely many sets in the collection $\{\sA_\idx\}_{\idx=1}^\infty$---they are the events that always remain in the union of an infinite family of sets no matter how many of the leading ones we remove (hence the name).
We will not use Borel--Cantelli directly, but they offer a probabilistic proof of a key result (\cref{cor:durrett-434}) which will in turn lead to the desired statement about tightness.
We formally state the first and second Borel--Cantelli lemmata below.
\begin{lemma}{Borel--Cantelli I}{bc1}
    If $\sum_{\idx=1}^\infty\probfunction(\sA_\idx) < \infty$, then $\probfunction(\sA_\idx \text{ i.o.})=0$.
\end{lemma}

\begin{lemma}{Borel--Cantelli II}{bc2}
    Assume $\{\sA_\idx\}$ is a sequence of independent events, then $\sum_{\idx=1}^\infty \probfunction(\sA_\idx) = \infty \Rightarrow \probfunction(\sA_\idx \text{ i.o.})=1$.
\end{lemma}

Using the Borel--Cantelli lemmata, we can prove the following useful fact.
\begin{corollary}{}{durrett-434}
    Given a sequence $\{p_\idx\}$ where $p_\idx\in[0,1)$. Then,
    \begin{equation}
        \prod_{\idx=1}^\infty (1-p_\idx) = 0 \Longleftrightarrow \sum_{\idx=1}^\infty p_\idx=\infty.
    \end{equation}
\end{corollary}
To show \cref{cor:durrett-434}, we first show the following simple consequence of Borel--Cantelli.

\begin{corollary}{}{io-implies-div}
    If $\probfunction(\sA_n \text{ i.o.})=1$, then $\sum_{n=1}^\infty \probfunction(\sA_n) = \infty$.
\end{corollary}
\begin{proof}
    Suppose to the contrary that $\sum_{n=1}^\infty \probfunction(\sA_n) < \infty$, then, by Borel--Cantelli I (\cref{lem:bc1}), $\probfunction(\sA_n \text{ i.o.})=0$, which contradicts the assumption.
    Hence, $\sum_{n=1}^\infty \probfunction(\sA_n) = \infty$.
\end{proof}

\begin{proof}
    We can use a product measure to construct a sequence of independent events $\{\sA_n\}_{n=1}^{\infty}$ such that $\probfunction(\sA_n)=p_n$.
    (The product measure ensures independence.)
    Then, by definition in \cref{eq:defn-io},
    \begin{align}
        \setcomplement{\{\sA_n \text{ i.o.}\}} = \bigcup_{m=1}^\infty \bigcap_{n\geq m} \setcomplement{\sA_n}
    \end{align}
    So,
    \begin{subequations}
        \begin{align}
            1-\probfunction(\sA_n \text{ i.o.}) & =\probfunction\left(\bigcup_m \bigcap_{n\geq m} \setcomplement{\sA_n}\right)                                                                   \\
                                                & =\lim_{m\to\infty} \probfunction\left(\bigcap_{n\geq m} \setcomplement{\sA_n}\right)                                                           \\
                                                & =\lim_{m\to\infty} \prod_{n\geq m} \probfunction(\setcomplement{\sA_n})              & \justification{$\sA_n$ are independent by construction} \\
                                                & =\lim_{m\to\infty} \prod_{n\geq m} (1-p_n)
        \end{align}
    \end{subequations}

    \paragraph{$(\Rightarrow)$:} Assume $\prod_{n=1}^\infty (1-p_n) = 0$. Then, for any $m$,
    \begin{equation}
        0= \prod_{n\geq 1} (1-p_n) =\underbrace{\left(\prod_{1\leq n<m} (1-p_n)\right)}_{>0}
        \left(\prod_{n\geq m} (1-p_n)\right)
    \end{equation}
    So it must the case that, for any $m$, $\prod_{n\geq m} (1-p_n)=0$. Therefore,
    \begin{equation}
        1-\probfunction(\sA_n \text{ i.o.}) = \lim_{m\rightarrow \infty} \prod_{n\geq m} (1-p_n) = 0
    \end{equation}
    which implies $\probfunction(\sA_n \text{ i.o.})=1$.
    \Cref{cor:io-implies-div} implies that $\sum_{n=1}^{\infty} p_n=\infty$.

    \paragraph{$(\Leftarrow)$:} Assume $\sum_{n=1}^\infty p_n=\infty$.
    Then by Borel--Cantelli II (\cref{lem:bc2}), $\probfunction(\sA_n \text{ i.o.})=1$ which implies
    \begin{equation}
        0 = 1-\probfunction(\sA_n \text{ i.o.}) = \lim_{m\rightarrow \infty} \prod_{n\geq m} (1-p_n)
    \end{equation}
    Observe that $\Big\{\prod_{n\geq m} (1-p_n)\Big\}_m$ is a non-decreasing sequence in $m$; to see this, note that as $m$ grows larger we multiply strictly fewer values $(1 - p_n) \in (0, 1]$.
    However, since we know the sequence is non-negative and tends to $0$, it follows that for \emph{any} $m$, we have
    \begin{align}
        \prod_{n\geq m} (1-p_n) =0.
    \end{align}
    It follows that, for any $m$, we have
    \begin{equation}
        \prod_{n=1}^\infty (1-p_n) = \prod_{n<m} (1-p_n) \underbrace{\prod_{n\geq m} (1-p_n)}_{=0} = \prod_{n<m} (1-p_n) \cdot 0 = 0.
    \end{equation}
\end{proof}

We now turn to proving a more general version of \cref{prop:div-implies-tight}, which would imply its converse.
First, we define the following quantity
\begin{align}\label{eq:ptildeeos-set}
    \pdensTildeEOS(\tstep)\defeq\probfunction(\sA_\tstep\mid \setcomplement{\sA_1}\cap \dotsm \cap \setcomplement{\sA_{\tstep-1}})
\end{align}
which can be viewed as the \eos probability at step $\tstep$, given that \eos was not generated at any earlier step.
One can also show that, when $\pdensTildeEOS(\tstep)$ is defined, it has the same value as
\begin{align}\label{eq:ptildeeos}
    \pdensTildeEOS(\tstep)=\frac{
        \sum_{\bomega\in \alphabet^{\tstep-1}} \pLN(\bomega)\pLN(\eos\mid\bomega)}
    {\sum_{\bomega\in \alphabet^{\tstep-1}} \pLN(\bomega)\phantom{\pLN(\eos\mid\bomega)}},
\end{align}
which one can see as the weighted average probability of terminating at a string of length $\tstep$.

We can now completely characterize the tightness of an \LNMAcronym{} with the following theorem.
\begin{theorem}{A sufficient condition for tightness}{lm-tight-main}
    An \LNMAcronym{} is tight if and only if $\pdensTildeEOS(\tstep)=1$ for some $\tstep$ or $\sum_{\tstep=1}^\infty \pdensTildeEOS(t)=\infty$.
\end{theorem}
\begin{proof}
    Recall the definition of $\pdensTildeEOS$, as previously defined in \cref{eq:ptildeeos-set}, is
    \begin{align}
        \pdensTildeEOS(\tstep)\defeq\probfunction(\sA_\tstep\mid \setcomplement{\sA_1}\cap \dotsm \cap \setcomplement{\sA_{\tstep-1}}).
    \end{align}

    \paragraph{Case 1.} Suppose that $\pdensTildeEOS(\tstep)<1$ for all $\tstep$. Consider the termination probability again:
    \begin{subequations}
        \begin{align}
            \probfunction\left(\bigcap_{\tstep=1}^\infty \setcomplement{\sA_\tstep}\right)
             & = \lim_{\finaltstep\to\infty} \probfunction\left(\bigcap_{\tstep=1}^\finaltstep \setcomplement{\sA_\tstep}\right)                                                               \\
             & = \lim_{\finaltstep\to\infty} \prod_{\tstep=1}^\finaltstep \probfunction(\setcomplement{\sA_\tstep} \mid \setcomplement{\sA_1} \cap \dotsm \cap \setcomplement{\sA_{\tstep-1}}) \\
             & = \lim_{\finaltstep\to\infty} \prod_{\tstep=1}^\finaltstep (1-\widetilde{p}_\eos(\tstep))                                                                                       \\
             & = \prod_{\tstep=1}^\infty (1-\widetilde{p}_\eos(\tstep)). \label{eq:term-prob-inf-prod}
        \end{align}
    \end{subequations}
    In the above, we have assumed that $\probfunction(\setcomplement{\sA_1} \cap \dotsm \cap \setcomplement{\sA_{\tstep}})>0$ for all $\tstep$, which is true by assumption that $\pdensTildeEOS(\tstep)<1$.
    Hence, by \cref{cor:durrett-434}, \cref{eq:term-prob-inf-prod} is $0$ if and only if $\sum_\tstep \widetilde{p}_\eos(\tstep)=\infty$.

    \paragraph{Case 2.} If $\widetilde{p}_\eos(\tstep)=1$ is true for some $\tstep=\tstep_0$, then $\probfunction(\setcomplement{\sA_1} \cap \dotsm \cap \setcomplement{\sA_{\tstep_0}})=0$ and hence $\probfunction\left(\bigcap_{t=1}^\infty \setcomplement{\sA_\tstep}\right)=0$ and such a language model is guaranteed to terminate at $\tstep_0$.
\end{proof}

The first condition intuitively says that there exists a step $\tstep$ at which the \LNMAcronym{} will stop with probability 1.
If the first case of the condition does not hold, the second case can be checked since its summands will be well-defined (the conditional probabilities in \cref{eq:ptildeeos} will not divide by $0$).
We remark that \cref{thm:lm-tight-main} is a generalization of \cref{prop:div-implies-tight} since if $\pdensTildeEOS(\tstep)$ is lower-bounded by $f(\tstep)$ whose series diverges, its own series would also diverge.
However, since $\pdensTildeEOS(\tstep)$ involves the computation of a partition function in its denominator, it is most likely intractable to calculate \citep{Lin2021, Lin2022}.
Hence, \cref{prop:div-implies-tight} will be the main tool for determining tightness when we explore concrete language modeling frameworks later.

\Anej{Add example of $\pdens\left(a \mid a^{t}\right) = 1 - \frac{1}{t + 1}$ and $\pdens\left(\eos \mid a^{t}\right) = \frac{1}{t + 1}$ as a tight model and $t^2$ as non-tight.}

We have now very thoroughly defined the notion of language model tightness and provided sufficient and necessary conditions for an \LNMAcronym{} or a sequence model to be tight.
In the next sections, we start our exploration of concrete computational models of language, from the very simple and historically important finite-state language models, their neural variants, to the modern Transformer architectures.
For each of them, we will also individually discuss their tightness results and conditions.
\newpage{}

\chapter{Modeling Foundations} \label{chapter:modeling-founds}

The previous chapter introduced the fundamental measure-theoretic characteristics of language modeling.
We will revisit those over and over as they will serve as the foundations on which subsequent concepts are built.

In this chapter, we turn our attention to \emph{modeling} foundations, that is, the decisions we face when we want to \emph{build} a distribution over strings and \emph{learn} the appropriate parameters for that distribution.
We first discuss \emph{how} to parameterize a distribution over strings (\cref{sec:representation-based-modeling}), what it means to \emph{learn} good parameters, and how this can be done with modern optimization techniques and objectives (\cref{sec:learning}).

Continuing our framing of the notes in terms of questions, we will try to address the following:
\begin{aquestion}{Parametrizing a sequence model}{parametrization}
    How can a sequence model be parameterized?
\end{aquestion}\clara{its not quite obvious what it means to be parameterized. I'd recommend giving a concrete defintion, like function that's described by free parameters $\theta$\response{Anej} I'd keep very formal definitions outside of the intro section. Do you think the added sentence below is enough?}
We introduce a more formal definition of a ``parametrized model'' later.
For now, you can simply think of it as a function $\pdens_\params: \kleene{\eosalphabet} \to \R$ described by some \emph{free parameters} $\params \in \paramspace$ from a parameter space $\paramspace$.
This means that the values that $\pdens_\params$ maps its inputs to might depend on the choice of the parameters $\params$---the presence of parameters in a model, therefore,  allows us to \emph{fit} them, which in our context specifically, means choosing them to maximize some objective with respect to data.
This raises the following question:
\begin{aquestion}{Training a model}{}
    Given a parameterized model and a  dataset, how can model parameters be chosen to reflect the  dataset as well as possible?
\end{aquestion}
We begin with \cref{qt:parametrization}.
\newpage{}

\section{Representation-based Language Models}
\label{sec:representation-based-modeling}
\label{sec:general-framework}

Most modern language models are defined as locally normalized models.
However, in order to define locally normalized language model, we first define a sequence model $\pLNSM\left(\eossym\mid\eosstr\right)$.
Then, we prove that the specific parameterization used in $\pLNSM\left(\eossym\mid\eosstr\right)$ encodes a tight locally normalized language model.
However, as we demonstrated in \cref{ex:non-tight-2-gram}, not all sequence models encode tight locally normalized language models in the sense of \cref{def:tightness}.
So far, however, we have only talked about this process abstractly.
For example, we have proven that every language model can be locally normalized and we have also given necessary and sufficient conditions for when a sequence model encodes a tight locally normalized language model.
In this section, we start making the abstraction more concrete by considering a very general framework for parameterizing a locally normalized language model through sequence models $\pLNSM\left(\eossym\mid\eosstr\right)$.
We will call this the \defn{representation-based language modeling} framework.

In the representation-based language modeling framework, each conditional distribution in a sequence model $\pLNSM\left(\eossym\mid\eosstr\right)$ directly models the probability of the next symbol $\eosstr \in \eosalphabet$ given the context $\eosstr$---in other words, it tells us how likely $\eossym$ is to appear in the context of $\eosstr$.\footnote{Unless explicitly stated otherwise, we use the phrase ``in the context of'' to imply given \emph{prior} context---i.e., when discussing probability distributions,  this refers to the distribution $\pLNSM\left(\eossym_\tstep \mid \eosstr_{<\tstep}\right)$ with $\eossym = \eossym_\tstep$.
    We will also see examples of models which specify the conditional probabilities in terms of symbols that do not necessarily appear before the current one.}
For example, given the string $\eosstr = \textexample{Papa eats caviar with a}$,
we would like $\pLN\left(\eossym\mid \eosstr\right)$ to capture that \textexample{spoon}
is more likely than \textexample{fork} At the same time, since eating caviar with a fork is technically possible, we would also like $\pLN\left(\eossym\mid \eosstr\right)$ to capture that \textexample{fork} is likelier than, for example, \textexample{pencil}.

However, it is not \textit{a-priori} clear how we should model $\pLNSM\left(\eossym\mid\eosstr\right)$ concretely.
We want to define a function that can map contexts $\eosstr$ to a distribution over possible continuations $\eossym$ with the caveat that this distribution can be easily adjusted, i.e., we can optimize its parameters with some objective in mind (cf. \cref{sec:learning}).
We will do this by adopting a very general idea of defining $\pLNSM\left(\eossym\mid\eosstr\right)$ in terms of similarity between representations that represent the symbol $\eossym$ and the context $\eosstr$.
The more compatible the symbol $\eossym$ is with the context $\eosstr$, the more probable it should be.
Intuitively, going from the example above, this means that \textexample{spoon} should be more similar to \textexample{Papa eats caviar with a} than \textexample{fork} should be, and that should still be more similar than \textexample{pencil}.
On the other hand, notice that this also means that \textexample{spoon} and \textexample{fork} should be closer together than any of them to \textexample{pencil}.

One possibility for doing this is by \emph{embedding} individual symbols $\eossym$ and all possible contexts $\eosstr$ as vectors in a Hilbert space, i.e., a complete vector space endowed with an inner product\index{vector representation}.
Once we embed the symbols and contexts in such a space, we can talk about how similar they are.
We will first describe how this can be done abstractly \cref{sec:representations} and then discuss how exactly vector representations can be used when defining \emph{discrete} probability distributions over the symbols in \cref{sec:representation-lms} by taking into account the notion of similarities between vectors.
We discuss methods for \emph{learning} representations later in this chapter (\cref{sec:learning}) and in \cref{ch:nn-lms}.

\subsection{Vector Space Representations} \label{sec:representations}

It is not immediately obvious how to measure the similarity or compatibility between two symbols, two contexts or a symbol and a context\ryan{We should standardize whether it's called context or history.}.
However, such a notion is required as part of our intuitive desiderata for $\pLNSM\left(\eossym\mid \eosstr\right)$.
We begin by stating an important guiding principle, which we describe in detail next and use heavily throughout the rest of the notes.
\begin{principle}{Representation Learning}{repr-learning}
    The \defn{good representation principle} states that the success of a machine learning model depends---in great part---on the representation that is chosen (or learned) for the objects that are being modeled.
    In the case of language modeling, the two most salient choice points are the representations chosen for the symbols, elements of $\eosalphabet$, and the representations chosen for the contexts, elements of $\kleene{\eosalphabet}$.
\end{principle}
Learning vector representations from data where individual entities are represented in some \defn{representation space}\index{representation space} (i.e., a Hilbert space) has a rich history in NLP and machine learning in general \citep{bengio2013representation}.

To discuss the representations of symbols and strings more formally, we first introduce the notion of a \defn{Hilbert space}, which leads us to a useful geometric manner to discuss the similarity and compatibility of symbols and contexts.
We first start with some more basic definitions.
A vector space over a field $\field$ is a set $\vectorspace$ together with two binary operations that satisfy certain axioms.
The elements of $\field$ are often referred to as \defn{scalars} and the elements of $\vectorspace$ as \defn{vectors}.
The two operations in the definition of a vector space are the \emph{addition of vectors} and \emph{scalar multiplication of vectors}.
\begin{definition}{Vector space}{}
    A \defn{vector space} over a field $\field$ is a set $\vectorspace$ together with two binary operations that satisfy the following axioms:
    \begin{enumerate}
        \item \textbf{Associativity} of vector addition: for all $\vv, \vu, \vq \in \vectorspace$
              \begin{equation}
                  \left(\vv + \vu\right) + \vq = \vv + \left(\vu + \vq\right)
              \end{equation}
        \item \textbf{Commutativity} of vector addition: for all $\vv, \vu \in \vectorspace$
              \begin{equation}
                  \vv + \vu = \vu + \vv
              \end{equation}
        \item \textbf{Identity} element of vector addition: there exists $\zero \in \vectorspace$ such that for all $\vv \in \vectorspace$
              \begin{equation}
                  \vv + \zero = \vv
              \end{equation}
        \item \textbf{Inverse} elements of vector addition: for every $\vv \in \vectorspace$ there exists a $-\vv \in \vectorspace$ such that
              \begin{equation}
                  \vv + (-\vv) = \zero
              \end{equation}
        \item \textbf{Compatibility} of scalar multiplication with field multiplication: for all $\vv \in \vectorspace$ and $x, y \in \field$
              \begin{equation}
                  x \left(y \vv\right) = \left(xy\right) \vv
              \end{equation}
        \item \textbf{Identity} element of scalar multiplication: for all $\vv \in \vectorspace$
              \begin{equation}
                  1 \vv = \vv
              \end{equation}
              where $1$ is the multiplicative identity in $\field$.
        \item \textbf{Distributivity} of scalar multiplication with respect to vector addition: for all $x \in \field$ and all $\vu, \vv \in \vectorspace$
              \begin{equation}
                  x \left(\vv + \vu\right) = x \vv + x \vu
              \end{equation}
        \item \textbf{Distributivity} of scalar multiplication with respect to field addition: for all $x, y \in \field$ and all $\vv \in \vectorspace$
              \begin{equation}
                  \left(x + y\right) \vv = x \vv + y \vv
              \end{equation}
    \end{enumerate}
\end{definition}
In almost all practical cases, $\field$ will be $\R$ and $\vectorspace$ will be $\R^\hiddDim$ for some $\hiddDim \in \N$.

An important characteristic of a vector space is its \defn{dimensionality}, which, informally, corresponds to the number of independent directions---\defn{basis vectors}---in the space.
Any $\vv \in \vectorspace$ can be expressed as a \emph{linear combination} of the $\vspacedim$ basis vectors.
The coefficients of this linear combination can then be combined into a $\vspacedim$-dimensional \defn{coordinate vector} in $\field^\vspacedim$.
Vector spaces, therefore, allow us to talk about their elements in terms of their expressions with respect to the basis vectors.
Inner product spaces additionally define an \defn{inner product}, mapping pairs of elements of the vector space to scalars.
More formally, it is a vector space together with a map $\innerProd{\cdot}{\cdot}$ (the inner product) defined as follows.
\begin{definition}{Inner product space}{}
    An \defn{inner product space} is a vector space $\vectorspace$ over a field $\field$ coupled with a map
    \begin{equation}
        \innerProd{\cdot}{\cdot}: \vectorspace \times \vectorspace \to \field
    \end{equation}
    such that the following axioms hold
    \begin{enumerate}
        \item \textbf{Conjugate symmetry}: for all $\vv, \vu \in \vectorspace$
              \begin{equation}
                  \innerProd{\vv}{\vu} = \overline{\innerProd{\vu}{\vv}}
              \end{equation}
              where $\overline{x}$ denotes the \defn{conjugate} of the element $x \in \field$.
        \item \textbf{Linearity} in the first argument: for all $\vv, \vu, \vz \in \vectorspace$ and $x, y \in \field$
              \begin{equation}
                  \innerProd{x \vv + y \vu}{\vz} = x\innerProd{\vv}{\vz} + y\innerProd{\vu}{\vz}
              \end{equation}
        \item \textbf{Positive-definiteness}: for all $\vv \neq 0$
              \begin{equation}
                  \innerProd{\vv}{\vv} > 0
              \end{equation}
    \end{enumerate}
\end{definition}

Inner products are often defined such that they capture some notion of similarity of the vectors in $\vectorspace$.
We will use this when formally defining $\pLNSM\left(\eossym\mid\eosstr\right)$ in \cref{sec:next-symbol-compat}.

Every inner product on a real or complex vector space induces a vector norm defined as follows.
\begin{definition}{Norm}{}
    Given a vector space $\vectorspace$ over $\R$ or $\C$ and an inner product $\innerProd{\cdot}{\cdot}$ over it, the \defn{norm} induced by the inner product is defined as the function $\norm{\cdot}: \vectorspace \to \Rnonnegative$ where
    \begin{equation}
        \norm{\vv} \defeq \sqrt{\innerProd{\vv}{\vv}}.
    \end{equation}
\end{definition}

A Hilbert space is then an inner product space in which all sequences of elements satisfy a useful property with respect to the norm defined by the inner product: every convergent series with respect to the norm converges to a vector in $\vectorspace$.\clara{would add notation here ($\vv_i$)}\ryan{This confuses me a bit. We should discuss it.}
\begin{definition}{Hilbert space}{}
    A \defn{Hilbert space} is an inner product space that is \defn{complete} with respect to the norm defined by the inner product.
    An inner product space is complete with respect to the norm if every Cauchy sequence (an absolutely convergent sequence, i.e., a sequence whose elements become arbitrarily close to each other) converges to an element in $\vectorspace$.
    More precisely, an inner product space is complete if, for every series
    \begin{equation}
        \sum_{\idx = 1}^{\infty} \vv_\idx
    \end{equation}
    such that
    \begin{equation}
        \sum_{\idx = 1}^{\infty} \norm{\vv_\idx} < \infty,
    \end{equation}
    it holds that
    \begin{equation}
        \sum_{\idx = 1}^{\infty} \vv_\idx \in \vectorspace.
    \end{equation}
\end{definition}

Note that even if an inner product space $\vectorspace$ is not necessarily a Hilbert space, $\vectorspace$ can always be \emph{completed} to a Hilbert space.
\begin{theorem}{Completion theorem for inner product spaces}{}
    Any inner product space can be \emph{completed} into a Hilbert space.
\end{theorem}
We omit the proof for this theorem\clara{is it because its a proof found in many textbooks? Should probably point to one of those}.
More precisely, the inner product space can be completed into a Hilbert space by completing it with respect to the norm induced by the inner product on the space.
For this reason, inner product spaces are also called pre-Hilbert spaces.

To motivate our slightly more elaborate treatment of representation spaces, we consider an example of a model which falls under our definition of a representation-based language model but would be ill-defined if it worked\clara{what does it mean to 'work' under a space?} under any space with fewer axioms than a Hilbert space.

\begin{example}{A series of representations}{sqrt2}
    Recurrent neural networks are a type of neural network that sequentially process their input and compute the output (context representation) at time step $\tstep$ based on the output at time step $\tstep - 1$: $\hiddState_\tstep = \vfunc\left(\hiddState_{\tstep - 1}, \sym_\tstep \right)$.
    A formal definition of a recurrent neural network, which we provide in \cref{sec:rnns}, is not required at the moment.
    However, note that a recurrent neural network with one-dimensional representations $h$ could, for example, take the specific form
    \begin{equation} \label{eq:sqrt2}
        h_\tstep = \frac{1}{2}h_{\tstep - 1} + \frac{1}{h_{\tstep - 1}}
    \end{equation}
    with $h_0 = 2$.

    Suppose we chose the inner product space $\Q$ over the field $\Q$ for our representation space.
    All elements of the sequence $h_\tstep$ are indeed rational numbers.
    However, the limit of the sequence, which can be shown to be $\sqrt{2}$, is \emph{not} in the inner product space!
    This shows that $\Q$ is not a Hilbert space and that we must, in full generality, work with Hilbert spaces whenever we are dealing with possibly infinite sequences of data.
    The reason this is especially relevant for language modeling is the need to consider arbitrarily long strings (contexts), whose representations we would like to construct in a way similar to \cref{eq:sqrt2}.
    Such representations can, therefore, approach a limiting representation \emph{outside} the space whenever the representation space does not satisfy the axioms of a Hilbert space.
\end{example}

A summary of the utilities of the three algebraic spaces introduced in this subsection is summarized in \cref{tab:vector-inner-hilbert}.
\begin{table}
    \centering
    \begin{tabular}{c p{0.65\textwidth}}
        \toprule
        Space               & Utility                                                                                                                                                 \\
        \midrule
        Vector space        & A space in which representations of symbols and string live. It also allows the expression of the vector representations in terms of the basis vectors. \\
        Inner product space & Defines an inner product, which defines a norm and can measure similarity.                                                                              \\
        Hilbert space       & There are no ``holes'' in the representation space with respect to the defined norm, since all convergent sequences converge into $\vectorspace$.       \\
        \bottomrule
    \end{tabular}
    \caption{The utility of different spaces introduced in this section.}
    \label{tab:vector-inner-hilbert}
\end{table}

\subsubsection{Representation Functions}

We can now introduce the notion of a general representation function.
\begin{definition}{Representation function}{} \label{def:rep-function}
    Let $\sS$ be a set and
    $\vectorspace$ a Hilbert space over some field $\field$.
    A \defn{representation function} $\vfunc$ for the elements of $\sS$ is a function of the form
    \begin{equation}
        \vfunc\colon \sS \mapsto \vectorspace.
    \end{equation}
\end{definition}
The dimensionality of the Hilbert space of the representations, $\hiddDim$, is determined by the modeler.
In NLP, $\hiddDim$ usually ranges between $10$ to $10000$.

Importantly, in the case that $\sS$ is finite, we can represent a representation function as a \emph{matrix} $\embedMtx \in \R^{|\sS| \times \hiddDim}$ (assuming $\vectorspace = \R^{\hiddDim}$ where the $\idx^{\text{th}}$ row corresponds to the representation of the $\idx^{\text{th}}$ element of $\sS$.
This method for representing $\vfunc$ is both more concise and will be useful for integrating the symbol representation function into a model, where matrix multiplications are often the most efficient way to implement such functions on modern hardware.

This is the case for the representations of the individual symbols $\eossym$ from $\eosalphabet$, where the representation function, which we will denote as $\embedding{\cdot}$, is implemented as a lookup into the embedding matrix $\embedMtx \in \R^{|\eosalphabet| \times \hiddDim}$, i.e., $\embedding{\eossym} = \embedMtx_\eossym$.\footnote{Here, we use the notation $\embedMtx_{\eossym}$ to refer to the lookup of the row in $\embedMtx$ corresponding to $\eossym$.}
In this case, we will also refer to $\embedding{\cdot}$ as the embedding function.
\begin{definition}{Symbol embedding function}{output-symbol-embedding-function}
    Let $\alphabet$ be an alphabet.
    An \defn{embedding function} $\embedding{\cdot}\colon \eosalphabet \to \R^\hiddDim$ is a representation function of individual symbols $\eossym \in \eosalphabet$.
\end{definition}
The representations $\embedding{\eossym}$ are commonly referred to as \defn{embeddings}, but, for consistency, we will almost exclusively use the term representations in this text. Let us first consider possibly the simplest way to represent discrete symbols with real-valued vectors: \defn{one-hot encodings}.

\begin{example}{One-hot encodings}{one-hot-encoding}
    Let  $\ordering: \eosalphabet \to \left\{1, \ldots, |\eosalphabet|\right\}$ be a bijection (i.e., an ordering of the alphabet, assigning an index to each symbol in $\alphabet$).
    A one-hot encoding $\onehot{\cdot}$ is a representation function which assigns the symbol $\eossym \in \eosalphabet$ the $\ordering\!\left(\eossym\right)^\text{th}$ basis vector:
    \begin{equation}
        \onehot{\sym} \defeq \vd_{\ordering\left(\sym\right)},
    \end{equation}
    where here $\vd_{\idx}$ is the $\idx^{\text{th}}$ canonical basis vector, i.e., a vector of zeros with a $1$ at position $\idx$.

\end{example}

While one-hot encodings are an easy way to create vector representations of symbols, they have a number of drawbacks. First, these representations are relatively large---we have $\hiddDim=|\eosalphabet|$---and \emph{sparse}, since only one of the dimensions is non-zero. Second, such representations are not ideal for capturing the variation in the \emph{similarity} between different words. For example, the cosine similarity---a metric we will motivate in the next section for measuring the similarity between symbol representations---between symbols' one-hot encodings is zero for all non-identical symbols.
Ideally, we would like symbol representations to encode semantic information, in which case, a metric such as cosine similarity could be used to quantify semantic similarity.
This motivates the use of more complex representation functions, which we subsequently discuss.

While most systems use this standard way of defining individual symbol representations using the embedding matrix, the way that the \emph{context} is encoded (and what even is considered as context) is really the major difference between the different architectures which we will consider later.
Naturally, since the set of all contexts is infinite, we cannot simply represent the representation function with a matrix.
Rather, we define the representation of a context $\eosstr$ through an encoding function.
\begin{definition}{Context encoding function}{encoding-function}
    Let $\alphabet$ be an alphabet.
    A \defn{context encoding function} $\enc\left(\cdot\right)\colon \kleene{\eosalphabet} \to \R^\hiddDim$ is a representation function of strings $\eosstr \in \kleene{\eosalphabet}$.\footnote{Note that, to be completely consistent, the encoding function should be defined over the set $\kleene{\left(\eosalphabet \cup \set{\bos}\right)}$ to allow for the case when $\sym_0 = \bos$. However, unlike $\eos$, we do not necessarily require $\bos$ in any formal setting, which is why we leave it out. We apologize for this inconsistency.}
\end{definition}
We will refer to $\enc(\eosstr)$ as the \defn{encoding} of $\eosstr \in \kleene{\alphabet}$.
In the general framework, we can simply consider the encoding function $\enc$ to be a black box---however, a major part of \cref{ch:nn-lms} will concern defining specific functions $\enc$ and analyzing their properties.

With this, we now know how we can represent the discrete symbols and histories as real-valued vectors.
We next consider how to use such representations for defining probability distributions over the next symbol.


\subsection{Compatibility of Symbol and Context} \label{sec:next-symbol-compat}


Inner products naturally give rise to the geometric notion of angle, by giving us the means to measure the similarity between two representations.
Concretely, the smaller the angle between the two representations is, the more similar the two representations are.
In a Hilbert space, we \emph{define} the cosine of the angle $\theta$ between the two representations
\begin{equation}
    \cos\left(\theta\right) \defeq \frac{\innerProd{\vu}{\vv}}{\norm{\vu}\norm{\vv}}.
\end{equation}
The Cauchy--Schwartz inequality immediately gives us that $\cos\left(\theta\right) \in [-1, 1]$ since
$-\norm{\vu}\norm{\vv}\leq \innerProd{\vu}{\vv} \leq \norm{\vu}\norm{\vv}$.
Traditionally, however, we take the \emph{unnormalized} cosine similarity as our measure of similarity, which simply corresponds to the inner product of the Hilbert space.
\Anej{Add geometric similarity interpretation example, e.g., embeddings and encodings on a plane}


Given a context representation $\enc\left(\eosstr\right)$, we can compute its inner products with all symbol representations $\embedding{\eossym}$:
\begin{equation}
    \innerProd{\embedding{\eossym}}{\enc\left(\eosstr\right)}.
\end{equation}
which can be achieved simply with a matrix-vector product:\ryan{I would choose $\mathbb{E}$ such that $\mathbb{E}^{\top}$ is what is pre-multiplied.}
\begin{equation}\label{eq:prob-similarity-1}
    \embedMtx \, \enc\left(\eosstr\right).
\end{equation}
$\embedMtx \, \enc\left(\eosstr\right) \in \R^{|\eosalphabet|}$, therefore, has the nice property that each of the individual entries corresponds to the similarities of a particular symbol to the context $\eosstr$.
For reasons that will become clear soon, the entries of the vector $\embedMtx \, \enc\left(\eosstr\right)$ are often called \defn{scores} or \defn{logits}.
This brings us almost to the final formulation of the probability distribution $\pLNSM\left(\eossym\mid\eosstr\right)$.

If $\embedMtx \, \enc\left(\eosstr\right)$ encodes similarity or compatibility, then a natural way to model the probability distribution $\pLNSM\left(\eossym\mid\eosstr\right)$ would be as proportional to the inner product between $\embedding{\eossym}$ and $\enc\left(\eosstr\right)$.
However, the inner product $ \innerProd{\embedding{\eossym}}{\enc\left(\eosstr\right)}$ may be negative; further, the sum over the similarity between a context and all tokens is not necessarily $1$.
To resolve this, we have to introduce the last piece of the puzzle: transforming $\embedMtx \, \enc\left(\eosstr\right)$ into a valid discrete probability distribution by using a projection function.


\subsection{Projecting onto the Simplex} \label{sec:representation-lms}

In the previous subsections we discussed how to encode symbols and contexts in a Hilbert space and
how an inner product gives us a natural notation of similarity between a potentially infinite number of items. We can now finally discuss how to create the conditional distribution $\pLNSM\left(\eossym\mid\eosstr\right)$, i.e., how we can  \emph{map}  the real-valued $\embedMtx \, \enc\left(\eosstr\right)$ that encodes symbol--context similarities to a valid probability distribution---a vector on the \defn{probability simplex}.
\subsubsection{Projection Functions: Mapping Vectors onto the Probability Simplex}
$\pLNSM\left(\eossym\mid\eosstr\right)$ is  a \defn{categorical distribution}\index{categorical distribution} with $|\eosalphabet|$ categories, i.e., a vector of probabilities whose components correspond to the probabilities of individual categories.
Perhaps the simplest way to represent a categorical distribution is as a vector on a probability simplex.
\begin{definition}{Probability Simplex}{prob_simplex}
    A \defn{probability simplex} $\Simplexdminus$ is the set of non-negative vectors $\R^{\hiddDim}$ whose components sum to $1$:
    \begin{equation}\label{eq:prob_simplex}
        \Simplexdminus \defeq \left\{  \vx \in \R^{\hiddDim} \mid \evx_{\idxd} \geq 0, \idxd = 1, \ldots, \hiddDim \text{ and }  \sum_{\idxd=1}^{\hiddDim} \evx_\idxd = 1 \right\}
    \end{equation}
\end{definition}
So far, we have framed $\pLNSM$ as a function assigning the conditional distribution over $\eossym$ to each string $\eosstr$. The definition of a simplex means that we can more formally express $\pLNSM$ as a \defn{projection}\index{projection}
from the Hilbert space of the context representations to $ \SimplexEosalphabetminus$, i.e., $\pLNSM\colon \vectorspace \to \SimplexEosalphabetminus$.
Yet all we have discussed so far is creating a vector  $\embedMtx \, \enc\left(\eosstr\right)$ that encodes symbol--context similarities---$\embedMtx \, \enc\left(\eosstr\right)$ is not necessarily on the probability simplex $\SimplexEosalphabetminus$. To address this issue, we turn to projection functions:

\begin{definition}{Projection Function}{projec-func}
    A \defn{projection function} $\projfunc$ is a mapping from a real-valued Hilbert space $\R^{\hiddDim}$ to the probability simplex $\Simplexdminus$
    \begin{equation}
        \projfunc\colon \R^{\hiddDim} \to\Simplexdminus.
    \end{equation}
\end{definition}
\noindent which allows us to define a probability distribution according to $\embedMtx \, \enc\left(\eosstr\right)$:
\begin{equation}\label{eq:proj-vy}
    \pLNSM\left(\eossym\mid\eosstr\right) =
    \projfuncEosalphabetminus\left(\embedMtx \, \enc\left(\eosstr\right)\right)_{\eossym}
\end{equation}

Clearly, we still want the projection of $\embedMtx \, \enc\left(\eosstr\right)$ onto $\SimplexEosalphabetminus$ to maintain several attributes of the original vector---otherwise, we will lose the notion of compatibility that $\embedMtx \, \enc\left(\eosstr\right)$ inherently encodes.
However, $\projfuncEosalphabetminus$ must satisfy several additional criteria in order to map onto a valid point in $\SimplexEosalphabetminus$.
For example, the inner product of two vectors (and consequently $\embedMtx \, \enc\left(\eosstr\right)$) is not necessarily positive---yet all points in $\SimplexEosalphabetminus$ are positive (see \cref{def:prob_simplex}).
These characteristics motivate the use of a projection function that is both monotonic and positive everywhere.
Thus, one clear choice is to base our chosen projection function on the exponential function, i.e., \clara{perhaps use f' here? not sure...}
\begin{equation}\label{eq:propto-exp}
    \projfuncEosalphabetminus(\embedMtx \, \enc\left(\eosstr\right)) \propto \exp\left ( \embedMtx \, \enc\left(\eosstr\right)\right).
\end{equation}
To make a function of the form in \cref{eq:propto-exp} a valid projection function, we now simply have to  ensure that the output of $\vfunc_{\Simplexdminus}$ sums to $1$, which can easily be accomplished by \emph{re-normalizing} the vector of exponentiated values by their sum.
This brings us to the main star of this subsection: the $\softmax$.

While we simply motivated its introduction by chasing our goal of ending up on the probability simplex, the origin of the $\softmax$ function goes back to the Boltzmann distribution from statistical mechanics introduced in the mid-1800s by \citet{boltzmann1868studien}.
It was then studied intensely and popularized by \citet{Gibbs1902}.
It was originally introduced as a way to convert the energy function of the Boltzmann distribution
into a probability distribution.\footnote{This is precisely the connection we mentioned in \cref{def:energy-function}.} Yet now, for reasons we will see in this subsection, the $\softmax$ is the predominant choice of projection function in machine learning applications.

Formally, the $\softmax$ is often defined in terms of a temperature parameter $\tempParam$ as follows.
\begin{definition}{Softmax}{softmax}
    Let $\tempParam \in \Rplus$ be the \defn{temperature}.
    The \defn{softmax}\index{softmax} at temperature $\tempParam$  is the projection function defined as:
    \begin{align}\label{eq:softmax}
        \softmaxfunc{\vx}{\idxd} & \defeq \frac{\exp\!\left[ \frac{1}{\tempParam} \evx_\idxd \right]}{\sum_{\idxj=1}^{\hiddDim} \exp\!\left[ \frac{1}{\tempParam} \evx_\idxj \right]}, \textnormal{ for } \idxd = 1,\ldots,\hiddDim
    \end{align}
\end{definition}
where the temperature parameter $\tempParam$ gives us a mechanism for controlling the entropy of the $\softmax$ function by \emph{scaling} the individual scores in the input vector before their exponentiation.
In the context of the Boltzmann distribution, it was used to control the ``randomness'' of the system:
When the temperature is high, the $\softmax$ function outputs a more uniform probability distribution whose probabilities are relatively evenly spread out among the different categories.
When the temperature is low, the $\softmax$ function outputs a peaked probability distribution, where the probability mass is concentrated on the most likely category.
In the limit, as we take $\tempParam$ to the edge of the possible values it can assume, the following properties hold:
\begin{theorem}{Limiting behavior of the softmax function}{softmax-limits}
    \begin{align}
        \lim_{\tempParam \to \infty} \softmax\left(\vx\right) & = \frac{1}{\hiddDim} \one        \\
        \lim_{\tempParam \to 0^+} \softmax\left(\vx\right)    & = \ve_{\argmax\left(\vx\right)},
    \end{align}
    where $\ve_\idxd$ denotes the $\idxd^\text{th}$ basis vector in $\R^\hiddDim$, $\one \in \R^\hiddDim$ the vector of all ones, and
    \begin{equation}
        \argmax\left(\vx\right) \defeq \min{\left\{\idxd \mid \evx_\idxd = \max_{\idxd = 1, \ldots, \hiddDim}\left(\evx_\idxd\right)\right\}},
    \end{equation}
    i.e., the index of the maximum element of the vector $\vx$ (with the ties broken by choosing the lowest such index).
    In words, this means that the output of the $\softmax$ approaches the uniform distribution as $\tempParam \to \infty$ and towards a single mode as $\tempParam \to 0^+$.\footnote{$\tempParam \to 0^+$ denotes the limit from above.}
\end{theorem}
\begin{proof}
    Let us first consider the case of $\tempParam \to 0^+$. Without loss of generality, let us consider a 2-dimensional vector $\vx = [x_1, x_2]^\top$
    \begin{align}
        \lim_{\tempParam \to 0^+}{\softmaxfunc{\vx}{1}}
         & = \lim_{\tempParam \to 0^+}{\frac{\exp{(\frac{x_1}{\tempParam})}}{\exp{(\frac{x_1}{\tempParam})}+ \exp{(\frac{x_2}{\tempParam})}}}                                                                            \\
         & = \lim_{\tempParam \to 0^+}{\frac{\exp{(\frac{x_1}{\tempParam})}\exp{(-\frac{x_1}{\tempParam})}}{\left(\exp{(\frac{x_1}{\tempParam})}+ \exp{(\frac{x_2}{\tempParam})}\right)\exp{(-\frac{x_1}{\tempParam})}}} \\
         & = \lim_{\tempParam \to 0^+}{\frac{1}{1 + \exp{(\frac{x_2-x_1}{\tempParam})}}}\label{eq:softmax_lim}
    \end{align}

    which leads us to the following definition for element-wise values:
    \begin{equation}
        \lim_{\tempParam \to 0^+}{\exp{\left(\frac{x_2-x_1}{\tempParam}\right)}} = \left\{ \begin{array}{ll}
            0,      & \mathrm{if}\,x_1 > x_2 \\
            1,      & \mathrm{if}\,x_1 = x_2 \\
            \infty, & \mathrm{o.w.}
        \end{array} \right.
    \end{equation}

    Then the limit of softmax as $\tempParam \to 0^+$ is given as

    \begin{equation}
        \lim_{\tempParam \to 0^+}{\softmax(\vx)} = \left\{ \begin{array}{ll}
            \left[1,0\right]^\top,                     & \mathrm{if}\,x_1 > x_2 \\
            \left[\frac{1}{2},\frac{1}{2}\right]^\top, & \mathrm{if}\,x_1 = x_2 \\
            \left[0,1 \right]^\top,                    & \mathrm{o.w.}
        \end{array} \right.
    \end{equation}
    which is equivalent to the $\argmax$ operator over $\vx$. The proof extends to arbitrary $\hiddDim$-dimensional vectors.

    The case of $\tempParam \to \infty$ follows similar logic, albeit $\lim_{\tempParam \to \infty}{\exp{\left(\frac{x_2-x_1}{\tempParam}\right)}} = 1$ in all cases. Hence, we get $\lim_{\tempParam \to \infty}{\softmax(\vx)} = \frac{1}{\hiddDim} \one$.
\end{proof}
The second property, specifically, shows that the $\softmax$ function resembles the $\argmax$ function as the temperature approaches $0$---in that sense, a more sensible name for the function would have been ``soft\emph{arg}max''.
We will most often simply take $\tempParam$ to be $1$.
However, different values of the parameter are especially useful when \emph{sampling} or generating text from the model, as we discuss subsequently.

The output of the $\softmax$ is equivalent to the solution to a particular optimization problem, giving it a variational interpretation.
\begin{theorem}{Variational characterization of the softmax}{softmax-opt}
    Given a set of real-valued scores $\vx$, the following equality holds
    \begin{align}\label{eq:softmax-opt}
        \softmaxfunc{\vx}{} & = \argmax_{\rvp \in\Simplexdminus} \left( \rvp^{\top} \vx  -
        \tempParam \sum_{\idxd =1}^\hiddDim \ervp_\idxd\log \ervp_\idxd  \right)           \\
                            & = \argmax_{\rvp \in\Simplexdminus} \left( \rvp^{\top} \vx  +
        \tempParam\ent(\rvp) \right)
    \end{align}
    This tells us that softmax can be given a variational characterization, i.e., it can be viewed as the solution to an optimization problem.
\end{theorem}
\begin{proof}
    \cref{eq:softmax-opt} can equivalently be written as
    \begin{align}
        \softmaxfunc{\vx}{} & = \argmax \left( \rvp^{\top} \vx  -
        \tempParam \sum_{\idxd =1}^\hiddDim \ervp_\idxd\log \ervp_\idxd  \right)       \\
                            & \quad\quad\quad\quad\text{s.t.} \sum_{\idxd} p_\idxd = 1
    \end{align}

    \noindent from which we can clearly see that the  Lagrangian of this optimization problem is  $\Lambda = \rvp^{\top} \vx  -
        \tempParam\sum_{\idxd =1}^\hiddDim \ervp_\idxd\log \ervp_\idxd + \lambda \sum_{\idxd} p_\idxd$. Taking the derivative of $\Lambda$ with respect to $p_\idxd$, we see that the optimum of is reached when
    \begin{equation}
        \frac{\partial \Lambda}{\partial p_\idxd} = v_\idxd - \tempParam ( \log p_\idxd + 1) + \lambda = 0
    \end{equation}
    Solving for $p_\idxd$ gives us $p_\idxd = Z\exp(\frac{x_i}{\tempParam})$, where $Z$ is the normalizing constant that ensures $\sum_{\idxd} p_\idxd=1$. This solution is equivalent to performing the $\softmax$ operation over $\vx$, as desired.
\end{proof}

\cref{thm:softmax-opt} reveals an interpretation of the softmax as the projection $\rvp\in \Simplexdminus$ that has the maximal similarity with $\vx$ while being regularized to produce a solution with high entropy.
Further, from both \cref{def:softmax} and \cref{eq:softmax-opt}, we can see that $\softmax$ leads to non-sparse solutions as an entry $\softmaxfunc{\vx}{\idxi}$ can only be $0$ if $\evx_\idxd\!=\!-\infty$.

In summary, the $\softmax$ has a number of desirable properties for use in machine learning settings.
\begin{theorem}{Desirable properties of the softmax function}{softmax-desirable}
    The $\softmax$ function with temperature parameter $\tempParam$ exhibits the following properties.
    \begin{enumerate}
        \item In the limit  as $\tempParam\to 0^+$ and $\tempParam\to\infty$, the $\softmax$ recovers the $\argmax$ operator and projection to the center of the probability simplex (at which lies the uniform distribution), respectively.
        \item $\softmax(\vx + c\one) = \softmax(\vx)$ for $c\in\R$, i.e., the $\softmax$ is invariant to adding the same constant to all coordinates in $\vx$.
        \item The derivative of the $\softmax$ is continuous and differentiable everywhere; the value of its derivative can be explicitly computed.
        \item For all temperatures $\tempParam \in \Rplus$, if $x_\idxi \leq x_\idxj$, then $\softmax(\vx)_\idxi \leq \softmax(\vx)_\idxj$. In words, the $\softmax$ maintains the rank of $\vx$.
    \end{enumerate}
\end{theorem}
\begin{proof}
    Property 1. is simply a restatement of \cref{thm:softmax-limits}. The proof for property 2. can be shown using simple algebraic manipulation:
    \begin{equation}
        \resizebox{0.8\textwidth}{!}{
            $\softmax(\vx + c\one)_\idxd = \frac{\exp\!\left[ \frac{1}{\tempParam} \evx_\idxd + c\right]}{\sum_{\idxj=1}^{\hiddDim} \exp\!\left[ \frac{1}{\tempParam} \evx_\idxj + c\right]} = \frac{\exp\!\left[ \frac{1}{\tempParam} \evx_\idxd \right]\cdot \exp{c}}{\sum_{\idxj=1}^{\hiddDim} \exp\!\left[ \frac{1}{\tempParam} \evx_\idxj \right]\cdot \exp{c}} = \softmax(\vx )_\idxd$
        }
    \end{equation}
    The derivative of the $\softmax$ at position $\idxi$ with respect to the variable at position $\idxj$ is given by
    \begin{equation}\label{eq:softmax-deriv}
        \frac{\partial \softmax(\vx)_\idxi}{x_\idxj} =\frac{ \delta_{i}(j)\cdot\exp(x_\idxi)\sum_{k}\exp(x_k) - \exp(x_\idxi)\cdot\exp(x_\idxj)}{(\sum_{k}\exp(x_k))^2}
    \end{equation}
    where  $\delta_{i}(j)$ is the Dirac Delta function, defined as $\delta_{i}(j) = \begin{cases} 1 \,\,\textbf{if}\,\, i = j\\ 0 \,\,\textbf{else} \end{cases}$. Clearly, \cref{eq:softmax-deriv} is continuous. Further, it takes on values for all $
        \vx\in\Rd$.\clara{perhaps need to say that denominator is non-zero?}
    Lastly, property 4. follows from the monotonicity of the $\exp$ function.

\end{proof}

There are many other valid projection functions that one could choose from. For example, \cite{sparsemax} introduce the \defn{sparsemax}, which \emph{can} output sparse distributions:
\begin{equation}\label{eq:spmax}
    \sparsemaxfunc{\vx}{} \defeq \argmin_{\rvp\in \Simplexdminus} ||\rvp- \vx ||^2_2
\end{equation}\ryan{We need a theorem (from the appendix of the softmax paper with citation) about how to compute the derivation of sparsemax. there be details.\response{clara} If anything, wouldnt adding that be distracting? We never actually use the sparsemax}
In words, sparsemax directly maps $\vx$ onto the probability simplex, which often leads to solutions on the boundary, i.e., where at least one entry of $\rvp$ is $0$. \cite{sparsemax} provide a method for computing the closed form solution of this optimization problem in Alg. 1 of their work.\todo{include alg} \cite{pmlr-v89-blondel19a} later introduced a framework that encompasses many different projection functions, which they term regularized prediction functions. Essentially, this framework considers the subset of  projection functions that can be written as:
\begin{equation}\label{eq:proj-opt}
    \projfuncEosalphabetminusFunc{\vx}{} \defeq \argmax_{\rvp \in\Simplexdminus} \left( \rvp^{\top} \vx  -
    \Omega(\rvp)\right)
\end{equation}
where $\Omega\colon\R^{\hiddDim}\to\R$ is regularization term. For certain choices of $\Omega$, there are straightforward closed-form solutions to \cref{eq:proj-opt}. For example, as we can see from \cref{eq:softmax-opt}, \cref{eq:proj-opt} is equivalent to the $\softmax$ when $ \Omega(\rvp) = -\ent(\rvp)$, meaning we can compute its closed form using \cref{eq:softmax}. Further, we recover the $\sparsemax$ when $ \Omega(\rvp) = -||\rvp||_2^2$, which likewise has a closed-form solution. The notion of regularizing $\rvp$ may be unintuitive at first, but we can view it as trying to balance out the ``suitability'' term $\rvp^{\top} \vx$ with a ``confidence'' term $\Omega(\rvp)$, which should be smaller when $\rvp$ is ``uncertain.'' We point the interested reader to the comprehensive work of \cite{pmlr-v89-blondel19a} for further elaboration.

So why aren't these other projection functions more widely employed in machine learning frameworks? First, not all choices of $\Omega$  lead to closed-form solutions; further, not all meet the desirable criterion listed in \cref{thm:softmax-desirable}. For example, the $\sparsemax$ is not everywhere differentiable, meaning that one could not simply use out-of-the-box automatic differentiation frameworks when training a model using the sparsemax as its projection function. Rather one would have to specify its gradient explicitly.
\begin{theorem}{Deterivative of the sparsemax function}{}
    The derivative of the the $\sparsemax$ with respect to its input $\vx$ is as follows:
    \begin{equation}
        \frac{\partial\sparsemax(\vx)_\idxi}{\partial x_\idxj} = \begin{cases}
            \delta_{\idxi \idxj} - \frac{1}{S(\vx)} \quad & \textbf{if}\,\, \idxi, \idxj \in S(\vx) \\
            0                                             & \textbf{else}
        \end{cases}
    \end{equation}
\end{theorem}
\begin{proof}
    See \citet{sparsemax}.
\end{proof}

To conclude, projection functions, together with symbol representations and the representation function $\enc$, give us the tools to define a probability distribution over next symbols that encodes complex linguistic interactions.
We now bring all the components together into the locally normalized modeling framework in the next section.


\subsection{Representation-based Locally Normalized Models}
With these tools at hand, we now define representation-based locally normalized language models.
\begin{definition}{Representation-Based Locally Normalized Model}{locally-normalized-rep-based}
    Let $\enc$ be an encoding function.
    A \defn{representation-based locally normalized model} is a model of the following form:
    \begin{align} \label{eq:softmax-sequence-model}
        \pLNSM(\eossym_\tstep \mid \eosstr_{<\tstep})
         & \defeq \projfuncEosalphabetminusFunc{\embedMtx \,\enc(\eosstr_{<\tstep})}_{\eossym_\tstep}
    \end{align}
    \noindent where unless otherwise stated, we assume $\projfuncEosalphabetminus = \softmax$.
    It defines the probability of an entire string $\str \in \kleene{\alphabet}$ as
    \begin{equation} \label{eq:full-string-softmax}
        \pLN\left(\str\right) \defeq \pLNSM\left(\eos \mid \str \right) \prod_{\tstep=1}^{\strlen} \pLNSM(\sym_\tstep \mid \str_{<\tstep})
    \end{equation}
    where $\sym_0 \defeq \bos$.
\end{definition}\ryan{We should probably add a caveat about tightness. \response{anej} what do you mean?}
Alternatively, we could also include an additive \defn{bias} term $\bias$ as part of the projection function $\projfuncEosalphabetminus$ in the definition of the conditional distribution $\pLNSM(\eossym_\tstep \mid \eosstr_{<\tstep})$, i.e., $\pLNSM(\eossym_\tstep \mid \eosstr_{<\tstep}) = \projfuncEosalphabetminusFunc{\embedMtx \,\enc(\eosstr_{<\tstep}) + \bias}_{\eossym_\tstep}$.
However, note that the bias term can be absorbed into the encoding function $\enc$, meaning that we can assume the form \cref{eq:softmax-sequence-model} without loss of generality.
In representation-based language models, $\embedding{\eossym}$ and $\enc(\str)$ carry all the necessary information to determine how probable individual symbols $\sym$ are given the context $\str$.
Therefore, the design choices of $\embedding{\eossym}$ and $\enc(\str)$ are crucial when building language models this way.
Indeed, a large portion of the discussion in the remainder of the notes will center around how to build good representations of the context and individual symbols.

\subsection{Tightness of Softmax Representation-based Models} \label{sec:rep-based-tightness}
Having introduced representation-based language models, we can now state a very general result about the tightness of such models.
It connects the notion of tightness to the intuition about the ``compatibility'' of symbols to the context---namely, the compatibility of the $\eos$ symbol to the context (compared to the compatibility of all other symbols).
The compatibility is here captured by the distance of the representation of the $\eos$ symbol to the representation of the other symbols---if this distance grows slowly enough with respect to $\tstep$ (modulo the norm of the context representation), the model is tight.
\begin{theorem}{Proposition 5.9 in \citealp{Du2023}}{leo-rep-tightness}
    Let $\pLNSM$ be a representation-based sequence model over the alphabet $\alphabet$, as defined in \cref{def:locally-normalized-rep-based}.
    Let
    \begin{equation}
        s \defeq \sup_{\sym \in \alphabet} \norm{\embedding{\sym} - \embedding{\eos}}_2,
    \end{equation}
    i.e, the largest distance to the representation of the $\eos$ symbol, and
    \begin{equation}
        z_\text{max} \defeq \max_{\str \in \alphabet^{\tstep}}{\norm{\enc\left(\str\right)}_2},
    \end{equation}
    i.e., the maximum attainable context representation norm for contexts of length $\tstep$.
    Then the locally normalized model $\pLN$ induced by $\pLNSM$ is tight if
    \begin{equation}
        s z_\text{max} \leq \log \tstep.
    \end{equation}
\end{theorem}
\begin{proof}
    Let $\rx_\tstep(\bomega)$ be the random variable that is equal to the $\tstep$\textsuperscript{th} token in an outcome $\bomega \in \samplespace$.
    Then for an arbitrary $\tstep \in \N$ and any $\str \in \alphabet^\tstep$, we have:
    \begin{subequations}
        \begin{align}
            \probfunction(\rx_{\tstep} = \eos \mid \rvx_{<\tstep} = \str)
             & = \frac{\exp\left[\embedding{\eos}^{\top} \encfunc{\str}\right]}{\sum_{\sym \in \eosalphabet} \exp \left[\embedding{\sym}^{\top} \encfunc{\str}\right]}                                              \\
             & = \frac{1}{\frac{\sum_{\sym \in \eosalphabet} \exp \left[\embedding{\sym}^{\top} \encfunc{\str}\right]}{\exp \left[\embedding{\eos}^{\top} \encfunc{\str}\right]}}                                   \\
             & = \frac{1}{1 + \sum_{\sym \in \alphabet} \exp\left[ (\embedding{\sym} - \embedding{\eos})^{\top} \encfunc{\str}\right]}                                                                              \\
             & \geq \frac{1}{1 + \sum_{\sym \in \alphabet} \exp\left[\Vert\embedding{\sym} - \embedding{\eos}\Vert_2 \Vert\encfunc{\str}\Vert_2\right]}                           & \justification{Cauchy--Schwarz} \\
             & \geq \frac{1}{1 + \sum_{\sym \in \alphabet} \exp\left[k\Vert\encfunc{\str}\Vert_2\right]}                                                                                                            \\
             & = \frac{1}{1 + |\alphabet| \exp\left[\,\Vert\encfunc{\str}\Vert_2\right]}
        \end{align}
    \end{subequations}
    Now define $z_\text{max} \defeq \sup_{\str \in \alphabet^\tstep} \Vert\encfunc{\str}\Vert_2$.
    We then have that $\forall \tstep \in \N$ and $\forall \str \in \alphabet^\tstep$:
    \begin{align}
        \probfunction(\rx_{\tstep} = \eos \mid \rvx_{<\tstep} = \str) \ge \frac{1}{1 + |\alphabet| \exp(k z_\text{max})}
    \end{align}

    Now, by \cref{prop:div-implies-tight}, we have that if $\sum_{\tstep = 1}^\infty \frac{1}{1 + |\alphabet| \exp(k\,z_\text{max})}$ diverges, then the language model is tight.
    We will show that if we have that $\exists N \in \N$ such that $\forall \tstep \geq N$, $kz_\text{max} \le \log \tstep$, then the sequence model must be tight.

    First, note that $\lim_{\tstep \to \infty} \frac{1}{\tstep}\frac{1 + |\alphabet| \tstep}{1} = \lim_{\tstep \to \infty} \frac{1}{\tstep} + |\alphabet| = |\alphabet| \in (0, \infty)$.
    Hence, by the limit comparison test, since $\sum_{\tstep=1}^\infty \frac{1}{\tstep}$ diverges, this means $\sum_{\tstep=1}^\infty \frac{1}{1 + |\alphabet| \tstep}$ must also diverge.

    Now, suppose that $k \, z_\text{max} \le \log \tstep$ for all $\tstep \ge N$.
    This implies that for $\tstep \ge N$ we have $\frac{1}{1 + |\alphabet| \exp( k z_\text{max})} \ge \frac{1}{1 + |\alphabet| \tstep}$, which combined with the above and the comparison test, implies that
    $\sum_{\tstep=N}^\infty \frac{1}{1 + |\alphabet| \exp( k z_\text{max})}$ diverges.
    This in turn means that $\sum_{\tstep=1}^\infty \frac{1}{1 + |\alphabet| \exp( k z_\text{max})}$ diverges.
    Hence, if $k \, z_\text{max} \le \log \tstep$ for all $\tstep \ge N$ for some $N \in \N$, then the language model is tight.

\end{proof}

\cref{thm:leo-rep-tightness} is a generalization of the following result from \citet{welleck-etal-2020-consistency}.
\begin{theorem}{Representation-based language models with bounded encodings are tight}{bounded}
    A locally-normalized representation-based language model, as defined in \cref{def:locally-normalized-rep-based}, with uniformly bounded $||\enc(\str)||_p$ (for some $p\geq 1$) is tight.
\end{theorem}
For most of the language models that we consider, $\enc(\str)$ is bound due to the choice of activation functions.
In turn, $\embedMtx \,\encfunc{\eosstr_{<\tstep}}$ is bounded for all $\eosstr$.
Further, by the definition of the softmax, $\projfuncEosalphabetminusFunc{\embedMtx \,\enc(\eosstr_{<\tstep})}_{\eos} > \eta$ for some constant $\eta$.

This concludes our investigation of general representation-based models.
The next section discusses \emph{learning} parametrized models (as a special case, also symbol and context representations).
\newpage{}

\section{Estimating a Language Model from Data}
\label{sec:learning}

\newcommand{\hypspace}{{\color{MacroColor}H}}
\newcommand{\loglike}{{\color{MacroColor}\mathcal{L}}}
\newcommand{\vthetamle}{{\color{MacroColor}\modelparams{\mathrm{MLE}}}}
\newcommand{\corpus}{{\color{MacroColor}\mathcal{D}}}
\newcommand{\corpustrain}{{\color{MacroColor}\mathcal{D}_{\scaleto{\text{train}}{4pt}}}}
\newcommand{\corpustest}{{\color{MacroColor}\mathcal{D}_{\scaleto{\text{test}}{4pt}}}}
\newcommand{\corpusval}{{\color{MacroColor}\mathcal{D}_{\scaleto{\text{val}}{4pt}}}}
\newcommand{\calX}{{\color{MacroColor}\mathcal{X}}}
\newcommand{\objective}{{\color{MacroColor}O}}
\newcommand{\datasize}{{\color{MacroColor}N}}
\newcommand{\dataidx}{{\color{MacroColor}n}}
\newcommand{\numtimesteps}{{\color{MacroColor}S}}
\newcommand{\timeidx}{{\color{MacroColor}s}}
\newcommand{\desiderata}{{\color{MacroColor}C}}
\newcommand{\learningrate}{{\color{MacroColor}\eta}}
\newcommand{\learningratesched}{{\color{MacroColor}\boldsymbol\eta}}
\newcommand{\updatedir}{{\color{MacroColor}U}}
\newcommand{\langdist}{{\color{MacroColor}p_{\scaleto{\modelparamstrue}{4pt}}}}
\newcommand{\emplangdist}{{\color{MacroColor}\widetilde \langdist}}
\newcommand{\arbitrarymetric}{{\color{MacroColor}M}}
\newcommand{\iidsim}{{\color{MacroColor}\overset{\scaleto{\text{i.i.d.}}{3pt}}{\sim}}}

\Anej{I changed all the macro colors to the default color. The ones that were red are still there in the commented definitions in the file.}

The \defn{language modeling task}\index{language modeling task} refers to any attempt to estimate the parameters\footnote{Most of this course focuses on the parametric case, i.e., where $\pM$ is governed by a set of parameters $\modelparams$. However, we will briefly touch upon various non-parametric language models.\looseness=-1} of a model $\pM$ of the ground-truth probability distribution over natural language strings $\pLM$ using data $\corpus = \{ \str^{(\dataidx)}\}_{\dataidx=1}^{\datasize}$, where we assume samples $\str^{(\dataidx)}$ were generated according to  $\pLM$.
This task is often treated as an optimization problem.
Here we will discuss the various components of this optimization problem, primarily the objective and the algorithm used to perform optimization.
Note that the material covered here corresponds to what is colloquially referred to as pre-training.
The learning paradigm for fine-tuning a language model for a downstream task will be covered later in the course.\ryan{We don't have time for this, but I am thinking the Bayesian non-parametric ones and the nearest neighbor ones.\response{clara} do you want that done in this chapter or at some other point? \response{ryan} a new section, I think. Not sure where, though. Maybe after the transformer section?
    \response{clara} ok, keeping this here as a reminder}

\subsection{Data}
In this course, we consider objectives that are defined in terms of data $\corpus$. Therefore, we will first discuss the nature of this data which, more precisely, is a corpus of texts.
Following the notation used throughout the rest of these notes, let $\alphabet$ be an alphabet.
A \defn{corpus}\index{corpus} $\corpus = \{ \str^{(\dataidx)}\}_{\dataidx=1}^{\datasize} \subset \kleene{\alphabet}$ is a collection of $N$ strings. We will use the terms corpus and dataset interchangeably throughout this section. We make the following assumption about the data-generating process of $\corpus$:
\begin{assumption}{Independently and identically distributed assumption}{iid}
    The strings $\str^{(\dataidx)}$ in our corpus $\corpus$ are generated independently and identically distributed (i.i.d.) by some unknown distribution $\pLM$.
\end{assumption}
Note that $\str^{(\dataidx)}$ are strings of an arbitrary length; they can be single words, sentences, paragraphs, or even entire documents depending on how we choose $\alphabet$.
For example, often our models' architectural designs make them unable to process document-length strings efficiently, e.g., they might not fit into a context window that can be reasonably processed by a transformer language model; we will elaborate on this statement in our discussion of transformers in \cref{sec:transformers}.
Thus in practice, we often chunk documents into paragraphs that we treat as separate data points.\footnote{This practice technically breaks \cref{as:iid}, yet the negative (empirically-observed) effects of this violation are minimal and perhaps outweighed by the additional data it allows us to make use of.} This means that our model may not be able to learn properties of language such as discourse structure.

\ryan{I would avoid this term. I know it's in vogue, but I don't know what it means and can't really explain it. Language modeling is a generative modeling task like any other.\response{clara} Removed the mention of 'self-supervised' but I think its important to mention the implication that these models are using unlabeled data to learn, which has alleviated the data bottleneck present in most ML tasks}

\subsection{Language Modeling Objectives}

Similarly to many other machine learning tasks, we can cast our problem as the \emph{search} for the best model $\pM$ of the ground-truth distribution over strings $\pLM$.
In order to make this search tractable, we must limit the models $\pM$ that we consider.
Explicitly, we make the following assumption:
\begin{assumption}{Parametrized model}{hypspace}
    $\pLM$ is a member of the parameterized family of models $\{\model \mid \modelparams\in\paramspace\}$, the set of all distributions representable by parameters $\modelparams$ in a given parameter space $\paramspace$.
\end{assumption}
\noindent As concrete examples, $\modelparams$ could be the conditional probabilities in a simple, standard \ngram{} model for a given prefix of size $n-1$, i.e., $\modelparams$ is $n-1$ simplices of size $|\alphabet|$.\footnote{One might be tempted to assume we only need $|\alphabet|-1$ parameters per simplex, but we condition over $\eosalphabet$ classes per prefix position.}
As another example, $\modelparams$ could be the weights of a neural network; the set $\paramspace$ would then cover all possible valid weight matrices that could parameterize our model.

\cref{as:hypspace} implies that we can equivalently write $\pLM$ as $\langdist$ for certain (unknown) parameters $\modelparamstrue\in\paramspace$.\footnote{We discuss the implications of the case that $\pLM\notin\{\model \mid \modelparams\in\paramspace\}$ later in this section. }
Further, an arbitrary model $\pM$ from this hypothesis space with parameters $\modelparams$ can be written as $\model$; we will use this notation for the remainder of the chapter  to make the parameterization of our distribution explicit.
We now turn to the general framework for choosing the best parameters $\modelparams\in\paramspace$ so that our model $\model$ serves as a good approximation of $\langdist$.\footnote{The modeling paradigms that we will discuss in this section  are predominantly generative, i.e., these models try to learn the underlying distribution of the data rather than the boundaries between different classes or categories. The implication is that parameter estimation in language modeling typically makes use of \emph{unannotated} text data, and is therefore sometimes referred to as \defn{self-supervised}.}\looseness=-1

\subsubsection{General Framework}
We search for model parameters $\modelparamsopt \in \paramspace$ such that the model induced by those parameters maximizes a chosen objective, or alternatively, minimizes some loss function $\loss: \paramspace \times \paramspace \rightarrow \mathbb{R}_{\geq 0}$.
This loss can be used to measure the quality of this model as an approximation to $\langdist$.
In simple math, we search for the solution.
\begin{equation}
    \modelparamsopt \defeq \argmin_{\modelparams \in \paramspace} \loss(\modelparamstrue, \modelparams)
    \label{eq:opt}
\end{equation}

\noindent where our loss function is chosen with the following principle in mind
\clara{cant think of a better name...}
\begin{principle}{Proximity Principle}{}
    We seek a model $\model$ that is ``close'' to $\langdist$.
\end{principle}
\noindent That is, we choose our loss function to be a measure $\arbitrarymetric$ of the difference between a distribution parameterized by $\modelparams$ and one parameterized by the true $\modelparamstrue$, i.e., those of our ground-truth distribution. Yet we are immediately faced with a problem: computing an arbitrary $\arbitrarymetric$ between $\modelparams$ and  $\modelparamstrue$ (or at least the distributions induced by these sets of parameters) requires knowledge of both, the latter for which  we only have samples $\str^{(\dataidx)}\in\corpus$.
We will therefore use our corpus $\corpus$ as an approximation to $\langdist$, which is typically implemented by representing  $\corpus$ as an empirical distribution---a collection of Dirac Delta functions---which we will denote as $\emplangdist$.
Formally, we define
\begin{equation}
    \emplangdist(\str) \defeq \frac{1}{\datasize}\sum_{\dataidx=1}^\datasize\delta_{\str^{(n)}}(\str)
\end{equation}
\noindent where the Dirac Delta function $\delta_{x'}(x) = \begin{cases} 1 \,\,\textbf{if}\,\, x = x'\\ 0 \,\,\textbf{else} \end{cases}$ is essentially a point mass with all probability on $x'$.
We can decompose this definition over symbols in our strings as well. I.e., we can compute
\begin{equation}\label{eq:one-hot-token}
    \emplangdist(\sym_\tstep \mid\str_{<\tstep}) = \frac{1}{\datasize_{\str_{<\tstep}}}\sum_{\dataidx=1}^\datasize\delta _{\sym_\tstep^{(\dataidx)}\mid \str_{<\tstep}^{(\dataidx)}}(\sym_\tstep \mid\str_{<\tstep})
\end{equation}
\noindent where $\datasize_{\str_{<\tstep}}\defeq \sum_{\dataidx=1}^\datasize \mathbbm{1}\{\str_{<\tstep}^{(\dataidx)} = \str_{<\tstep}\}$. Note that we can likewise define \cref{eq:one-hot-token} in terms of the one-hot encodings of symbols, i.e., using the definition in \cref{ex:one-hot-encoding}: $\emplangdist(\cdot \mid\str_{<\tstep}) = \frac{1}{\datasize_{\str_{<\tstep}}}\sum_{\dataidx=1}^\datasize \onehot{\sym_\tstep^{(\dataidx)}} \mathbbm{1}\{\str_{<\tstep}^{(\dataidx)} = \str_{<\tstep}\}$. In fact, the empirical distribution is often also referred to in machine learning as the one-hot encoding of a dataset.

Now that we are equipped with methods for representing both $\langdist$ and $\model$, we can define a loss function for approximating $\langdist$ using $\model$.

\paragraph{Cross-Entropy.} A natural choice for a loss function is cross-entropy, a measure of the difference between two probability distributions, which has its roots in information theory \citep{shannon}. Specifically, in \cref{eq:opt}, we take $\loss(\modelparamstrue, \modelparams) = \ent(\emplangdist, \model)$ where the definition of the cross-entropy $\ent$ between distributions $p_1$ (with support $\gY$) and $p_2$ is as follows:
\begin{equation}\label{eq:cross-ent}
    \ent(p_1, p_2) = -\sum_{\str\in\gY}p_1(\str)\log p_2(\str)
\end{equation}
Further, most of the models that we will encounter in this course are locally normalized.
Thus, it is more common to see cross-entropy  expressed as
\begin{equation}\label{eq:cross_decomp}
    \ent(p_1, p_2) = -\sum_{\str\in\gY}\sum_{t=1}^T p_1(\sym_\tstep^{(\dataidx)})\log p_2(\sym_\tstep\mid \str_{<\tstep}).
\end{equation}

Note that cross-entropy is not symmetric, i.e., $\ent(p_1, p_2)\neq \ent(p_2, p_1)$. To motivate cross-entropy as a loss function, as well as the intuitive difference between the two argument orderings, we turn to coding theory, a sub-field of information theory. In words, the cross-entropy between two probability distributions is the expected number of bits needed to encode an event $\str\in\gY$ from $p_1$ when using the optimal encoding scheme corresponding to distribution $p_2$. Importantly, the optimal encoding scheme for $p_1$ uses $\log p_1(\str)$ bits to encode an event $\str$ that occurs with probability $p_1(\str)$\clara{should I make this a proposition? Also, what would be the correct citation if I dont want to write out the proof...}, implying that the minimal cross-entropy is achieved when $p_1 = p_2$.  This characteristic of cross-entropy motivates another metric: the $\mathrm{KL}$ divergence $\KL$.

\paragraph{$\mathrm{KL}$ Divergence.}
A \defn{divergence measure}\index{divergence measure} is a measure of statistical distance\footnote{Divergences are not technically distances because they are not symmetric, i.e., it may be the case for divergence measure $D$ and probability distributions $p$ and $q$ that $D(p \mid\mid q) \neq D(p \mid\mid q)$. However, they do meet the criteria that $D(p \mid\mid q) \geq 0 \ \forall p, q$ and $D(p \mid\mid q) = 0 \iff p = q$.} between two probability distributions. The $\mathrm{KL}$ divergence is defined as:
\begin{equation}\label{eq:kl}
    \KL(p_1\mid\mid p_2) = \sum_{\str\in\gY}p_1(\str)\log p_2(\str)  - p_1(\str)\log p_1(\str)
\end{equation}
The  $\mathrm{KL}$ divergence can intuitively be viewed as the cross-entropy shifted by the expected number of bits used by the optimal encoding scheme for $p_1$, i.e., it is the \emph{additional} number of expected bits needed to encode events from $p_1$ when using our encoding scheme from $p_2$.  Indeed, taking $\loss(\modelparamstrue, \modelparams) = \KL(\emplangdist \mid\mid \model)$ should lead to the same solution as taking $\loss(\modelparamstrue, \modelparams) = \ent(\emplangdist, \model)$ because the $\emplangdist(\str)\log \emplangdist(\str)$ term is constant with respect to model parameters $\modelparams$.

\subsubsection{Relationship to Maximum Likelihood Estimation}\label{sec:standard-objective}
An alternative way that we could frame our search for model parameters $\modelparamsopt \in \paramspace$ is in terms of data likelihood. Formally, the \defn{likelihood}\index{likelihood} of the corpus $\corpus$ under the distribution $\model$ is the joint probability of all $\str^{(\dataidx)}$:
\begin{equation}\label{eq:mle}
    L(\modelparams) = \prod_{\dataidx=1}^{\datasize} \model(\str^{(\dataidx)}).
\end{equation}
The principle of maximum likelihood then dictates:
\begin{principle}{Maximum Likelihood}{}
    The optimal parameters for a model are those that maximize the likelihood of observing the given data under that model. Formally:
    \begin{equation}\label{eq:optmle}
        \modelparamsopt_{\mathrm{MLE}} \defeq \argmax_{\modelparams \in \paramspace} \gL(\modelparams)
    \end{equation}
\end{principle}

Note that in practice, we typically work with the \defn{log-likelihood}\index{likelihood!log} $\gL(\modelparams) = \log L(\modelparams)$ rather than the likelihood for a number of reasons, e.g., it is convex and more numerically stable given the small probabilities we encounter when using $L$ and the finite precision of the computing frameworks that we employ.
Since $\log$ is a monotonically increasing function, this would not change the solution to \cref{eq:optmle}. Further, as is the case with \cref{eq:cross_decomp}, we decompose our loss over symbol-level distributions.

Notably,  in our setting, finding parameters that maximize data log-likelihood is equivalent to finding those that minimize cross-entropy.
We show this equivalence below.
\begin{proposition}{}{}
    The optimal parameters under \cref{eq:optmle} are equivalent to the optimal parameters when solving for \cref{eq:opt} with the cross-entropy loss between the empirical distribution $\emplangdist$ and the model $\model$.
\end{proposition}
\begin{proof}
    Under the standard practice of taking $0\log(0) = 0$, the only elements of $\gY$ that make a nonzero contribution to $\ent(\emplangdist, \model)$ are sequences in the support of $\emplangdist$, making summing over $\gY$  equivalent to summing over $\corpus$:
    \begin{align}
        \ent(\emplangdist, \model) & = -\sum_{\str\in\alphabet^*}\emplangdist(\str)\log \model(\str)                                                              \\
                                   & = -\sum_{\str\in\alphabet^*}\frac{1}{\datasize}\sum_{\dataidx=1}^\datasize\delta _{\str^{(\dataidx)}}(\str)\log \model(\str) \\
                                   & = -\sum_{\str\in\alphabet^*}\frac{1}{\datasize}\mathbbm{1}\{\str^{(\dataidx)}\in \corpus \}\log \model(\str)                 \\
                                   & \propto -\sum_{\str\in\corpus}\log \model(\str)                                                                              \\
                                   & = -\gL(\modelparams)
    \end{align}
    Thus, we can see that the objectives are equivalent, up to a multiplicative constant that is independent of model parameters.
\end{proof}
The equivalence of cross-entropy, $\KL$ divergence , and maximum likelihood as learning objectives provides intuition about our many goals when learning $\model$: (1) we want a close (w.r.t. a given metric) approximation of the data-generating distribution, and (2) this approximation should place high probability on samples of real language data.

\paragraph{Properties of $\modelparamsopt$ under the cross-entropy loss.}
\cref{as:hypspace} may feel quite strong, as it implies we know a great deal about the nature of $\pLM$. However, it allows us to prove the optimality of $\modelopt$ under certain conditions.
\begin{theorem}{Maximum likelihood estimate is consistent}{}
    Consider that our loss function $\loss(\modelparamstrue, \modelparams) = \ent(\emplangdist, \model)$ (or equivalently that $\loss(\modelparamstrue, \modelparams) = \KL(\emplangdist\mid\mid\model)$). Given \cref{as:iid} and that the minimizer $\modelopt$ of $ \ent(\emplangdist, \model)$ is unique, then under certain (quite strong) regularity conditions on $\{\model \mid \modelparams\in\paramspace\}$,  $\modelparamsopt$ is a consistent estimator, i.e., it converges to  $\modelparamstrue$ in probability as $n\rightarrow\infty$.
\end{theorem}

Arguably, in practice, \cref{as:hypspace} does not hold; we often make some incorrect modeling assumptions. Naturally, this raises the following question: If we \emph{mis}specify the family of models that $\pLM$ belongs to, i.e., $\pLM\notin\{\model \mid \modelparams\in\paramspace\}$, then is our optimal model $\modelopt$ under the cross-entropy loss at all meaningful? Fortunately, the answer here is yes.   In this case, we can interpret $\modelopt$ as a projection of $\pLM$ onto the manifold of parametric
models $\{\model \mid \modelparams\in\paramspace\}$. This projection is formally known as an \defn{information projection} \citep{information_projection}, which while we do not cover formally here, we can intuit as a mapping of $\pLM$ onto its ``closest'' point in $\{\model \mid \modelparams\in\paramspace\}$.  In this setting, using different metrics $\arbitrarymetric$ leads to different definitions of closeness, which in turn means that optimal models under different $\arbitrarymetric$ exhibit different properties.

\paragraph{Potential drawbacks of cross-entropy loss.}
A closer inspection of \cref{eq:cross-ent} reveals that, when we use $\ent(\emplangdist, \model)$ as our loss function, $\model$ must put probability mass on \emph{all} samples $\str^{(\dataidx)}$ in the support of $\emplangdist$; otherwise, our loss is infinite. Since the model is not explicitly penalized for extraneous coverage, it will thus resort to placing mass over all of $\kleene{\alphabet}$ to avoid such gaps;\footnote{This behavior can also be (at least partially) attributed to the softmax used to transform model outputs into a probability distribution over symbols. Since the softmax maps to the interior of the probability simplex, no symbol can be assigned a probability of exactly $0$. } this is sometimes referred to as \emph{mean-seeking} behavior. In practice, this means that sequences of symbols that one might qualitatively describe as gibberish are assigned nonzero probability by $\model$. It is unclear whether this is a desirable property under a language model. While perhaps useful when using such a model to assign probabilities to strings---in which case, we might be more interested in how strings' probabilities rank against each other and may not want to write off any string as completely improbable---it could prove problematic when generating strings from these models, a topic covered later in this course.

\paragraph{Teacher Forcing.}
The loss functions that we have considered thus far are all based on our model's predictions conditioned on  prior context.  Here we are faced with a choice during training: we could either use the model's predictions from the previous time step(s) $\model(\cdot \mid \widehat\str_{<\tstep})$ (e.g., the most probable symbols) as the prior context or use the ground-truth prior context from our data  $\model(\cdot \mid\str_{<\tstep})$. The latter method is often referred to as \defn{teacher forcing}\index{teacher forcing}: Even if our model makes an incorrect prediction at one step of training, we intervene and provide the correct answer for it to make subsequent predictions with.\looseness=-1

From a theoretical perspective, training with the cross-entropy loss mandates that we should use the teacher-forcing approach since each conditional distribution is defined with respect to the ground-truth context; this is elucidated, for example, in \cref{eq:cross_decomp}.
Yet such meticulous guidance can lead to poor performance in tasks where the model is required to accurately predict an entire sequence of symbols on its own.
For example, in language generation, since the model is not exposed to its own generations during training, small errors in predictions can compound, leading to degenerate text. This problem is known as \defn{exposure bias}\index{exposure bias}.
Only the other hand, using previous model outputs  in order to make subsequent predictions can lead to serious instability during training, especially if implemented from the start of training. Methods for alleviating exposure bias have been proposed with more stable training dynamics, such as scheduled sampling \cite{bengio_scheduled_sampling}, which we discuss in \cref{sec:alt-objectives}.


\subsubsection{Alternative Objectives}\label{sec:alt-objectives}
\paragraph{Masked Language Modeling.}
So far, our parameter estimation strategies have made use of the decomposition of $\model(\str)$ into individual symbol probabilities, conditioned on \emph{prior} symbols, i.e.,  $\model(\str) = \prod_{t=1}^T \model(\sym_\tstep\mid \str_{<\tstep})$. In other words, we do not give a model both sides of a symbol's context when asking it to estimate the probability distribution over that symbol. While this paradigm might be more realistic when using a language model for tasks such as generation---for which we may want to generate outputs sequentially to mimic human language production---access to both sides of a symbol's context could be critical when using the model for tasks such as acceptability judgments or classification. This motivates the use of an alternative objective for parameter estimation.

Similarly to the maximum likelihood objective in \cref{eq:mle}, we can choose model parameters by optimizing for the per-symbol log-likelihood of a dataset $\corpus$, albeit in this case, using \emph{both} sides of the symbol's context:
\begin{equation}\label{eq:mlm}
    \gL_{\texttt{MLM}}(\modelparams) = \sum_{\dataidx=1}^{\datasize}\sum_{\tstep=1}^\strlen \log \model(\sym_\tstep^{(\dataidx)}\mid \str_{<\tstep}^{(\dataidx)}, \str_{>\tstep}^{(\dataidx)})
\end{equation}
\cref{eq:mlm} is sometimes referred to as the \defn{pseudo(log)likelihood}\index{likelihood!pseudo} \citep{besag_pseudolikelihood}, since it gives us an approximation of the true log-likelihood, i.e., $\sum_{\tstep=1}^\strlen \log \model(\sym_\tstep\mid \str_{<\tstep}, \str_{>\tstep})\approx \log\model(\str)$. Pseudolikelihood  has its origins in thermodynamics, where it was used as an approximate inference technique for parameter estimation in Ising models. In such situations, computing $ \model(\sym_\tstep\mid \str_{\neq\tstep})$ often proved computationally easier than computing the exact set of of conditional probabilities whose product equaled the marginal.

Using \cref{eq:mlm} as a model's training objective is also motivated by psychological tests of language understanding---specifically, the Cloze \citep{cloze} task in psychology, in which the goal is to predict the omitted symbol from a piece of text that constitutes a logical and coherent completion. For example, in the string
\begin{example}{The Close task}{}
    The students \texttt{[MASK]} to learn about language models.
\end{example}
\noindent we  predict \textit{want} or \textit{like} with high probability for the \texttt{[MASK]} position.
When used as an objective in NLP, estimating the probability distribution over symbols at the masked position is referred to as masked language modeling; BERT \citep{devlin-etal-2019-bert} is one well known example of a masked language model.
In practice, typically only the distributions over symbols at a percentage of randomly-chosen positions in $\corpus$ are estimated during training.
As mentioned in \cref{sec:lm-tightness}, a model whose parameters are estimated with the masked language modeling objective is not a valid language model in the sense of \cref{def:language-model} because it does not provide a valid distribution over $\kleene{\alphabet}$.
Yet, masked language models have become increasingly popular as base models for fine-tuning on certain downstream tasks, where they sometimes lead to superior performance over standard language models.

\paragraph{Other Divergence Measures.}
From a given hypothesis space (see \cref{as:hypspace}), the distribution  that  minimizes a given divergence measure with $\langdist$ exhibits certain properties with respect to how probability mass is spread over the support of that distribution. For example, the model $\model$ that minimizes $\KL(\langdist \mid\mid \model)$ exhibits \emph{mean-seeking} behavior, as discussed earlier in this section. These properties  have been studied in depth by a number of works \citep{minka_divergence,theis_note_2016,huszar_how_2015,labeau-cohen-2019-experimenting}.
The implication of these findings is that, depending on the use case for the model, other divergence measures may be better suited as a learning objective. For example, prior work has noted frequency biases in models estimated using the standard log-likelihood objective, i.e., these models exhibit an inability to accurately represent the tails of probability distributions \citep{gong_frequency}. This is particularly relevant in the case of language modeling, as symbol-usage in natural language tends to follow a power-law distribution \citep{zipf}.
Consequently, when we care particularly about accurately estimating the probability of rare words, we may wish to instead use a loss function that prioritizes good estimation of probability distribution tails.  On the other hand, in the case of language generation, we may desire models that only assign probability mass to outputs that are highly-likely according to $\langdist$, even if this means assigning probabilities of $0$ to some outcomes possible under $\langdist$. In other words, we may want a model with \emph{mode-seeking} behavior, which is characteristic of models trained to minimize $\KL(\model \mid\mid \langdist)$.
However, there are  a number of computational issues with using other divergence measures---such as general power divergences, reverse $\KL$ divergence, and total variation distance---for training neural probabilistic models over large supports, making them difficult to work with in practice\todo{give citations for this}. For example, we can compute a Monte Carlo estimate of the forward $\KL$\todo{could illustrate this more explicitly} divergence simply by using samples from $\langdist$, which is exactly what we have in our dataset. However an unbiased estimator of the reverse $\KL$ divergence would require the ability to query $\langdist$ for probabilities, which we  do not have.\clara{perhaps talk about the relationship of reverse KL and reinforcement learning?}

\paragraph{Scheduled Sampling and Alternative Target Distributions.}
Scheduled sampling \citep{bengio_scheduled_sampling} is an algorithm proposed with the goal of alleviating exposure bias: after an initial period of training using the standard teacher forcing approach, some percentage of the models' predictions are conditioned on prior model outputs, rather than the ground-truth context. However, under this algorithm, $\modelparamsopt$ does not lead to a consistent estimator of $\modelparamstrue$ \citep{huszar_how_2015}.
Other methods likewise aim to alleviate the discrepancy between settings during parameter estimation and those at inference time by specifying an alternative target distribution, for example, one that ranks ``higher-quality'' text as more probable than average-quality text. \todo{bit of a rough transition here. Smooth over}Ultimately, these methods often make use of techniques developed for reinforcement learning, i.e., the REINFORCE algorithm. These methods fall under the category of fine-tuning criterion, which are discussed later in this course.

\paragraph{Auxiliary Prediction Tasks.} Certain works jointly optimize for an additional objective when performing parameter estimation. For example, the parameters for BERT were learned using both the masked language modeling objective as well as a task referred to as next sentence prediction, i.e., given two sentences, estimating the probability that the second sentence followed the first in a document. A number of similar auxiliary tasks have subsequently been proposed, such as symbol frequency prediction or sentence ordering (see \cite{aroca-ouellette-rudzicz-2020-losses} for summary). However, these tasks do not have a formal relationship to language modeling and it is unclear what their effects are on a model's ability to serve as a valid probability distribution over strings. They likely lead to models that no longer fulfill the formal criteria of \cref{sec:tightness}.


\subsection{Parameter Estimation} \label{sec:param-estimation}
Given a loss function $\loss$ and a parameter space $\paramspace$ from which to choose model parameters, we are now tasked with \emph{finding} the parameters $\modelparamsopt$, i.e., solving \cref{eq:opt}. For the class of models that we consider (those parameterized by large neural networks), finding an exact solution analytically would be impractical, if not impossible. Thus, we must resort to numerical methods, where we find approximately optimal parameters by iterating over solutions. This is known as \defn{parameter estimation}\index{parameter estimation}, or more colloquially as \defn{training}\index{training} our model.

Here we will review the various components of training a language model from start to finish. Many of the techniques used for training language models are generally applicable machine learning techniques, e.g., gradient-descent algorithms. Further, these techniques are constantly evolving and often viewed as trade secrets, meaning  that entities building and deploying models may not reveal the combination of components that they employed. Thus, we give a more general overview of the design choices involved in parameter estimation, along with the characteristics common to most components.


\subsubsection{Data Splitting}
In any machine learning setting, we may overestimate model quality if we evaluate solely on its performance w.r.t. the data on which its parameters were estimated. While we can often construct a model that performs arbitrarily well on a given dataset, our goal is to build a model that generalizes to unseen data.  Thus, it is important to measure the final performance of a model on data  that has had no influence on the choice of model parameters.

This practice can be accomplished simply by splitting the data into several sets. The two basic data splits are a training set $\corpustrain$ and test set $\corpustest$; as the names imply, the training set is used during parameter estimation while the test set is used for evaluating final performance. When samples from $\corpustest$ can be found in $\corpustrain$, we call this \defn{data leakage}\index{data leakage}. The training set can be further divided to produce a validation set $\corpusval$. Typically, $\corpusval$ is not used to define the objective for which parameters are optimized. Rather, it serves as a check during training for the generalization abilities of a model, i.e., to see whether the model has started overfitting to the training data. The validation set can be used, e.g., to determine when to stop updating parameters.

\subsubsection{Numerical Optimization} \label{sec:numerical-optimization}
From a starting point $\modelparams_0 \in \paramspace$ chosen according to our initialization strategy, we want to find $\modelparamsopt$ in an efficient manner.
This is where numerical \defn{optimization algorithms}\index{optimization algorithms} come into play---a precise set of rules for choosing how to move within $\paramspace$ in order to find our next set of parameters.
The output of a numerical optimization algorithm is a sequence of iterates $\{\modelparams_\timeidx\}_{t=0}^T$, with the property that as $T \rightarrow \infty$ we find the minimizer of our objective $\loss$.
Ideally, even after a finite number of iterations, we will be sufficiently close to $\modelparamsopt$.

The basic algorithm for searching the parameter space for $\modelparamsopt$ follows a simple formula: starting from $\modelparams_0 \in \paramspace$, we iteratively compute $\modelparams_1, \modelparams_2, \ldots$ as
\begin{equation}
    \modelparams_{\timeidx + 1} = \modelparams_\timeidx + \text{update magnitude} \times \text{update direction}.
\end{equation}
where the update added to $\modelparams_\timeidx$ to obtain $\modelparams_{\timeidx+1}$ is intended to move us closer to $\modelparamsopt$. Once some maximum number of updates $\numtimesteps$ or a pre-defined desideratum has been met, e.g., our loss has not improved in subsequent iterations, we stop and return the current set of parameters. Many of the numerical optimization techniques in machine learning are gradient-based, i.e., we use the gradient of the objective with respect to current model parameters (denoted as $\nabla_{\modelparams_\timeidx} \loss({\modelparams_\timeidx})$) to determine our update direction. Standard vanilla gradient descent takes the form of \cref{alg:param-opt}, where the learning rate schedule $\learningratesched = \langle\learningrate_0, \cdots, \learningrate_T\rangle$ determines the step size of our parameter update in the loss minimizing direction---there is an inherent trade-off between the rate of convergence and overshooting---and the stopping criterion $\desiderata$ determines whether we can terminate parameter updates before our maximum number of iterations $\numtimesteps$.
\begin{algorithm}[h]\label{alg:param-opt}
    \textbf{Input:} $\loss$ objective\\
    \hspace*{3.2em} $\modelparams_0$ initial parameters \\
    \hspace*{3.2em} $\learningratesched\!$ learning rate schedule \\
    \hspace*{3.2em} $\desiderata \!: \loss \times \paramspace \times \paramspace \rightarrow \{\mathrm{True}, \mathrm{False}\}$ stopping criterion
    \begin{algorithmic}[1]
        \caption{Gradient descent for parameter optimization. }
        \For{$\timeidx = 0, \cdots, \numtimesteps$}
        \State{$\modelparams_{\timeidx + 1} \gets \modelparams_\timeidx - \learningrate_{\timeidx} \cdot \nabla_{\modelparams} \loss( \modelparams_\timeidx)$}
        \If{$\desiderata(\loss, \modelparams_\timeidx, \modelparams_{\timeidx-1})$}
        \State \textbf{break}
        \EndIf
        \EndFor
        \State \Return $\modelparams_{\timeidx}$
    \end{algorithmic}
\end{algorithm}
In vanilla gradient descent, we set $\learningratesched = c\cdot \mathbf{1}$ for some constant $c$ and $\desiderata(\loss, \modelparams_\timeidx, \modelparams_{\timeidx-1}) = \mathbbm{1}\{| \loss(\modelparams_\timeidx) - \loss( \modelparams_{\timeidx-1})| < \epsilon\}$\clara{not sure whether to use indicator variable here or just the expression on its own} for user-chosen $\epsilon$---in words, we stop when the change in loss between parameter updates is below a chosen threshold. In practice, more sophisticated learning rate schedules $\learningrate$, e.g., square-root functions of the timestep \citep{squareroot_lr} or adaptive functions that take into account model parameter values \citep{duchi_adagrad}, and stopping criterion $\desiderata$ are employed.

Modern training frameworks rely on backpropagation---also known as reverse-mode automatic differentiation \citep{griewank2008evaluating}\clara{would you like other citations? Bauer?}---to compute gradients efficiently (and, as the name implies, automatically!).
In fact, gradients can be computed using backpropogation in the same complexity as evaluation of the original function. We do not provide a formal discussion of backpropagation here but see \citet{griewank2008evaluating} for this material.\clara{@Ryan, do you want an explicit mention of the NLP course?}

Recall that our loss function---and consequently the gradient of our loss function---is defined with respect to the \emph{entire} dataset. Vanilla gradient descent therefore requires iterating through all of $\corpustrain$  in order to determine the direction to move parameters $\updatedir$, which is an incredibly time-consuming computation for the large datasets employed in modern machine learning settings. Rather, an optimization algorithm would likely take much less time to converge if it could rapidly compute estimates of the gradient at each step. This is the motivation behind  perhaps the most widely employed class of optimization algorithms in machine learning:  variations of \defn{stochastic gradient descent} (SGD), such as mini-batch gradient descent. Explicitly, these algorithms make use of the fact that $\mathbb{E}_{  \corpus'\iidsim\corpus}\nabla_\modelparams(\modelparams, \corpus') = \nabla_\modelparams(\modelparams, \corpus)$, where in slight abuse of notation, we use $\corpus'\iidsim\corpus$ to signify that the multi-set $\corpus'$ consists of random i.i.d. samples from $\corpus$.\clara{should we make this a principle?} Thus we can instead base our loss $\loss$, and consequently $\updatedir$, off of a randomly selected subset of the data.\footnote{While this logic holds even for samples of size 1 (which is the sample size for standard SGD by definition), basing updates off of single samples can lead to noisy updates. Depending on resource constraints, batch sizes of a few hundred are often used, leading to much more stable training (although in the face of memory constraints, larger batch sizes can be mimicked by accumulating, i.e., averaging, gradients across multiple batches when computing update directions). Batch size itself is often viewed as a model hyperparameter that can have a significant effect on model performance. } In practice though, this sample is taken \emph{without} replacement, which breaks the i.i.d. assumption. This in turn implies this our gradient estimates are biased  under the mini-batch  gradient descent algorithm. However, this bias does not seem to empirically harm the performance of such optimization strategies. Indeed, an entire branch of machine learning called \defn{curriculum learning} focuses on trying to find an optimal data ordering with which to train models to achieve desirable characteristics such as generalization abilities. Even when orderings are randomly selected, the chosen ordering can have a large impact on model performance \cite{Dodge2020FineTuningPL}.

A number of optimization algorithms have since iterated on SGD, e.g., the momentum algorithm \citep{POLYAK19641}. In short,  the momentum algorithm computes an exponentially decaying moving average of past gradients, and continues updating parameters in this direction, which can drastically speed up convergence. A widely-employed optimization algorithm called ADAM \citep{adam} takes a similar approach. Just as in momentum, it computes update directions using a moving average (first moment) of gradients, albeit it additionally makes use of the variance of gradients (second moment) when computing update directions. ADAM is one of the most popular optimization algorithms used for training large language models in modern ML frameworks.\clara{are there LM specific optimization algs? perhaps there are ones for when output dim is large?}

\subsubsection{Parameter Initialization}
Our search for (approximately) optimal model parameters must start from some point in the parameter space, which we denote as $\modelparams_0$. Ideally, starting from any point would lead us to the same solution, or at least to solutions of similar quality. Unfortunately, this is not the case: both training dynamics and the performance of the final model can depend quite heavily on the chosen initialization strategy, and can even have high variance between different runs of the same strategy. This makes sense at some intuitive level though: depending on the learning algorithm, an initial starting point can heavily dictate the amount of searching we will have to do in order to find $\modelparamsopt$, and how many local optima are on the route to $\modelparamsopt$.  Consequently, a poor initial starting point may lead to models that take longer to train and/or may lead our learning algorithm to converge to sub-optimal solutions (i.e., an alternative local optimum)  \citep{Dodge2020FineTuningPL,sellam2022the}. This can be the case even when only estimating the final layer of a network, e.g., when building a classifier by appending a new layer to a pretrained model---a recent, widely-adopted practice in NLP \citep{Dodge2020FineTuningPL}.


Methods for initializing the parameters of neural language models are largely the same as those for initializing other neural networks. Perhaps the simplest approach is to randomly initialize all parameters, e.g., using a uniform or normal random variable generator. The parameters of these generators (mean, standard deviation, bounds) are considered hyperparameters of the learning algorithm.  Subsequent methods have iterated on this strategy to develop methods that take into account optimization dynamics or model architectures. One consideration that is particularly relevant for language models is that the input and output sizes of the embedding layer and the fully connected layer can be very different; this exacerbate the problem of vanishing or exploding gradients during training. For example, \cite{pmlr-v9-glorot10a} proposed Xavier init, which keeps the variance of the input and output of all layers within a similar range in order to prevent vanishing or exploding gradients; \cite{he_initialization} proposed a uniform initialization strategy specifically designed to work with ReLU activation units. Using uniform random variables during parameter initialization can likewise alleviate the problem of vanishing gradients. While most deep learning libraries use thoughtfully-selected initialization strategies for neural networks, it is important to internalize the variance in performance that different strategies can cause.\clara{what more can be said here?}

\subsubsection{Early Stopping}
As previously discussed, performance on $\corpustrain$ is not always the best indicator of model performance. Rather, even if our  objective continues to increase as we optimize over model parameters, performance on held-out data, i.e., $\corpustest$ or even $\corpusval$, may suffer as the model starts to overfit to the training data. This phenomenon inspires a practice called \defn{early stopping}\index{early stopping}, where we stop updating model parameters before reaching  (approximately) optimal model parameter values w.r.t.  $\corpustrain$. Instead, we base our stopping criterion $\desiderata$ off of model performance on $\corpusval$ as a quantification of generalization performance, a metric other than that which model parameters are optimized for, or just a general slow down in model improvement on the training objective.

Early stopping  sacrifices better training performance for better generalization performance; in this sense, it can also be viewed as a regularization technique, a topic which we discuss next. As with many regularization techniques, early stopping can have adverse effects as well. Recent work suggests that many models may have another period of learning after an initial period of plateauing train/validation set performance. Indeed, a sub-field has recently emerged studying the ``grokking'' phenomenon \citep{grokking}, when validation set performance suddenly improves from mediocre to near perfect after a long period in which it appears that model learning has ceased, or even that the model has overfit to the training data. Thus, it is unclear whether early stopping is always a good practice.\clara{more to say here?}

\subsection{Regularization Techniques}
Our goal during learning is to produce a model $\model$ that generalizes beyond the observed data; a model that perfectly fits the training data but produces unrealistic estimates for a new datapoint is of little use.
Exactly fitting the empirical distribution is therefore perhaps not an ideal goal. It can lead to \defn{overfitting}\index{overfitting}, which we informally define as the situation when a model uses spurious relationships between inputs and target variables observed in training data in order to make predictions. While this behavior decreases training loss, it generally harms the model's ability to perform on unseen data, for which such spurious relationships likely do not hold.

To prevent overfitting, we can apply some form of regularization.
\begin{principle}{Regularization}{}
    Regularization is a modification to a learning algorithm that is intended to increase a model's generalization performance, perhaps at the cost of training performance.\footnote{Adapted from \citet{Goodfellow-et-al-2016}, Ch. 7.}
\end{principle}
There are many ways of implementing regularization, such as smoothing a distribution towards a chosen baseline or adding a penalty to the  loss function to reflect a prior belief that we may have about the values model parameters should take on \cite{hastie01statisticallearning,bishop}. Further, many regularization techniques are formulated for specific model architecture: for example, the count-based smoothing methods used  $n$-gram language models \citep{NEY19941,Gale1995GoodTuringFE}.
Here we specifically consider the forms of regularization often used in the estimation of neural language models.
Most fall into two categories: methods that try to ensure a model's robustness to yet unseen (or rarely seen) inputs---e.g., by introducing noise into the optimization process---or methods that add a term to our loss function that reflects biases we would like to impart on our model. This is by no means a comprehensive discussion of regularization techniques, for which we refer the reader to Ch.7 of \citet{Goodfellow-et-al-2016}.

\subsubsection{Weight Decay}

A bias that we may wish to impart on our model is that not all the variables available to the model may be necessary for an accurate prediction. Rather, we hope for our model to learn the simplest mapping from inputs to target variables, as this is likely the function that will be most robust to statistical noise.\footnote{This philosophy can be derived from Occam's Razor, i.e., the principle that one should search for explanations constructed using the smallest possible set of elements.} This bias can be operationalized using regularization techniques such as weight decay \citep{Goodfellow-et-al-2016}---also often referred to as $\ell_2$ regularization. In short, a penalty for the $\ell_2$ norm of $\modelparams$ is added to $\loss$. This should in theory discourage the learning algorithm from assigning high values to model parameters corresponding to variables with  only a noisy relationship with the output, instead assigning them a value close to 0 that reflects the a non-robust relationship.


\subsubsection{Entropy Regularization}
One sign of overfitting in  a language model $\model$ is that it places effectively all of its probability mass on a single symbol.\footnote{The softmax transformation serves as somewhat of a regularizer against this behavior since it does not allow any symbol be assigned a probability of $0$. } Rather, we may want the distributions output by our model to generally have higher entropy, i.e., following the principle of maximum entropy: ``the probability distribution which best represents the current state of knowledge about a system is the one with largest entropy, in the context of precisely stated prior data'' \cite{jaynes}. Several regularization techniques, which we refer to as \defn{entropy regularizers}\index{entropy regularizer}, explicitly penalize the model for low entropy distributions.

Label smoothing \citep{label_smoothing} and the confidence penalty \citep{confidence_penalty} add terms to $\loss$ to penalize the model for outputting peaky distributions. Explicitly, label smoothing reassigns a portion of probability mass in the reference distribution from the ground truth symbol to all other symbols in the vocabulary. It is equivalent to adding a term $\KL(u \mid\mid \model)$ to $\loss$, where $u$ is the uniform distribution. The confidence penalty regularizes against low entropy distribution by adding a term $\ent(\model)$ to $\loss$ that encourage high entropy in model outputs. The general class of entropy regularizers have proven effective in training neural models \cite{meister+al.acl20}.

\subsubsection{Dropout}

Regularization also encompasses methods that expose a model to noise that can occur in the data at inference time. The motivation behind such methods is both to penalize a model for being overly dependent on any given variable (whether directly from the input or somewhere further along in the computational graph) for making predictions. 
Dropout does this explicitly by randomly ``dropping''  variables from a computation in the network \citep{dropout}.

More formally, consider a model defined as a series of computational nodes\clara{@Ryan should we cite notes from NLP class?}, where any given node is the product of the transformation of previous nodes. When dropout is applied to the module that contains that node, then the node is zeroed out with some percentage chance, i.e., it is excluded in all functions that may make use of it to compute the value of future nodes. In this case, the model will be penalized if it relied completely on the value of that node for any given computation in the network. Dropout can be applied to most variables within a model, e.g., the inputs to the model itself, the inputs to the final linear projection in a feed-forward layer, or the summands in the attention head of a Transformer.
Note that at inference time, all nodes are used to compute the model's output.\footnote{Some form of renormalization is typically performed to account for the fact that model parameters are learned with only  partial availability of variables. Thus when all variables are used in model computations, the scale of the output will (in expectation) be larger than during training, potentially leading to poor estimates.}

\subsubsection{Batch and Layer Normalization}
Rescaling variables within a network helps with training stability, and further, with generalization by keeping variables within the same range and with unit variance. Specifically, batch normalization helps regularize the problem of covariate shift, where the distribution of features (both the input features and the variables corresponding to transformed features within a network) differs between the training data and the data at inference time. Batch normalization alleviates this problem by recentering (around 0) and scaling (such that data points have unit variance) data points such that the data flowing between intermediate layers of the network follows approximately the same distribution between batches. Layer normalization likewise performs centering and rescaling, albeit across features rather than across data points. Specifically, normalization is performed so that all of the feature values within a data point have mean 0 and unit variance. \todo{link more explicitly to regularization?}
\newpage{}

\chapter{Classical Language Models} \label{chapter:classical-lms}

Next, we turn to two classical language modeling frameworks: finite-state language models (a natural generalization of the well-known \ngram{} models) in \cref{sec:finite-state} and pushdown language models \cref{sec:pushdown-lms}.
Although the most successful approaches to language modeling are based on neural networks, the study of older approaches to language modeling is invaluable.
First, due to the simplicity of the models, learning how they work helps distill concepts.
And, moreover, they often serve as important baselines in modern NLP and provide very useful insights into the capabilities of modern architectures as we will see when we discuss modern architectures in \cref{ch:nn-lms}.\anej{fix reference}

In the spirit of our question-motivated investigation, we will focus on the following two questions.
\begin{aquestion}{Representing conditional distributions}{}
    How can we tractably represent all conditional distributions of the form $\pLNSM\left(\sym\mid\str\right)$ in a simple way?
\end{aquestion}
\begin{aquestion}{Representing hierarchical structure}{}
    How can we tractably represent the hierarchical structure of human language?
\end{aquestion}

\section{Finite-state Language Models}
\label{sec:finite-state}
After rigorously defining what language models are (and what they are not) and discussing how we can estimate them, it is time to finally introduce our first class of language models---those based on finite-state automata.
Language models derived from probabilistic finite-state automata are some of the simplest classes of language models because they definitionally distinguish a \emph{finite} numbers of contexts when modeling the conditional distribution of the next possible symbol $\pM\left(\sym \mid \str\right)$.
We first give an intuitive definition of a finite-state language model and then introduce a more formal definition, which we will use throughout the rest of the section.
\begin{definition}{Informal definition of a finite-state language model}{}
    A language model $\pLM$ is \defn{finite-state} if it defines only finitely many unique conditional distributions $\pLM\left(\sym\mid\str\right)$.
    In other words, there are only finitely many contexts $\str$ which define the distribution over the next symbol, $\pLM\left(\sym\mid\str\right)$.
\end{definition}
Intuitively, this framework might be useful because it \emph{bounds} the number of unique conditional distributions we have to learn.
However, as we will see later in this chapter, finite-state language models are not sufficient for modeling human language.
Nevertheless, they can still offer a baseline for modeling more complex phenomena.
They also offer a useful theoretical tool in the understanding of neural language models, which we will discuss in \cref{ch:nn-lms}.

\subsection{Weighted Finite-state Automata} \label{sec:wfsa}
Before we introduce finite-state \emph{language models}, we go on a brief detour into the theory of finite-state \emph{automata}.
As we will see, finite-state automata are a tidy and well-understood formalism.
As we will see later in \cref{sec:rnn-expressiveness}, they also provide a solid and convenient theoretical framework for understanding modern neural language models, e.g., those based on recurrent neural networks and transformers.
We, therefore, begin by briefly introducing the theory of finite-state automata with real-valued weights.

\subsubsection{Finite-state Automata}
In words, finite-state automata are one of the simplest devices for defining a formal language (cf. \cref{def:formal-language}).
We give a formal definition below.
\begin{definition}{Finite-state Automata}{fsa}
    A \defn{finite-state automaton}\index{finite-state automaton} (FSA) is a 5-tuple $\fsatuple$ where
    \begin{itemize}
        \item $\alphabet$ is an alphabet;
        \item $\states$ is a finite set of states;
        \item $\initial \subseteq \states$ is the set of initial states;
        \item $\final \subseteq \states$ the set of final or accepting states;
        \item A finite multiset $\trans \subseteq \states \times \left( \alphabet \cup  \set{\eps} \right) \times \states$.\footnote{The fact that it is a multiset reflects that it can contain multiple copies of the same element (i.e., transitions between the same pair of states with the same symbol).}
              Elements of $\trans$ are generally called \defn{transitions}\index{transition}.
    \end{itemize}
\end{definition}
The name, \emph{finite-state} automaton, stems from the requirement that the set of states $\states$ is finite, which stands in contrast to the remaining formalisms we will cover in this course, e.g., pushdown automata and recurrent neural networks.
We will denote a general finite-state automaton with a (subscripted) $\automaton$.
We will also adopt a more suggestive notation for transitions by denoting a transition $\left(\stateq_1, a, \stateq_2 \right)$ as $\uwedge{\stateq_1}{a}{\stateq_2}$.

An FSA can be graphically represented as a labeled, directed multi-graph.\footnote{The \emph{multi-} aspect of the multi-graph refers to the fact that we can have multiple transitions from any pair of states and labeled refers to the fact that we label those transitions with symbols from the alphabet $\alphabet$.}
The vertices in the graph represent the states $\stateq \in \states$ and the (labeled) edges between them the transitions in $\trans$.
The labels on the edges correspond to the input symbols $\syma \in \alphabet$ which are consumed when transitioning over the edges.
The initial states $\qinit \in \initial$ are marked by a special incoming arrow while the final states $\qfinal \in \final$ are indicated using a \emph{double} circle.

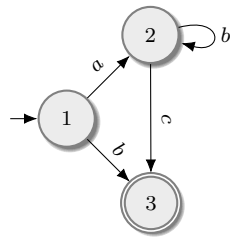
\begin{figure}[ht!]
    \centering
    \begin{tikzpicture}[node distance=8mm]
        \footnotesize
        \node[state, initial] (q1) [] { 1 };
        \node[state] (q2) [above right = of q1] { 2 };
        \node[state, accepting] (q3) [below right = of q1] { 3 };
        \draw[transition] (q1) edge[auto, sloped] node[minimum size=4mm]{ $\syma$ } (q2)
        (q1) edge[auto, sloped] node[minimum size=4mm]{ $\symb$ } (q3)
        (q2) edge[auto, sloped] node[minimum size=4mm]{ $\symc$ } (q3)
        (q2) edge[loop right] node{ $\symb$ } (q2);
    \end{tikzpicture}
    \caption{Example of a simple FSA.}
    \label{fig:fsa-example}
\end{figure}

\begin{example}{An example of a finite-state automaton}{}
    An example of an FSA can be seen in \cref{fig:fsa-example}.
    Formally, we can specify it as
    \begin{itemize}
        \item $\alphabet = \left\{\syma, \symb, \symc \right\}$
        \item $\states = \left\{1, 2, 3 \right\}$
        \item $\initial = \set{1}$
        \item $\final = \set{3}$
        \item $\trans = \{\left(1, \syma, 2\right), \left(1, \symb, 3\right), \left(2, \symb, 2\right), \left(2, \symc, 3\right)\}$
    \end{itemize}
\end{example}

A finite-state automaton sequentially reads in individual symbols of an \defn{input string}\index{input string} $\str \in \kleene{\alphabet}$ and transitions from state to state according to the transition function $\trans$.
The traversal through the automaton starts in a state $\qinit \in \initial$ (more precisely, it acts as if starting from all of them in parallel).
It then transitions from state $\stateq$ into the state $\stateq'$ upon reading the symbol $\syma$ if and only if $\uwedge{\stateq}{\syma}{\stateq'} \in \trans$.
$\eps$-labeled transitions, however, allow a finite-state machine to transition to a new state without consuming a symbol.
This is in line with $\eps$'s definition as an empty string.

A natural question to ask at this point is what happens if for a state--symbol pair $(\stateq, \syma)$ there is \emph{more than one} possible transition allowed under the relation $\trans$.
In such a case, we take \emph{all} implicit transitions simultaneously, which leads us to a pair of definitions.
\begin{definition}{Deterministic finite-state automaton}{fsa-deterministic}
    A FSA $\automaton = \fsatuple$ is \defn{deterministic}\index{deterministic FSA} if
    \begin{itemize}
        \item it does not have any $\eps$-transitions;
        \item for every $(\stateq, a) \in \states \times \alphabet$, there is at most one $\stateq' \in \states$ such that $\uwedge{\stateq}{a}{\stateq'} \in \trans$;
        \item there is a single initial state, i.e., $|\initial| = 1$.
    \end{itemize}
    Otherwise, $\automaton$ is \defn{non-deterministic}\index{non-deterministic FSA}.
\end{definition}
An important, and perhaps not entirely obvious, result is that the classes of deterministic and non-deterministic FSA are \emph{equivalent}, in the sense that you can always represent a member of one class with a member of the other.

If the automaton ends up, after reading in the last symbol of the input string, in one of the final states $\qfinal \in \final$, we say that the automaton \defn{accepts}\index{finite-state automaton!string acceptance} that string.
A finite-state automaton is therefore a computational device that determines whether a string satisfies a condition (namely, the condition that the automaton, by starting in an initial state and following one of the paths labeled with that string, ends in a final state).
A string that satisfies this condition is said to be \emph{recognized} by the automaton and the set of all strings satisfying this condition form the \emph{language} of the automaton.\footnote{We also say that the automaton \emph{recognizes} this set of strings (language).}
\begin{definition}{Language of a finite-state automaton}{}
    Let $\automaton = \fsatuple$ be an finite-state automaton.
    The \defn{language}\index{finite-state automaton!recognized language} of $\automaton$, $\lang\left(\automaton\right)$ is defined as
    \begin{equation}
        \lang\left(\automaton\right) \defeq \left\{\str \mid \ \str \text{ is recognized by } \automaton \right\}
    \end{equation}
\end{definition}
Abstractly, a finite-state automaton is hence a specification of a set of \emph{rules} that strings must satisfy to be included in its language.
The set of languages that finite-state automata can recognize is known as the class of \defn{regular languages}.
\begin{definition}{Regular language}{}
    A language $\lang\subseteq\kleene{\alphabet}$ is \defn{regular}\index{regular language} if and only if it can be recognized by an unweighted finite-state automaton, i.e., if there exists a finite-state automaton $\automaton$ such that $\lang = \lang\left(\automaton\right)$.
\end{definition}

\begin{figure}[ht!]
    \centering
    \begin{subfigure}[\tstep]{0.45\textwidth}
        \centering
        \begin{tikzpicture}[node distance=8mm]
            \footnotesize
            \node[state, initial] (q1) [] { 1 };
            \node[state] (q2) [above right = of q1] { 2 };
            \node[state] (q3) [below right = of q1] { 3 };
            \node[state] (q4) [right = of q2] { 4 };
            \node[state] (q5) [right = of q3] { 5 };
            \node[state, accepting] (q6) [below right = of q4] { 6 };
            \draw[transition] (q1) edge[auto, sloped] node[minimum size=4mm]{ $\syma$ } (q2)
            (q1) edge[auto, sloped] node[minimum size=4mm]{ $\symb$ } (q3)
            (q2) edge[auto, sloped] node[minimum size=4mm]{ $\symc$ } (q4)
            (q2) edge[loop below] node{ $\symb$ } (q2)
            (q3) edge[auto, sloped] node[minimum size=4mm]{ $\symc$ } (q4)
            (q3) edge[auto, sloped] node[minimum size=4mm]{ $\symb$ } (q5)
            (q4) edge[auto, sloped] node[minimum size=4mm]{ $\syma$ } (q6)
            (q5) edge[auto, sloped] node[minimum size=4mm]{ $\syma$ } (q6);
        \end{tikzpicture}
        \subcaption{A deterministic FSA, $\automaton_1$. Each state only has one outgoing transition labeled with the same symbol.} \label{fig:fsa-det}
    \end{subfigure}
    \hfill
    \begin{subfigure}[\tstep]{0.45\textwidth}
        \centering
        \begin{tikzpicture}[node distance=8mm]
            \footnotesize
            \node[state, initial] (q1) [] { 1 };
            \node[state] (q2) [above right = of q1] { 2 };
            \node[state] (q3) [below right = of q1] { 3 };
            \node[state] (q4) [right = of q2] { 4 };
            \node[state] (q5) [right = of q3] { 5 };
            \node[state, accepting] (q6) [below right = of q4] { 6 };
            \draw[transition] (q1) edge[auto, sloped] node{ $\syma$ } (q2)
            (q1) edge[auto, sloped] node{ $\syma$ } (q3)
            (q2) edge[auto, sloped] node{ $\symc$ } (q4)
            (q2) edge[loop below] node{ $\symb$ } (q2)
            (q3) edge[auto, sloped] node{ $\symb$ } (q4)
            (q3) edge[auto, sloped] node{ $\symb$ } (q5)
            (q4) edge[auto, sloped] node{ $\syma$ } (q6)
            (q5) edge[auto, sloped] node{ $\syma$ } (q6);
        \end{tikzpicture}
        \subcaption{A non-deterministic FSA, $\automaton_2$. State $1$ has two outgoing transitions labeled with $a$ whereas state $3$ has two outgoing transitions labeled with $\symb$.} \label{fig:fsa-non-det}
    \end{subfigure}
    \caption{Examples of a deterministic and a non-deterministic FSA.}
    \label{fig:fsa-examples}
\end{figure}
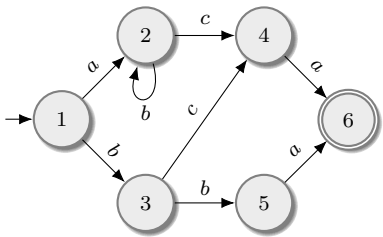
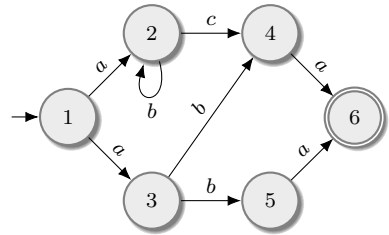

\begin{example}{Additional examples of finite-state automata}{}
    Additional simple examples of FSAs are shown in \cref{fig:fsa-examples}.
    The FSA in \cref{fig:fsa-det}, for example, can formally be defined with
    \begin{itemize}
        \item $\alphabet = \left\{\syma, \symb, \symc \right\}$
        \item $\states = \left\{1, 2, 3, 4, 5, 6 \right\}$
        \item $\initial = \set{1}$
        \item $\final = \set{6}$
        \item $\trans = \{\left(1, \syma, 2\right), \left(1, \symb, 3\right), \left(2, \symb, 2\right),\allowbreak \left(2, \symc, 4\right), \left(3, \symc, 4\right), \left(3, \symb, 5\right), \left(4, \syma, 6\right), \left(5, \syma, 6\right)\}$
    \end{itemize}
    The FSA in \cref{fig:fsa-det} is deterministic while the one in \cref{fig:fsa-non-det} is non-deterministic.

    A few examples of strings accepted by the $\automaton_1$ include $\symb\symb\syma$, $\symb\symc\syma$, $\syma\symc\syma$, $\syma\symb\symc\syma$, $\syma\symb\symb\symc\syma$, $\syma\symb\symb\symb\symc\syma, \dots$.
    In fact, due to the self-loop at state $2$, the symbol $\symb$ can appear an arbitrary number of times at position 2 in the accepted string $\syma\symb\symc\syma$.
    Notice that, starting from the state $1$ and following the transitions dictated by any of the accepted strings, we always end up in the only final state, state $6$.
    In particular, the string ``$\syma\symb\symb\symc\syma$'' is accepted with the following set of transitions in $\automaton_1$:
    \begin{equation*}
        \uwedge{1}{\syma}{2}, \uwedge{2}{\symb}{2}, \uwedge{2}{\symb}{2}, \uwedge{2}{\symc}{4}, \uwedge{4}{\syma}{6}.
    \end{equation*}

\end{example}

\subsubsection{Weighted Finite-state Automata}
A common and very useful augmentation to finite-state automata is through the addition of \emph{weights} on the transitions.
The general theory of weighted automata makes use of semiring theory, which is beyond the scope of this course.\footnote{Semirings and semiring-weighted formal languages are covered in detail in the Advanced Formal Language Theory course offered at ETH as well.}
In this course, we will limit ourselves to the study of automata with real-valued weights.
\begin{definition}{Real-weighted Finite-State Automaton}{}
    A \defn{real-weighted finite-state automaton}\index{weighted finite-state automaton} (WFSA) $\automaton$ is a 5-tuple $\wfsatuple$ where
    \begin{itemize}
        \item $\alphabet$ is a finite alphabet;
        \item $\states$ is a finite set of states;
        \item $\trans \subseteq \states \times \left( \alphabet \cup \left\{ \eps \right\} \right) \times \R \times \states$ a finite multiset of transitions;\footnote{Again, we use the notation $\edge{\stateq}{\syma}{w}{\stateq'}$ to denote $\left(\stateq, \syma, w, \stateq'\right) \in \trans$.}
        \item $\initf : \states \rightarrow \R$ a weighting function over $\states$;
        \item $\finalf : \states \rightarrow \R$ a weighting function over $\states$.
    \end{itemize}
\end{definition}
Notice that we omit the initial and final state sets from the definition of WFSAs.
Those can implicitly be specified by the states given non-zero initial or final weights by the $\initf$ and $\finalf$ functions, i.e., $\initial = \left\{\stateq \in \states \mid \initf\left(\stateq\right) \neq 0\right\}$ and $\final = \left\{\stateq \in \states \mid \finalf\left(\stateq\right) \neq 0\right\}$.
We might refer to them in the text later for notational convenience and clarity of exposition.
We will also sometimes denote transition weights with $\transitionWeight\left(\edge{\stateq}{\syma}{w}{\stateq'}\right) \defeq w$.

\begin{figure}[ht]
    \centering
    \centering
    \begin{tikzpicture}[node distance = 1cm]
        \footnotesize
        \node[state, initial, initial text = $0.3$] (q1) [] { $1$ };
        \node[state] (q2) [above right = of q1] { $2$ };
        \node[state] (q3) [below right = of q1] { $3$ };
        \node[state] (q4) [right = of q2] { $4$ };
        \node[state] (q5) [right = of q3] { $5$ };
        \node[state, accepting] (q6) [below right = of q4] { $6/\textcolor{ETHRed}{\frac{1}{e}}$ };
        \draw[transition] (q1) edge[auto, sloped] node{ $\syma/0.5$ } (q2)
        (q1) edge[auto, sloped] node{ $\symb/\frac{1}{\pi}$ } (q3)
        (q2) edge[loop above] node{ $\symb/0.63$ } (q2)
        (q2) edge[auto, sloped] node{ $\symc/0.9$ } (q4)
        (q3) edge[auto, sloped] node{ $\symc/0.21$ } (q4)
        (q3) edge[auto, sloped] node{ $\symb/0.13$ } (q5)
        (q4) edge[auto, sloped] node{ $\syma/\frac{1}{\pi \cdot e}$ } (q6)
        (q5) edge[auto, sloped] node{ $\syma/0.29$ } (q6);
    \end{tikzpicture}
    \caption{The WFSA corresponding to the FSA from \cref{fig:fsa-det}.}
    \label{fig:weighted-fsa}
\end{figure}
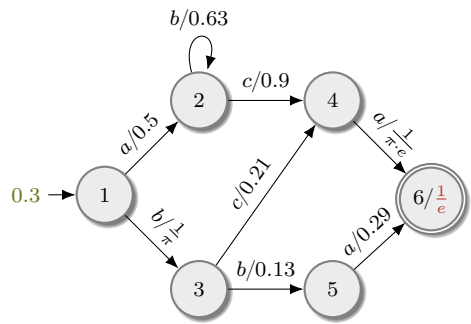

Graphically, we write the transition weights on the edges of the graph representing the WFSA after the output symbol, separated by a ``/''.
The same separator is also used to separate the state name from its \textcolor{ETHRed}{final weight}, which is written in the node.
The \textcolor{ETHGreen}{initial weights}, however, are written on the incoming arrow denoting initial states.
\begin{example}{An example of a weighted finite-state automaton}{}
    \cref{fig:weighted-fsa} shows a weighted version of the FSA from \cref{fig:fsa-det} above.
\end{example}

The connection of WFSAs to graphs makes it natural to define a set of transition matrices specified by a WFSA.
\begin{definition}{Transition matrix}{}
    Let $\wfsa = \wfsatuple$ be a WFSA.
    For any $\syma \in \alphabet$, we define the \defn{symbol-specific transition matrix} $\transMtx^{\left(\syma\right)}$ as the transition matrix of the graph restricted to $\syma$-labeled transitions.
    We also define the (full) \defn{transition matrix}\index{weighted finite-state automaton!transition matrix} as $\transMtx \defeq \sum_{\syma \in \alphabet}\transMtx^{\left(\syma\right)}$.
\end{definition}

\begin{example}{Examples of transition matrices}{}
    Consider the WFSA $\wfsa$ in \cref{fig:weighted-fsa}. The (symbol-specific) transition matrices for $\wfsa$ are
    \begin{align*}
        \transMtx^{\left(\syma\right)}                                                                              & = \begin{pmatrix}
                                                                                                                            0 & 0.5 & 0 & 0 & 0 & 0                     \\
                                                                                                                            0 & 0   & 0 & 0 & 0 & 0                     \\
                                                                                                                            0 & 0   & 0 & 0 & 0 & 0                     \\
                                                                                                                            0 & 0   & 0 & 0 & 0 & \frac{1}{\pi \cdot e} \\
                                                                                                                            0 & 0   & 0 & 0 & 0 & 0.29                  \\
                                                                                                                            0 & 0   & 0 & 0 & 0 & 0                     \\
                                                                                                                        \end{pmatrix}                    \\
        \transMtx^{\left(\symb\right)}                                                                              & = \begin{pmatrix}
                                                                                                                            0 & 0    & \frac{1}{\pi} & 0 & 0    & 0 \\
                                                                                                                            0 & 0.63 & 0             & 0 & 0    & 0 \\
                                                                                                                            0 & 0    & 0             & 0 & 0.13 & 0 \\
                                                                                                                            0 & 0    & 0             & 0 & 0    & 0 \\
                                                                                                                            0 & 0    & 0             & 0 & 0    & 0 \\
                                                                                                                            0 & 0    & 0             & 0 & 0    & 0 \\
                                                                                                                        \end{pmatrix}                        \\
        \transMtx^{\left(\symc\right)}                                                                              & = \begin{pmatrix}
                                                                                                                            0 & 0 & 0 & 0    & 0 & 0 \\
                                                                                                                            0 & 0 & 0 & 0.9  & 0 & 0 \\
                                                                                                                            0 & 0 & 0 & 0.21 & 0 & 0 \\
                                                                                                                            0 & 0 & 0 & 0    & 0 & 0 \\
                                                                                                                            0 & 0 & 0 & 0    & 0 & 0 \\
                                                                                                                            0 & 0 & 0 & 0    & 0 & 0 \\
                                                                                                                        \end{pmatrix}                                       \\
        \transMtx = \transMtx^{\left(\syma\right)} + \transMtx^{\left(\symb\right)}+ \transMtx^{\left(\symc\right)} & = \begin{pmatrix}
                                                                                                                            0 & 0.5  & \frac{1}{\pi} & 0    & 0    & 0                     \\
                                                                                                                            0 & 0.63 & 0             & 0.9  & 0    & 0                     \\
                                                                                                                            0 & 0    & 0             & 0.21 & 0.13 & 0                     \\
                                                                                                                            0 & 0    & 0             & 0    & 0    & \frac{1}{\pi \cdot e} \\
                                                                                                                            0 & 0    & 0             & 0    & 0    & 0.29                  \\
                                                                                                                            0 & 0    & 0             & 0    & 0    & 0                     \\
                                                                                                                        \end{pmatrix}
    \end{align*}
\end{example}

\subsubsection{Paths and Path Weights}
A path is an important concept when talking about (weighted) finite-state automata as it defines the basic structure by which a string is recognized or weighted.
We now give a formal definition of a path and discuss how to weight paths.

\begin{definition}{Path}{}
    A \defn{path}\index{path} $\apath$ is an element of $\kleene{\trans}$ with \emph{consecutive transitions}, meaning that it is of the form $\left( \edge{\stateq_1}{\bullet}{\bullet}{\stateq_2}, \edge{\stateq_2}{\bullet}{\bullet}{\stateq_3} \cdots  \edge{\stateq_{n-1}}{\bullet}{\bullet}{\stateq_n}\right) $, where $\bullet$ is a placeholder.\footnote{Notice we use the Kleene closure on the set $\trans$ here. It thus represents any sequence of transitions $\in \trans$}
    The \defn{length}\index{path!length} of a path is the number of transition in it; we denote the length as $|\apath|$.
    We use $\prevq\left(\apath\right)$ and $\nextq\left(\apath\right)$ to denote the origin and the destination of a path, respectively.
    The \defn{yield}\index{path!yield} of a path is the concatenation of the input symbols on the edges along the path, which we will mark with $\yield\left(\apath\right)$.
    Furthermore, we denote sets of paths with capital $\paths$.
    Throughout the text, we will use a few different variants involving $\paths$ to avoid clutter:
    \begin{itemize}
        \item $\paths(\automaton)$ as the set of all paths in automaton $\automaton$;
        \item $\paths(\automaton, \str)$ as the set of all paths in automaton $\automaton$ with yield $\str \in \kleene{\alphabet}$;
        \item $\paths(\automaton, \stateq, \stateq')$ as the set of all paths in automaton $\automaton$ from state $\stateq$ to state $\stateq'$.\looseness=-1
    \end{itemize}
\end{definition}

One of the most important questions when talking about weighted formalisms like weighted finite-state automata is how to \emph{combine} weights of atomic units like transitions into weights of complete \emph{structures}.\footnote{In the case of WFSAs, a structure is a path. In the next section, we will see how to combine weights from basic units into \emph{trees}.}
We begin by multiplicatively combining the weights of individual transitions in a path into the weights of the full path.
\begin{definition}{Path Weight}{}
    The \defn{inner path weight}\index{path!inner weight} $\innerweight\left(\apath\right)$ of a path $\apath = \edge{\stateq_1}{a_1}{w_1}{\stateq_2} \cdots \edge{\stateq_{\pathlen-1}}{a_\pathlen}{w_\pathlen}{\stateq_\pathlen}$ is defined as
    \begin{equation}
        \innerweight\left(\apath\right) = \prod_{\idx = 1}^\pathlen w_\idx.
    \end{equation}
    The \defn{(full) path weight}\index{path!weight} of the path $\apath$ is then defined as
    \begin{equation} \label{eq:path-weight}
        \weight\left(\apath\right) = \initf \left( \prevq \left( \apath \right) \right) \innerweight \left( \apath \right) \finalf \left( \nextq \left( \apath \right) \right).
    \end{equation}
    A path $\apath$ is called \defn{accepting}\index{path!accepting} or \defn{successful}\index{path!successful} if $\weight\left(\apath\right) \neq \zero$.
\end{definition}
The inner path weight is therefore the product of the weights of the transitions on the path, while the (full) path weight is the product of the transition weights as well as the initial and final weights of the origin and the destination of the path, respectively.

\subsubsection{String Acceptance Weights and Weighted Regular Languages}
When we introduced unweighted finite-state automata, we defined the important concept of recognizing a string and recognizing a language.
We generalize these concepts to the very natural quantity of the weight assigned by a WFSA to a string $\str \in \kleene{\alphabet}$, i.e., its acceptance weight, or stringsum, as the sum of the weights of the paths that yield $\str$.
\begin{definition}{Stringsum}{}
    The \defn{stringsum}\index{weighted finite-state automaton!stringsum}, string weight, or acceptance weight of a string $\str \in \kleene{\alphabet}$ under a WFSA $\automaton$ is defined as
    \begin{equation} \label{eq:acceptance-weight}
        \automaton \left( \str \right) \defeq \sum_{\apath \in \paths\left( \automaton, \str \right) }  \weight \left( \apath \right).
    \end{equation}
\end{definition}
This naturally generalizes the notion of acceptance by an unweighted FSA---whereas an unweighted FSA only makes a binary decision of accepting or rejecting a string, a weighted FSA always accepts a string with a specific weight.
This leads to the definition of the \emph{weighted language} of the WFSA.
\begin{definition}{Weighted language of a weighted finite-state automaton}{}
    Let $\wfsa$ be a WFSA.
    Its \defn{(weighted) language} is defined as
    \begin{equation}
        \lang\left(\wfsa\right) \defeq \left\{\left(\str, \wfsa\left(\str\right)\right) \mid \str \in \kleene{\alphabet}\right\}.
    \end{equation}
\end{definition}
We say a language is a weighted regular language if it is a language of some WFSA:
\begin{definition}{Weighted regular language}{}
    A weighted language $\lang$ is a weighted regular language if there exists a WFSA $\wfsa$ such that $\lang = \lang\left(\wfsa\right)$.
\end{definition}

Lastly, we also define the full and state-specific allsum of the automaton.
The former refers to the total weight assigned to \emph{all} possible strings, or all possible paths whereas the latter refers to the sum of the path weights of the paths stemming from a specific state.
\begin{definition}{State-specific allsum}{}
    Let $\wfsa = \wfsatuple$ be a WFSA.
    The \defn{allsum}\index{weighted finite-state automaton!state-specific allsum} of a state $\stateq \in \states$ is defined as
    \begin{equation}\label{eq:wfsa-state-allsum}
        \allsum\left(\wfsa, \stateq\right) = \sum_{\substack{\apath \in \paths\left(\wfsa\right) \\ \stateq_1 = \stateq}}         \innerweight\left(\apath\right) \finalf\left(\nextq\left(\apath\right)\right).
    \end{equation}
\end{definition}
State-specific allsums are also referred to as the \defn{backward values} in the literature and are often denoted as $\backwardFun{\stateq}$.
\begin{definition}{WFSA allsum}{}
    Let $\wfsa = \wfsatuple$ be a WFSA.
    The \defn{allsum}\index{weighted finite-state automaton!allsum} of $\wfsa$ is defined as
    \begin{equation}\label{eq:wfsa-allsum}
        \allsum\left(\automaton\right) = \sum_{\str \in \kleene{\alphabet}} \wfsa\left(\str\right) = \sum_{\str \in \kleene{\alphabet}} \sum_{\apath \in \paths\left( \automaton, \str \right) }  \weight \left( \apath \right) = \sum_{\apath \in \paths\left(\wfsa\right)} \weight\left(\apath\right).
    \end{equation}
\end{definition}
The second equality in \cref{eq:wfsa-allsum} comes from the crucial observation that the double sum in the second term sums over precisely \emph{all paths} of the automaton $\wfsa$, which is where the name of the quantity comes from \emph{allsum}.\footnote{Analogously, given some (implicitly defined) set of paths $\sS$, we will name the sum over the weights of the paths in $\sS$ the allsum over $\sS$}
This is easy to see if we consider that by summing over all possible strings, we enumerate all possible path yields, and each path in the automaton has a yield $\in \kleene{\alphabet}$.
$\allsum\left(\automaton\right)$ is again the result of summing over infinitely many terms (whether the set of strings in $\kleene{\alphabet}$ of the infinitely many paths in a cyclic WFSA), and might therefore not necessarily be finite.
For reasons which will become clear shortly, we will say that a WFSA $\wfsa$ is \defn{normalizable}\index{weighted finite-state automaton!normalizable} if $\allsum\left(\wfsa\right) < \infty$.

Note that the sum in \cref{eq:acceptance-weight} only contains one term if the automaton is deterministic.
Whenever the automaton is non-deterministic, or when we are interested in the sum of paths with \emph{different} yields as in \cref{eq:wfsa-allsum}, the interactions (namely, the \emph{distributive law}) between the \emph{sum} over the different \emph{paths} and the \emph{multiplications} over the \emph{transitions} in the paths play an important role when designing efficient algorithms.
Indeed, many algorithms defined for WFSAs rely on decompositions of such sums enabled by the distributive law.\footnote{Many such examples are covered in the Advanced Formal Language Theory course.}

\subsubsection{Accessibility and Probabilistic Weighted Finite-state Automata}

An important property of states of a WFSA which we will need when investigating the tightness of finite-state language models is accessibility.
\begin{definition}{(Co)-Accessible and useful states}{acc-co-acc}
    A state $\stateq\in \states$ of a WFSA is \defn{accessible}\index{accessible state} if there is a non-zero-weighted path to $\stateq$ from some state $\qinit$ with $\initf\left(\qinit\right) \neq 0$; it is \defn{co-accessible state}\index{co-accessible} if there is a non-zero-weighted path from $\stateq$ to some state $\qfinal$ with $\finalf\left(\qfinal\right) \neq 0$.
    It is \defn{useful}\index{useful state} if it is both accessible and co-accessible, i.e., $\stateq$ appears on some non-zero-weighted accepting path.
\end{definition}

\begin{definition}{Trim automaton}{trim}
    \defn{Trimming}\index{weighted finite-state automaton!trim} a WFSA means removing its useless states.\footnote{This does not affect the weights of the strings with $\weight\left(\str\right) \neq 0$.}
    Removing the non-useful states means removing their rows and columns from $\transMtx$ as well as their rows from $\initfVect$ and $\finalfVect$, yielding possibly smaller $\transMtx', \initfVect'$ and $\finalfVect'$.
\end{definition}

We will use WFSAs to specify language models.
However, not every WFSA is a language model, i.e., a distribution over strings.
Generally, the weight of a string could be negative if we allow arbitrary real weights.
Thus, a restriction we will impose on \emph{all} weighted automata that represent finite-state language models is that the weights be \emph{non-negative}. \ryan{We should mention this is not necessary. \response{Anej} In what sense? Technically, they could be negative, but that would }

Furthermore, a special class of WFSAs that will be of particular interest later is probabilistic WFSAs.\anej{do the initial weights have to sum to $1$?}
\begin{definition}{Probabilistic Weighted Finite-State Automaton}{stochastic-wfsa}
    A WFSA $\wfsa = \wfsatuple$ is \defn{probabilistic}\index{probabilistic finite-state automaton} (a PFSA) if
    \begin{equation}
        \sum_{\stateq \in \states} \initf\left(\stateq\right) = 1
    \end{equation}
    and, for all $\stateq \in \states$ and all outgoing transitions $\edge{\stateq}{\syma}{w}{\stateq'} \in \trans$ it holds that
    \begin{align}
        \initf\left(\stateq\right)  & \geq 0 \\
        \finalf\left(\stateq\right) & \geq 0 \\
        w                           & \geq 0
    \end{align}
    and
    \begin{equation}
        \sum_{\edge{\stateq}{\syma}{w}{\stateq'}} w + \finalf\left(\stateq\right) = 1.
    \end{equation}
\end{definition}
This means that the initial weights of all the states of the automaton form a probability distribution (the initial weight of a state corresponds to the \emph{probability} of starting in it), as well as that, for any state $\stateq$ in the WSFA, the weights of its outgoing transitions (with any label) together with its final weight form a valid discrete probability distribution.
In a certain way, probabilistic finite-state automata naturally correspond to \emph{locally normalized} language models, as we explore in the next subsection.

\paragraph{The $\eos$ symbol and the final weights.}
Notice that the final weights in a PFSA play an analogous role to the $\eos$ symbol: the probability of ending a path in a specific state $\stateq$---and therefore ending a string---is $\stateq$'s final weight!
That is, the probability $\finalf\left(\qfinal\right)$ for some $\qfinal \in \states$, representing the probability of ending the path in $\qfinal$, is analogous to the probability of ending a string $\str$, $\pLNSM\left(\eos \mid \str\right)$, where $\qfinal$ ``represents'' the string (history) $\str$.\footnote{Due to the possible non-determinism of WFSAs, the connection is of course not completely straightforward, but the point still stands.}
When modeling language with weighted finite-state automata, we will therefore be able to avoid the need to specify the special symbol and rather rely on the final weights, which are naturally part of the framework.

\subsection{Finite-state Language Models}
We can now formally define what it means for a language model to be finite-state:
\begin{definition}{Finite-state language models}{}
    A language model $\pLM$ is \defn{finite-state}\index{language model!finite-state} if it can be represented by a weighted finite-state automaton, i.e., if there exists a WFSA $\automaton = \wfsatuple$ such that $\lang\left(\automaton\right) = \lang\left(\pLM\right)$.
    Equivalently, we could say that $\pLM$ is finite-state if its language is a weighted regular language.
\end{definition}

On the other hand, given a WFSA $\wfsa$, there are two established ways of defining a \emph{probability} of  string.
\subsubsection{String Probabilities in a Probabilistic Finite-state Automaton} \label{sec:pfsa-str-prob}
In a probabilistic FSA (cf. \cref{def:stochastic-wfsa}), any action from a state $\stateq \in \states$ is associated with a probability.
Since the current state completely encodes all the information of the input seen so far in a finite-state automaton, it is intuitive to see those probabilities as conditional probabilities of the next symbol given the input seen so far.
One can, therefore, define the probability of a path as the product of these individual ``conditional'' probabilities.
\begin{definition}{Path probability in a PFSA}{}
    We call the weight of a path $\apath \in \paths\left(\wfsa\right)$ in a probabilistic FSA the \defn{probability} of the path $\apath$.
\end{definition}
This alone is not enough to define the probability of any particular string $\str \in \kleene{\alphabet}$ since there might be multiple accepting paths for $\str$.
Naturally, we define the probability of $\str$ as the sum of the individual paths that recognize it:
\begin{definition}{String probability in a PFSA}{}
    We call the stringsum of a string $\str \in \kleene{\alphabet}$ in a probabilistic FSA the \defn{probability} of the string $\str$:
    \begin{equation}
        \pdens_\wfsa\left(\str\right) \defeq \wfsa\left(\str\right).
    \end{equation}
\end{definition}

Crucially, notice that these two definitions did not require any \emph{normalization} over all possible paths or strings.
This closely resembles the way we defined locally normalized models based on the conditional probabilities of a sequence model.
Again, such definitions of string probabilities are attractive as the summation over all possible strings is avoided.
However, a careful reader might then ask themself: do these probabilities actually sum to $1$, i.e., is a probabilistic FSA \emph{tight}?
As you might guess, they might \emph{not}.\footnote{Notice that, however, whenever a PFSA is tight, its allsum is $1$.}
We explore this question in \cref{sec:tight-pfsas}.

\subsubsection{String Probabilities in a General Weighted Finite-state Automaton}
To define string probabilities in a general weighted FSA, we use the introduced notions of the stringsum and the allsum.
The allsum allows us to tractably \emph{normalize} the stringsum to define the globally normalized probability of a string $\str$ as the \emph{proportion} of the total weight assigned to all strings that is assigned to $\str$.\footnote{We will see how the allsum can be computed tractably in \cref{sec:normalizing-finite-state-lms}.}
\begin{definition}{String probability in a WFSA}{}
    Let $\wfsa = \wfsatuple$ be a normalizable WFSA with non-negative weights.
    We define the probability of a string $\str \in \kleene{\alphabet}$ under $\wfsa$ as
    \begin{equation} \label{eq:wfsa-str-prob}
        \pdens_\wfsa\left(\str\right) \defeq \frac{\automaton\left(\str\right)}{\allsum\left(\automaton\right)}.
    \end{equation}
\end{definition}

\subsubsection{Language Models Induced by a WFSA}
With the notions of string probabilities in both probabilistic and general weighted FSAs, we can now define the language model induced by $\wfsa$ as follows.
\begin{definition}{A language model induced by a WFSA}{}
    Let $\wfsa = \wfsatuple$ be a WFSA.
    We define the \defn{language model induced by $\wfsa$}\index{weighted finite-state automaton!induced language model} as the following probability distribution over $\kleene{\alphabet}$
    \begin{equation} \label{eq:wfsa-lm}
        \pLM_\wfsa\left(\str\right) \defeq \pdens_\wfsa\left(\str\right).
    \end{equation}
\end{definition}

It is easy to see that while global normalization requires the computation of the allsum, language models induced by weighted FSAs through \cref{eq:wfsa-str-prob} are \emph{globally normalized} and thus always tight.
In the next subsection, we consider \emph{how} the quantities needed for computing \cref{eq:wfsa-str-prob} can be computed.
Of particular interest will be the quantity $\allsum\left(\wfsa\right)$, as it involves the summation over possibly infinitely many terms and therefore requires some clever tricks to be computed.

\subsection{Normalizing Finite-state Language Models} \label{sec:normalizing-finite-state-lms}
In this subsection, we develop an algorithm for normalizing a globally normalized language model (cf. \cref{def:globally-normalized-model}) defined by a WFSA, i.e., an algorithm for computing the allsum $\allsum\left(\wfsa\right)$ whenever this quantity is finite.
Moreover, the derivation will also reveal necessary and sufficient conditions for WFSAs to be normalizable.

\paragraph{Converting a matrix of pairwise pathsums to the allsum.}
Before we consider how to compute $\allsum\left(\automaton\right)$, let us first consider a much simpler problem.
Suppose we had a matrix $\mM$, which contained at the entry $\mM_{\idxi \idxj}$ the sum of all the inner weights over all paths between the states $\idxi$ and $\idxj$, i.e., \ryan{Why is the $w_{\mathrm{I}}$ here? \response{Anej} $\mM$ contains the pathsums without the initial/final weights, that's why the inner weight, right?}
\begin{equation*}
    \mM_{\idxi \idxj} = \sum_{\apath \in \paths\left( \automaton,  \idxi, \idxj \right) }  \innerweight \left( \apath \right).
\end{equation*}
How could we then compute the quantity $\allsum\left(\wfsa\right)$?
\begin{align}
    \allsum\left(\wfsa\right) & = \sum_{\apath \in \paths\left( \wfsa \right) }  \weight \left( \apath \right)                                                                                                                                                               \\
                              & = \sum_{\apath \in \paths\left( \wfsa \right) }  \initf \left( \prevq \left( \apath \right) \right) \innerweight \left( \apath \right) \finalf \left( \nextq \left( \apath \right) \right)                                                   \\
                              & = \sum_{\idxi, \idxj \in \states}\sum_{\apath \in \paths\left( \automaton, \idxi, \idxj \right) }  \initf \left( \prevq \left( \apath \right) \right) \innerweight \left( \apath \right) \finalf \left( \nextq \left( \apath \right) \right) \\
                              & = \sum_{\idxi, \idxj \in \states}\sum_{\apath \in \paths\left( \automaton, \idxi, \idxj \right) }  \initf \left( \idxi \right) \innerweight \left( \apath \right) \finalf \left( \idxj \right)                                               \\
                              & = \sum_{\idxi, \idxj \in \states} \initf \left( \idxi \right) \left(\sum_{\apath \in \paths\left( \automaton, \idxi, \idxj \right) }  \innerweight \left( \apath \right) \right) \finalf \left( \idxj \right)                                \\
                              & = \sum_{\idxi, \idxj \in \states} \initf \left( \idxi \right) \mM_{\idxi \idxj} \finalf \left( \idxj \right)                                                                                                                                 \\
                              & = \initfVect \mM \finalfVect,
\end{align}
where $\initfVect$ and $\finalfVect$ denote the vectors resulting from the ``vectorization'' of the functions $\initf$ and $\finalf$, i.e., $\initfVect_{\idx} = \initf\left(\idx\right)$ and $\finalfVect_{\idx} = \finalf\left(\idx\right)$.
This also explains the naming of the functions $\initf$ and $\finalf$: the initial weights function $\initf$, ``\underline{l}ambda'' appears on the \underline{l}eft side of the closed form expression for $\allsum\left(\wfsa\right)$ and the definition of the path weight (cf. \cref{eq:path-weight}), whereas the final weights function $\finalf$, \underline{r}ho, appears on the \underline{r}ight side of the expression and the definition of the path weight.

\paragraph{Computing the matrix of pairwise pathsums.}
Let $\transMtx$ be the transition matrix of the automaton $\wfsa$.
Notice that the entry $\transMtx_{\idxi\idxj}$ by definition contains the sum of the inner weights of all paths of length exactly $1$ (individual transitions) between the states $\idxi$ and $\idxj$.
We also define $\transMtx^0 = \mI$, meaning that the sum of the weights of the paths between $\idxi$ and $\idxj$ of length zero is $0$ if $\idxi \neq \idxj$ and $1$ (the unit for multiplication) if $\idxi = \idxj$.
This corresponds to not transitioning, i.e., staying in place, if $\idxi = \idxj$.
We next state a basic result from graph theory.
\begin{lemma}{}{matrix-prod-allsum}
    Let $\transMtx$ be the transition matrix of some weighted directed graph $\graph$.
    Then the matrix $\transMtx^d$ contains the allsum of all paths of length \emph{exactly} $d$, i.e.,
    \begin{equation}
        \transMtx^d_{\idxi, \idxj} = \sum_{\substack{\apath \in \paths\left(\automaton, \idxi, \idxj\right)\\|\apath| = d}} \innerweight\left(\apath\right).
    \end{equation}
\end{lemma}
\begin{proof}
    By induction on the path length.
    Left as an exercise for the reader.
\end{proof}
It follows directly that the matrix
\begin{equation*}
    \transMtx^{\leq d} \defeq \sum_{\idxk = 1}^d \transMtx^{\idxk}
\end{equation*}
contains the pairwise pathsums of paths of length \emph{at most} $d$.

In general, the WFSA representing a \ngram{} language model can of course be cyclic.
This means that the number of paths in $\paths\left(\wfsa\right)$ might be infinite and they might be of arbitrary length (which is the result of looping in a cycle arbitrarily many times).
To compute the pairwise pathsums over \emph{all} possible paths, we, therefore, have to compute
\begin{equation} \label{eq:transMtx-closure}
    \kleene{\transMtx} \defeq \lim_{d \rightarrow \infty} \transMtx^{\leq d} = \sum_{d = 0}^\infty \transMtx^d.
\end{equation}
This is exactly the matrix form of the \emph{geometric sum}.
Similarly to the scalar version, we can manipulate the expression \cref{eq:transMtx-closure} to arrive to a closed-form expression for computing it:
\begin{align}
    \kleene{\transMtx} & = \sum_{d = 0}^\infty \transMtx^d                       \\
                       & = \mI + \sum_{d = 1}^\infty \transMtx^d                 \\
                       & = \mI + \sum_{d = 1}^\infty \transMtx \transMtx^{d - 1} \\
                       & = \mI + \transMtx \sum_{d = 1}^\infty \transMtx^{d - 1} \\
                       & = \mI + \transMtx \sum_{d = 0}^\infty \transMtx^{d}     \\
                       & = \mI + \transMtx \kleene{\transMtx}.
\end{align}
If the inverse of $\left(\mI - \transMtx\right)$ exists, we can further rearrange this equation to arrive at
\begin{align}
    \kleene{\transMtx}                                & = \mI + \transMtx \kleene{\transMtx}  \\
    \kleene{\transMtx} - \transMtx \kleene{\transMtx} & = \mI                                 \\
    \kleene{\transMtx} - \kleene{\transMtx} \transMtx & = \mI                                 \\
    \kleene{\transMtx} \left(\mI - \transMtx\right)   & = \mI                                 \\
    \kleene{\transMtx}                                & = \inv{\left(\mI - \transMtx\right)}.
\end{align}

This means that, if $\left(\mI - \transMtx\right)$ exists, we can compute the pairwise pathsums by simply inverting it!
Using the remark above on how to convert a matrix of pairwise pathsums into the full allsum, we can therefore see that we can globally normalize an \ngram{} language model by computing a matrix inversion!
Since the runtime of inverting a $\mtxDim \times \mtxDim$ matrix is $\bigo\left(\mtxDim^3\right)$, and $\mtxDim = |\states|$ for a transition matrix of a WFSA with states $\states$, we can globally normalize a \ngram{} language model in time cubic in the number of its states.
This is a special case of the general algorithm by \citet{lehmann1977algebraic}.
Note, however, that this might still be prohibitively expensive: as we saw, the number of states in a \ngram{} model grows exponentially with $\ngr$, and even small $\ngr$'s and reasonable alphabet sizes might result in a non-tractable number of states in the WFSA with the cubic runtime.

We still have to determine when the infinite sum in \cref{eq:transMtx-closure} converges.
One can see by writing out the product $\transMtx^d$ in terms of its eigenvalues that the entries of $\transMtx^d$ diverge towards $\pm \infty$ as soon as the magnitude of any of $\transMtx$'s eigenvalues is larger than $1$.
This means that $\norm{\transMtx}_2 < 1$ (spectral norm) is a necessary condition for the infinite sum to exist.
This is, however, also a \emph{sufficient} condition: if $\norm{\transMtx}_2 < 1$, all of $\transMtx$'s eigenvalues are smaller than $1$ in magnitude, meaning that the eigenvalues of $\mI - \transMtx$ are strictly positive and the matrix $\mI - \transMtx$ is invertible.\footnote{$1 - \lambda$ is an eigenvalues of $\mI - \transMtx$ iff $\lambda$ is an eigenvalue of $\transMtx$.}

\subsubsection{Speed-ups of the Allsum Algorithm}
The introduced algorithm for computing the allsum in a WFSA can, therefore, be implemented as a matrix inverse.
This means that its runtime is $\bigo\left(\nstates^3\right)$, which can be relatively expensive.
Fortunately, faster algorithms exist for WFSAs with more structure (in their transition functions)---for example, the allsum can be computed in time \emph{linear} in the number of transitions if the automaton is acyclic using a variant of the Viterbi algorithm \citep{eisner-2016-inside}.
Furthermore, if the automaton ``decomposes'' into many smaller strongly connected components (i.e., subgraphs that are cyclic), but the components are connected sparsely and form an acyclic graph of components, the allsum can also be computed more efficiently using a combination of the algorithms described above and the algorithm for acyclic WFSA, resulting in a possibly large speedup over the original algorithm.

Importantly, the allsum algorithm and all the speed-ups are \emph{differentiable}, meaning that they can be used during the \emph{gradient-based training} (cf. \cref{sec:numerical-optimization}) of a finite-state language model, where the weights are parametrized using some learnable parameters---we will return to this point shortly.

\subsubsection{Locally Normalizing a Globally Normalized Finite-state Language Model}
As shown in \cref{thm:locally-normalizing-lms}, any language model (and thus, any globally-normalized model with a normalizable energy function) can also be locally normalized.
In the case of finite-state language models, we can actually explicitly construct the WFSA representing the locally normalized variant using a procedure that is conceptually similar to the allsum algorithm described here.
In contrast to the procedure we presented here, however, the local normalization algorithm computes the pathsums of the paths stemming from every possible state $\stateq$ \emph{individually} and then ``reweights'' the \emph{transitions} depending on the pathsums of their target states $\stater$.
You can think of this as computing the contributions to the entire allsum from $\stateq$ made by all the individual outgoing transitions from $\stateq$ and then normalizing those contributions.
This is an instance of the more general \defn{weight pushing}\index{weight pushing} algorithm.\footnote{See \citet{Mohri2008} for a more thorough discussion of weight pushing.}
This can be summarized by the following theorem:
\begin{theorem}{PFSAs and WFSAs are equally expressive}{locally-normalizing-wfsas}
    Normalizable weighted finite-state automata with non-negative weights and tight probabilistic finite-state automata are equally expressive.
\end{theorem}
In the proof of this theorem, we will make use of the following lemma.
\begin{lemma}{}{allsum-state-decomposition}
    Let $\wfsa = \wfsatuple$ and $\stateq \in \states$.
    Then
    \begin{equation}
        \allsum\left(\wfsa, \stateq\right) = \sum_{\edge{\stateq}{\syma}{w}{\stateq'}\in \trans_{\wfsa_L}} \transitionWeight\left(\edge{\stateq}{\syma}{\cdot}{\stateq'}\right) \allsum(\wfsa, \stateq') + \finalf\left(\stateq\right)
    \end{equation}
\end{lemma}
\begin{proof}
    You are asked to show this in \cref{exercise:state-allsum-decomposition}.
\end{proof}

We can now prove \cref{thm:locally-normalizing-wfsas}
\begin{proof}

    To prove the theorem, we have to show that any WFSA can be written as a PFSA and vice versa.\footnote{By ``written as'', we mean that the weighted language is the same.}

    $\Leftarrow$
    Since any tight probabilistic FSA is simply a WFSA with $\allsum\left(\wfsa\right) = 1$, this holds trivially.

    $\Rightarrow$
    Local normalization is a general property of automata resulting from weight pushing.
    Here, we describe the construction in the special case of working with real-valued weights.
    See \citet{Mohri2008} for a general treatment.

    Let $\wfsa = \wfsatuple$ be a normalizable WFSA with non-negative weights.
    We now show that, for any WFSA, there exists a PFSA encoding the same language model.
    Let $\wfsa_G = \wfsatuple$ be a trim WFSA that encodes a distribution over $\kleene{\alphabet}$ using \cref{eq:wfsa-str-prob}.
    We now construct a tight probabilistic finite-state automaton $\wfsa_L = \left(\alphabet, \states, \trans_{\wfsa_L}, \initf_{\wfsa_L}, \finalf_{\wfsa_L}\right)$ whose language is identical.
    We define the initial and final weights of the probabilistic FSA as follows.
    \begin{align}
        \initf_{\wfsa_L}\left(\stateq\right)  & \defeq \initf\left(\stateq\right) \frac{\allsum\left(\wfsa, \stateq\right)}{\allsum\left(\wfsa\right)} \\
        \finalf_{\wfsa_L}\left(\stateq\right) & \defeq \frac{\finalf\left(\stateq\right)}{\allsum\left(\wfsa, \stateq\right)}
    \end{align}
    We define the transitions of the probabilistic FSA as follows.
    \begin{equation}
        \transitionWeight_{\wfsa_L}\left(\edge{\stateq}{\syma}{\cdot}{\stateq'}\right) \defeq \frac{\transitionWeight\left(\edge{\stateq}{\syma}{\cdot}{\stateq'}\right) \allsum(\wfsa, \stateq')}{\allsum(\wfsa, \stateq)}
    \end{equation}
    This means that $\wfsa_L$ contains the same transitions as $\wfsa$, they are simply reweighted.
    Note that the assumption that $\wfsa$ is trimmed means that all the quantities in the denominators are non-zero.

    It is easy to see that the weights defined this way are non-negative due to the non-negativity of $\wfsa$'s weights.
    Furthermore, the weights of all outgoing arcs from any $\stateq \in \states$ and its final weight sum to $1$:
    \begin{align}
         & \sum_{\edge{\stateq}{\syma}{w}{\stateq'}\in \trans_{\wfsa_L}} w + \finalf_{\wfsa_L}\left(\stateq\right)                                                                                                                                                                &                                                       \\
         & = \sum_{\edge{\stateq}{\syma}{w}{\stateq'}\in \trans_{\wfsa_L}} \frac{\transitionWeight\left(\edge{\stateq}{\syma}{\cdot}{\stateq'}\right) \allsum(\wfsa, \stateq')}{\allsum(\wfsa, \stateq)} + \frac{\finalf\left(\stateq\right)}{\allsum\left(\wfsa, \stateq\right)} & \justification{definition of $\trans_{\wfsa_L}$}      \\
         & = \frac{1}{\allsum\left(\wfsa, \stateq\right)} \left(\sum_{\edge{\stateq}{\syma}{w}{\stateq'}\in \trans_{\wfsa_L}} \transitionWeight\left(\edge{\stateq}{\syma}{\cdot}{\stateq'}\right) \allsum(\wfsa, \stateq') + \finalf\left(\stateq\right) \right)                 &                                                       \\
         & = 1                                                                                                                                                                                                                                                                    & \justification{\cref{lem:allsum-state-decomposition}}
    \end{align}
    It is also easy to see that the initial weights form a probability distribution over the states of the constructed automaton.
    \begin{align}
        \sum_{\stateq \in \states} \initf_{\wfsa_L}\left(\stateq\right) & = \sum_{\stateq \in \states} \initf\left(\stateq\right) \frac{\allsum\left(\wfsa, \stateq\right)}{\allsum\left(\wfsa\right)}   \\
                                                                        & = \frac{1}{\allsum\left(\wfsa\right)} \sum_{\stateq \in \states} \initf\left(\stateq\right) \allsum\left(\wfsa, \stateq\right) \\
                                                                        & = \frac{1}{\allsum\left(\wfsa\right)} \allsum\left(\wfsa\right) = 1
    \end{align}

    We now have to show that the probabilities assigned by these two automata match.
    We will do that by showing that the probabilities assigned to individual \emph{paths} match, implying that stringsums match as well.
    The probability of a path is defined analogously to a probability of a string, i.e., $\pdens_\wfsa\left(\apath\right) = \frac{\weight\left(\apath\right)}{\allsum\left(\wfsa\right)}$ (where $\allsum\left(\wfsa\right) = 1$ for tight probabilistic FSAs).
    Let then $\apath = \left(\edge{\stateq_1}{\syma_1}{w_1}{\stateq_2}, \ldots, \edge{\stateq_{\pathlen - 1}}{\syma_{\pathlen - 1}}{w_{\pathlen - 1}}{\stateq_\pathlen}\right) \in \paths\left(\wfsa\right) = \paths\left(\wfsa_L\right)$.
    Then, by the definitions of $\transitionWeight_{\wfsa_L}$, $\initf_{\wfsa_L}$, and $\finalf_{\wfsa_L}$
    \begin{align} \label{eq:wfsa-local-norm-derivation}
        \pdens_{\wfsa_L}\left(\apath\right) & = \initf_{\wfsa_L}\left(\stateq_1\right) \left(\prod_{\idx = 1}^{\pathlen - 1} w_\idx\right) \finalf_{\wfsa_L}\left(\stateq_\pathlen\right)                                                                                                                                                                                                                                                     \\
                                            & = \initf\left(\stateq_1\right) \frac{\allsum\left(\wfsa, \stateq_1\right)}{\allsum\left(\wfsa\right)} \prod_{\idx = 1}^{\pathlen - 1} \frac{\transitionWeight\left(\edge{\stateq_{\idx}}{\syma}{\cdot}{\stateq_{\idx + 1}}\right) \allsum(\wfsa, \stateq_{\idx + 1})}{\allsum(\wfsa, \stateq_{\idx})} \frac{\finalf\left(\stateq_\pathlen\right)}{\allsum\left(\wfsa, \stateq_\pathlen\right)}.
    \end{align}
    Notice that the state-specific allsums of all the inner states of the path (all states apart from $\stateq_1$ and $\stateq_\pathlen$) \emph{cancel out} as the product moves over the transitions of the path.
    Additionally, the terms $\allsum\left(\wfsa, \stateq_1\right)$ and $\allsum\left(\wfsa, \stateq_\pathlen\right)$ cancel out with the definitions of $\initf_{\wfsa_L}$ and $\finalf_{\wfsa_L}$.
    This leaves us with
    \begin{equation}
        \pdens_{\wfsa_L}\left(\apath\right) = \initf\left(\stateq_1\right) \frac{1}{\allsum(\wfsa)} \prod_{\idx = 1}^{\pathlen - 1} \transitionWeight\left(\edge{\stateq_{\idx}}{\syma}{\cdot}{\stateq_{\idx + 1}}\right) \finalf\left(\stateq_\pathlen\right) = \pdens_{\wfsa}\left(\apath\right),
    \end{equation}
    finishing the proof.
\end{proof}
While \cref{thm:locally-normalizing-lms} shows that any language model can be locally normalized, \cref{thm:locally-normalizing-wfsas} shows that in the context of finite-state language models, the locally normalized version of a globally-normalized model is \emph{also} a finite-state model.

\subsubsection{Defining a Parametrized Globally Normalized Language Model} \label{sec:parametrized-wfsa-lms}
Having learned how an arbitrary normalizable finite-state language model can be normalized, we now discuss how models in this framework can be \emph{parametrized} to enable fitting them to some training data.
Crucial for parameterizing a globally normalized model is a \defn{score function}\index{score function} $\func^\trans_\params: \states \times \alphabet \times \states \to \R$, which parametrizes the transitions between the states and thus determines the weights of the (accepting) paths.
Additionally, we also parameterized the initial and final functions $\func^\initf_\params$ and $\func^\finalf_\params$.\ryan{This definition relates to the chat that Anej and I had in Zulip. We need to map it back to section 2.4 \response{Anej} How about this?}
These parametrized functions then define the automaton $\wfsa_\params \defeq \left(\alphabet, \states, \trans_\params, \initf_\params, \finalf_\params\right)$, where $\trans_\params \defeq \left\{\edge{\stateq_1}{\sym}{\func^\trans_\params\left(\stateq_1, \sym, \stateq_2\right)}{\stateq_2}\right\}$, $\initf_\params\left(\qinit\right) \defeq \func^\initf_\params\left(\qinit\right)$, and $\finalf_\params\left(\qfinal\right) \defeq \func^\finalf_\params\left(\qfinal\right)$.
Note that we can parametrize the function $\func_\params$ in any way we want; for example, the function could be a neural network using distributed representations (we will see a similar example at the end of this section), or it could simply be a lookup table of weights.
The fact that the function $\func_\params: \left(\stateq_1, \sym, \stateq_2 \right)$ can only ``look at'' the identities of the states and the symbol might seem limiting; however, the states alone can encode a lot of information: for example, in \ngram{} models we describe below, they will encode the information about the previous $\ngr - 1$ symbols and the transitions will then encode the probabilities of transitioning between such sequences of symbols.

The globally parametrized model then simply takes in any string $\str \in \kleene{\alphabet}$ and computes its stringsum value under the parametrized automaton, which in turn, as per \cref{eq:wfsa-lm}, defines probabilities of the strings.
The quantity $\allsum\left(\wfsa_\params\right)$ can be computed with the allsum algorithm discussed in \cref{sec:normalizing-finite-state-lms}.
Importantly, since the algorithms for computing the string probabilities are differentiable, the model defined this way can also be \emph{trained} with gradient-based learning as described in \cref{sec:numerical-optimization}.

You might notice that this formulation does not exactly match the formulation of globally normalized models from \cref{def:globally-normalized-model}---the function $\wfsa: \kleene{\alphabet} \to \R$ does not exactly match the form of an energy function as its values are not exponentiated as in \cref{eq:globally-normalized-formulation}.
However, we tie this back to the definition of globally normalized models by defining an actual energy function as a simple \emph{transformation} of the stringsum given by $\wfsa_\params$.
We can define the globally normalizing energy function $\unnormalizedpGN^{\wfsa_\params}$ as
\begin{equation}
    \unnormalizedpGN^{\wfsa_\params}\left(\str\right) \defeq -\log\left(\wfsa\left(\str\right)\right),
\end{equation}
which can be easily seen to, after exponentiating it as in \cref{eq:globally-normalized-formulation}, result in the same expression as \cref{eq:wfsa-lm}.
With this, we have formulated finite-state language models as general globally normalized models.

Having introduced WFSAs as a formal and abstract computational model which can define a set of weighted strings, we now show how it can be used to explicitly model a particularly simple family of languages.
We arrive at this family of language models when we impose a specific assumption on the set of \emph{conditional distributions} of the language models that ensures that they are finite-state: the \ngram{} assumption.

\subsection{Tightness of Finite-state Models} \label{sec:tight-pfsas}
Any normalizable globally normalized finite-state language model is tight by definition because the sum of the scores over all \emph{finite} strings is finite, and since they are normalized, they sum to $1$.
We, therefore, focus on \emph{locally} normalized finite-state models and provide necessary and sufficient conditions for their tightness.
Locally normalized finite-state models are exactly probabilistic WFSAs (\cref{def:stochastic-wfsa}).
Luckily, the tightness of probabilistic WFSAs can be easily characterized, as the following theorem shows.
\begin{theorem}{A sufficient condition for tightness of finite-state language models}{sfslm-tight}
    A probabilistic FSA is tight if and only if all accessible states are also co-accessible.
\end{theorem}
\begin{proof}
    We prove each direction in turn.
    \paragraph{($\Rightarrow$):}  Assume the WFSA is tight. Let $\stateq\in \states$ be an accessible state, which means  $\stateq$ can be reached after a finite number of steps with positive probability.
    By tightness assumption, then there must be a positive probability path from $\stateq$ to termination, or else the WFSA will not be able to terminate after reaching $\stateq$, resulting in non-tightness.
    This means that $\stateq$ is also co-accessible.
    So, assuming that the WFSA is tight, every accessible state is also co-accessible.

    \paragraph{($\Leftarrow$):}
    Assume that all accessible states are co-accessible.
    First, one may consider a Markov chain consisting only of the set of accessible states $\states_A\subseteq \states$, since all other states will have probability 0 at every step. Recall a fundamental result in finite-state Markov chain theory which states that, if there exists a unique absorbing state which is reachable from every state, then the Markov process is absorbed by this state with probability 1 (see, e.g., Theorem 11.3 in \citealp{grinstead1997}). We already have that
    \begin{itemize}
        \item \eos{} is an absorbing state, and that
        \item by assumption, every state in $\states_A$ is co-accessible which implies that they can reach \eos.
    \end{itemize}
    Hence, it remains to show that \eos{} is the \emph{unique} absorbing state.
    Suppose there is another state (or group of states) in $\states_A$ distinct from \eos{} that is absorbing, i.e., cannot leave once entered.
    Then, these states cannot reach \eos{} by assumption, which means they are not co-accessible, contradicting the assumption that every state in $\states_A$ is co-accessible.
    Hence, \eos{} is the only absorbing state in $\states_A$ and by the property of an absorbing Markov chain, the process is absorbed by \eos{} with probability 1. In other words, the WFSA is tight.\looseness=-1
\end{proof}

Notice that trimming a PFSA results in a model that satisfies $\finalf\left(\stateq\right) + \sum_{\edge{\stateq}{\syma}{w}{\stateq'}} w \leq 1$, but might no longer achieve equality as required by \cref{def:stochastic-wfsa}.
We call such models substochastic WFSAs.
\begin{definition}{Substochastic Weighted Finite-State Automaton}{substochastic-wfsa}
    A WFSA $\wfsa = \wfsatuple$ is \defn{substochastic}\index{weighted finite-state automaton!substochastic} if for all $\stateq \in \states$ and all outgoing transitions $\edge{\stateq}{\syma}{w}{\stateq'} \in \trans$ it holds that
    \begin{align}
        \initf\left(\stateq\right)  & \geq 0 \\
        \finalf\left(\stateq\right) & \geq 0 \\
        w                           & \geq 0
    \end{align}
    and
    \begin{equation}
        \finalf\left(\stateq\right) + \sum_{\edge{\stateq}{\syma}{w}{\stateq'}} w \leq 1.
    \end{equation}
\end{definition}

We can then express the termination probability of a WFSA in simple linear algebra terms.
\begin{theorem}{A sufficient condition for the tightness of a sub-stochastic WFSA}{sub-fslm-tight}
    Let $\transMtx'$ be the transition sum matrix of a trimmed substochastic WFSA.
    Then $\mI-\transMtx'$ is invertible and $\pdens\left(\rvx \in \kleene{\alphabet}\right)=\initfVect'^\top (\mI-\transMtx')^{-1} \finalfVect' \leq 1$.
\end{theorem}

In the following, we will make use of the spectral radius of a matrix.
\begin{definition}{Spectral radius}{}
    The \defn{spectral radius}\index{spectral radius} of a matrix $\mM \in \C^{\mtxDim \times \mtxDim}$ with eigenvalues $\lambda_1, \ldots, \lambda_\mtxDim$ is defined as
    \begin{equation}
        \spectRad\left(\mM\right) \defeq \max\left\{|\lambda_1|, \ldots, |\lambda_\mtxDim|\right\}.
    \end{equation}
\end{definition}

To prove \cref{thm:sub-fslm-tight}, we will make use of the following useful lemma.

\begin{lemma}{}{sub-fslm-eigen}
    Let $\transMtx'$ be the transition sum matrix of a trimmed substochastic WFSA, then $\spectRad(\transMtx')<1$.
\end{lemma}
To begin with, we wish to apply the following result which connects the row sums of a matrix to its spectral radius. Below, $\sM_\mtxDim$ denotes the set of $\mtxDim \times \mtxDim$ matrices, and $\norm{\mA}_\infty=\max_{1\leq \idxn \leq \mtxDim} \sum_{\idxi=1}^\mtxDim |\mA_{\idxn \idxi}|$ denotes the infinity matrix norm.
\begin{proposition}{\S6.2.P8; \citealp{horn2013}}{hj-628}
    For any $\mA\in \sM_\mtxDim$, $\spectRad(\mA)\leq \norm{\mA}_\infty$. Additionally, if $\mA$ is irreducible and not all absolute row sums of $\mA$ are equal, then $\spectRad(\mA)<\norm{\mA}_\infty$.
\end{proposition}
However, the transition sum matrix $\mP$ of a substochastic WFSA may be reducible whereas the irreducibility condition in \cref{prop:hj-628} cannot be dropped.
Hence, we need to ``decompose'' $\transMtx'$ in a way to recover irreducibility.
We use the \emph{Frobenius normal form} (also known as \emph{irreducible normal form}) to achieve this.
\begin{proposition}{\S8.3.P8; \citealp{horn2013}}{hj838}
    Let $\mA\in \sM_\mtxDim$ be non-negative. Then, either $\mA$ is irreducible or there exists a permutation matrix $\mP$ such that
    \begin{align}
        \mP^\top \mA \mP =
        \begin{bmatrix}
            \mA_1      &        & \ast  \\
                       & \ddots &       \\
            \mathbf{0} &        & \mA_K
        \end{bmatrix}
    \end{align}
    is block upper triangular, and each diagonal block is irreducible (possibly a 1-by-1 zero matrix). This is called an \defn{Frobenius normal form}\index{Frobenius normal form} (or \defn{irreducible normal form}) of $A$. Additionally, $\Lambda(A)=\Lambda(\mA_1)\cup\dotsm\cup\Lambda(\mA_K)$ where $\Lambda(\cdot)$ denotes the set of eigenvalues of a matrix.
\end{proposition}\ryan{Does $\lambda$ return a set? I am super confused by this issue. It needs to change. \response{Anej} I changed it to $\Lambda$.}
We now proceed to the proof of \cref{lem:sub-fslm-eigen}.
\begin{proof}
    Notice that, by way of a similarity transformation via a permutation matrix, the Frobenius normal form is equivalent to a relabeling of the states in the trimmed WFSA in the sense of
    \begin{subequations}
        \begin{align}
            (\mP^\top\initfVect')^\top (\mP^\top \transMtx' \mP)^K (\mP^\top\finalfVect')
             & = (\initfVect'^\top \mP) (\mP^\top \transMtx'^K \mP)(\mP^\top \finalfVect') \\
             & = \initfVect'^\top \transMtx'^K \finalfVect'
        \end{align}
    \end{subequations}
    where the equalities follow from the fact that the inverse of a permutation matrix $\mP$ is its transpose.
    Hence, with an appropriate relabeling, we may assume without loss of generality that $\mP$ is already put into a Frobenius normal form
    \begin{align}
        \transMtx' = \begin{bmatrix}
                         \transMtx'_1 &        & \ast         \\
                                      & \ddots &              \\
                         \mathbf{0}   &        & \transMtx'_K
                     \end{bmatrix}
    \end{align}
    where each $\transMtx'_k$ is irreducible.

    Since the transition sum matrix $\transMtx'$ of a trimmed substochastic WFSA is a substochastic matrix, each $\transMtx'_k$ is also substochastic.
    In fact, each $\transMtx'_k$ is \emph{strictly} substochastic, meaning that there is at least a row that sums to less than 1.
    To see this, suppose to the contrary that there is a probabilistic $\transMtx'_k$.
    Since the WFSA is trimmed, every state is both accessible and co-accessible.
    Being accessible implies that there is a positive probability of reaching every state in $\transMtx'_k$.
    However, the probabilisticity of $\transMtx'_k$ forces the corresponding $\finalfVect'$ entries to be 0.
    Hence, none of these states can transition to \eos, meaning that they're not co-accessible, contradicting the assumption.
    Hence, every $\transMtx'_k$ is strictly substochastic and has at least one strictly less than 1 row sum.
    Then, either all row sums of $\transMtx'_k$ are less than 1 or some row sums are 1 and some are less than 1.
    In either cases, \cref{prop:hj-628} implies that $\spectRad(\transMtx'_k)<1$ for all $1\leq k\leq K$.
    Finally, as \cref{prop:hj838} entails, $\spectRad(\transMtx')=\max\{\spectRad(\transMtx'_1),\dots,\spectRad(\transMtx'_K)\}$ where each $\spectRad(\transMtx'_k)<1$.
    Hence, $\spectRad(\transMtx')<1$.
\end{proof}

We now use the stated results to finally prove \cref{thm:sub-fslm-tight}.
\begin{proof}
    By \cref{lem:sub-fslm-eigen}, $\spectRad(\transMtx')<1$, in which case $\mI-\transMtx'$ is invertible and the Neumann series $\mI+\transMtx'+\transMtx'^2+\dotsm$ converges to $(\mI-\transMtx')^{-1}$ \citep[\S5.6,][]{horn2013}. Hence, we can write $(\mI - \transMtx')^{-1}=\sum_{k = 0}^\infty \transMtx'^k$. Then,
    \begin{subequations}
        \begin{align}
            \pdens(\alphabet^\ast) & = \sum_{k = 0}^\infty P(\alphabet^k)                                        \\
                                   & = \sum_{k = 0}^\infty \initfVect'^\top \transMtx'^k \finalfVect'            \\
                                   & = \initfVect'^\top \left(\sum_{k=0}^\infty \transMtx'^k\right) \finalfVect' \\
                                   & = \initfVect'^\top (\mI - \transMtx')^{-1} \finalfVect'.
        \end{align}
    \end{subequations}
\end{proof}


\subsection{The \textit{n}-gram Assumption and Subregularity} \label{sec:n-gram-models}
We now turn our attention to one of the first historically significant language modeling frameworks: \ngram{} models.
While they are often taught completely separately from (weighted) finite-state automata, we will see shortly that they are simply a special case of finite-state language models and thus all results for the more general finite-state language models also apply to the specific \ngram{} models as well.

As we saw in \cref{thm:locally-normalizing-lms}, we can factorize the language model $\pLM$ for $\str = \sym_1 \ldots \sym_\strlen \in \kleene{\alphabet}$ as
\begin{equation} \label{eq:locally-normalized-lm}
    \pLM\left(\str\right) = \pLN\left(\str\right) = \pLNSM(\eos \mid \str) \prod_{\tstep=1}^{\strlen} \pLNSM(\sym_\tstep \mid \str_{<\tstep}),
\end{equation}
where $\pLNSM(\sym \mid \str)$ are specified by a locally normalized model (\cref{def:locally-normalized-model}).

Recall that \SMAcronym{}s specify individual conditional distributions of the next symbol $\sym_\tstep$ given the previous $\tstep - 1$ symbols for \emph{all possible} $\tstep$.
However, as $\tstep$ grows and the history of seen tokens accumulates, the space of possible histories (sequences of strings to condition on) grows very large (and indeed infinite as $\tstep \rightarrow \infty$).
This makes the task of modeling individual conditional distributions for large $\tstep$ computationally infeasible.
One way to make the task more manageable is by using the \ngram{} assumption.
\begin{assumption}{\ngram{} assumption}{ngram} \label{def:ngram}
    In words, \defn{\ngram{} assumption} states that the conditional probability of the symbol $\sym_\tstep$ given $\strlt$ only depends on $\ngr-1$ previous symbols $\str^{\tstep - 1}_{\tstep - \ngr + 1} \defeq \sym_{\tstep-1},\ldots,\sym_{\tstep-\ngr+1}$:\looseness=-1
    \begin{equation} \label{eq:n-gram-assumption}
        \pLNSM\left(\sym_\tstep\mid \str_{<\tstep}\right) = \pLNSM\left(\sym_\tstep \mid \str^{\tstep - 1}_{\tstep - \ngr + 1}\right).
    \end{equation}
\noindent We will refer to $\str^{\tstep - 1}_{\tstep - \ngr + 1}$ as the \defn{history} of $\sym_\tstep$.
    The sequence $\sym_{\tstep-1} \cdots \sym_{\tstep-\ngr+1})$ is often called the \defn{history} or the \defn{context}.
\end{assumption}

\begin{figure}
    \centering
    \begin{tikzpicture}
        \node[align=center] at (0, 0.6) {The \quad quick \quad brown \quad fox \quad jumps \quad over \ldots};

        \draw[draw=none, rounded corners, fill=DarkBlue!15] (-4, -0.3) rectangle (0.9, 0.3);
        \draw[draw=none, rounded corners, fill=DarkBlue!30] (-0.5 + 0.45, -0.3) rectangle (0.9, 0.3);
        \draw[draw=none, rounded corners, fill=AccentBlue!15] (-
        3.1, -0.3-0.61) rectangle (2.3, 0.3-0.61);
        \draw[draw=none, rounded corners, fill=AccentBlue!30] (0.9, -0.3-0.61) rectangle (2.3, 0.3-0.61);
        \draw[draw=none, rounded corners, fill=LightBlue!15] (-1.6, -0.3-0.61-0.61) rectangle (3.5, 0.3-0.61-0.61);
        \draw[draw=none, rounded corners, fill=LightBlue!30] (2.3, -0.3-0.61-0.61) rectangle (3.5, 0.3-0.61-0.61);

        \node[align=center] at (-2, 0) {$\pLM\left(\text{\small fox} \mid \text{\small The quick brown}\right)$};
        \node[align=center] at (-1.05, -0.61) {$\pLM\left(\text{\small jumps} \mid \text{\small quick brown fox}\right)$};
        \node[align=center] at (0.35, -0.61-0.61) {$\pLM\left(\text{\small over} \mid \text{\small brown fox jumps}\right)$};
        \node[align=center] at (0.45, 0) {$\cdot$};
        \node[align=center] at (1.6, -0.61) {$\cdot$};
        \node[align=center] at (2.9, -0.61-0.61) {$\cdot$};
    \end{tikzpicture}
    \caption{An illustration of how an $4$-gram LM computes the probability of a string.
        All conditional probabilities can be computed in parallel and then multiplied into the probability of the entire string.}
    \label{fig:example-ngram}
\end{figure}
In plain English, this means that the probability of a token only depends on the previous $\ngr-1$ tokens.
\ngram{} assumption is, therefore, an alias of $(\ngr-1)^\text{th}$ order Markov assumption in the language modeling context.

\paragraph{Handling edge cases by padding.}
Given our definition in \cref{eq:n-gram-assumption} where the conditional probability $\pLNSM\left(\sym_\tstep \mid \str_{\tstep - \ngr - 1: \tstep - 1}\right)$ depends on exactly $\ngr - 1$ previous symbols, we could run into an issue with negative indices for $\tstep < \ngr$.
To handle edge cases for $\tstep < \ngr$, we will \defn{pad}\index{padding} the sequences with the $\bos$ symbols at the beginning, that is, we will assume that the sequences $\eossym_1 \ldots \eossym_\tstep$ for $\tstep < \ngr - 1$ are ``transformed'' as
\begin{equation}
    \eossym_1 \eossym_2 \ldots \eossym_\tstep \mapsto \underbrace{\bos \ldots \bos}_{\ngr - 1 - \tstep \text{ times}} \eossym_{1} \eossym_{2} \ldots \eossym_{\tstep}
\end{equation}
Notice that with such a transformation, we always end up with strings of length $\ngr - 1$, which is exactly what we need for conditioning in an \ngram{} model.
In the following, we will assume that all such sequences are already transformed, but at the same time, we will assume that
\begin{equation}
    \pLNSM\left(\eossym \mid \underbrace{\bos \ldots \bos}_{\ngr - 1 - \tstep \text{ times}} \eossym_{1} \eossym_{2} \ldots \eossym_{\tstep} \right) = \eossym_0 \eossym_{1} \eossym_{2} \ldots \eossym_{\tstep}
\end{equation}

By definition, \ngram{} language models can only model dependencies spanning $\ngr$ tokens or less.
By limiting the length of the relevant context when determining $\pLNSM\left(\sym_\tstep\mid\str_{<\tstep}\right)$ to the previous $\ngr$ tokens, the \ngram{} assumption limits the number of possible probability \emph{distributions} that need to be tracked to $\bigo\left(|\alphabet|^{\ngr - 1}\right)$.

Despite their simplicity, \ngram LMs have a storied place in language modeling \citep{6773024,1162650,10.5555/907280,1454428,4767370,10.5555/108235.108270,NIPS2000_728f206c,10.5555/944919.944966,Bengio2006,SCHWENK2007492,heafield-2011-kenlm,heafield-etal-2013-scalable}.
Because the conditional probabilities of \ngram LMs only depend on the previous $\ngr - 1$ symbols, different parts of the string can be processed independently, i.e., in parallel.
This facilitates a natural connection to transformer LMs since parallelizability is a prevalent feature of the architecture and one of its main advantages over other neural LMs such as RNN LMs \citep{Vaswani2017}.\looseness=-1

A particularly simple case of the \ngram{} model is the \defn{bigram}\index{bigram model} model where $\ngr = 2$, which means that the probability of the next word only depends on the previous one, i.e., \ryan{I would use $y$, $y'$ here.} $\pLNSM \left( \sym_{\tstep} \mid \str_{<\tstep} \right) = \pLNSM \left( \sym_{\tstep} \mid \sym_{\tstep - 1} \right)$.\footnote{What would the uni-gram ($\ngr = 1$) model look like? What conditional dependencies between words in a sentence could be captured by it?}

\begin{example}{A simple bigram model}{}
    Let us look at a specific example of a simple bigram model.
    Suppose our vocabulary consists of the words \textexample{large}, \textexample{language}, and \textexample{models}, thus, $|\alphabet|=3$.
    To specify the bigram model, we have to define the conditional probabilities $\pM \left( \sym_{\idxj} \mid \sym_{\idxi} \right)$ for $\sym_{\idxi} \in \alphabet \cup \set{\bos, \eos}$ and $\sym_{\idxj} \in \alphabet \cup \set{\eos}$ (remember that we do not have to model the probability of the next token being $\bos$).
    In the case of bigrams, we can represent those in a table, where the entry at position $\idxi,\idxj$ represents the probability $\pM \left( \sym_{\idxj} \mid \sym_{\idxi} \right)$:

    {
    \centering
    \[\begin{array}{c|cccc}
                                   & \textexample{large} & \textexample{language} & \textexample{models} & \textexample{$\eos$} \\
            \hline
            \text{$\bos$}          & 0.4                 & 0.2                    & 0.2                  & 0.2                  \\
            \textexample{large}    & 0.1                 & 0.4                    & 0.2                  & 0.3                  \\
            \textexample{language} & 0.1                 & 0.1                    & 0.4                  & 0.4                  \\
            \textexample{models}   & 0.2                 & 0.2                    & 0.1                  & 0.5                  \\
            \textexample{$\eos$}   & 0.4                 & 0.2                    & 0.2                  & 0.2                  \\
        \end{array}\]
    }

    Under our model, the probability of the sentence ``large language models'' would be
    \begin{align*}
                 & \pLNSM \left( \textexample{large} \mid \bos \right)                    \\
        \cdot \, & \pLNSM \left( \textexample{language} \mid \textexample{large} \right)  \\
        \cdot \, & \pLNSM \left( \textexample{models} \mid \textexample{language} \right) \\
        \cdot \, & \pLNSM \left( \eos \mid \textexample{models} \right)                   \\
        = \,     & 0.4 \cdot 0.4 \cdot 0.4 \cdot 0.5 = 0.032
    \end{align*}
    while the probability of the sentence ``large large large" would be
    \begin{align*}
                 & \pLNSM \left( \textexample{large} \mid \bos \right)                \\
        \cdot \, & \pLNSM \left( \textexample{large} \mid \textexample{large} \right) \\
        \cdot \, & \pLNSM \left( \textexample{large} \mid \textexample{large} \right) \\
        \cdot \, & \pLNSM \left( \eos \mid \textexample{large} \right)                \\
        = \,     & 0.4 \cdot 0.1 \cdot 0.1 \cdot 0.3 = 0.0012.
    \end{align*}
    Note that the probabilities in the above table are made up and not completely reasonable.
    A real \ngram{} model would not allow for probabilities of exactly $0$ to avoid pathological behavior.
\end{example}

\subsubsection{Representing \ngram{} Models as WFSAs} \label{sec:ngram-wfsa}
We define \ngram{} language models as models that only consider a finite amount of context when defining the conditional probabilities of the next token.
This means that the set of possible conditional distributions $\pLNSM\left(\sym\mid \str\right)$ is also finite which very naturally connects them to weighted finite-state automata---indeed, every \ngram{} language model is a WFSA---specifically, a probabilistic finite-state automaton (or a substochastic one).
We will make this connection more formal in this subsection, thus formally showing that \ngram{} models are indeed finite-state.
Note that this is different from \cref{sec:parametrized-wfsa-lms}, where we discussed how to parametrize a general WFSA and use it as a globally normalized model---in contrast, in this section, we consider how to fit a (locally normalized) \ngram{} model into the finite-state framework.

The intuition behind the connection is simple: the finite length of the context implies a finite number of histories we have to model.
These histories represent the different states the corresponding automaton can reside in at any point.
Given any history $\str$ with $|\str| < \ngr$ and the state $\stateq \in \states$ representing $\str$, then, the conditional distribution of the next token given $\str$ dictate the transition weights into the next states in the WFSA, representing the new, updated history of the input.


Importantly, since we want PFSAs to represent globally-normalized models, we will also remove the $\eos$ symbol from the \ngram{} model before transforming it into a PFSA---as the remark above about the relationship between the $\eos$ symbol and the final states hints, the latter will fill in the role of the $\eos$ symbol.
The way we do that is the following.
From the semantics of the $\eos$ symbol discussed in the section on tightness (cf. \cref{eq:rv}), we also know that to model the probability distribution over finite strings in $\kleene{\alphabet}$, we only require to keep track of strings up to the first occurrence of the $\eos$ symbol.
Therefore, when converting a given \ngram{} model to a WFSA, we will only model sequences up to the first occurrence of the special symbol, meaning that $\eos$ will never occur in the \emph{context} of any conditional distribution $\pLNSM\left(\eossym\mid\str\right)$.
We now detail this construction.

Let $\pLN$ be a well-defined \ngram{} language model specified by conditional distributions $\pLNSM$ as defined by \cref{def:ngram}.
We will now construct a WFSA representing $\pLN$.
Intuitively, its states will represent all possible sequences of words of length $\ngr$ while the transitions between the states $\stateq_1$ and $\stateq_2$ will correspond to the possible transitions between the \ngram{}s which those represent.
This means that the only possible (positively weighted) transitions will be between the \ngram{}s which can follow each other, i.e. $\str_{\tstep-\ngr: \tstep-1}$ and $\str_{\tstep-\ngr+2: \tstep}$ for some $\sym_{\tstep - \ngr}, \sym_{\tstep} \in \eosalphabet$ (until the first occurrence of $\eos$).
The transition's weight will depend on the probability of observing the ``new" word $\sym_{0}$ in the second \ngram{} given the starting \ngram{} $\sym_{-\ngr} \sym_{-(\ngr - 1)} \dots \sym_{-1}$.
Further, the final weights of the states will correspond to \emph{ending} the string in them.
In $\pLN$, this is modeled as the probability of observing $\eos$ given the context $\str_{\tstep-\ngr: \tstep-1}$---this, therefore, is set as the final weight of the state representing the history $\str_{\tstep-\ngr: \tstep-1}$.
Formally, we can map a \ngram{} model into a WFSA $\automaton = \left(\alphabet_{\wfsa}, \states_{\wfsa}, \trans_{\wfsa}, \initf_{\wfsa}, \finalf_{\wfsa}\right)$ by constructing $\wfsa$ as follows.
\begin{itemize}
    \item Automaton's alphabet:
          \begin{equation}
              \alphabet_{\wfsa} \defeq \alphabet
          \end{equation}
    \item The set of states:
          \begin{equation}
              \states_{\wfsa} \defeq \bigcup_{\tstep = 0}^{\ngr - 1} \{\bos\}^{\ngr-1-\tstep} \times \alphabet^{\tstep}
          \end{equation}
    \item The transitions set
          \begin{align}
              \trans_{\wfsa} \defeq \{ & \edge{\str_{\tstep-\ngr: \tstep-1}}{\sym_\tstep}{\pLNSM \left( \sym_\tstep \mid \str_{\tstep-\ngr: \tstep-1} \right)}{\str_{\tstep-\ngr + 1: \tstep}} \mid \\ \nonumber & \str_{\tstep - \ngr + 1: \tstep - 1} \in \bigcup_{\tstep = 0}^{\ngr-2}{\{\bos\}^{\ngr-2-\tstep} \times \alphabet^{\tstep}; \sym_{\tstep - \ngr}, \sym_{\tstep} \in \alphabet } \}
          \end{align}
    \item The initial function:
          \begin{equation}
              \initf_{\wfsa}: \str \mapsto \begin{cases} \one & \text{ if } \str = \underbrace{\bos \dots \bos}_{\ngr -1 \text{ times}} \\ \zero & \text{ otherwise} \end{cases}
          \end{equation}
    \item The final function
          \begin{equation}
              \finalf_{\wfsa}: \str \mapsto \pLNSM\left(\eos \mid \str \right), \ \str \in \states_{\wfsa}
          \end{equation}
\end{itemize}
The definition of the states set $\states_{\wfsa}$ captures exactly the notion of padding with the $\bos$ symbol for handling the edge cases we described above.
This shows that \ngram{} language models are indeed finite-state (we leave the formal proof showing that $\lang\left(\automaton\right) = \lang\left(\pLN\right)$ to the reader.

\paragraph{Defining a \ngram{} language model through a parametrized WFSA.}
We now consider how we can use the framework of WFSA to define a more ``flexible'' \emph{parametrized} \emph{globally} normalized model.
In this case, we do not start from an existing locally normalized set of distributions forming $\pLNSM$.
Rather, we would like to model the ``suitability'' of different \ngram{}s following each other---that is, we would like to somehow \emph{parametrize} the probability that some \ngram{} $\str'$ will follow an \ngram{} $\str$ without having to worry about normalizing the model at every step.
This will allow us to then fit the probability distributions of the model to those in the data, e.g., with techniques described in \cref{sec:param-estimation}.
Luckily, the flexibility of the WFSA modeling framework allows us to do exactly that.

\subsubsection{Subregularity} \label{sec:subregular}
We saw that language models implementing the very natural \ngram{} assumption can be represented using weighted finite-state automata.
However, \ngram{} models do not ``need the full expressive power'' of WFSAs---they can actually be modeled using even simpler machines than finite-state automata.
This, along with several other examples of simple families of formal languages, motivates the definition of \emph{subregular} languages.
\begin{definition}{Subregular language}{}
    A language is \defn{subregular}\index{subregular language} if it can be recognized by a finite-state automaton or any weaker machine.
\end{definition}
Most subregular languages can indeed be recognized by formalisms which are much simpler than FSAs.
Many useful and interesting classes of subregular languages have been identified---recently, especially in the field of phonology.
Naturally, due to their simpler structure, they also allow for more efficient algorithms---this is why we always strive to represent a language with the simplest formalism that still captures it adequately.
See \citet{Jager2012,Avcu2017SubregularCA} for comprehensive overviews of subregular languages.

Subregular languages actually form multiple hierarchies of complexity within regular languages.
Interestingly, \ngram{} models fall into the simplest level of complexity in one of the hierarchies, directly above \emph{finite} languages.
This class of subregular languages is characterized by patterns that depend solely on the blocks of symbols that occur \emph{consecutively} in the string, which each of the blocks considered independently of the others---it is easy to see that \ngram{} models intuitively fall within such languages.
This family of subregular languages is suggestively called \emph{strictly local languages}.
\begin{definition}{Strictly local languages}{}
    A language $\lang$ is \defn{strictly $\ngr$-local} \index{strictly $\ngr$-local}  ($\nStrictLocal$) if, for every string $\str$ of length $|\str| = \ngr - 1$, and all strings $\strx_1, \strx_2, \strz_1, \strz_2 \in \kleene{\alphabet}$, it holds that if $\strx_1 \str \strz_1 \in \lang$ and $\strx_2 \str \strz_2 \in \lang$, then also $\strx_1 \str \strz_2 \in \lang$ (and $\strx_2 \str \strz_1 \in \lang$).

    A language is \defn{strictly local}\index{strictly local} ($\strictLocal$) if it is strictly $\ngr$-local for any $\ngr$.
\end{definition}
Note that we could of course also define this over with the $\eos$-augmented alphabet $\eosalphabet$.
You can very intuitively think of this definition as postulating that the history more than $\ngr$ symbols back does not matter anymore for determining or specifying whether a string is in a language (or its weight, in the weighted case)---this is exactly what the \ngram{} assumption states.

\subsection{Representation-based \textit{n}-gram Models} \label{sec:bengio2003-model}
\ryan{Can we call this ``Representation-based $n$-gram models and tie it back to 3.1? \response{ANej} I think it works better now.}
So far, we have mostly talked about the conditional probabilities and the WFSA weights defining a language model very abstractly.
Apart from describing how one can generally parametrize the weights of the underlying WFSA with the scoring function in \cref{sec:ngram-wfsa}, we only discussed what values the weights can take for the language model to be well-defined and what implications that has on the distribution defined by the WFSA.
In this section, we consider for the first time what an actual implementation of a finite-state, or more precisely, a \ngram{} language model might look like.
Concretely, we will define our first parameterized language model in our General language modeling framework (cf. \cref{sec:general-framework}) by defining a particular form of the encoding function $\enc$ as a simple multi-layer feed-forward neural network.\anej{fix reference}\footnote{While we introduce particular architectures of neural networks, for example, recurrent neural networks and transformers later in \cref{chapter:classical-lms}, we assume some familiarity with neural networks in general. See Chapter 6 of \citet{Goodfellow-et-al-2016} for an introduction.}

However, before we dive into that, let us consider as an alternative possibly the simplest way to define a (locally normalized) \ngram{} language model: by directly parametrizing the probabilities of each of the symbols $\sym$ in the distribution $\pLNSM\left(\eossym\mid \eosstr\right)$ for any context $\eosstr$, that is
\begin{equation}\label{eq:naive-ngram}
    \params \defeq \left\{\param_{\sym\mid \str} \defeq \pLNSM\left(\sym\mid \str\right) \mid \sym \in \eosalphabet, \str \in \eosalphabet^{\ngr - 1}, \param_{\sym\mid \str} \geq 0, \sum_{\sym' \in \eosalphabet}\param_{\sym\mid \str} = 1\right\}.
\end{equation}
The following proposition shows that the maximum likelihood solution (\cref{eq:optmle}) to this parametrization is what you would probably expect.
\begin{proposition}{}{ngram-mle}
    The MLE solution
    of \cref{eq:naive-ngram} is
    \begin{equation}
        \pLNSM\left(\sym_\ngr\mid \str_{<\ngr}\right) = \frac{\strCount\left(\sym_1,\ldots,\sym_\ngr\right)}{\strCount\left(\sym_1,\ldots,\sym_{\ngr-1}\right)}
    \end{equation}
    whenever the denominator $>0$, where $\strCount\left(\sym_1,\ldots,\sym_\ngr\right)$ denotes the number of occurrences of all possible strings of the form $\sym_1,\ldots,\sym_\ngr$ and $\strCount\left(\sym_1,\ldots,\sym_\ngr\right)$ denotes the number of occurrences of all possible strings of the form $\sym_1,\ldots,\sym_{\ngr-1}$.
\end{proposition}
\begin{proof}
    Let $\dataset \defeq \left\{\str^{\left(1\right)}, \ldots, \str^{\left(M\right)}\right\}$ be the training dataset.
    The log-likelihood of a single example $\str^{\left(\idxm\right)}$ is
    \begin{align}
        \log\left(\pLN\left(\str\right)\right) &
        = \log\left(\prod_{\tstep = 1}^{|\str^{\left(\idxm\right)}|} \pLNSM\left(\sym_{\tstep}^{\left(\idxm\right)}\mid \sym_{\tstep-\ngr: \tstep - 1}^{\left(\idxm\right)} \right)\right)                             \\
                                               & = \sum_{\tstep = 1}^{|\str^{\left(\idxm\right)}|} \log \pLNSM\left(\sym_{\tstep}^{\left(\idxm\right)}\mid \sym_{\tstep-\ngr: \tstep - 1}^{\left(\idxm\right)} \right)
    \end{align}
    which means that the log-likelihood of the entire dataset is
    \begin{align}
        \loglikelihood\left(\dataset\right)
         & = \sum_{\idxm = 1}^{M} \sum_{\tstep = 1}^{|\str^{\left(\idxm\right)}|} \log \pLNSM\left(\sym_{\tstep}^{\left(\idxm\right)}\mid \sym_{\tstep-\ngr: \tstep - 1}^{\left(\idxm\right)} \right) \\
         & = \sum_{\idxm = 1}^{M} \sum_{\tstep = 1}^{|\str^{\left(\idxm\right)}|} \log \param_{\sym_\ngr \mid \str{< \ngr}}.
    \end{align}
    \cref{exercise:ngram-mle} asks you to show that this can be rewritten with the \defn{token to type switch}\index{token to type switch} as
    \begin{equation}
        \loglikelihood\left(\dataset\right) = \sum_{\substack{\str\\|\str| = \ngr}} \strCount\left(\str\right) \param_{\sym_\ngr \mid \str{< \ngr}}.
    \end{equation}

    The maximum likelihood parameters can then be determined using Karush--Kuhn--Tucker (KKT) conditions\footnote{See \url{https://en.wikipedia.org/wiki/Karush\%E2\%80\%93Kuhn\%E2\%80\%93Tucker_conditions}.} to take into account the non-negativity and local normalization constraints:
    \begin{equation} \label{eq:kkt-ngram}
        \nabla_{\params} \left(\loglikelihood\left(\dataset\right) - \sum_{\substack{\str \in \kleene{\alphabet} \\ |\str| = \ngr - 1}} \evlambda_{\str \sym} \left(\sum_{\sym \in \alphabet} \param_{\sym\mid\str} - 1\right) - \sum_{\substack{\str \in \kleene{\alphabet} \\ |\str| = \ngr - 1}} \eveta_{\str \sym} \param_{\sym\mid\str}\right) = 0.
    \end{equation}
    Recall that the KKT conditions state that a $\params$ is an optimal solution of $\loglikelihood$ if and only if $\left(\params, \left\{\evlambda_{\str \sym'}\right\}_{\str \in \alphabet^{\ngr - 1}, \sym \in \alphabet}, \left\{\eveta_{\str \sym'}\right\}_{\str \in \alphabet^{\ngr - 1}, \sym \in \alphabet}\right)$ satisfy \cref{eq:kkt-ngram}.
    Since this is simply a sum over the dataset with no interactions of parameters for individual contexts $\str$ with $|\str| = \ngr - 1$ in $\param_{\sym\mid\str}$, it can be solved for each context $\str$ individually.

    Moreover, as you are asked to show \cref{exercise:ngram-count-sum}, it holds that
    \begin{equation}
        \sum_{\sym' \in \alphabet}\strCount\left(\sym_1 \ldots \sym_{\ngr - 1} \sym'\right) = \strCount\left(\sym_1 \ldots \sym_{\ngr - 1}\right)
    \end{equation}
    for any $\str = \sym_1 \ldots \sym_{\ngr - 1} \in \alphabet^{\ngr - 1}$.
    This leaves us with the following system for each $\str \in \alphabet^{\ngr - 1}$:
    \begin{equation}
        \sum_{\sym' \in \alphabet} \strCount\left(\str \str'\right) \log\param_{\sym'\mid\str} - \evlambda_{\str} \left(\sum_{\str'\in \alphabet}\param_{\sym' \mid \str} - 1\right) - \sum_{\sym' \in \alphabet}\eveta_{\str \sym'} \param_{\sym'\mid\str}.
    \end{equation}
    It is easy to confirm that $\param_{\sym\mid\str} = \frac{\strCount\left(\str \sym\right)}{\strCount\left(\str\right)}$ with $\evlambda_\str = \strCount\left(\str\right)$ and $\eveta_{\str \sym'} = 0$ is a saddle point of \cref{eq:kkt-ngram}.
    This means that $\param_{\sym\mid\str} = \frac{\strCount\left(\str \sym\right)}{\strCount\left(\str\right)}$ is indeed the maximum likelihood solution.
\end{proof}
This results in a locally normalized \ngram{} model.
To avoid issues with division-by-zero and assigning 0 probability to unseen sentences, we can employ methods such as \emph{smoothing} and \emph{backoff}, which are beyond the scope of the course.\footnote{See \citep{chen-goodman-1996-empirical} and Chapter 4 in \citep{jurafsky_martin_2009}.}

While this model might seem like an obvious choice, it comes with numerous drawbacks.
To see what can go wrong, consider the following example.
\begin{example}{\ngram{} model}{n-gram-independent-words}
    Suppose we have a large training corpus of sentences, among which sentences like {\fontfamily{ppl}\selectfont ``We are going to the shelter to adopt a dog.''}, {\fontfamily{ppl}\selectfont ``We are going to the shelter to adopt a puppy.''}, and {\fontfamily{ppl}\selectfont ``We are going to the shelter to adopt a kitten.''}, however, without the sentence {\fontfamily{ppl}\selectfont ``We are going to the shelter to adopt a cat.''}
    Fitting an \ngram{} model using the count statistics and \emph{individual} tables of conditional probabilities $\pLNSM\left(\sym\mid\str\right)$, we would assign the probability $$\pLNSM\left(\sym_\tstep = \text{cat}\mid\str_{< \tstep} = \text{\fontfamily{ppl}\selectfont We are going to the shelter to adopt a}\right)$$ the value $0$ (or some ``default'' probability if we are using smoothing).
    However, the words ``dog'', ``puppy'', ``kitten'', and ``cat'' are semantically very similar---they all describe pets often found in shelters.
    It would therefore be safe to assume that the word ``cat'' is similarly probable given the context {\fontfamily{ppl}\selectfont ``We are going to the shelter to adopt a''} as the other three words observed in the training dataset.
    However, if we estimate all the conditional probabilities \emph{independently}, we have no way of using this information---the words have no relationship in the alphabet, they are simply different indices in a lookup table.
    Additionally, statistics gathered for the sentences above will not help us much when encountering very similar sentences, such as {\fontfamily{ppl}\selectfont ``We went to a nearby shelter and adopted a kitten.''}
    The issue is that there are simply many ways of expressing similar intentions.
    We would thus like our language models to be able to generalize across different surface forms and make use of more ``semantic'' content of the sentences and words.
    However, if the model is parametrized as defined in \cref{eq:naive-ngram}, it is not able to take advantage of any such relationships.
\end{example}

The model defined by \cref{eq:naive-ngram} is therefore unable to take into account the relationships and similarities between words.
The general modeling framework defined in \cref{sec:general-framework} allows us to remedy this using the \defn{distributed word representations}\index{distributed word representations}.
Recall that, in that framework, we associate each word $\sym$ with its vector representation $\embedding{\sym}$ (its \emph{embedding}), and we combine those into the embedding matrix $\embedMtx$.
Importantly, word embeddings are simply additional parameters of the model and can be \emph{fit} on the training dataset \emph{together} with the language modeling objective.
One of the first successful applications of $\enc(\cdot)$ is due to \citet{Bengio2003}, which we discuss next.

To be able to use the embeddings in our general framework, we now just have to define the concrete form of the context-encoding function $\enc$.
In the case of the neural \ngram{} model which we consider here and as defined by \citep{Bengio2003}, the representations of the context $\str_{<\tstep}$, $\enc\left(\str_{<\tstep}\right)$, are defined as the output of a neural network which looks at the previous $\ngr - 1$ words in the context:
\begin{equation}\label{eq:bengio-2003-model}
    \enc(\str_{< \tstep}) \defeq \enc(\sym_{\tstep-1}, \sym_{\tstep-2}, \ldots, \sym_{\tstep-\ngr+1}),
\end{equation}
where $\enc$ is a neural network we define in more detail shortly.
The full language model is therefore defined through the conditional distributions \ryan{This is also sloppy. The notation we should use is as follows. Please makes sure this is consistent with 3.1. Softmax is a vector-to-vector transformation.
    Therefore, the equation should read as
    \begin{equation}
        \pLNSM\left(\str\right) = \prod_{\tstep = 1}^{\strlen + 1} \softmax\left(\enc\left(\sym_\tstep, \sym_{\tstep-1}, \ldots, \sym_{\tstep-\ngr+1}\right)^\top \embedding{\sym_\tstep} + \biasStr \right)_{y_t}
    \end{equation}
    However, surely the left-hand side of the below is wrong so I updated it.

    Furthermore, I don't see why we need MLP at all at this point. We should just use enc.
    \response{Anej} I removed the MLP formulation.
    \response{Anej} I think its smoother now.
}
\begin{equation}
    \pLNSM\left(\eossym_\tstep\mid \eosstr_{<\tstep}\right) \defeq \softmax\left(\enc\left(\eossym_{\tstep - 1}, \eossym_{\tstep-2}, \ldots, \eossym_{\tstep-\ngr+1}\right)^\top \embedMtx + \biasVector \right)_{\eossym_\tstep}
\end{equation}
resulting in the locally normalized model
\begin{align}
    \pLN\left(\str\right) = & \softmax\left(\enc\left(\eossym_\strlen, \eossym_{\strlen-1}, \ldots, \eossym_{\strlen-\ngr+2}\right)^\top \embedMtx + \biasVector \right)_{\eos}                                   \\
                            & \cdot \prod_{\tstep = 1}^{\strlen} \softmax\left(\enc\left(\sym_{\tstep-1}, \sym_{\tstep-2}, \ldots, \sym_{\tstep-\ngr+1}\right)^\top \embedMtx + \biasVector \right)_{\sym_\tstep}
\end{align}
for $\str \in \kleene{\alphabet}$.

Importantly, notice that although this is a \emph{neural} model, it is nonetheless still an \ngram{} model with finite context---\cref{eq:bengio-2003-model} is simply a restatement of the \ngram{} assumption in terms of the neural encoding function $\enc$.
It therefore still suffers from some of the limitations of regular \ngram{} models, such as the inability to model dependencies spanning more than $\ngr$ words.
However, it solves the problems encountered in \cref{ex:n-gram-independent-words} by considering word similarities and \emph{sharing} parameters across different contexts in the form of an encoding function rather than a lookup table.

While encoding function $\enc$ in \cref{eq:bengio-2003-model} could in principle take any form, the original model defined in \citet{Bengio2003} defines the output as for the string $\str = \sym_\tstep, \sym_{\tstep-1}, \ldots, \sym_{\tstep-n+1}$ as
\begin{equation} \label{eq:bengio-2003-equation}
    \enc\left(\sym_\tstep, \sym_{\tstep-1}, \ldots, \sym_{\tstep-n+1}\right) \defeq \vb + \mW \rvx + \mU \tanh\left(\rvd + \mH \rvx\right),
\end{equation}
where $\rvx \defeq \text{concat}\left(\embedding{\sym_\tstep}, \embedding{\sym_{\tstep-1}}, \ldots, \embedding{\sym_{\tstep-n+1}}\right)$ denotes the concatenation of the context symbol embeddings into a long vector of size $\left(\ngr - 1\right)\cdot \embedDim$, and $\vb$, $\rvd$, $\mW$, and $\mU$ define the parameters of the encoding function.
This completes our definition of the model in the general language modeling framework---the model can then simply be trained on the language modeling objective as defined in \cref{sec:standard-objective}.

We can also see that such a model also reduces the number of parameters required to specify a \ngram{} model: whereas a lookup-table-based \ngram{} model with no parameter sharing requires $\bigo\left(|\alphabet|^\ngr\right)$ parameters to be defined, the number of parameters required by a representation-based \ngram{} model scales \emph{linearly} with $\ngr$---all we have to do is add additional rows to the matrices defined in \cref{eq:bengio-2003-equation}.
We will later see how this can be reduced to a \emph{constant} number of parameters w.r.t. the sequence length in the case of recurrent neural networks in \cref{sec:rnns}.\anej{fix reference}

\begin{figure}
    \centering
    \includegraphics[width=\textwidth]{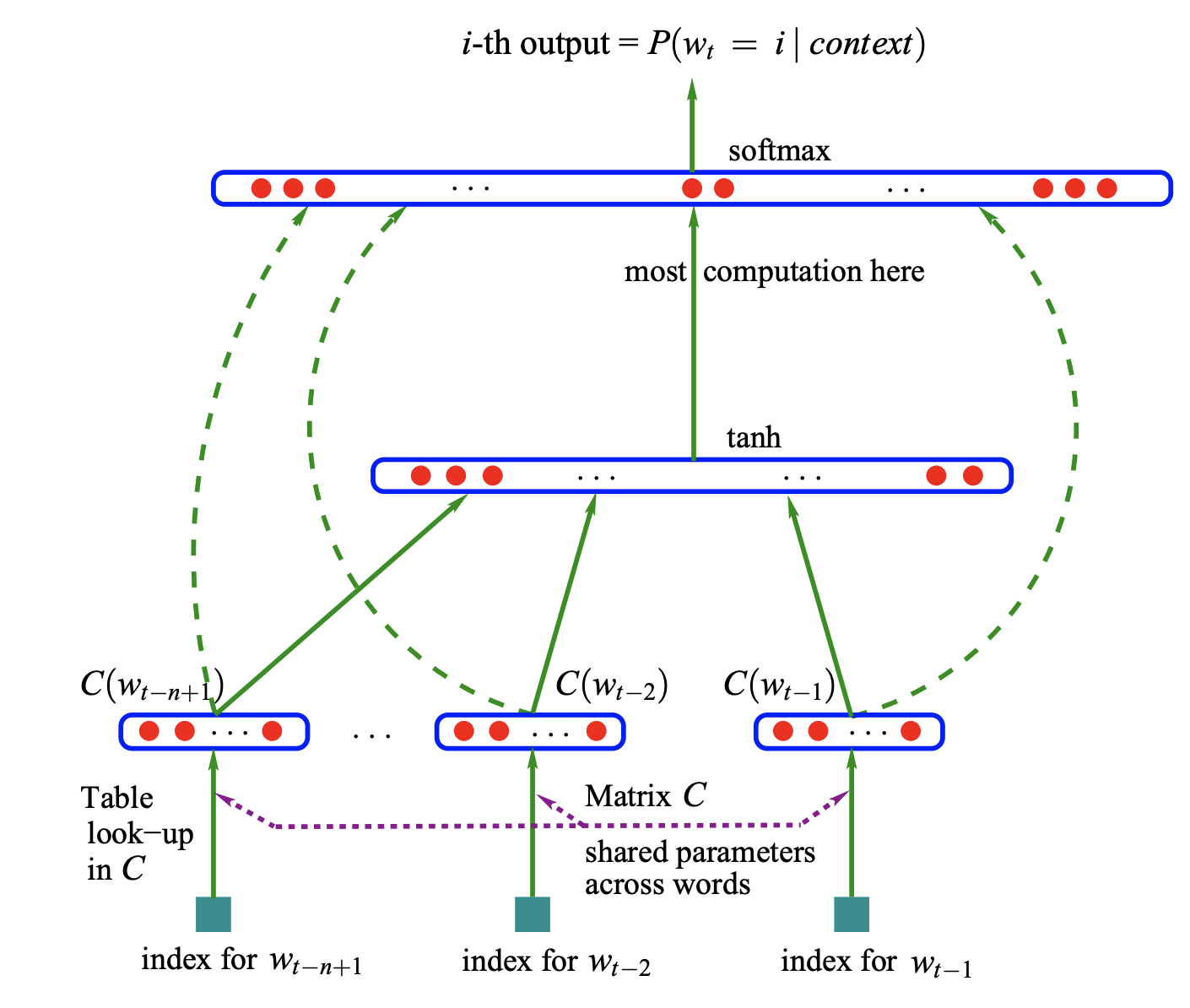}
    \caption{A pictorial depiction of the \ngram{} neural language model from the original publication \citep{Bengio2003}. Note that the quantity $C\left(w\right)$ corresponds to $\embedding{\sym}$ for a word $\sym$ in our notation.}
    \label{fig:bengio-2003-model}
\end{figure}

Pictorially, we can imagine the model as depicted in \cref{fig:bengio-2003-model} (taken from the original publication).
This shows that the \ngram{} modeling framework is not limited to counting co-occurrence statistics.
The model from \cref{eq:bengio-2003-model} can also be represented by a WFSA just like the simpler models we discussed above, with the weights on the transitions parametrized by the neural network.
This allows us to both understand well with insights from formal language theory, as well as to train it in a flexible way allowed for by the non-linear encoding function.
However, the model from \cref{eq:bengio-2003-model} is still limited to statistics of the last $\ngr$ tokens or less.
If we want to model arbitrarily long dependencies and hierarchical structures, we have to leave the space of finite-state languages behind and develop formalisms capable of modeling more complex languages.
The next section explores the first of such frameworks: context-free languages with the computational models designed to model them.
\newpage{}

\section{Pushdown Language Models}
\label{sec:pushdown-lms}
\label{sec:context-free}

An strong limitation of finite-state language models is that they can definitionally only distinguish a finite set of contexts.
However, human language has inherently more structure than what a finite set of contexts can encode.
For example, human language contains arbitrarily deep recursive structures which cannot be captured by a finite set of possible histories---we will see an example of this soon in \cref{sec:human-language-not-finite-stete}.

To be able to model these structures we are climbing a rung higher on the ladder of the hierarchy of formal languages: we are going to consider \defn{context-free languages}\index{context-free language}, a larger class of languages than regular languages.
Luckily, we will see that a lot of the formal machinery we introduce in this section closely follows analogs from the finite-state section and we invite the reader to pay close attention to the parallels.
For example, similarly to how we weighted a string in a regular language by summing over the weights of the paths labeled with that string, we will weight strings in context-free languages by summing over analogous structures.

To be able to recognize context-free languages, will have to extend finite-state automata from \cref{sec:wfsa} with an additional data structure---the stack.
Finite-state automata augmented with a stack are called pushdown automata.
We introduce them in \cref{sec:wpda}.
Before giving a formal treatment of pushdown automata, however, we will discuss an arguably more natural formalism for generating the context-free languages---context-free grammars.\footnote{You might wonder what non-context-free grammars are: a superclass of context-free grammars is that of context-sensitive grammars, in which a production rule may be surrounded by a left and right context.
    They are still however a set of restricted cases of general grammars, which are grammars that can emulate Turing machines.}

In the last part of the section, we will then further extend the regular pushdown automaton with an \emph{additional} stack.
Interestingly, this will make it more powerful: as we will see, it will raise its expressive power from context-free languages to all computable languages, as it is Turing complete.
While this augmentation will not be immediately useful from a language modeling perspective, we will then later use this machine to prove some theoretical properties of other modern language models we consider later in the course.

\subsection{Human Language Is not Finite-state} \label{sec:human-language-not-finite-stete}
As hinted above, human language contains structures that cannot be modeled by finite-state automata.
Before we introduce ways of modeling context-free languages, let us, therefore, first motivate the need for a more expressive formalism by more closely considering a specific phenomenon often found in human language: recursive hierarchical structure.
We discuss it through an example, based on \citet{jurafsky_martin_2009}.\anej{This is centre embedding}
\begin{example}{Center embeddings}{human-language-nonregular}

    Consider the sentence:
    \begin{center}
        \textexample{The cat likes to cuddle.}
    \end{center}
    It simply describes a preference of a cat.
    However, we can also extend it to give additional information about the cat:
    \begin{center}
        \textexample{The cat the dog barked at likes to cuddle.}
    \end{center}
    This sentence, in turn, can be extended to include additional information about the dog:
    \begin{center}
        \textexample{The cat the dog the mouse startled barked at likes to cuddle.}
    \end{center}
    Of course, we can continue on:
    \begin{center}
        \textexample{The cat the dog the mouse the rat frightened startled barked at likes to cuddle.}
    \end{center}
    and on:
    \begin{center}
        {\footnotesize \textexample{The cat the dog the mouse the rat the snake scared frightened startled barked at likes to cuddle.}}
    \end{center}
    In theory, we could continue like this for as long as we wanted---all these sentences are \emph{grammatically correct}---this is an instance of the so-called \defn{center embeddings}.

    Crucially, such sentences cannot be captured by a regular language, i.e., a language based on an automaton with finitely many states.
    While we would need formal machinery beyond the scope of this course to formally prove this, the intuition is quite simple.
    By adding more and more ``levels'' of recursion to the sentences (by introducing more and more animals in the chain), we unboundedly increase the amount of information the model has to ``remember'' about the initial parts of the sentence while processing it sequentially, to be able to process or generate the matching terms on the other end of the sentence correctly.
    Because such hierarchies can be arbitrarily deep (and thus the sentences arbitrarily long), there is no bound on the number of states needed to remember them, which means they cannot be captured by a finite-state automaton.

    Note that this example also touches upon the distinction of the grammatical \emph{competence} versus grammatical \emph{performance} \citep{Chomsky1959,Chomsky1963,Chomsky1965}.
    The former refers to the purely theoretical properties of human language, for example, the fact that such hierarchical structures can be arbitrarily long and still grammatically correct.
    Grammatical performance, on the other hand, studies language grounded more in the way people actually use it.
    For example, nested structures like the one above are never very deep in day-to-day speech---indeed, you probably struggled to understand the last few sentences above.
    We rarely come across nestings of depth more than three in human language \citep{Miller1963,jin-etal-2018-depth,Karlsson2007}.
\end{example}

\subsection{Context-free Grammars}
How can we capture recursive structures like those in \cref{ex:human-language-nonregular} and the long-term dependencies arising from them?
The first formalism modeling such phenomena we will introduce is context-free grammars: a \emph{generative} formalism which can tell us how to generate or ``compose`` strings in the language it describes.
Later in the section (\cref{sec:wpda}), we will introduce the context-free analog of finite-state automata, which will tell us how to \emph{recognize} whether a string is in a context-free language (rather than generate a string): pushdown automata.

\begin{definition}{Context-free Grammar}{}
    A \defn{context-free grammar}\index{context-free grammar} (CFG) is a 4-tuple $\grammar = \cfgtuple$ where $\alphabet$ is an alphabet of terminal symbols, $\nonterm$ is a non-empty set of non-terminal symbols with $\nonterm \cap \alphabet = \emptyset$, $\start\in\nonterm$ is the designated start non-terminal symbol and $\rules$ is the set of production rules, where each rule $\arule \in \rules$ is of the form $\production{\NTX}{\cfgstr}$ with $\NTX\in\nonterm$ and $\cfgstr\in\kleene{(\nonterm\cup \alphabet)}$.\footnote{As is the case for initial states in FSAs, multiple start symbols could be possible. However we consider only one for the sake of simplicity.}\index{context-free grammar!production}\index{context-free grammar!terminal}\index{context-free grammar!non-terminal}\index{context-free grammar!start-symbol}
\end{definition}
\begin{example}{A simple context-free grammar}{simple-cfg}
    Let $\grammar = \cfgtuple$ be defined as follows:
    \begin{itemize}
        \item $\alphabet = \set{\syma, \symb}$
        \item $\nonterm = \set{\NTX}$
        \item $\start = \NTX$
        \item $\rules = \set{\production{\NTX}{\syma \NTX \symb}, \production{\NTX}{\eps}}$
    \end{itemize}
    This defines a simple context-free grammar.
    We will return to it later, when we will formally show that it generates the language $\lang = \set{\syma^n \symb^n \mid n \in \Nzero}$.
\end{example}

\subsubsection{Rule Applications and Derivations}
Context-free grammars allow us to generate strings $\str\in\kleene{\alphabet}$ by \emph{applying} production rules on its non-terminals.
We apply a production rule $\production{\NTX}{\cfgstr}$ to $\NTX\in\nonterm$ in a rule $\arule$ by taking $\NTX$ on the right-hand side of $\arule$ and replacing it with $\cfgstr$.\footnote{We say that $\NTX$ is on the right-hand side of a rule $\arule$ if $\arule$ takes the form $\arule=(\production{\NTY}{\cfgstr \, \NTX \vgamma})$, where $\cfgstr,\vgamma\in\kleene{(\nonterm\cup \alphabet)}$.
    We will sometimes refer to $\NTX$ as the \emph{head} of the production rule $\production{\NTX}{\cfgstr}$, and the right-hand side $\cfgstr$ as the \emph{body} of the production rule.}
\begin{definition}{Rule Application}{}
    A production rule $\production{\NTY}{\vbeta}, \vbeta\in\kleene{(\nonterm\cup\alphabet)},$ is \defn{applicable}\index{context-free grammar!applicable production} to $\NTY$ in a rule $\arule$, if $\arule$ takes the form
    \begin{equation*}
        \production{\NTX}{\cfgstr \, \NTY \, \vgamma}, \quad \cfgstr,\vgamma\in\kleene{(\nonterm\cup\alphabet)}.
    \end{equation*}
    The \defn{result of applying} $\production{\NTY}{\vbeta}$ to $\cfgstr \, \NTY \, \vgamma$ is $\cfgstr \, \vbeta \, \vgamma$.
\end{definition}

Starting with $\start$, we apply $\production{\start}{\cfgstr}$ to $\start$ for some $(\production{\start}{\cfgstr})\in\rules$, then take a non-terminal in $\cfgstr$ and apply a new production rule.\footnote{We will write $\NTX\in\cfgstr$, which formally means a substring of $\cfgstr$ with length $1$. Unless otherwise stated, $\NTX$ can be either a non-terminal or a terminal.}
To generate a string we follow this procedure until all non-terminal symbols have been transformed into terminal symbols.
The resulting string, i.e., the \defn{yield}\index{production!yield}, will be the string taken by concatenating all terminal symbols read from left to right.
More formally, a derivation can be defined as follows.
\begin{definition}{Derivation}{}
    A \defn{derivation} \index{context-free grammar!derivation}
    in a grammar $\grammar$ is a sequence $\cfgstr_1,\dots,\cfgstr_M$, where $\cfgstr_1\in\nonterm$, $\cfgstr_2,\dots,\cfgstr_{M-1}\in\kleene{(\nonterm\cup \alphabet)}$ and $\cfgstr_M\in\kleene{\alphabet}$, in which each $\cfgstr_{\idxm+1}$ is formed by applying a production rule in $\rules$ to $\cfgstr_{\idxm}$.

\end{definition}

We say that $\cfgstr\in\kleene{(\nonterm\cup \alphabet)}$ is derived from $\NTX\in\nonterm$ if we can apply a finite sequence of production rules to generate $\cfgstr$ starting from $\NTX$.
We will denote this as $\NTX \derives_\grammar \cfgstr$.
See the following formal definition.
\begin{definition}{Derives}{}
    Let $\grammar \defeq \cfgtuple$ be a CFG.
    We say that $\NTX$ \defn{derives}\index{context-free grammar!derive} $\vbeta$ under the grammar $\grammar$, denoted as $\NTX \derivesbase_{\scaleto{\grammar}{5pt}} \vbeta$ if $\exists \arule \in \rules$ such that $\arule=(\production{\NTX}{\cfgstr \, \vbeta \, \vgamma})$, $\cfgstr,\vgamma\in\kleene{(\nonterm\cup\alphabet)}$ and $\vbeta\in\kleene{(\nonterm\cup\alphabet)}\setminus\{\eps\}$.
    The special case $\NTX \derivesbase_{\scaleto{\grammar}{5pt}} \eps$ holds iff $\production{\NTX}{\eps} \in \rules$.
    We denote the reflexive transitive closure of the $\derivesbase_{\scaleto{\grammar}{5pt}}$ relation as $\derives_{\scaleto{\grammar}{5pt}}$.
    We say that $\vbeta$ is \defn{derived from} $\NTX$ if $\NTX \derives_{\scaleto{\grammar}{5pt}} \vbeta$.
\end{definition}

The (context-free) language of a CFG $\grammar$ is defined as all the strings $\str\in\kleene{\alphabet}$ that can be derived from the start symbol $\start$ of $\grammar$, or alternatively, the set of all yields possible from derivations in $\grammar$ that start with $\start$. We will denote the language generated by $\grammar$ as $\lang(\grammar)$.
\begin{definition}{Language of a Grammar}{}
    The \defn{language}\index{context-free grammar!language} of a context-free grammar $\grammar$ is
    \begin{equation}
        \lang(\grammar)=\{\str\in\kleene{\alphabet}\mid \start \derives_{\scaleto{\grammar}{5pt}} \str \}
    \end{equation}
\end{definition}

\subsubsection{Parse Trees and Derivation Sets}
A natural representation of a derivation in a context-free grammar is a \defn{derivation tree} $\tree$\index{context-free grammar!derivation tree}\index{context-free grammar!parse tree} (also known as a parse tree).
A derivation tree represents the sequence of applied rules in a derivation with a directed tree.
The tree's internal nodes correspond to the non-terminals in the derivation, and each of their children corresponds to a symbol (from $\alphabet \cup \nonterm$) on the right side of the applied production in the derivation.
The leaves, representing terminal symbols, ``spell out'' the derived string---the tree's yield.
More formally, for each production rule $\production{\NTX}{\cfgstr}$, the node corresponding to the specific instance of the non-terminal $\NTX$ in the derivation is connected to the nodes corresponding to $\NTY \in \cfgstr$ where $\NTY \in \alphabet \cup \nonterm$.

We will mostly be interested in representing derivations starting with $\start$---the root node of a tree representing any such derivation will correspond to $\start$.
We will denote the string generated by a tree $\tree$---its yield---by $\yield\left(\tree\right)$.
See \cref{fig:trees_example1} for examples of parse trees for the grammar from \cref{ex:simple-cfg}.
\begin{figure}[t]
    \center
    \Tree [.$\NTX$ $\eps$ ]
    \hspace{10mm}
    \Tree [.$\NTX$ $\syma$ [.$\NTX$ $\eps$ ] $\symb$ ]
    \hspace{10mm}
    \Tree [.$\NTX$ $\syma$ [.$\NTX$ $\syma$ [.$\NTX$ $\eps$ ] $\symb$ ] $\symb$ ]
    \hspace{10mm}
    \Tree [.$\NTX$ $\syma$ [.$\NTX$ $\syma$ [.$\NTX$ $\syma$ [.$\NTX$ $\eps$ ] $\symb$ ] $\symb$ ] $\symb$ ]
    \caption{A sequence of derivation trees for the strings in $\{\syma^n\symb^n \mid n=0,1,2,3\}$ in the grammar from \cref{ex:simple-cfg}.}
    \label{fig:trees_example1}
\end{figure}

\begin{figure}[t]
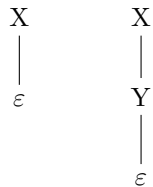

    \center
    \Tree [.$\NTX$ $\eps$ ]
    \hspace{10mm}
    \Tree [.$\NTX$ [.$\NTY$ $\eps$ ] ]
    \caption{Two parse trees in the modified grammar $\grammar$ yielding $\eps$.}
    \label{fig:two-eps-derivations}
\end{figure}

Importantly, a grammar may in fact admit \emph{multiple} derivations and hence multiple derivation trees for any given string.
\begin{example}{Multiple derivation strings}{}
    It is relatively easy to see that in the grammar $\grammar$, each string $\syma^n \symb^n$ is only generated by a single derivation tree---each new pair of symbols $\syma$ and $\symb$ can only be added by applying the rule $\production{\NTX}{\syma \NTX \symb}$ and the string $\syma^n \symb^n$ can only be generated by the application of the rule $\production{\NTX}{\syma \NTX \symb}$ $n$ times and the rule $\production{\NTX}{\eps}$ once in this order.

    However, we can modify $\grammar$ by adding, for instance, a non-terminal $\NTY$ and rules $\production{\NTX}{\NTY}, \production{\NTY}{\eps}$.
    The empty string $\eps$ may then be derived either by $(\production{\NTX}{\eps})$, or $(\production{\NTX}{\NTY}),(\production{\NTY}{\eps})$, corresponding to two separate derivation trees, as shown in \cref{fig:two-eps-derivations}.
    The set of these two trees comprises what we call the derivation set of $\eps$.
\end{example}

We denote a derivation set of a string $\str$, generated by the grammar $\grammar$, as $\derivationset{\grammar}{\str}$.
\begin{definition}{String derivation set}{}
    Let $\str \in \kleene{\alphabet}$.
    Its \defn{derivation set}, denoted by $\derivationset{\grammar}{\str}$ is defined as
    \begin{equation}
        \derivationset{\grammar}{\str} \defeq \set{\tree \mid \yield\left(\tree\right) = \str}.
    \end{equation}
\end{definition}\anej{think about how to constrain trees. maybe we should define the derivation set of a grammar first, like the set of paths in fsas, and then define this...}
We say that a grammar is \defn{unambiguous} if, for every string that can be generated by the grammar, there is only one associated derivation tree.
\begin{definition}{Unambiguity}{}
    A grammar $\grammar$ is \defn{unambiguous} if for all $\str\in\lang(\grammar)$, $|\derivationset{\grammar}{\str}|=1$.
\end{definition}
The converse holds for \defn{ambiguous} grammars.
\begin{definition}{Ambiguity}{}
    A grammar $\grammar$ is \defn{ambiguous}  if $\exists\str\in\lang(\grammar)$ such that $|\derivationset{\grammar}{\str}|>1$.
\end{definition}

The set of all derivation trees in a grammar is its derivation set.
\begin{definition}{Grammar derivation set}{}
    The \defn{derivation set of a grammar}, $\grammarDerivationset{\grammar}$, is the set of all derivations possible under the grammar.
    More formally, it can be defined as the union over the derivation set for the strings in its language,
    \begin{equation}
        \grammarDerivationset{\grammar} \defeq \bigcup_{\str'\in\lang(\grammar)}\derivationset{\grammar}{\str'}
    \end{equation}
\end{definition}
\begin{definition}{Non-terminal derivation set}{}
    The \defn{derivation set of a non-terminal} $\NTY\in\nonterm$ in $\grammar$, denoted $\derivationset{\grammar}{\NTY}$, is defined as the set of derivation subtrees with root node $\NTY$.
\end{definition}
Note that $\grammarDerivationset{\grammar}$ could be defined as $\derivationset{\grammar}{\start}$.
For a terminal symbol $\syma\in\alphabet$, we trivially define the derivation set $\derivationset{\grammar}{\syma}$ to be empty.\footnote{Empty derivation sets for terminal symbols is defined solely for ease of notation later.}

In cases where it is irrelevant to consider the order of the production rules in a derivation tree, we will write $(\production{\NTX}{\cfgstr})\in\tree$ to refer to specific production rules in the tree---viewing trees as multisets (or bags) over the production rules they include.

\begin{figure}[t]
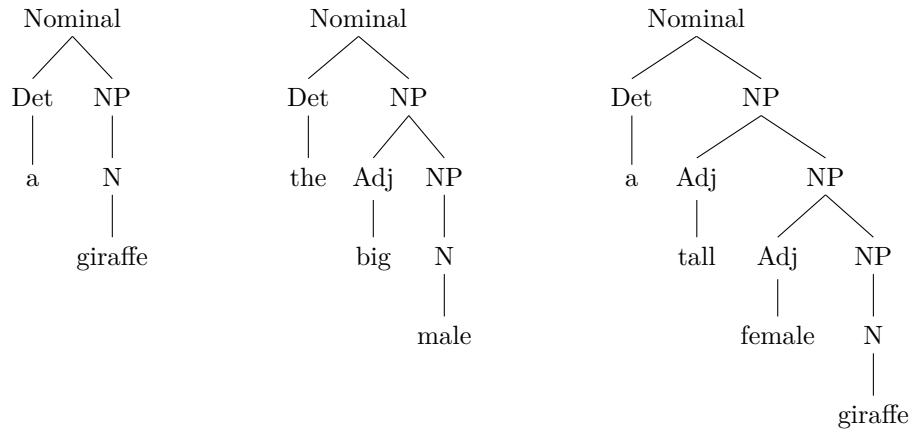

    \center
    \Tree [.$\NT{Nominal}$ [.$\NT{Det}$ a ] [.$\NT{NP}$ [.$\NT{N}$ giraffe ] ] ]
    \hspace{15mm}
    \Tree [.$\NT{Nominal}$ [.$\NT{Det}$ the ] [.$\NT{NP}$ [.$\NT{Adj}$ big ] [.$\NT{NP}$ [.$\NT{N}$ male ] ] ] ]
    \hspace{15mm}
    \Tree [.$\NT{Nominal}$ [.$\NT{Det}$ a ] [.$\NT{NP}$ [.$\NT{Adj}$ tall ] [.$\NT{NP}$ [.$\NT{Adj}$ female ] [.$\NT{NP}$ [.$\NT{N}$ giraffe ] ] ] ] ]
    \caption{Derivation trees for natural language nominal phrases.}
    \label{fig:ex_np_grammar}
\end{figure}

\begin{example}{Nominal Phrases}{cfg}
    CFGs are often used to model natural languages. Terminals would then correspond to words in the natural language, strings would be text sequences and non-terminals would be abstractions over words. As an example, consider a grammar $\grammar$ that can generate a couple of nominal phrases. We let $\nonterm=\{\NT{Adj}, \NT{Det}, \NT{N}, \NT{Nominal}, \NT{NP}\}$, $\alphabet=\{\text{a, big, female, giraffe, male, tall, the}\}$, $\start=\NT{Nominal}$ and define the following production rules:
    \begin{align*}
         & \production{\NT{Nominal}}{\NT{Det}\; \NT{NP}}                                          \\
         & \production{\NT{NP}}{\NT{N} \mid \NT{Adj}\; \NT{NP}}                                   \\
         & \production{\NT{Det}}{\text{a}\mid \text{the}}                                         \\
         & \production{\NT{N}}{\text{female} \mid \text{giraffe} \mid \text{male}}                \\
         & \production{\NT{Adj}}{\text{big} \mid \text{female} \mid \text{male} \mid \text{tall}}
    \end{align*}
    See \cref{fig:ex_np_grammar} for a few examples of derivation trees in this grammar.
    \label{ex:nominal_grammar}
\end{example}

\begin{example}{The generalized Dyck languages $\dyck{\nBracketTypes}$}{dyck}
    A very widely studied family of context-free languages are the Dyck-$\nBracketTypes$ languages, $\dyck{\nBracketTypes}$, the languages of well-nested brackets of $\nBracketTypes$ types.
    They are, in some ways, archetypal context-free languages \citep{Chomsky1963}.
    Formally, we can define them as follows.
    \begin{definition}{$\dyck{\nBracketTypes}$ languages}{}
        Let $\nBracketTypes \in \N$.
        The $\dyck{\nBracketTypes}$ language is the language of the following context-free grammar $\grammar \defeq \cfgtuple$
        \begin{itemize}
            \item $\alphabet \defeq \left\{\openBr{\idx} \mid \idx = 1, \ldots, \nBracketTypes\right\} \cup \left\{\closeBr{\idx} \mid \idx = 1, \ldots, \nBracketTypes\right\}$
            \item $\nonterm \defeq \left\{\start\right\}$
            \item $\start \defeq \start$
            \item $\rules \defeq \left\{\start \to \eps, \start \to \start \start\right\} \cup \left\{ \start \to \openBr{\idx} \start \closeBr{\idx} \mid \idx = 1, \ldots, \nBracketTypes\right\}$
        \end{itemize}
    \end{definition}

    Examples of strings in the language $\dyck{3}$ would be $\openBr{3} \closeBr{3}\openBr{2} \closeBr{2}\openBr{1} \closeBr{1}$, $\openBr{3} \closeBr{3}\openBr{1} \openBr{2} \openBr{2} \closeBr{2} \closeBr{2} \closeBr{1}$, and $\openBr{1} \openBr{2} \openBr{2} \closeBr{2} \closeBr{2} \openBr{3} \openBr{1} \closeBr{1} \closeBr{3}\closeBr{1}$.
    The string $\openBr{2} \openBr{2} \closeBr{1} \closeBr{2}$ is not in the language $\dyck{3}$.
\end{example}

To give you a taste of what formally working with context-free grammars might look like, we now formally show that the grammar from \cref{ex:simple-cfg} really generates the language $\lang = \set{\syma^n\symb^n\mid n \in \Nzero}$, as we claimed.
\begin{example}{Recognizing $\syma^n \symb^n$}{ab_grammar}
    The language $\lang=\{\syma^n \symb^n\mid n\in\mathbb N\}$ is not regular.\footnote{Again, while the intuition behind this is similar to our reasoning from \cref{ex:human-language-nonregular}, this would have to be proven using the so-called pumping lemma for regular languages.}
    However, we can show that it is context-free and recognized exactly by the simple grammar from \cref{ex:simple-cfg}.
    We restate it here for convenience: $\grammar = \cfgtuple$ with $\nonterm = \{\NTX\}$, $\alphabet = \{\syma, \symb\}$, $\start=\NTX$, $\rules=\{\production{\NTX}{\syma \NTX \symb}, \production{\NTX}{\eps}\}$.
    \begin{lemma}{}{}
        Given the grammar $\grammar$ defined above, we have $\lang\left(\grammar\right) = \{\syma^n \symb^n\mid n\in\mathbb N\}$.
    \end{lemma}
    \begin{proof}
        We will show that $\lang=\lang(\grammar)$ in two steps: (i) showing that $\lang\subseteq\lang(\grammar)$ and (ii) showing that $\lang(\grammar)\subseteq\lang$. Define $\str_n=\syma^n\symb^n$.

        (i) We first need to show that each $\str\in\lang$ can be generated by $\grammar$, which we will do by induction.
        \paragraph{Base case ($n=0$)}
        We have that $\str_0=\eps$, which is generated by $\tree=(\production{\NTX}{\eps})$.
        \paragraph{Inductive step ($n>1$)}
        We have that $\str_n$ is generated by $$\tree=\underbrace{(\production{\NTX}{\syma\NTX\symb})\cdots(\production{\NTX}{\syma\NTX\symb})}_{n\text{ times}}(\production{\NTX}{\eps}).$$
        It is then easy to see that $\str_{n+1}$ is generated by the derivation we get by replacing the last rule $(\production{\NTX}{\eps})$ with $(\production{\NTX}{\syma\NTX\symb})(\production{\NTX}{\eps})$---they are exactly the trees illustrated in \cref{fig:trees_example1}.

        (ii) Next, we show that for each $\tree\in\grammarDerivationset{\grammar}$, we have that $\str(\tree)\in\lang$.
        \paragraph{Base case ($\tree=(\production{\NTX}{\eps})$)}
        It is trivial to see that the derivation $\tree=(\production{\NTX}{\eps})$ yields $\str(\tree)=\eps$.

        \paragraph{Inductive step}
        Now observe that $\rules$ only contains two production rules and one non-terminal.
        Starting with $\NTX$, we can either apply $\production{\NTX}{\syma\NTX\symb}$ to get one new non-terminal $\NTX$, or apply $\production{\NTX}{\eps}$ to terminate the process.
        Hence, if we fix the length of the sequence of production rules, there is no ambiguity in which string will be generated.
        Thus, by induction, we conclude that if we have a derivation tree given by $\underbrace{(\production{\NTX}{\syma\NTX\symb}),\dots,(\production{\NTX}{\syma\NTX\symb})}_{n\text{ times}},(\production{\NTX}{\eps})$ generating $\syma^n\symb^n$, the derivation tree given by $\underbrace{(\production{\NTX}{\syma\NTX\symb}),\dots,(\production{\NTX}{\syma\NTX\symb})}_{n+1 \text{ times}},(\production{\NTX}{\eps})$ will generate $\syma^{n+1}\symb^{n+1}$.
    \end{proof}
\end{example}

\subsubsection{Reachable Non-terminals and Pruning}
Similarly to how some states in a WFSA can be useless in the sense that they are not accessible from an initial state or might not lead to a final state, so too can non-terminals in a CFG be useless by not beaing reachable from the start symbol or might not lead to any string of terminals.
In the context of CFGs, we typically use a different terminology: ``reachable'' instead of ``accessible'' and ``generating'' instead of ``co-accessible''.
\begin{definition}{Accessibility for CFGs}{}
    A symbol $\NTX \in \nonterm\cup\alphabet$ is \defn{reachable} (or accessible) if $\exists \cfgstr, \cfgstr' \in \kleene{\left(\nonterm \cup \alphabet\right)}$ such that $\start \derives \cfgstr\NTX\cfgstr'$.
\end{definition}
\begin{definition}{Co-accessibility for CFGs}{}
    A non-terminal $\NTY$ is \defn{generating} (or co-accessible) if $\exists\str\in\kleene{\alphabet}$ such that $\NTY \derives \str$.
\end{definition}

In words, reachable symbols are those that can be derived from the start symbol, whereas generating non-terminals are those from which at least one string (including the empty string) can be derived.
Note that we define reachable for both non-terminals and terminals while generating is only defined for non-terminals.

This allows us to define a pruned context-free grammar, which is the CFG version of a trimmed WFSA.
\begin{definition}{Pruned CFG}{}
    A CFG is \defn{pruned} (or trimmed) if it has no useless non-terminals, i.e. all non-terminals are both reachable and generating.
    \defn{Pruning} (or trimming) refers to the removal of useless non-terminals.
\end{definition}

\subsection{Weighted Context-free Grammars}
\ryan{I would add the example from Booth and Thompson that a PCFG can't recognize all distributions over strings.
    They have a simple one with a Poisson. \response{Anej} This example is at the end of the section.}
As we did with finite-state automata, we will augment the classic, unweighted context-free grammars with real-valued weights.
We do that by associating with each rule $\production{\NTX}{\cfgstr}$ a weight $\productionWeight(\production{\NTX}{\cfgstr})\in\R$.

\begin{definition}{Weighted Context-free Grammar}{}
    A \defn{real-weighted context-free grammar} \index{weighted context-free grammar} is a 5-tuple $\wcfgtuple$ where $\alphabet$ is an alphabet of terminal symbols, $\nonterm$ is a non-empty set of non-terminal symbols with $\nonterm \cap \alphabet = \emptyset$, $\start\in\nonterm$ is the designated start non-terminal symbol, $\rules$ is the set of production rules, and $\productionWeight$ a function $\productionWeight\colon \rules \to \R$, assigning each production rule a real-valued weight.
\end{definition}
For notational brevity, we will denote rules $\arule \in \rules$ as $\arule = \wproduction{\NTX}{\cfgstr}{w}$ for $\NTX\in\nonterm$, $\cfgstr\in\kleene{(\nonterm\cup \alphabet)}$ and $w=\productionWeight(\production{\NTX}{\cfgstr})\in\R$.
\begin{example}{A simple weighted context-free grammar}{simple-wcfg}
    Consider the grammar $\grammar = \wcfgtuple$ defined as follows:
    \begin{itemize}
        \item $\alphabet = \set{\syma, \symb}$
        \item $\nonterm = \set{\NTX}$
        \item $\start = \NTX$
        \item $\rules = \set{\production{\NTX}{\syma \NTX \symb}, \production{\NTX}{\eps}}$
        \item $\productionWeight = \set{\production{\NTX}{\syma \NTX \symb}\mapsto \frac{1}{2}, \production{\NTX}{\eps}\mapsto \frac{1}{2}}$
    \end{itemize}
    This defines a simple weighting of the CFG from \cref{ex:simple-cfg}.
\end{example}

Weights assigned to productions by WFCGs can be arbitrary real numbers.
Analogous to probabilistic WFSAs (\cref{def:stochastic-wfsa}) describing locally normalized finite-state language models, we also define probabilistic WCFGs, where the weights of \emph{applicable production rules} to any non-terminal form a probability distribution.
\begin{definition}{Probabilistic Context-free grammar}{probabilistic-cfg}
    A weighted context-free grammar $\grammar = \wcfgtuple$ is \defn{probabilistic} \index{probabilistic context-free grammar} if the weights of the productions of every non-terminal are non-negative and sum to $1$, i.e., for all $\NTX \in \nonterm$, it holds that
    \begin{equation}
        \forall \ \production{\NTX}{\cfgstr} \in \rules, \ \productionWeight\left(\production{\NTX}{\cfgstr}\right) \geq 0
    \end{equation}
    and
    \begin{equation}
        \sum_{\production{\NTX}{\cfgstr} \in \rules} \productionWeight\left(\production{\NTX}{\cfgstr}\right) = 1
    \end{equation}
\end{definition}
Intuitively, this means that all the production weights are non-negative and that, for any left side of a production rule $\NTX$, the weights over all production rules $\production{\NTX}{\cfgstr}$ sum to $1$.
The grammar from \cref{ex:simple-wcfg} is, therefore, also probabilistic.

Again analogously to the WFSA case, we say that a string $\str$ is in the language of WCFG $\grammar$ if there exists a derivation tree $\tree$ in $\grammar$ containing only non-zero weights with yield $\yield\left(\tree\right)=\str$.

\subsubsection{Tree Weights, String Weights, and Allsums}
In the case of regular languages, we discussed how individual strings are ``produced'' by paths in the automaton (in the sense that each path \emph{yields} a string). \ryan{I don't know what captured means here. \response{Anej} Is it better now?}
As \cref{ex:cfg} showed, the structures that ``produce'' or yield strings in a context-free grammar are \emph{trees}---those, therefore, play an analogous role in context-free grammars to paths in finite-state automata.

Just like we asked ourselves how to combine individual transition weights in a WFSA into weights of entire paths and later how to combine those into weights of strings, we now consider the questions of how to combine the weights of individual production rules into the weight of entire trees and later also individual strings.
We start by giving a definition of the weight of a tree as the product over the weights of all the rules in the tree, i.e., as a \emph{multiplicatively decomposable} function over the weights of its rules.
As you can probably foresee, we will then define the weight of a string as the \emph{sum} over all the trees that yield that string.
\begin{definition}{Weight of a derivation tree}{}
    The \defn{weight of a derivation tree}\index{weighted context-free grammar!derivation tree weight} $\tree\in\grammarDerivationset{\grammar}$ defined by a WCFG $\grammar$ is
    \begin{equation}
        \weight(\tree)=\prod_{(\production{\NTX}{\cfgstr})\in \tree}\productionWeight(\production{\NTX}{\cfgstr}).
    \end{equation}
\end{definition}
The stringsum or the string acceptance weight of a particular string under a grammar is then defined as follows:
\begin{definition}{Stringsum in a context-free grammar}{}
    The \defn{stringsum}\index{weighted context-free grammar!stringsum} $\grammar \left(\str\right)$ of a string $\str$ generated by a WCFG $\grammar$ is defined by
    \begin{align}
        \grammar \left(\str\right) & = \sum_{\tree\in\derivationset{\grammar}{\str}}\weight(\tree)                                                                            \\
                                   & = \sum_{\tree\in\derivationset{\grammar}{\str}}\prod_{(\production{\NTX}{\cfgstr})\in\tree}\productionWeight(\production{\NTX}{\cfgstr})
    \end{align}
\end{definition}

Lastly, analogously to the allsum in WFSAs, an \defn{allsum} is the sum of the weights of all the trees in a WCFG.
We first define the allsum for symbols (non-terminals and terminals).
\begin{definition}{Nonterminal allsum in a context-free grammar}{}
    The \defn{allsum}\index{weighted context-free grammar!non-terminal allsum} for a non-terminal $\NTY$ in a grammar $\grammar$ is defined by
    \begin{align}
        \allsum \left(\grammar, \NTY\right) & = \sum_{\tree\in\derivationset{\grammar}{\NTY}} \weight(\tree)                                                                          \\
                                            & =\sum_{\tree\in\derivationset{\grammar}{\NTY}}\prod_{(\production{\NTX}{\cfgstr})\in\tree}\productionWeight(\production{\NTX}{\cfgstr})
    \end{align}
    The allsum for a terminal $\syma\in\alphabet\cup\{\eps\}$ is defined to be
    \begin{equation}
        \allsum \left(\syma\right) \defeq \one.
    \end{equation}
    \label{def:cfg-allsum}
\end{definition}
The allsum for a grammar is then simply the allsum for its start symbol.
\begin{definition}{Allsum in a context-free grammar}{}
    The \defn{allsum}\index{weighted context-free grammar!allsum} of a weighted context-free grammar $\grammar=\wcfgtuple$ is
    \begin{align}
        \allsum \left(\grammar\right) & =\allsum \left(\grammar, \start\right)                                                                                                    \\
                                      & = \sum_{\tree\in\derivationset{\grammar}{\start}} \weight(\tree)                                                                          \\
                                      & =\sum_{\tree\in\derivationset{\grammar}{\start}}\prod_{(\production{\NTX}{\cfgstr})\in\tree}\productionWeight(\production{\NTX}{\cfgstr})
    \end{align}
\end{definition}

When the grammar $\grammar$ we refer to is clear from context, we will drop the subscript and write e.g. $\allsum(\start)$.

Although we can in some cases compute the allsum of a WCFG in closed form, as we will see in the example below, we generally require some efficient algorithm to be able to do so.

\begin{example}{Geometric Series as an Allsum}{geometric_allsum}
    Consider the WCFG $\grammar=\wcfgtuple$, given by $\nonterm=\{\NTX\}$, $\alphabet=\{\syma\}$, $\start=\NTX$, and the rules:
    \begin{align*}
         & \wproduction{\NTX}{\syma\, \NTX}{1/3} \\
         & \wproduction{\NTX}{\eps}{1}
    \end{align*}
    The language generated by $\grammar$ is $\lang(\grammar)=\{\syma^n\mid n\geq 0\}$. Further note that this grammar is unambiguous -- each string $\str=\syma^m$, for some $m\geq 0$, is associated with the derivation tree given by $\underbrace{(\wproduction{\NTX}{\syma\, \NTX}{1/3}),\dots,(\wproduction{\NTX}{\syma\, \NTX}{1/3})}_{m\text{ times}},(\wproduction{\NTX}{\eps}{1})$. Due to the multiplicative decomposition over the weights of the rules, the weight associated with each derivation tree $\tree$ will hence be
    \begin{equation*}
        \weight(\tree)=\left(\frac{1}{3}\right)^m\times 1 = \left(\frac{1}{3}\right)^m
    \end{equation*}
    Accordingly, we can compute the allsum of $\grammar$ using the closed-form expression for geometric series:
    \begin{equation*}
        \allsum\left(\grammar\right)=\sum_{m=0}^\infty\left(\frac{1}{3}\right)^m = \frac{1}{1-1/3} = \frac{3}{2}
    \end{equation*}
\end{example}

Just like we defined normalizable WFSAs, we also define normalizable WCFSs in terms of their allsum.
\begin{definition}{Normalizable Weighted Context-free Grammar}{}
    A weighted context-free grammar $\grammar$ is \defn{normalizable}\index{weighted context-free grammar!normalizable} if $\allsum \left(\grammar\right)$ is finite, i.e., $\allsum \left(\grammar\right) < \infty$.
\end{definition}

\subsection{Context-free Language Models}

This brings us to the definition of context-free language models.
\begin{definition}{Context-free language model}{}
    A language model $\pLM$ is \defn{context-free}\index{language model!context-free} if its weighted language equals the language of some weighted context-free grammar, i.e., if there exists a weighted context-free grammar $\grammar$ such that $\lang\left(\grammar\right) = \lang\left(\pLM\right)$.
\end{definition}

Going the other way---defining string probabilities given a weighted context-free grammar---there are again two established ways of defining the probability of a string in its language.
\subsubsection{String Probabilities in a Probabilistic Context-free Grammar}
In a probabilistic CFG (cf. \cref{def:probabilistic-cfg}), any production from a non-terminal $\NTX \in \nonterm$ is associated with a probability.
As the probabilities of continuing a derivation (and, therefore, a derivation tree) depend solely on the individual terminals (this is the core of \emph{context-free} grammars!), it is intuitive to see those probabilities as conditional probabilities of the new symbols given the output generated so far.
One can, therefore, define the probability of a path as the product of these individual ``conditional'' probabilities.
\begin{definition}{Tree probability in a PCFG}{}
    We call the weight of a tree $\tree \in \grammarDerivationset{\grammar}$ in a probabilistic CFG the \defn{probability} of the tree $\tree$.
\end{definition}
This alone is not enough to define the probability of any particular string $\str \in \kleene{\alphabet}$ since there might be multiple derivations of $\str$.
Naturally, we define the probability of $\str$ as the sum of the individual trees that generate it:
\begin{definition}{String probability in a PCFG}{}
    We call the stringsum of a string $\str \in \kleene{\alphabet}$ in a probabilistic CFG $\grammar$ the \defn{probability} of the string $\str$:
    \begin{equation}
        \pdens_\grammar\left(\str\right) \defeq \grammar\left(\str\right).
    \end{equation}
\end{definition}
These definitions and their affordances mirror the ones in probabilistic finite-state automata (cf. \cref{sec:pfsa-str-prob}): they again do not require any \emph{normalization} and are therefore attractive as the summation over all possible strings is avoided.
Again, the question of \emph{tightness} of such models comes up: we explore it question in \cref{sec:tight-cfgs}.

\subsubsection{String Probabilities in a General Weighted Context-free Grammar}
To define string probabilities in a general weighted CFG, we use the introduced notions of the stringsum and the allsum---we \emph{normalize} the stringsum to define the globally normalized probability of a string $\str$ as the \emph{proportion} of the total weight assigned to all strings that is assigned to $\str$.
\begin{definition}{String probability in a WCFG}{}
    Let $\grammar = \wcfgtuple$ be a normalizable WCFG with non-negative weights.
    We define the probability of a string $\str \in \kleene{\alphabet}$ under $\grammar$ as
    \begin{equation} \label{eq:wcfg-str-prob}
        \pdens_\grammar\left(\str\right) \defeq \frac{\grammar\left(\str\right)}{\allsum\left(\grammar\right)}.
    \end{equation}
\end{definition}

\subsubsection{Language Models Induced by a Weighted Context-free Grammar}
With the notions of string probabilities in both probabilistic and general weighted CFGs, we can now define the language model induced by $\grammar$ as follows.
\begin{definition}{A language model induced by a WCFG}{}
    Let $\grammar = \wcfgtuple$ be a WCFG.
    We define the \defn{language model induced by $\grammar$}\index{weighted context-free grammar!induced language model} as the following probability distribution over $\kleene{\alphabet}$
    \begin{equation} \label{eq:wcfg-lm}
        \pLM_\grammar\left(\str\right) \defeq \pdens_\grammar\left(\str\right).
    \end{equation}
\end{definition}

Again, it is easy to see that while global normalization requires the computation of the allsum, language models induced by weighted FSAs through \cref{eq:wcfg-str-prob} are \emph{globally normalized} and thus always tight.
The tightness of \emph{probabilistic} WCFGs is discussed next, after which we investigate the relationship between globally- and locally-normalized context-free grammars.

\subsection{Tightness of Context-free Language Models} \label{sec:tight-cfgs}

Again, an advantage of globally normalized context-free language models (grammars) is that they are always tight, as the derivation trees are explicitly normalized with the global normalization constant such that they sum to $1$ over the set of possible sentences.

In this section, we, therefore, consider the tightness of probabilistic context-free grammars.
We follow the exposition from \citet{Booth1973}.
The proof requires the use of multiple new concepts, which we first introduce below.
\begin{definition}{Generation level}{}
    We define the \defn{level of a generation sequence} inductively as follows.
    The zeroth level $\stackseq_0$ of a generation sequence is defined as $\start$.
    Then, for any $\idx > 0$, $\stackseq_\idx$ corresponds to the string is obtained by applying the applicable productions onto \emph{all} nonterminals of $\stackseq_{\idx - 1}$.
\end{definition}
\begin{example}{Generation levels}{}
    Let $\grammar = \cfgtuple$ with $\alphabet = \set{\syma, \symb}$, $\nonterm = \set{\start, \NTX, \NTY}$, \newline and $\rules = \set{\production{\start}{\syma\, \NTX\, \NTY}, \production{\NTX}{\NTY\, \NTX}, \production{\NTX}{\symb \NTY\, \NTY}, \production{\NTY}{\syma \, \syma \, \NTY}, \production{\NTY}{\syma}}$.
    Then the generation sequence of the string $\syma \syma \symb \syma \syma \syma \syma \syma$ would be
    \begin{align*}
        \stackseq_0 & = \start                                                      & \justification{definition}                                                                                                                   \\
        \stackseq_1 & = \syma \NTX \NTY                                             & \justification{applying $\production{\start}{\syma \, \NTX \, \NTY}$}                                                                        \\
        \stackseq_2 & = \syma \NTY \NTX \syma \syma \NTY                            & \justification{applying $\production{\NTX}{\NTY \, \NTX}$, $\production{\NTY}{\syma \, \syma \, \NTY}$}                                      \\
        \stackseq_3 & = \syma \syma \symb \NTY \NTY \syma \syma \syma \syma \NTY    & \justification{applying $\production{\NTY}{\syma}$, $\production{\NTX}{\symb \, \NTY \, \NTY}$, $\production{\NTX}{\syma \, \syma \, \NTY}$} \\
        \stackseq_3 & = \syma \syma \symb \syma \syma \syma \syma \syma \syma \syma & \justification{applying $\production{\NTY}{\syma}$, $\production{\NTY}{\syma}$,  $\production{\NTY}{\syma}$}
    \end{align*}
\end{example}
We will also rely heavily on generating functions.
A generating function is simply a way of \emph{representing} an infinite sequence by encoding its elements as the coefficients of a \emph{formal power series}.
Unlike ordinary series such as the geometric power series from \cref{ex:geometric_allsum}, a formal power series does not need to converge: in fact, at its core a generating function is not actually regarded as a \emph{function}---its ``variables'' are indeterminate and they simply serve as ``hooks'' for the numbers in the sequence.
\begin{definition}{Production generating function}{}
    Let $\grammar \defeq \wcfgtuple$ be a PCFG and $N \defeq |\nonterm|$.
    For each $\NTX_\idx \in \nonterm$, define its \defn{production generating function}\index{production generating function} as
    \begin{equation}
        \funcg\left(s_1, \ldots, s_N\right) \defeq \sum_{\production{\NTX_\idx}{\cfgstr}} \productionWeight\left(\production{\NTX_\idx}{\cfgstr}\right) s_1^{\rr_1\left(\cfgstr\right)} s_2^{\rr_2\left(\cfgstr\right)} \cdot \cdots \cdot s_N^{\rr_N\left(\cfgstr\right)},
    \end{equation}
    where $\rr_\idxm\left(\cfgstr\right)$ denotes the number of times the nonterminal $\NTX_\idxm \in \nonterm$ appears in $\cfgstr \in \kleene{\left(\alphabet \cup \nonterm\right)}$.
\end{definition}
\begin{example}{Tightness of a context-free grammar}{cfg-tight-example-2}
    Let $\grammar = \cfgtuple$ with $\alphabet = \set{\syma, \symb}$, $\nonterm = \set{\start, \NTX}$, and $\rules = \set{\production{\start}{\syma \, \start \, \NTX}, \production{\start}{\symb}, \production{\NTX}{\syma \NTX\, \NTX}, \production{\NTX}{\syma \syma}}$.
    Then
    \begin{align*}
        \funcg_1\left(s_1, s_2\right) & = \productionWeight\left(\production{\start}{\syma \start \NTX} \right) s_1 s_2 + \productionWeight\left(\production{\start}{\symb}\right) \\
        \funcg_2\left(s_1, s_2\right) & = \productionWeight\left(\production{\NTX}{\syma \NTX \NTX} \right) s_2^2 + \productionWeight\left(\production{\NTX}{\syma \syma}\right)
    \end{align*}
\end{example}
\begin{definition}{Generating function}{}
    The \defn{generating function} of the $\idxl^\text{th}$ level \index{generating function} is defined as
    \begin{align}
        \genFunc_0\left(s_1, \ldots, s_N\right)     & \defeq s_1                                                                                                                    \\
        \genFunc_1\left(s_1, \ldots, s_N\right)     & \defeq \funcg_1\left(s_1, \ldots, s_N\right)                                                                                  \\
        \genFunc_\idxl\left(s_1, \ldots, s_N\right) & \defeq \genFunc_{\idxl - 1}\left(\funcg_1\left(s_1, \ldots, s_N\right), \ldots, \funcg_N\left(s_1, \ldots, s_N\right)\right),
    \end{align}
    that is, the $\idxl^\text{th}$-level generating function is defined as the $\idxl - 1^\text{st}$-level generating function applied to production generating functions as arguments.
\end{definition}
\begin{example}{Tightness of a context-free grammar}{}
    For the grammar from \cref{ex:cfg-tight-example-2}, we have
    \begin{align*}
        \genFunc_0\left(s_1, s_2\right) & = s_1                                                                                                                                                                                                                    \\
        \genFunc_1\left(s_1, s_2\right) & = \funcg\left(s_1, s_2\right) = \productionWeight\left(\production{\start}{\syma \start \NTX} \right) s_1 s_2 + \productionWeight\left(\production{\start}{\symb}\right)                                                 \\
        \genFunc_2\left(s_1, s_2\right) & = \productionWeight\left(\production{\start}{\syma \start \NTX} \right) \left[\funcg_1\left(s_1, s_2\right)\right] \left[\funcg_2\left(s_1, s_2\right)\right] + \productionWeight\left(\production{\start}{\symb}\right) \\
                                        & = \productionWeight\left(\production{\start}{\syma \start \NTX} \right)^2 \productionWeight\left(\production{\NTX}{\syma \NTX \NTX} \right) s_1 s_2^3                                                                    \\
                                        & + \productionWeight\left(\production{\start}{\syma \start \NTX} \right)^2 \productionWeight\left(\production{\NTX}{\syma \syma} \right) s_1 s_2                                                                          \\
                                        & + \productionWeight\left(\production{\start}{\syma \start \NTX} \right) \productionWeight\left(\production{\start}{\symb} \right) \productionWeight\left(\production{\NTX}{\syma \NTX \NTX} \right) s_2^2                \\
                                        & + \productionWeight\left(\production{\start}{\syma \start \NTX} \right) \productionWeight\left(\production{\start}{\symb} \right) \productionWeight\left(\production{\NTX}{\syma \syma} \right)                          \\
                                        & + \productionWeight\left(\production{\start}{\symb} \right)
    \end{align*}
\end{example}

We can see that a generating function $\genFunc_\idxl\left(s_1, \ldots, s_N\right)$ can be expressed as
\begin{equation}
    \genFunc_\idxl\left(s_1, \ldots, s_N\right) = D_\idxl\left(s_1, \ldots, s_N\right) + C_\idxl
\end{equation}
where the polynomial $D_\idxl\left(s_1, \ldots, s_N\right)$ does not contain any constant terms.
It is easy to see that the constant $C_\idxl$ then corresponds to the probability of all strings that can be derived in $\idxl$ levels or fewer.
This brings us to the following simple lemma.
\begin{lemma}{}{}
    A PCFG is tight if and only if
    \begin{equation} \label{eq:pcfg-tightness-limit}
        \lim_{\idxl \to \infty} C_\idxl = 1.
    \end{equation}
\end{lemma}
\begin{proof}
    Suppose that $\lim_{\idxl \to \infty} C_\idxl < 1$. This means that the generation process can enter a generation sequence that has a non-zero probability of not terminating---this corresponds exactly to it not being tight.

    On the other hand, $\lim_{\idxl \to \infty} C_\idxl = 1$ implies that no such sequence exists, since the limit represents the probability of all strings that can be generated by derivations of a finite number of production rules.
\end{proof}

The rest of the section considers necessary and sufficient conditions for \cref{eq:pcfg-tightness-limit} to hold.
For this, we first define the first-moment matrix of a PCFG.
\begin{definition}{First-moment matrix}{}
    Let $\grammar \defeq \wcfgtuple$ be a PCFG.
    We define its \defn{first-moment matrix} (its mean matrix) $\mE \in \R^{N \times N}$ as
    \begin{equation}
        \emE_{\idxn\idxm} \defeq \frac{\partial\funcg_\idxn\left(s_1, \ldots, s_N\right)}{\partial s_\idxm}\Bigg|_{s_1, \ldots, s_N = 1}.
    \end{equation}
\end{definition}
Note that $\emE_{\idxn\idxm}$ represents the \emph{expected number of occurrences} of the non-terminal $\NTX_\idxm$ in the set of sequences $\cfgstr$ with $\NTX_\idxn \derivesbase_{\scaleto{\grammar}{5pt}} \cfgstr$, i.e., the set of sequences $\NTX_\idxn$ can be rewritten into:
\begin{equation}
    \emE_{\idxn\idxm} = \sum_{\production{\NTX_\idxn}{\cfgstr}} \productionWeight\left(\production{\NTX_\idxn}{\cfgstr}\right) \rr_\idxm\left(\cfgstr\right).
\end{equation}
The informal intuition behind this is the following: each of the terms $\productionWeight\left(\production{\NTX_\idx}{\cfgstr}\right) s_1^{\rr_1\left(\cfgstr\right)} s_2^{\rr_2\left(\cfgstr\right)} \cdot \cdots \cdot s_N^{\rr_N\left(\cfgstr\right)}$ in $\funcg_\idxn$ contains the information about \emph{how many times} any non-terminal $\NTX_\idxm$ appears in the production rule $\production{\NTX_\idxn}{\cfgstr}$ as well as what the probability of ``using'' or applying that production rule to $\NTX_\idxn$ is.
Differentiating $\productionWeight\left(\production{\NTX_\idx}{\cfgstr}\right) s_1^{\rr_1\left(\cfgstr\right)} s_2^{\rr_2\left(\cfgstr\right)} \cdot \cdots \cdot s_N^{\rr_N\left(\cfgstr\right)}$ w.r.t. $s_\idxm$ then ``moves'' the coefficient $\rr_\idxm$ corresponding to the number of occurrences of $\NTX_\idxm$ in $\production{\NTX_\idxn}{\cfgstr}$ in front of the term $\productionWeight\left(\production{\NTX_\idx}{\cfgstr}\right) s_1^{\rr_1\left(\cfgstr\right)} s_2^{\rr_2\left(\cfgstr\right)} \cdot \cdots \cdot s_N^{\rr_N\left(\cfgstr\right)}$ in $\funcg_\idxn$, effectively multiplying the probability of the occurrence of the rule with the number of terms $\NTX_\idxm$ in the rule---this is exactly the expected number of occurrences of $\NTX_\idxm$ for this particular rule, averaging over all possible rules that could be applied.
Summing over all applicable production rules for $\NTX_\idxn$ (which form a probability distribution) gives us the total expected number of occurrences of $\NTX_\idxm$.
This brings us to the core theorem of this section characterizing the tightness of PCFGs.
\begin{theorem}{A sufficient condition for the tightness of probabilistic context-free grammars}{Tightness of PCFGs}
    A PCFG is tight if $|\evalmax| < 1$ and is non-tight if $|\evalmax| > 1$, where $\evalmax$ is the eigenvalue of $\mE$ with the largest absolute value.
\end{theorem}
\begin{proof}
    The coefficient of the term $s_1^{\rr_1} s_2^{\rr_2} \cdot \cdots \cdot s_N^{\rr_N}$ in the generating function $\genFunc_\idxl\left(s_1, \ldots, s_N\right)$ corresponds to the probability that there will be $r_1$ non-terminal symbols $\NTX_1$, \ldots, $r_N$ non-terminal symbols $\NTX_N$ in the $\idxl^\text{th}$ level of the generation sequence.
    In particular, if the grammar is tight, this means that
    \begin{equation}
        \lim_{\idxl \to \infty} \genFunc_\idxl\left(s_1, \ldots, s_N\right) = \lim_{\idxl \to \infty} \left[D_\idxl\left(s_1, \ldots, s_N\right) + C_\idxl\right] = 1.
    \end{equation}
    This, however, is only true if
    \begin{equation}
        \lim_{\idxl \to \infty} D_\idxl\left(s_1, \ldots, s_N\right) = 0
    \end{equation}
    and this, in turn, can only be true if $\lim_{\idxl \to \infty }\rr_\idxn = 0$ for all $\idxn = 1, \ldots N$.
    The expected value of $\rr_\idxn$ at level $\idxl$ is
    \begin{equation}
        \overline{\rr}_{\idxl, \idxn} = \frac{\partial\genFunc_\idxl\left(s_1, \ldots, s_N\right)}{\partial s_\idxn}\Bigg|_{s_1, \ldots, s_N = 1}.
    \end{equation}
    Reasoning about this is similar to the intuition behind the first-moment matrix, with the difference that we are now considering the number of occurrences after a sequence of $\idxl$ applications.
    Denoting
    \begin{equation}
        \overline{\rr}_\idxl \defeq \left[\overline{\rr}_{\idxl, 1}, \ldots, \overline{\rr}_{\idxl, N}\right]
    \end{equation}
    we have
    \begin{align}
        \overline{\rr}_\idxl & = \Bigg[ \sum_{\idxj = 1}^{N}     & \frac{\partial \genFunc_{\idxl - 1}\left(\funcg_1\left(s_1, \ldots, s_N\right), \ldots, \funcg_N\left(s_1, \ldots, s_N\right)\right)}{\partial \funcg_\idxj} \\
                             &                                   & \cdot \frac{\partial \funcg_\idxj}{\partial s_\idxn}\left(s_1, \ldots, s_N\right) \mid \idxn = 1, \ldots, N \Bigg]\Bigg|_{s_1, \ldots, s_N = 1}              \\
                             & = \overline{\rr}_{\idxl - 1} \mE. &
    \end{align}
    Applying this relationship repeatedly, we get
    \begin{equation}
        \overline{\rr}_\idxl = \overline{\rr}_0 \mE^{\idxl} = \left[1, 0, \ldots, 0\right] \mE^\idxl,
    \end{equation}
    meaning that
    \begin{equation}
        \lim_{\idxl \to \infty} \overline{\rr}_\idxl = \zero \text{ iff } \lim_{\idxl \to \infty} \mE_\idxl = \zero.
    \end{equation}
    The matrix $\mE$ satisfies this condition if $|\evalmax| < 1$.
    On the other hand, if $|\evalmax| > 1$, the limit diverges.
\end{proof}
Note that the theorem does not say anything about the case when $|\evalmax| = 1$.

We conclude the subsection by noting that, interestingly, weighted context-free grammars \emph{trained} on data with maximum likelihood are \emph{always} tight \citep{chi-geman-1998-estimation,Chi1999}.
This is not the case for some models we consider later, e.g., recurrent neural networks (cf. \cref{sec:rnns}).

\subsection{Normalizing Weighted Context-free Grammars}
Having investigated probabilistic context-free grammars in terms of their tightness, we now turn out attention to general weighted context-free grammars, which define string probabilities using global normalization (cf. \cref{eq:wcfg-str-prob}).
To be able to compute these probabilities, require a way to compute the normalizing constant $\allsum\left(\grammar\right)$ and the stringsum $\grammar\left(\str\right)$.
In the section on finite-state automata, we explicitly presented an algorithm for computing the normalizing constant $\allsum\left(\wfsa\right)$.
The derivation of a general allsum algorithm for weighted context-free grammars, on the other hand, is more involved and beyond the scope of this course.\footnote{The allsums of individual non-terminals can be expressed as solutions to a nonlinear set of equations. Again, the interested reader should have a look at the Advanced Formal Language Theory course.}
Here, we simply assert that there are ways of computing the quantities in \cref{eq:wcfg-str-prob} and only consider the following result:
\begin{theorem}{PCFGs and WCFGs are equally expressive \citep{smith2007}}{globally-locally-normalized-wcfgs}
    Normalizable weighted context-free grammars with non-negative weights and tight probabilistic context-free grammars are equally expressive.

\end{theorem}
\begin{proof}
    To prove the theorem, we have to show that any WCFG can be written as a PCFG and vice versa.\footnote{Again, by ``written as'', we mean that the weighted language is the same.}

    $\Leftarrow$
    Since any tight probabilistic context-free grammar is simply a WCFG with $\allsum\left(\grammar\right) = 1$, this holds trivially.

    $\Rightarrow$
    We now show that, for any WCFG, there exists a PCFG encoding the same language model.
    Let $\grammar_G = \wcfgtuple$ be a pruned WCFG that encodes a distribution over $\kleene{\alphabet}$ using \cref{eq:wcfg-str-prob}.
    We now construct a tight probabilistic context-free grammar $\grammar_L = \left(\alphabet, \nonterm, \start, \rules, \productionWeight_L\right)$ whose language is identical.
    Notice that all components of the grammar remain identical apart from the weighting function.
    This means that the derivations of the strings in the grammars remain the same (i.e., $\grammarDerivationset{\grammar_G} = \grammarDerivationset{\grammar_L}$)---only the weights of the derivations change, as we detail next.
    We define the production weights of the probabilistic CFG as follows.
    \begin{equation}
        \productionWeight_{\grammar_L}\left(\production{\NTX}{\cfgstr}\right) \defeq \frac{\productionWeight\left(\production{\NTX}{\cfgstr}\right)\prod_{\NTY \in \cfgstr} \allsum(\grammar, \NTY)}{\allsum(\grammar, \NTX)}
    \end{equation}
    Remember that $\allsum\left(\syma\right) = 1$ for $\syma \in \alphabet$.
    Note that the assumption that $\grammar$ is pruned means that all the quantities in the denominators are non-zero.

    It is easy to see that the weight defined this way are non-negative due to the non-negativity of $\grammar$'s weights.
    Furthermore, the weights of all production rules for any non-terminal $\NTX \in \nonterm$ sum to $1$, as by the definitions of $\productionWeight_{\grammar_L}$ and $\allsum(\grammar, \NTX)$ we have
    \begin{align}
        \sum_{\production{\NTX}{\cfgstr}} \productionWeight_{\grammar_L}\left(\production{\NTX}{\cfgstr}\right) & = \sum_{\production{\NTX}{\cfgstr}} \frac{\productionWeight\left(\production{\NTX}{\cfgstr}\right)\prod_{\NTY \in \cfgstr} \allsum(\grammar, \NTY)}{\allsum(\grammar, \NTX)}    \\
                                                                                                                & = \frac{1}{\allsum(\grammar, \NTX)}\sum_{\production{\NTX}{\cfgstr}} \productionWeight\left(\production{\NTX}{\cfgstr}\right)\prod_{\NTY \in \cfgstr} \allsum(\grammar, \NTY) & \\
                                                                                                                & = \frac{1}{\allsum(\grammar, \NTX)} \allsum(\grammar, \NTX)                                                                                                                     \\
                                                                                                                & = 1
    \end{align}

    We now have to show that the probabilities assigned by these two grammars match.
    We will do that by showing that the probabilities assigned to individual \emph{derivations} match, implying that stringsums match as well.
    The probability of a derivation is defined analogously to a probability of a string, i.e., $\pdens_\grammar\left(\tree\right) = \frac{\weight\left(\tree\right)}{\allsum\left(\grammar\right)}$ (where $\allsum\left(\grammar\right) = 1$ for tight probabilistic grammars).
    Let then $\tree \in \grammarDerivationset{\grammar} = \grammarDerivationset{\grammar_L}$.
    Then
    \begin{align} \label{eq:wcfg-local-norm-derivation}
        \pdens_{\grammar_L}\left(\tree\right) & = \prod_{\production{\NTX}{\cfgstr} \in \tree}\productionWeight_{\grammar_L}\left(\production{\NTX}{\cfgstr}\right)                                                                     &                                                                 \\
                                              & = \prod_{\production{\NTX}{\cfgstr} \in \tree} \frac{\productionWeight\left(\production{\NTX}{\cfgstr}\right)\prod_{\NTY \in \cfgstr} \allsum(\grammar, \NTY)}{\allsum(\grammar, \NTX)} & \justification{definition of $\productionWeight_{\grammar_L}$}.
    \end{align}
    Notice that by multiplying over the internal nodes of the derivation tree, \cref{eq:wcfg-local-norm-derivation} includes the non-terminal allsum of each \emph{internal} (non-root and non-leaf) non-terminal in the derivation \emph{twice}: once as a parent of a production in the denominator, and once as a child in the numerator.
    These terms, therefore, all cancel out in the product.
    The only terms which are left are the allsums of the leaf nodes---the terminals---which are $1$, and the allsum of the root node---$\start$---which equals $\allsum\left(\grammar_G\right)$ and the weights of the individual productions, which multiply into the weight assigned to $\tree$ by the original grammar $\grammar_G$.
    This means that
    \begin{equation}
        \pdens_{\grammar_L}\left(\tree\right) = \frac{1}{\allsum(\grammar, \NTX)}\prod_{\production{\NTX}{\cfgstr} \in \tree} \productionWeight\left(\production{\NTX}{\cfgstr}\right) = \frac{1}{\allsum(\grammar, \NTX)}\weight\left(\tree\right) = \pdens_{\grammar_G}\left(\tree\right),
    \end{equation}
    finishing the proof.
\end{proof}
This means that the classes of probabilistic and weighted context-free grammars are in fact \emph{equally expressive}.
In other words, this result is analogous to \cref{thm:locally-normalizing-wfsas} in WFSAs: it shows that in the context of context-free language models, the locally normalized version of a globally-normalized model is \emph{also} context-free.


\subsection{Pushdown Automata}
We presented context-free grammars as a formalism for specifying and representing context-free languages.
Many algorithms for processing context-free languages, for example, the allsum algorithms and their generalizations, can also be directly applied to context-free grammars.
However, it is also convenient to talk about processing context-free languages in terms of computational models in the form of automata, i.e., the \emph{recognizer} of the language.\footnote{This relationship between a formalism specifying how to \emph{generate} (i.e., a grammar) and a model of \emph{recognizing} a language can be seen in multiple levels of the hierarchy of formal languages. In the case of context-free languages, the former are context-free grammars, while the latter are pushdown automata discussed in this subsection. Regular languages as introduced in the previous section, however, are simply defined in terms of their recognizers---finite-state automata.}
As we mentioned, the types of automata we considered so far, (weighted) finite-state automata, can only recognize regular languages.
To recognize context-free languages, we must therefore extend finite-state automata.\footnote{Formally, we would of course have to prove that finite-state automata cannot model context-free languages. This can be done with the so-called pumping lemma, which are outside the scope of this class.}
We do that by introducing \defn{pushdown automata} (PDA), a more general and more expressive type of automata.

\subsubsection{Single-stack Pushdown Automata} \label{sec:wpda}
Pushdown automata augment finite-state automata by implementing an additional \emph{stack} memory structure for storing \emph{arbitrarily long} strings from a designated alphabet, which allows them to work with unbounded memory effectively.
Abstractly, this unbounded memory is the only difference to finite-state automata.
However, the definition looks slightly different:
\begin{definition}{Pushdown automaton}{pda}
    A \defn{pushdown automaton}\index{pushdown automaton} (PDA) is a tuple $\pushdown \defeq \pdatuple$, where:
    \begin{itemize}
        \item $\states$ is a finite set of states;
        \item $\alphabet$ is a finite set of input symbols called the input alphabet;
        \item $\stackalphabet$ is a finite set of stack symbols called the stack alphabet;
        \item $\trans \subseteq \states \times \kleene{\stackalphabet} \times \left( \alphabet \cup  \set{\eps} \right) \times \states \times \kleene{\stackalphabet}$ is a multiset representing the transition function;
        \item $\pdaconfig{\initstack}{\qinit}$ is called the initial configuration and $\pdaconfig{\finalstack}{\qfinal}$ is called the final configuration, where $\qinit, \qfinal \in \states$ and $\initstack, \finalstack \in \kleene{\stackalphabet}$ .
    \end{itemize}
\end{definition}
The initial and final configurations in pushdown play analogous roles to the sets of initial and final sets of finite-state automata.
Compared to the latter, they also allow for different starting configurations of the stack coupled with each possible initial or final state.

Stacks are represented as strings over $\stackalphabet$, from bottom to top.
Thus, in the stack $\stackseq = \stacksymbol{X_1} \stacksymbol{X_2} \cdots \stacksymbol{X_n}$, the symbol $\stacksymbol{X_1}$ is at the bottom of the stack, while $\stacksymbol{X_n}$ is at the top.
$\stackseq = \emptystack$ denoses the empty stack.

\begin{definition}{Configuration of a pushdown automaton}{}
    A \defn{configuration}\index{pushdown automaton!configuration} of a PDA is a pair $\pdaconfig{\stackseq}{\stateq}$, where $\stateq \in \states$ is the current state and ${\stackseq} \in \kleene{\stackalphabet}$ is the current contents of the stack.
\end{definition}

The initial and final configurations of a PDA are examples of configurations; it is possible to generalize the initial and final stacks to (say) regular expressions over $\stackalphabet$, but the above definition suffices for our purposes.

A PDA moves from configuration to configuration by following transitions of the form $\pdaEdgenoweight{\stateq}{a}{\stater}{\stackseq_1}{\stackseq_2}$, which represents a move from the state $\stateq$ to state $\stater$, while popping the sequence of symbols $\stackseq_1 \in \kleene{\stackalphabet}$ from the top of the stack and pushing the sequence $\stackseq_2 \in \kleene{\stackalphabet}$.
The PDA transition function therefore not only depends on the current state $\stateq$ and input symbol $\syma$, but also on some \emph{finite} sequence of symbols on the \emph{top} of the stack.
The stack hence determines the behavior of the automaton, and since the set of possible configurations of the stack is infinite, the set of configurations of the automaton is infinite, in contrast to finite-state automata.

To describe how pushdown automata process strings, we introduce the concepts of scanning and runs.\alexandra{(scanning) runs?}
\begin{definition}{Scanning}{}
    We say that $\atrans = \left(\statep, \stackseq_1, \syma, \stateq, \stackseq_2 \right) \in \trans$ \defn{scans}\index{pushdown automaton!scan} $a$, and if $a \neq \eps$, we call $\atrans$ \defn{scanning}\index{pushdown automaton!scanning transition}; otherwise, we call it \defn{non-scanning}\index{pushdown automaton!non-scanning transition}.
\end{definition}

\begin{definition}{Pushdown automaton transitions}{}
    If $\pdaconfig{\stackseq \stackseq_1}{\stateq_1}$ and $\pdaconfig{\stackseq \stackseq_2}{\stateq_2}$ are configurations, and $\atrans$ is a transition $\pdaEdgenoweight{\stateq_1}{a}{\stateq_2}{\stackseq_1}{\stackseq_2}$, we write $\pdaconfig{\stackseq \stackseq_1}{\stateq_1} \Rightarrow_\atrans \pdaconfig{\stackseq \stackseq_2}{\stateq_2}$.
\end{definition}

Since the behavior of a pushdown automaton does not only depend on the states encountered by it but also on the content of the stack, we generalize the notion of a path to include the \emph{configuration} of the automaton.
This is called a run.\ryan{Is this right that it alternates?}
\begin{definition}{Run of a pushdown automaton}{}
    A \defn{run}\index{pushdown automaton!run} of a PDA $\pushdown$ is a sequence of configurations and transitions \[\arun=\pdaconfig{\stackseq_0}{\stateq_0}, \atrans_1, \pdaconfig{\stackseq_1}{\stateq_1}, \ldots, \atrans_n, \pdaconfig{\stackseq_\pathlen}{\stateq_\pathlen}\] where, for $\idx = 1, \ldots, \pathlen$, we have $\pdaconfig{\stackseq_{\idx-1}}{\stateq_{\idx-1}} \Rightarrow_{\atrans_\idx} \pdaconfig{\stackseq_\idx}{\stateq_\idx}$.\footnote{Sometimes it will be convenient to treat $\arun$ as a sequence of only configurations or only transitions.}
    A run is called \defn{accepting}\index{pushdown automaton!accepting run} if $\pdaconfig{\stackseq_0}{\stateq_0}$ is the initial configuration and $\pdaconfig{\stackseq_\pathlen}{\stateq_\pathlen}$ is the final configuration.
    If, for $\idx=1, \ldots, \pathlen$, $\atrans_\idx$ scans $a_\idx$, then we say that $\arun$ scans the string $a_1 \cdots a_\pathlen$.
    We write $\runs \left(\pushdown, \str \right)$ for the set of runs that scan $\str$ and $\runs \left( \pushdown \right)$ for the set of all accepting runs of $\pushdown$.
\end{definition}

\begin{definition}{Recognition of a string by a pushdown automaton}{}
    We say that the PDA $\pushdown$ \defn{recognizes}\index{pushdown automaton!recognized string} the string $\str$ if $\runs\left(\pushdown, \str\right) \neq \emptyset$, i.e., if there exists an accepting run with the yield $\str$.
    The set of all strings recognized by $\pushdown$ is the \defn{language recognized by $\pushdown$}\index{pushdown automaton!recognized language}, which we denote by $\lang\left(\pushdown\right)$, i.e.,
    \begin{equation}
        \lang\left(\pushdown\right) \defeq \left\{\str \mid \runs\left(\pushdown, \str\right) \neq \emptyset\right\}.
    \end{equation}
\end{definition}
\alexandra{should mention the initial and final configurations as otherwise it is unclear why the path is accepting}
\begin{example}{Example of a pushdown automaton}{}
    \cref{fig:pda_example} shows an example of a pushdown automaton $\pushdown$ accepting the language $\lang\left(\pushdown\right) = \{a^n b^n \mid n \in \mathbb{N}\}$.
    $\left( \pdaEdgenoweight{1}{a}{1}{\eps}{\NTX},  \pdaEdgenoweight{1}{\eps}{2}{\eps}{\eps} \right) $ is a run of $\pushdown$; $\left(\pdaEdgenoweight{1}{a}{1}{\eps}{\NTX},  \pdaEdgenoweight{1}{\eps}{2}{\eps}{\eps}, \pdaEdgenoweight{2}{b}{2}{\NTX}{\eps} \right) $ is an \emph{accepting} run of $\pushdown$.
\end{example}
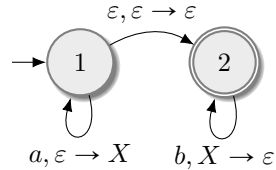
\begin{figure}[ht]
    \centering
    \begin{tikzpicture}[node distance = 10mm]
        \node[state, initial] (q1) [] { $1$ };
        \node[state, accepting] (q2) [right = of q1] { $2$ };
        \draw[transition] (q1) edge[bend left, above] node{ $\eps, \eps \rightarrow \eps$ } (q2)
        (q1) edge[loop below] node{ $a, \eps \rightarrow X$ } (q1)
        (q2) edge[loop below] node{ $b, X \rightarrow \eps$ } (q2) ;
    \end{tikzpicture}
    \caption{The PDA that accepts the language $\{a^n b^n \mid n \in \mathbb{N}\}$.}
    \label{fig:pda_example}
\end{figure}

Lastly, we define deterministic pushdown automata, analogously to their finite-state version (\cref{def:fsa-deterministic}). Recall that in the case of finite-state automata, a deterministic machine has at most one possible next move for each state. Similalry, a deterministic pushdown automaton has at most one possible next move for each \emph{configuration}.
\begin{definition}{Deterministic pushdown automaton}{pda-deterministic}
    A PDA $\pushdown = \pdatuple$ is \defn{deterministic}\index{deterministic PDA} if
    \begin{itemize}
        \item there are no transitions of the type $\left(q,\eps,\stackseq,p,\stackseq\right)$;
        \item for every $\left(q, a, \stackseq \right) \in \states \times \alphabet \cup \{ \eps\} \times \kleene{\stackalphabet}$, there is at most one transition $\left(q, a, \stackseq, p, \stackseq' \right) \in \trans$;
        \item if there is a transition $\left(q, a, \stackseq, p, \stackseq' \right) \in \trans$ for some $a\in\alphabet$, then there is no transition $\left(q,\eps,\stackseq,p,\stackseq'' \right)\in\trans$.
    \end{itemize}
    Otherwise, $\pushdown$ is \defn{non-deterministic}\index{non-deterministic PDA}.
\end{definition}
Importantly, \emph{not all} context-free languages can be recognized by deterministic pushdown automata.
That is, in contrast to finite-state automata, where deterministic machines are just as powerful as non-deterministic ones (at least in the unweighted case---interestingly, some weighted non-deterministic FSAs cannot be determinized), non-deterministic pushdown automata are more expressive than deterministic ones.
Specifically, as stated in \cref{thm:cfgs-pdas}, non-deterministic pushdown automata recognize exactly context-free languages, while deterministic pushdown automata only recognize a subset of them \citep{Sipser2013}.

\subsubsection{Weighted Pushdown Automata}
Analogously to the finite-state case, and the case of context-free grammars, we now also extend the definition of a pushdown automaton to the weighted case.
The formal definition is:
\begin{definition}{Weighted pushdown automaton}{}
    A \defn{real-weighted pushdown automaton}\index{weighted pushdown automaton} (WPDA) is a tuple $\pushdown = (\states, \alphabet, \stackalphabet, \trans, \pdaconfig{\initstack}{\qinit}, \pdaconfig{\finalstack}{\qfinal})$, where:
    \begin{itemize}
        \item $\states$ is a finite set of states;
        \item $\alphabet$ is a finite set of input symbols called the input alphabet;
        \item $\stackalphabet$ is a finite set of stack symbols called the stack alphabet;
        \item $\trans \subseteq \states \times \kleene{\stackalphabet} \times \left( \alphabet \cup  \set{\eps} \right) \times \states \times \kleene{\stackalphabet} \times \R$ is a multi-set representing the transition weighting function;
        \item $\pdaconfig{\initstack}{\qinit}$ is called the initial configuration and $\pdaconfig{\finalstack}{\qfinal}$ is called the final configuration, where $\qinit, \qfinal \in \states$ and $\initstack, \finalstack \in \kleene{\stackalphabet}$ .
    \end{itemize}
\end{definition}
As you can see, the only difference between the weighted and the unweighted case is the transition function, which in the weighted case weights the individual transitions instead of specifying the set of possible target configurations.

As with WFSAs (\cref{def:stochastic-wfsa}) and WCFGs (\cref{def:probabilistic-cfg}), we now define probabilistic WPDAs.
This definition, however, is a bit more subtle.
Notice that the transition weighting ``function'' $\trans$ in a WPDA is crucially still a \emph{finite}---there is only a finite number of actions we can ever do.
Similarly, when defining a \emph{probabilistic} PDA, we have to limit ourselves to a finite number of configurations over which we define probability distributions over the next possible actions.
We define a probabilistic pushdown automaton given an equivalence relation as follows.
\begin{definition}{Probabilistic pushdown automaton}{}
    A WPDA $\pushdown = (\states, \alphabet, \stackalphabet, \trans, \pdaconfig{\initstack}{\qinit}, \pdaconfig{\finalstack}{\qfinal})$ is \defn{probabilistic}\index{probabilistic pushdown automaton} if it holds that
    \begin{equation}
        \forall \ \pdaEdge{w}{\stateq}{a}{\stater}{\stackseq_1}{\stackseq_2} \in \trans: \  w \geq 0
    \end{equation}
    and for any $q\in\states$ and $\stackseq\in \kleene{\stackalphabet}$
    \begin{equation}
        \sum_{\substack{\pdaEdge{w}{\stateq}{a}{\stater}{\stackseq_1}{\stackseq_2} \\ \text{s.t. } \suffixOf{\stackseq_1}{\stackseq}}} w = 1.
    \end{equation}
\end{definition}

\begin{definition}{Transitions of a weighted pushdown automaton}{}
    If $\pdaconfig{\stackseq \stackseq_1}{\stateq_1}$ and $\pdaconfig{\stackseq \stackseq_2}{\stateq_2}$ are configurations, and $\atrans$ is a transition $\pdaEdge{w}{\stateq_1}{a}{\stateq_2}{\stackseq_1}{\stackseq_2}$ with $w \neq 0$, we write $\pdaconfig{\stackseq \stackseq_1}{\stateq_1} \Rightarrow_\atrans \pdaconfig{\stackseq \stackseq_2}{\stateq_2}$.
\end{definition}

\begin{definition}{Transition weights in a pushdown automaton}{}
    If $\trans(\statep,\stackseq_1,a,\stateq,\stackseq_2) = w$, then we usually write
    \begin{equation}
        \transweight{\statep}{a}{\stateq}{\stackseq_1}{\stackseq_2} = w
    \end{equation}
    or that $\trans$ has transition $(\pdaEdge{w}{\stateq}{a}{\statep}{\stackseq_1}{\stackseq_2})$.
    We sometimes let $\atrans$ stand for a transition, and we define $\trans(\atrans) = w$.
\end{definition}

And again, just like we combined the weights of individual transitions into the weights of paths in WFSAs, and we combined the weights of production rules into the weights of the trees in WCFGs, we now multiplicatively combine the weights of individual transitions in a run to define the weight of a run in a WPDA:
\begin{definition}{Run weight}{}
    The \defn{weight}\index{weighted pushdown automaton!run weight} $\weight\left(\arun\right)$ of a run \[\arun=\pdaconfig{\stackseq_0}{\stateq_0}, \atrans_1, \pdaconfig{\stackseq_1}{\stateq_1}, \ldots, \atrans_\pathlen, \pdaconfig{\stackseq_\pathlen}{\stateq_\pathlen}\] is the multiplication of the transition weights, i.e.,
    \begin{equation}
        \weight\left(\arun\right)\defeq \prod_{\idx = 1}^\pathlen \trans\left(\atrans_\idx\right)
    \end{equation}
\end{definition}

Analogously to a stringsum in WFSAs, we define the stringsum for a string $\str$ in a WPDA $\pushdown$ as the sum over the weights of all runs scanning $\str$.
\begin{definition}{Stringsum in a pushdown automaton}{}
    Let $\pushdown$ be a WPDA and $\str \in \kleene{\alphabet}$ a string.
    The \defn{stringsum}\index{weighted pushdown automaton!stringsum} for $\str$ in $\pushdown$ is defined as
    \begin{equation}
        \pushdown\left(\str\right) \defeq \sum_{\arun \in \runs\left(\pushdown, \str\right)} \weight\left(\arun\right)
    \end{equation}
\end{definition}

\begin{definition}{Recognition by a weighted pushdown automaton}{} \label{def:pda-recognize-string}
    We say that the PDA $\pushdown$ \defn{recognizes} the string $\str$ with the weight $\pushdown\left(\str\right)$.
\end{definition}

With this, we can define the weighted language defined by a WPDA.
\begin{definition}{Weighted language of a weighted pushdown automaton}{}
    Let $\pushdown$ be a WPDA.
    The \defn{(weighted) language} $\lang\left(\pushdown\right)$ of $\pushdown$ is defined as
    \begin{equation}
        \lang\left(\pushdown\right) \defeq \left\{\left(\str, \pushdown\left(\str\right)\right)\mid \str \in \kleene{\alphabet}\right\}
    \end{equation}
\end{definition}

Finally, we also define the WPDA allsum and normalizable WPDAs.
\begin{definition}{Allsum of a weighted pushdown automaton}{}
    The \defn{allsum}\index{weighted pushdown automaton!allsum} of a WPDA $\pushdown$ is defined as
    \begin{equation}
        \allsum\left(\pushdown\right) \defeq \sum_{\arun \in \runs\left(\pushdown\right)} \weight\left(\arun\right)
    \end{equation}
\end{definition}
\begin{definition}{Normalizable weighted pushdown automaton}{}
    A WPDA $\pushdown$ is \defn{normalizable}\index{weighted pushdown automaton!normalizable} if $\allsum\left(\pushdown\right)$ is finite, i.e., if $\allsum\left(\pushdown\right) < \infty$.
\end{definition}

\subsubsection{Relationship to Context-free Grammars} \label{sec:wcfg-vs-wpda}

We motivated the introduction of pushdown automata as a means of recognizing context-free languages.
However, this correspondence is not obvious from the definition!
Indeed, the equivalence of the expressive power of context-free grammars and pushdown automata is a classic result in formal language theory, and it is summarised by the theorem below:
\begin{theorem}{Context-free grammars and pushdown automata are equally expressive}{cfgs-pdas}
    A language is context-free if and only if some pushdown automaton recognizes it.
\end{theorem}
\begin{proof}
    See Theorem 2.20 in \citet{Sipser2013}.
\end{proof}

This result extends to the probabilistic case.
\begin{theorem}{Probabilistic context-free grammars and probabilistic pushdown automata are equally expressive}{}
    A language is generated by a probabilistic context-free grammar if and only if some probabilistic pushdown automaton recognizes it.
\end{theorem}
\begin{proof}
    See Theorems 3 and 7 in \citet{Abney1999}.
\end{proof}

Lastly, analogously to how \cref{thm:globally-locally-normalized-wcfgs} showed that weighted context-free grammars are equally expressive as probabilistic context-free grammars, the following theorem asserts the same about pushdown automata:
\begin{theorem}{Globally normalized weighted pushdown automata can be locally normalized}{globally-locally-normalized-wpdas}
    Any globally normalized weighted pushdown automaton can be locally normalized.
    More precisely, this means the following.
    Let $\pushdown$ be a weighted pushdown automaton.
    Then, there exists a probabilistic pushdown automaton $\pushdown_p$ such that \begin{equation} \label{eq:globally-locally-normalized-wpdas}
        \pushdown_p\left(\str\right) = \frac{\pushdown\left(\str\right)}{\allsum\left(\pushdown\right)}
    \end{equation}
    for all $\str \in \kleene{\alphabet}$.
\end{theorem}
\begin{proof}
    The proof is not straightforward: it can be shown that one cannot simply convert an arbitrary weighted pushdown automaton into a locally-normalized one directly.
    Rather, the construction of the latter goes through their \emph{context-free grammars}: given a WPDA $\pushdown$, one first constructs the WCFG equivalent to $\pushdown$, and then converts that to a locally normalized one (cf. \cref{thm:globally-locally-normalized-wcfgs}), i.e., a PCFG.
    Then, this PCFG can be converted to a structurally quite different probabilistic pushdown automaton $\pushdown_p$, which nevertheless results in the language we require (\cref{eq:globally-locally-normalized-wpdas}).

    This construction is described in more detail in \citet{Abney1999} and \citet{Butoi2022}.
\end{proof}

\subsection{Pushdown Language Models}

We can now define pushdown language models, the title of this section.
\begin{definition}{Pushdown language model}{}
    A \defn{pushdown language model}\index{language model!pushdown} is a language model whose weighted language equals the language of some weighted pushdown automaton, i.e., if there exists a weighted pushdown automaton $\pushdown$ such that $\lang\left(\pushdown\right) = \lang\left(\pLM\right)$.
\end{definition}

Similarly, pushdown automata also induce language models.
\begin{definition}{Language model induced by a pushdown automaton}{}
    Let $\pushdown$ be a weighted pushdown automaton.
    We define the \defn{language model induced by $\pushdown$}\index{weighted pushdown automaton!induced language model} as the probability distribution induced by the probability mass function
    \begin{equation}
        \pLM_{\pushdown} \left(\str\right) \defeq \frac{\pushdown\left(\str\right)}{\allsum\left(\pushdown\right)},
    \end{equation}
    for any $\str \in \kleene{\alphabet}$.
\end{definition}
You might wonder why we specifically define pushdown language models and models induced by them if WPDAs are equivalent to WCFGs (cf. \cref{sec:wcfg-vs-wpda}).
In that sense, a language model is context-free if and only if it is a pushdown language model.
However, this holds only for single-stack pushdown automata which we have discussed so far.
We make this explicit distinction of pushdown language models with an eye to the next section, in which we introduce \emph{multi-stack} WPDAs.
Those are, as it turns out, much more powerful (expressive) than context-free grammars.
We will, however, reuse this definition of a pushdown language model for those more powerful machines.

\subsection{Multi-stack Pushdown Automata}
We now consider an extension of (weighted) pushdown automata, namely, machines that employ \emph{multiple} stacks.
While this might not seem like an important distinction, we will see shortly that this augmentation results in a big difference in the expressiveness of the framework!
\begin{definition}{Two-stack pushdown automaton}{two-stack-pda}
    A \defn{two-stack pushdown automaton} \index{two-stack pushdown automaton} (2-PDA) is a tuple $\pushdown = (\alphabet, \states, \stackalphabet_1, \stackalphabet_2, \trans, \twoPdaconfig{\initstack_1}{\initstack_2}{\qinit}, \twoPdaconfig{\finalstack_1}{\finalstack_2}{\qfinal})$, where:
    \begin{itemize}
        \item $\alphabet$ is a finite set of input symbols called the input alphabet;
        \item $\states$ is a finite set of states;
        \item $\stackalphabet_1$ and $\stackalphabet_2$ are finite sets of stack symbols called the stack alphabets;
        \item $\trans \subseteq \states \times \kleene{\stackalphabet_1} \times \kleene{\stackalphabet_2} \times \left( \alphabet \cup  \set{\eps} \right) \times \states \times \kleene{\stackalphabet_1} \times \kleene{\stackalphabet_2}$ is a multiset representing the transition function;
        \item $\twoPdaconfig{\initstack_1}{\initstack_2}{\qinit}$ is called the initial configuration and $\twoPdaconfig{\finalstack_1}{\finalstack_2}{\qfinal}$ is called the final configuration, where $\qinit, \qfinal \in \states$, $\initstack_1, \finalstack_1 \in \kleene{\stackalphabet_1}$, and $\initstack_2, \finalstack_2 \in \kleene{\stackalphabet_2}$.
    \end{itemize}
\end{definition}
Note that we could more generally define a $k$-stack PDA by including $k$ stacks in the definition, but the restriction to two stacks will be sufficient for our needs, as we will see in the next subsection.
The transition function now depends on the values stored in \emph{both} of the stacks.
The definitions of the configuration and run of a two-stack PDA are analogous to the single-stack variant, with the addition of the two stacks.
We again extend this definition to the weighted and the probabilistic case.
\begin{definition}{Two-stack weighted pushdown automaton}{two-stack-wpda}
    A \defn{two-stack real-weighted pushdown automaton} \index{two-stack weighted pushdown automaton} (2-WPDA) is a tuple $\pushdown = (\alphabet, \states, \stackalphabet_1, \stackalphabet_2, \trans, \twoPdaconfig{\initstack_1}{\initstack_2}{\qinit}, \twoPdaconfig{\finalstack_1}{\finalstack_2}{\qfinal})$, where:
    \begin{itemize}
        \item $\alphabet$ is a finite set of input symbols called the input alphabet;
        \item $\states$ is a finite set of states;
        \item $\stackalphabet_1$ and $\stackalphabet_2$ are finite sets of stack symbols called the stack alphabets;
        \item $\trans \subseteq \states \times \kleene{\stackalphabet_1} \times \kleene{\stackalphabet_2} \times \left( \alphabet \cup  \set{\eps} \right) \times \states \times \kleene{\stackalphabet_1} \times \kleene{\stackalphabet_2} \times \R$ is a multiset representing the transition weighting function;
        \item $\twoPdaconfig{\initstack_1}{\initstack_2}{\qinit}$ is called the initial configuration and $\twoPdaconfig{\finalstack_1}{\finalstack_2}{\qfinal}$ is called the final configuration, where $\qinit, \qfinal \in \states$, $\initstack_1, \finalstack_1 \in \kleene{\stackalphabet_1}$, and $\initstack_2, \finalstack_2 \in \kleene{\stackalphabet_2}$.
    \end{itemize}
\end{definition}

And lastly, we define probabilistic two-stack PDAs:
\begin{definition}{Probabilistic two-stack pushdown automaton}{}
    A 2-WPDA $\pushdown = (\states, \alphabet, \stackalphabet_1, \stackalphabet_2, \trans, \twoPdaconfig{\initstack_1}{\initstack_2}{\qinit}, \twoPdaconfig{\finalstack_1}{\finalstack_2}{\qfinal})$ is \defn{probabilistic} \index{probabilistic two-stack pushdown automaton} if for any configuration $\twoPdaconfig{\stackseq_1}{\stackseq_2}{\stateq}$ it holds that
    \begin{equation}
        \forall \ \twoPdaEdge{w}{\stateq}{a}{\stater}{\stackseq_1}{\stackseq_1'}{\stackseq_2}{\stackseq_2'} \in \trans: \  w \geq 0
    \end{equation}
    and for any $q\in\states$ and $\stackseq\in \kleene{\stackalphabet_1}$, $\stackseq'\in \kleene{\stackalphabet_2}$
    \begin{equation}
        \sum_{\substack{\twoPdaEdge{w}{\stateq}{a}{\stater}{\stackseq_1}{\stackseq_1'}{\stackseq_2}{\stackseq_2'} \\ \text{s.t. } \suffixOf{\stackseq_1}{\stackseq} \text{ and } \suffixOf{\stackseq_2}{\stackseq'}}} w = 1 .
    \end{equation}
\end{definition}

\subsubsection{Turing Completeness of Multi-stack Pushdown Automata}
\anej{we should expand this section: why is Turing completeness interesting? what even is Turing completeness?}
Besides modeling more complex languages than finite-state language models from \cref{sec:finite-state}, (multi-stack) pushdown automata will also serve an important part in analyzing some modern language models that we introduce later.
Namely, we will show that Recurrent neural networks (cf. \cref{sec:rnns}) can \emph{simulate} any two-stack PDA.
This will be useful when reasoning about the computational expressiveness of recurrent neural networks because of a fundamental result in the theory of computation, namely, that two-stack PDAs are \emph{Turing complete}:
\begin{theorem}{Two-stack pushdown automata are Turing complete}{}
    Any 2-stack pushdown automaton is Turing complete.
    \label{thm:turing-completeness-PDA}
\end{theorem}
\begin{proof}
    The equivalence is quite intuitive: the two stacks (which are infinite in one direction) of the 2-PDA can simulate the tape of a Turing machine by popping symbols from one stack and pushing symbols onto the other one simultaneously.
    The head of the Turing machine then effectively reads the entries at the top of one of the two stacks.
    For a formal proof, see Theorem 8.13 in \citet{hopcroft79}.
\end{proof}
This is also the reason why we only have to consider two-stack PDAs---they can compute everything that can be computed, meaning that additional stacks do not increase their expressiveness!

Since unweighted pushdown automata are simply special cases of weighted PDAs, which are equivalent to probabilistic PDAs, we can therefore also conclude:\anej{Is it safe to assert the equivalence in the multi-stack case as well?}
\begin{corollary}{}{}
    Any weighted 2-stack pushdown automaton is Turing complete.
\end{corollary}
\begin{corollary}{}{}
    Any probabilistic 2-stack pushdown automaton is Turing complete.
\end{corollary}

A straightforward consequence of the Turing completeness of two-stack PPDAs is that their \emph{tightness} is undecidable.
\begin{theorem}{Tightness of 2-PPDA is undecidable}{}
    The tightness of a probabilistic two-stack pushdown automaton is undecidable.
\end{theorem}
\begin{proof}
    We start with a simple observation: a pushdown automaton $\pushdown$ is tight if and only if it \emph{halts} on inputs with measure $1$ (given the probability measure on $\kleene{\alphabet} \cup \alphabet^\infty$ defined in \cref{sec:tightness}), as this, by the definition of the language accepted by the WPDA (cf. \cref{def:pda-recognize-string}), corresponds to its language only containing \emph{finite} strings with probability $1$.\anej{This has to be checked thoroughly.}

    Let $\tm$ then be a Turing machine and $\pushdown$ be a 2-PPDA which simulates it.
    Then, $\pushdown$ is tight if and only if it halts with probability $1$ (again, based on the probability measure from above).
    This is equivalent to the problem of $\tm$ halting with probability $1$---this, however, is a variant of the \defn{halting problem}\index{halting problem}, which is one of the fundamental undecidable problems.
    We have therefore reduced the problem of determining the tightness of 2-PPDAs to the halting problem, implying that the former is undecidable.
\end{proof}

You might wonder what this means for the (weighted) languages recognized by multiple-stack (weighted) automata.
Turing machines can recognize \emph{recursively enumerable languages}.
This means that weighted multi-stack pushdown automata model \emph{distributions} over recursively enumerable languages.
To see why this might be useful, let us finish the discussion of context-free languages with an example of a language model that is \emph{not} context-free:\anej{move this to CFGs?}
\begin{example}{Example of a non-context-free distribution over strings}{}
    Let $\alphabet = \left\{\syma\right\}$ and $\pLM\left(\syma^n\right) = e^{-\lambda}\frac{\lambda^n}{n!}$ for $n \in \Nzero$, i.e., $\lang\left(\pLM\right)\ni\str = \syma^n \sim \text{Poisson}\left(\lambda\right)$ for some $\lambda > 0$.
    This language is not context-free: the proof, however, is not trivial.
    We direct the reader to \citet{Icard2020} for one.
\end{example}
\newpage{}

\section{Exercises}
\begin{Exercise}[label={exercise:state-allsum-decomposition}]
    Prove the following lemma.
    \begin{lemma}{}{}
        Let $\wfsa = \wfsatuple$ and $\stateq \in \states$.
        Then
        \begin{equation}
            \allsum\left(\wfsa, \stateq\right) = \sum_{\edge{\stateq}{\syma}{w}{\stateq'}\in \trans_{\wfsa_L}} \transitionWeight\left(\edge{\stateq}{\syma}{\cdot}{\stateq'}\right) \allsum(\wfsa, \stateq') + \finalf\left(\stateq\right)
        \end{equation}
    \end{lemma}
\end{Exercise}

\begin{Exercise}[label={exercise:ngram-mle}]
    Show that the expression for the log-likelihood of the \ngram{} model can be rewritten as
    \begin{equation}
        \loglikelihood\left(\dataset\right) = \sum_{\idxm = 1}^{M} \sum_{\tstep = 1}^{|\str^{\left(\idxm\right)}|} \log \param_{\sym_\ngr \mid \str{< \ngr}} = \sum_{\substack{\str\\|\str| = \ngr}} \strCount\left(\str\right) \param_{\sym_\ngr \mid \str{< \ngr}}
    \end{equation}
    with the quantities as defined in \cref{prop:ngram-mle}.
    This is a common trick.
    It is also known as the \defn{token to type switch} because we switch from counting over the individual tokens to counting over their identities (types)
\end{Exercise}

\begin{Exercise}[label={exercise:ngram-count-sum}]
    Let $\strCount\left(\str\right)$ be the string occurrence count for $\str \in \kleene{\alphabet}$ occurrence count as defined in \cref{prop:ngram-mle}.
    Show (or simply convince yourself) that, in a given training corpus $\dataset$
    \begin{equation}
        \sum_{\sym' \in \alphabet}\strCount\left(\sym_1 \ldots \sym_{\ngr - 1} \sym'\right) = \strCount\left(\sym_1 \ldots \sym_{\ngr - 1}\right)
    \end{equation}
\end{Exercise}
\newpage{}

\chapter{Neural Network Language Models} \label{ch:nn-lms}
\Cref{chapter:classical-lms} introduced two classical language modeling frameworks: finite-state language models and context-free language models.
While those served as a useful introduction to the world of language modeling, most of today's state-of-the-art language models go beyond the modeling assumptions of these two frameworks.
This chapter dives into the diverse world of modern language modeling architectures, which are based on neural networks.
We define two of the most common architectures---recurrent neural networks and transformers---and some of their variants.
The focus is again on rigorous formalization and theoretical understanding---we analyze the introduced models in terms of the theoretical foundations so far (e.g., expressiveness and tightness)---but, due to their practical applicability, we also study some practical aspects of the models.

We begin with recurrent neural networks.

\newpage{}

\section{Recurrent Neural Language Models}\label{sec:rnn}

The first neural language modeling architecture we consider is one based on recurrent neural networks.
Recurrent neural networks capture the idea of the \emph{sequential} processing of strings relatively naturally while also making decisions based on an \emph{infinite} context.
Before delving into the technical details of recurrent neural networks (RNNs), however, we first motivate the introduction of modeling contexts of unbounded length.
Then, we formally define recurrent neural networks and devote a large portion of the section to their theoretical properties.
The most important of those will be the Turing completeness of this architecture, as it has numerous consequences on the solvability of many of the tasks we might be interested in, such as finding the most probable string in the language model represented by a recurrent neural network and determining whether an RNN is tight.

\subsection{Human Language is Not Context-free}
Recall that we motivated the introduction of context-free languages by observing that finite memory is insufficient to model all formal phenomena of human language, e.g., infinite recursion (cf. \cref{ex:human-language-nonregular}).
Context-free languages described by context-free grammars and pushdown automata were able to capture those.
However, human language is more expressive than that---it includes linguistic phenomena that cannot be described by context-free grammars.
A typical example is called \defn{cross-serial dependencies}\index{cross-serial dependency}, which are common in \emph{Swiss German}.
\begin{example}{\citealp[Cross-Serial Dependencies in Swiss German,][]{shieber1985}}{human-language-noncf}
    Swiss German is a textbook example of a language with grammatical cross-serial dependencies, i.e., dependencies in which the arcs representing them, cross.
    In the example sentence below, the words connected with arcs are objects and verbs belonging to the same predicates (verb phrases).
    Because of that, they have to agree on the form---they depend on one another.
    As we show next, context-free languages cannot capture such dependencies. \\
    \begin{dependency}[theme=simple,align=center]
        \begin{deptext}[column sep=.1cm,font=\small]
            ...mer \& {\color{ETHGreen} d'chind} \& {\color{ETHRed} em Hans} \& {\color{ETHBlue} s' huus}  \& {\color{ETHGreen} l\"ond} \& {\color{ETHRed} h\"alfe} \& {\color{ETHBlue} aastriiche} \\
            ...we \& {\color{ETHGreen} the children} \& {\color{ETHRed} Hans} \& {\color{ETHBlue} the house} \& {\color{ETHGreen} let} \& {\color{ETHRed} help} \& {\color{ETHBlue} paint} \\
        \end{deptext}
        \depedge[-,ETHBlue,thick,edge height=8ex]{4}{7}{}
        \depedge[-,ETHRed,thick,edge height=8ex]{3}{6}{}
        \depedge[-,ETHGreen,thick,edge height=8ex]{2}{5}{}
    \end{dependency}
\end{example}

\paragraph{Why are cross-serial dependencies non-context-free?}
Before reasoning about the phenomenon of cross-serial dependencies, we revisit \cref{ex:human-language-nonregular} with a somewhat more formal approach.
The arbitrarily deep nesting can, for example, be abstractly represented with the expression
\begin{equation}
    x \textcolor{ETHBronze}{A^{n}} \textcolor{ETHBronze}{B^{n}} y
\end{equation}
with\footnote{In this case, we of course only consider an arbitrarily long sequence of barking dogs.}
\begin{align*}
    x                        & = \textexample{The cat} \nonumber          \\
    \textcolor{ETHBronze}{A} & = \textexample{the dog} \nonumber          \\
    \textcolor{ETHBronze}{B} & = \textexample{barked at} \nonumber        \\
    y                        & = \textexample{likes to cuddle}. \nonumber
\end{align*}
From this abstract perspective, center embeddings are very similar to the $\dyck{1}$ language (\cref{ex:dyck}), in that every noun phrase \textexample{the dog} has to be paired with a verb phrase \textexample{barked at}, which cannot be represented by any regular language.

In a similar fashion, \cref{ex:human-language-noncf} can abstractly be represented with the expression
\begin{equation} \label{eq:context-sensitive-expression}
    x \textcolor{ETHGreen}{A^{m}} \textcolor{ETHRed}{B^{n}} \textcolor{ETHGreen}{C^{m}} y \textcolor{ETHRed}{D^{n}} z
\end{equation}
with
\begin{align*}
    x                       & = \textexample{...mer} \nonumber      \\
    \textcolor{ETHGreen}{A} & = \textexample{d'chind} \nonumber     \\
    \textcolor{ETHRed}{B}   & = \textexample{em Hans} \nonumber     \\
    y                       & = \textexample{s' huus} \nonumber     \\
    \textcolor{ETHGreen}{C} & = \textexample{l\"ond} \nonumber      \\
    \textcolor{ETHRed}{D}   & = \textexample{h\"alfe} \nonumber     \\
    z                       & = \textexample{aastriiche}. \nonumber
\end{align*}
Admittedly, this is a relatively uncommon formulation even with $n = m = 1$.
It should be taken with a grain of salt, as the title of the original publication discussing this phenomenon, \emph{Evidence against context-freeness} \citep{shieber1985}, also suggests.
However, theoretically, the number of repetitions of \textexample{d'chind} and \textexample{l\"ond}, as well as  \textexample{em Hans} and \textexample{h\"alfe}, can be increased arbitrarily.
Repeating the former would correspond to having many groups of children.
The last of the groups would let Hans help paint the house, whereas each of the previous groups would let the group after them either let Hans paint the house or recurse onto another group of children.
Similarly, repeating \textexample{em Hans} and \textexample{h\"alfe} would correspond to a number of Hanses, each either helping another Hans or helping paint the house.
Then, using the pumping lemma for \emph{context-free} languages, it can be shown that the expressions of the form in \cref{eq:context-sensitive-expression} cannot be recognized by any context-free grammar.
We refer the readers to \citet[Example 7.20]{hopcroft79} for detailed proof.

\cref{ex:human-language-noncf} means that to model a human language formally, we need more expressive formalisms than context-free grammars or pushdown automata as described in the previous sections.\footnote{On the other hand, note that we would ideally also like to upper-bound the expressive power of the formal models, as this introduces useful inductive biases for learning and sparks insights into how humans process language.
    This means that we would not simply like to jump to Turing-complete models in such an exploration of language models.}
However, instead of defining a more expressive formalism motivated by formal language theory (like we did with context-free grammars and center embeddings), we now introduce recurrent neural networks, which, as we will see, under certain assumptions, have the capacity to model all computable languages (i.e., they are Turing complete).
Moreover, they can also model \emph{infinite} lengths of the context $\strlt$ in a very flexible way.
In the next section, we define them formally.

\subsection{Recurrent Neural Networks} \label{sec:rnns}
As discussed, natural languages are beyond the descriptive power of regular and context-free languages.
Now, we turn to a class of models that is theoretically capable of recognizing all computable languages: \defn{recurrent neural networks} (RNNs)\index{recurrent neural network}\index{RNN}.\footnote{In this subsection, we focus on the applications of recurrent neural networks to language modeling.
    However, recurrent neural networks have been widely used to process sequential data and time series, thanks to their power of taking in arbitrary-length inputs.}

\subsubsection{An Informal Introduction to Recurrent Neural Networks}
Human language is inherently sequential: we produce and consume both spoken as well as written language as a stream of units.\footnote{So far, we have simply referred to those units as \emph{symbols}.}
This structure is reflected in some of the algorithms for processing language we have seen so far.
For example, finite-state automata (cf. \cref{sec:finite-state}) process the input string one symbol at a time and build the representations of the string seen so far in the current state of the automaton.
Pushdown automata function similarly, but additionally keep the stack as part of the configuration.


Recurrent neural networks are neural networks that capture the same idea of iterative processing of the input but do so in a more flexible way than the finite-memory finite-state automata and the stack-based pushdown automata.
Very abstractly, a recurrent neural network sequentially processes a sequence of inputs and, while doing so, produces a sequence of \defn{hidden states}\index{hidden state}, which we will denote as $\hiddState$, based on a transition function in form of a \defn{recurrent dynamics map}\index{dynamics map}, which acts similarly to a (deterministic) transition function in a finite-state machine: given the current hidden state and an input symbol, it (deterministically) determines the next hidden state.\anej{for our exploration of non-deterministic RNNs, we should rethink this...}
The hidden states play, as we will see, an analogous role to the states of a finite-state automaton or the configuration of a pushdown automaton: The current hidden state of a recurrent neural network at time $\tstep$ determines, together with the input at time $\tstep$, through the dynamics map, the hidden state at time $\tstep + 1$---indeed, very similar to how finite-state automata process strings and transition between their states.
Again, the hidden state can be thought of as a compact (constant-size) summary of the input $\str_{\leq \tstep}$ seen so far and should ideally characterize $\str_{\leq \tstep}$ as well as possible (in the sense of retaining all information required for continuing the string).
Remember from \cref{sec:human-language-not-finite-stete} that the finite number of states of a finite-state automaton presented a serious limitation to its ability to model human language.
As we will see, the main difference between (weighted) finite-state automata and RNNs is that the latter can work with \emph{infinite} state spaces, for example, $\R^\hiddDim$ in the abstract formulation, or $\Q^\hiddDim$ in a digital computing system, such as a computer.
This, together with the flexibility of the transition function between hidden states, will allow RNNs to represent more complex languages than those recognized by finite-state automata or context-free grammars.
In fact, the large state space and the flexible transition functions endow RNNs, under some assumptions, with the possibility to model infinitely-long-term dependencies on the input string, distinguishing them from the Markovian \ngram{} models.

You might wonder why we refer to the current state of an RNN as \emph{hidden} states instead of only as \emph{states}, as with finite-state automata.
Indeed, when analyzing recurrent neural networks in terms of their expressivity and connections to classical models of computation, we will regard the hidden states as completely analogous to states in a finite-state or pushdown automaton.
The hidden part comes from the fact that the hidden states $\hiddState$ are usually not what we are interested in when modeling language with an RNN.
Rather, $\hiddState$ is simply seen as a component in a system that produces individual \emph{conditional probabilities} over the next symbol, as in sequence models (cf. \cref{def:sequence-model})---these conditional probabilities are the actual ``visible'' parts, while the ``internal'' states are, therefore, referred to as hidden.

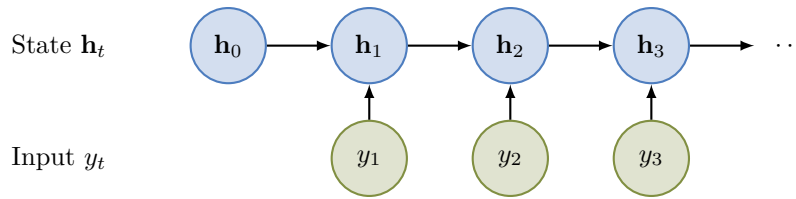
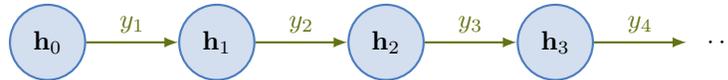
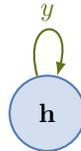
\begin{figure}
    \centering
    \begin{subfigure}{\textwidth}
        \centering
        \begin{tikzpicture}[x=1.5cm, y=1.5cm,>=latex]

            \foreach \i in {1,...,3}
            \node[circle, draw=ETHGreen!80!white, thick, fill=ETHGreen!20, inner sep=0pt, minimum size=10mm] (y\i) at (1.25*\i, 0) {$\sym_\i$};

            \foreach \i in {0,...,3}
            \node[circle, draw=ETHBlue!80!white, thick, fill=ETHBlue!20, inner sep=0pt, minimum size=10mm] (H\i) at (1.25*\i, 1) {$\hiddState_\i$};

            \node[circle, draw=none, minimum size=10mm] (Hminusone) at (-0.9, 1) {};
            \node[circle, draw=none, minimum size=10mm] (Hend) at (5, 1) {$\cdots$};

            \foreach \i in {1,...,3}
                {
                    \draw[->, thick] (y\i) -- (H\i);
                }
            \draw[->, thick] (H0) -- (H1);
            \draw[->, thick] (H1) -- (H2);
            \draw[->, thick] (H2) -- (H3);
            \draw[->, thick] (H3) -- (Hend);

            \node at (-1.5, 1) {State $\hiddState_\tstep$};
            \node at (-1.5, 0) {Input $\sym_\tstep$};

        \end{tikzpicture}
        \caption{An abstract depiction of how an RNN processes one symbol in a string.
            The hidden state $\hiddStatet$ summarizes the inputs $\sym_1\sym_2\ldots\symt$.
        }
        \label{fig:rnn-abstract}
    \end{subfigure}

    \vspace{6mm}

    \begin{subfigure}{\textwidth}
        \centering
        \begin{tikzpicture}[x=1.5cm, y=1.5cm,>=latex]

            \foreach \i in {0,...,3}
            \node[circle, draw=ETHBlue!80!white, thick, fill=ETHBlue!20, inner sep=0pt, minimum size=10mm] (H\i) at (1.5*\i, 1) {$\hiddState_\i$};

            \node[circle, draw=none, minimum size=10mm] (Hminusone) at (-0.9, 1) {};
            \node[circle, draw=none, minimum size=10mm] (Hend) at (6, 1) {$\cdots$};

            \draw[->, thick, color=ETHGreen] (H0) to node[above] {$\sym_1$} (H1);
            \draw[->, thick, color=ETHGreen] (H1) to node[above] {$\sym_2$} (H2);
            \draw[->, thick, color=ETHGreen] (H2) to node[above] {$\sym_3$} (H3);
            \draw[->, thick, color=ETHGreen] (H3) to node[above] {$\sym_4$} (Hend);

        \end{tikzpicture}
        \caption{An abstract depiction of an RNN as an automaton. The transitions between the possibly infinitely-many hidden states are determined by the dynamics map.}
        \label{fig:rnn-abstract-automaton}
    \end{subfigure}

    \vspace{2mm}

    \begin{subfigure}{\textwidth}
        \centering
        \begin{tikzpicture}[x=1.5cm, y=1.5cm,>=latex]

            \node[circle, draw=ETHBlue!80!white, thick, fill=ETHBlue!20, inner sep=0pt, minimum size=10mm] (Ht) at (0, 0) {$\hiddState$};

            \path[->, thick, color=ETHGreen, every loop/.style={looseness=8}] (Ht) edge [loop above] node {$\sym$} (Ht);

        \end{tikzpicture}
        \caption{An abstract depiction of an RNN as a system updating the hidden state $\hiddState$ depending on the input $\sym$.}
        \label{fig:rnn-abstract-circle}
    \end{subfigure}
    \caption{Different possible depictions of an abstract RNN model.
        The way that the hidden states are updated based on the input symbol $\sym_\tstep$ is abstracted away.}
    \label{fig:rnn-abstract-whole}
\end{figure}

RNNs are abstractly illustrated in different ways in the literature.
Often, they are represented as a sequence of hidden states and the input symbols consumed to arrive at those states---this is shown in \cref{fig:rnn-abstract}.
They can also be presented more similarly to automata, with (a possibly infinite) labeled graph, where the transition labels again correspond to the symbols used to enter the individual states.
This is presented in \cref{fig:rnn-abstract-automaton}.
Lastly, due to the infinite state space, one can also think of an RNN as a system that keeps the most current hidden state in memory and updates it as new symbols are consumed---this is shown in \cref{fig:rnn-abstract-circle}.\footnote{More precisely, these illustrations correspond to \emph{first-order} RNNs, which are by far the most common. Later, we will also briefly consider higher-order RNNs, whose hidden state update depends on multiple previous hidden states.}\anej{We have to agree on t vs t-1. I think the standard way to describe RNNs in practice is $h_t = f(h_{t -1}, y_t)$ and not $y_{t-1}$.}\tianyu{$\vht$ has the information of $\symt$, and $\str = \symone\cdots$ \response{Anej} I agree but Ryan wrote it differently in the lecture :)}\tianyu{I removed the arrow pointing into h0.}
\franz{Maybe calling $y_t$ the output in a) is more intuitive}





\subsubsection{A Formal Definition of Recurrent Neural Networks}
Having introduced RNNs and their motivations informally, we now move to their formal definition.
Our definition and treatment of recurrent neural networks might differ slightly from what you might normally encounter in the literature.
Namely, we define RNNs below as abstract systems transitioning between possibly infinitely-many states.
Our definition will allow for an intuitive connection to classical language models such as finite-state and pushdown language models as well as for tractable theoretical analysis in some special cases.
Specifically, when analyzing RNNs theoretically, we will make use of their connections to automata we saw in \cref{chapter:classical-lms}.

In an abstract sense, recurrent neural networks can be defined as a system transitioning between possibly infinitely-many states, which we will assume to be vectors in a vector space.
Specifically, we will distinguish between \emph{real} and \emph{rational} recurrent neural networks.
\begin{definition}{Real-valued Recurrent Neural Network}{rnn}
    Let $\alphabet$ be an alphabet.
    A (deterministic) \defn{real-valued recurrent neural network}\index{real-valued recurrent neural network} $\rnn$ is a four-tuple $\generalrnntuple$ where
    \begin{itemize}
        \item $\alphabet$ is the alphabet of input symbols;
        \item $\hiddDim$ is the dimension of $\rnn$;
        \item $\dyMap\colon \R^\hiddDim \times \alphabet \to \R^\hiddDim$ is the dynamics map, i.e., a function defining the transitions between subsequent states;
        \item $\hiddStateZero \in \R^\hiddDim$ is the \defn{initial state}\index{initial state}.
    \end{itemize}
\end{definition}
We analogously define \defn{rational-valued recurrent neural networks}\index{rational-valued recurrent neural network} as recurrent neural networks with the hidden state space $\Q^\hiddDim$ instead of $\R^\hiddDim$.
You might wonder why we make the distinction.
Soon, when we take on theoretical analysis of RNNs, it will become important over which state spaces the models are defined.
RNNs implemented in a computer using floating-point numbers, of course, cannot have irrational-valued weights---in this sense, all implemented recurrent neural networks are rational.
However, defining the models over the real numbers crucially allows us to perform operations from calculus for which some sort of continuity and smoothness is required, for example, differentiation for gradient-based learning (cf. \cref{sec:param-estimation}).

\begin{example}{A rational-valued RNN}{example-rnns}
    An example of a rational-valued RNN is the series
    \begin{equation}
        h_\tstep = \frac{1}{2}h_{\tstep - 1} + \frac{1}{h_{\tstep - 1}}
    \end{equation}
    which we considered in \cref{ex:sqrt2}.
    In this case
    \begin{itemize}
        \item $\alphabet = \set{\syma}$
        \item $\hiddDim = 1$
        \item $\dyMap\colon \left(x, \syma\right) \mapsto \frac{1}{2}x + \frac{1}{x}$
        \item $\hiddStateZero = 2$
    \end{itemize}
\end{example}
\begin{example}{Another example of an RNN}{}
    The tuple $\rnn = \generalrnntuple$ where
    \begin{itemize}
        \item $\alphabet = \set{\syma, \symb}$
        \item $\hiddDim = 2$
        \item $\dyMap\colon \left(\vx, \sym\right) \mapsto \begin{cases}
                      \begin{pmatrix}
                          \cos\phi  & -\sin\phi \\
                          \sin \phi & \cos \phi
                      \end{pmatrix} \vx & \textbf{ if } $\sym = \syma$ \\
                      \begin{pmatrix}
                          \cos\psi  & -\sin\psi \\
                          \sin \psi & \cos \psi
                      \end{pmatrix} \vx & \textbf{ otherwise}
                  \end{cases}$
        \item $\hiddStateZero = \begin{pmatrix}
                      1 \\ 1
                  \end{pmatrix}$
    \end{itemize}
    is an example of a real-valued RNN which rotates the current hidden state by the angle $\phi$ if the input symbol is $\syma$ and rotates it by $\psi$ if the symbol is $\symb$.
\end{example}
\begin{example}{Another example of an RNN}{}
    Another example of an RNN would be the tuple
    \begin{itemize}
        \item $\alphabet = \textnormal{GOOD} \cup \textnormal{BAD} = \set{\textexample{great}, \textexample{nice}, \textexample{good}} \cup \set{\textexample{awful}, \textexample{bad}, \textexample{abysmal}}$
        \item $\hiddDim = 2$
        \item $\dyMap\colon \left(\hiddState, \syma\right) \mapsto \begin{cases}
                      \hiddState + \begin{pmatrix}
                                       1 \\ 0
                                   \end{pmatrix} \ifcondition \syma \in \textnormal{GOOD} \\
                      \hiddState + \begin{pmatrix}
                                       0 \\ 1
                                   \end{pmatrix} \otherwisecondition
                  \end{cases}$
        \item $\hiddStateZero = \begin{pmatrix}
                      0 \\ 0
                  \end{pmatrix}$
    \end{itemize}
    which counts the number of occurrences of positive and negative words.
\end{example}
To define \emph{language models} using recurrent neural networks, we will use them as the encoder functions $\enc$ in our general language modeling framework (cf. \cref{sec:general-framework}).
To connect \cref{def:rnn} with the general LM framework, we define the RNN encoding function.
\begin{definition}{Recurrent Neural Encoding Function}{rnn-encoding-fn}
    Let $\rnn = \generalrnntuple$ be a recurrent neural network.
    A \defn{recurrent neural encoding function}\index{recurrent neural encoding function} $\encRNN$ is a representation function (cf. \cref{def:rep-function}) that recursively encodes strings of arbitrary lengths using its dynamics map $\dyMap$:
    \begin{equation}
        \encRNN\left(\strltplus\right) \defeq \dyMap(\encRNNfunc{\strlt}, \symt) \in \R^\hiddDim
    \end{equation}
    and
    \begin{equation}
        \encRNNfunc{\str_{<1}} \defeq \hiddStateZero \in \R^\hiddDim
    \end{equation}
\end{definition}
Intuitively, an RNN $\rnn$ takes an input string $\str$ and encodes it with the encoding function $\encRNN$ by sequentially applying its dynamics map $\dyMap$.
The representations of individual prefixes (cf. \cref{def:str-subelements}) of the input string are called hidden states.
\begin{definition}{Hidden State}{rnn-hidden-state}
    Let $\rnn = \generalrnntuple$ be an RNN.
    The hidden state $\hiddStatet \in \R^\hiddDim$ describes state of $\rnn$ after reading $\symt$.
    It is recursively computed according to the dynamics map $\dyMap$ as follows:
    \begin{equation} \label{eq:rnn-update-step}
        \vht \defeq \encRNNfunc{\strltplus} = \dyMap(\vhtminus, \symt)
    \end{equation}
\end{definition}




\begin{example}{Hidden states}{}
    The hidden states of the RNN from \cref{ex:example-rnns} are the individual values $h_\tstep$, which, as $\tstep$ increases, approach $\sqrt{2}$.
\end{example}

\subsubsection{Recurrent Neural Sequence Models}
\begin{figure}
    \centering
    \begin{subfigure}{\textwidth}
        \centering
        \begin{tikzpicture}[x=1.5cm, y=1.5cm,>=latex]

            \foreach \i in {1,...,2}
            \node[circle, draw=ETHGreen!80!white, thick, fill=ETHGreen!20, inner sep=0pt, minimum size=10mm] (y\i) at (2.25*\i, 0) {$\sym_\i$};

            \foreach \i in {0,...,2}
            \node[circle, draw=ETHBlue!80!white, thick, fill=ETHBlue!20, inner sep=0pt, minimum size=10mm] (H\i) at (2.25*\i, 1.3) {$\hiddState_\i$};

            \node[circle, draw=none, minimum size=10mm] (Hminusone) at (-0.9, 1.3) {};
            \node[circle, draw=none, minimum size=10mm] (Hend) at (5.5, 1.3) {$\cdots$};
            \node[circle, draw=none, minimum size=10mm] (yend) at (5.5, 0) {$\cdots$};

            \foreach \i in {1,...,2}
                {
                    \draw[->, thick] (y\i) -- (H\i);
                }
            \draw[->, thick, color=ETHGreen, dotted] (H0) to node[above, sloped] {\scriptsize $\sym_1 \sim \pLNSM\left(\cdot \mid \hiddState_0\right)$} (y1);
            \draw[->, thick, color=ETHGreen, dotted] (H1) to node[above, sloped] {\scriptsize $\sym_2 \sim \pLNSM\left(\cdot \mid \hiddState_1\right)$} (y2);
            \draw[->, thick] (H0) -- (H1);
            \draw[->, thick] (H1) -- (H2);
            \draw[->, thick] (H2) -- (Hend);

            \node at (-1.5, 1.3) {State $\hiddState_\tstep$};
            \node at (-1.5, 0) {Input $\sym_\tstep$};

        \end{tikzpicture}
        \caption{An abstract depiction of how an RNN generates a string one symbol at a time.
            The hidden state $\hiddStatet$ summarizes the string $\sym_1\sym_2\ldots\sym_\tstep$ generated so far.
            The dotted lines denote the sampling steps.}
        \label{fig:rnn-abstract-generative}
    \end{subfigure}

    \vspace{6mm}

    \begin{subfigure}{\textwidth}
        \centering
        \begin{tikzpicture}[x=1.5cm, y=1.5cm,>=latex]

            \foreach \i in {0,...,2}
            \node[circle, draw=ETHBlue!80!white, thick, fill=ETHBlue!20, inner sep=0pt, minimum size=10mm] (H\i) at (2.25*\i, 1) {$\hiddState_\i$};

            \node[circle, draw=none, minimum size=10mm] (Hminusone) at (-0.9, 1) {};
            \node[circle, draw=none, minimum size=10mm] (Hend) at (6, 1) {$\cdots$};

            \draw[->, thick, color=ETHGreen] (H0) to node[above] {\scriptsize $\sym_1 \sim \pLNSM\left(\cdot \mid \hiddState_0\right)$} (H1);
            \draw[->, thick, color=ETHGreen] (H1) to node[above] {\scriptsize $\sym_2 \sim \pLNSM\left(\cdot \mid \hiddState_1\right)$} (H2);
            \draw[->, thick, color=ETHGreen] (H2) to node[above] {} (Hend);

        \end{tikzpicture}
        \caption{An abstract depiction of a generative RNN as an automaton.}
        \label{fig:rnn-abstract-automaton-generative}
    \end{subfigure}

    \vspace{2mm}

    \begin{subfigure}{\textwidth}
        \centering
        \begin{tikzpicture}[x=1.5cm, y=1.5cm,>=latex]

            \node[circle, draw=ETHBlue!80!white, thick, fill=ETHBlue!20, inner sep=0pt, minimum size=10mm] (Ht) at (0, 0) {$\hiddStatet$};

            \path[->, thick, color=ETHGreen, every loop/.style={looseness=8}] (Ht) edge [loop above] node {$\sym_\tstep \sim \pLNSM\left(\cdot \mid \hiddState_\tstep\right)$} (Ht);

        \end{tikzpicture}
        \caption{An abstract depiction of an RNN as a system updating the hidden state $\hiddStatet$ depending on the generated symbol $\symt$.}
        \label{fig:rnn-abstract-circle-generative}
    \end{subfigure}
    \caption{Different possible depictions of an abstract RNN model \emph{generating} symbols.
        The way that the hidden states are updated based on the input symbol $\sym_\tstep$ is abstracted away.}
    \label{fig:rnn-abstract-whole-generative}
\end{figure}
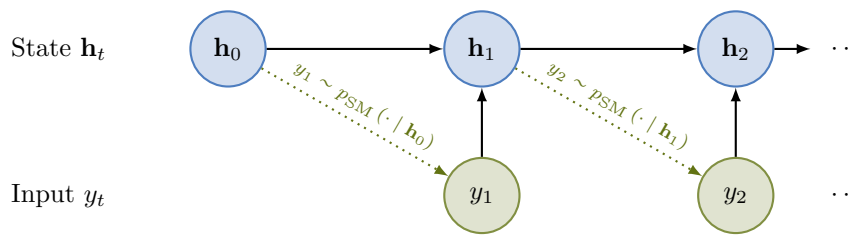
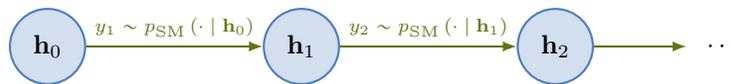
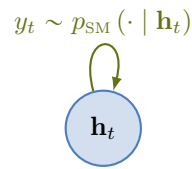

A recurrent neural network based on \cref{def:rnn} on its own does not yet define a sequence model, but simply a \emph{context encoding function} $\encRNN \colon \kleene{\eosalphabet} \to \R^\hiddDim$.
To define a sequence model based on an RNN, we simply plug in the RNN encoding function \cref{def:rnn-encoding-fn} into the General language modeling framework from \cref{sec:general-framework}.
\begin{definition}{Recurrent neural sequence model}{recurrent-neural-language-model}
    Let $\rnn = \generalrnntuple$ be a recurrent neural network and $\outMtx \in \R^{|\eosalphabet| \times \hiddDim}$ a symbol representation matrix.
    A $\hiddDim$-dimensional \defn{recurrent neural sequence model}\index{recurrent neural sequence model} over an alphabet $\alphabet$ is a tuple $\rnntuple$ defining the sequence model of the form
    \begin{equation} \label{eq:rnn-sequence-model}
        \pLNSM\left(\symt \mid \strlt\right) \defeq \projfuncEosalphabetminusFunc{\embedMtx \, \encRNNfunc{\strlt}}_{\symt} = \projfuncEosalphabetminusFunc{\embedMtx \, \vhtminus}_{\symt}.
    \end{equation}
    By far the most common choice of the projection function is the $\softmax{}$ yielding the sequence model
    \begin{equation} \label{eq:rnn-sequence-model-softmax}
        \pLNSM\left(\symt \mid \strlt\right) \defeq \softmaxfunc{\embedMtx \, \encRNNfunc{\strlt}}{\symt} = \softmaxfunc{\embedMtx \, \vhtminus}{\symt}.
    \end{equation}
    For conciseness, we will refer to RNN sequence models whose next-symbol probability distributions are computed using the $\softmax$ function as \defn{$\softmax$ RNN sequence models}.
\end{definition}
From this perspective, we see that RNNs are simply a special case of our general language modeling framework with parameterized representations of tokens $\sym \in \eosalphabet$ and the history $\str \in \kleene{\alphabet}$ (cf. \cref{sec:general-framework})---an RNN simply defines how the encoding function $\enc$ is specified.
The three figures from \cref{fig:rnn-abstract-whole} are presented again with this probabilistic perspective in \cref{fig:rnn-abstract-whole-generative}.

\subsubsection{A Few More Definitions}

In the following, we will often use the so-called one-hot encodings of symbols for concise notation.
We define them here.
\begin{definition}{One-hot encoding}{one-hot}
    Let $\alphabet$ be an alphabet and $\ordering: \alphabet \to \left\{1, \ldots, |\alphabet|\right\}$ a bijection (i.e., an ordering of the alphabet, assigning an index to each symbol in $\alphabet$).
    A \defn{one-hot encoding}\index{one-hot encoding} $\onehot{\cdot}$ is a representation function of the symbols in $\alphabet$ which assigns the symbol $\sym \in \alphabet$ the $\ordering\!\left(\sym\right)^\text{th}$ basis vector:
    \begin{equation}
        \onehot{\sym} \defeq \vd_{\ordering\left(\sym\right)},
    \end{equation}
    where here $\vd_{\idx}$ is the $\idx^{\text{th}}$ canonical basis vector, i.e., a vector of zeros with a $1$ at position $\idx$. 
\end{definition}

\begin{example}{One-hot encoding}{}
    Let $\alphabet = \{ \textexample{large}, \textexample{language}, \textexample{models} \}$ and $\ordering = \{ \textexample{large}\colon 1, \textexample{language}\colon 2, \textexample{models}\colon 3   \}$.
    The one-hot encoding of the vocabulary is:
    \begin{equation} \textstyle
        \onehot{\textexample{large}} = \begin{pmatrix}1\\0\\0\end{pmatrix},
        \onehot{\textexample{language}} = \begin{pmatrix}0\\1\\0\end{pmatrix},
        \onehot{\textexample{models}} = \begin{pmatrix}0\\0\\1\end{pmatrix}
    \end{equation}
\end{example}

Many specific variants of recurrent neural networks define the dynamics map $\dyMap$ in a specific way: the output of the function is some element-wise (non-linear) transformation of some ``inner'' function $\vfuncg$.
The dynamics map of such an RNN is then the composition of $\vfuncg$ and the non-linearity.
\begin{definition}{Activation function}{activation-function}
    Let $\rnn = \rnntuple$ be an RNN.
    If the hidden states $\hiddStatet$ of the RNN are computed as
    \begin{equation}
        \hiddStatet = \sigmoid\!\left(\vfuncg\!\left(\hiddState_{\tstep - 1}, \sym\right)\right)
    \end{equation}
    for some function $\vfuncg\colon \R^\hiddDim \times \alphabet \to \R^\hiddDim$ and some function $\sigmoid\colon \R \to \R$ which is computed \emph{element-wise} (that is, $\sigmoid\!\left(\vx\right)_\idxd = \sigmoid\!\left(\evx_\idxd\right)$ for all $\idxd = 1, \ldots, \hiddDim$ and $\vx \in \R^\hiddDim$), we call $\sigmoid$ an \defn{activation function}\index{activation function}.
\end{definition}

This finishes our formal definition of recurrent neural networks.
We next consider some of their theoretical properties, starting with tightness.


\subsection{General Results on Tightness} \label{sec:rnn-tightness}
We now discuss a general result on the tightness of recurrent neural sequence models, as defined in \Cref{def:recurrent-neural-language-model}.
The analysis is straightforward and is a translation of the generic results on tightness (cf. \cref{sec:rep-based-tightness}) to the case of the norm of the hidden states of an RNN, $\hiddStatet$ as the encodings of the prefixes $\str_{\leq \tstep}$, but it requires us to focus specifically on $\softmax$ RNN sequence models.
\begin{theorem}{Tightness of Recurrent Neural Sequence Model}{rnn-tightness}
    A $\softmax$ recurrent neural sequence model is tight if for all time steps $\tstep$ it holds that
    \begin{equation} \label{eq:rnn-tight-condition}
        s \normsubtwo{\vht} \leq \log \tstep,
    \end{equation}
    where $s \defeq \max_{\sym \in \alphabet} \normsubtwo{\embedSym - \embedEOS} $.
\end{theorem}
\begin{proof}
    This is simply a restatement of \cref{thm:leo-rep-tightness} for the case when $\enc$ takes the form of a general RNN encoding function, $\encRNN$.
\end{proof}

\anej{do we talk about activation functions or general dynamics maps here?}
\begin{corollary}{RNNs with bounded dynamics maps are tight}{bounded-rnn-tight}
    A $\softmax$ recurrent neural sequence model $\rnn = \generalrnntuple$ with a bounded dynamics map $\dyMap$, i.e, with a dynamics map $\dyMap$ such that
    \begin{equation}
        |\dyMap\!\left(\vx\right)_{\idxd}| \leq \constbound
    \end{equation}
    for some $\constbound \in \R$, for all $\idxd = 1, \ldots, \hiddDim$ and all $\vx \in \R^\hiddDim$, is tight.
\end{corollary}
\begin{proof}
    If the dynamics map is bounded, the norm of the hidden state, $\norm{\vht}_2$, is bounded as well.
    This means that the left-hand-side of \cref{eq:rnn-tight-condition} is \emph{constant} with respect to $\tstep$ and the condition holds trivially.
\end{proof}
A special case of \cref{cor:bounded-rnn-tight} is RNNs with bounded \emph{activation} functions (cf. \cref{def:activation-function}).
Those are tight if the activation function itself is bounded.
This implies that all standard sigmoid and $\tanh$ activated recurrent neural networks are tight.
However, the same does not hold for RNNs with unbounded activation functions, which have lately been more popular
(one of the reasons for this is the vanishing gradient problem
\citep{Rectifier-Bengio}).

\anej{Add complete derivation}
\begin{example}{RNNs with unbounded activation functions may not be tight}{nontight-rnn-example}
    A very popular unbounded activation function is the so-called \defn{rectified linear unit}\index{rectified linear unit}\index{ReLU} ($\ReLU$), defined as
    \begin{equation}
        \ReLU\left(x\right) \defeq \max\left(0, x\right).
    \end{equation}
    This function is clearly unbounded.

    Now suppose we had the following RNN over the simple alphabet $\alphabet = \set{\syma}$.
    \begin{equation} \label{eq:nontight-rnn-example}
        \hiddState_\tstep =
        \hiddState_{\tstep - 1} + \begin{pmatrix}
            1
        \end{pmatrix},
    \end{equation}
    initial state
    \begin{equation}
        \hiddStateZero = \begin{pmatrix}
            0
        \end{pmatrix}
    \end{equation}
    and the output matrix
    \begin{equation}
        \outMtx = \begin{pmatrix}
            -1 \\
            1
        \end{pmatrix}
    \end{equation}
    where the top row of $\outMtx$ computes the logit of the $\eos$ symbol and the bottom one the one of $\syma$.
    It is easy to see that
    \begin{equation}
        \hiddStatet = \begin{pmatrix}
            \tstep
        \end{pmatrix}.
    \end{equation}
    This already does not look promising for tightness---the norm of the hidden state, which is, in this case, $\norm{\begin{pmatrix} \tstep \end{pmatrix}} = \tstep$, and is, therefore, increasing at a much higher rate than $\bigo\left(\log \tstep\right)$ required by \cref{thm:rnn-tightness}.
    We encounter a similar hint against tightness if we compute the conditional probabilities of the $\eos$ symbol and the symbol $\syma$.
    \begin{align}
        \pLNSM\left(\eos \mid \str_{<\tstep}\right) & = \softmaxfunc{\begin{pmatrix}
                                                                             -1 \\
                                                                             1
                                                                         \end{pmatrix} \begin{pmatrix}
                                                                                           \tstep - 1
                                                                                       \end{pmatrix}}{\eos} = \frac{\exp\left[-\tstep + 1\right]}{\exp\left[-\tstep + 1\right] + \exp\left[\tstep - 1\right]} \\
        \pLNSM\left(\eos \mid \str_{<\tstep}\right) & = \softmaxfunc{\begin{pmatrix}
                                                                             -1 \\
                                                                             1
                                                                         \end{pmatrix} \begin{pmatrix}
                                                                                           \tstep - 1
                                                                                       \end{pmatrix}}{\syma} = \frac{\exp\left[\tstep - 1\right]}{\exp\left[-\tstep + 1\right] + \exp\left[\tstep - 1\right]}
    \end{align}
    The probability of ending the string at time step $\tstep$ is, therefore
    \begin{equation} \label{eq:non-tight-rnn-eos-prob}
        \pLNSM\left(\eos \mid \str_{<\tstep}\right) = \frac{1}{1 + \exp\left[2\left(\tstep - 1\right)\right]}.
    \end{equation}
    Intuitively, this means that the probability of ending the string (generating $\eos$) diminishes rapidly with $\tstep$---in this case much faster than any diverging sum required by \cref{thm:lm-tight-main}.
    All signs, thus, point towards the RNN from \cref{eq:nontight-rnn-example} not being tight.
    Indeed, for this specific case, one can show using some algebraic manipulations that
    \begin{equation}
        \sum_{\str \in \kleene{\alphabet}} \pLN\left(\str\right) = \sum_{\idx \in \Nzero} \pLN\left(\syma^\idx\right) < 0.15
    \end{equation}
    where $\pLN$ is the locally normalized model induced by the RNN.
    This means that the RNN from \cref{eq:nontight-rnn-example} assigns less than $0.15$ probability to finite strings---all other probability mass leaks to infinite sequences.
\end{example}

\begin{example}{RNNs with unbounded activation functions can still be tight}{unbounded-tight-rnn}
    \cref{ex:nontight-rnn-example} showed that RNNs with unbounded activation functions can indeed result in non-tight sequence models.
    However, this is not necessarily the case, as this simple modification of the RNN from \cref{ex:nontight-rnn-example} shows.
    The only aspect of the RNN that we modify is the output matrix $\outMtx$, which we change by flipping its rows:
    \begin{equation}
        \outMtx = \begin{pmatrix}
            1 \\
            -1
        \end{pmatrix}
    \end{equation}

    Now the probability of ending the string at time step $\tstep$ is
    \begin{equation} \label{eq:tight-rnn-eos-prob}
        \pLNSM\left(\eos \mid \str_{<\tstep}\right) = \frac{\exp\left[\tstep - 1\right]}{\exp\left[-\tstep + 1\right] + \exp\left[\tstep - 1\right]} = \frac{1}{\exp\left[-2\left(\tstep - 1\right)\right] + 1}.
    \end{equation}
    Compared to \cref{eq:non-tight-rnn-eos-prob}, the probability of $\eos$ in \cref{eq:tight-rnn-eos-prob} does not diminish.
    Indeed, since $\frac{1}{\exp\left[-2\left(\tstep - 1\right)\right] + 1} > \frac{1}{2}$ for all $\tstep$, the sum
    \begin{equation}
        \sum_{\tstep = 0}^\infty \pLNSM\left(\eos \mid \str_{<\tstep}\right)
    \end{equation}
    diverges, which, according to \cref{prop:div-implies-tight} implies that the sequence model is tight.
\end{example}

\subsection{Elman and Jordan Networks} \label{sec:elman-jordan-rnns}
The characterization of dynamics maps we gave in \cref{def:recurrent-neural-language-model} allows for $\dyMap$ to be an arbitrary mapping from the previous state and the current input symbol to the new state.
In this section, we introduce two seminal and particularly simple parameterizations of this map---the simplest recurrent neural sequence models.
We term them Elman sequence models and Jordan sequence models, as each is inspired by architectures proposed by \citet{elman1990} and \citet{Jordan1986}, respectively.
The definitions we present here are slightly different than those found in the original works---most notably, both Elman and Jordan networks were originally defined for \emph{transduction} (mapping an input string to an output string, as with translation) rather than language modeling.

Put simply, these two models \emph{restrict} the form of the dynamics map $\dyMap$ in the definition of an RNN (cf. \cref{def:rnn}).
They define particularly simple relationships between the subsequent hidden states, which are composed of affine transformations of the previous hidden state and the representation of the current input symbol passed through a non-linear activation function (cf. \cref{def:activation-function}).
The affine transformations are performed by different matrices and bias vectors---the parameters of the model (cf. \cref{as:hypspace})---each transforming a separate part of the input to the dynamics map.

\begin{definition}{Elman Sequence Model \citep{elman1990}}{elman-rnn}
    An \defn{Elman sequence model}\index{Elman sequence model} $\rnn = \elmanrnntuple$ is a $\hiddDim$-dimensional recurrent neural sequence model over an alphabet $\alphabet$ with the following dynamics map\looseness=-1
    \begin{equation} \label{eq:elman-update-rule}
        \vht = \sigmoid\left(\recMtx \vhtminus + \inMtx \inEmbedSymt + \biasVech \right).
    \end{equation}
    Here, $\inEmbedding{\cdot}\colon \alphabet \to \R^\embedDim$ is the input symbol \defn{embedding function}\index{embedding function} which represents each symbol $\sym \in \alphabet$ as a $\embedDim$-dimensional vector and $\sigmoid$ is an element-wise non-linearity.\footnote{The symbol representations $\inEmbedding{\sym}$ are often also referred to as \defn{static symbol embeddings} because they do not depend on the string surrounding or preceding $\sym$. Here, we treat them as any other parameters of the model which can be learned using gradient-based learning (cf. \cref{sec:learning}). However, note that learning good static embeddings was a very active field before the emergence of large end-to-end systems we see today. Very popular examples include Word2Vec~\citep{mikolov2013efficient}, GloVe~\citep{pennington-etal-2014-glove}, and FastText~\citep{bojanowski-etal-2017-enriching}.}
    $\biasVech \in \R^{\hiddDim}$, $\recMtx \in \R^{\hiddDim \times \hiddDim}$, and $\inMtx \in \R^{\hiddDim \times \embedDim}$.
\end{definition}
Due to its simplicity, the Elman RNN is also known as the \emph{vanilla RNN} variant, emphasizing it is one of the most fundamental variants of the framework.

\paragraph{On the symbol representations.}
Notice that in \cref{eq:elman-update-rule}, the input symbols $\symt$ are first transformed into their vector representations $\inEmbedSymt$ and then additionally linearly transformed using the matrix $\inMtx$.
This results in an over-parametrized network---since the symbols are already embedded using the representation function $\inEmbedding{}$, the matrix $\inMtx$ is theoretically superfluous and could be replaced by the identity matrix.
However, the matrix $\inMtx$ could still be useful if the representations $\inEmbedSymt$ are fixed---in this case, the matrix can be used by the RNN to transform the representations during training to fit the training data better.
This is especially useful if the symbol representations $\inEmbedSymt$ already represent the input symbols in a compact representation space in which the parameters can be shared across different symbols.
Alternatively, we could represent the symbols using their one-hot encodings, i.e., $\inEmbedSymt = \onehot{\symt}$, in which case the columns of the matrix $\inMtx$ would correspond to the symbol representations (analogously to the representation matrix $\embedMtx$ from \cref{eq:softmax-sequence-model}).
However, notice that in this case, the representations on the symbols do not share any parameters, and each column of the matrix is therefore an unconstrained vector.
Such matrix-lookup-based \emph{input} symbol representations from \cref{eq:elman-update-rule} are sometimes \defn{tied}\index{embedding tying}, i.e., $\inEmbedding{\cdot} = \embedding{\cdot}$, with the \emph{output} symbol representations from the embedding matrix $\embedMtx$ in the definition of the sequence model induced by an RNN (cf. \cref{def:locally-normalized-rep-based,eq:rnn-sequence-model}).

However, embedding tying is non-essential to representation-based LMs.
The input symbol embedding function can always be chosen independently with the output symbol embedding function \cref{def:output-symbol-embedding-function}.

The Jordan network is somewhat different in that it feeds the \emph{output} logits computed through the output matrix $\outMtx$ into the computation of the next state, and not directly the hidden state.
\begin{definition}{Jordan Sequence Model \citep{Jordan1986}}{jordan-rnn}
    A \defn{Jordan sequence model}\index{Jordan sequence model} is a $\hiddDim$-dimensional recurrent neural sequence model over an alphabet $\alphabet$ with the following dynamics map\looseness=-1
    \begin{align}
        \vht       & = \sigmoid\!\left(\recMtx \vr_{\tstep - 1} + \inMtx \inEmbedSymt + \biasVech \right)  \label{eqn:jordan-update-rule} \\
        \vr_\tstep & = \sigmoid_o\!\left(\outMtx \vht \right)
    \end{align}
    Again, $\inEmbedding{\cdot}\colon \alphabet \to \R^\embedDim$ is the input symbol embedding function which represents each symbol $\sym \in \alphabet$ as a $\embedDim$-dimensional vector while $\sigmoid$ and $\sigmoid_o$ are element-wise non-linearities.
    $\biasVech \in \R^{\hiddDim}$, $\recMtx \in \R^{\hiddDim \times \hiddDim}$, and $\inMtx \in \R^{\hiddDim \times \embedDim}$.
\end{definition}
Notice that the hidden state $\hiddStatet$ in \cref{eqn:jordan-update-rule} is not computed based on the previous hidden state $\hiddState_{\tstep - 1}$, but rather on the transformed outputs $\vr_{\tstep - 1}$---this is analogous to feeding back in the logits computed in \cref{eq:rnn-sequence-model} into the computation of $\hiddState$ rather than the previous hidden state.
The sequence model induced by a Jordan network is then directly induced by the logits $\vr_\tstep$ (i.e., the conditional probabilities are computed by putting $\rv_\tstep$ through the $\softmax$.

In both architectures, the activation function $\sigmoid$ can be any suitable element-wise function.
The canonical choices for it have been the sigmoid and $\tanh$ functions, however, a more common choice nowadays is the $\ReLU$ function or any of its more modern variants.\footnote{See \citet[\S\~6.3.1]{Goodfellow-et-al-2016} for an overview of modern activation functions used in neural networks.}

Since we will refer to the individual matrices defining the dynamics maps in Elman and Jordan networks quite a lot in the next subsections, we give them specific names.
The matrix $\recMtx$, which linearly transforms the previous hidden state (or the output) is the \defn{recurrence matrix}\index{recurrence matrix}.
The matrix $\inMtx$, which linearly transforms the representations of the input symbol, is called the \defn{input matrix}\index{input matrix}.
Lastly, the matrix which linearly transforms the hidden state before computing the output values $\vr_\tstep$ with an activation function is called the \defn{output matrix}\index{output matrix}.
$\biasVech$ is the hidden \defn{bias vector}\index{bias vector}.

\paragraph{Tightness of Elman and Jordan Recurrent Neural Networks}
As a simple corollary of \cref{cor:bounded-rnn-tight}, we can characterize the tightness of Elman and Jordan recurrent neural networks as follows.
\begin{corollary}{Tightness of simple RNNs}{}
    Elman and Jordan RNNs with a bounded activation function $\sigmoid$ and the $\softmax$ projection function are tight.
\end{corollary}

\newpage{}


\subsection{Variations on Recurrent Networks}\label{sec:var-rnn}

\newcommand{\hatvht}{{\color{MacroColor} \hat{\vh}_\tstep}}

\newcommand{\vit}{{\color{MacroColor} \boldsymbol{i}_t}}
\newcommand{\vft}{{\color{MacroColor} \boldsymbol{f}_t}}
\newcommand{\vvot}{{\color{MacroColor} \boldsymbol{o}_t}}
\newcommand{\vgt}{{\color{MacroColor} \boldsymbol{g}_t}}
\newcommand{\vct}{{\color{MacroColor} \boldsymbol{c}_t}}
\newcommand{\vctminus}{{\color{MacroColor} \boldsymbol{c}_{t-1}}}

\newcommand{\vrt}{{\color{MacroColor} \boldsymbol{r}_t}}
\newcommand{\vzt}{{\color{MacroColor} \boldsymbol{z}_t}}

\newcommand{\InpFunc}[1]{{\color{MacroColor} \boldsymbol{i}\!\left(#1\right)}}
\newcommand{\FgtFunc}[1]{{\color{MacroColor} \boldsymbol{f}\!\left(#1\right)}}
\newcommand{\CelFunc}[1]{{\color{MacroColor} \boldsymbol{g}\!\left(#1\right)}}
\newcommand{\OutFunc}[1]{{\color{MacroColor} \boldsymbol{o}\!\left(#1\right)}}
\newcommand{\CFunc}[1]{{\color{MacroColor} \boldsymbol{c}\!\left(#1\right)}}

\newcommand{\ResetFunc}[1]{{\color{MacroColor} \boldsymbol{r}\!\left(#1\right)}}
\newcommand{\UpdateFunc}[1]{{\color{MacroColor} \boldsymbol{z}\!\left(#1\right)}}

\newcommand{\iInpMtx}{{\color{MacroColor} \inMtx^{i}}}
\newcommand{\hInpMtx}{{\color{MacroColor} \recMtx^{i}}}
\newcommand{\InpBias}{{\color{MacroColor} \bias^{i}}}

\newcommand{\iResetMtx}{{\color{MacroColor} \inMtx^{r}}}
\newcommand{\hResetMtx}{{\color{MacroColor} \recMtx^{r}}}
\newcommand{\ResetBias}{{\color{MacroColor} \bias^{r}}}

\newcommand{\iUpdateMtx}{{\color{MacroColor} \inMtx^{z}}}
\newcommand{\hUpdateMtx}{{\color{MacroColor} \recMtx^{z}}}
\newcommand{\UpdateBias}{{\color{MacroColor} \bias^{z}}}

\newcommand{\iFgtMtx}{{\color{MacroColor} \inMtx^{f}}}
\newcommand{\hFgtMtx}{{\color{MacroColor} \recMtx^{f}}}
\newcommand{\FgtBias}{{\color{MacroColor} \bias^{f}}}

\newcommand{\iCelMtx}{{\color{MacroColor} \inMtx^{g}}}
\newcommand{\hCelMtx}{{\color{MacroColor} \recMtx^{g}}}
\newcommand{\CelBias}{{\color{MacroColor} \bias^{g}}}

\newcommand{\iOutMtx}{{\color{MacroColor} \inMtx^{o}}}
\newcommand{\hOutMtx}{{\color{MacroColor} \recMtx^{o}}}
\newcommand{\OutBias}{{\color{MacroColor} \bias^{o}}}

\newcommand{\losststep}{{\color{MacroColor} \ell}}

In the previous sections we introduced the two simplest RNN variants: the Elman \citep{elman1990}  and Jordan \citep{JORDAN1997471} networks.
Even though such simple RNNs in theory are all we need to model any computable language, empirically those architectures face many challenges.
One of the biggest are the vanishing and exploding gradient problems \citep{lstm}, which in practice is linked with the issue of learning long-term dependencies in language.

In this subsection, we expand our repertoire of RNN variants by going beyond the simple recurrent dynamics defined by the Elman and Jordan update rules.
To do so, we take a step back and return to \cref{def:rnn} of a recurrent neural network $\rnn$ as the tuple $\generalrnntuple$ .
We will define more elaborate dynamics maps $\dyMap$ which both aim to tackle some of the (empirically encountered) challenges of simpler variants as well as improve some theoretical aspects of the networks.
Importantly, keep in mind that the only aspect of the RNN we will strive to modify is the dynamics map---that is, the mapping from $\hiddStatetminus$ to $\hiddStatet$.
Given a hidden state, the definition of a sequence model will remain identical.

A common component of the more complex dynamics maps we explore in this section is the gating mechanism, which is why we start with it.





\subsubsection{Gating}
The update equations of Elman and Jordan RNNs define relatively simple transformations of the hidden states as an affine transformation of the previous hidden state and the new input, followed by some form of non-linearity.
In this sense, the interaction between the previous hidden state and the input symbol is relatively limited---the hidden state is transformed by the recurrence matrix $\recMtx$ at every time step invariant to the input symbol being read.
To see why this could be a limiting factor, consider the following example.
\anej{finish \response{Clemente} I have added the last sentence, I think it helps t make sense of the example}
\begin{example}{RNN Gates}{rnn-gates}
    Consider the language $\lang = \set{\syma^n \symb^n \symc^n x \syma^m \symb^m \symc^m\mid n, m \in \Nzero}$.
    It intuitively consists of two-part strings, where the two parts are separated by a symbol $\symx$.
    The part on the left side of $\symx$ contains a sequence of $n$ $\syma$'s followed by $n$ $\symb$'s, which is followed by $n$ $\symc$'s.
    The substring on the right side of $x$ contains a sequence of $m$ $\syma$'s which is again followed by $m$ $\symb$'s, and later by $m$ $\symc$'s.
    Both parts of the string can be arbitrarily long, and, intuitively, to correctly recognize a string in this language, a computational model has to keep the information about the number of $\syma$'s while reading in $\symb$'s to be able to ensure there is a correct number of $\symc$'s as well.
    This creates a long-term dependency across the entire block of $\symb$'s.
    However, notice that, after reading the symbol $x$, the information about the number of $\syma$'s becomes irrelevant to the recognition of the string: the model can, therefore, discard it and solely focus on modeling the rest of the string, which again requires keeping track of the number of the symbol occurrences.
    In other words: after a certain amount of time, previous information becomes irrelevant, and we may want to design a network that is able to select which information is important to \emph{keep around}.
\end{example}
To enable richer interaction between the transformation of the RNN hidden state and the input symbols, we introduce the gating mechanism.
Intuitively, the gating mechanism enables more fine-grained control over the transformations of the hidden state by ``selecting'' which aspects of the hidden state should be retained, which should be modified, and which should be deleted---in general, based on both the previous hidden state as well as the current input symbol.
Such transformations are defined using gates and gating functions.
\begin{definition}{Gate}{}
    A \defn{gate} is a real-valued vector $\gate \in \R^\hiddDim$, such that $\evg_\idxd \in [0,1]$ for all $\idxd \in \{1, \ldots, \hiddDim\}$.
    Gates are computed using \defn{gating functions}\index{gating functions}, i.e., functions whose outputs live in $\left[0, 1\right]^\hiddDim$.
\end{definition}
The fact that every dimension in a gate $\gatet$ takes a value between $0$ and $1$ invites a natural interpretation of the values as \emph{soft switches}, analogously to how switches are used in the electrical engineering context.
Intuitively, in the context of RNNs, where the information is passed around in the hidden states $\hiddStatet$, a gate of the same dimensionality as the hidden state can control which aspects (dimensions) of the hidden state should be forgotten (switched off) or and which ones retained (kept on)---a gate value close to $0$ can be interpreted as a signal that the information captured in the corresponding dimension of the hidden state should be ``forgotten'', and a gate value close to $1$ as the opposite.
Such modifications of the hidden state can be performed using an \emph{element-wise multiplication} of the hidden state $\hiddState$ and the gate $\gate$, which we denote with $\hiddState \odot \gate$.

Importantly, the gates can be computed based on the information about the string seen so far as well as the new input symbol---this means that the decision on what should be remembered and what should be forgotten can be made for each situation individually.
This allows RNN variants using gating to implement mechanisms to tackle challenges as the one described in \cref{ex:rnn-gates}.
Furthermore this not only enables RNNs to selectively keep information about the string, but also combat the vanishing and exploding gradient problems \citep[Appendix 2]{lstm}.
We next consider two of the best-known gated RNNs: Long Short-Term Memory and Gated Recurrent Unit networks .


\subsubsection{Long Short-term Memory Networks}
Long Short-term Memory Networks \citep[LSTM,][]{lstm} are perhaps the best-known type of a gated recurrent network.
They were introduced specifically to combat the vanishing gradient problem in the famous paper with more than $80\,000$ citations.
The somewhat unusual name comes from connections to human memory, in which short-term memory is characterized by evanescent neural activations, and long-term memory is based on the growth and structural change in neuron connections \citep{hebb1949organization}. 

\paragraph{The LSTM unit.}
LSTM RNNs are built from the LSTM units, which implement the RNN dynamics map and therefore perform the RNN update step.
To transform the hidden state $\hiddStatet$ at each time step, an LSTM network additionally keeps \emph{another} running summary of the string $\strlt$, on which the recurrent update depends---this is the so-called memory cell which we will denote by $\vct$.
Informally, one can think of the context as the information needed to decide on \emph{how} to transform the hidden state at each individual time step, depending on the input string.
The formal definition of the LSTM cell is the following.
\begin{definition}{Long Short-Term Memory}{lstm}
    A \defn{long short-term memory unit}\index{long short-term memory unit}\index{LSTM} is a recurrent neural network with the dynamics map defined through the following sequence of computations:
    \begin{align*} \label{eq:lstm-update-rule}
        \vit  & = \sigmoid \left( \hInpMtx\vhtminus + \iInpMtx \inEmbedding{\symt} + \InpBias  \right)  \tag{\text{input gate}} \\
        \vft  & = \sigmoid \left( \hFgtMtx\vhtminus + \iFgtMtx \inEmbedding{\symt} + \FgtBias \right) \tag{\text{forget gate}}  \\
        \vvot & = \sigmoid \left( \hOutMtx\vhtminus + \iOutMtx \inEmbedding{\symt} + \OutBias \right) \tag{\text{output gate}}  \\
        \vgt  & = \tanh (\hCelMtx\vhtminus + \iCelMtx \inEmbedding{\symt} + \CelBias ) \tag{\text{candidate vector}}            \\
        \vct  & = \vft \odot \vctminus + \vit \odot \vgt  \tag{\text{memory cell}}                                              \\
        \vht  & = \vvot \odot \tanh{(\vct)} \tag{\text{hidden state}}
    \end{align*}
    $\vit, \vft, \vot$ are the \defn{input}, \defn{forget}, and \defn{output} gates, $\vct$ is the \defn{memory cell} vector, and $\vgt$ is the \defn{candidate vector}.
    Here, $\sigmoid$ refers to the original sigmoid function.
\end{definition}
As we can see, the update rule of an LSTM network is considerably more complex than that of an Elman RNN.
It is also computationally more expensive, as it involves more matrix multiplications.
However, LSTMs have consistently shown improved performance compared to vanilla RNNs and are therefore considered together with GRUs the go-to choice for an RNN architecture \citep{Goodfellow-et-al-2016}. 
The theoretical reason of their success is that their gating mecahnism helps to reduce the Vanishing/Exploding gradient problem, and thus to learn long-term dependencies \citep[Appendix 2]{lstm}.

\begin{figure}
    \centering
    \newcommand{\empt}[2]{$#1_{#2}$}
\begin{tikzpicture}[
    font=\sf \scriptsize,
    >=LaTeX,
    cell/.style={
        rectangle, 
        rounded corners=5mm, 
        draw,
        very thick,
        },
    operator/.style={
        circle,
        draw,
        inner sep=-0.5pt,
        minimum height =.2cm,
        },
    function/.style={
        ellipse,
        draw,
        inner sep=1pt
        },
    ct/.style={
        circle,
        draw,
        line width = .75pt,
        minimum width=1cm,
        inner sep=1pt,
        },
    gt/.style={
        rectangle,
        draw,
        minimum width=4mm,
        minimum height=3mm,
        inner sep=1pt
        },
    mylabel/.style={
        font=\scriptsize\sffamily
        },
    ArrowC1/.style={
        rounded corners=.25cm,
        thick,
        },
    ArrowC2/.style={
        rounded corners=.5cm,
        thick,
        },
    ]

    \node [cell, minimum height =4cm, minimum width=6cm] at (0,0){} ;

    \node [gt, label={[mylabel]above: \hspace{-6 mm} \empt{f}{t}}] (ibox1) at (-2,-0.75) {$\sigma$};
    \node [gt] (ibox2) at (-1.5,-0.75) {$\sigma$};
    \node [gt, label={[mylabel]above:\quad\quad\quad\empt{g}{t}} ,minimum width=1cm] (ibox3) at (-0.5,-0.75) {Tanh};
    \node [gt] (ibox4) at (0.5,-0.75) {$\sigma$};

    \node [operator] (mux1) at (-2,1.5) {$\times$};
    \node [operator] (add1) at (-0.5,1.5) {+};
    \node [operator, label={[mylabel]left: \hspace{-12 mm}\empt{i}{t}}] (mux2) at (-0.5,0) {$\times$};
    \node [operator,label={[mylabel]left: \hspace{-12 mm}\empt{o}{t}} ] (mux3) at (1.5,0) {$\times$};
    \node [function] (func1) at (1.5,0.75) {Tanh};

    \node[ct, label={[mylabel]Cell}] (c) at (-4,1.5) {\empt{c}{t-1}};
    \node[ct, label={[mylabel]Hidden}] (h) at (-4,-1.5) {\empt{h}{t-1}};
    \node[ct, label={[mylabel]left:Input}] (x) at (-2.5,-3) {e'\empt{y}{t}};

    \node[ct ] (c2) at (4,1.5) {\empt{c}{t}};
    \node[ct ] (h2) at (4,-1.5) {\empt{h}{t}};
    \node[ct ] (x2) at (2.5,3) {\empt{h}{t}};

    \draw [ArrowC1] (c) -- (mux1) -- (add1) -- (c2);

    \draw [ArrowC2] (h) -| (ibox4);
    \draw [ArrowC1] (h -| ibox1)++(-0.5,0) -| (ibox1); 
    \draw [ArrowC1] (h -| ibox2)++(-0.5,0) -| (ibox2);
    \draw [ArrowC1] (h -| ibox3)++(-0.5,0) -| (ibox3);
    \draw [ArrowC1] (x) -- (x |- h)-| (ibox3);

    \draw [->, ArrowC2] (ibox1) -- (mux1);
    \draw [->, ArrowC2] (ibox2) |- (mux2);
    \draw [->, ArrowC2] (ibox3) -- (mux2);
    \draw [->, ArrowC2] (ibox4) |- (mux3);
    \draw [->, ArrowC2] (mux2) -- (add1);
    \draw [->, ArrowC1] (add1 -| func1)++(-0.5,0) -| (func1);
    \draw [->, ArrowC2] (func1) -- (mux3);

    \draw [-, ArrowC2] (mux3) |- (h2);
    \draw (c2 -| x2) ++(0,-0.1) coordinate (i1);
    \draw [-, ArrowC2] (h2 -| x2)++(-0.5,0) -| (i1);
    \draw [-, ArrowC2] (i1)++(0,0.2) -- (x2);

\end{tikzpicture}
    \caption{A pictorial depiction of the LSTM cell in action. The input $\vit$, forget $\vft$, and output $\vvot$ gates control which information of the input and of the previous hidden state is retained in the memory cell, and which information is passed to next the hidden state. }
    \label{fig:lstm-structure}
\end{figure}
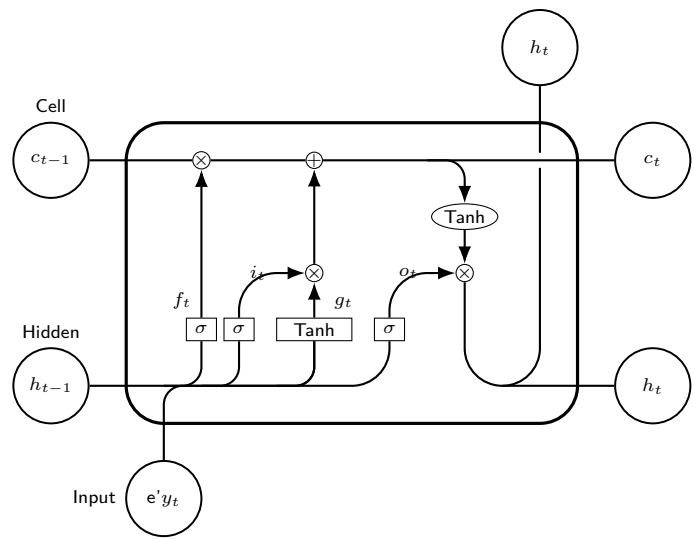

The names of the different quantities computed in \cref{def:lstm} reflect their intuitive interpretations.
The input, forget, and output vectors are all gates: they control the information which will be added to the memory cell based on the new input, the information which will be retained or forgotten from the previous memory cell, and the information which will be transferred from the memory cell to the hidden state, respectively.
Notice the identical nature in which all three gates are computed: they are non-linear transformations of affine transformations of the previous hidden state and the input symbol representations.
Their parameter matrices define the way in which the gates will influence the memorization, forgetting, and addition of information.
The additional information added to the memory cell in the form of the candidate vector $\vgt$ is computed similarly, with the only difference being the activation function.
This is the step that bears the most resemblance to the update step of the vanilla RNN (\cref{eq:elman-update-rule}).
However, compared to the latter, only parts of this transformation are kept (based on the input and forget vectors $\vit$ and $\vft$).
The memory cell $\vct$ then combines the old memory content $\vctminus$ with the newly integrated information in $\vgt$ to form the new memory content, which is then transformed using the $\tanh$ function and combined with the output gate to produce the hidden state $\hiddStatet$.
This is pictorially presented in \cref{fig:lstm-structure}.

As mentioned, the LSTM update step is noticeably more computationally complex than that of a vanilla RNN.
This has led to a line of work trying to combine the efficiency of vanilla RNNs and the empirical performance of gated RNNs.
In the next subsection, we consider Gated Recurrent Units, one of the best-known compromises found in this domain.

\subsubsection{Gated Recurrent Units}
The Gated Recurrent Unit \citep[GRU,][]{cho-etal-2014-learning,gru} provides a compromise between the simplicity of vanilla recurrent neural networks and the empirical success of being able to model long-term dependencies with LSTMs.
It defines a gated recurrent update unit that implements a simpler dynamics map by removing the memory component $\vct$ in the LSTM cell and combining the input and forget gates $\vit, \vft$ into one update gate.
These changes make GRU more memory efficient and easier to train than LSTM in practice.
The full GRU update step is defined as follows.
\begin{definition}{Gated Recurrent Units}{}
    A \defn{gated recurrent unit} defines a dynamics map in which a new hidden state is computed as:
    \begin{align*}
        \vrt & = \sigmoid \left( \hResetMtx\vhtminus + \iResetMtx \inEmbedding{\symt} + \ResetBias \right)  \tag{\text{reset gate}}                    \\
        \vzt & = \sigmoid \left( \hUpdateMtx\vhtminus + \iUpdateMtx \inEmbedding{\symt} + \UpdateBias \right) \tag{\text{update gate}}                 \\
        \vgt & = \tanh \left(\hCelMtx \left( \vrt\odot\vhtminus\right) + \iCelMtx \inEmbedding{\symt} + \CelBias \right) \tag{\text{candidate vector}} \\
        \vht & = (\one -\vzt) \odot \vgt + \vzt \odot \vhtminus
    \end{align*}
    $\vrt, \vzt$ are known as the \defn{reset} and \defn{update} gates, and $\vgt$ as the \defn{candidate} vector.
\end{definition}

Intuitively, the update gate works like a hot/cold water mixing valve: it is trained to find the \emph{optimum} blend of information of the candidate vector with that coming from the previous hidden state. The reset gate instead, can zero the information of the previous hidden state, when computing the candidate vector. This allows to \emph{forget} past information that becomes irrelevant, exactly like in the LSTM architecture.

\subsubsection{Parallelizability: The Achilles' Heel of Recurrent Neural Networks} \label{sec:rnn-parallelization}
It is easy to see from the definition of the RNN hidden state (cf. \cref{def:rnn-hidden-state}) that, to compute $\hiddStatet$, we have to compute $\hiddState_{\tstep'}$ for all $\tstep' < \tstep$ first.
Another way to say this is that RNNs are inherently \emph{sequential} models, processing the input string one symbol at a time to update their hidden state $\hiddStatet$, and using this hidden state in turn to compute $\hiddState_{\tstep + 1}$.
This results in perhaps the biggest shortcoming of the architecture for its applicability to real-world language modeling: The inability to efficiently \emph{parallelize} the processing (encoding) of long strings.
Let us first consider what we mean by the parallelizability of a language model architecture.\footnote{This section provides a very brief and intuitive treatment of parallelization. Our main goal is simply to point out this shortcoming of RNN LMs and with it motivate the next neural architecture we will introduce: transformers.}
Due to this sequential nature, the training procedure of RNNs is difficult to \emph{parallelize} effectively, leading to slower training times.
This characteristic poses a significant challenge when modeling long strings, as the computation for each element is dependent on the computation of the previous element, leading to a bottleneck in the training process.

In short, in our specific use case of language modeling, parallelization refers to the division of the processing of a specific string across multiple computational nodes, such that any specific node only performs a subset of operations required to process the entire string---the results of the subsets of the operations are then combined to build the representation of the entire string.
Importantly, the computations should be performed \emph{independently} between nodes in the sense that no node has to wait for any other node to provide it the results of its computation.
Being able to parallelize large models across computational nodes has led to some of the biggest advancements in modern deep learning.
As such, parallelizability is a crucial feature of any successful deep learning architecture.

However, notice that any dependence between the computations performed by the nodes defeats the purpose of parallelization---if the nodes have to wait for each other to finish computations, the same operations might as well be performed by a single node.
This is where recurrent neural networks fall short: the computations required to encode a string $\str$ into the hidden state $\hiddState_{|\str|}$ will always be sequential, preventing their distribution across different nodes.

\paragraph{Parallelizability in language modeling.}
When talking about parallelizing language models, it is important to think about \emph{which parts} can actually be parallelized.
In the case of RNNs, we saw that no part of the processing can be (besides the matrix multiplication in a single update rule)---the length of the longest chain of dependent computation will always scale linearly with the length of the string.
In the next section, we introduce transformers, a recent neural network architecture first introduced for processing text.
One of the big contributions of transformers is their parallelizability \emph{during training}---it enables their training on extremely large corpora and is thus one of the main reasons that they are behind the success of many of the most successful modern large language models.
Parallelizability \emph{during training} is crucial---notice that parallelization is, in fact, \emph{not possible} during generation from a locally normalized language model (cf. \cref{def:locally-normalized-model})---by definition, such models will generate \emph{one symbol at a time}.
To compute the representation of the new sentence (or the new prefix), which is required for the generation of the next symbol, the generated (sampled) symbol has to first be determined, which leaves their generation process inherently sequential.
In that respect, RNNs are as parallelizable as they can be during generation.
However, the sequential computation of the hidden states prevents the parallelization of computations of $\encRNNfunc{\strlet}$ even if the whole string is already \emph{given}.

We will see that the big parallelizability improvements of other architectures only come into play during training when the model is given the whole string in advance (such that no part of the string has to be sequentially generated) and can compute the (log-)likelihood of the given \emph{ground-truth} next symbols given the context.
That is, during training and given a string $\str \in \kleene{\alphabet}$, the model simply has to compute $\pLNSMFun{\eossym_\tstep}{\strlt}$ for all $\tstep = 1, \ldots, \strlen$---in representation based language models (cf. \cref{def:locally-normalized-rep-based}) this depends on $\encfunc{\strlt}$.
Crucially, computing $\pLNSMFun{\eossym_\tstep}{\strlt}$ is all that we need for training a language model (in the simplest case that we analyze)---the computed log-likelihood of the ground-truth next character $\eossym_\tstep$ is used for computing the loss function during training and used for gradient-based updates to the parameters as discussed in \cref{sec:numerical-optimization}.
In the next section, we will see how $\encfunc{\strlt}$ can be computed \emph{without} sequential dependencies.
However, in the case of RNNs, $\encfunc{\strlt}$ can only be computed sequentially---even if the entire string is known in advance.
This results in a crucial bottleneck in training RNNs on large corpora and vastly limits their applicability to implementing large language models.








\section{Representational Capacity of Recurrent Neural Networks} \label{sec:rnn-expressiveness}
Recurrent neural networks are one of the fundamental and most successful neural language model architectures.
In this section, we study some theoretical explanations behind their successes as well as some of their theoretical limitations.
Answering this question is essential whenever we require formal guarantees of the correctness of the outputs generated by an LM.
For example, one might ask a language model to solve a mathematical problem based on a textual description \citep{shridhar-etal-2023-distilling} or ask it to find an optimal solution to an everyday optimization problem \citep{lin-etal-2021-limitations}.
If such problems fall outside the theoretical capabilities of the LM, we have no ground to believe that the result provided by the model is correct.
The question also follows a long line of work on the linguistic capabilities of LMs, as LMs must be able to implement mechanisms of recognizing specific syntactic structures to generate grammatical sequences \citep[][\textit{inter alia}]{olmpics-Talmor-goldberg,hewitt-manning-2019-structural,jawahar-etal-2019-bert,liu-etal-2019-linguistic,Icard2020-ICACGM,Emrgent-Linguistic-Structure-Manning,A-Primer-in-Bertology-Rogers,belinkov-2022-probing,Deletang2022NeuralNA}.

One way of quantifying the expressive power of computational models is with the complexity of formal languages they can recognize \citep{Deletang2022NeuralNA}---we, too, will study the classes of (weighted) formal languages (such as the regular languages and the Turing computable languages) they can express.
Through this, diverse formal properties of modern LM architectures have been shown
\citep[e.g.,][inter alia]{Siegelmann-Sontag-Computational-Power,hao-etal-2018-context,korsky2019computational,merrill-2019-sequential,merrill-etal-2020-formal,hewitt-etal-2020-rnns,merrill-etal-2022-saturated,merrill-etal-2022-entailment}.
Inspecting complex models such as recurrent neural networks through the lens of formal language theory allows us to apply the well-studied theoretical results and understanding from the field to the recently more successful neural models.
While studying neural language models, we will revisit various aspects of the classical language models introduced in \cref{chapter:classical-lms}---indeed, this was also our main motivation for studying those closely.

Specifically, we will focus mainly on the \emph{Elman} recurrent neural networks due to their simplicity and their role as the ``fundamental'' RNN, capturing their recurrent nature.
We will also briefly touch upon the computational power of LSTM networks due to their somewhat different theoretical properties.
However, note that most of the results presented in the section generalize to other architectures as well.
We begin by investigating Elman RNNs in a practical setting, that is, under relatively strict and realistic assumptions, such as fixed-point arithmetic.
We show that Elman RNNs under such a regime are in fact equivalent to weighted finite-state automata.
Next, in the second subsection, we show that under more permissive assumptions of infinite precision and unbounded computation time, Elman RNNs are Turing complete.

\iftoggle{publish-notes}
{}
{
    \subsection{Equivalence of Formalisms and Language Homomorphisms}

    \begin{example}{A non-determinizable WFSA}{}
        The WFSA in \cref{fig:nondeterminizable-wfsa} is not determinizable.
        While the formal proof (and discussion of what it even means to be determinizable formally) is beyond the scope of this work\footnote{As always, more on this can be found in the Advanced formal language theory course.}, note that the intuition behind the issue is that we can arrive at the states $\stateq_1$ and $\stateq_2$ with the same string ($\syma$), and then \emph{loop} over the self-loop $\symb$ in both states.
        However, these self-loops have different weights, which means that they result in a path with the same label but with different weights.
        Because we can take the self-loop infinitely-many times, there are infinitely many paths \emph{with the same label} yet with different weights.
        If we wanted the WFSA to be deterministic, these paths with the same label should lead to the same state (otherwise, we would create non-determinism at some point).
        However, since we can construct infinitely many paths with the same label but with different weights, this means that we cannot ``group'' them into the same state.
        That is why the WFSA cannot be determinized.
    \end{example}
    \begin{figure}[t]
        \centering
        \begin{tikzpicture}[node distance = 17mm]
            \node[state, initial] (q0) [] { $\stateq_{0} / 1$ };
            \node[state] (q1) [right = of q0, yshift=14mm] { $\stateq_{1}$ };
            \node[state] (q2) [right = of q0, yshift=-14mm] { $\stateq_{2}$ };
            \node[state, accepting] (q3) [right = of q1, yshift=-14mm] { $\stateq_{3} / 1$ };
            \draw[transition] (q0) edge[bend left, above, sloped] node{ $\syma/0.5$ } (q1)
            (q0) edge[bend right, above, sloped] node{ $\syma/0.5$ } (q2)
            (q1) edge[above, loop above] node{ $\symb/0.9$ } (q1)
            (q2) edge[below, loop below] node{ $\symb/0.1$ } (q2)
            (q1) edge[bend left, above, sloped] node{ $\symc/0.1$ } (q3)
            (q2) edge[bend right, above, sloped] node{ $\symc/0.9$ } (q3) ;
        \end{tikzpicture}
        \caption{A non-determinizable WFSA.}
        \label{fig:nondeterminizable-wfsa}
    \end{figure}
}

\subsection{RNNs and Weighted Regular Languages }
Analyzing complex systems with intricate interactions between inputs and parameters and temporal dependencies can be tricky.
This is a common issue when studying neural networks in general.
In fact, most, if not all, theoretical frameworks for analyzing neural models such as RNNs rely on various assumptions about their components to make the analysis feasible.
For example, theoretical results on neural networks (for example, optimization guarantees or function/system identifiability guarantees) often make the assumption that the activation functions are linear or of some other easy-to-analyze form.
Similarly, a fruitful manner to analyze the expressivity of recurrent neural networks specifically is by making (somewhat different) simplifying assumptions on the non-linear activation functions, since those are what often make analysis difficult.
A common simplification is the use of the Heaviside activation function.
\begin{definition}{Heaviside function}{heaviside}
    The \defn{Heaviside} function is defined as
    \begin{equation} \label{eq:heaviside}
        \heaviside(x) = \begin{cases} 1 & \ifcondition x > 0  \\
              0 & \otherwisecondition
        \end{cases}
    \end{equation}
\end{definition}
In words, the Heaviside function maps every real value either $0$ or $1$, depending on whether it is greater than or less than zero.
In the following, we will refer to the set $\set{0, 1}$ as $\B \defeq \set{0, 1}$.
\begin{definition}{Heaviside Elman Network}{}
    A \defn{Heaviside Elman network} (\hernnAcr{}) is an Elman network with Heaviside function $\heaviside$ as the non-linearity.
\end{definition}

\paragraph{Elman network parameters.}
Importantly, note that the \emph{parameters} of the network do not have to be elements of $\B$---we assume those can take arbitrary real (or rational) values.
Indeed, networks constrained to parameters $\modelparam \in \B$ would only be able to recognize unweighted languages.
Furthermore, for this section, we expand our definition of a real- or rational-weighted RNN to be able to contain \emph{weights} $\infty$ and $-\infty$.
While those are not real (or rational) numbers, we will see they become useful when we want to explicitly exclude specific sequences from the support of the model, i.e., when we want to assign probability $0$ to them.

Before we move to the central result of the subsection, we first introduce a fact that makes it easier to talk about how an RNN language models can simulate a deterministic PFSAA.
We will be interested in conjoining elements of vectors in $\B^\hiddDim$, which can be performed by an Elman RNN with appropriately set parameters.
\begin{fact}{Performing the \texttt{AND} operation with a neural network}{and}
    Let $\idxm \in \NTo{\hiddDim}$, $\idxi_1, \ldots, \idxi_\idxm\in \NTo{\hiddDim}$, and $\vx, \vv \in \B^\hiddDim$ with $\evv_{\idxi} = \ind{\idxi \in \set{\idxi_1, \ldots, \idxi_\idxm}}$.
    Then, $\heaviside\left(\vv^\top \vx - \left(\idxm - 1\right) \right) = 1$ if and only if $\evx_{\idxi_\idxk} = 1$ for all $\idxk = 1, \ldots, \idxm$.
    In other words, $\heaviside\left(\vv^\top \vx - \left(\idxm - 1\right) \right) = \evx_{\idxi_1} \wedge \cdots \wedge \evx_{\idxi_\idxm}$.
\end{fact}

\paragraph{The central result.}
The central result of this section is captured in the following theorem from \citet{svete-cotterell-2023-recurrent}.
\begin{theorem}{Equivalence of Heaviside Elman RNNs and WFSAs}{elman-pfsa-euqivalent}
    Heaviside Elman RNNs are equivalent to deterministic probabilistic finite-state automata.
\end{theorem}
Notice that we only make the claim for \emph{probabilistic} WFSA.
This is without loss of generality, as, from \cref{thm:locally-normalizing-wfsas}, we know we can assume $\wfsa$ is locally normalized.
We will prove \cref{thm:elman-pfsa-euqivalent} by showing that an RNN with Heaviside activations is at most regular, and then showing how such an RNN can in fact simulate any deterministic PFSA.
We show each direction as its own lemma.

\begin{lemma}{}{heaviside-rnns-regular}
    The distribution represented by a recurrent neural network with a Heaviside non-linearity $\heaviside$ is regular.
\end{lemma}
\begin{proof}
    Let $\rnn = \elmanrnntuple$ be a \hernnAcr{} defining the conditional probabilities $\pLNSM$.
    We construct a deterministic PFSA $\wfsa = \wfsatuple$ defining the same string probabilities.
    Let $\hToQ: \B^\hiddDim \to \Z_{2^\hiddDim}$ be a bijection.
    Now, for every state $\stateq \defeq \hToQFun{\vh} \in \states \defeq \B^\hiddDim$, construct a transition $\edge{\stateq}{\sym}{w}{\stateq'}$ where $\stateq' = \elmanUpdate{\vh}{\sym}$ with the weight $w = \pLNSM\left(\sym \mid \vh\right) = \projfuncEosalphabetminusFunc{\outMtx \, \vh}_{\sym}$.
    We define the initial function as $\initf\left(\hToQFun{\vh}\right) = \ind{\vh = \hiddStateZero}$ and final function $\finalf$ with $\finalf\left(\stateq\right) \defeq \pLNSM\left(\eos \mid \hToQFun{\stateq}\right)$.

    It is easy to see that $\wfsa$ defined this way is deterministic.
    We now prove that the weights assigned to strings by $\wfsa$ and $\rnn$ are the same.
    Let $\str \in \kleene{\alphabet}$ with $|\str| = \strlen$ and
    \[
        \apath = \left(\edge{\hToQFun{\hiddStateZero}}{\sym_1}{w_1}{\stateq_1}, \ldots, \edge{\qT}{\sym_{\strlen}}{w_{\strlen}}{\stateq_{\strlen + 1}}\right)
    \]
    the $\str$-labeled path starting in $\hToQFun{\hiddStateZero}$ (such a path exists since we the defined automaton is \emph{complete}---all possible transitions are defined for all states).
    \begin{align*}
        \wfsa\left(\str\right) = & \initf\left(\hToQFun{\hiddStateZero}\right) \cdot \left[\prod_{\tstep = 1}^{\strlen} w_\tstep\right] \cdot \finalf\left(\stateq_{\strlen + 1}\right)                \\
        =                        & 1 \cdot \prod_{\tstep = 1}^{\strlen}\pLNSM\left(\symt \mid \hToQinvFun{\stateq_\tstep}\right) \cdot \pLNSM\left(\eos \mid \hToQinvFun{\stateq_{\strlen + 1}}\right) \\
        =                        & \pLN\left(\str\right)
    \end{align*}
    which is exactly the weight assigned to $\str$ by $\rnn$.
    Note that all paths not starting in $\hToQFun{\hiddStateZero}$ have weight $0$ due to the definition of the initial function.
\end{proof}

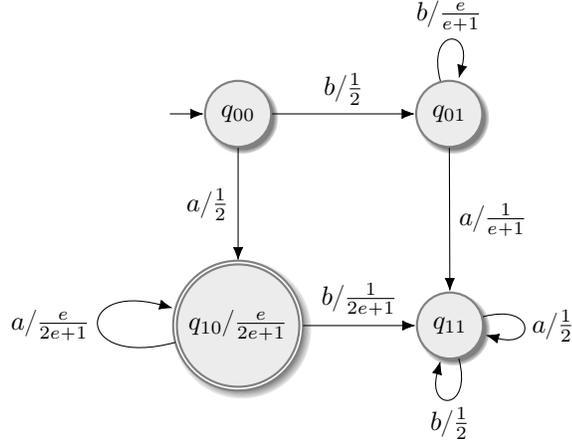
\begin{figure}[t]
    \centering
    \begin{tikzpicture}[node distance = 15mm]
        \node[state, initial] (q00) [] { $\stateq_{00}$ };
        \node[state] (q01) [right = of q00, xshift=4mm] { $\stateq_{01}$ };
        \node[state, accepting] (q10) [below = of q00] { $\stateq_{10} / \frac{e}{2e + 1}$ };
        \node[state] (q11) [right = of q10] { $\stateq_{11}$ };
        \draw[transition]  (q00) edge[left] node{ $\syma/\frac{1}{2}$ } (q10)
        (q00) edge[above] node{ $\symb/\frac{1}{2}$ } (q01)
        (q01) edge[right] node{ $\syma/\frac{1}{e + 1}$ } (q11)
        (q01) edge[loop above] node{ $\symb/\frac{e}{e + 1}$ } (q01)
        (q10) edge[loop left] node{ $\syma/\frac{e}{2e+1}$ } (q10)
        (q10) edge[above] node{ $\symb/\frac{1}{2e+1}$ } (q11)
        (q11) edge[loop right] node{ $\syma/\frac{1}{2}$ } (q11)
        (q11) edge[loop below] node{ $\symb/\frac{1}{2}$ } (q11) ;
    \end{tikzpicture}
    \caption{The WFSA corresponding to the RNN defined in \cref{eq:wfsa-rnn-example}.}
    \label{fig:rnn-wfsa}
\end{figure}

Let us look at an example of the construction above.
\begin{example}{A PFSA simulating an RNN}{}
    Let $\rnn = \rnntuple$ be a Heaviside RNN sequence model with the parameters
    \begin{align} \label{eq:wfsa-rnn-example}
        \alphabet                            & = \set{\syma, \symb}                                                                          \\
        \hiddDim                             & = 2                                                                                           \\
        \dyMap\left(\hiddStatet, \sym\right) & = \heaviside\left(\begin{pmatrix}
                                                                     1 & 0 \\ 0 & 1
                                                                 \end{pmatrix} \hiddState_{\tstep - 1} + \begin{pmatrix}
                                                                                                             1 & 0 \\ 0 & 1
                                                                                                         \end{pmatrix} \onehot{\sym} \right) \\
        \outMtx                              & = \begin{pmatrix}
                                                     1 & 0 \\ 0 & 1 \\ 1 & -\infty
                                                 \end{pmatrix}                                                                \\
        \hiddStateZero                       & = \begin{pmatrix}
                                                     0 \\ 0
                                                 \end{pmatrix}
    \end{align}
    and $\ordering\left(\syma\right) = 1$, $\ordering\left(\symb\right) = 2$, and $\ordering\left(\eos\right) = 3$.
    The automaton corresponding to this RNN contains the states $\stateq_{ij}$ corresponding to the hidden states $\hiddState = \begin{pmatrix}
            i \\ j
        \end{pmatrix}$.
    It is shown in \cref{fig:rnn-wfsa}; as we can see, the automaton indeed has an exponential number of useful states in the dimensionality of the hidden state, meaning that the RNN is a very compact way of representing it.
\end{example}

To show the other direction of \cref{thm:elman-pfsa-euqivalent}, we now give a variant of a classic theorem originally due to \citet{Minsky1954} but with a \emph{probabilistic} twist, allowing us to model \emph{weighted} languages with Elman RNNs.
\begin{lemma}{Elman RNNs can encode PFSAs}{minsky-constr}
    Let $\wfsa = \wfsatuple$ be a tight probabilistic deterministic finite-state automaton.
    Then, there exists a Heaviside-activated Elman network with a hidden state of size $\hiddState = |\alphabet||\states|$ that encodes the same distribution as $\wfsa$.
\end{lemma}

We give proof by construction: Given a deterministic PFSA $\wfsa = \wfsatuple$, we construct an Elman RNN $\rnn = \elmanrnntuple$ accepting the same weighted language as $\wfsa$: $\lang\left(\wfsa\right) = \lang\left(\rnn\right)$ by defining the elements of the tuple $\elmanrnntuple$.
In the rest of the section, we will first intuitively describe the construction, and then formally prove the central results that the construction relies on.
Let $\ordering\colon \states \times \alphabet \rightarrow \Zmod{\nstates\nsymbols }$ be a bijection, i.e., an ordering of $\states \times \alphabet$, $\symordering\colon \alphabet \rightarrow \Zmod{|\alphabet|}$ an ordering of $\alphabet$, and $\eossymordering\colon \eosalphabet \rightarrow \Zmod{|\eosalphabet|}$ an ordering of $\eosalphabet$; these mappings assign each element in their pre-image an integer which can then be used to index into matrices and vectors as we will see below.
We use $\ordering$, $\symordering$, and $\eossymordering$ to define the one-hot encodings $\onehot{\cdot}$ of state-symbol pairs and of the symbols.
That is, we assume that $\onehot{\stateq, \sym}_{\idxd} = \ind{\idxd = \ordering\left(\stateq, \sym\right)}$, and similar for $\onehot{\sym}$.
Similarly to the proof of \cref{lem:heaviside-rnns-regular}, we denote with $\qToH{}$ and $\qToHinv{}$ the mappings from $\states \times \eosalphabet$ to the hidden state space of the RNN and its inverse.
The alphabet of the RNN of course matches the one of the WFSA.

\paragraph{\hernnAcr{}'s hidden states.}
The hidden states of the RNN live in $\B^{|\states| |\alphabet|}$.
A hidden state $\hiddStatet$ encodes the state $\qt$ the simulated $\wfsa$ is in at time $\tstep$ \emph{and} the transition symbol $\symt$ with which $\wfsa$ ``arrived'' at $\qt$ as a \emph{one-hot encoding} of the pair $\left(\qt, \symt\right)$.
Formally,
\begin{equation}
    \hiddStatet = \onehot{\left(\qt, \symt\right)} \in \B^{\nstates \nsymbols }.
\end{equation}
This also means that $\hiddDim = |\states| |\alphabet|$.
There is a small caveat: how do we set the incoming symbol of $\wfsa$'s (sole) initial state $\qinit$ (the first time it is entered)?
A straightforward solution would be to augment the alphabet of the RNN with the $\bos$ symbol (cf. \cref{sec:global-local-normalization}), which we define to be the label of the incoming arc denoting the initial state (this would be the only transition labeled with $\bos$).
However, as we show later, the symbol used to arrive into $\statep$ does not have an effect on the \emph{subsequent} transitions---it is only needed to determine the \emph{target} of the current transition.
Therefore, we can simply represent the initial state $\hiddStateZero$ of $\rnn$ with the one-hot encoding of \emph{any} pair $\left(\qinit, \syma\right)$, where $\qinit$ is the initial state of the WFSA and $\syma \in \alphabet$.

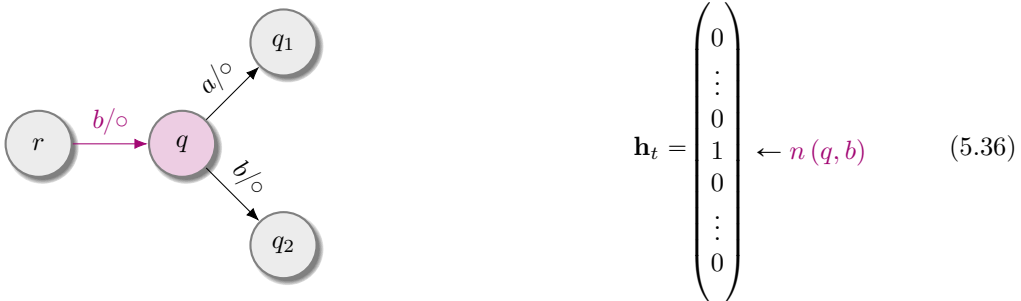
\begin{figure}[htbp]
    \centering
    \begin{minipage}{0.5\linewidth}
        \centering
        \begin{tikzpicture}[node distance = 10mm]
            \node[state] (r) [] { $\stater$ };
            \node[state, fill=ETHPurple!20] (p) [right = of r] { $\stateq$ };
            \node[state] (q1) [above right = of p] { $\stateq_1$ };
            \node[state] (q2) [below right = of p] { $\stateq_2$ };
            \draw[transition]  (r) edge[above, ETHPurple] node{ $\symb/\circ$ } (p)
            (p) edge[above, sloped] node{ $\syma/\circ$ } (q1)
            (p) edge[above, sloped] node{ $\symb/\circ$ } (q2);
        \end{tikzpicture}
        \caption{A fragment of a WFSA.}
        \label{fig:wfsa-fragment-misky-1}
    \end{minipage}
    \hfill
    \begin{minipage}{0.45\linewidth}
        \centering
        \vspace{-10mm}
        \begin{equation} \label{eq:minsky-const-example-1}
            \hiddState_\tstep =
            \begin{blockarray}{cc}
                & \\
                \begin{block}{(c)c}
                    & \\
                    0 & \\
                    \vdots & \\
                    0 & \\
                    1 & \leftarrow \textcolor{ETHPurple}{\ordering\left(\stateq, \symb\right)}\\
                    0 & \\
                    \vdots & \\
                    0 & \\
                    & \\
                \end{block}
            \end{blockarray}
        \end{equation}
    \end{minipage}
\end{figure}

For example, for the fragment of a WFSA in \cref{fig:wfsa-fragment-misky-1}, the hidden state encoding the current state $\stateq$ and the incoming arc $\symb$ is of the form presented in \cref{eq:minsky-const-example-1}.

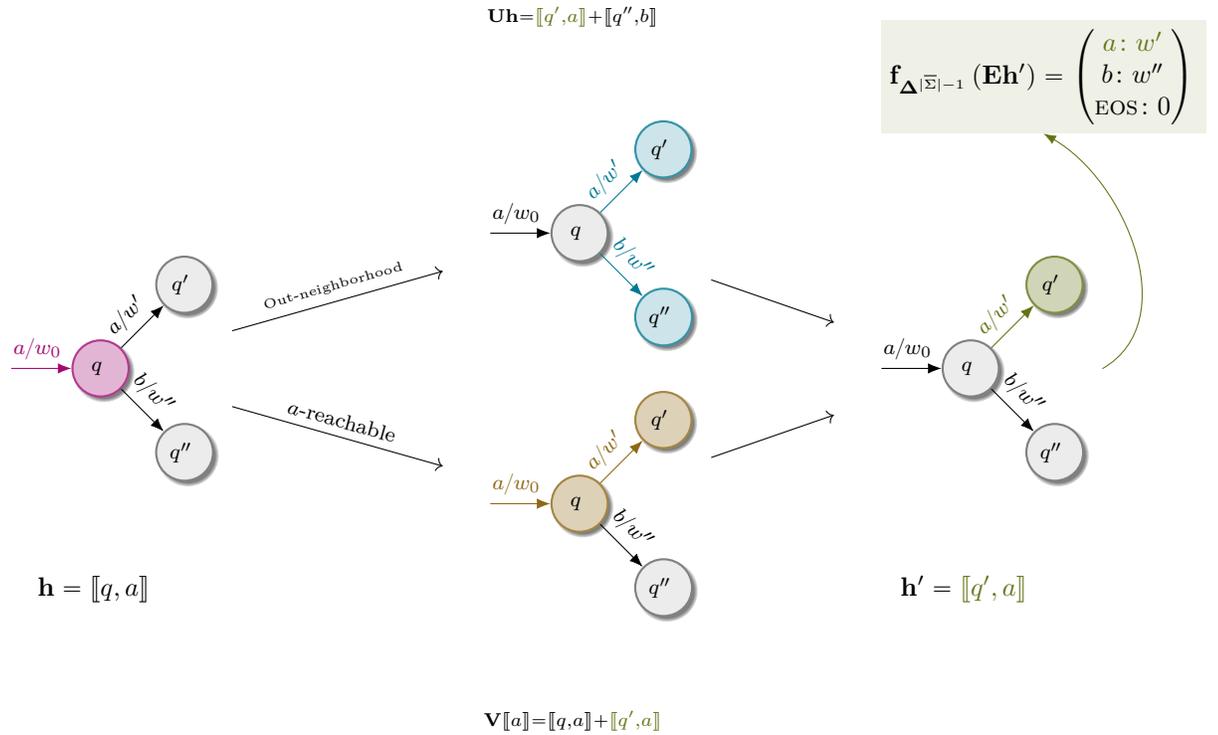
\begin{figure*}
    \centering
    \begin{tikzpicture}

        \node[align=center] (a1) {
            \begin{tikzpicture}[node distance = 8mm, minimum size = 5mm]
                \footnotesize
                \node[state, fill=ETHPurple!30, draw=ETHPurple!80] (q){ $\stateq$ };
                \node[draw=none, fill=none] (q0) [left = of q] { };
                \node[state] (q1) [above right = of q] { $\stateq'$ };
                \node[state] (q2) [below right = of q] { $\stateq''$ };
                \draw[transition]   (q0) edge[ETHPurple, above, sloped] node{ $\syma/w_0$ } (q)
                (q) edge[above, sloped] node{ $\syma/w'$ } (q1)
                (q) edge[above, sloped] node{ $\symb/w''$ } (q2);
            \end{tikzpicture}
        };
        \node[align=center, below = of a1, yshift=1cm] (e1) { $\hiddState = \onehot{\stateq, \syma}$ };

        \node[align=center, right = of a1, xshift=8mm, yshift=18mm] (a2) {
            \begin{tikzpicture}[node distance = 8mm, minimum size = 5mm]
                \footnotesize
                \node[state] (q){ $\stateq$ };
                \node[draw=none, fill=none] (q0) [left = of q] { };
                \node[state, fill=ETHPetrol!20, draw=ETHPetrol!80] (q1) [above right = of q] { $\stateq'$ };
                \node[state, fill=ETHPetrol!20, draw=ETHPetrol!80] (q2) [below right = of q] { $\stateq''$ };
                \draw[transition]   (q0) edge[above, sloped] node{ $\syma/w_0$ } (q)
                (q) edge[ETHPetrol, above, sloped] node{ $\syma/w'$ } (q1)
                (q) edge[ETHPetrol, above, sloped] node{ $\symb/w''$ } (q2);
            \end{tikzpicture}
        };
        \node[align=center, above = of a2, yshift=-1cm] (e2) { $\scriptstyle \recMtx \hiddState = \textcolor{ETHGreen}{\onehot{\stateq', \syma}} + \onehot{\stateq'', \symb}$ };

        \node[align=center, right = of a1, xshift=8mm, yshift=-18mm] (a3) {
            \begin{tikzpicture}[node distance = 8mm, minimum size = 5mm]
                \footnotesize
                \node[state, fill=ETHBronze!30, draw=ETHBronze!80] (q){ $\stateq$ };
                \node[draw=none, fill=none] (q0) [left = of q] { };
                \node[state, fill=ETHBronze!30, draw=ETHBronze!80] (q1) [above right = of q] { $\stateq'$ };
                \node[state] (q2) [below right = of q] { $\stateq''$ };
                \draw[transition]   (q0) edge[ETHBronze, above, sloped] node{ $\syma/w_0$ } (q)
                (q) edge[ETHBronze, above, sloped] node{ $\syma/w'$ } (q1)
                (q) edge[above, sloped] node{ $\symb/w''$ } (q2);
            \end{tikzpicture}
        };
        \node[align=center, below = of a3, yshift=1cm] (e3) { $\scriptstyle \inMtx \onehot{\syma} = \onehot{\stateq, \syma} + \textcolor{ETHGreen}{\onehot{\stateq', \syma}}$ };

        \node[align=center, right = of a1, xshift=60mm] (a4) {
            \begin{tikzpicture}[node distance = 8mm, minimum size = 5mm]
                \footnotesize
                \node[state] (q4f){ $\stateq$ };
                \node[draw=none, fill=none] (q40) [left = of q4f] { };
                \node[state, fill=ETHGreen!30, draw=ETHGreen!80] (q41) [above right = of q4f] { $\stateq'$ };
                \node[state] (q42) [below right = of q4f] { $\stateq''$ };
                \draw[transition]   (q40) edge[above, sloped] node{ $\syma/w_0$ } (q4f)
                (q4f) edge[ETHGreen, above, sloped] node(greenEdge){ $\syma/w'$ } (q41)
                (q4f) edge[above, sloped] node{ $\symb/w''$ } (q42);
            \end{tikzpicture}
        };

        \node[align=center, below = of a4, yshift=1cm] (e4) { $\hiddState' = \textcolor{ETHGreen}{\onehot{\stateq', \syma}}$ };

        \draw[->] (a1) edge[above, sloped] node{\tiny Out-neighborhood} (a2)
        (a1) edge[above, sloped] node{\footnotesize $\syma$-reachable} (a3)
        (a2) edge (a4)
        (a3) edge (a4) ;





        \node[align=center, above = of a4, xshift=10mm, yshift=-5mm, fill=ETHGreen!10] (sm) {
            $\projfuncEosalphabetminus\left(\outMtx \hiddState'\right) = \begin{pmatrix}
                    \textcolor{ETHGreen}{\syma\colon w'} \\
                    \symb\colon w''                      \\
                    \eos\colon 0
                \end{pmatrix}$
        };

        \draw[-Latex, ETHGreen, out = 30, in = -30] (a4.east) to (sm.south);

    \end{tikzpicture}
    \caption{A high-level illustration of how the transition function of the FSA is simulated in Minsky's construction on a fragment of an FSA \textcolor{ETHPurple}{starting at $\stateq$} (encoded in $\hiddState$) and reading the symbol $\syma$.
        The top path disjoins the representations of the states in the \textcolor{ETHPetrol}{out-neighborhood} of $\stateq$, whereas the bottom path disjoins the representations of states \textcolor{ETHBronze}{reachable by} an $\syma$-transition.
        The Heaviside activation \textcolor{ETHGreen}{conjoins} these two representations into $\hiddState'$ (rightmost fragment).
        Projecting $\outMtx \hiddState'$ results in the vector defining the same probability distribution as the outcoming arcs of $\stateq$ (\textcolor{ETHGreen}{green box}).}
    \label{fig:minsky-overview}
\end{figure*}

\paragraph{Encoding the transition function.}
The idea of defining $\recMtx$, $\inMtx$, and $\biasVech$ is for the Elman update rule to perform, upon reading $\symtplus$, element-wise conjunction between the representations of the out-neighborhood of $\qt$ and the representation of the states $\wfsa$ can transition into after reading in $\symtplus$ from \emph{any state}.
The former is encoded in the recurrence matrix $\recMtx$, which has access to the current hidden state that encodes $\qt$ while the latter is encoded in the input matrix $\inMtx$, which has access to the one-hot representation of $\symtplus$.
Conjugating the entries in those two representations will, due to the determinism of $\wfsa$, result in a single non-zero entry: one representing the state which can be reached from $\qt$ (1\textsuperscript{st} component) using the symbol $\symtplus$ (2\textsuperscript{nd} component); see \cref{fig:minsky-overview}.

The recurrence matrix $\recMtx$ lives in $\B^{|\alphabet||\states| \times |\alphabet||\states|}$.
The main idea of the construction is for each column $\recMtx_{\colon, \ordering\left(\stateq, \sym\right)}$ of the matrix to represent the ``out-neighborhood'' of the state $\stateq$ in the sense that the column contains $1$'s at the indices corresponding to the state-symbol pairs $\left(\stateq', \sym'\right)$ such that $\wfsa$ transitions from $\stateq$ to $\stateq'$ after reading in the symbol $\sym'$.
That is, for $\stateq, \stateq' \in \states$ and $\sym, \sym' \in \alphabet$, we define
\begin{equation} \label{eq:minsky-recmtx}
    \eRecMtx_{\ordering(\stateq', \sym'), \ordering(\stateq, \sym)} \defeq \ind{\edge{\qt}{\sym'}{\circ}{\stateq'} \in \trans}.
\end{equation}
Since $\sym$ is free, each column is repeated $\nsymbols$-times: once for every $\sym \in \alphabet$---this is why, after entering the next state, the symbol used to enter it does not matter anymore and, in the case of the initial state, any incoming symbol can be chosen to represent $\hiddStateZero$.

For example, for the fragment of a WFSA in \cref{fig:wfsa-fragment-misky-1}, the recurrence matrix would take the form
\begin{equation}
    \recMtx =
    \begin{blockarray}{ccccccc}
        & & & {\ordering\left(\stateq, \symb\right)} & & \\
        & & & \textcolor{ETHPurple}{\downarrow} & & \\
        \begin{block}{(cccccc)c}
            & & & 0 & & & \\
            & & & \vdots & & & \\
            & & & 1 & & & \leftarrow \ordering\left(\stateq_1, \syma\right)\\
            & \cdots & & \vdots & \cdots & & \\
            & & & 1 & & &  \leftarrow \ordering\left(\stateq_2, \symb\right) \\
            & & & \vdots & & & \\
            & & & 0 & & & \\
        \end{block}
    \end{blockarray}
\end{equation}
and the matrix-vector product $\recMtx \hiddState_\tstep$ with $\hiddStatet$ from before results in
\begin{equation}
    \recMtx \hiddState_\tstep =
    \begin{blockarray}{cc}
        & \\
        \begin{block}{(c)c}
            & \\
            0 & \\
            \vdots & \\
            1 & \leftarrow \ordering\left(\stateq_1, \syma\right)\\
            \vdots & \\
            1 &  \leftarrow \ordering\left(\stateq_2, \symb\right) \\
            \vdots & \\
            0 & \\
            & \\
        \end{block}
    \end{blockarray}
\end{equation}

The input matrix $\inMtx$ lives in $\B^{|\alphabet||\states| \times |\alphabet|}$ and encodes the information about which states can be reached by which symbols (from \emph{any} state in $\wfsa$).
The non-zero entries in the column corresponding to $\sym' \in \alphabet$ correspond to the state-symbol pairs $\left(\stateq', \sym'\right)$ such that $\stateq'$ is reachable with $\sym'$ from \emph{some} state:
\begin{equation}\label{eq:minsky-inmtx}
    \eInMtx_{\ordering\left(\stateq', \sym'\right), \symordering\left(\sym'\right)} \defeq \ind{\edge{\circ}{\sym'}{\circ}{\stateq'} \in \trans}.
\end{equation}
For example, for the fragment of a WFSA in \cref{fig:wfsa-fragment-misky-2}, the input matrix would take the form
\begin{equation}
    \inMtx =
    \begin{blockarray}{ccccccc}
        & & & {\symordering\left(\symb\right)} & & \\
        &  & & {\downarrow} & & \\
        \begin{block}{(cccccc)c}
            & & & 0 & & & \\
            & & & \vdots & & & \\
            & & & 1 & & & \textcolor{ETHPurple}{\leftarrow\ordering\left(\statep, \symb\right)}\\
            & \cdots & & \vdots & \cdots & & \\
            & & & 1 & & &  \textcolor{ETHPurple}{\leftarrow\ordering\left(\stateq_2, \symb\right) }\\
            & & & \vdots & & & \\
            & & & 0 & & & \\
        \end{block}
    \end{blockarray}
\end{equation}
and the matrix-vector product $\inMtx \ve_{\symordering\left(\syma\right)}$ and $\inMtx \ve_{\symordering\left(\symb\right)}$ would take the form (see also \cref{fig:wfsa-fragment-misky-3})
\begin{equation}
    \inMtx \ve_{\symordering\left(\syma\right)} =
    \begin{blockarray}{cc}
        \begin{block}{(c)c}
            & \\
            0 & \\
            \vdots & \\
            1 & \textcolor{ETHGreen}{\leftarrow \ordering\left(\stateq_1, \syma\right)}\\
            \vdots &  \\
            0 & \\
            & \\
        \end{block}
    \end{blockarray}
    \quad \quad \quad
    \inMtx \ve_{\symordering\left(\symb\right)} =
    \begin{blockarray}{cc}
        \begin{block}{(c)c}
            & \\
            0 & \\
            \vdots & \\
            1 & \textcolor{ETHPurple}{\leftarrow \ordering\left(\statep, \symb\right)}\\
            \vdots &  \\
            1 & \textcolor{ETHPurple}{\leftarrow \ordering\left(\stateq_2, \symb\right) }\\
            \vdots & \\
            0 & \\
            & \\
        \end{block}
    \end{blockarray}
\end{equation}
\begin{figure}[t]
    \centering
    \begin{subfigure}{0.45 \textwidth}
        \centering
        \begin{tikzpicture}[node distance = 10mm]
            \node[state] (r) [] { $\stater$ };
            \node[state, fill=ETHPurple!20] (p) [right = of r] { $\statep$ };
            \node[state] (q1) [above right = of p] { $\stateq_1$ };
            \node[state, fill=ETHPurple!20] (q2) [below right = of p] { $\stateq_2$ };
            \draw[transition]  (r) edge[above, ETHPurple] node{ $\symb/\circ$ } (p)
            (p) edge[above, sloped] node{ $\syma/\circ$ } (q1)
            (p) edge[above, ETHPurple, sloped] node{ $\symb/\circ$ } (q2);
        \end{tikzpicture}
        \caption{An example of a fragment of a WFSA.}
        \label{fig:wfsa-fragment-misky-2}
    \end{subfigure}
    \hfill
    \begin{subfigure}{0.45 \textwidth}
        \centering
        \begin{tikzpicture}[node distance = 7mm]
            \node[state] (r) [] { $\stater$ };
            \node[state, fill=ETHPurple!20] (p) [right = of r] { $\statep$ };
            \node[state, fill=ETHGreen!20] (q1) [above right = of p] { $\stateq_1$ };
            \node[state, fill=ETHPurple!20] (q2) [below right = of p] { $\stateq_2$ };
            \draw[transition]  (r) edge[above, ETHPurple] node{ $\symb/\circ$ } (p)
            (p) edge[above, ETHGreen, sloped] node{ $\syma/\circ$ } (q1)
            (p) edge[above, ETHPurple, sloped] node{ $\symb/\circ$ } (q2);
        \end{tikzpicture}
        \caption{An example of a fragment of a WFSA.}
        \label{fig:wfsa-fragment-misky-3}
    \end{subfigure}
\end{figure}
Lastly, we define the bias as $\biasVech \defeq -\one \in \R^{\nstates\nsymbols }$, which allows the Heaviside function to perform the needed conjunction.

To put these components together, consider that, at each step of the computation, $\rnn$ computes $\hiddState_{\tstep + 1} = \heaviside\left(\recMtx \hiddState_{\tstep} + \inMtx \ve_{\syma} + \biasVech \right)$ where $\sym_{\tstep + 1} = \syma$.
The input to the non-linearity is computed as follows:
\begin{equation}
    \recMtx \hiddState_\tstep + \inMtx \ve_{\symordering\left(\syma\right)} + \biasVech =
    \begin{blockarray}{cc}
        & \\
        \begin{block}{(c)c}
            & \\
            0 & \\
            \vdots & \\
            \textcolor{ETHRed}{1} & \leftarrow \textcolor{ETHRed}{\ordering\left(\stateq_1, \syma\right)}\\
            \vdots & \\
            1 &  \leftarrow \ordering\left(\stateq_2, \symb\right) \\
            \vdots & \\
            0 & \\
            & \\
        \end{block}
    \end{blockarray}
    +
    \begin{blockarray}{cc}
        \begin{block}{(c)c}
            & \\
            0 & \\
            \vdots & \\
            \textcolor{ETHRed}{1} & \textcolor{ETHRed}{\leftarrow \ordering\left(\stateq_1, \syma\right)}\\
            \vdots &  \\
            0 & \\
            & \\
        \end{block}
    \end{blockarray}
    +
    \begin{blockarray}{c}
        \begin{block}{(c)}
            \\
            -1 \\
            \vdots \\
            -1 \\
            \vdots \\
            -1 \\
            \\
        \end{block}
    \end{blockarray}
\end{equation}

The following lemma proves that the construction described correctly implements the transition function of the PFSA.
\begin{lemma}{}{minsky-transitions}
    Let $\wfsa = \wfsatuple$ be a deterministic PFSA, $\str = \sym_1\ldots\sym_\strlen \in \kleene{\alphabet}$, and $\qt$ the state arrived at by $\wfsa$ upon reading the prefix $\strlet$.
    Let $\rnn$ be the \hernnAcr{} specified by the Minsky construction for $\wfsa$, $\ordering$ the ordering defining the one-hot representations of state-symbol pairs by $\rnn$, and $\hiddStatet$ $\rnn$'s hidden state after reading $\strlet$.
    Then, it holds that $\hiddStateZero = \onehot{\left(\qinit, \sym\right)}$ where $\qinit$ is the initial state of $\wfsa$ and $\sym \in \alphabet$ and $\hiddStateT = \onehot{\left(\qT, \sym_\strlen\right)}$.
\end{lemma}
\begin{proof}
    Define $\hToQFun{\hiddState = \onehot{\left(\stateq, \sym\right)}} \defeq \stateq$.
    We can then restate the lemma as $\hToQFun{\hiddStateT} = \qT$ for all $\str \in \kleene{\alphabet}$, $|\str| = \strlen$.
    Let $\apath$ be the $\str$-labeled path in $\wfsa$.
    We prove the lemma by induction on the string length $\strlen$.
    \paragraph{Base case: $\strlen = 0$.}
    Holds by the construction of $\hiddStateZero$.
    \paragraph{Inductive step: $\strlen > 0$.}
    Let $\str \in \kleene{\alphabet}$ with $|\str| = \strlen$ and assume that $\hToQFun{\hiddState_{\strlen - 1}} = \stateq_{\strlen - 1}$.

    We prove that the specifications of $\recMtx$, $\inMtx$, and $\biasVech$ ensure that $\hToQFun{\hiddStateT} = \qT$.
    By definition of the recurrence matrix $\recMtx$ (cf. \cref{eq:minsky-recmtx}), the vector $\recMtx \hiddStateTminus$ will contain a $1$ at the entries $\ordering\left(\stateq', \sym'\right)$ for $\stateq' \in \states$ and $\sym' \in \alphabet$ such that $\edge{\qTminus}{\sym'}{\circ}{\stateq'} \in \trans$.
    This can equivalently be written as $\recMtx \hiddStateTminus = \bigvee_{\edge{\qTminus}{\sym'}{\circ}{\stateq'} \in \trans} \onehot{\left(\stateq', \sym'\right)}$, where the disjunction is applied element-wise.

    On the other hand, by definition of the input matrix $\inMtx$ (cf. \cref{eq:minsky-inmtx}), the vector $\inMtx \onehot{\symT}$ will contain a $1$ at the entries $\ordering\left(\stateq', \symT\right)$ for $\stateq' \in \states$ such that $\edge{\circ}{\symT}{\circ}{\stateq'} \in \trans$.
    This can also be written as $\inMtx \onehot{\symT} = \bigvee_{\edge{\circ}{\symT}{\circ}{\stateq'} \in \trans} \onehot{\left(\stateq', \symT\right)}$.

    By \cref{fact:and}, $\heaviside\left(\recMtx \hiddStateTminus + \inMtx \onehot{\symT} + \biasVech\right)_{\ordering\left(\stateq', \sym'\right)} = \heaviside\left(\recMtx \hiddStateTminus + \inMtx \onehot{\symT} - \one \right)_{\ordering\left(\stateq', \sym'\right)} = 1$ holds if and only if $\left(\recMtx \hiddStateTminus\right)_{\ordering\left(\stateq', \sym'\right)} = 1$ and $\left(\inMtx \onehot{\symT}\right)_{\ordering\left(\stateq', \sym'\right)} = 1$.
    This happens if
    \begin{equation}
        \edge{\qTminus}{\sym'}{\circ}{\stateq'} \in \trans \text{ and } \edge{\circ}{\symT}{\circ}{\stateq'} \in \trans \iff \edge{\qTminus}{\symT}{\circ}{\stateq'},
    \end{equation}
    i.e., if and only if $\wfsa$ transitions from $\qTminus$ to $\qT$ upon reading $\symT$ (it transitions only to $\qT$ due to determinism).

    Since the string $\str$ was arbitrary, this finishes the proof.
\end{proof}

\paragraph{Encoding the transition probabilities.}
We now turn to the second part of the construction: encoding the string acceptance weights given by $\wfsa$ into the probability distribution defined by $\rnn$.
We present two ways of doing that: using the more standard softmax formulation, where we make use of the extended real numbers, and with the sparsemax.

The conditional probabilities assigned by $\rnn$ are controlled by the $|\eosalphabet| \times \nstates\nsymbols$-dimensional output matrix $\outMtx$.
Since $\hiddStatet$ is a one-hot encoding of the state-symbol pair $\qt, \symt$, the matrix-vector product $\outMtx \hiddStatet$ simply looks up the values in the $\ordering\left(\qt, \symt\right)^\text{th}$ column.
After being projected to $\SimplexEosalphabetminus$, the entry in the projected vector corresponding to some $\symtplus \in \eosalphabet$ should match the probability of that symbol given that $\wfsa$ is in the state $\qt$.
This is easy to achieve by simply encoding the weights of the outgoing transitions into the $\ordering\left(\qt, \symt\right)^\text{th}$ column, depending on the projection function used.
This is especially simple in the case of the sparsemax formulation.
By definition, in a PFSA, the weights of the outgoing transitions and the final weight of a state $\qt$ form a probability distribution over $\eosalphabet$ for every $\qt \in \states$.
Projecting those values to the probability simplex, therefore, leaves them intact.
We can therefore define
\begin{equation} \label{eq:minsky-outmtx}
    \outMtx_{\eossymordering\left(\sym'\right) \ordering\left(\stateq, \sym\right)} \defeq
    \begin{cases} \transitionWeightFun{\edge{\stateq}{\sym'}{w}{\circ}} & \mid \textbf{if } \sym' \in \alphabet \\
              \finalf\left(\stateq\right)                           & \mid \otherwisecondition\end{cases}.
\end{equation}
Projecting the resulting vector $\outMtx \hiddStatet$, therefore, results in a vector whose entries represent the transition probabilities of the symbols in $\eosalphabet$.

In the more standard softmax formulation, we proceed similarly but log the non-zero transition weights.
Defining $\log{0} \defeq -\infty$,\footnote{
    Note that the $-\infty$ entries are only needed whenever the original WFSA assigns $0$ probability to some transitions.
    In many implementations using $\softmax$-activated probabilities, this would not be required.
} we set
\begin{equation}
    \outMtx_{\eossymordering\left(\sym'\right) \ordering\left(\stateq, \sym\right)} \defeq
    \begin{cases} \log{\transitionWeightFun{\edge{\stateq}{\sym'}{w}{\circ}}} & \mid \textbf{if } \sym' \in \alphabet \\
              \log{\finalf\left(\stateq\right)}                           & \mid \otherwisecondition\end{cases}.
\end{equation}
It is easy to see that the entries of the vector $\softmaxfunc{\outMtx \hiddStatet}{}$ form the same probability distribution as the original outgoing transitions out of $\stateq$.
Over the course of an entire input string, these weights are multiplied as the RNN transitions between different hidden states corresponding to the transitions in the original PFSA $\wfsa$.

\begin{figure}[t]
    \centering
    \begin{tikzpicture}[node distance = 10mm]
        \node[state] (r) [] { $\stater$ };
        \node[state, fill=ETHPurple!20] (p) [right = of r] { $\statep$ };
        \node[state] (q1) [above right = of p] { $\stateq_1$ };
        \node[state] (q2) [below right = of p] { $\stateq_2$ };
        \draw[transition] (r) edge[above, ETHPurple] node{ $\symb/\circ$ } (p)
        (p) edge[above, sloped] node{ $\syma/w_1$ } (q1)
        (p) edge[above, sloped] node{ $\symb/w_2$ } (q2);
    \end{tikzpicture}
    \caption{An example of a fragment of a WFSA.}
    \label{fig:wfsa-fragment-misky-4}
\end{figure}
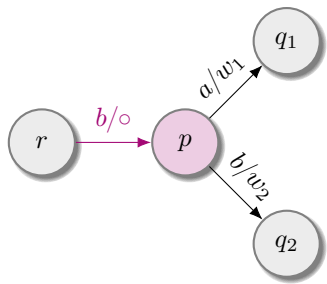

For example, for the fragment of a WFSA in \cref{fig:wfsa-fragment-misky-4}, the output matrix would take the form
\begin{equation}
    \outMtx =
    \begin{blockarray}{ccccccc}
        & & & \textcolor{ETHPurple}{\ordering\left(\stateq, \symb\right)} & & \\
        & & & \textcolor{ETHPurple}{\downarrow} & & \\
        \begin{block}{(cccccc)c}
            & & & -\infty & & & \\
            & & & \vdots & & & \\
            & & & \log w_1 & & & \leftarrow \eossymordering\left(\syma\right)\\
            & \cdots & & \vdots & \cdots & & \\
            & & & \log w_2 & & &  \leftarrow \eossymordering\left(\symb\right) \\
            & & & \vdots & & & \\
            & & & -\infty & & & \\
        \end{block}
    \end{blockarray}
\end{equation}

This means that, if $\hiddStatet$ encodes the state-symbol pair $\left(\stateq, \sym\right)$, the vector $\outMtx \hiddStatet$ will copy the selected column in $\outMtx$ which contains the output weight for all out symbols $\sym;$ of $\stateq$, i.e., the entry $\outMtx \hiddState_{\eossymordering\left(\sym'\right)}$ contains the weight on the arc $\edge{\stateq}{\sym'}{w}{\circ}$.
Over the course of an entire input string $\str$, these probabilities are simply multiplied as the RNN transitions between different hidden states corresponding to the transitions in the original WFSA $\wfsa$.

For example, for the fragment of a WFSA in \cref{fig:wfsa-fragment-misky-4}, the matrix-vector product $\outMtx \hiddState_\tstep$ would take the form
\begin{equation}
    \outMtx \hiddState_\tstep =
    \begin{blockarray}{cc}
        \begin{block}{(c)c}
            -\infty & \\
            \vdots & \\
            \log w_1 & \leftarrow \eossymordering\left(\syma\right)\\
            \vdots &  \\
            \log w_2 & \leftarrow \eossymordering\left(\symb\right) \\
            \vdots & \\
            -\infty & \\
        \end{block}
    \end{blockarray}
\end{equation}

The equivalence of the produced RNN LM to the PFSA is shown in the following lemma.
\begin{lemma}{}{minsky-probs}
    Let $\wfsa = \wfsatuple$ be a deterministic PFSA, $\str = \sym_1\ldots\sym_\strlen \in \kleene{\alphabet}$, and $\qt$ the state arrived at by $\wfsa$ upon reading the prefix $\strlet$.
    Let $\rnn$ be the \hernnAcr{} specified by the Minsky construction for $\wfsa$, $\outMtx$ the output matrix specified by the generalized Minsky construction, $\ordering$ the ordering defining the one-hot representations of state-symbol pairs by $\rnn$, and $\hiddStatet$ $\rnn$'s hidden state after reading $\strlet$.
    Then, it holds that $\pLN\left(\str\right) = \wfsa\left(\str\right)$.
\end{lemma}
\begin{proof}
    Let $\str \in \kleene{\alphabet}$, $|\str| = \strlen$ and let $\apath$ be the $\str$-labeled path in $\wfsa$.
    Again, let $\overline{\pdens}\left(\str\right) \defeq \prod_{\tstep = 1}^{|\str|} \pLNSMFun{\symt}{\strlt}$.
    We prove $\overline{\pdens}\left(\str\right) = \prod_{\tstep = 1}^\strlen w_\tstep$ by induction on $\strlen$.
    \paragraph{Base case: $\strlen = 0$.}
    In this case, $\str = \eps$, i.e., the empty string, and $\wfsa\left(\eps\right) = 1$.
    $\rnn$ computes $\overline{\pdens}\left(\eps\right) = \prod_{\tstep =1 }^{0} \pLNSM\left(\symt\mid\strlt\right) = 1$.
    \paragraph{Inductive step: $\strlen > 0$.}
    Assume that the $\overline{\pdens}\left(\sym_1 \ldots \symTminus\right) = \prod_{\tstep = 1}^{\strlen - 1} w_\tstep$.
    By \cref{lem:minsky-transitions}, we know that $\hToQFun{\hiddStateTminus} = \qTminus$ and $\hToQFun{\hiddStateT} = \qT$.
    By the definition of $\outMtx$ for the specific $\projfuncEosalphabetminus$, it holds that $\projfuncEosalphabetminusFunc{\outMtx \hiddState_{\strlen - 1}}_{\symordering\left(\sym\right)} = \transitionWeightFun{\edge{\hToQFun{\hiddState_{\strlen-1}}}{\sym}{w_\strlen}{\hToQFun{\hiddState_{\strlen}}}} = w_\strlen$.
    This means that $\overline{\pdens}\left(\str_{\leq \strlen}\right) = \prod_{\tstep = 1}^{\strlen} w_\tstep$, which is what we wanted to prove.

    Clearly, $\pLN\left(\str\right) = \overline{\pdens}\left(\str\right) \pLNSM\left(\eos \mid \str\right)$.
    By the definition of $\outMtx$ (cf. \cref{eq:minsky-outmtx}), $\left(\outMtx \hiddStateT\right)_{\symordering\left(\eos\right)} = \finalf\left(\hToQFun{\hiddStateT}\right)$, meaning that
    \begin{equation*}
        \pLN\left(\str\right) = \overline{\pdens}\left(\str\right) \pLNSM\left(\eos \mid \str\right) = \prod_{\tstep = 1}^{\strlen} w_\tstep\finalf\left(\hToQFun{\hiddStateT}\right) = \wfsa\left(\str\right).
    \end{equation*}
    Since $\str \in \kleene{\alphabet}$ was arbitrary, this finishes the proof.
\end{proof}

\begin{figure}[t]
    \centering
    \begin{tikzpicture}[node distance = 15mm]
        \node[state, initial, initial text = $1$] (q0) [] { $\stateq_0$ };
        \node[state] (q1) [right = of q0, xshift = 2.5 mm, yshift = 15 mm] { $\stateq_1$ };
        \node[state] (q2) [right = of q0, yshift = -15 mm] { $\stateq_2/0.5$ };
        \draw[transition] (q0) edge[above, bend left, sloped] node{ $\syma/0.1$ } (q1)
        (q0) edge[above, bend right, sloped] node{ $\symb/0.9$ } (q2)
        (q1) edge[above, bend left, sloped] node{ $\syma/0.5$ } (q0)
        (q1) edge[above, sloped] node{ $\symb/0.5$ } (q2)
        (q2) edge[loop right, above, sloped] node{ $\symb/0.5$ } (q2);
    \end{tikzpicture}
    \caption{The WFSA $\wfsa$.}
    \label{fig:minsky-const-example}
\end{figure}

We now walk through an example of the Minsky construction.
\begin{example}{Minsky construction}{}
    Let $\wfsa = \wfsatuple$ be a WFSA as shown in \cref{fig:minsky-const-example}.
    Since $\wfsa$ has $|\states| = 3$ states and an alphabet of $|\alphabet| = 2$ symbols, the hidden state of the representing RNN $\rnn$ will be of dimensionality $3 \cdot 2 = 6$.
    Assume that the set of state-symbol pairs is ordered as $\left(\stateq_0, \syma\right), \left(\stateq_0, \symb\right), \left(\stateq_1, \syma\right), \left(\stateq_1, \symb\right), \left(\stateq_2, \syma\right), \left(\stateq_2, \symb\right)$.
    The initial state can be represented (choosing $\syma$ as the arbitrary ``incoming symbol'') as
    \begin{equation}
        \hiddStateZero = \begin{pmatrix}
            1 \\ 0 \\ 0 \\ 0 \\ 0 \\ 0
        \end{pmatrix}.
    \end{equation}
    The recurrent matrix $\recMtx$ of $\rnn$ is
    \begin{equation}
        \recMtx = \begin{pmatrix}
            0                     & 0                     & \textcolor{ETHRed}{1} & \textcolor{ETHRed}{1} & 0                     & 0                     & \\
            0                     & 0                     & 0                     & 0                     & 0                     & 0                     & \\
            \textcolor{ETHRed}{1} & \textcolor{ETHRed}{1} & 0                     & 0                     & 0                     & 0                     & \\
            0                     & 0                     & 0                     & 0                     & 0                     & 0                     & \\
            0                     & 0                     & 0                     & 0                     & 0                     & 0                     & \\
            \textcolor{ETHRed}{1} & \textcolor{ETHRed}{1} & \textcolor{ETHRed}{1} & \textcolor{ETHRed}{1} & \textcolor{ETHRed}{1} & \textcolor{ETHRed}{1} &
        \end{pmatrix},
    \end{equation}
    the input matrix $\inMtx$
    \begin{equation}
        \inMtx = \begin{pmatrix}
            \textcolor{ETHRed}{1} & 0                     \\
            0                     & 0                     \\
            \textcolor{ETHRed}{1} & 0                     \\
            0                     & 0                     \\
            0                     & 0                     \\
            0                     & \textcolor{ETHRed}{1}
        \end{pmatrix},
    \end{equation}
    and the output matrix $\outMtx$ is
    \begin{equation}
        \outMtx = \begin{pmatrix}
            \log\left(0.1\right) & \log\left(0.1\right) & \log\left(0.5\right) & \log\left(0.5\right) & -\infty              & -\infty              \\
            \log\left(0.9\right) & \log\left(0.9\right) & \log\left(0.5\right) & \log\left(0.5\right) & \log\left(0.5\right) & \log\left(0.5\right) \\
            -\infty              & -\infty              & -\infty              & -\infty              & \log\left(0.5\right) & \log\left(0.5\right)
        \end{pmatrix},
    \end{equation}
    where the last row corresponds to the symbol $\eos$.
    The target of the $\symb$-labeled transition from $\stateq_0$ ($\edge{\stateq_0}{\symb}{0.9}{\stateq_2}$) is computed as follows:
    \begin{align*}
        \hiddState_1 & = \heaviside\left(\recMtx \hiddStateZero + \inMtx \onehot{\symb} + \biasVech\right) \\
                     & = \heaviside\left(\begin{pmatrix}
                                                 \textcolor{ETHGreen}{0} & 0 & 1 & 1 & 0 & 0 & \\
                                                 \textcolor{ETHGreen}{0} & 0 & 0 & 0 & 0 & 0 & \\
                                                 \textcolor{ETHGreen}{1} & 1 & 0 & 0 & 0 & 0 & \\
                                                 \textcolor{ETHGreen}{0} & 0 & 0 & 0 & 0 & 0 & \\
                                                 \textcolor{ETHGreen}{0} & 0 & 0 & 0 & 0 & 0 & \\
                                                 \textcolor{ETHGreen}{1} & 1 & 1 & 1 & 1 & 1 &
                                             \end{pmatrix}
        \begin{pmatrix}
                \textcolor{ETHGreen}{1} \\ 0 \\ 0 \\ 0 \\ 0 \\ 0
            \end{pmatrix} +
        \begin{pmatrix}
                1 & \textcolor{ETHGreen}{0} \\
                0 & \textcolor{ETHGreen}{0} \\
                1 & \textcolor{ETHGreen}{0} \\
                0 & \textcolor{ETHGreen}{0} \\
                0 & \textcolor{ETHGreen}{0} \\
                0 & \textcolor{ETHGreen}{1}
            \end{pmatrix}
        \begin{pmatrix}
                0 \\ \textcolor{ETHGreen}{1}
            \end{pmatrix}  +
        \begin{pmatrix}
                -1 \\
                -1 \\
                -1 \\
                -1 \\
                -1 \\
                -1
            \end{pmatrix}  \right)                                                                             \\
                     & = \heaviside\left(\begin{pmatrix}
                                                 0                       \\
                                                 0                       \\
                                                 \textcolor{ETHGreen}{1} \\
                                                 0                       \\
                                                 0                       \\
                                                 \textcolor{ETHGreen}{1}
                                             \end{pmatrix} +
        \begin{pmatrix}
                0 \\
                0 \\
                0 \\
                0 \\
                0 \\
                \textcolor{ETHGreen}{1}
            \end{pmatrix}  +
        \begin{pmatrix}
                -1 \\
                -1 \\
                -1 \\
                -1 \\
                -1 \\
                -1
            \end{pmatrix}
        \right)
        = \heaviside\left(\begin{pmatrix}
                                  -1                     \\
                                  -1                     \\
                                  \textcolor{ETHBlue}{0} \\
                                  -1                     \\
                                  -1                     \\
                                  \textcolor{ETHGreen}{1}
                              \end{pmatrix}
        \right) = \begin{pmatrix}
                      0 \\
                      0 \\
                      0 \\
                      0 \\
                      0 \\
                      \textcolor{ETHGreen}{1}
                  \end{pmatrix},
    \end{align*}
    which corresponds exactly the configuration in which $\wfsa$ is in state $\stateq_2$ which it arrived to by reading in the symbol $\symb$.

    The probability of the string $\str = \symb$ under the locally-normalized model induced by $\rnn$ can be computed as
    \begin{align*}
        \pLN\left(\str\right) & = \pLN\left(\symb\right) = \pLNSM\left(\symb\mid\bos\right) \pLNSM\left(\eos\mid\symb\right) = \pLNSM\left(\symb\mid\hiddStateZero\right) \pLNSM\left(\eos\mid\hiddState_1\right) \\
                              & = \softmax\left(\outMtx \hiddStateZero\right)_\symb \softmax\left(\outMtx \hiddState_1\right)_\eos                                                                                \\
                              & = \softmax\left(\begin{pmatrix}
                                                        \textcolor{ETHGreen}{\log\left(0.1\right)} & \cdots & -\infty              \\
                                                        \textcolor{ETHGreen}{\log\left(0.9\right)} & \cdots & \log\left(0.5\right) \\
                                                        \textcolor{ETHGreen}{-\infty}              & \cdots & \log\left(0.5\right)
                                                    \end{pmatrix}
        \begin{pmatrix}
                \textcolor{ETHGreen}{1} \\ 0 \\ 0 \\ 0 \\ 0 \\ 0
            \end{pmatrix}
        \right)_\symb \cdot                                                                                                                                                                                       \\
                              & \phantom{=} \softmax\left(\begin{pmatrix}
                                                                  \log\left(0.1\right) & \cdots & \textcolor{ETHGreen}{-\infty}              \\
                                                                  \log\left(0.9\right) & \cdots & \textcolor{ETHGreen}{\log\left(0.5\right)} \\
                                                                  -\infty              & \cdots & \textcolor{ETHGreen}{\log\left(0.5\right)}
                                                              \end{pmatrix}
        \begin{pmatrix}
                0 \\ 0 \\ 0 \\ 0 \\ 0 \\ \textcolor{ETHGreen}{1}
            \end{pmatrix}
        \right)_\eos                                                                                                                                                                                              \\
                              & = \softmax\begin{pmatrix}
                                              {\log\left(0.1\right)} \\
                                              {\log\left(0.9\right)} \\
                                              {-\infty}
                                          \end{pmatrix}_\symb \cdot \softmax\begin{pmatrix}
                                                                                {-\infty}              \\
                                                                                {\log\left(0.5\right)} \\
                                                                                {\log\left(0.5\right)}
                                                                            \end{pmatrix}_\eos = 0.9 \cdot 0.5 = 0.45.
    \end{align*}
\end{example}

\paragraph{Implications for recurrent neural language models.}

\cref{lem:heaviside-rnns-regular,lem:minsky-constr} formalize the equivalence between \hernnAcr{}s and deterministic PFSAs.
A direct corollary of this result is that \hernnAcr{}s are at most as expressive as deterministic PFSAs and, therefore, strictly less expressive as general, non-deterministic, PFSAs.\footnote{General PFSAs are, in turn, equivalent to probabilistic regular grammars and discrete Hidden Markov Models \citep{Icard2020}.}
An example of a very simple non-deterministic PFSA, i.e., a PFSA whose distribution cannot be expressed by an \hernnAcr{} LM, is shown in \cref{fig:nondet-pfsa}.
\begin{figure}[t]
    \centering
    \begin{tikzpicture}[node distance = 15mm]
        \node[state, initial] (q0) [] { $\stateq_{0} / {1}$ };
        \node[state] (q1) [right = of q0, yshift=10mm] { $\stateq_{1}$ };
        \node[state] (q2) [right = of q0, yshift=-10mm] { $\stateq_{2}$ };
        \node[state, accepting] (q3) [right = of q1, yshift=-10mm] { $\stateq_{3} / {1}$ };
        \draw[transition]  (q0) edge[bend left, above, sloped] node{ $\syma/{0.5}$ } (q1)
        (q0) edge[bend right, above, sloped] node{ $\syma/{0.5}$ } (q2)
        (q1) edge[above, loop above] node{ $\symb/{0.9}$ } (q1)
        (q2) edge[below, loop below] node{ $\symb/{0.1}$ } (q2)
        (q1) edge[bend left, above, sloped] node{ $\symc/{0.1}$ } (q3)
        (q2) edge[bend right, above, sloped] node{ $\symc/{0.9}$ } (q3) ;
    \end{tikzpicture}
    \caption{A non-determinizable PFSA. It assigns the string $\syma \symb^n \symc$ the probability $\wfsa\left(\syma \symb^n \symc\right) = 0.5 \cdot 0.9^n \cdot 0.1 + 0.5 \cdot 0.1^n \cdot 0.9$, which can not be expressed as a single term for arbitrary $n \in \Nzero$.}
    \label{fig:nondet-pfsa}
\end{figure}
Furthermore, even if a non-deterministic PFSA can be determinized, the number of states of the determinized machine can be exponential in the size of the non-deterministic one \citep{Buchsbaum1998}.
In this sense, non-deterministic PFSAs can be seen as exponentially compressed representations of finite-state LMs.
However, the compactness of this non-deterministic representation must be ``undone'' using determinization before it can be encoded by an \hernnAcr{}.

While \cref{lem:heaviside-rnns-regular} focuses on \hernnAcr{} LMs and shows that they are finite-state, a similar argument could be made for any RNN whose activation functions map onto a finite set.
This is the case with any implementation of an RNN on a computer with finite-precision arithmetic---in that sense, all deployed RNNLMs are finite-state, albeit very large in the sense of encoding possibly very large weighted finite-state automata.
However, there are a few important caveats with this: firstly, notice that, although finite, the number of states represented by an RNN is \emph{exponential} in the size of the hidden state.
Even for moderate hidden state dimensionalities, this can be very large (hidden states can easily be of size $100$--$1000$).
In other words, one can view RNNs as very compact representations of large deterministic probabilistic finite-state automata whose transition functions are represented by the RNN's update function.
Furthermore, since the topology of this implicit WFSA is completely determined by the update function of the RNN, it can be \emph{learned} very flexibly yet efficiently based on the training data---this is made possible by the \emph{sharing} of parameters across the entire graph of the WFSA instead of explicitly parametrizing every possible transition, as, for example, in \cref{sec:parametrized-wfsa-lms}, or hard-coding the allowed transitions as in \cref{sec:n-gram-models}.
This means that the WFSA is not only represented, but also \emph{parametrized} very efficiently by an RNN.
Nevertheless, there is an important detail that we have somewhat neglected so far: this is the requirement that the simulated WFSA be \emph{deterministic}.

\subsection{Addendum to Minsky's Construction: Lower Bounds on the Space Complexity of Simulating PFSAs with RNNs}
\cref{lem:minsky-constr} shows that \hernnAcr{} LMs are at least as powerful as \dpfsaAcr{}s.
More precisely, it shows that any \dpfsaAcr{} $\wfsa = \wfsatuple$ can be simulated by an \hernnAcr{} LM of size $\bigO{\nstates \nsymbols}$.
In this section, we address the following question: How large does an \hernnAcr{} LM have to be such that it can correctly simulate a \dpfsaAcr{}?
We study the asymptotic bounds with respect to the size of the set of states, $\nstates$, as well as the number of symbols, $\nsymbols$.\footnote{This section is based on the survey by \citet{svete2023recurrent}.}

\paragraph{Asymptotic Bounds in $\nstates$.}
Intuitively, the $2^\hiddDim$ configurations of a $\hiddDim$-dimensional \hernnAcr{} hidden state could represent $2^\hiddDim$ states of a (P)FSA.
One could therefore expect that we could achieve exponential compression of a \dpfsaAcr{} by representing it as an \hernnAcr{} LM.
Interestingly, this is not possible in general: extending work by \citet{Dewdney1977}, \citet{Indyk1995} shows that, to represent an unweighted FSA with an \hernnAcr{}, one requires an \hernnAcr{} of size $\Omega\left(\nsymbols \sqrt{\nstates}\right)$.
This lower bound can be achieved.
For completeness, we present constructions by \citet{Dewdney1977, Indyk1995}, which represent an unweighted FSA with a \hernnAcr{} of size $\bigO{\nsymbols\nstates^{\frac{3}{4}}}$ and $\bigO{\nsymbols\sqrt{\nstates}}$, respectively, next, before giving a lower bound in for the probabilistic case.

\cref{lem:minsky-constr} gives a relatively simple construction of an RNN recognizing a weighted regular language.
However, the resulting RNN is relatively large, with a hidden state of size linear in the number of states of the (deterministic) WFSA recognizing the language, with the additional multiplicative factor in the size of the alphabet.
Note that constructions resulting in smaller RNNs exist, at least for the unweighted case.
For example, for an arbitrary WFSA $\wfsa = \wfsatuple$, \citet{Dewdney1977,Noga1991} present a construction of an RNN with a hidden state of size $\bigo\left(|\states|^{\frac{3}{4}}\right)$ simulating $\wfsa$, whereas \citet{Indyk1995} provides a construction of an RNN with a hidden state of size $\bigo\left(|\states|^{\frac{1}{2}}\right)$.
The latter is also provably a \emph{lower bound} on the number of neurons required to represent an arbitrary unweighted FSA with a Heaviside-activated recurrent neural network \citep{Indyk1995}.
It is not yet clear if this can be generalized to the weighted case or if Minsky's construction is indeed optimal in this setting.
This is quite interesting since one would expect that an RNN with a hidden state of size $\hiddDim$ can represent up to $2^\hiddDim$ individual states (configurations of the $\hiddDim$-dimensional vector).
However, the form of the transition function with the linear transformation followed by a Heaviside activation limits the number of transition functions that can be represented using $\hiddDim$ dimensions, resulting in the required exponential increase in the size of the hidden state.

Minsky's construction (\cref{lem:minsky-constr}) describes how to represent a \dpfsaAcr{} $\wfsa$ with a \hernnAcr{} of size linear in the number of $\wfsa$'s states.
Importantly, the encoding of the FSA transition function (taken from Minsky's original construction) is decoupled from the parameter defining the probability distribution, $\outMtx$.
This section describes two asymptotically more space-efficient ways of constructing the component simulating the transition function.
They originate in the work by \citet{Dewdney1977}, who showed that an unweighted FSA $\wfsa = \fsatuple$ can be represented by an \hernnAcr{} of size $\bigO{\nsymbols\nstates^{\frac{3}{4}}}$.
Using the same ideas, but a specific trick to compress the size of the processing layer of the RNN further, \citet{Indyk1995} reduced this bound to $\bigO{\nsymbols \sqrt{\nstates}}$, which, as discussed in \cref{sec:space-bounds}, is asymptotically optimal.
Naturally, as shown in \cref{sec:space-bounds}, the space-efficiency gain can not be carried over to the weighted case---that is, the space-efficiency is asymptotically overtaken by the output matrix $\outMtx$.
Nevertheless, for a more complete treatment of the subject, we cover the two compressed constructions of the \hernnAcr{} simulating an unweighted FSA in this section in our notation.
Importantly, given a \dpfsaAcr{}, we focus only on the \emph{underlying FSA}, i.e., the unweighted transition function of the automaton, since by \cref{thm:pfsa-rnn-lower-bound}, the compression can only be achieved with components representing that part of the automaton.

\subsubsection{Dewdney's Construction} \label{sec:dewdney}
This section describes the construction due to \citet{Dewdney1977} in our notation.
Since some of the parts are very similar to the construction due to \citet{Indyk1995}, those parts are reused in \cref{sec:indyk} and introduced more generally.

\paragraph{Representing states of the FSA.}
Let $\wfsa = \fsatuple$ be a deterministic FSA.
Recall that Minsky's construction encodes the $\wfsa$'s current state as a one-hot encoding of the state-symbol pair.
The construction due to \citet{Dewdney1977}, on the other hand, represents the states separately from the symbols.
It encodes the states with \emph{two-hot representations} by using the coefficients of what we call a square-root state representation.
This results in representations of states of size $\bigO{\sqrt{\nstates}}$.
The input symbols are incorporated into the hidden state \emph{separately}.\footnote{This again adds a factor $\nsymbols$ to the size of the hidden state, as we discuss later.}
\begin{definition}{Square-root state representation}{}
    Let $\wfsa = \fsatuple$ be an FSA and $\srNstates \defeq \ceil{\sqrt{\nstates}}$.
    We define the \defn{square-root state representation} of $\wfsa$'s states $\stateq \in \states$ as\footnote{Notice that $\phitwoFun{\stateq}$ represents the coefficients of the expression of $\stateq \in \N$ in base $\srNstates$.}
    \begin{equation}
        \phitwoFun{\stateq} \defeq \left(\floor{\frac{\stateq}{\srNstates}}, \stateq\!\mod\!\srNstates\right).
    \end{equation}
    We denote the inverse of $\phitwo$ with $\phitwoInv$ and further define for $k \in \Zmod{\srNstates}$
    \begin{equation}
        \phitwoInvFun{k, \cdot} \defeq \set{\stateq \in \states \mid \evvarphi_0 = k \text{ where } \phiCoor = \phitwoFun{\stateq}}
    \end{equation}
    and $\phitwoInvFun{\cdot, k}$ analogously.
\end{definition}
Specifically, we will denote $\phitwoInvFun{k, \cdot}$ and $\phitwoInvFun{\cdot, k}$ with $k$ in the $j\textsuperscript{th}$ position (with $\idxj \in \Zmod{2}$, $0$ for $\phitwoInvFun{k, \cdot}$ and $1$ for $\phitwoInvFun{\cdot, k}$) as $\compMtx{\idxk}{\idxj}$.

We can think of the function $\phitwo$ as representing states of the FSA in a two-dimensional space $\Zmod{\srNstates} \times \Zmod{\srNstates}$.
However, to efficiently simulate $\wfsa$ with an \hernnAcr{}, it is helpful to think of $\phitwoFun{\stateq}$ in two different ways: as a vector $\vv \in \Nzero^{2 \nstates}$, or as a matrix in $\BQQ$ in the following sense.
\begin{definition}{Vector and matrix state representations}{vector-matrix-state-representation}
    Given a square-root state representation function $\phitwo$, we define the \defn{vector representation} of the state $\stateq \in \states$ as the vector $\qvectv{\stateq} \in \BtwoQ$ with
    \begin{align}
        \qvectv{\stateq}_{\evvarphi_0}              & = 1  \\
        \qvectv{\stateq}_{\srNstates + \evvarphi_1} & = 1,
    \end{align}
    where $\phiCoor = \left(\evvarphi_0, \evvarphi_1\right) = \phitwoFun{\stateq}$, and all other entries $0$.
    Furthermore, we define the \defn{matrix representation} of the state $\stateq \in \states$ as the matrix $\mB \in \BQQ$ with
    \begin{equation}
        \qMtx{\stateq}_{\evvarphi_0 \evvarphi_1} = 1
    \end{equation}
    and all other entries $0$.
\end{definition}
Dewdney's construction also heavily relies on the representations of sets of states.
We define those additively.
\begin{definition}{Matrix and vector representation of state sets}{matrix-state-set-representation}
    Let $\stateSet \subseteq \states$ be a set of states.
    We define the vector representation of $\stateSet$ as the vector
    \begin{equation}
        \qvectv{\stateSet} \defeq \bigvee_{\stateq \in \states} \qvectv{\stateq}.
    \end{equation}
    Similarly, we define the matrix representation of $\stateSet$ as the matrix
    \begin{equation}
        \qMtx{\stateSet} \defeq \bigvee_{\stateq \in \states} \qMtx{\stateq}.
    \end{equation}
\end{definition}

\begin{figure}[t]
    \centering
    \begin{tikzpicture}[node distance = 10mm]
        \node[state] (q0) [] { $0$ };
        \node[state] (q1) [right = of q0, yshift=-10mm] { $1$ };
        \node[state] (q2) [right = of q0, yshift=10mm] { $2$ };
        \draw[transition]  (q0) edge[below, sloped, bend right] node{ $\syma$ } (q1)
        (q0) edge[above, sloped, bend left] node{ $\symb$ } (q2)
        (q1) edge[right, bend right] node{ $\syma$ } (q2)
        (q2) edge[above, sloped, bend left] node{ $\symb$ } (q0)
        (q1) edge[right, loop right] node{ $\symb$ } (q1) ;
    \end{tikzpicture}
    \caption{An example of a fragment of an FSA.}
    \label{fig:dewdney-def-fsa}
\end{figure}
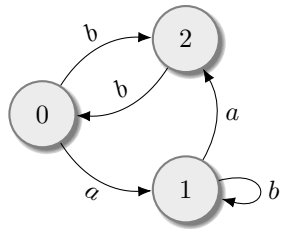

To help understand the above definitions, we give an example of an FSA and the representations of its states.
\begin{example}{Dewdney's construction}{}
    Consider the FSA in \cref{fig:dewdney-def-fsa}, for which $\srNstates = \ceil{\sqrt{\nstates}} = \ceil{\sqrt{3}} = 2$, meaning that
    \begin{equation}
        \phitwoFun{0} = \left(0, 0\right)
        \qquad \phitwoFun{1} = \left(0, 1\right)
        \qquad \phitwoFun{2} = \left(1, 0\right),
    \end{equation}
    resulting in the state-to-vector mapping\footnote{Despite the notation $(\ldots \mid \ldots)$, we assume we are working with column vectors.}
    \begin{align}
        \qvectv{0} & = \begin{pmatrix}
                           1 & 0 & \mid & 1 & 0
                       \end{pmatrix} \\
        \qvectv{1} & = \begin{pmatrix}
                           1 & 0 & \mid & 0 & 1
                       \end{pmatrix} \\
        \qvectv{2} & = \begin{pmatrix}
                           0 & 1 & \mid & 1 & 0
                       \end{pmatrix},
    \end{align}
    and the state-to-matrix mapping
    \begin{equation}
        \qMtx{0} = \begin{pmatrix}
            1 & 0 \\
            0 & 0
        \end{pmatrix}
        \qquad \qMtx{1} = \begin{pmatrix}
            0 & 1 \\
            0 & 0
        \end{pmatrix}
        \qquad \qMtx{2} = \begin{pmatrix}
            0 & 0 \\
            1 & 0
        \end{pmatrix}.
    \end{equation}
    The two components of the vector representations separated by ``$\mid$'' denote the two halves of the representation vectors, corresponding to the two components of $\phitwoFun{\stateq}$.

\end{example}

\paragraph{High-level idea of Dewdney's construction.}
Given these definitions, the intuition behind Dewdney's construction of an \hernnAcr{} simulating an FSA $\wfsa$ is the following:
\begin{enumerate}
    \item Represent $\wfsa$'s states as vectors in $\Btwos$, or, equivalently, matrices in $\Bss$.
    \item For each $\stateq \in \states$, construct the matrix representation of the set of $\sym$-predecessors $\qMtx{\qPreds{\stateq; \sym}}$ for all $\sym \in \alphabet$.
    \item To simulate $\wfsa$'s transition function $\trans$, compare the representation of the current state $\qt$ with all constructed predecessor matrices $\qMtx{\qPreds{\stateq; \symt}}$ given the current input symbol $\symt$.
          Activate the two-hot representation of the (unique) state $\qtplus$ for which the representation of $\qt$ was detected in $\qtplus$'s predecessor matrix for symbol $\symt$, $\qMtx{\qPreds{\qtplus; \symt}}$.
\end{enumerate}

\paragraph{Simulating the transition function of an FSA by detecting preceding states.}
We elaborate on the last point above since it is the central part of the construction.\footnote{Later, we will see that \citet{Indyk1995} uses the exact same idea for simulating $\trans$.}
The idea of simulating the transition function $\trans$ is reduced to detecting \emph{whose predecessor} given the current input symbol $\symt$ is currently active---naturally, this should be the state active at $\tstep + 1$.
Concretely, consider again the FSA $\wfsa$ in \cref{fig:dewdney-def-fsa}.
The predecessors of the three states, indexed by the incoming symbols are: for $0$ $\left\{\symb: 2\right\}$, for $1$ $\left\{\syma: 1, \symb: 0\right\}$, and for $2$ $\left\{\syma: 1, \symb: 0\right\}$.
Suppose that at some time $\tstep$, $\wfsa$ is in state $0$ and is reading in the symbol $\symb$.
Then, since the state $0$ is the $\symb$-predecessor of the state $2$, we know that at time $\tstep + 1$, $\wfsa$ will be in state $2$.
This principle can be applied more generally: to determine the state of an FSA at time $\tstep + 1$, we simply have to somehow detect whose \emph{predecessor} is active at time $\tstep$ \emph{given the current input symbol} at time $\tstep$.

The crux of Dewdney's construction is then the following:\footnote{Again, the same applies to \citet{Indyk1995}.} How do we, using only the Elman update rule, determine whose $\symt$-predecessor is active at time $\tstep$?
This can be done by \emph{detecting} which predecessor matrix $\qMtx{\qPreds{\stateq; \symt}}$ the representation of the current state $\qt$ is included in in the sense that if $\phitwoFun{\qt} = \phiCoor$, it holds that $\qMtx{\qPreds{\stateq; \symt}}_{\evvarphi_0\evvarphi_1} = 1$.
To be able to formally talk about the detection of a representation in a set of predecessors, we define several notions of \defn{matrix detection}.

Informally, we say that a matrix is easily detectable if the presence of its non-zero elements can be detected using a single neuron in the hidden layer of a \hernnAcr{}.
\begin{definition}{Easily detectable matrices}{}
    Let $\mB \in \BDD$ be a binary matrix.
    We say that $\mB$ is \defn{easily detectable} if there exist $\vw \in \Q^{2 D}$ and $b \in \Q$ (neuron coefficients) such that
    \begin{equation}
        \sigmoidFun{\innerProd{\ve_{ij}}{\vw} + b} = 1 \iff \emB_{ij} = 1,
    \end{equation}
    where $\ve_{ij} = \begin{pmatrix} \ve_i & \mid & \ve_j \end{pmatrix}$ refers to the $2 D$-dimensional vector with $1$'s at positions $i$ and $D + j$.
    In words, this means that the neuron defined by $\vw, b$ fires on the input $\ve_{ij}$ if and only if $\emB_{ij} = 1$.
\end{definition}
We define detectable matrices as the matrices which can be detected using a conjunction of two neurons.
\begin{definition}{Detectable matrices}{}
    Let $\mB \in \BDD$ be a binary matrix.
    We say that $\mB$ is \defn{detectable} if there exist $\vw_1, \vw_2 \in \Q^{2 D}$ and $b_1, b_2 \in \Q$ such that
    \begin{equation}
        \sigmoidFun{\innerProd{\ve_{ij}}{\vw_1} + b_1} = 1 \wedge \sigmoidFun{\innerProd{\ve_{ij}}{\vw_2} + b_2} = 1 \iff \emB_{ij} = 1.
    \end{equation}
\end{definition}
Furthermore, we say that a matrix is (easily) \defn{permutation-detectable} if there exist permutation matrices $\mP$ and $\mQ$ such that $\mP \mB \mQ$ is (easily) detectable.

Intuitively, this means that one can effectively replace an easily detectable matrix $\mB$ with \emph{a single neuron}: instead of specifying the matrix explicitly, one can simply detect if an entry $\emB_{ij}$ of $\mB$ is $1$ by passing $\ve_{ij}$ through the neuron and seeing if it fires.
This reduces the space complexity from $D^2$ to $2D$.
Similarly, one can replace a detectable matrix with \emph{two} neurons.
As shown in \cref{fact:and}, the required conjunction of the two resulting neurons can then easily be performed by a third (small) neuron, meaning that a detectable matrix is effectively represented by a two-layer MLP.

An example of easily detectable matrices are the so-called \defn{northwestern} matrices.
\begin{definition}{Northwestern matrix}{}
    A matrix $\mB \in \BDD$ is \defn{northwestern} if there exists a vector $\balpha$ with $|\balpha| = D$ and $D \geq \alpha_1 \geq \ldots \geq \alpha_D \geq 0$ such that
    \begin{equation}
        \emB_{ij} = 1 \iff j \leq \alpha_i.
    \end{equation}
\end{definition}
Intuitively, northwestern matrices contain all their ones contiguously in their upper left (northwest) corner.
An example of a northwestern matrix for $\balpha = \begin{pmatrix}
        2 & 1 & 1
    \end{pmatrix}$ is
\begin{equation}
    \mB = \begin{pmatrix}
        1 & 1 & 0 \\ 1 & 0 & 0 \\ 1 & 0 & 0
    \end{pmatrix}.
\end{equation}

\begin{lemma}{}{nw-matrices-easily-detectable}
    Northwestern matrices are easily detectable.
\end{lemma}
\begin{proof}
    Define
    \begin{equation*}
        \vw \defeq \begin{pmatrix}
            \balpha & \mid & D & \ldots & 1
        \end{pmatrix}
    \end{equation*}
    and $b = -D$.
    It is easy to see that for any $\ve_{ij}$ where $\mB_{ij} = 1$, it holds that
    \begin{align*}
         & \innerProd{\ve_{ij}}{\vw} = \alpha_i + (D - j + 1) \geq j + D - j + 1 = D + 1                             \\
         & \implies \heavisideFun{\innerProd{\ve_{ij}}{\vw} + b} = \heavisideFun{\innerProd{\ve_{ij}}{\vw} - D} = 1.
    \end{align*}

    On the other hand, for $\emB_{ij} = 0$, we have
    \begin{align*}
         & \innerProd{\ve_{ij}}{\vw} = \alpha_i + (D - j + 1) < j + D - j + 1 = D                                    \\
         & \implies \heavisideFun{\innerProd{\ve_{ij}}{\vw} + b} = \heavisideFun{\innerProd{\ve_{ij}}{\vw} - D} = 0.
    \end{align*}
\end{proof}

A more general useful class of detectable matrices are line matrices \citep{Dewdney1977}.
\begin{definition}{Line matrix}{line-matrix}
    A binary matrix $\mB \in \BDD$ is a \defn{line matrix} if any of the following conditions hold:
    \begin{enumerate}
        \item All $\mB$'s ones lie either in the same row ($\mB$ is a \defn{row matrix}) or in the same column ($\mB$ is a \defn{column matrix}).
        \item $\mB$ is a \emph{transversal}, i.e., a matrix in which there is at most one $1$ in any column and row.
    \end{enumerate}
\end{definition}

\begin{lemma}{}{lines-detectable}
    Row and column matrices are easily permutation-detectable.
\end{lemma}
\begin{proof}
    Let $i, N \in \Zmod{D}$ and $\mB$ be a row matrix with $\mB_{ij_n} = 1$ for $n \in \Zmod{N}$, i.e., a row matrix with all its ones in the $i^\text{th}$ row.
    Define $\mP \in \BDD$ as $\emP_{1 i} = 1$ and $0$ elsewhere and $\mQ \in \BDD$ with $\mQ_{j_n n} = 1$ and $0$ elsewhere.
    Then, $\mP \mB \mQ$ contains all its $1$ in its northwestern corner (contiguously in the first row) and is thus easily detectable.
    Let $\vw \defeq \begin{pmatrix} \balpha & \mid & D & \ldots & 1 \end{pmatrix}$, $b = D$ be the neuron weights from \cref{lem:nw-matrices-easily-detectable}.
    Define $\vw' \defeq \begin{pmatrix} \mP^\top \balpha & \mid & \mQ(D & \ldots & 1) \end{pmatrix}$, $b' = D$.
    It is easy to see that this ``rearranges'' the components of the neuron recognizing the northwestern matrix $\mP \mB \mQ$ to make them recognize the original matrix, meaning that the neuron defined by $\vw'$ and $b'$ recognizes the line matrix.
    The proof for a column matrix is analogous.
\end{proof}

\begin{lemma}{}{trans-detectable}
    Transversals are permutation-detectable.
\end{lemma}
\begin{proof}
    The core idea of this proof is that every transversal can be permuted into a diagonal matrix, which can be written as a Hadamard product of a lower-triangular and an upper-triangular matrix.

    Let $\mB$ be a transversal.
    Pre-multiplying $\mB$ with its transpose $\mP \defeq \mB^\top$ results in a diagonal matrix.
    It is easy to see that $\mP \mB$ can be written as a Hadamard product $\mH_1 \otimes \mH_2$ of a lower-triangular matrix $\mH_1$ and an upper-triangular matrix $\mH_2$.
    Both are easily permutation detectable.
    A conjunction of the neurons detecting $\mH_1$ and $\mH_2$ (again, performed by another neuron) detects the original matrix $\mB$.
    In the following, we will refer to $\mH_1$ and $\mH_2$ as the \emph{factors} of the transversal.
\end{proof}

Crucially, any binary matrix $\mB \in \BDD$ can be decomposed into a set of line matrices $\sB$ whose disjunction is $\mB$: $\bigvee_{\mM \in \sB} \mM = \mB$.
It is easy to see that $\mB_{ij} = 1$ if and only if there exists $\mM \in \sB$ such that $\mM_{ij} = 1$.
This means that non-zero entries of \emph{any} $\mB \in \BDD$ decomposed into the set of line matrices $\sB$ can be detected using an MLP in two steps:
\begin{enumerate}
    \item Detect the non-zero entries of the individual line matrices from the decomposition $\sB$ (which are, as shown above, detectable).
    \item Take a disjunction of the detections of the individual line matrices to result in the activation of the original matrix.
\end{enumerate}
The disjunction can again be performed by applying another 2-layer MLP to the activations of the line matrices.
An important consideration in both Dewdney's as well as Indyk's construction later will be \emph{how large} $\sB$ has to be.

\paragraph{Using matrix decomposition and detection for simulating the transition function.}
We now describe how Dewdney's construction uses matrix detection based on the decomposition of matrices into line matrices to simulate an FSA using an \hernnAcr{}.
From a high level, the update steps of the \hernnAcr{} will, just like in Minsky's construction, simulate the transition function of the simulated FSA.
However, in contrast to the Minsky construction, in which each transition step in the FSA was implemented by a \emph{single} application of the Elman update rule, here, a single transition in the FSA will be implemented using \emph{multiple} applications of the Elman update rule, the end result of which is the activation of the two-hot representation of the appropriate next state.
Nonetheless, there are, abstractly, two sub-steps of the update step, analogous to the Minsky construction (cf. \cref{fig:minsky-overview}):
\begin{enumerate}
    \item Detect the activations of all possible next states, considering any possible input symbol (performed by the term $\recMtx \hiddStatet$ in Minsky's construction).
    \item Filter the activations of the next states by choosing only the one transitioned into by a $\symt$-transition (performed by conjoining with the term $\inMtx \onehot{\symt}$ in Minsky's construction).
\end{enumerate}
The novelty of Dewdney's construction comes in the first sub-step: How can the Elman update step be used to activate the two-hot representation of $\qt$'s out-neighborhood?
As alluded to, this relies on the pre-computed predecessor matrices $\qPreds{\stateq; \sym}$ (cf. \cref{def:vector-matrix-state-representation}).
The predecessor matrices of individual states are compressed (disjoined) into component-activating matrices, the representation matrices of the predecessors of specific \emph{sets} of states (cf. \cref{def:matrix-state-set-representation}), defined through the function $\phitwo$ in the following sense.
\begin{definition}{Component-activating matrix}{}
    A \defn{component-activating matrix} is the representation matrix $\mB_{\idxj, \sym, \idxk} \defeq \qMtx{\qPreds{\compMtx{\idxk}{\idxj}; \sym}}$ for some $\idxk \in \Zmod{\frNstates}$ and $\idxj \in \Zmod{2}$.
\end{definition}
Intuitively, the component-activating matrix $\mB_{\idxj, \sym, \idxk}$ is the result of the disjunction of the matrix representations of all $\sym$-predecessors $\stateq$ of all states $\stateq'$ whose $\idxj\textsuperscript{th}$ component of the vector $\phitwoFun{\stateq'}$ equals $\idxk$.
This results in $2 \nsymbols \srNstates$ matrices.
They can be pre-computed and naturally depend on the transition function $\trans$.
The name component-activating matrix is inspired by the fact that each of the matrices ``controls'' the activation of one of the $2 \nsymbols \srNstates$ neurons in a specific sub-vector of the \hernnAcr{} hidden state.
That is, each component-activating matrix controls a particular dimension, indexed by the tuple $\left(\idxj, \sym, \idxk\right)$ for $\idxj \in \B, \sym \in \alphabet, \idxk \in \Zmod{\srNstates}$, in the \defn{data sub-vector} of the \hernnAcr{} hidden state.
As we will see shortly, they contain all the information required for simulating $\wfsa$ with a \hernnAcr{}.

\begin{figure*}
    \centering
    \begin{subfigure}{\textwidth}
        \includegraphics[width=0.9\textwidth]{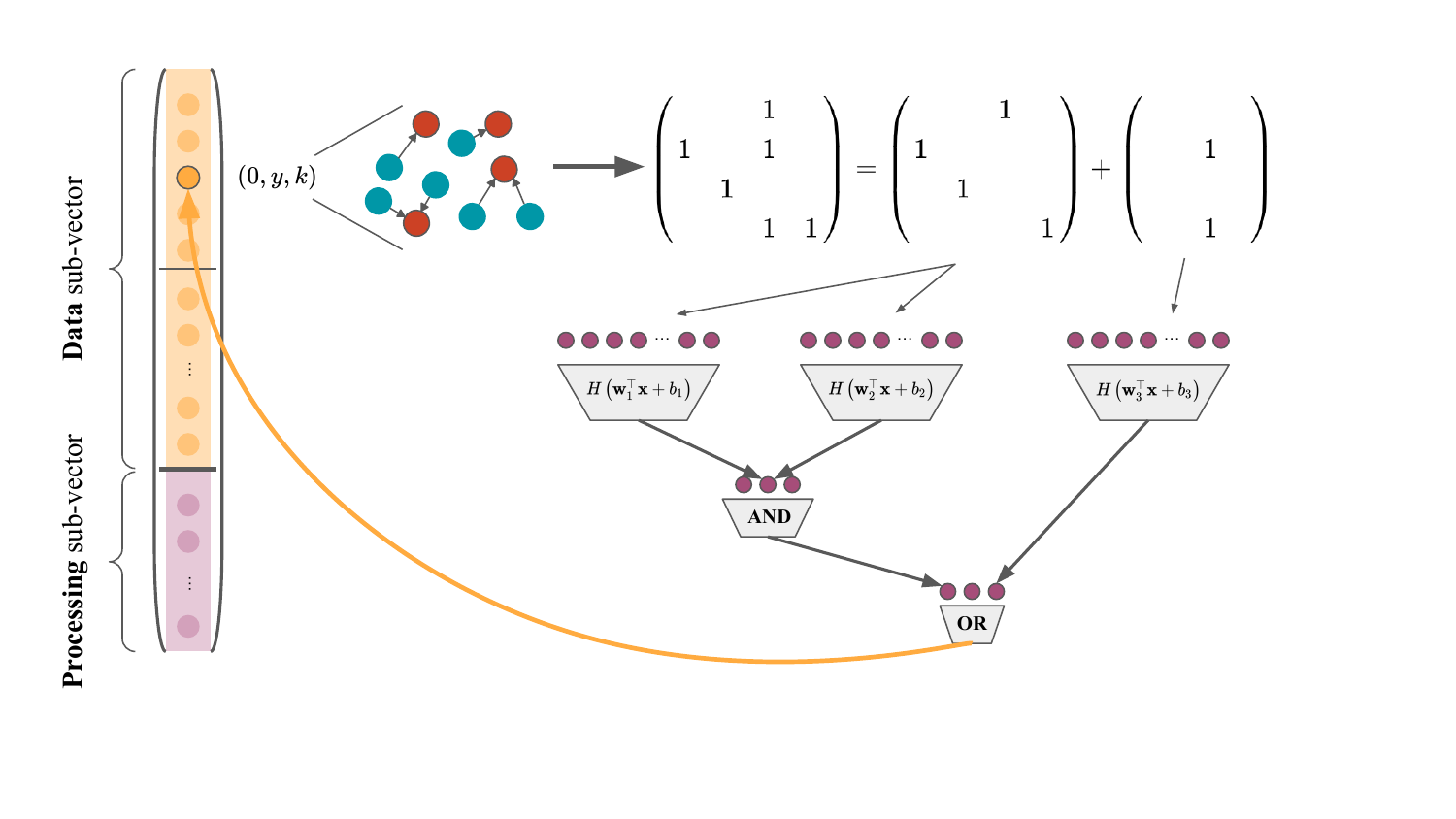}
        \caption{High-level overview of Dewdney's construction.
            The highlighted \textcolor{ETHBronze}{orange} neuron in the representation of the state from the data sub-vector corresponds to the activation of one of the components of the \textcolor{ETHRed}{red} states (which have in common that their $0\textsuperscript{th}$ component of $\phitwoFun{\stateq}$ is the same).
            The matrix corresponding to the disjunction of the representations of their \emph{$\sym$-predecessors} (\textcolor{ETHBlue}{blue} states) is decomposed into two line matrices---a transversal and a column matrix.
            The non-zero elements of the former can be detected by a conjunction of two neurons while the non-zero elements of the latter can be detected directly by a single neuron.
            Those activations are then disjoined to result in the activation in the \textcolor{ETHBronze}{orange} neuron.
            The \textcolor{ETHPurple}{purple} neurons in the processing sub-vector are composed of the neurons in the networks implementing the detection of line matrices and their conjunctions and disjunctions (also shown in \textcolor{ETHPurple}{purple}).}
        \label{fig:dewdney-high-level}
    \end{subfigure}
    \begin{subfigure}{0.9\textwidth}
        \centering
        \begin{tikzpicture}
            \matrix[matrix of math nodes, column sep=0.6cm, row sep=5mm, nodes={minimum width=1.5cm}] (mat) {
                \begin{tikzpicture}[node distance = 6mm]
                    \footnotesize
                    \node[state, fill=ETHPurple!30, draw=ETHPurple!80] (q){ $\stateq$ };
                    \node[state] (q1) [above right = of q] { $\stateq'$ };
                    \node[state] (q2) [below right = of q] { $\stateq''$ };
                    \draw[transition]   (q) edge[above, sloped] node{ $\syma$ } (q1)
                    (q) edge[above, sloped] node{ $\symb$ } (q2);
                \end{tikzpicture}
                                            &
                                            &
                \begin{tikzpicture}[node distance = 6mm]
                    \footnotesize
                    \node[state] (q){ $\stateq$ };
                    \node[state, fill=ETHPetrol!20] (q1) [above right = of q] { $\stateq'$ };
                    \node[state, fill=ETHPetrol!20] (q2) [below right = of q] { $\stateq''$ };
                    \draw[transition]   (q) edge[ETHPetrol, above, sloped] node{ $\syma$ } (q1)
                    (q) edge[ETHPetrol, above, sloped] node{ $\symb$ } (q2);
                \end{tikzpicture} &
                \begin{tikzpicture}[node distance = 6mm]
                    \footnotesize
                    \node[state] (q){ $\stateq$ };
                    \node[state, fill=ETHGreen!30, draw=ETHGreen!80] (q1) [above right = of q] { $\stateq'$ };
                    \node[state] (q2) [below right = of q] { $\stateq''$ };
                    \draw[transition]   (q) edge[ETHGreen, above, sloped] node{ $\syma$ } (q1)
                    (q) edge[above, sloped] node{ $\symb$ } (q2);
                \end{tikzpicture}   \\
                \begin{pmatrix}
                    \textcolor{ETHRed}{\qvectv{\stateq}} \\ \vp
                \end{pmatrix} &
                \begin{pmatrix}
                    \qvectv{\stateq} \\ \textcolor{ETHRed}{\vp'}
                \end{pmatrix} &
                \begin{pmatrix}
                    \textcolor{ETHRed}{\qvectv{\set{\stateq', \stateq''}}} \\ \vp''
                \end{pmatrix} &
                \begin{pmatrix}
                    \textcolor{ETHRed}{\qvectv{\stateq'}} \\ \vp'''
                \end{pmatrix}   \\
            };

            \foreach \i in {1,...,4} {
                    \node at ([yshift=0.25cm]mat-2-\i.north) {\small Phase \i};
                }

            \foreach \i in {1,...,3} {
                    \draw [->,thick] (mat-2-\i.east) -- (mat-2-\the\numexpr\i+1\relax.west);
                }
        \end{tikzpicture}
        \caption{A high-level illustration of how the transition function of the FSA is implemented in Dewdney's construction on an example of an FSA fragment, where the simulated automaton is initially in the state $\stateq$ and reads the symbol $\syma$, transitioning to $\stateq'$.
            The components whose changes are relevant at a given step are \textcolor{ETHRed}{highlighted}.
            Starting in the state $\stateq$, which is stored in the data sub-vector $\qvectv{\stateq}$, in the first sub-step, the processing bits of the appropriate line matrices are activated ($\vp'$).
            Next, the activated line matrices are used to activate the representations of all the states in the out-neighborhood of $\stateq$ in the data sub-vector ($\qvectv{\set{\stateq', \stateq''}}$).
            Lastly, these representations are conjoined with the states reachable by the symbol $\syma$, resulting in the representation of the state $\stateq$ in the data sub-vector ($\qvectv{\stateq}$).
        }
        \label{fig:dewdney-phases}
    \end{subfigure}
\end{figure*}

To define the transition function of the \hernnAcr{} simulating $\wfsa$, all $2 \nsymbols \srNstates$ component-activating matrices are decomposed into permutation-detectable line matrices (cf. \cref{def:line-matrix}) whose activations are combined (disjoined) into the activations of individual component-activating matrices.
Analogously to above, we will denote the sets of line matrices decomposing the component-activating matrices as $\sB_{\idxj, \sym, \idxk}$, i.e., $\mB_{\idxj, \sym, \idxk} = \bigvee_{\mM \in \sB_{\idxj, \sym, \idxk}} \mM$.
The dimensions of the hidden state corresponding to the activations of the line matrices before they are combined into the activations of the component-activating matrices form the \defn{processing sub-vector} of the \hernnAcr{} hidden state since they are required in the pre-processing steps of the update step to determine the activation of the actual hidden state.
This is schematically drawn in \cref{fig:dewdney-high-level}.

For any component-activating matrix $\mB$ decomposed into the set of line matrices $\sB$, we know by \cref{lem:lines-detectable,lem:trans-detectable} that all $\mM \in \sB$ are detectable by a single-layer MLP.
By adding an additional layer to the MLP, we can disjoin the detections of $\mM \in \sB$ into the detection of $\mB$.
More abstractly, this MLP, therefore, detects the activation of \emph{one} of the $2 \nstates \srNstates$ cells of the data sub-vector of the \hernnAcr{} hidden state---all of them together then form the two-hot encoding of all possible next states of the FSA (before taking into account the input symbol).
Designing $2 \nstates \srNstates$ such single-values MLPs, therefore, results in an MLP activating the two-hot representations of all possible next states of the simulated FSA.
Conjoining these activations with the input symbol, analogously to how this is done in the Minsky construction, results in the activation of the two-hot representation of only the actual next state of the simulated FSA.
This is illustrated in \cref{fig:dewdney-phases}.

\paragraph{High-level overview of simulating a transition.}
In summary, after decomposing all the component-activating matrices into the sets $\sB_{\idxj, \sym, \idxk}$, the detection of all candidate next states (before considering the input symbol) in the update step of \hernnAcr{} is composed of the following sub-steps.
\begin{enumerate}
    \item Compute the activations of the two \emph{factors} of all the transversals in $\sB_{\idxj, \sym, \idxk}$ for all $\idxj, \sym, \idxk$ (\cref{lem:trans-detectable}).
    \item Conjoin the activations of the two factors into the activations of the transversals (\cref{lem:trans-detectable}).
    \item Compute the activations of the column and row matrices in $\sB_{\idxj, \sym, \idxk}$ for all $\idxj, \sym, \idxk$  (\cref{lem:lines-detectable}).
    \item Disjoin of the activations of all the line matrices (transversals, row, and column matrices) in $\sB_{\idxj, \sym, \idxk}$ for all $x, \sym, \idxk$ to compute the activations of all $2 \nsymbols \srNstates$ component-activatimg matrices.
\end{enumerate}
This results in the activation of the two-hot representations of all possible next states (i.e., the entire out-neighborhood of $\qt$).
In the last sub-step of the \hernnAcr{} update step, these are conjoined with the representation of the current input symbol.
This step is very similar to the analogous stage in Minsky's construction, with the difference that here, the non-zero entries of the vector $\inMtx \hiddStatet$ must cover the two-hot representations of the states with an incoming $\symt$-transition.
This conjunction then ensures that among all the states in the out-neighborhood of $\qt$, only the one reached by taking the $\symt$-transition will be encoded in $\hiddStatetplus$.
The construction just described can be summarized by the following lemma.\footnote{To formally prove it is correct, we would have to follow a similar set of steps to how the correctness of Minsky's construction (\cref{lem:minsky-transitions}) was proved. We omit this for conciseness.}
\begin{lemma}{}{}
    Let $\wfsa = \fsatuple$ be a deterministic FSA.
    Then, Dewdney's construction results in a \hernnAcr{} correctly simulating $\wfsa$'s transition function, i.e, $\hToQFun{\hiddStatet} = \qt$ for all $\tstep$.
\end{lemma}

This shows that Dewdeny's construction correctly encodes the FSA in a \hernnAcr{}.
However, its space efficiency remains to be determined.
As mentioned above, working with two-hot representations of the states means that the data sub-vector is of size $\bigO{\nsymbols \sqrt{\nstates}}$.
However, the construction also requires a number of processing dimensions in the processing sub-vector.
To understand the full complexity of the construction, we have to determine the maximal number of processing bits in the \hernnAcr{}.
The first step to the answer is contained in the following lemma, which describes the number of line matrices required to cover an arbitrary binary matrix.
It lies in the core of the efficiency of Dewdney's construction.
\begin{lemma}{}{line-decomp-lower-bound}
    Let $\mB \in \BDD$ with $N^2$ elements equalling $1$.
    Then, there exists a decomposition $\sB$ of $\mB$ into at most $2 N$ line matrices such that $\bigvee_{\mM \in \sB} \mM = \mB$.
\end{lemma}
\begin{proof}
    Based on \citet{Dewdney1977}.
    Define the sequence of transversals $\mT_1, \mT_2, \ldots$ where $\mT_\idxi$ is the transversal containing the maximum number of ones in the matrix $\mB_\idxi \defeq \mB - \bigvee_{\idxj = 1}^{\idxi - 1} \mB_{\idxj}$.
    The transversal containing the maximal number of ones can be found using the maximum matching algorithm.
    Continue this sequence until there are no more ones in $\mB_\idxi$.
    The number of ones in the matrices $\mB_\idxi$, $\norm{\mB_\idxi}_1$, forms a (weakly) decreasing sequence.

    If there are at most $2 N$ transversals in the sequence, the lemma holds.
    Otherwise, we compare the functions $\func\left(\idxi\right) \defeq \norm{\mT_\idxi}_1$ and $\funcg\left(\idxi\right) \defeq 2N - \idxi$.
    \begin{itemize}
        \item If $\func\left(\idxi\right) > \funcg\left(\idxi\right)$ for all $\idxi = 1, \ldots, N$, then $\sum_{\idxi = 1}^{N}\func\left(\idxi\right) = \sum_{\idxi = 1}^{N}\norm{\mT_\idxi}_1 > \sum_{\idxi = 1}^{N} 2N - \idxi = 2N^2 - \frac{1}{2} N (N + 1) \geq N^2$.
              However, the transversals in the decomponsition cannot contain more ones than the original matrix.
        \item We conclude that for some $\idxi \leq N$, $\func\left(\idxi\right) \leq \funcg\left(\idxi\right)$.
              Let $\idxi_0$ be the first such index in $1, \ldots, N$ and $\sL_1 \defeq \set{\mT_1, \ldots, \mT_k}$.
              Since the maximum number of independent ones (in the sense that at most one appears in a single row/column) in $\mB_{\idxi_0 - 1}$ is $\norm{\mT_{\idxi_0}}_1 \leq 2N - \idxi_0$ (those are chosen by the maximum transversal $\mT_{\idxi_0}$).
              By K\"onig's theorem \citep{szárnyas2020graphs}, there is a set of at most $2N - \idxi_0$ column or row matrices $\sL_2 \defeq \set{\mL_{1}, \ldots \mL_{\idxk}}$ with $k \leq 2N - \idxi_0$ which cover $\mB_{\idxi_0 - 1}$.\footnote{Intuitively, since all ones are contained within $\leq 2N - \idxi_0$ rows or columns, they can be simply covered by matrices containing those.}
              Therefore, $\sL \defeq \sL_1 \cup \sL_2$ constitutes a valid cover of $\mB$ with $\leq N + 2N - \idxi_0 = \bigO{N}$ matrices.
    \end{itemize}
\end{proof}
We will denote the number of matrices in the line decomposition of a matrix $\mB$ constructed by the greedy procedure from \cref{lem:line-decomp-lower-bound} as $\nLines{\mB}$.
Connecting this lemma to Dewdney's construction, this shows that the number of neurons required to detect the activation of a \emph{single set} $\qPreds{k; \sym}$ grows asymptotically as the square root of the number of ones in the representation matrix $\qMtx{\qPreds{k; \sym}}$---this is how many line matrices the matrix will decompose into.
The size of each neuron is $2 \nsymbols \srNstates$.

This allows us to show how many neurons the entire \hernnAcr{} simulating $\wfsa$ has.
Since we know that the data sub-vector will always have exactly $2 \nsymbols \srNstates$ cells, we characterize the number of processing cells in the following lemma.
\begin{lemma}{}{}
    Let $\wfsa = \fsatuple$ be a deterministic FSA.
    Then, Dewdney's construction results in a \hernnAcr{} with a hidden state of size $\bigO{\nsymbols \nstates^{\frac{3}{4}}}$.
\end{lemma}
\begin{proof}
    The number of cells in the entire processing sub-vector is simply the sum of the processing neurons of all the data components.
    In the worst case, a single component-activating matrix $\mB$ requires $2 \nLines{\mB} + 1$ neurons ($2$ for each transversal in the decomposition of $\mB$ and an additional one for their disjunction).
    Therefore, enumerating the set of matrices $\set{\mB_{\idxj, \sym, \idxk} \mid \idxj \in \Zmod{2}, \sym \in \alphabet, \idxk \in \Zmod{\srNstates}}$ with $\mB_\idxn$ for $\idxn = 1, \ldots, 2 \nsymbols \srNstates$, the number of neurons required by all component-activating matrices is bounded as follows.
    \begin{equation}
        \sum_{\idxn = 1}^{2 \nsymbols \srNstates} 2 \nLines{\mB_\idxn} + 1 \leq
        \sum_{\idxn = 1}^{2 \nsymbols \srNstates} 2 \left(2 \ceil{\sqrt{\norm{\mB_\idxn}_1}}\right) + 1 \defeq
        \sum_{\idxn = 1}^{2 \nsymbols \srNstates} 4 m_\idxn + 1  \label{eq:n-components}
    \end{equation}
    Since the matrices $\mB_\idxn$ contain one non-zero entry for each state-symbol pair, it holds that
    \begin{equation} \label{eq:n-components-constraint}
        \sum_{\idxn = 1}^{2 \nsymbols \srNstates} \norm{\mB_\idxn}_1 \leq \sum_{\idxn = 1}^{2 \nsymbols \srNstates} m^2_\idxn = \nsymbols \nstates
    \end{equation}
    Pretending that $m_\idxn$ can take real values, the value of \cref{eq:n-components} is maximized under the constraint from \cref{eq:n-components-constraint} when all $m_\idxn$ are equal with $m_\idxn = \sqrt{2 \srNstates}$.
    This means that
    \begin{equation}
        \sum_{\idxn = 1}^{2 \nsymbols \srNstates} 4 m_\idxn + 1 \leq \sum_{\idxn = 1}^{2 \nsymbols \srNstates} 4 \sqrt{2 \srNstates} + 1
        = 8 \nsymbols \srNstates \sqrt{2 \srNstates} + 1
        = \bigO{\nsymbols \nstates^{\frac{3}{4}}},
    \end{equation}
    finishing the proof.
\end{proof}

All results stated in this section can be summarized in the following theorem.
\begin{theorem}{\citet{Dewdney1977}}{}
    Let $\wfsa = \fsatuple$ be a deterministic FSA.
    Then, there exists a \hernnAcr{} of size $\bigO{\nsymbols \nstates^{\frac{3}{4}}}$ correctly simulating $\wfsa$.
\end{theorem}

\subsubsection{Indyk's Construction} \label{sec:indyk}
\cref{sec:dewdney} describes a construction of an \hernnAcr{} of size $\bigO{\nsymbols \nstates^\frac{3}{4}}$ simulating an FSA.
While this improves the space efficiency compared to Minsky's construction, it is not asymptotically optimal.
\citet{Indyk1995} proved that a \hernnAcr{} simulating an FSA $\wfsa = \fsatuple$ over a binary alphabet $\alphabet = \B$ requires at least $\Omega\left(\sqrt{\nstates}\right)$ hidden dimensions.
He also provided a construction that achieves this lower bound.
This construction is conceptually very similar to Dewdney's in that it works by activating neurons corresponding to some form of compressed predecessor matrices (component-activating matrices) and then selecting the transition which matches the input symbol.
Again, it additively covers these matrices with components that are easy to detect, similar to how Dewdney's construction uses line matrices.
However, Indyk's construction defines component-activating matrices based on different sets of states and covers them with a different decomposition---these are the two crucial differences allowing the construction to achieve the optimal lower bound.

\begin{figure*}
    \centering
    \includegraphics[width=\textwidth]{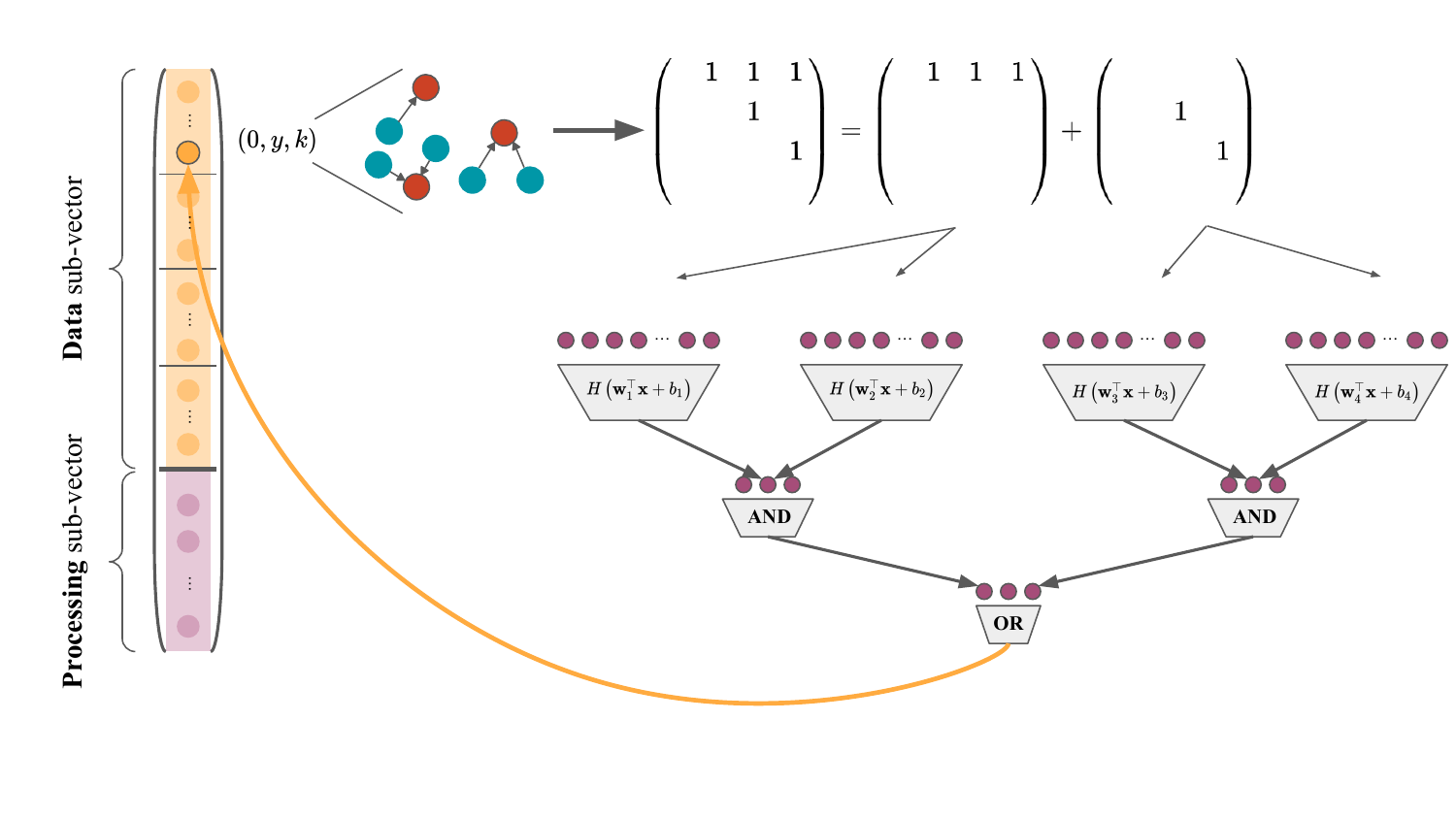}
    \caption{High-level overview of Indyk's construction.
        The highlighted \textcolor{ETHBronze}{orange} neuron in the representation of the state from the data sub-vector corresponds to the activation of one of the components of the \textcolor{ETHRed}{red} states (which have in common that their $0\textsuperscript{th}$ component of $\phifourFun{\stateq}$ is the same).
        The matrix corresponding to the disjunction of the representations of their \emph{$\sym$-predecessors} (\textcolor{ETHBlue}{blue} states) is decomposed into two non-decreasing matrices.
        The non-zero elements of both can be detected by a conjunction of two neurons; here, $\func_1 = \begin{pmatrix}
                0         & 1 & 2 & 3 \\
                \emptyset & 0 & 0 & 0
            \end{pmatrix}$ and $\func_2 = \begin{pmatrix}
                0         & 1         & 2 & 3 \\
                \emptyset & \emptyset & 1 & 2
            \end{pmatrix}$, meaning that $\vw_1 = \begin{pmatrix}
                3 & 0 & 0 & 0 & \mid & 0 & 1 & 1 & 1
            \end{pmatrix}$, $\vw_2 = \begin{pmatrix}
                0 & 3 & 2 & 0 & \mid & 0 & 0 & 1 & 2
            \end{pmatrix}$, and $b_1 = b_2 = 4$.
        Those activations are then disjoined to result in the activation in the \textcolor{ETHBronze}{orange} neuron.
        The \textcolor{ETHPurple}{purple} neurons in the processing sub-vector are composed of the neurons in the networks implementing the detection of line matrices and their conjunctions and disjunctions (also shown in \textcolor{ETHPurple}{purple}).
        Note that even if the second matrix were not non-decreasing in itself (i.e., the columns of the two ones would be flipped), one could still transform it into a non-decreasing matrix by permuting the columns and permuting the corresponding neurons.}
    \label{fig:indyk-high-level}
\end{figure*}

We first define the component-activating matrices and their role in updating the hidden state of the \hernnAcr{}.
In Indyk's construction, the component-activating matrices are based on \emph{four-hot} rather than two-hot encodings of states.
\begin{definition}{Four-hot representation of a state}{}
    Let $\wfsa = \fsatuple$ be an FSA, $\frNstates \defeq \ceil{\nstates^\frac{1}{4}}$, and $\perm$ a permutation of $\states = \NTo{\nstates}$.\footnote{The exact form of $\perm$ will be important later. For now, one can think of $\perm$ as the identity function.}
    We define the \defn{four-hot representation} of $\stateq \in \states$ as
    \begin{equation}
        \phifour(\stateq) = \left(\ell_1, \ell_2, \ell_3, \ell_4\right)
    \end{equation}
    where
    \begin{equation}
        \ell_\idxj = \frac{\perm\left(\stateq\right)}{\frNstates^{\idxj - 1}} \mod \frNstates.
    \end{equation}
    We denote the inverse of $\phifour$ with $\phifourInv$ and further define for $k \in \Zmod{\frNstates}$
    \begin{equation}
        \phifourInvFun{k, \cdot, \cdot, \cdot} \defeq \set{\stateq \in \states \mid \phifourFun{\stateq}_1 = k}
    \end{equation}
    and $\phifourInvFun{\cdot, k, \cdot, \cdot}$, $\phifourInvFun{\cdot, \cdot, k, \cdot}$, and $\phifourInvFun{\cdot, \cdot, \cdot, k}$ analogously.
\end{definition}
We will denote $\phifourInvFun{\ldots, \idxk, \ldots}$ with $\idxk$ in $\idxj\textsuperscript{th}$  position (with $\idxj \in \Zmod{4}$) as $\compMtx{\idxk}{\idxj}$.
Despite using the four-hot representations, Indyk's construction still requires the two-hot representations based on $\phitwo$ as before.
In this case, however, they again depend on the chosen permutation $\perm$.
This allows us to define the component-activating matrices as follows.
\begin{definition}{Component-activating matrix}{}
    A \defn{component-activating matrix} in Indyk's construction is the representation matrix $\qMtx{\qPreds{\compMtx{\idxk}{\idxj}; \sym}}$ for some $\idxk \in \Zmod{\frNstates}$, $\idxj \in \Zmod{4}$, and $\sym \in \alphabet$.
\end{definition}
For efficient detection, the component-activating matrices are covered by so-called non-decreasing matrices.
\begin{definition}{Non-decreasing matrix}{}
    We say that $\mB \in \BDD$ is \defn{non-decreasing} if there exists a \emph{non-decreasing} (partial) function $\func\colon \Zmod{D} \to \Zmod{D}$ (from columns to rows) such that
    \begin{equation}
        \mB_{\idxi \idxj} = 1 \iff \func\left(\idxj\right) = \idxi
    \end{equation}
    \textbf{and}, if $\func$ is defined for some $\idxj \in \Zmod{D}$, it is also defined for all $\idxj' \geq \idxj$.
\end{definition}
\begin{example}{Non-decreasing matrices}{}
    An example of a non-decreasing matrix is
    \begin{equation}
        \mB = \begin{pmatrix}
            0 & 1 & 0 & 0 \\
            0 & 0 & 0 & 0 \\
            0 & 0 & 1 & 1 \\
            0 & 0 & 0 & 0
        \end{pmatrix}.
    \end{equation}
    The function $\func$ defining the non-decreasing matrix $\mB$ is $\func = \begin{pmatrix}
            0         & 1 & 2 & 3 \\
            \emptyset & 0 & 1 & 1
        \end{pmatrix}$, where $\emptyset$ denotes that the function is not defined.
\end{example}
Again, clearly, any matrix $\mB \in \BDD$ can be (non-uniquely) decomposed into at most $D$ non-decreasing matrices.
Moreover, non-decreasing matrices are detectable.
\begin{lemma}{}{}
    Non-decreasing matrices are detectable.
\end{lemma}
\begin{proof}
    Let $\mB \in \BDD$ be a non-decreasing matrix defined by the partial function $\func$.
    Divide the domain of $\func$ into the set of intervals in which the function is constant, with $I(\idxj)$ denoting the interval of $\idxj \in \Zmod{\frNstates^2}$ for $\idxj$ such that $\func\left(\idxj\right)$ is defined.
    Then, it is easy to see that $\emB_{\idxi \idxj} = 1 \iff \idxi = \func\left(\idxj\right)$, meaning that by defining the parameters $\vw$ and $b$ as
    \begin{align}
        \evw_{\func\left(\idxj\right)} & \defeq \frNstates^2 - I\left(\idxj\right) \\
        \evw_{\frNstates^2 + \idxj}    & \defeq I(\idxj)                           \\
        b                              & \defeq -\frNstates^2
    \end{align}
    and other elements as $0$, we get that
    \begin{equation} \label{eq:indyk-equality}
        \emB_{\idxi \idxj} = 1 \iff \idxi = \func\left(\idxj\right) \iff \evw_\idxi + \evw_\idxj + b = 0.
    \end{equation}

    Compared to earlier, where component-activating matrices were detected by testing an \emph{inequality}, detecting a non-decreasing matrix requires testing an \emph{equality}.
    Since all terms in the equality are integers, testing the equality can be performed with the Heaviside activation function by conjoining two neurons; one testing the inequality $\evw_\idxi + \evw_\idxj + b - 1 < 0$ and another one testing the inequality $\evw_\idxi + \evw_\idxj + b + 1 > 0$.
    Both can individually be performed by a single neuron and then conjoined by an additional one.
\end{proof}
With this, the high-level idea of Indyk's construction is outlined in \cref{fig:indyk-high-level}.
After constructing the component-activating matrices based on $\phifour$ and decomposing them into non-decreasing matrices, the rest of Indyk's construction is very similar to Dewdney's construction, although the full update step of the \hernnAcr{} requires some additional processing.
To test the equality needed to detect non-decreasing matrices in the decomposition, \cref{eq:indyk-equality}, the four-hot representations are first converted into two-hot ones.
This can be done by a simple conjunction of the first two and the last two components of the four-hot representation.
Then, the activations of the non-decreasing matrices can be computed and disjoined into the representations of the component-activating matrices.
These form the $4 \nsymbols \frNstates$ components of the data sub-vector of the \hernnAcr{} hidden state.
They contain the activations of all possible next states, i.e., the out-neighborhood of the current state of $\wfsa$.
These are then conjoined with the representation of the current input symbol in the same way as in Dewdney's construction but adapted to the four-hot representations of the states.
The process is thus very similar to the phases of Dewdeney's construction illustrated in \cref{fig:dewdney-phases}.

Indyk's construction can be summarized by the following lemma.\footnote{Again, to formally prove it is correct, we would have to follow a similar set of steps to how the correctness of Minsky's construction (\cref{lem:minsky-transitions}) was proved. We omit this for conciseness.}
\begin{lemma}{}{}
    Let $\wfsa = \fsatuple$ be a deterministic FSA.
    Then, Indyk's construction results in a \hernnAcr{} correctly simulating $\wfsa$'s transition function, i.e, $\hToQFun{\hiddStatet} = \qt$ for all $\tstep$.
\end{lemma}

The only remaining thing to show is that Indyk's construction achieves the theoretically optimal lower bound on the size of the \hernnAcr{} simulating a deterministic FSA.
All previous steps of the construction were valid no matter the chosen permutation $\perm$.
The permutation, however, matters for space efficiency: intuitively, it determines how efficiently one can decompose the resulting component-activating matrices (which depend on the permutation) into non-decreasing matrices in the sense of how many non-decreasing matrices are required to cover it.
Indyk, therefore, proved that there always exists, with non-zero probability, a permutation in which the decomposition across all states is efficient enough to achieve the minimum number of neurons required.
This is formalized by the following lemma, whose proof can be found in \citet[][Lemma 6]{Indyk1995}.
\begin{lemma}{}{}
    Let $\wfsa = \fsatuple$ be a deterministic FSA.
    There exists a permutation of $\states$ such that Indyk's construction results in a \hernnAcr{} of size $\bigO{\nsymbols \sqrt{\nstates}}$.
\end{lemma}
This concludes our presentation of Indyk's construction.
All results stated in this section can be summarized by the following theorem.
\begin{theorem}{\citet{Indyk1995}}{}
    Let $\wfsa = \fsatuple$ be a deterministic FSA.
    There exists a \hernnAcr{} of size $\bigO{\nsymbols \sqrt{\nstates}}$ correctly simulating $\wfsa$.
\end{theorem}

\subsection{Lower Bound in the Probabilistic Setting} \label{sec:space-bounds}
We now ask whether the same lower bound can also be achieved when simulating \dpfsaAcr{}s.
We find that the answer is negative: \dpfsaAcr{}s may require an \hernnAcr{} LMs of size $\Omega\left(\nsymbols\nstates\right)$ to faithfully represent their probability distribution.
Since the transition function of the underlying FSA can be simulated with more efficient constructions, the bottleneck comes from defining the same probability distribution.
Indeed, as the proof of the following theorem shows, the issue intuitively arises in the fact that, unlike in an \hernnAcr{} LM, the local probability distributions of the different states in a PFSA are completely arbitrary, whereas they are defined by shared parameters (the output matrix $\outMtx$) in an \hernnAcr{} LM.
\begin{theorem}{A lower bound on the size of the RNN simulating a PFSA}{pfsa-rnn-lower-bound}
    Let $\wfsa = \wfsatuple$ be a minimal \dpfsaAcr{} and $\elmanrnntuple$ an \hernnAcr{} LM defining the same LM.
    Then, $\hiddDim$ must scale linearly with $\nstates$.
\end{theorem}
\begin{proof}
    Without loss of generality, we work with $\Rex$-valued hidden states.
    Let $\wfsa$ be a minimal deterministic PFSA and $\rnn = \elmanrnntuple$ a \hernnAcr{} with $\pLN\left(\str\right) = \wfsa\left(\str\right)$ for every $\str \in \kleene{\alphabet}$.
    Let $\str_{<\strlen} \in \kleene{\alphabet}$ and $\str_{\leq \strlen} \defeq \str_{<\strlen} \sym$ for some $\sym \in \alphabet$.
    Define $\overline{\pdens}\left(\str\right) \defeq \prod_{\tstep = 1}^{|\str|} \pLNSMFun{\symt}{\strlt}$.
    It is easy to see that $\overline{\pdens}\left(\str_{<\strlen} \sym_\strlen\right) = \overline{\pdens}\left(\str_{<\strlen}\right)\pLNSMFun{\symt}{\str_{<\strlen}}$.
    The conditional distribution $\pLNSM\left(\cdot \mid \str_{<\strlen}\right)$ are proportional to the values in $\outMtx \hiddState_{\strlen - 1}$.
    By definition of the deterministic PFSA, there are $\nstates$ such conditional distributions.
    Moreover, these distributions (represented by vectors $\in \SimplexEosalphabetminus$) can generally be \emph{linearly independent}.
    This means that for any $\stateq$, the probability distribution of the outgoing transitions can not be expressed as a linear combination of the probability distributions of other states.
    To express the probability vectors for all states, the columns of the output matrix $\outMtx$, therefore, have to span $\Rex^{\nstates}$, implying that $\outMtx$ must have at least $\nstates$ columns.
    This means that the total space complexity (and thus the size of the \hernnAcr{} representing the same distribution as $\wfsa$) is $\Omega\left(\nstates\right)$.
\end{proof}

\paragraph{Asymptotic Bounds in $\nsymbols$}
Since each of the input symbols can be encoded in $\log \nsymbols$ bits, one could expect that the linear factor in the size of the alphabet from the constructions above could be reduced to $\bigO{\log\nsymbols}$.
However, we again find that such reduction is in general not possible---the set of FSAs presented next is an example of a family that requires an \hernnAcr{} whose size scales linearly with $\nsymbols$ to be simulated correctly.
We also provide a sketch of the proof of why a compression in $\nsymbols$ is not possible.

Let $\wfsa_N = \left(\alphabet_N, \set{0, 1}, \set{0}, \set{1}, \trans_N\right)$ be an FSA over the alphabet $\alphabet_N = \set{\sym_1, \ldots, \sym_N}$ such that $\trans_N = \set{\uwedge{0}{\sym_1}{1}} \cup \set{\uwedge{0}{\sym_\idx}{2}\mid \idx = 2, \ldots N}$ (see \cref{fig:log-sigma-counterexample}).
\begin{figure}[t]
    \centering
    \begin{tikzpicture}[node distance = 16mm]
        \node[state, initial] (q0) [] { $0$ };
        \node[state] (q1) [right = of q0, yshift=10mm] { $1$ };
        \node[state, accepting] (q2) [right = of q0, yshift=-10mm] { $2$ };
        \draw[transition]  (q0) edge[below, sloped, bend right] node{ $\sym_2,\ldots,\sym_N$ } (q2)
        (q0) edge[below, sloped, bend left] node{ $\sym_1$ } (q1);
    \end{tikzpicture}
    \caption{The FSA $\wfsa_N$.}
    \label{fig:log-sigma-counterexample}
\end{figure}

Clearly, to be able to correctly represent all local distributions of the \dpfsaAcr{}, the \hernnAcr{} LM must contain a representation of each possible state of the \dpfsaAcr{} in a unique hidden state.
On the other hand, the only way that the \hernnAcr{} can take into account the information about the current state $\qt$ of the simulated FSA $\wfsa$ is through the hidden state $\hiddStatet$.
The hidden state, in turn, only interacts with the recurrence matrix $\recMtx$, which does not have access to the current input symbol $\symtplus$.
The only interaction between the current state and the input symbol is thus through the addition in $\recMtx \hiddStatet + \inMtx \onehot{\symtplus}$.
This means that, no matter how the information about $\qt$ is encoded in $\hiddStatet$, in order to be able to take into account all possible transitions stemming in $\qt$ (before taking into account $\symtplus$), $\recMtx \hiddStatet$ must activate \emph{all} possible next states, i.e., the entire out-neighborhood of $\qt$.
On the other hand, since $\inMtx \onehot{\symtplus}$ does not have precise information about $\qt$, it must activate all states which can be entered with an $\symtplus$-transition, just like in Minsky's construction.

In Minsky's construction, the recognition of the correct next state was done by keeping a separate entry (one-dimensional sub-vector) for each possible pair $\qtplus, \symtplus$.
However, when working with compressed representations of states (e.g., in logarithmic space), a single common sub-vector of size $<\nsymbols$ (e.g., $\log \nsymbols$) has to be used for all possible symbols $\sym \in \alphabet$.
Nonetheless, the interaction between $\recMtx \hiddStatet$ and $\inMtx \onehot{\symtplus}$ must then ensure that only the correct state $\qtplus$ is activated.
For example, in Minsky's construction, this was done by simply taking the conjunction between the entries corresponding to $\stateq, \sym$ in $\recMtx \hiddStatet$ and the entries corresponding to $\stateq', \sym'$ in $\inMtx \onehot{\sym'}$, which were all represented in individual entries of the vectors.
On the other hand, in the case of the $\log$ encoding, this could intuitively be done by trying to match the $\log \nsymbols$ ones in the representation $\left(\vp\left(\sym\right) \mid \one - \vp\left(\sym\right)\right)$, where $\vp\left(\sym\right)$ represent the binary encoding of $\sym$.
If the $\log \nsymbols$ ones match (which is checked simply as it would result in a large enough sum in the corresponding entry of the matrix-vector product), the correct transition could be chosen (to perform the conjunction from \cref{fact:and} correctly, the bias would simply be set to $\log\nsymbols - 1$).
However, an issue arises as soon as \emph{multiple} dense representations of symbols in $\inMtx \onehot{\sym}$ have to be activated against the same sub-vector in $\recMtx \hiddStatet$---the only way this can be achieved is if the sub-vector in $\recMtx \hiddStatet$ contains the disjunction of the representations of all the symbols which should be activated with it.
If this sets too many entries in $\recMtx \hiddStatet$ to one, this can result in ``false positives''.
This is explained in more detail for the \dpfsaAcr{}s in \cref{fig:log-sigma-counterexample} next.

Let $\vr_\idx$ represent any dense encoding of $\sym_\idx$ in the alphabet of $\wfsa_N$ (e.g., in the logarithmic case, that would be $\left(\vp\left(\idx\right) \mid \one - \vp\left(\idx\right)\right)$).
Going from the intuition outlined above, any \hernnAcr{} simulating $\wfsa_N$, the vector $\recMtx \hiddStateZero$ must, among other things, contain a sub-vector corresponding to the states $1$ and $2$.
The sub-vector corresponding to the state $2$ must activate (through the interaction in the Heaviside function) against any $\sym_\idx$ for $\idx = 2, \ldots, N$ in $\wfsa_N$.
This means it has to match all representations $\vr_\idx$ for all $\idx = 2, \ldots, N$.
The only way this can be done is if the pattern for recognizing state $2$ being entered with any $\sym_\idx$ for $\idx = 2, \ldots, N$ is of the form $\vr = \bigvee_{\idx = 2}^N \vr_\idx$.
However, for sufficiently large $N$, $\vr = \bigvee_{\idx = 2}^N \vr_\idx$ will be a vector of all ones---including all entries active in $\vr_1$.
This means that \emph{any} encoding of a symbol will be activated against it---among others, $\sym_1$.
Upon reading $\sym_1$ in state $1$, the network will therefore not be able to deterministically activate only the sub-vector corresponding to the correct state $1$.
This means that the linear-size encoding of the symbols is, in general, optimal for representing \dpfsaAcr{}s with \hernnAcr{} LMs.
This discussion implies the following theorem.
\begin{theorem}{A lower bound on the size of the RNN simulating a PFSA}{pfsa-rnn-lower-bound-alphabet}
    Let $\wfsa = \wfsatuple$ be a minimal \dpfsaAcr{} and $\elmanrnntuple$ an \hernnAcr{} LM defining the same LM.
    Then, $\hiddDim$ must scale linearly with $\nsymbols$.
\end{theorem}

Based on the challenges encountered in the example above, we can devise a simple sufficient condition for a logarithmic compression w.r.t. $\nsymbols$ to be possible: namely, that for any pair of states $\stateq, \stateq' \in \states$, there is at most a single transition leading from $\stateq$ to $\stateq'$.
This intuitive characterization can be formalized by a property we call $\log\nsymbols$-separability.
\begin{definition}{$\log\nsymbols$-separable finite-state automaton}{}
    An FSA $\wfsa = \fsatuple$ is \defn{$\log\nsymbols$-separable} if it is deterministic and, for any pair $\stateq, \stateq' \in \states$, there is at most one symbol $\sym \in \alphabet$ such that $\uwedge{\stateq}{\sym}{\stateq'} \in \trans$.
\end{definition}

$\log\nsymbols$-separability is a relatively restrictive condition.
To amend that, we introduce a simple procedure which, at the expense of enlarging the state space by a factor of $\alphabet$, transforms a general deterministic (unweighted) FSA into a $\log\nsymbols$-separable one.
We call this \defn{$\log\nsymbols$-separation}.
Intuitively, it augments the state space by introducing a new state $\left(\stateq, \sym\right)$ for every outgoing transition $\uwedge{\stateq}{\sym}{\stateq'}$ of every state $\stateq \in \states$, such that $\left(\stateq, \sym\right)$ simulates the only state the original state $\stateq$ would transition to upon reading $\sym$.
Due to the determinism of the original FSA, this results in a $\log\nsymbols$-separable FSA with at most $\nstates \nsymbols$ states.

While the increase of the state space might seem like a step backward, recall that using Indyk's construction, we can construct an \hernnAcr{} simulating an FSA whose size scales with the square root of the number of states.
And, since the resulting FSA is $\log\nsymbols$-separable, we can reduce the space complexity with respect to $\alphabet$ to $\log \nsymbols$.
This is summarized in the following theorem, which characterizes how compactly general deterministic FSAs can be encoded by \hernnAcr{}s.
To our knowledge, this is the tightest bound on simulating general unweighted deterministic FSAs with \hernnAcr{}s.
\begin{theorem}{Efficiently simulating general FSAs}{}
    Let $\wfsa = \fsatuple$ be a minimal FSA recognizing the language $\lang$.
    Then, there exists an \hernnAcr{} $\rnn = \elmanrnntuple$ accepting $\lang$ with $\hiddDim = \bigO{\log\nsymbols \sqrt{\nsymbols \nstates}}$.
\end{theorem}

The full $\log\nsymbols$-separation procedure is presented in \cref{algo:separation}.
It follows the intuition of creating a separate ``target'' for each transition $\uwedge{\stateq}{\sym}{\stateq'}$ for every state $\stateq \in \states$.
To keep the resulting FSA deterministic, a new, artificial, initial state with no incoming transitions is added and is connected with the augmented with the out-neighborhood of the original initial state.

\begin{algorithm}
    \begin{algorithmic}[1]
        \Func{\separationAlgoName{}($\wfsa=\fsatuple$)}
        \State $\wfsa' \gets \left(\alphabet, \states' = \states \times \alphabet \cup \set{\qinit'}, \trans'=\varnothing, \initial'=\set{\qinit'}, \final'=\varnothing\right)$
        \LineComment{Connect the out-neighborhood of the original initial state $\qinit$ with the new, aritificial, initial state.}
        \For{$\sym \in \alphabet$}
        \For{$\uwedge{\qinit}{\sym'}{\stateq'} \in \trans$}
        \State \textbf{add} $\uwedge{\qinit'}{\sym}{\left(\stateq', \sym'\right)}$ \textbf{to} $\trans'$
        \EndFor
        \EndFor
        \For{$\stateq \in \states, \sym \in \alphabet$}
        \For{$\uwedge{\stateq}{\sym'}{\stateq'} \in \trans$}
        \State \textbf{add} $\uwedge{\left(\stateq, \sym\right)}{\sym'}{\left(\stateq', \sym'\right)}$ \textbf{to} $\trans'$
        \EndFor
        \EndFor
        \LineComment{Add all state-symbol pairs with a state from the original set of final states to the new set of final states.}
        \For{$\qfinal \in \final, \sym \in \alphabet$}
        \State \textbf{add} $\left(\qfinal, \sym\right)$ \textbf{to} $\final'$
        \EndFor
        \If{$\qinit \in \initial$} \Comment{Corner case: if the original initial state $\qinit$ is an initial state, make the artificial initial state $\qinit'$ final.}
        \State \textbf{add} $\qinit'$ \textbf{to} $\final'$
        \EndIf
        \State \Return $\wfsa'$
        \EndFunc
    \end{algorithmic}
    \caption{}
    \label{algo:separation}
\end{algorithm}

The following simple lemmata show the formal correctness of the procedure and show that it results in a $\log\nsymbols$-separable FSA, which we need for compression in the size of the alphabet.

\begin{lemma}{}{sep-lemma1}
    For any $\sym \in \alphabet$, $\uwedge{\left(\stateq, \sym\right)}{\sym'}{\left(\stateq', \sym'\right)} \in \trans'$ if and only if $\uwedge{\stateq}{\sym'}{\stateq'} \in \trans$.
\end{lemma}
\begin{proof}
    Ensured by the loop on Line 3.
\end{proof}

\begin{lemma}{}{}
    $\log\nsymbols$-separation results in an equivalent FSA.
\end{lemma}
\begin{proof}
    We have to show that, for any $\str \in \kleene{\alphabet}$, $\str$ leads to a final state in $\wfsa$ if and only if $\str$ leads to a final state in $\wfsa'$.
    For the string of length $0$, this is clear by Lines 13 and 14.
    For strings of length $\geq 1$, it follows from \cref{lem:sep-lemma1} that $\str$ leads to a state $\stateq$ in $\wfsa$ if and only if $\exists \sym \in \alphabet$ such that $\str$ leads to $\left(\stateq, \sym\right)$ in $\wfsa'$.
    From Lines 11 and 12, $\left(\stateq, \sym\right) \in \final'$ if and only if $\stateq \in \final$, finishing the proof.
\end{proof}

\begin{lemma}{}{}
    $\log\nsymbols$-separation results in a $\log\nsymbols$-separable FSA.
\end{lemma}
\begin{proof}
    Since the state $\left(\stateq', \sym'\right)$ is the only state in $\states'$ transitioned to from $\left(\stateq, \sym\right)$ after reading $\sym'$ (for any $\sym \in \alphabet$), it is easy to see that $\wfsa'$ is indeed $\log\nsymbols$-separable.
\end{proof}

\paragraph{Discussion and the practical applicability of these result.}
This section showed that Heaviside-activated RNNs are equivalent to WFSAs.
This might come as a bit of a surprise considering that we introduced RNNs with the goal of overcoming some limitations of exactly those models, e.g., the finite context length.
However, note that to arrive at this result, we considerately restricted the form of a recurrent neural network.
While on the one hand restriction to the Heaviside activation function means that all the RNNs we considered in this section can be implemented and represented in a computer, the RNN sequence models that we usually deal with are much more complex than this analysis allowed for.
Furthermore, note that the RNNs in practice do not learn sparse hidden states of the form considered in the construction in the proof of \cref{lem:minsky-constr}---indeed, networks with Heaviside activation functions are not trainable with methods discussed in \cref{sec:learning} as the gradient on the entire parameter space would be either $\zero$ or undefined and in this sense, the trained networks would never have such hidden state dynamics.
The dynamics of RNNs in practice result in \emph{dense} hidden states, i.e., states in which many dimensions are non-zero.
Nonetheless, keep in mind that theoretically, due to the finite-precision nature of our computers, all models we ever consider will be at most finite-state---the differentiating factor between them will be how appropriately to the task they are able to learn the topology (transitions) of the finite-state automaton they represent and how efficiently they are able to learn it.
Lastly, note that, by considering special classes of (sub-)regular languages, one can arrive at neural representations far smaller than those described by the constructions above \citep{hewitt-etal-2020-rnns,svete2024theoretical}.

\pagebreak
\subsection{Turing Completeness of Recurrent Neural Networks} \label{sec:rnn-turing-completeness}
We now turn to the (purely theoretical) treatment of the expressive capacity of recurrent neural networks in which we take the liberty of making somewhat unrealistic assumptions.
Specifically, in practice, RNNs usually have the following properties:
\begin{itemize}
    \item The weights and intermediate computations are done with finite floating point precision;
    \item An RNN always operates in \emph{real-time}, meaning that it performs a constant number of operations before consuming/outputting a symbol.
\end{itemize}
Under these assumptions, we saw that RNNs with Heaviside activations in a practical setting lie at the bottom of the weighted Chomsky hierarchy, being able to only recognize regular languages.
However, if we relax these two assumptions, allowing for arbitrary precision and unbounded computation time between symbols, RNNs jump directly to the top of the hierarchy: they become Turing complete.

\begin{figure}[h!]
    \centering
    \begin{tikzpicture}[scale=2]

        \draw [<->] (0,1.5) node [left] {$\sigmoid\left(x\right)$} |- (2,0) node [below] {$x$};

        \draw [-, ETHPetrol, thick, domain=-1:0] plot (\x,0);
        \draw [-, ETHPetrol, thick, domain=0:1] plot (\x,\x);
        \draw [-, ETHPetrol, thick, domain=1:2] plot (\x,1);

        \node [below left] at (0,0) {$0$};
        \node [below] at (1,0) {$1$};
        \node [left] at (0,1) {$1$};

    \end{tikzpicture}

    \caption{The saturated sigmoid.}
    \label{fig:saturated-sigmoid}
\end{figure}
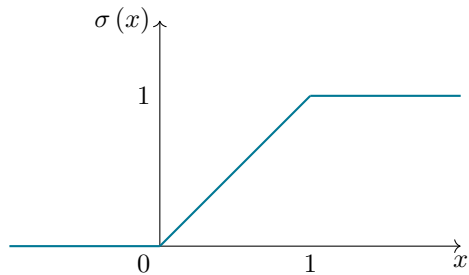

We start by introducing the saturated sigmoid, one of the building blocks we will use to show this.
\begin{definition}{Saturated Sigmoid function}{saturated-sigmoid}
    The \defn{saturated sigmoid} is defined as
    \begin{equation}
        \sigmoid\left(x\right)=
        \begin{cases}
            0 & \ifcondition x  \le 0     \\
            x & \ifcondition 0 < x \le  1 \\
            1 & \ifcondition x  > 1
        \end{cases}.
    \end{equation}
\end{definition}
Intuitively, the saturated sigmoid clips all negative values to $0$, all values larger than $1$ to $1$, and leaves the elements of $\left[0, 1\right]$ intact.
The graph of this function is shown in \cref{fig:saturated-sigmoid}.

The central result of this subsection is then summarized in the following theorem, summarized from \citet{nowak-etal-2023-representational}.
\begin{theorem}{Saturated Sigmoid Elman RNNs are Turing complete}{}
    Elman recurrent neural network sequence models with the saturated sigmoid activation functions are Turing complete.
\end{theorem}
By the end of this subsection, we will have proven this result by showing that Saturated Sigmoid Elman RNNs can encode two-stack pushdown automata which are computationally equivalent to Turing machines (cf. \cref{def:two-stack-pda,def:two-stack-wpda}).
We start with a simpler construction: building on our placement of RNNs on at least the regular rung of the ladder of formal language complexity (cf. \cref{lem:minsky-constr}), we take one step up the ladder and show that RNNs can simulate a \emph{single-stack} pushdown automaton (cf. \cref{def:pda}).
This will help us gain the intuition behind how RNNs can use infinite precision arithmetic to simulate a stack.
We will then simply generalize this construction to the two-stack case.

Let us begin by considering the problem of representing a stack---an arbitrarily long sequence of symbols, accessible in a last-in-first-out (LIFO) fashion---in a vector of constants size, e.g., the hidden state of a recurrent neural network.
For simplicity, but without the loss of generality, assume that we are working with a simple two-letter stack alphabet $\stackalphabet = \set{0, 1}$.
Any stack sequence $\stackseq$ will be a member of $\kleene{\stackalphabet}$, i.e., a string of $0$'s and $1$'s.
If we think of the stack \emph{symbols} as \emph{numbers} for a moment, there is a natural correspondence between the possible stack configurations and \emph{numbers} expressed in base $2$.
By convention, we will represent a string of stack symbols $\stackseq$ with numbers after the decimal point, rather than as integers.
Assuming infinite precision, we can therefore simply represent each stack configuration as a single number (of course, the stack alphabet does not have to be exactly $\stackalphabet = \set{0, 1}$---we can always map symbols from any alphabet into their numeric representations in some base large enough to allow for the entire alphabet).
Notice that in this case, pushing or popping from the stack can be performed by division and multiplication of the value representing the stack---if we want to push a value $x \in \set{0, 1}$, we can \emph{divide} the current representation (by $2$) and append $x$ to the right side of the new representation and if we want to pop any value, we simply have to \emph{multiply} the current representation by $2$. \franz{This is not really clear - are we working with binary numbers or decimals here?}
This also gives us an idea of how to represent a stack in the hidden state of an RNN: the \emph{entire} stack sequence will simply be represented in a single dimension of the hidden state, and the value stored in the cell will be updated according to the transitions defined by the simulated automaton.
Note, however, that the RNN will not only have a single dimension in the hidden state: other dimensions will contain values that will be required to control the RNN updates correctly.

In our proofs, we consider a special type of pushdown automata, as defined in \cref{def:pda}: we will use pushdown automata which only consider the \emph{topmost} element of the stack when defining the possible transitions from a configuration and can only push one stack symbol at a time.
More formally, this means that in the tuple $\pushdown = \pdatuple$, we have that $\trans \subseteq \states \times \stackalphabet \times \left( \alphabet \cup  \set{\eps} \right) \times \states \times \stackalphabet$ rather than the more general $\trans \subseteq \states \times \kleene{\stackalphabet} \times \left( \alphabet \cup  \set{\eps} \right) \times \states \times \kleene{\stackalphabet}$.
Furthermore, we assume that $\initstack = \eps$ and $\finalstack = \eps$, that is, the PDA starts off with an empty stack and has to empty it again to arrive at a final configuration.
Note that these restrictions can be done without loss of generality---that is, such pushdown automata are as powerful as the unrestricted versions \citep{Sipser2013}.
With this in mind, we can show that arbitrary precision RNNs are capable of recognizing at least deterministic context-free languages:
\begin{theorem}{RNNs can recognize deterministic context-free languages}{rnn-pda}
    Elman recurrent neural networks can recognize deterministic context-free languages.
\end{theorem}

Before we continue to the proof of \cref{thm:rnn-pda}, let us remark on three simple but important intuitions which will be crucial for understanding the construction of the Elman RNN, both in the single- as well as the two-stack variants of PDAs.
Multiple times in the construction, we will be faced with the task of \emph{moving} or copying the value from some dimension $\idxi$ to the dimension $\idxj$ in the vector.
The following fact shows how this can be done using simple matrix multiplication with a specific matrix.

\begin{fact}{Copying elements of a vector}{mm-copy}
    Let $\vx \in \R^\hiddDim$ and $\mM \in \R^{\hiddDim \times \hiddDim}$ such that $\mM_{\idxi, \colon} = \basisVect_\idxi$, where $\mM_{\idxi, \colon}$ denotes the $\idxi^\text{th}$ row of $\mM$ and $\basisVect_\idxi$ denotes the $\idxi^\text{th}$ basis vector.
    Then, it holds that $\left(\mM \vx\right)_{\idxi} = \evx_\idxi$.
\end{fact}
Also, note that setting the row $\mM_{\idxi, \colon}$ to the zero vector $\zero_{\in\R^\hiddDim}$ sets the entry $\evx_\idxi$ to $0$, i.e., it erases the entry.

Furthermore, we will use the saturated sigmoid function multiple times to \emph{detect} whether a number of dimensions of a vector are set to one at the same time.
Given the recurrent dynamics of the Elman RNN (cf. \cref{eq:elman-update-rule}), we can perform this check as follows.
\begin{fact}{Detecting the activation of multiple values in the hidden state}{multi-hot-to-one-hot}
    Let $\sigmoid$ be the saturated sigmoid from \cref{def:saturated-sigmoid}, $\idxm \in \set{1, \ldots, \hiddDim}$, $\idxi_1, \ldots, \idxi_\idxm, \idxj \in \set{1, \ldots, \hiddDim}$, $\vx \in \R^\hiddDim$, $\bias \in \R^\hiddDim$, and $\mM \in \R^{\hiddDim \times \hiddDim}$ such that
    \begin{equation*}
        \mM_{\idxj, \idxi} =
        \begin{cases}
            1 & \textbf{if } \idxi \in \set{\idxi_1, \ldots, \idxi_\idxm} \\
            0 & \otherwisecondition
        \end{cases}
    \end{equation*}
    and $\evb_\idxj = -(\idxm - 1)$.
    Then, it holds that $\left(\sigmoid\left(\mM \vx + \bias \right)\right)_{\idxj} = 1$ if and only if $\evx_{\idxi_\idxk} = 1$ for all $\idxk = 1, \ldots, \idxm$.\footnote{Note that this is simply a restatement of \cref{fact:and}, which we include here for clarity and to make the connection with the construction that follows clearer.}
\end{fact}

Lastly, we will sometimes have to \emph{turn off} certain dimensions of the hidden state if any of the other dimensions are active.
Using the dynamics of Elman RNNs and the saturated sigmoid, this can be done as follows.
\begin{fact}{Turning off dimensions in the hidden state}{}
    Let $\sigmoid$ be the saturated sigmoid from \cref{def:saturated-sigmoid}, $\idxm \in \set{1, \ldots, \hiddDim}$, $\idxi_1, \ldots, \idxi_\idxm, \idxj \in \set{1, \ldots, \hiddDim}$, $\vx \in \R^\hiddDim$, $\bias \in \R^\hiddDim$, and $\mM \in \R^{\hiddDim \times \hiddDim}$ such that
    \begin{equation*}
        \mM_{\idxj, \idxi} =
        \begin{cases}
            -1 & \textbf{if } \idxi \in \set{\idxi_1, \ldots, \idxi_\idxm} \\
            0  & \otherwisecondition
        \end{cases}
    \end{equation*}
    and $\evb_\idxj = 1$.
    Then, it holds that $\left(\sigmoid\left(\mM \vx + \bias \right)\right)_{\idxj} = 0$ if and only if $\evx_{\idxi_\idxk} = 1$ for some $\idxk = 1, \ldots, \idxm$.
\end{fact}

With these intuitions in mind, we now prove \cref{thm:rnn-pda}.
Due to the relatively elaborate construction, we limit ourselves to pushdown automata with a two-symbol input alphabet $\alphabet = \set{\syma, \symb}$ as well as a two-symbol stack alphabet $\stackalphabet = \set{0, 1}$.
Note, however, that this restriction can be done without the loss of generality, meaning that this is enough to prove the Turing completeness of RNNs in general.\footnote{To simulate an arbitrary Turing machine with a machine with the binary alphabet, we simply have to encode each of the finitely-many symbols of the simulated machine using binary encoding.}
\begin{proof}
    We show this by constructing, for a given deterministic pushdown automaton $\pushdown$ recognizing a deterministic context-free language, an Elman RNN simulating the steps performed by $\pushdown$.

    Let $\pushdown = \pdatuple$ be such a deterministic pushdown automaton.
    We now define the parameters of the RNN $\rnn = \elmanrnntuple$ such that the updates to $\rnn$'s hidden state will correspond to the configuration changes in $\pushdown$.

    The construction is more involved than the one in Minsky's theorem (cf. \cref{lem:minsky-constr}).
    We, therefore, first intuitively describe the semantics of the different \emph{components} of the hidden state of the RNN.
    Then, we describe the submatrices of the parameters $\recMtx$, $\inMtx$, and $\bias$ that \emph{control} these components of the vector.
    The hidden state $\hiddState$ of the RNN will altogether have \emph{five} components.
    \begin{itemize}
        \item \textbf{Component 1: Data component}: This component, consisting of three cells, will contain the actual numerical representation of the stack, $\stackCell{}$, as well as two additional ``buffer'' cells, $\bufferCell{1}$ and $\bufferCell{2}$, which will be used for intermediate copies of the stack values during the computation of the new state.
        \item \textbf{Component 2: Top of stack component}: This component contains three cells, each corresponding to a flag denoting that \textit{(a)} the stack is empty ($\stackTopCellEmpty$), \textit{(b)} the top element of the stack is a $0$ ($\stackTopCellZero$), or \textit{(c)} the top element of the stack is a $1$ ($\stackTopCellOne$).
        \item \textbf{Component 3: Configuration component}: This component encodes the current configuration of the stack (Component 2) together with the current input symbol.
              Note that, while we assume that the input PDA works with the two-symbol alphabet $\alphabet = \set{\syma, \symb}$, the sequence model defined by the RNN requires an $\eos$ symbol to be able to terminate generation (cf. \cref{eq:rv}): $\rnn$, therefore, defines the conditional probabilities over the set $\eosalphabet = \set{\syma, \symb, \eos}$.
              With this, there are nine possible configurations $\left(\sym, \stacksymbol{\gamma}\right)$ for $\stacksymbol{\gamma} \in \set{\eps, 0, 1}$ and $\sym \in \set{\syma, \symb, \eos}$, meaning that there are nine cells in this configuration, $\stackInputConfig{\stacksymbol{\gamma}}{\sym}$, each corresponding to one of these configurations.
        \item \textbf{Component 4: Computation component}: This component contains four cells in which the computation of the next value of the stack is computed.
              There are five cells $\stackComp{\texttt{action}}{\stacksymbol{\gamma}}$ because \emph{all} possible actions ($\pushOp$ 0, $\pushOp$ 1, $\popOp$ 0, $\popOp$ 1, and $\noOp$) are performed simultaneously, and only the correct one is copied into the data component (Component 1) in the end.
        \item \textbf{Component 5: Acceptance component}: This component contains a single cell, $\acceptCell$, signaling whether the RNN accepts the string $\str$ after reading in the input $\str\ \eos$.
    \end{itemize}
    Altogether, the hidden state of $\rnn$ contains $3 + 3 + 9 + 5 + 1 = 21$ dimensions.
    The \emph{initial hidden state} $\hiddStateZero$ is a vector with a single non-zero component, whose value is $1$: the cell $\stackTopCellEmpty$ since we assume that the stack of the simulated automaton is empty at the beginning of the execution.
    We now intuitively describe the dynamics that these components define.

    \paragraph{The full update step of the network.}
    The RNN will compute the next hidden state corresponding to the new stack configuration by applying the Elman update rule (cf. \cref{eq:elman-update-rule}) four times to complete four discrete sub-steps of the computation.
    We first define
    \begin{equation}
        \hiddState^{\left(0\right)}_{\tstep + 1} \defeq \hiddStatet
    \end{equation}
    and
    \begin{equation} \label{eq:tupda-atomic-update}
        \hiddState^{\left(\idx\right)}_{\tstep + 1} \defeq \sigmoid\left(\recMtx \hiddState^{\left(\idx - 1\right)}_{\tstep + 1} + \inMtx \embedSymt + \biasVech \right)
    \end{equation}
    for $\idx = 1, 2, 3, 4$.
    Then
    \begin{equation} \label{eq:tupda-final-phase}
        \hiddState_{\tstep + 1} \defeq \hiddState^{\left(4\right)}_{\tstep + 1}.
    \end{equation}
    Intuitively, each of the four stages of computation of the actual next hidden state ``detects'' some parts of the pattern contributing to the transition in the pushdown automaton.
    We describe those patterns next intuitively before talking about the submatrices (or subvectors) of the RNN parameters corresponding to the specific parts that update the individual components of the hidden state.

    \paragraph{Data component.}
    The cells of the data component form a queue of three components: the $\stackCell{}$ cell forms the head of the queue, followed by $\bufferCell{1}$ and $\bufferCell{2}$.
    The values in the cells are updated at each execution of \cref{eq:tupda-atomic-update} by moving the currently stored values into the next cell in the queue.
    By doing so, the entry in $\bufferCell{2}$ gets discarded.
    The value of $\stackCell{}$ is copied from the cells of the \emph{computation} component by \emph{summing} them.
    We will see later that at any point of the computation (when it matters), only one of the computation components will be non-zero, which means that the summation simply corresponds to copying the non-zero computation component.
    All these operations can be performed by matrix multiplication outlined in \cref{fact:mm-copy}.

    \paragraph{Encoding the stack sequence.}
    While we outlined a possible encoding of a stack sequence above, the encoding we use in this construction is a bit different.
    Remember that for a stack sequence $\stackseq \in \kleene{\stackalphabet}$ of length $N$, the \emph{right-most} symbol $\stacksymbol{\gamma}_N$ denotes the top of the stack.
    We encode the stack sequence $\stackseq \in \kleene{\stackalphabet}$ as follows:
    \begin{equation}
        \stackToNumeric{\stacksymbol{\gamma}_1\ldots\stacksymbol{\gamma}_N} \defeq \sum_{\idxn = 1}^{N} \stackSymToDigit{\stacksymbol{\gamma}_\idxn} 10^{N - \idxn - 1}
    \end{equation}
    where $\stackSymToDigit{\stacksymbol{\gamma}} \defeq \begin{cases}
            1 & \textbf{ if } \stacksymbol{\gamma} = 0 \\
            3 & \textbf{ otherwise }
        \end{cases}.$
    \begin{example}{Scalar stack representations}{}
        For example, the stack sequence $\stackseq = 00110111$ would be represented with $\stackToNumeric{00110111} = 0.33313311$.
        Notice the ``opposite orientation'' of the two strings: the top of the stack in $\stacksymbol{\gamma}$ is the right-most symbol, while it is the left-most digit in the numerical representation.
    \end{example}
    Note that the digits $1$ and $3$ in the definition of $\stackSymToDigit{\cdot}$ are chosen somewhat arbitrarily---the encoding could also have been chosen differently.
    Similarly, a different (non-decimal) base could have been chosen.


    \paragraph{Top of stack component.}
    As mentioned, the $\stackTopCellEmpty$ cell of this component is one of the two cells set to $1$ in the initial state of the RNN.
    The individual cells of this component then get updated according to the top symbol on the stack encoded in the $\stackCell{}$ cell by taking into account how the stack is represented by $\stackCell{}$.
    Specifically, the parameters of the RNN are defined such that $\hiddState_\stackTopCellEmpty = 1$
    if previously $\hiddState_\stackCell{} = 0$, $\hiddState_\stackTopCellZero = 1$ if previously $\hiddState_\stackCell{} = 0.1\ldots$, and $\hiddState_\stackTopCellOne = 1$ if previously $\hiddState_\stackCell{} = 0.3\ldots$.

    \paragraph{Configuration component.}
    The cells of the configuration component combine the pattern captured by the top of the stack component with the input symbol at the current time step to activate only the appropriate cell $\stackInputConfig{\stacksymbol{\gamma}}{\sym}$.
    This can be done by incorporating the information from the top of the stack component with the information about the current input symbol from $\inMtx \onehot{\sym_\tstep}$.
    More precisely, the parameters of $\rnn$ are set such that $\hiddState_\stackInputConfig{\stacksymbol{\gamma}}{\sym} = 1$ if at the previous one of the four sub-steps of the computation of the next hidden state, $\hiddState_\stackTopCell{\stacksymbol{\gamma}} = 1$ \emph{and} the input symbol is $\sym$.

    \paragraph{Computation component.}
    The computation component contains the cells in which the results of all the possible actions on the stack are executed.
    The parameters of the computation component are set such that, given that the previous stack configuration is $\hiddState_\stackCell{} = x_1 x_2 \ldots x_N$ and the input symbol is $\sym$, cells of the computation component are set as
    \begin{align*}
        \hiddState_\stackComp{\popOp}{\stacksymbol{\gamma}}  & = x_2 \ldots x_N                         \\
        \hiddState_\stackComp{\pushOp}{\stacksymbol{\gamma}} & = \stackSymToDigit{\sym} x_1 \ldots x_N.
    \end{align*}

    \paragraph{Acceptance component.}
    The cell in the acceptance component is activated if and only if the current input symbol is $\eos$ (denoting the end of the string whose recognition should be determined) and the stack is empty, i.e., the $\stackTopCellEmpty$ cell is activated.

    More precisely, the dynamics described here are implemented by the four steps of the hidden state update as follows (where $\hiddState^{\left(1\right)}_\tstep = \hiddStatet$).\franz{I think it should be $\hiddState^{(0)}$ to be consistent with above}
    \begin{itemize}
        \item In \emph{phase 1}, the configuration of the stack is determined by setting the top of the stack component in $\hiddState^{\left(2\right)}_\tstep$.
        \item In \emph{phase 2}, the configuration of the stack and the input symbol are combined by setting the configuration component in $\hiddState^{\left(3\right)}_\tstep$.
        \item In \emph{phase 3} all possible operations on the stack are performed in the computation component, and, at the same time, the results of all invalid operations (only one operation is valid at each time step due to the deterministic nature of $\pda$) are zeroed-out in $\hiddState^{\left(4\right)}_\tstep$.
              This is done by setting the entries of the recurrence matrix $\recMtx$ such that only the valid action is not zeroed out.
        \item In \emph{phase 4} the result of the executed operations (only one of which is non-zero) is copied over to the $\stackCell{}$ cell in the hidden state in $\hiddState_{\tstep + 1}$.
    \end{itemize}

    Having defined the intuition behind the dynamics of the hidden state updates, we now formally define how the parameters of the RNN are set to enable them.
    Whenever an entry of a matrix or vector is not set explicitly, it is assumed that it is $0$ (that is, we only explicitly set the non-zero values).
    Again, we define them for each component in turn.

    \paragraph{The data component.}
    The values of the parameters in the data component are set as follows.
    \begin{align}
        \eRecMtx_{\bufferCell{1}, \stackCell{}}         & = 1                                                                                                                                                     \\
        \eRecMtx_{\bufferCell{2}, \bufferCell{1}}       & = 1                                                                                                                                                     \\
        \eRecMtx_{\stackCell{}, \stackComp{\pushOp}{0}} & = \eRecMtx_{\stackCell{}, \stackComp{\pushOp}{1}} = \eRecMtx_{\stackCell{}, \stackComp{\popOp}{0}} = \eRecMtx_{\stackCell{}, \stackComp{\popOp}{1}} = 1
    \end{align}
    The first two elements correspond to moving the values to the next element in the data component queue, while the entries in the last row correspond to \emph{summing up} the values from the computation component to move them into the stack cell after the computation has been completed.
    Note that, of course, the elements of the computation component are \emph{always} summed up and written in the $\stackCell{}$ cell, no matter what the values there are.
    However, the division of the computation of the next hidden state into phases ensures that \emph{when it matters}, i.e., after the third phase, there is only a single computation component that is non-zero, and that one is copied into the $\stackCell{}$ component in the fourth computation sub-step.
    All other parameters (in $\inMtx$ and $\biasVech$) are $0$.

    \paragraph{The top of the stack component.}
    The parameters setting the top of the stack component are set as follows:
    \begin{align}
        \eRecMtx_{\stackTopCellEmpty, \stackCell{}} & = -10 \\
        \eRecMtx_{\stackTopCellZero, \stackCell{}}  & = -10 \\
        \eRecMtx_{\stackTopCellOne, \stackCell{}}   & = 10  \\
        \evb_{\stackTopCellEmpty}                   & = 1   \\
        \evb_{\stackTopCellZero}                    & = 3   \\
        \evb_{\stackTopCellOne}                     & = -2.
    \end{align}
    Other parameters ($\inMtx$) are $0$.
    The reasoning behind these parameters is the following.
    The cell $\stackCell{}$ contains the numeric encoding of the stack content.
    We distinguish three cases.
    \begin{itemize}
        \item If the stack is \emph{empty}, $\hiddState_\stackCell{} = 0$.
              Therefore, using the parameters above, the value of the cell $\stackTopCellOne$ after the sub-step update will be $0$, while the cells $\stackTopCellEmpty$ and $\stackTopCellZero$ will be $1$ due to the positive bias term.
              This might not be what you would expect---it might seem like, in this case, this step erroneously signals both an empty stack and a stack whose top component is $0$.
              This, however, is corrected for in the configuration component, as we discuss below.
        \item If the top of the stack is the symbol $\stacksymbol{0}$, $\hiddState_\stackCell{} = 0.1\ldots$.
              This means that $10 \cdot \hiddState_\stackCell{}\leq 1$ and, therefore, after the update rule application, $\hiddState_\stackTopCellOne = 0$.
              It is easy to see that the setting of the parameters also implies $\hiddState_\stackTopCellEmpty = 0$.
              However, since $-10 \cdot \hiddState_\stackCell{} \geq -2$, we have that $\hiddState_\stackTopCellZero = 1$.
        \item Lastly, if the top of the stack is the symbol $\stacksymbol{1}$, $\hiddState_\stackCell{} = 0.3\ldots$.
              Therefore, $10\cdot \hiddState_\stackCell{} \geq 3$, meaning that after the update rule application, $\hiddState_\stackTopCellOne = 1$.
              Again, it is easy to see that the setting of the parameters also implies $\hiddState_\stackTopCellEmpty = 0$.
              On the other hand, since $-10 \cdot \hiddState_\stackCell{} \leq -3$, it also holds that $\hiddState_\stackTopCellZero = 0$.
    \end{itemize}

    \paragraph{The configuration component.}
    The configuration component is composed of the most cells of any component:
    \begin{align}
        \eRecMtx_{\stackInputConfig{\stacksymbol{\gamma}}{\sym}, \stackTopCell{\stacksymbol{\gamma}}} & = 1 \textbf{ for } \stacksymbol{\gamma} \in \set{\eps, 0, 1}, \sym \in \set{\eos, \syma, \symb}  \\
        \eRecMtx_{\stackInputConfig{\stacksymbol{\gamma}}{0}, \stackTopCell{\stacksymbol{\eps}}}      & = -1 \textbf{ for } \sym \in \set{\eos, \syma, \symb}                                            \\
        \eInMtx_{\stackInputConfig{\stacksymbol{\gamma}}{\sym}, \symordering\left(\sym\right)}        & = 1 \textbf{ for } \stacksymbol{\gamma} \in \set{\eps, 0, 1}, \sym \in \set{\eos, \syma, \symb}  \\
        \evb_{\stackInputConfig{\stacksymbol{\gamma}}{\sym}}                                          & = -1 \textbf{ for } \stacksymbol{\gamma} \in \set{\eps, 0, 1}, \sym \in \set{\eos, \syma, \symb}
    \end{align}
    Here, the first, third, and fourth terms together ensure that the cell $\stackInputConfig{\stacksymbol{\gamma}}{\sym}$ is activated if the current input symbol is $\syma$ ($\eInMtx_{\stackInputConfig{\stacksymbol{\gamma}}{\sym}, \symordering\left(\sym\right)}$) and the top of the stack is $\stacksymbol{\gamma}$ ($\eRecMtx_{\stackInputConfig{\stacksymbol{\gamma}}{\sym}, \stackTopCell{\stacksymbol{\gamma}}}$).
    $\evb_{\stackInputConfig{\stacksymbol{\gamma}}{\sym}}$ ensures that both conditions have to be met.
    The second term, $\eRecMtx_{\stackInputConfig{\stacksymbol{0}}{\sym}, \stackTopCell{\stacksymbol{\eps}}}$, on the other hand, takes care of an edge case: as shown above, $\evb_{\stackTopCellZero} = 0$, which means that $\stackTopCellZero$ is, by default, set to $1$.
    The negative weight $\eRecMtx_{\stackInputConfig{\stacksymbol{0}}{\sym}, \stackTopCell{\stacksymbol{\eps}}} = -1$ ensures that, if the stack is indeed empty, the effect of this default value is ``canceled out'', i.e., the configuration cell is not activated by mistake.

    \paragraph{The computation component.}
    This is the most complicated component.
    The computation components are manipulated with the following parameters:
    \begin{align}
        \eRecMtx_{\pushOp{0}, \bufferCell{2}}                                                                   & = \eRecMtx_{\pushOp{1}, \bufferCell{2}} = \frac{1}{10}                                                                                                                   \\
        \eRecMtx_{\popOp{0}, \bufferCell{2}}                                                                    & = \eRecMtx_{\popOp{1}, \bufferCell{2}} = 10                                                                                                                              \\
        \eRecMtx_{\noOp, \bufferCell{2}}                                                                        & = 1                                                                                                                                                                      \\
        \eRecMtx_{\texttt{A}, \stackInputConfig{\stacksymbol{\gamma}}{\sym}}                                    & = -10 \textbf{ for } \texttt{A} \in \set{\stackComp{\pushOp}{0}, \stackComp{\pushOp}{1}, \stackComp{\popOp}{0}, \stackComp{\popOp}{1}, \stackCompNoOp}                   \\
                                                                                                                & \phantom{= -10 \textbf{ for }} \stacksymbol{\gamma} \in \set{0, 1}, \sym \in \set{\syma, \symb} \nonumber                                                                \\
        \eRecMtx_{\stackComp{\texttt{A}}{\stacksymbol{\gamma'}}, \stackInputConfig{\stacksymbol{\gamma}}{\sym}} & = 0 \textbf{ for } \pdaEdgenoweight{\stateq}{\sym}{\stateq}{\stacksymbol{\gamma}}{\stacksymbol{\gamma'}} \in \trans,                                                     \\
                                                                                                                & \phantom{= 0 \textbf{ for }} \texttt{A} \in \set{\stackComp{\pushOp}{0}, \stackComp{\pushOp}{1}, \stackComp{\popOp}{0}, \stackComp{\popOp}{1}, \stackCompNoOp} \nonumber \\
        \eRecMtx_{\stackCompNoOp, \stackInputConfig{\stacksymbol{\gamma}}{\eos}}                                & = 0 \textbf{ for } \stacksymbol{\gamma} \in \set{\eps, 0, 1}                                                                                                             \\
        \evb_{\stackComp{\pushOp}{0}}                                                                           & = \frac{1}{10}                                                                                                                                                           \\
        \evb_{\stackComp{\pushOp}{1}}                                                                           & = \frac{3}{10}                                                                                                                                                           \\
        \evb_{\stackComp{\popOp}{0}}                                                                            & = -1                                                                                                                                                                     \\
        \evb_{\stackComp{\popOp}{1}}                                                                            & = -3                                                                                                                                                                     \\
        \evb_{\stackCompNoOp}                                                                                   & = 0
    \end{align}
    The first three parameters above concern \emph{copying} the value of the stack encoded by the previous hidden state into the computation component and preparing it for modification.
    They work together with the corresponding entries in the bias vector $\biasVech$.
    For example, a value can be \emph{pushed} onto the stack by dividing the value of the stack encoding by $10$ and adding either $0.1$ or $0.3$, depending on whether $0$ or $1$ is being pushed.
    This is encoded by the first setting above and $\evb_{\stackComp{\pushOp}{0}}$ and $\evb_{\stackComp{\pushOp}{1}}$.
    Similarly, a value can be popped from the stack by multiplying the stack encoding with $10$ and then subtracting the appropriate value according to the bias entry.
    The $\noOp$ action is implemented simply by copying the values of the stack into its cell.
    The remaining three parameter settings above ensure that, after executing all possible stack actions, \emph{only the appropriate computation} is kept and all others are zeroed out.
    The fourth row above ensures that ``by default'', all computation cells are reset to $0$ after every update.
    However, the next row ``removes'' the negative weights (sets them to $0$) for the changes in the configuration which correspond to the \emph{valid transitions}, or valid actions, in the pushdown automaton.
    That is, setting those values of the matrix $\recMtx$ to $0$ disables ``erasing'' the entry $\stackComp{\texttt{A}}{\stacksymbol{\gamma'}}$ in the hidden state by the configuration $\stackInputConfig{\stacksymbol{\gamma}}{\sym}$ \emph{if} the transition from the configuration with  the top of the stack $\stacksymbol{\gamma}$ to $\stacksymbol{\gamma'}$ with the action \texttt{A} upon reading $\sym$ is encoded by the original automaton.
    The last remaining row simply ensures that reading in the $\eos$ symbol results in the $\noOp$ action being executed ($\eos$ actions are not encoded by the original pushdown automaton).

    \paragraph{The acceptance component.}
    Lastly, the acceptance component is controlled by the following parameters:
    \begin{align}
        \eRecMtx_{\acceptCell, \texttt{A}}                                    & = -10 \textbf{ for } \texttt{A} \in \set{\stackComp{\pushOp}{0}, \stackComp{\pushOp}{1}, \stackComp{\popOp}{0}, \stackComp{\popOp}{1}, \stackCompNoOp} \\
        \eRecMtx_{\acceptCell, \stackInputConfig{\stacksymbol{\gamma}}{\sym}} & = -10 \textbf{ for } \stacksymbol{\gamma} \in \set{\stacksymbol{\eps}, \stacksymbol{0}, \stacksymbol{1}}, \sym \in \set{\eos, \syma, \symb}            \\
        \evb_{\acceptCell}                                                    & = 1
    \end{align}
    The entry $\evb_{\acceptCell}$ ensures that, by default, the value of $\acceptCell$ is set to $1$.
    However, the other parameters ensure that, as soon as any part of the configuration is not compatible with the acceptance state (the read symbol is not $\eos$ or the stack is not empty), the acceptance bit is turned off.

    A full proof of the theorem would now require us to show formally that the update rule \cref{eq:tupda-final-phase} results in the correct transitions in the PDA.
    We, however, leave the proof with the intuitive reasoning behind the setting of the parameters and leave this as an exercise for the reader.
    The proof is also demonstrated in the \texttt{python} implementation of the constructions here: \url{https://github.com/rycolab/rnn-turing-completeness}.
\end{proof}

The construction described in the proof of \cref{thm:rnn-pda} is demonstrated in the following example.
\begin{example}{Siegelmann's construction}{}
    Let $\pushdown$ be a single-stack PDA presented in \cref{fig:single-stack-pda-siegelmann-example}.
    We now simulate the recognition of the string $\str = \syma \symb$, which is accepted by $\pushdown$.
    The initial state $\hiddStateZero$ has a single non-zero cell, $\stackTopCell{\eps}$.
    The four phases of the processing of the first input symbol $\syma$ are shown in \cref{tab:single-stack-pda-simulation-1}.
    The four phases of the processing of the second input symbol $\symb$ are shown in \cref{tab:single-stack-pda-simulation-2}.

\end{example}

\begin{figure}[t]
    \centering
    \begin{tikzpicture}[node distance = 15mm]
        \node[state, initial, accepting] (q0) [] { $\stateq$ };
        \draw[transition]  (q0) (q0) edge[above, loop above, align=center] node{ $\syma, \stacksymbol{\eps} \to \stacksymbol{0}$ \\ $\syma, \stacksymbol{0} \to \stacksymbol{1}$ \\ $\syma, \stacksymbol{1} \to \stacksymbol{\eps}$} (q0)
        (q0) edge[right, loop right, align=center] node{ $\symb, \stacksymbol{\eps} \to \stacksymbol{1}$ \\ $\symb, \stacksymbol{0} \to \stacksymbol{\eps}$ \\ $\symb, \stacksymbol{1} \to \stacksymbol{1}$} (q0);
    \end{tikzpicture}
    \caption{The single-stack pushdown automaton $\pushdown$.}
    \label{fig:single-stack-pda-siegelmann-example}
\end{figure}
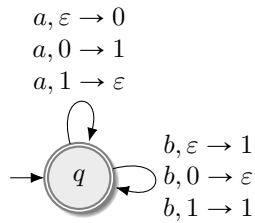
\begin{table}[]
    \centering
    \begin{tabular}{cccccc}
        \toprule
                                         & Initial state & Phase 1        & Phase 2        & Phase 3        & Phase 4                            \\
        \midrule
        $\stackCell{}$                   & $0$           & $0$            & $\frac{2}{5}$  & $\frac{2}{5}$  & $\textcolor{ETHRed}{\frac{1}{10}}$ \\
        $\bufferCell{1}$                 & $0$           & $0$            & $0$            & $\frac{2}{5}$  & $\frac{2}{5}$                      \\
        $\bufferCell{2}$                 & $0$           & $0$            & $0$            & $0$            & $\frac{2}{5}$                      \\
        $\stackTopCellEmpty$             & $1$           & $1$            & $1$            & $0$            & $0$                                \\
        $\stackTopCellZero$              & $0$           & $1$            & $1$            & $0$            & $0$                                \\
        $\stackTopCellOne$               & $0$           & $0$            & $0$            & $1$            & $1$                                \\
        $\stackInputConfig{\eos}{\syma}$ & $0$           & $0$            & $1$            & $1$            & $0$                                \\
        $\stackInputConfig{\eos}{\symb}$ & $0$           & $0$            & $0$            & $0$            & $0$                                \\
        $\stackInputConfig{0}{\syma}$    & $0$           & $0$            & $0$            & $0$            & $0$                                \\
        $\stackInputConfig{0}{\symb}$    & $0$           & $0$            & $0$            & $0$            & $0$                                \\
        $\stackInputConfig{1}{\syma}$    & $0$           & $0$            & $0$            & $0$            & $1$                                \\
        $\stackInputConfig{1}{\symb}$    & $0$           & $0$            & $0$            & $0$            & $0$                                \\
        $\stackInputConfig{\eps}{\eos}$  & $0$           & $0$            & $0$            & $0$            & $0$                                \\
        $\stackInputConfig{0}{\eos}$     & $0$           & $0$            & $0$            & $0$            & $0$                                \\
        $\stackInputConfig{1}{\eos}$     & $0$           & $0$            & $0$            & $0$            & $0$                                \\
        $\stackComp{\pushOp}{0}$         & $0$           & $\frac{1}{10}$ & $\frac{1}{10}$ & $\frac{1}{10}$ & $\frac{1}{10}$                     \\
        $\stackComp{\pushOp}{1}$         & $0$           & $\frac{3}{10}$ & $\frac{3}{10}$ & $0$            & $0$                                \\
        $\stackComp{\popOp}{0}$          & $0$           & $0$            & $0$            & $0$            & $0$                                \\
        $\stackComp{\popOp}{1}$          & $0$           & $0$            & $0$            & $0$            & $0$                                \\
        $\stackCompNoOp$                 & $0$           & $0$            & $0$            & $0$            & $0$                                \\
        $\acceptCell$                    & $0$           & $1$            & $0$            & $0$            & $0$                                \\
        \bottomrule
    \end{tabular}
    \caption{The simulation of the processing of the first symbol $\syma$ by the RNN simulating the PDA in \cref{fig:single-stack-pda-siegelmann-example}. After the fourth phase, the stack cell contains the encoding of the stack as $0.1$.}
    \label{tab:single-stack-pda-simulation-1}
\end{table}
\begin{table}[]
    \centering
    \begin{tabular}{cccccc}
        \toprule
                                         & Initial state & Phase 1  & Phase 2  & Phase 3  & Phase 4                   \\
        \midrule
        $\stackCell{}$                   & $ 1/10 $      & $ 1/10 $ & $ 1 $    & $ 2/5 $  & $ \textcolor{ETHRed}{0} $ \\
        $\bufferCell{1}$                 & $ 2/5 $       & $ 1/10 $ & $ 1/10 $ & $ 1 $    & $ 2/5 $                   \\
        $\bufferCell{2}$                 & $ 2/5 $       & $ 2/5 $  & $ 1/10 $ & $ 1/10 $ & $ 1 $                     \\
        $\stackTopCellEmpty$             & $ 0 $         & $ 0 $    & $ 0 $    & $ 0 $    & $ 0 $                     \\
        $\stackTopCellZero$              & $ 0 $         & $ 1 $    & $ 1 $    & $ 0 $    & $ 0 $                     \\
        $\stackTopCellOne$               & $ 1 $         & $ 0 $    & $ 0 $    & $ 1 $    & $ 1 $                     \\
        $\stackInputConfig{\eos}{\syma}$ & $ 0 $         & $ 0 $    & $ 0 $    & $ 0 $    & $ 0 $                     \\
        $\stackInputConfig{\eos}{\symb}$ & $ 0 $         & $ 0 $    & $ 0 $    & $ 0 $    & $ 0 $                     \\
        $\stackInputConfig{0}{\syma}$    & $ 0 $         & $ 0 $    & $ 0 $    & $ 0 $    & $ 0 $                     \\
        $\stackInputConfig{0}{\symb}$    & $ 0 $         & $ 0 $    & $ 1 $    & $ 1 $    & $ 0 $                     \\
        $\stackInputConfig{1}{\syma}$    & $ 1 $         & $ 0 $    & $ 0 $    & $ 0 $    & $ 0 $                     \\
        $\stackInputConfig{1}{\symb}$    & $ 0 $         & $ 1 $    & $ 0 $    & $ 0 $    & $ 1 $                     \\
        $\stackInputConfig{\eps}{\eos}$  & $ 0 $         & $ 0 $    & $ 0 $    & $ 0 $    & $ 0 $                     \\
        $\stackInputConfig{0}{\eos}$     & $ 0 $         & $ 0 $    & $ 0 $    & $ 0 $    & $ 0 $                     \\
        $\stackInputConfig{1}{\eos}$     & $ 0 $         & $ 0 $    & $ 0 $    & $ 0 $    & $ 0 $                     \\
        $\stackComp{\pushOp}{0}$         & $ 1/10 $      & $ 0 $    & $ 0 $    & $ 0 $    & $ 0 $                     \\
        $\stackComp{\pushOp}{1}$         & $ 0 $         & $ 0 $    & $ 0 $    & $ 0 $    & $ 0 $                     \\
        $\stackComp{\popOp}{0}$          & $ 0 $         & $ 0 $    & $ 0 $    & $ 0 $    & $ 0 $                     \\
        $\stackComp{\popOp}{1}$          & $ 0 $         & $ 1 $    & $ 0 $    & $ 0 $    & $ 0 $                     \\
        $\stackCompNoOp$                 & $ 0 $         & $ 0 $    & $ 2/5 $  & $ 0 $    & $ 0 $                     \\
        $\acceptCell$                    & $ 0 $         & $ 0 $    & $ 0 $    & $ 0 $    & $ 0 $                     \\
        \bottomrule
    \end{tabular}
    \caption{The simulation of the processing of the second symbol $\symb$ by the RNN simulating the PDA in \cref{fig:single-stack-pda-siegelmann-example}. After the fourth phase, the stack cell contains the encoding of the empty stack.}
    \label{tab:single-stack-pda-simulation-2}
\end{table}

\cref{thm:rnn-pda} shows that Elman RNNs are theoretically at least as expressive as deterministic CFGs.
We now return to the main result of this subsection: the Turing completeness of RNNs.
Luckily, \cref{thm:rnn-pda} gets us most of the way there!
Recall that by \cref{thm:turing-completeness-PDA}, \emph{two-stack} PDA are Turing complete.
We make use of this fact by generalizing the construction in the proof of \cref{thm:rnn-pda} to the two-stack case.
This will prove that RNNs can in fact simulate any Turing machine, and are, therefore, Turing complete.
\begin{lemma}{}{}
    Let $\pda = \twostackpdatuple$ be a two-stack pushdown automaton.
    Then, there exists an Elman RNN $\rnn$ simulating $\pushdown$, i.e., $\lang\left(\rnn\right) = \lang\left(\pda\right)$.
\end{lemma}
\begin{proof}
    Again, given a two-stack PDA $\pda = \twostackpdatuple$ with $\alphabet = \set{\syma, \symb}$, $\stackalphabet_1 = \set{0, 1}$, and $\stackalphabet_2 = \set{0,1}$, we construct an Elman RNN $\rnn = \elmanrnntuple$ which recognizes the same language as $\pda$.

    The hidden state of $\rnn$ will contain the same components as in the proof of \cref{thm:rnn-pda}.
    Moreover, their dynamics will be exactly the same---they will simply be \emph{larger} to account for more possible configurations of the two stacks together.
    For example, the top of the stack component will now consist of cells $\stackTopCell{\stacksymbol{\gamma_1}\stacksymbol{\gamma_2}}$ for $\stacksymbol{\gamma_1} \in \stackalphabet_1$ and $\stacksymbol{\gamma_2} \in \stackalphabet_2$ flagging that the top symbol on stack $1$ is $\stacksymbol{\gamma_1}$ and the top symbol of stack $2$ is $\stacksymbol{\gamma_2}$.
    Furthermore, the configuration component will contain cells of the form $\stackComp{\text{action}}{\stacksymbol{\gamma}_1\stacksymbol{\gamma}_2}$ with an analogous interpretation.
    Lastly, all the computation and data component cells would be duplicated (with one sub-component for each of the stacks), whereas the acceptance component stays the same.
    We now again describe these components and their dynamics intuitively, whenever there is any difference to the single-stack version.

    \paragraph{Data component.}
    Instead of having a \emph{single} queue of data cells, $\rnn$ now has \emph{two} queues, one for each of the stacks.
    The first queue will be formed by $\stackCell{1}$, $\bufferCell{11}$, and $\stackCell{21}$, and the second one by $\stackCell{2}$, $\bufferCell{12}$, and $\stackCell{22}$.
    Each of the queues acts exactly the same as in the single-stack version, and they act independently based on the computations done in the computation cells of the respective stacks.
    Each of the stacks is also encoded as a numeric sequence in the same way as in the single-stack version.


    \paragraph{Top of stack component.}
    Again, $\rnn$ starts in an initial state in which the cell $\stackTopCell{\eps\eps}$ is $1$ and all others are $0$.
    The individual cells of this component then get updated according to the top symbols on the stacks encoded in the $\stackCell{1}$ and $\stackCell{2}$ cells.

    \paragraph{Configuration component.}
    The cells of the configuration component combine the pattern captured by the top of \emph{both} stack components with the input symbol at the current time step to activate only the appropriate cell $\stackInputConfig{\stacksymbol{\gamma_1}\stacksymbol{\gamma_2}}{\sym}$.

    \paragraph{Computation component.}
    Again, the computation component contains the cells in which the results of all the possible actions on both stacks are executed.
    They execute the actions on both stacks independently.

    \paragraph{Acceptance component.}
    The acceptance component functions identically to the single-stack case.

    Using these components, the RNN then transitions between the phases exactly like in the single-stack case.
    We leave the specifications of the matrix parameter values to the reader.
    They again follow those presented in the single-stack case but treat the transitions and configurations of both stacks.

\end{proof}

\subsection{The Computational Power of RNN Variants}

Most of the current section was devoted to understanding the computational power of simple Elman RNN language models due to their simplicity which allows an easy connection to formal models of computation.
However, as mentioned in \cref{sec:var-rnn}, gated RNN variants such as LSTMs and GRUs have become the standard in modern natural language processing tasks, showing more better and more reliable performance on a variety of tasks.
Interestingly, besides their better resilience to the vanishing gradient problem, LSTM-based language models are also provably more \emph{expressive} than simple RNN language models.
On the other hand, the simpler GRU-based language models are in some ways only as powerful as simple Elman RNN language models.
To give an intuition behind this, this subsection provides a short overview the results considering the computational power of RNN variants.

\citet{weiss-etal-2018-practical} compare the practical computational power of different RNN variants under the constraint of bounded computation time and limited precision.
Interestingly they empirically find that LSTMs can learn to recognize languages that require some form of \emph{counting}, like $\set{a^nb^n\mid n \in \N}$ or $\set{a^nb^nc^n\mid n \in \N}$, while GRUs struggle to do so.
This invites the comparison of LSTMs with \defn{counter machines} \citep{hopcroft79,Fischer1968CounterMA}, a class of formal computational models with the ability to count.
Simply put, counter machines are finite-state automata with an additional (unbounded) counter cell, which they can manipulate by incriminating and decrementing the value stored in it.\footnote{We only consider counter machines with a \emph{single} counter cell.}
Counter machines present an interesting addition to the traditional hierarchy of computational models, since they in some ways \emph{cut across} it, being able to recognize some, but not all, context-free languages, while also being able to recognize some \emph{context-sensitive} languages.
Among others, they can for example recognize the context-free language $\set{a^nb^n\mid n \in \N}$ and the context-sensitive language $\set{a^nb^nc^n\mid n \in \N}$, while not being able to recognize the Dyck context-free languages $\dyck{\nBracketTypes}$ with $\nBracketTypes$ different parenthesis types (intuitively, this is because recognizing Dyck languages requires keeping track of the order in which the parantheses appeared, not only their counts).

Counter machines can recognize languages like $\set{a^nb^nc^n\mid n \in \N}$, by counting the number of $a$s and making sure that it matches the number of $b$s.
Further, analyzing the activation of the memory cell and of the hidden state, they find that LSTMs that one can recognize the use of a counting mechanism implemented by the network by using one or more dimensions of the memory cell as counters.
This result is particularly interesting as GRUs are often considered an equivalent variant to the LSTM one, with the same computational power, but smaller computation overhead.
However, GRUs seem to lack this counting behavior, which can be backed by theoretical analysis.

\citet{merrill-etal-2020-formal} take a look at the computational power of the different RRN variant from another perspective.
They consider space complexity and whether the networks are rationally recurrent, i.e. whether their hidden state update function can be expressed in terms of finite state machine computations.
Making the assumption of saturated networks\footnote{Informally, a saturated network is a neural network in which the norms of the parameters are taken to $\infty$.
    This has the effect of pushing all the squashing functions of the network to one of their extreme values, in the case of the sigmoid $\{ 0,1 \}$ and $\{-1,1 \}$ in the case of the hyperbolic tangent}, they find that while GRUs and Elman networks are rationally recurrent and therefore at most regular, LSTMs are not, meaning that their hidden state update function cannot be expressed by means of finite-state machines.

\begin{figure}[t]
    \center
    \includegraphics[scale=0.33]{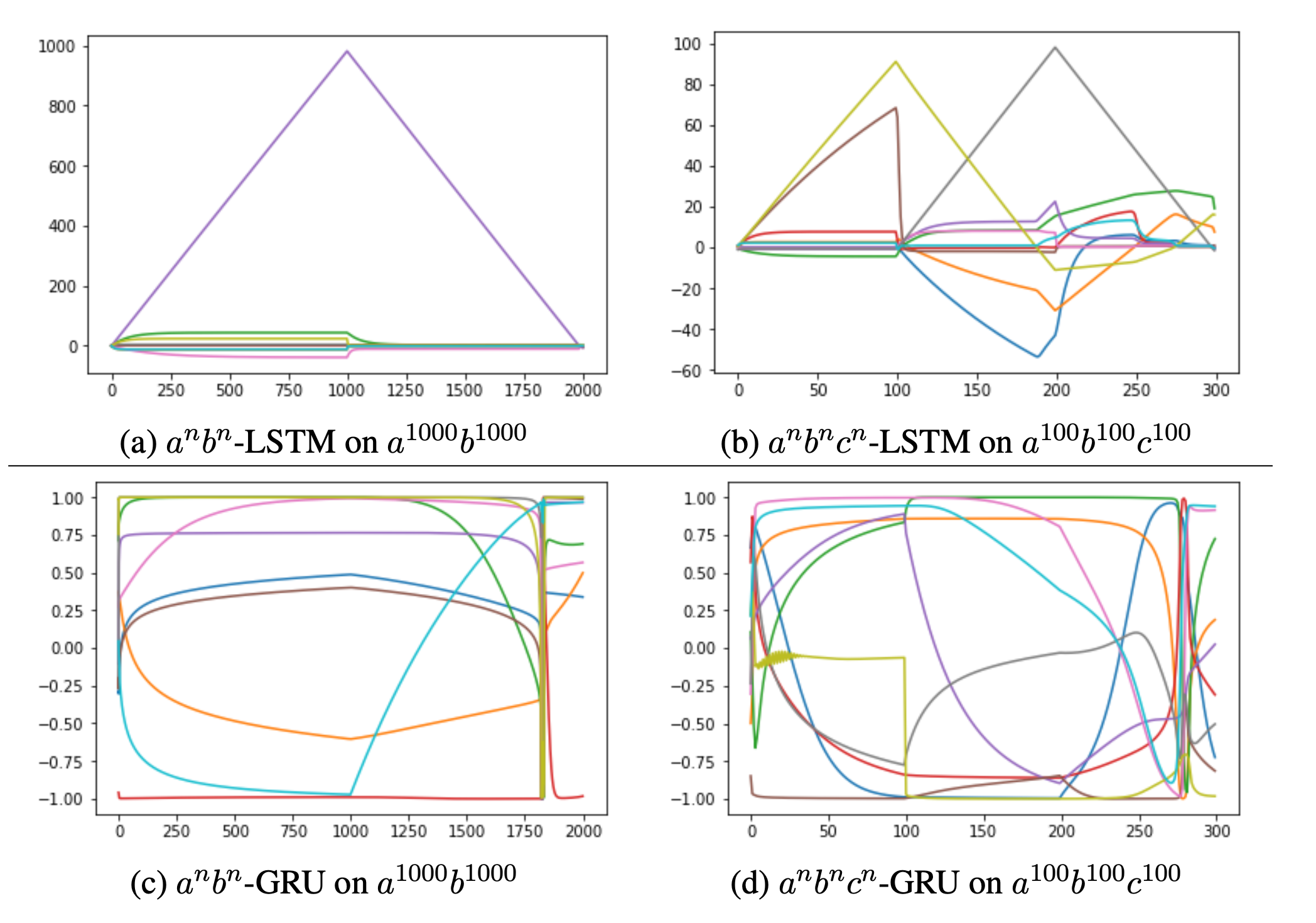}
    \caption{Picture from \citet{weiss-etal-2018-practical}: Plot of the activation value of the memory cell (LSTM) and of the hidden state (GRU), versus the indices of the input.
        The Networks have been trained to recognize either the languages $\set{a^nb^n\mid n \in \N}$ or $\set{a^nb^nc^n\mid n \in \N}$.
        As you can notice, in both cases the LSTM has learned to use one or more dimension of the memory cell as a \emph{counter}, which allows it to count how many $a$ and $b$ have been have been consumed so far.
        Conversely, the GRU hasn't developed this counter mecahnism, and in fact empirical evidence shows that it struggles to to recognize the described languages.}
\end{figure}

\subsection{Consequences of the Turing completeness of recurrent neural networks}
The section above outlines \citeposs{Siegelmann-Sontag-Computational-Power} construction of encoding a Turing machine in an RNN.
While Turing completeness means that RNNs are in many ways computational very powerful (as powerful as they can be), it also brings with it many computational challenges faced when working with Turing machines.
Computability theory, for example, defines many problems related to Turing machines and their properties which are \emph{not computable}, or \emph{undecidable}, meaning that no algorithm (or, equivalently, a Turing machine) can solve them.\footnote{The notion of solving a problem computationally is quite nuanced and we only provide very broad intuitions for now. Readers who want to delve deeper into the topic of computability and with it the implications of Turing completeness are encouraged to look at some classical texbooks on the material, for example \citet[][Part Two]{Sipser2013}.}
The most classical and fundamental of such problems is the halting problem \citep{turing-1937}.
\begin{definition}{Halting problem}{}
    Let $\tm$ be a Turing machine over the input alphabet $\alphabet$ and $\str \in \kleene{\alphabet}$.
    The \defn{halting problem} is the problem of deciding whether $\tm$ (halts and) accepts $\str$.\footnote{A formal definition of the halting problem would require us to define the formal language $\lang$ of Turing machine and input tuples $\left(\tm, \str\right)$ and asking if there exists a Turing machine $\tm'$ that accepts $\lang$. However, to keep the discussion brief, we define the problem more intuitively.}
\end{definition}

The halting problem is the foundation of the results presented by \citet{Chen2018}, which building on \citet{Siegelmann-Sontag-Computational-Power} considers its implications on the theory of RNN language models.
For example, they show that determining many practically useful properties of general rationally weighted RNNs is \emph{undecidable}.
Such properties include the tightness of a general RNN,\footnote{Note that the results from \cref{sec:rnn-tightness} consider only RNN LM with the $\softmax$ projection function.} the equivalence of two RNN language models, the minimal size (in the size of the hidden state) of an RNN defining the same language model as some given RNN, and the highest probability string in an RNN, i.e., $\argmax_{\str \in \kleene{\alphabet}} \pLN\left(\str\right)$.
By simulating a Turing machine, we can encode the halting problem as solving any of those tasks.\footnote{In computability theory, this is known as \emph{reducing} the halting problem to the given one}.
This means that if we could solve these problems for RNNs, we could also solve the halting problem, which is provably impossible, meaning that these problems are not computable either.
We briefly outline the individual findings of \citet{Chen2018} below, sketching the rough idea behind the proofs.\footnote{The interested reader is encouraged to see the original paper for the details.}
\begin{theorem}{Tightness}{}
    Determining tightness of an RNN language model is undecidable.
\end{theorem}
\begin{proof}
    Note first that not all RNNs are tight language models.
    As mentioned, this is not a contradiction to the results from \cref{sec:rnn-tightness}, which considered softmax-projected RNN language models.
    Indeed, the RNN we constructed in \cref{sec:rnn-turing-completeness} is not such an RNN.
    That construction shows that, given an arbitrary Turing machine $\tm$ and input $\str$, we can construct an RNN that simulates $\tm$ running on $\str$.
    Using this construction, we can reduce the problem of deciding the tightness of a general (rational weighted Elman) RNN to the halting problem, i.e., the problem of answering ``Does $\tm$ halt on $\str$?''.
    We do that by constructing an RNN that simulates the given Turing machine $\tm$ on the input $\str$, ending generation if $\tm$ halts, and, at the same time, produces strings according to a distribution that is tight for finite length strings, on the condition that no infinite length strings can be produced.
    Now if and only if $\tm$ halts on $\str$, the language model is tight.
    Therefore, deciding is at least as hard as solving the halting problem, and since that is undecidable, so is tightness.
\end{proof}
\begin{theorem}{Highest weighted string}{}
    Finding the highest weighted string of an RNN LM is undecidable.
\end{theorem}
\begin{proof}
    Once more, we reduce the halting problem to the given task on the RNN.
    The idea behind this is to again simulate an arbitrary Turing machine by constructing an RNN LM which is not tight unless the simulated Turing machine halts.
    One can create such an RNN with a one-symbol output alphabet such that its highest-weighted output is an infinite string.
    Now, if we again enforce that the RNN ends producing outputs once the simulated Turing machine halts, it can be shown that there exists a language in which each string has a probability of less than $0.12$ if and only if the Turing machine does not halt.
    On the other hand, if the Turing machine does halt after $T$ steps, producing a string which has length $3T-5$ has a probability of $\geq 0.25$.
    Therefore, the weight of the highest probability string depends on the question of whether the simulated Turing machine halts, which is undecidable.
\end{proof}
\begin{theorem}{Equivalence}{}
    Equivalence between two RNN LMs in the sense of defining the same language model is undecidable.
\end{theorem}
\begin{proof}
    The proof of this claim is again a reduction from the halting problem.
    We construct an RNN which simulates a given arbitrary Turing machine until it halts and has the same outputs as some other RNN.
    As soon as the Turing machine halts, the outputs differ, so the RNNs are equivalent if and only if the Turing machine does not halt.
    Hence, equivalence is undecidable.
\end{proof}
\begin{theorem}{Minimization}{}
    Finding the RNN with the minimum number of hidden layer neurons defining the same language model as a given RNN LM is undecidable.
\end{theorem}
\begin{proof}
    We can reduce the halting problem to the following problem: For some RNN LM and an integer $\hiddDim$, return yes if there is another RNN LM with $\leq \hiddDim$ hidden units that generates the same weighted language.
    Assume that there is a Turing machine $\tm$ that can decide this problem.
    Now, for another Turing machine $\tm'$ and input $\str$, construct a one symbol RNN LM, $\rnn$, that simulates $\tm'$ running on $\str$ and stops generating if $\tm'$ halts.
    We assume without loss of generality that $\tm'$ runs for more than one computation step.
    Now we run $\tm$ on the input $\left(\rnn, 0\right)$, which checks whether there is another RNN LM generating the same weighted language as $\rnn$ and has no hidden units.
    Having no hidden units means that the output probabilities of each symbol would have to be constant for each time step.
    If $\tm$ decides minimization returns true, that means the output probabilities of $\rnn$ cannot change over time, which means that $\tm'$ has to run forever.
    Conversely, if $\tm$ returns false, the output probabilities change when $\tm'$ halts.
    Therefore, $\tm$ deciding the minimal hidden states problem on $\left(\rnn, 0\right)$ is equivalent to it deciding the Halting problem for $\left(\tm',\str\right)$.
\end{proof}

This concludes our investigation of the formal properties of recurrent neural language models.
The sequential nature of the architecture and the relatively simple transition functions in the vanilla RNN architectures made the link to automata from formal language theory relatively straightforward, which allowed relatively strong theoretical insights.
However, it was exactly this sequential nature and the issues associated with it of RNNs that eventually led to them being overtaken by another neural architecture, which is now at the core of most if not all, modern state-of-the-art language models: the transformer.\footnote{We will not discuss the issues with the training speed and parallelization of RNNs in detail. Some of these issues will be highlighted in the latter parts of the course.}
We introduce them and discuss their theoretical underpinnings in the next section.

\newpage{}

\section{Transformer-based Language Models}\label{sec:transformers}
In the previous section (\cref{sec:rnns}), we introduced and studied RNN language models as of language models capable of storing an arbitrarily long context in its encoding $\encfunc{\strlt}$ by updating their hidden state $\hiddStatet$ an arbitrary number of times.
As we saw in their theoretical analysis, this mechanism gives them, under some assumptions, a lot of expressive power.
Nevertheless, as we also discussed, RNN LMs come with their distinct set of drawbacks.
Some of them, e.g., the exploding and vanishing gradient problems, can be amended using specific mechanisms resulting in more complex recurrent neural networks, such as LSTMs and GRUs (cf. \cref{sec:var-rnn}).
As discussed in \cref{sec:rnn-parallelization}, a more fundamental issue that cannot be avoided is the difficulty of parallel training, which is particularly noticeable on the vast internet-scale corpora used nowadays to train language models.
Can we do anything about that?
As discussed, the inherently sequential nature of RNNs suggests strict limits on this front.
Motivated by this limitation, in this section, we present a newer architecture that took over the field of language modeling (and NLP in general)---transformers~\citep{Vaswani2017}.
It was originally introduced for machine translation, but it can easily be applied to language modeling and has led to the success of models such as GPT-n.

The structure of this section will be a bit different from that of the other sections in the notes.
We will first give a formal definition of a transformer model in \cref{sec:transformers-formal} and, based on this definition, derive a number of results analogous to those for RNN LMs from \cref{sec:rnn-expressiveness}.
However, due to the current practical relevance of transformers in language modeling, we then devote a significant portion of the section to more practical aspects of transformer models and introduce a number of modifications used in modern language modeling systems.

\subsection{Informal Motivation of the Transformer Architecture}
\label{sec:transformers-informal}
Before we introduce transformers, let us consider another practical drawback of RNN LMs, which will give us more clues on how to improve them and motivate the architectural decisions behind transformers.
Luckily, the simplest patch-up of this issue will also lend itself naturally to parallelization, as we will see shortly.
The main characteristic of RNNs is the use of a single hidden state $\hiddStatet$ to represent an arbitrary prefix of any string $\strlt$ up to the current time step $\tstep$.
While this allows RNNs to model strings of any length, it also means that arbitrarily long strings must be \emph{compressed} into this hidden vector of fixed size.
Intuitively, this becomes increasingly difficult as the length of the context grows: As the amount of information to be compressed into the hidden state increases with the prefix length, the hidden state may struggle to model the entirety of the preceding context.
How can we amend that?
The simplest na{\"i}ve way to go about this is to retain the contextual encodings of \emph{all} prefixes of the string.
In this case, it is actually more natural to talk about contextual encodings not of full prefixes, but simply of the individual symbols in the string.\footnote{For example, looking back to RNNs, we could consider $\hiddStatet$ to simply be an encoding of the symbol $\symt$ augmented with the information from $\strlt$.}
Here, contextual means that the symbol encodings are augmented with the information from the rest of the string (in most cases, about the preceding context, as with the hidden states of an RNN).
With this, we avoid the need to summarize the entire context into a single state.
Note that our infinite-precision RNNs from the previous section implicitly did that as well---for example, by storing the information in the ``stack'' neurons, they could, in principle, store the entire history of the string.
However, storing all the encodings explicitly makes their utilization more direct and thus easier.
This of course leaves the model with the issue of remembering increasingly large amounts of information as the length of the context increases, but we will, for the moment, assume that we can always store enough information to process any string in this way.

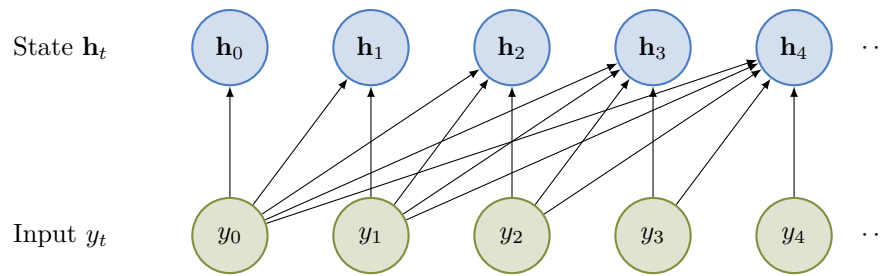
\begin{figure}
    \centering
    \begin{tikzpicture}[x=1.5cm, y=2.5cm,>=latex]

        \foreach \i in {0,...,4}
        \node[circle, draw=ETHGreen!80!white, thick, fill=ETHGreen!20, inner sep=0pt, minimum size=10mm] (y\i) at (1.25*\i, 0) {$\sym_\i$};

        \foreach \i in {0,...,4}
        \node[circle, draw=ETHBlue!80!white, thick, fill=ETHBlue!20, inner sep=0pt, minimum size=10mm] (H\i) at (1.25*\i, 1) {$\hiddState_\i$};

        \foreach \i in {0,...,4} {
                \foreach \j in {0,...,4} {
                        \ifthenelse{\i < \j}{}{
                            \draw[->] (y\j) -- (H\i);}
                    }}

        \node at (-1.5, 1) {State $\hiddState_\tstep$};
        \node at (-1.5, 0) {Input $\sym_\tstep$};
        \node[circle, draw=none, minimum size=10mm] (yend) at (5.75, 0) {$\cdots$};
        \node[circle, draw=none, minimum size=10mm] (Hend) at (5.75, 1) {$\cdots$};

    \end{tikzpicture}
    \caption{An abstract depiction of how a transformer language model produces the contextual embeddings of all symbols in a string.
        The hidden state $\hiddStatet$ can ``attend to'' (the precise meaning of this term will be introduced soon in \cref{sec:transformers-formal}) all preceding symbols $\strlt$ and the current symbol $\symt$.}
    \label{fig:transformer-abstract}
\end{figure}

\begin{figure}
    \centering
    \begin{tikzpicture}[x=1.5cm, y=2.5cm,>=latex]

        \foreach \i in {0,...,3}
        \node[circle, draw=ETHGreen!80!white, thick, fill=ETHGreen!20, inner sep=0pt, minimum size=10mm] (y\i) at (1.55*\i, 0) {$\sym_\i$};

        \foreach \i in {0,...,2}
        \node[circle, draw=ETHBlue!80!white, thick, fill=ETHBlue!20, inner sep=0pt, minimum size=10mm] (H\i) at (1.55*\i, 1) {$\hiddState_\i$};

        \foreach \i in {0,...,2} {
                \foreach \j in {0,...,2} {
                        \ifthenelse{\i < \j}{}{
                            \draw[->] (y\j) -- (H\i);}
                    }}

        \node at (-1.5, 1) {State $\hiddState_\tstep$};
        \node at (-1.5, 0) {Input/Output $\sym_\tstep$};
        \node[circle, draw=none, minimum size=10mm] (yend) at (5.4, 0) {$\cdots$};
        \node[circle, draw=none, minimum size=10mm] (Hend) at (4.4, 1) {$\cdots$};

        \draw[->, thick, dotted, color=ETHGreen] (H0) to node[above, sloped] {\scriptsize $\sym_1 \sim \pLNSM\left(\cdot \mid \hiddState_0\right)$} (y1);
        \draw[->, thick, dotted, color=ETHGreen] (H1) to node[above, sloped] {\scriptsize $\sym_2 \sim \pLNSM\left(\cdot \mid \hiddState_1\right)$} (y2);
        \draw[->, thick, dotted, color=ETHGreen] (H2) to node[above, sloped] {\scriptsize $\sym_3 \sim \pLNSM\left(\cdot \mid \hiddState_2\right)$} (y3);

    \end{tikzpicture}

    \caption{An abstract depiction of how a transformer language model \emph{generates} a string one symbol at a time.
        The hidden state $\hiddStatet$ can attend to all previously generated symbols $\strlt$ to sample the next symbol $\symt$.
        The dotted lines denote the sampling steps.}
    \label{fig:transformer-abstract-generative}
\end{figure}
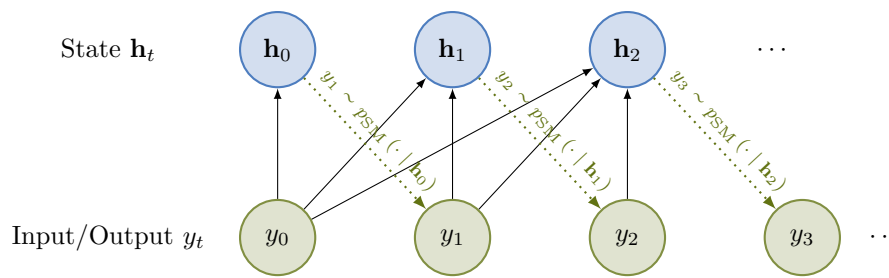

Having decided to keep around the encodings of all symbols in the string, let us think about parallelizing the process of encoding a string, i.e., computing $\encfunc{\str}$.
Remember the very general way in which RNNs build a representation of the string $\strlt$ by incrementally modifying $\hiddStatet$, which is illustrated in~\cref{fig:rnn-abstract}---this incrementality brings with it all the challenges of impossible parallelization.
The workaround for the issues with the sequential processing of RNNs is to process the context for each $\symt$ \emph{independently}, without relying on the encodings of the previous symbols, thus avoiding the sequential bottleneck.
Nevertheless, we still want the contextual encoding of $\symt$ to contain information about the rest of the string, i.e., the preceding context.
How can we achieve that without relying on recurrence?
Again, we grab onto the simplest solution: to compute the symbol encodings for each symbol $\symt$ ``from the ground up'' based only on the static symbol encodings $\inEmbedSymt$, which do not require any recurrence.
This is abstractly illustrated in~\cref{fig:transformer-abstract}, whereas~\cref{fig:transformer-abstract-generative} shows how this translates into the generative framework from \cref{def:locally-normalized-rep-based}, where individual symbols $\symt$ are both sequentially generated based on the encodings of the preceding context $\encfunc{\strlt}$ as well as used to build the representation of the context in the next time step $\encfunc{\strt}$.
Notice that instead of containing arcs denoting dependencies between the symbol encodings (``hidden states'') $\hiddStatet$, \cref{fig:transformer-abstract} and~\cref{fig:transformer-abstract-generative} contain arcs connecting each $\hiddStatet$ to all symbols $\sym_\idxj$ for all $\idxj \leq \tstep$.
Compare this to~\cref{fig:rnn-abstract}, where the arcs between $\hiddStatetminus$ and $\hiddStatet$ induce the temporal dependence, and carry the information about the symbols $\sym_\idxj$ for all $\idxj \leq \tstep$.

Clearly, this avoids the issues faced by RNNs due to their sequentiality.
However, it also introduces more work required to compute the individual contextual encodings from the static symbol representations.
The operations performed to do this by transformers, which are together known as the \defn{attention mechanism}, are introduced in the next subsection.
They represent possibly the most important component of the entire structure of the transformer---by ``attending'' to relevant preceding symbols when computing the symbol encodings (i.e., using them to compute $\encfunc{\strlt}$), the transformer can model long-range dependencies very effectively, and use them for appropriately modeling the distribution over the next word.

To recap, in this section, we informally motivated the new architecture, transformers, with the goals of
\begin{enumerate*}
    \item remembering the contextual encodings of all symbols explicitly and
    \item parallelizing the computation of the contextual symbol encodings.
\end{enumerate*}
The next subsection formally introduces the architecture, before we dive into their theoretical properties.

\subsection{A Formal Definition of Transformers} \label{sec:transformers-formal}
Having informally introduced the main two ideas behind the transformer architecture in the previous subsection, we now provide a formal definition of a transformer model, which we will then augment with more practical considerations in . \anej{add refs.}
\begin{definition}{Transformer network}{}
    A \defn{transformer network} $\transformernetwork$ is a tuple $\transformertuple$ where
    \begin{itemize}
        \item $\alphabet$ is the alphabet of input symbols,
        \item $\hiddDim$ is the dimension of $\transformernetwork$, and
        \item $\tfencfun$ is the transformer encoding function (cf. \cref{def:encoding-function}), which we define in more detail below (\cref{def:tf-hiddstate}).
    \end{itemize}
\end{definition}
From afar, the definition of a transformer network is therefore relatively simple; it is stated to make the transformer models fit well into our representation-based locally normalized language modeling framework (cf. \cref{def:locally-normalized-rep-based}).
The complexity of the models of course comes from the definition of the transformer encoding function $\tfencfun$, to which we devote the rest of the section.


Continuing in the framework of representation-based locally normalized language models, the hidden states of the transformer play an analogous role to those of RNNs, with the only difference being how they are computed.
\begin{definition}{Transformer hidden state}{tf-hiddstate}
    Let $\transformernetwork = \transformertuple$ be a transformer network.
    The hidden state $\hiddStatet \in \R^{\hiddDim}$ describes the state of $\transformernetwork$ after reading $\strlet$.
    It is defined with respect to the transformer encoding function $\tfencfun$ as follows:
    \begin{equation}
        \hiddStatet \defeq \tfencfun\left(\strlet\right)
    \end{equation}
\end{definition}
Crucially, as we will see shortly, the hidden state $\hiddStatet$ of the transformer does not have any dependence on the preceding hidden states themselves (although, as we will see, it is partially a function of the same \emph{inputs}).

As hinted above, with this, we can easily fit transformers into the representation-based locally normalized language modeling framework and define a sequence model based on the model.
\begin{definition}{Transformer sequence model}{transformer-plnsm}
    Let $\transformernetwork$  be a transformer network and  $\outMtx \in \R^{|\eosalphabet| \times \hiddDim}$ a symbol representation matrix.
    A $\hiddDim$-dimensional \defn{transformer sequence model} over the alphabet $\alphabet$ is a tuple $\tfseqmodel$ defining the sequence model of the form
    \begin{equation}
        \pLNSM\left(\eossymt \mid \strlt \right) \defeq \softmaxfunc{\embedMtx\,\hiddStatetminus}{\eossymt} = \softmaxfunc{\embedMtx\,\tfencfun(\strlt)}{\eossymt}
    \end{equation}
\end{definition}

Now that we have unified the transformer $\transformernetwork$ to the theoretical framework introduced so far in the course, we can jump in and look at the internal structure of the transformer encoder function, which is where the novelty of the transformer architecture comes from.

\subsubsection{The Attention Mechanism}
As we mentioned in the informal motivation, to avoid over-compressing information about sentences into a single vector, a transformer model retains the encodings (captured in the hidden states $\hiddStatet$) of \emph{all} possible prefixes of the string, which we can equivalently simply regard as encodings of individual symbols augmented with the information from the preceding string (see \cref{fig:transformer-abstract}).\footnote{From now on, we will talk about \defn{contextual symbol encodings}, which simply refers to the hidden states corresponding to individual symbols.}
However, rather than computing the encodings sequentially like an RNN, the encodings of the individual symbols are computed with the so-called attention mechanism.
\begin{definition}{Attention}{attention}
    Let $\tfscorefun: \R^{\hiddDim} \times \R^{\hiddDim} \to \R$ be a \defn{scoring function} and $\projfunc$ a projection function.
    Furthermore, let $\vq \in \R^{\hiddDim}$, $\mK_\tstep = \left(\vk_1^{\top}, \dots, \vk_\tstep^{\top}\right) \in \R^{\tstep \times \hiddDim}$ and $\mV_\tstep = \left(\vv_1^{\top}, \dots, \vv_\tstep^{\top}\right) \in \R^{\tstep \times \hiddDim}$.

    \defn{Attention} over $\mK_\tstep, \mV_\tstep$, also denoted by $\attn\left(\vq_\tstep, \mK_\tstep, \mV_\tstep\right)\colon \R^{\hiddDim} \times \R^{\tstep \times \hiddDim} \times \R^{\tstep \times \hiddDim} \to \R^{\hiddDim}$ is a function computing the vector $\va$ in the following two-step process:
    \begin{align}
        \vs_\tstep = \left(\evs_1, \dots, \evs_\tstep\right)       & \defeq \projfunc\left(\tfscorefun\left(\vq, \vk_1\right), \tfscorefun\left(\vq, \vk_2\right), \dots ,\tfscorefun\left(\vq, \vk_\tstep\right) \right) \label{eq:single_query_attn_score} \\
        \va_\tstep = \attn\left(\vq, \mK_\tstep, \mV_\tstep\right) & \defeq \evs_1\vv_1 + \evs_2\vv_2 + \dots + \evs_\tstep\vv_\tstep
    \end{align}
\end{definition}
$\vq$, $\mK$, and $\mV$ are commonly referred to as the \defn{query}, \defn{keys}, and \defn{values} of the attention mechanism, respectively.\anej{I would add a figure showing how the matrix of keys is compared to the query, then these scores are normalized and used to index the matrix of values.}
We talk about the parameters $\vq$, $\mK$, and $\mV$ completely abstractly for now.
However, to help you connect this to the representation-based language modeling framework, note that $\vq$ will later correspond to a query representing an individual symbol $\symt$, whereas $\mK$ and $\mV$ will contain the information from $\strlt$ used to compute $\hiddStatet$.

\paragraph{What the attention function computes.}
The scoring function $\tfscorefun$ is, abstractly, simply a parameter of the model which we can choose freely.
Intuitively, it should express the relevance of a particular key $\vk$ to the query $\vq$---the more the key is relevant to the query, the more ``attention'' the model will put to the \emph{value} associated to that key.
The projection function $\projfunc$ then transforms the computed scores ensuring that the transformed scores sum to $1$.\footnote{While the fact that the transformed scores sum to one invites their interpretation as probabilities, this is not their central role. Rather, the weights are simply used to define a convex combination of the values.}
The vector of transformed scores $\vs$ (\cref{eq:single_query_attn_score}) is then used to compute the result of the attention function---the vector $\va$.
$\va$ is a convex combination of the \emph{values} $\vv$ passed to the attention function.
Abstractly, therefore, the \emph{keys} contain the information used for ``indexing'' the \emph{values} with the specific \emph{query}.

\paragraph{The scoring function.}
As mentioned, the scoring function is supposed to measure the ``relevance'' of a particular value for a query $\vq$ through the values' key.
The most common choice for $\tfscorefun$ is the dot product between query and key, which is often \emph{scaled} by the square root of the vector dimensionality:
\begin{equation}
    \tfscorefun\left(\vq, \vk\right) = \frac{1}{\sqrt{\hiddDim}} \innerProd{\vq}{\vk}
\end{equation}

\paragraph{The projection function and soft and hard attention.}
The projection function used to transform the un-normalized attention scores is a crucial component of the transformer model.
By far the most commonly used projection function is again the softmax.
In this case, the attention function is referred to as soft attention.
\begin{definition}{Soft attention}{}
    The \defn{soft attention} is computed with the projection function $\projfunc = \softmax$.
\end{definition}
However, the softmax again makes the models difficult to analyze.
In our voyage to theoretically understand transformer-based language models, we will, therefore, again make specific (less frequently used) modeling choices, particularly in the case of the projection function.

Indeed, to be able to derive any interesting expressivity results (see \cref{sec:transformers-expressivity}), we jump to the other side of the spectrum and define hard attention.
Simply put, instead of spreading the attention across all values like softmax, hard attention puts all the mass on the element whose key maximizes the scoring function $\tfscorefun$.
One way to arrive at it from the definition of soft attention is by sending the temperature $\tempParam$ in the definition of the $\softmax$ function (cf. \cref{def:softmax}) to $0$.
Recall that that results in the output vector representing a uniform distribution over the elements that maximize the input vector.
This is known as \emph{averaging} hard attention.
\begin{definition}{Averaging hard attention}{}
    The \defn{averaging hard attention} is an attention mechanism with the projection function $\projfunc = \hardmaxAvg$, where $\hardmaxAvg$ is defined as:
    \begin{equation}
        \hardmaxAvg(\vx)_d \defeq \begin{cases}
            \frac{1}{r} \text{ if } d\in \argmax(\vx) \\
            0 \text{ otherwise}
        \end{cases}
        \text{, for } d = 1, \dots, \hiddDim
    \end{equation}
    where $\vx \in \R^\hiddDim$ and $r = |\argmax(x)|$ is the cardinality of the argmax set over $\vx$.
\end{definition}
Interestingly, there is another form of hard attention that results in a model with a different expressive capacity: the unique hard attention.
The difference lies exactly in how it handles \emph{ties} in the elements which maximize the scoring function.
Unique hard attention chooses only \emph{a single} element of those that maximize the score: it can be chosen randomly or deterministically (e.g., always the first one).
\begin{definition}{Unique hard attention}{}
    The \defn{unique hard attention} is an attention mechanism with the projection function $\projfunc = \hardmaxUni$, where $\hardmaxUni$ is defined as follows.
    For $\vx \in \R^\hiddDim$, sample $\hat{d} \sim \textnormal{Unif}\left(\argmax(\vx)\right)$ or choose some $\hat{d} \in \argmax(\vx)$ deterministically.
    Then
    \begin{equation}
        \hardmaxUni(\vx)_d \defeq \begin{cases}
            1 \text{ if } d = \hat{d} \\
            0 \text{ otherwise}
        \end{cases}
        \text{, for } d = 1, \dots, \hiddDim.
    \end{equation}
\end{definition}
While the difference between unique and averaging hard attention might seem subtle and marginal, it actually results in a large difference in the expressivity of transformer-based language models as we discuss in \cref{sec:transformers-expressivity}.
While we will investigate this in a lot more detail there, we just mention that the intuition behind the expressive difference is relatively straightforward: while the \emph{keys} maximizing the un-normalized scores might be the same (even though they necessarily don't have to be if $\tfscorefun$ is not injective), the \emph{values} (whose content is decoupled from the keys) that those keys index might not be---and in some cases, all those different values might be relevant for the task at hand.
Unique hard attention always allows us to only ``lookup'' a \emph{single} value associated with those keys, no matter how ``different'' and relevant all of those are.
It also does not allow the attention mechanism to sum over (``summarize'') across all the elements that maximize attention.
This is a very limiting characteristic, as many of the expressivity results that we will see later rely on summing over all the elements that maximize the attention scores.

\subsubsection{Transformer Blocks}
We have, so far, described the ``low-level'' details of how the attention function is computed.
We now combine the computations performed into larger blocks, showing how they are used to compute the string-augmented encodings of the individual symbols.
In particular, we have to connect the concepts of queries, keys, and values to the symbols and their (initial, static) encodings.
Intuitively, this is done by transforming the static encodings of those symbols using specific functions implemented through the attention mechanism and using the transformed encodings as queries, keys, and values, as we describe below.

We first abstract the attention mechanism from \cref{def:attention} a bit.
With this, we will, in a few steps, arrive at exactly how the hidden states $\hiddStatet$ or the contextual encodings are computed in a transformer.
At the core of this computation lies a repeated application of the same sequence of operations, which, as we will see, augment the ``current version'' of the contextual encodings with the current information from the preceding information.
We call a single sequence of operations a transformer layer.
\begin{definition}{Transformer layer}{transformer-layer}
    Let $\qTransf$, $\kTransf$, $\vTransf$, and $\oTransf$ be parametrized functions from $\R^{\hiddDim}$ to $\R^{\hiddDim}$.

    A \defn{transformer layer} is a function $\tflayer\colon \R^{\finaltstep \times \hiddDim} \to \R^{\finaltstep \times \hiddDim} $ that takes as input sequence of vectors $\tflayerinputmat = \tflayerinputseq$ and returns $\tflayeroutputmat = \tflayeroutputseq \in \R^{\finaltstep \times \hiddDim}$ according to the following steps:
    \begin{equation} \label{eq:attn-block-1}
        \va_\tstep = \underbrace{\attn\left(\qTransf(\tflayerinputsy_\tstep), \kTransf(\tflayerinputmat_\tstep), \vTransf(\tflayerinputmat_\tstep)\right)}_{\text{\cref{def:attention}}} + \tflayerinputsy_\tstep
    \end{equation}
    \begin{equation} \label{eq:attn-block-2}
        \tflayeroutputsy_\tstep = \oTransf(\va_\tstep) + \va_\tstep
    \end{equation}
    for $\tstep = 1, \ldots, \finaltstep$, so that $\tflayer(\tflayerinputmat) \defeq \tflayeroutputmat = \tflayeroutputseq \in \R^{\finaltstep \times \hiddDim}$.
\end{definition}
While we defined the transformer layer on a general matrix (with $\finaltstep$ columns), note that these $\finaltstep$ vectors will refer to the (current) symbol encodings of the symbols in the string up to the $\finaltstep\textsuperscript{th}$ symbol, i.e., $\str_{\leq \finaltstep}$.

What do these quantities correspond to?
\cref{eq:attn-block-1,eq:attn-block-2} outline a two-step process of computing the outputs of a \emph{single} transformer layer: $\tflayerinputmat = \tflayerinputseq$ represents the input to the layer, which $\tflayer$ transforms into the output sequence $\tflayeroutputmat = \tflayeroutputseq$.
Before being fed into the attention mechanism, the inputs $\tflayerinputmat$ are first \emph{transformed} into the quantities required by the attention mechanism: the query $\vq_\tstep$ (a single one for each $\vx_\tstep$), the matrix of keys $\mK_\tstep$, and the matrix of values $\mV_\tstep$---all of these are, therefore, \emph{transformations} of the input sequence of vectors.
The transformations $\qTransf$, $\kTransf$, and $\vTransf$ determine how these inputs are transformed into the (interpretable) quantities required by the attention mechanism.

The individual $\va_\tstep$ represent the ``intermediate'' results of the computation---the results of applying the actual attention mechanism (with a slight modification) from \cref{def:attention} onto the produced values of the query, the keys, and the values.

The modification mentioned refers to the addition of the inputs $\vx_\tstep$ to the output of the attention mechanism in \cref{eq:attn-block-1}.
This mechanism is known as adding \defn{residual connections} to the model.
First introduced by \citet{He2016Residual} in the context of deep convolutional neural network-based architectures, residual connections are now a common feature in many state-of-the-art deep learning architectures.
Note that their use is mostly motivated by empirically better performance---this is often attributed to the fact that, intuitively, residual connections allow gradients (i.e., learning signals) to flow through the network through a more direct route rather than all layers that can ``squish'' the signal similarly to the Elman RNN case (in that sense, they help mitigate the vanishing gradient issue).
However, as we will see later in our analysis, \anej{add ref} residual connections will also play an important role in determining the theoretical properties of transformers, particularly their computational expressive power.
The same mechanism is applied in the second step of the transformer layer, where the intermediate results $\va_\tstep$ are transformed by the output transformation $\oTransf$ into the final outputs of the layer.

In the simplest case, you can imagine the inputs $\tflayerinputmat$ to be the initial static embeddings of the symbols.
The application of the transformer layer, in this case, therefore, ``selectively'' (determined by the attention mechanism) augments the static embeddings with the information from the preceding context.
However, as we will see shortly, a transformer model will apply \emph{multiple} transformer blocks to the input sequence and thus transform it in multiple steps, analogously to how layers are composed in a regular feed-forward neural network.
In that sense, the inputs to the transformer blocks will refer to general intermediate representations produced from the initial static embeddings after some number of applications of transformer layers.

Lastly, let us consider how the current symbol representations $\mX$ are transformed into the queries, keys, and values using $\qTransf$, $\kTransf$, and $\vTransf$?
The original formulation \citep{Vaswani2017} and all standard implementations of the transformer architecture use one of the simplest possible mappings: a linear transformation implemented by matrix multiplication.
On the other hand, the final output of the transformer, computed with output mapping $\oTransf$, is usually implemented by a multi-layer perceptron.

The transformer layer puts the attention mechanism into a functional block that describes how a sequence of current symbol representations is transformed in a single step into a sequence of representations augmented with the information in the current set of values.
However, we are not done abstracting yet!
As mentioned, this process can be applied arbitrarily many times, resulting in ``deep'' encodings of individual symbols, which contain information from the preceding symbols in the string computed in a composite way.
This also answers the question of how the augmented symbol representations used in the language modeling formulation are computed from the initial symbol representations: they are the result of multiple applications of the transformer layer to the matrix of initial symbol representations.
That is: multiple transformer layer layers are stacked on top of one another so that the output of one layer becomes the input of the next.

We now have all the building blocks to define the full transformer architecture, which computes the encodings of the string prefixes (and thus the hidden states) in \cref{def:transformer-plnsm}.
\begin{definition}{Transformer}{transformer}
    For $\tfnumlayer \in \N$, we define a $\tfnumlayer$-layer \defn{transformer} model as a $\hiddDim$-dimensional transformer sequence model over an alphabet $\alphabet$ where the hidden state $\hiddStatet \defeq \tfencfun\left(\sym_1 \ldots \sym_\tstep\right) = \tfencfun\left(\str\right)$ is computed as follows.
    \begin{align}
         & \tflayerinputmat^1 \defeq \left(\inEmbeddingFun{\symzero}, \inEmbeddingFun{\symone}, \dots, \inEmbeddingFun{\symt}\right)        \\
         & \tflayeroutputmat^\tflayeridx = \tflayer_\tflayeridx(\tflayerinputmat^\tflayeridx) \text{ for } 1 \leq \tflayeridx < \tfnumlayer \\
         & \tflayerinputmat^{\tflayeridx+1}=\tflayeroutputmat^{\tflayeridx} \text{ for } 1 \leq \tflayeridx < \tfnumlayer                   \\
         & \hiddStatet = \fTransf(\tflayeroutputsy_\tstep^\tfnumlayer)
    \end{align}
    $\tflayer_\tflayeridx$ for $\tflayeridx = 1, \ldots, \tfnumlayer$ represent $\tfnumlayer$ different transformer layers with decoupled parameter (cf. \cref{def:transformer-layer}).
    $\fTransf\colon \R^{\hiddDim} \to \R^{\hiddDim}$ is a transformation function applied to the contextual encoding of the last symbol in the last ($\tfnumlayer\textsuperscript{th}$) layer and $\inEmbedding\colon \alphabet \to \R^{\hiddDim}$ is a symbol representation function computing the initial representations of the symbols passed to the first layer of the transformer.\footnote{The symbol representation function $\inEmbedding$ is often also implemented as a linear transformation of the one-hot representations (cf. \cref{def:one-hot}) of symbols, i.e., it is simply a table-lookup.}
\end{definition}
With this, the transformer model now fully specifies how to compute the representations required for the representation-based locally normalized sequence models from \cref{def:transformer-plnsm}---the representation function $\tfencfun$ is the composition of $\tfnumlayer$ transformer layers applied to the sequence of static encodings, followed by a final transformation $\fTransf$.

\subsubsection{Making Attention Work Fast Over Entire Strings}
\begin{figure}
    \centering
    \begin{tikzpicture}[x=1.5cm, y=1.5cm,>=latex]

        \foreach \i in {0,...,3}
        \node[circle, draw=ETHGreen!80!white, thick, fill=ETHGreen!20, inner sep=0pt, minimum size=10mm] (y\i) at (1.25*\i, 0) {$\sym_\i$};

        \node[draw=none] (p3) at (3.75, 3) {$\pLNSMFun{\eossym_4}{\hiddState_3}$};
        \node[draw=none] (pt) at (7.5, 3) {$\pLNSMFun{\eossym_{\tstep + 1}}{\hiddState_\tstep}$};
        \node[circle, draw=ETHBlue!80!white, thick, fill=ETHBlue!20, inner sep=0pt, minimum size=10mm] (H3) at (3.75, 2) {$\hiddState_3$};
        \node[circle, draw=ETHBlue!80!white, thick, fill=ETHBlue!20, inner sep=0pt, minimum size=10mm] (Ht) at (7.5, 2) {$\hiddState_\tstep$};
        \node[circle, draw=ETHGreen!80!white, thick, fill=ETHGreen!20, inner sep=0pt, minimum size=10mm] (ytminus) at (6.25, 0) {$\symtminus$};
        \node[circle, draw=ETHGreen!80!white, thick, fill=ETHGreen!20, inner sep=0pt, minimum size=10mm] (yt) at (7.5, 0) {$\symt$};

        \foreach \j in {0,...,3} {
                \draw[->, dashed] (y\j) -- (H3);
                \draw[->, dashed] (y\j) -- (Ht);
            }
        \draw[->, dashed] (ytminus) -- (Ht);
        \draw[->, dashed] (yt) -- (Ht);

        \draw[->] (H3) -- (p3);
        \draw[->] (Ht) -- (pt);

        \node[text width=3cm] at (-1.5, 3) {Locally-normalized distribution};
        \node[text width=3cm] at (-1.5, 2) {Encoding $\tfencfun$};
        \node[text width=3cm] at (-1.5, 1) {$\tfnumlayer$ applications of the transformer layer};
        \node[text width=3cm] at (-1.5, 0) {Input $\str$};
        \node[circle, draw=none, minimum size=10mm] (yend) at (5, 0) {$\cdots$};
        \node[circle, draw=none, minimum size=10mm] (Hend) at (5, 2) {$\cdots$};

    \end{tikzpicture}
    \caption{$\tfencfun\left(\strlet\right)$ is a function of the symbols $\sym_1, \ldots, \symt$ computed with multiple applications of the transformer block.
        Here, the dashed lines illustrate the dependencies of the outputs $\hiddStatet$ on the initial static encoding of the symbols $\sym$ denoted by green nodes.
        Na{\"i}vely, $\hiddState_3$ and $\hiddStatet$ could be computed by independently applying the attention mechanism from \cref{def:attention}.
        However, as we describe in the text, while the applications of the attention mechanism do not \emph{share} computations, they can be written concisely together.
    }
    \label{fig:transformer-single-query}
\end{figure}

Notice that, in the formulations so far, we always presented computations of the attention mechanism for individual queries $\vq_\tstep$.
This corresponds to the computation of the new version of the representation of a \emph{single} symbol in the string, with the keys and values representing the preceding symbols (including the symbol itself).
This is illustrated in \cref{fig:transformer-single-query}.
This could of course be applied $|\str|$-times to compute the representations of all $|\str|$ symbols in a single transformer block.
However, this would unnecessarily re-compute the keys and values of the symbols multiple times---and, as we motivated at the beginning, speedups were one of the main reasons to talk about transformers in the first place.
We can now show how the attention mechanism can be conveniently applied to entire strings at once.
Specifically, we focus on the case where the attention scoring function $\tfscorefun$ is implemented as a dot-product.\footnote{For conciseness, we will ignore the scaling factor, which could easily be added.}

\paragraph{What does the attention mechanism from \cref{def:attention} do in this case?}
Given a query $\vq_\tstep$ and a matrix of key values $\mK = \left(\vk_1^{\top}, \dots, \vk_\tstep^{\top}\right) \in \R^{\tstep \times \hiddDim}$, the scoring function simply computes\footnote{We switch notation from $\innerProd{\vq_\tstep}{\vk_\idxj}$ to $\vq_\tstep^\top\vk_\idxj$ to make the connection to matrix multiplication later clearer.}
\begin{equation*}
    \evu_\idxj=\tfscorefun\left(\vq_\tstep, \vk_\idxj\right) = \vq_\tstep^\top\vk_\idxj.
\end{equation*}
Notice that, in this case, the vector $\vu = \left(\evu_1, \dots, \evu_\tstep\right)$ of \emph{unnormalized} attention weights can simply be computed as a single matrix-vector product
\begin{equation*}
    \vu = \vq_\tstep^\top \mK^\top.
\end{equation*}
Furthermore, with this, attention can be easily extended to consider many queries in parallel by stacking multiple queries into a matrix $\mQ \defeq \left(\vq_1^{\top}, \vq_2^{\top}, \dots, \vq_\tstep^{\top}\right)$, as we detail now.\footnote{
    Note that, for easier presentation, we make a slight departure from the original definition of the attention mechanism, where the result of the attention mechanism for query $\tstep$ only depended on the keys and values $\idxj \leq \tstep$.
    For the rest of the paragraph, we assume that a query $\vq_\idxi$ with $\idxi < \tstep$ can consider keys and values $\vk_\idxj$ and $\vv_\idxj$ with $\idxj > \idxi$, which, in the interpretation of attention applied to strings, would mean that the symbols can ``look ahead'' in the string to their right.
    This will be addressed shortly with \emph{masking}.}
Consider now the product
\begin{equation*}
    \mU = \mQ \mK^\top.
\end{equation*}
Each entry of the resulting matrix $\emU_{\idxi \idxj}$ is exactly the dot-product between the query $\vq_\idxi$ and the key $\vk_\idxj$!
The \emph{rows} of $\mU$ then contain the unnormalized score vectors $\vu_\idxi$ from the definition of the attention mechanism.
This means that if we now apply the normalization function $\projfunc$ \emph{row-wise} (such that the sums of the elements in each row equal 1), we end up with exactly the required normalized scores required for combining the values from the value matrix.
With some abuse of notation, we will simply write that as
\begin{equation}
    \mS \defeq \left(\vs_1^\top, \ldots, \vs_\tstep^\top\right) \defeq \projfunc\left(\mU\right) = \projfunc\left(\mQ \mK^\top\right).
\end{equation}
The rows of $\projfunc\left(\mA\right)$, therefore, represent the normalized attention weights.
This brings us to the final step of the matrix-multiplication-based attention mechanism: Combining the values based on the computed attention weights.
Again, this can be performed by a single matrix multiplication.
Notice that the \emph{value} vectors are \emph{the same} for all queries---they are simply combined with different (attention) weights based on the query.
Right-multiplying the \emph{transposed} values matrix $\mV = \left(\vv_1^{\top}, \dots, \vv_\tstep^{\top}\right)$ with $\mS$, therefore, perform the convex combination of the value vector $\vv_1^{\top}, \dots, \vv_\tstep^{\top}$ such that
\begin{equation}
    \va_\idxi = \vs_\idxi \mV^\top = \mS_{\idxi, \colon} \mV^\top
\end{equation}
and thus
\begin{equation}
    \mA \defeq \left(\va_1, \ldots, \va_\tstep\right) = \mS \mV^\top.
\end{equation}
Altogether, this means that, given a sequence of (contextual) symbol encodings $\mX$, we can compute the attention values (i.e., the output of the attention mechanism) of \emph{all} queries (i.e., for all string in the string) with a single matrix multiplication, as long as the scoring function is the (scaled) dot-product.
We refer to this version of attention as an attention block, which, intuitively, simply replaces the \emph{element-wise} definition of the attention mechanism from \cref{def:attention} with a more efficient (and concise) definition through matrix multiplications.\footnote{Again, with the caveat that the attention weights are not confined to the preceding symbols but to all symbols in the string.}
\begin{definition}{Attention Block}{attention-block}
    Let $\qTransf$, $\kTransf$, and $\vTransf$ be parametrized functions from $\R^{\tfslen \times \hiddDim}$ to $\R^{\tfslen \times \hiddDim}$ and $\mX \in \R^{\tfslen \times \hiddDim}$ the matrix of input encodings.
    An \defn{attention block} is the function $\attnBlock\colon \R^{\tfslen \times \hiddDim} \to \R^{\tfslen \times \hiddDim}$ defined as
    \begin{equation}\label{eq:attention_block}
        \attnBlock(\mX) = \projfunc\left(\qTransf\left(\mX\right)\kTransf\left(\mX\right)^{\top}\right)\vTransf\left(\mX\right)
    \end{equation}
    Further, we define the \defn{attention matrix} as the square matrix $\mU \defeq \qTransf(\mX)\kTransf(\mX)^{\top} \in \R^{\tfslen \times \tfslen}$.
\end{definition}

As mentioned, the functions $\qTransf(\cdot)$, $\kTransf(\cdot)$, and $\vTransf(\cdot)$ are usually implemented as a linear transformation via matrix multiplication using weight matrices $\qMtx$, $\kMtx$, and $\vMtx$.
This means that the query matrix $\mQ$ can be computed as $\mQ = \mX\qMtx$, where $\qMtx \in \R^{\hiddDim \times \hiddDim}$ is a matrix of learnable parameters.
Since the attention block uses the same input matrix $\mX$ to encode queries, keys, and values, it is usually called \defn{self-attention}.

\begin{figure}
    \centering
    \begin{tikzpicture}[x=1.5cm, y=1.5cm,>=latex]


        \node[circle, draw=ETHBlue!80!white, thick, fill=ETHBlue!20, inner sep=0pt, minimum size=10mm] (Ht) at (3.75, 2) {$\hiddState_\tstep$};

        \node[circle, draw=ETHGreen!80!white, thick, fill=ETHGreen!20, inner sep=0pt, minimum size=10mm] (y0) at (0, 0) {$\sym_1$};
        \node[circle, draw=ETHGreen!80!white, thick, fill=ETHGreen!20, inner sep=0pt, minimum size=10mm] (y1) at (1.25, 0) {$\sym_2$};
        \node[circle, draw=none, minimum size=10mm] (ymid) at (2.5, 0) {$\cdots$};
        \node[circle, draw=ETHGreen!80!white, thick, fill=ETHGreen!20, inner sep=0pt, minimum size=10mm] (yj) at (3.75, 0) {$\sym_\tstep$};
        \node[circle, draw=none, minimum size=10mm] (yend) at (5, 0) {$\cdots$};
        \node[circle, draw=ETHGreen!80!white, thick, fill=ETHGreen!20, inner sep=0pt, minimum size=10mm] (ytminus) at (6.25, 0) {$\sym_{\strlen - 1}$};
        \node[circle, draw=ETHGreen!80!white, thick, fill=ETHGreen!20, inner sep=0pt, minimum size=10mm] (yt) at (7.5, 0) {$\sym_\strlen$};

        \node[draw=none] (k0) at (0, -0.75) {$\vk_1, \vv_1$};
        \node[draw=none] (k1) at (1.25, -0.75) {$\vk_2, \vv_2$};
        \node[draw=none] (kj) at (3.75, -0.75) {$\vk_\tstep, \vv_\tstep$};
        \node[draw=none] (ktminus) at (6.25, -0.75) {$\vk_{\strlen - 1}, \vv_{\strlen - 1}$};
        \node[draw=none] (kt) at (7.5, -0.75) {$\vk_\strlen, \vv_\strlen$};

        \foreach \j in {0,...,1} {
                \draw[->, dashed, ETHGreen] (y\j) -- (Ht);
            }
        \draw[->, dashed, ETHGreen] (yj) -- (Ht);
        \draw[->, dotted, ETHRed!80] (ytminus) -- (Ht);
        \draw[->, dotted, ETHRed!80] (yt) -- (Ht);

        \node[text width=3cm] at (-1.5, 2) {Encoding $\tfencfun$};
        \node[text width=3cm] at (-1.5, 1) {$\tfnumlayer$ applications of the transformer layer};
        \node[text width=3cm] at (-1.5, 0) {Input $\str$};

    \end{tikzpicture}
    \caption{
        In the context of language modeling, the attention mechanism is allowed to consider the symbols $\sym_\idxj$ and their keys/values for $\idxj \leq \tstep$ (the \textcolor{ETHGreen}{green dashed lines}) when computing the contextual encoding of the symbol $\symt$.
        In \cref{def:attention} this is enforced by the definition of the matrices $\mK_\tstep$ and $\mV_\tstep$.
        However, in the attention block formulation of the attention mechanism from \cref{def:attention-block}, since the matrices $\mK$ and $\mV$ contain the values corresponding to the entire string $\str$, the query $\vq_\tstep$ could, in principle, index into the values corresponding to the symbols $\sym_{\tstep + 1}, \ldots, \sym_\strlen$ (the \textcolor{ETHRed}{red dotted lines}).
        Masking prevents that by enforcing the attention weights $\eva_{\tstep + 1}, \ldots, \eva_\strlen$ to be $0$.
        In this sense, it removes the \textcolor{ETHRed}{red dotted lines}.
    }
    \label{fig:transformer-requires-masking}
\end{figure}
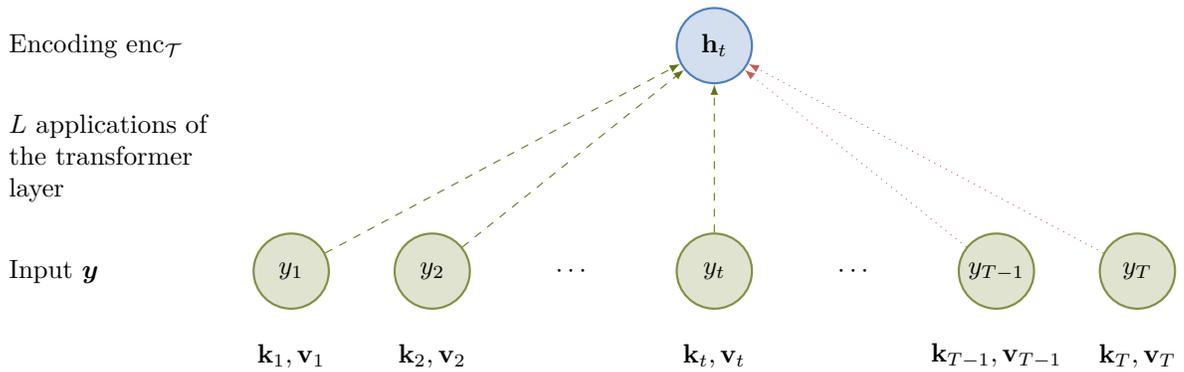

\paragraph{Confining attention to preceding symbols in the string.}
We now address the departure from the original definition of the attention mechanism in which the query $\vq_\tstep$ was only allowed to consider the keys and values $\vk_\idxj$ and $\vv_\idxj$ with $\idxj \leq \tstep$.
Notice that, in general, the version of attention of~\cref{eq:attention_block} allows each symbol to attend to \emph{any} symbol in the string, even those in later positions in the string.
Note that there is nothing inherently \emph{wrong} with that---the contextual symbol encodings could, in principle, depend on the information from the \emph{entire} string.
In fact, this is very common in the so-called \emph{masked language modeling}, which, importantly, despite its name, does not define language models in our sense of the word.
A very commonly used family of masked models is the BERT family of models.\footnote{In a very unfortunate, but also understandable, turn of events, we mention two completely different notions of masking in a single paragraph. Importantly, the masking in \emph{masked language models} (which, again, are not language models in our strict definition) has nothing to do with the ``causal'' masking relevant for autoregressive language modeling which we introduce in this section.}
However, in the case of locally normalized language models, ``looking ahead'' obviously violates the autoregressive structure of language modeling, i.e., violates our assumption that the context at time $\tstep$ forms $\strlt$.
This is illustrated in \cref{fig:transformer-requires-masking}.
To recover the autoregressive nature of the language model, we, therefore, posthoc modify~\cref{eq:attention_block} to allow each symbol to attend only to \emph{itself} and to \emph{preceding} symbols, while still being able to implement it using matrix multiplication.
We do that by adding a \emph{mask} to zero out the unwanted elements of $\mU$.
\begin{definition}{Masked Attention Block}{}
    Let $\qTransf(\cdot)$, $\kTransf(\cdot)$, and $\vTransf(\cdot)$ be parametrized functions from $\R^{\tfslen\times\hiddDim}$ to $\R^{\tfslen\times\hiddDim}$.
    A \defn{masked attention block} is a function $\attnBlock(\mX, \mM): \R^{\tfslen\times\hiddDim} \times \R^{\tfslen\times\hiddDim} \to \R^{\tfslen\times\hiddDim}$ defined as
    \begin{equation}\label{eq:masked_attention_block}
        \attnBlock(\mX, \mM) = \softmax(\qTransf(\mX)\kTransf(\mX)^{\top}\odot \mM)\vTransf(\mX)
    \end{equation}
    where $\odot$ is the element-wise product between matrices, and $\mM \in \R^{\ell \times \ell}$, the \defn{masking matrix}, is constructed as follows.
    \begin{equation}
        \emM_{\idxi,\idxj} = \begin{cases}
            1 \text{ if } \idxi \leq \idxj \\
            -\infty \text{ otherwise}
        \end{cases} \text{ for } 0 \leq \idxi, \idxj < \tfslen
    \end{equation}
\end{definition}
This implements a very easy ``fix'' to the looking-ahead problem by simply putting the normalized attention scores of the ``illegal'' elements to $0$.
In general, the exact value of the elements $\emM_{\idxi,\idxj}$ with $\idxi > \idxj$ of course depends on the projection function $\projfunc$---for simplicity, we only define $\mM$ for the case where $\projfunc = \softmax$.

\subsubsection{Bits and Bobs of the Transformer Architecture: Positional Encodings, Multiple Heads, and Layer Normalization}
Let us now take a step back and consider what the transformer model introduced so far does abstractly.
A transformer takes as input as string $\str \in \kleene{\alphabet}$, computes the initial symbol embeddings $\mX^1 = \mX$, and transforms those through a sequence of $\tfnumlayer$ applications of the transformer layer (cf. \cref{def:transformer}).
This results in the final augmented (contextual) symbol representations $\hiddStatet = \tfencfun\left(\strlet\right)$, which are then used to compute the conditional probabilities in the representation-based locally normalized language model defined by the transformer (cf. \cref{def:transformer-plnsm}), as illustrated on top of \cref{fig:transformer-single-query}.
In this subsection, which will finish off our formal definition of the architecture, we introduce the last three components often connected closely to the transformer model: Symbol positional encodings, multi-head attention, and layer normalization.

\paragraph{Adding positional information into the transformer architecture.}
There is an important omission we still have not addressed when talking about transformers: How does the model incorporate any notion of \emph{word order} into the contextual representations of symbols or the encodings of the context $\hiddStatet$?
The motivation is very clear: The meaning of a sentence depends on the word order.
The meaning of \textexample{A dog bit a man.} is not the same as \textexample{A man bit a dog.}.
This is one of the reasons why simple ``sentence encoding'' functions such as bag-of-words, which simply represent sentences with the number of individual words they contain, do not work well.
A careful reader might have noticed that at no point in our discussion about transformers and the attention mechanism did we say anything about the positions of the words.
Importantly, we did not talk about word positions in the case of RNNs either.
However, the sequential and incremental processing nature of RNNs makes it easy to ``manually'' keep track of the position of the current symbol of the string $\symt$, to the extent that the RNN variant is capable of ``counting'' (cf. \cref{sec:rnn-expressiveness}).
However, all operations composing the transformer model are \emph{position-agnostic}: The convex combination of the value vectors $\mV$ will be the same, no matter the permutation of the vectors (if we, of course, accordingly permute the keys).
The keys also cannot contain any positional information, since they are computed from position-agnostic static embeddings and a transformation function $\kTransf$ which does not depend on the position.

All that is to say that, to be able to take into account word order in a transformer, we have to \emph{explicitly provide} the positional information to the model.
The simplest way to do this is to \emph{augment} the static symbol encodings in the first transformer layer with \emph{positional encodings} in the form of vectors which can be added to or concatenated to the static encodings of symbols \citep{Vaswani2017}.\footnote{For simplicity, we assume the positional encodings are added to the static ones; notice that by dividing the $\hiddDim$-dimensional vectors into two components, one responsible for the static encodings and one for the positional ones (where the positional encoding component is zeroed out in the static encoding and vice-versa), one can easily implement ``concatenation'' of the two representations using only addition.}
\begin{definition}{Positional encoding}{positional-encodings}
    A \defn{positional encoding}\index{positional encoding} is a function $\posEnc\colon \N \to \R^\hiddDim$.
\end{definition}
This is a very simple definition: A positional encoding simply assigns a position in a string a vector representation.
A trivial example would be $\posEncFun{\tstep} = \left(\tstep\right)$.
This allows us to define a position-augmented symbol representation function.
\begin{definition}{Position-augmented representation function}{}
    Let $\inEmbedding\colon \eosalphabet \to \R^\hiddDim$ be a symbol representation function and $\posEnc\colon \N \to \R^\hiddDim$ a positional encoding.
    A \defn{position-augmented} representation function of a symbol $\symt$ in a string $\str$ is the representation function $\posInEmbedding\colon \eosalphabet \to \R^\hiddDim$ defined as
    \begin{equation}
        \posInEmbeddingFun{\symt} \defeq \inEmbeddingFun{\symt} + \posEncFun{\tstep}.
    \end{equation}
\end{definition}
To make the positional information available to the transformer model, we now simply pass the position-augmented ``static'' symbol encodings $\mX_\texttt{pos}$ to the model instead of the original ones $\mX$.
Apart from that, the transformer model can remain unaltered, and function simply as defined above, taking into account the positional information now included in its inputs.
Importantly, the intuitive notion of the importance of positional encodings for understanding natural language also transfers to the computational power of the model: Transformers as introduced in this section \emph{without} positional information are strictly less powerful than those with positional information \citep{perez_2021_turingtransformers}.
Again, this intuitively makes sense: Without positional information, a transformer model could not even recognize the simple (unweighted) regular language
\begin{equation*}
    \lang = \set{\syma \symb^n \mid n \in \N}
\end{equation*}
since it would have no way of knowing, provided that $\syma$ is in a given string $\str$, whether it appears in the first position or in any other position in the string.

\paragraph{Multiple heads.}
Importantly, the transformer introduced so far computes a single set of contextual representations---one for every input symbol (at every layer of the transformer).
However, we can easily extend the model to compute \emph{multiple} contextual representations for each symbol.
This is done using the so-called \defn{multi-head attention}, where a single attention block is called an \defn{attention head}.
This increases the representation space of the individual symbols and thus enables the model to capture more information about the symbols and the sentence.
The interpretation of computing \emph{multiple} representations (one for each head) independently also invites the interpretations that each of the heads ``focuses'' on a separate aspect of the text.
To be able to use the outputs of multi-head attention as inputs to the next block again, the outputs of the different attention heads are then concatenated and then projected down to the output size of a single attention block using an additional transformation.
\begin{definition}{Multi-Head Attention Block}{}
    Let $\tfheadnum \in \N$ be the number of attention heads, $\qTransf_h(\cdot)$, $\kTransf_h(\cdot)$, and $\vTransf_h(\cdot)$ be parametrized functions from $\R^{\tfslen\times\hiddDim}$ to $\R^{\tfslen\times\hiddDim}$ for $0 \leq h \leq \tfheadnum$, and $\tfheadCombine\colon \R^{\tfslen\cdot\tfheadnum\times\hiddDim} \to \R^{\tfslen\times\hiddDim}$ be a parametrized function.

    A \defn{multi-head attention block} is a function $\attnBlockMH(\mX): \R^{\tfslen\times\hiddDim} \to \R^{\tfslen\times\hiddDim}$ defined as
    \begin{equation}\label{eq:mh_attention_block}
        \attnBlockMH(\mX) = \tfheadCombine(\operatorname{concat}_{0 \leq h < \tfheadnum}\left(\softmax\left(\qTransf_h\left(\mX\right)\kTransf_h\left(\mX\right)^{\top})\vTransf_h\left(\mX\right)\right)\right)
    \end{equation}
\end{definition}
While multi-head attention is mostly motivated by empirically better performance (and the intuitive motivation of being able to separately focus on different notions of similarity), it will have some implications on the computational power of the model as well.
As we will see shortly, having multiple heads makes it very easy to reason about how a transformer model can simulate an \ngram{} model.\footnote{This does not, however, mean that multiple heads are \emph{required} for recognizing \ngram{} models. As we will see, under some caveats, single-head transformers are Turing complete.}

\paragraph{Layer normalization.}
As a final component of a transformer, we mention layer normalization.
Layer normalization, similar to the use of residual connections, represents a common ``trick'' in the deep learning space for ensuring more stable and reliable gradient-based learning---as such, it is not limited to transformers.
Formally, we can define layer normalization as follows \citep{ba2016layer}.
\begin{definition}{Layer normalization}{}
    Let $\vx, \vgamma, \vbeta \in \R^D$, and $\epsilon > 0$.
    The \defn{layer normalization} function $\layerNorm\colon \R^\hiddDim \to \R^\hiddDim$ is defined as
    \begin{equation}
        \layerNormFun{\vx;\vgamma, \vbeta} \defeq \frac{\vx - \mean{\vx}}{\sqrt{\var{\vx} + \epsilon}} \odot \vgamma + \vbeta,
    \end{equation}
    where $\mean{\vx}$ refers to the mean of the vector $\vx$ (and is subtracted from all elements of $\vx$ in the formulation above) and $\var{\vx}$ refers to the variance of elements of $\vx$.
    $\epsilon$ is added in the denominator to ensure stability if $\var{\vx} \ll 1$.
\end{definition}
Intuitively, the application of the layer normalization function ensures that the mean of the vector $\vx$ is (approximately) $\vbeta$ and its variance is controlled by $\vgamma$ (after being standardized by dividing by the standard deviation of $\vx$).
Most commonly, we simply set $\vgamma = \one \in \R^\hiddDim$ and $\vbeta = \zero \in \R^\hiddDim$.

Layer normalization is most commonly applied to the \emph{output of the transformer layer} (on every layer), i.e., to $\vz_\idxi$ in \cref{eq:attn-block-2}.
The full output of the transformer layer is therefore computed as
\begin{equation}
    \vz_\idxi \defeq \layerNormFun{\oTransf\left(\va_\idxi\right) + \va_\idxi;\vgamma, \vbeta}.
\end{equation}
Interestingly, although we mentioned that layer normalization is mostly motivated by the stability it brings to training and with it better performance, it does, just like multi-headed attention, seem to contribute to the computational expressivity of transformer models.
As we will see in \cref{sec:transformers-regular-languages}, layer normalization allows for a simple fix that solves one of the best known formal limitations of the transformer architecture (again, under some assumptions) \citep{hahn-2020-theoretical,chiang-cholak-2022-overcoming}.

\subsubsection{Connecting the Formal Definition Back to our Desiderata}
This brings us to the end of the formal definition of the transformer architecture.
As we saw, a transformer has many more moving parts than an RNN.
Is the additional complexity warranted?
Let us now return back to the informal motivations or desiderata that we laid out in \cref{sec:transformers-informal} and see how the components we defined in this section come through and ensure transformers fit them.
First of all, the transformer layers clearly store the representations of all symbols at all times---they are all needed to produce the query, key, and value matrices required by the attention mechanism.
As mentioned above, this allows us to easily store information about the entire string in a convenient and ``accessible'' format without having to compress it into a single hidden state.
Furthermore, the fact that the contextual representations $\mZ^(\ell + 1)_\tstep$ are computed from the representations $\vx^\ell_1, \ldots, \vx^\ell_\tstep$ directly at every time step, with no direct dependence between the different $\vz^(\ell + 1)_\idxi$, means that these computations can easily be parallelized.
More concretely, in the most common implementations of the transformer components, most operations take the form of matrix multiplications, which makes the computation and parallelization that much more efficient.\footnote{Note that, there is, of course, some sense of recurrence in the transformer---the composition of the transformer layers, which are stacked on top of each other, and of course require sequential computation.
    However, crucially, the number of layers does not depend on the \emph{length of the string}---the number of sequential steps required to process a string, therefore, does not depend on its length, which is what we wanted to achieve.}
Again, note that here, we are only interested in parallelizing the processing of \emph{entire} strings, as for example given in a training corpus.
As discussed in \cref{sec:rnn-parallelization}, there is an aspect of language modeling that is inherently sequential and even heavily parallelizable architectures such as the transformer cannot overcome: Generating stings one symbol at time.
While generating strings, even a transformer model will have to generate symbols one at a time and, therefore, recompute (parts of) the encoding $\tfencfun\left(\strlet\right)$ anew at every time step to generate the next symbol.
The advantages of parallelizability, therefore, come only at training time---however, given the vast corpora used for training today's models, this makes a crucial difference in the applicability of the architecture over recurrent ones.

Altogether, this means that transformers do, indeed, achieve the desiderata from our informal motivation!
This concludes our formal definition of transformers.
We move to analyze their theoretical properties.

\subsection{Tightness of Transformer-based Language Models}
Having introduced transformers formally, we can start investigating their formal properties.
As we did for RNN LMs, we first consider their tightness.
Specifically, in this subsection, we show that all \emph{soft attention-based} transformer language models are tight.
Key to our proof of the tightness of transformer language models, as well as the tightness of various other neural architectures, is the following basic fact in topology.
\begin{theorem}{Compactness}{compact}
    Let $\sX$ be a compact topological space and $\sY$ be any topological space.
    If $\func\colon \sX \to \sY$ is continuous, then $\func\left(\sX\right)\subseteq \sY$ is also compact.
\end{theorem}
\begin{proof}
    Let $\{\sU_\evs\}_{\evs\in\sA}$ be any open cover of $\func(\sX)$.
    By continuity, $\func^{-1}(\sU_\alpha)\subset \sX$ is open for any $\alpha\in\sA$, and hence $\{\func^{-1}(\sU_\evs)\}_{\alpha\in\sA}$ is also an open cover of $\sX$.
    By the compactness of $\sX$, there is a finite sub-cover $\{\func^{-1}(\sU_{\alpha_\idxn})\}_{\idxn=1}^N$, in which case $\{\sU_{\alpha_\idxn}\}_{\idxn=1}^N$ forms a finite sub-cover for $\func(\sX)$.
\end{proof}

We now further mathematically abstract transformers as a function on vector tuples,\footnote{Here $\left(\R^\hiddDim\right)^+$ is the set of nonempty tuples of vectors in $\R^\hiddDim$.  This is formally the disjoint union (coproduct) $\coprod_{\tstep\in\Z_{>0}}\R^{\tstep\times \hiddDim}$.} $\attnfunc\colon \left(\R^\hiddDim\right)^+ \to \left(\R^\hiddDim\right)^+$, that is \emph{length-preserving} in the sense that $\attnfunc\left(\R^{\tstep\times \hiddDim}\right)\subseteq \left(\R^{\tstep\times \hiddDim}\right)$ for all $\tstep > 0$.
Intuitively, this definition is saying that $\attnfunc$ is a function that maps a nonempty vector tuple $\{\vv_\idxj\}_{\idxj=1}^\tstep$ to another vector tuple $\{\bm{h}_\idxj\}_{\idxj=1}^\tstep$ of the same length,
\begin{align}\label{eq:transformer}
    \attnfunc(\vv_1,\dots,\vv_\tstep)=(\vh_1,\dots,\vh_\tstep)\in\R^{\tstep\times \hiddDim},
\end{align}
where $\vv_\idxj = \inEmbedding\left(\sym_\idxj\right) \in \R^\hiddDim$ are the initial representations of the input symbols $\sym_\idxj$.
In particular, we can take the function $\attnfunc\colon \left(\R^\hiddDim\right)^+ \to \left(\R^\hiddDim\right)^+$ to be the function defined by a stack of transformer layers, i.e., an attention block.
This setup will help us state the following.

\begin{lemma}{}{transformer-compact}
    Let $\attnfunc\colon \left(\R^\hiddDim\right)^+ \to \left(\R^\hiddDim\right)^+$ be the function defined by a $\tfnumlayer$ transformer layers with continuous functiona $\qTransf$, $\kTransf$, $\vTransf$, and $\oTransf$.
    Given a compact set $\sK\subset \R^\hiddDim$. Then, there exists a compact set $\sK'\subset\R^\hiddDim$ such that for every $t\in\Z_{>0}$,
    \begin{equation}
        \attnfunc\big(\sK^t\big)\subseteq \left(\sK'\right)^t.
    \end{equation}
\end{lemma}
\paragraph{Note.}
We make use of the following notations in the proof below: $B_r(\vz)=\{\vv\in\R^\hiddDim:\text{dist}(\vz,\vv)<r\}$ denotes the open ball centered at $\vz$ with radius $r$; $\overline{\sA}$ denotes the closure of set $\sA$.
\begin{proof} Let $\sK_0=\sK$.
    In an autoregressive transformer, each of the $\tfnumlayer$ layers consists of two blocks: a self-attention block and a feedforward block. We will use induction on the $2\tfnumlayer$ blocks to build up compact sets $\sK_1, \sK_2, \ldots, \sK_{2\tfnumlayer}$ that contain the output vectors of these respective blocks, and then take $\sK' = \sK_{2\tfnumlayer}$.

    The self-attention block is a function on $(\R^\hiddDim)^+\to(\R^\hiddDim)^+$.
    So, let $t\in\Z_{>0}$ be arbitrary and consider any sequence of input vectors $(\vv_1,\dots,\vv_t)$ such that for all $i$, $\vv_i\in \sK_0$.
    Denote the output vectors of the attention block with $(\vv_1', \dots,\vv_t')$.
    By definition of attention, each output vector $\vv_j'=\sum_{i=1}^t \evs^{(j)}_i\vv_i$ where $\vs^{(j)}\in\Simplextminus$ are the attention weight vectors obtained through the softmax function.
    Compact sets in $\R^\hiddDim$ are bounded (by the Heine--Borel theorem), and hence there exists $M>0$ such that $\sK_0\subseteq\overline{B_M(0)}$.
    Noting that the norm function $\Vert\cdot\Vert$ on $\R^\hiddDim$ is convex, we have the following
    \begin{subequations}
        \begin{align}
            \Vert\vv_j'\Vert & =\left\Vert\sum_{i=1}^t\evs^{(j)}_i\vv_i\right\Vert                               \\
                             & \leq\sum_{i=1}^t \evs^{(j)}_i \Vert\vv_j\Vert \label{eq:used-jensen} \tag{$\ast$} \\
                             & \leq\sum_{i=1}^t \evs^{(j)}_i M=M \label{eq:attn-output-bound}
        \end{align}
    \end{subequations}
    where (\ref{eq:used-jensen}) results from Jensen’s inequality. \cref{eq:attn-output-bound} shows that each of the output vectors $\vv_j'$ lies in $\overline{B_M(0)}$ which is compact. Hence, setting $\sK_1=\overline{B_M(0)}$, we have shown that, for any $t\in\Z_{>0}$, the attention block maps $\sK_0^t$ into $\sK_1^t$.

    Note that we \emph{cannot} use \cref{thm:compact} here because the attention block defines a different function on $\R^{t\times \hiddDim}\to\R^{t\times \hiddDim}$ for each $t$, and \cref{thm:compact} only implies that there exists a separate \emph{length-dependent} output compact set $\sK_t\subset\R^{t\times \hiddDim}$ for each $t$, which is different from this lemma's statement.

    The feedforward function is a continuous function on $\R^\hiddDim\to\R^\hiddDim$, and therefore, by \cref{thm:compact}, maps its input compact set $\sK_1$ to an output compact set, which we call $\sK_2$.

    Finally, residual connections and layer norms are also continuous functions acting on each of the input vectors, and hence by the same reasoning would also preserve compactness.

    Now we can use induction and show that there exist compact sets $\sK_3$, $\sK_4, \dots, \sK_{2\tfnumlayer-1}$, $\sK_{2\tfnumlayer}$ where $\sK_{2\tfnumlayer}$ contains the output set of the final layer. Set
    $\sK'=\sK_{2\tfnumlayer}$ and we have proven the statement.
\end{proof}

Now recall that a transformer language model with the softmax projection function (\cref{def:transformer-plnsm}) defines the conditional probabilities using the softmax transformation
\begin{align}\label{eq:transformer-softmax}
    \pLNSM(\eossymt\mid \strlt) = \frac{\exp(\embedding{\eossymt}^\top \hiddStatet)}{\sum_{\eossym'\in\eosalphabet}\exp(\embedding{\eossym'}^\top \hiddStatet)}
\end{align}
where $\embedding{\eossym}\in\R^\hiddDim$ is the output symbol embedding of $\eossym\in\eosalphabet$ and $\hiddStatet$ is defined from the input embeddings of $\strlt$ via \cref{eq:transformer}.
Using \cref{lem:transformer-compact}, together with the finiteness of the vocabulary $\alphabet$ and the continuity of the softmax transformation \labelcref{eq:transformer-softmax}, readily yields our main result on transformer language models.

\begin{theorem}{Transformer language models are tight}{transformer-main}
    The representation-based locally normalized language model (cf. \cref{def:transformer-plnsm}) defined by any (fixed-depth) transformer with soft attention is tight.
\end{theorem}
\begin{proof}
    Given the Transformer, there exists a fixed compact set $\sK$ that will contain all inputs $\vv_i \in \R^\hiddDim$ to the first layer.
    This is true because each $\vv_i$ is the sum of a word embedding, which falls in a finite set since $\eosalphabet$ is finite, and a position embedding, which lies in the compact set $[-1,1]^\hiddDim$.
    Hence, by \cref{lem:transformer-compact}, there exists a fixed compact set $\sK'$ that contains all output embedding vectors (regardless of how long the sequence is).

    The final output probability is given by a multiplication with the word embedding matrix followed by the softmax function as in \cref{eq:transformer-softmax}.
    This process amounts to composing two continuous functions.
    In particular, we can extract the \eos{} probability as a continuous $\R$-valued function $\funcg^\eos: \sK'\to\left(0,1\right)$ (neither 0 nor 1 is in the range of the softmax function).
    By continuity of $\funcg^\eos$ and \cref{thm:compact}, $\sK''\defeq \funcg^\eos(\sK')\subseteq(0,1)$ is compact.
    Since $\sK''$ is compact, and hence closed, $\inf \sK'' \in \sK''$.
    Thus $\inf \sK'' \in (0,1)$ and in particular $\inf \sK'' > 0$.
    Therefore, taking $\epsilon=\inf \sK''$, we have shown that the \eos{} probability of a Transformer is bounded below by some $\epsilon>0$ (regardless of the length of the sequence).
    Hence, by \cref{prop:div-implies-tight}, any transformer-based sequence model is tight and thus defines a language model.
\end{proof}

\section{Representational Capacity of Transformer Language Models} \label{sec:transformers-expressivity}
So far, we have introduced our formal definition of the transformer architecture and examined its tightness.
We now move on to the computational power of the architecture.
This section mirrors \cref{sec:rnn-expressiveness} and examines the expressivity of the transformer language model as defined in \cref{def:transformer-plnsm}.

Transformers are a much more recent architecture than recurrent neural language models, and our theoretical understanding of them is thus much more limited.
However, over the last few years, a series of results showing various properties of the transformer model have been established.
At first glance, one might find a number of contradictions among them: One of the first results shows that transformers are not even able to recognize the very simple \textsc{First} language recognized by the (unweighted) finite-state automaton shown in \cref{fig:first-fsa} nor the Dyck language.
\begin{figure}
    \centering
    \begin{tikzpicture}[node distance=8mm]
        \node[state, initial] (q1) [] { 1 };
        \node[state, accepting] (q2) [right = of q1] { 2 };
        \draw (q1) edge[auto] node[minimum size=4mm]{ $1$ } (q2)
        (q2) edge[auto, loop above] node[minimum size=4mm]{ $0, 1$ } (q2) ;
    \end{tikzpicture}
    \caption{An FSA recognizing the language $\textsc{First} = \{\str \in \kleene{\alphabet} \mid \alphabet =\{0,1\}, \text{ $\sym_1 = 1$} \}$.}
    \label{fig:first-fsa}
\end{figure}
On the other hand, there is work showing that transformers \emph{can} recognize the majority language (determining whether a string contains more symbols $\syma$ or $\symb$) and can even \emph{count}: Both of these languages are instances of non-regular languages.
Moreover, a fundamental result by \citet{perez_2021_turingtransformers} even shows that transformers are \emph{Turing complete}.
Upon closer inspection, the results can be explained by the different \emph{theoretical abstractions} of the original transformer model that the different works make and even different notions of equivalence.
Even very subtle differences in the model can lead to substantial differences in the expressivity of the model, as we will see below.
In this section, we present some original results which show that transformers can, with infinite precision, in fact, reach up to the top of the hierarchy of formal languages that we consider in these notes: They are Turing complete.
We also comment on the differences in our approach to the work so far (the main difference and novelty is that we embed the analysis into our language modeling framework) and try to unify it.
We will show that infinite-precision transformers can simulate the recognizers across the entire hierarchy: (weighted) finite-state automata, pushdown automata, and Turing machines.
Luckily, apart from the first construction, the proofs and constructions in our ascent up the hierarchy will be based on a unified approach in which we build on \citet{perez_2021_turingtransformers} and sequentially add components to be able to recognize more and more complex languages.

Besides being more novel and thus less researched, transformers are also less intuitive to think about as sequential machines transitioning between states as with finite-state or pushdown automata.
All classical computational models we introduced (finite-state automata, pushdown automata, and Turing machines) rely on some notion of an internal state which is sequentially updated, where the next state is determined based on the current configuration.
We also said in \cref{sec:rnn-parallelization} that this sequentiality is the Achilles' heel of the ability to parallelize and thus speed up the computations in a language model.
One of the main motivations for defining the transformer model is to avoid these sequential dependencies and to make sure the contextual representations of the individual symbols can be computed independently.
However, the lack of sequentiality in transformers makes it more difficult to compare to classical and well-understood models of computation---they simply do not define any notion of a configuration that would be passed over by reading a symbol at a time, and relating the configurations at different time points to the configuration of some classical model of computation was the main idea of most of the analyses in \cref{sec:rnn-expressiveness}.
This will not be possible with transformers, and we will have to be more clever about it to draw parallels to better-understood formalisms.
What is more, it seems like their parallelizable nature is one of the reasons for the lower (or, at least, ambiguous) computational power \emph{under some formalisms}, as covered in \citet{merrill-etal-2022-saturated,merrill-sabharwal-2023-parallelism}.

\Anej{They actually need infinite precision to even be finite state (positions)...}

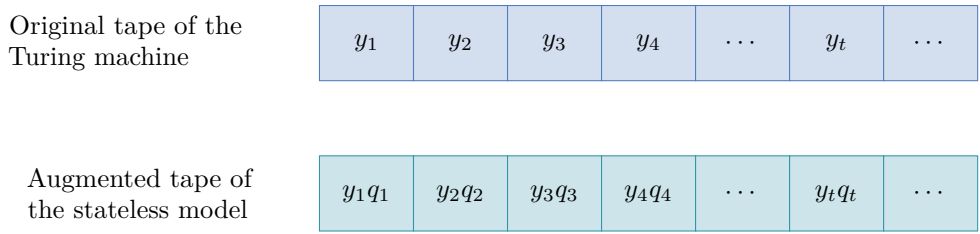
\begin{figure}
    \centering

    \begin{tikzpicture}[
            y=2cm,
            original tape node/.style={draw=ETHBlue!80,minimum size=1cm, minimum width=1.25cm, fill=ETHBlue!20},
            augmented tape node/.style={draw=ETHPetrol!80,minimum size=1cm, minimum width=1.25cm,fill=ETHPetrol!20}
        ]

        \node[draw=none, text width=3.5cm] at (-3, 1) {Original tape of the Turing machine};
        \node[draw=none, text width=3cm] at (-3, 0) {Augmented tape of the stateless model};

        \foreach \i/\y in {0/$\sym_1$,1/$\sym_2$,2/$\sym_3$,3/$\sym_4$,4/$\cdots$,5/$\symt$,6/$\cdots$} {
                \node[original tape node] (tape-\i) at (1.25*\i,1) {\y};
            }

        \foreach \i/\y in {0/$\sym_1\stateq_1$,1/$\sym_2\stateq_2$,2/$\sym_3\stateq_3$,3/$\sym_4\stateq_4$,4/$\cdots$,5/$\symt\stateq_\tstep$,6/$\cdots$} {
                \node[augmented tape node] (aug-tape-\i) at (1.25*\i,0) {\y};
            }

    \end{tikzpicture}
    \caption{An abstract illustration of how a model can keep track of its internal state by ``outputting'' it into the generated string.
        By reading the augmented symbol generated at time $\tstep$, the model can then determine its internal state.}
    \label{fig:transformer-augmented-alphabet}
\end{figure}

\paragraph{A word on model equivalence.}
\Anej{This should turn into its own subsection somewhere above at some point, and then this should just link back to the section on homomorphisms later.}
As mentioned above, the nature of the transformer architecture does not lend itself well to a straightforward comparison to classical models of computation.
To make the connection, we will have to be somewhat clever about the analysis.
As we will see shortly, we will mainly deal with this in two ways:
\begin{enumerate*}[label=\textit{(\roman*)}]
    \item By foregoing any notion of a state of a machine in case of \ngram{} language models\footnote{Reminder that \ngram{} language models are in fact \emph{subregular} (cf. \cref{sec:subregular}) and we make use of that in our analysis. Because their recognition relies purely on \emph{local} patterns in the strings, and a transformer model has the ability to consider large enough substrings, we will see that we can model an \ngram{} language model without keeping any notion of a state in a transformer} and
    \item by \emph{embedding} the state of a computational model \emph{into the alphabet itself}---the model will then use the augmented output alphabet to keep track of its state in the string itself without relying on any notion of its own internal state which would have to be updated sequentially.\footnote{Recall that, as discussed in \cref{sec:rnn-parallelization}, generation is inherently sequential. One can thus imagine augmenting the alphabet as a sort of exploitation of this sequential nature.}
\end{enumerate*}
How can this help us?
As will become clear in our analysis of the Turing completeness of a transformer model, the model can use the \emph{generated string} as a sort of a \emph{sequential memory structure}.
Because the transformer model can look back at the entirety of the string when computing $\tfencfun\left(\strlet\right)$ (where $\strlet$ is the augmented string generated so far), it is able to ``read off'' its internal state from the string.
Importantly, the generated string will still contain the information about the generated string, besides including the state of the computational model.
As the transformer will then compute the new embeddings $\tfencfun\left(\strlet\right)$, it will be able to account for the state it should be in.
This is illustrated in \cref{fig:transformer-augmented-alphabet}.

While this might seem like a convenient trick to achieve Turing completeness---and in many ways, it is---it is also, in a way, cheating.
This ``cheating'' can be described formally as the difference between model \emph{equivalence} and \emph{homomorphism equivalence}.
When we discussed the Turing completeness of RNN LMs, we showed they can model a Turing machine by directly recognizing the same strings (for the time being, we ignored the string weights).
This means that, for every Turing machine $\tm$, there exists an RNN $\rnn$ which recognizes the same language: $\lang\left(\rnn\right) = \lang\left(\tm\right)$.
However, we will not be able to make statements like this in the case of transformer models.
The augmented alphabet will instead bring us to a statement of the sort ``For every Turing machine $\tm$, there exists a transformer $\transformernetwork$ which recognizes the same language augmented with the state set of the Turing machine: $\lang_\extAlphabet\left(\transformernetwork\right) = \lang\left(\tm\right)$,'' where $\lang_\extAlphabet$ refers to the language of strings where each symbol is additionally augmented with the state of the Turing machine.
This might seem like a small difference, but, in formal language theory, homomorphism equivalence refers to a different problem to that of normal model equivalence \citep{CULIK1978163} and thus has to be considered differently.
Intuitively, it additionally allows additional information to be stored in the strings (in our case, that will be the state of the Turing machine) while still considering some models to be ``equivalent''.
Formally, model equivalence asks the following question.
\begin{definition}{Model equivalence}{model-equivalence}
    Two computational models $\compModel_1$ and $\compModel_2$ are \defn{equivalent} if
    \begin{equation}
        \lang\left(\compModel_1\right) = \lang\left(\compModel_2\right).
    \end{equation}
\end{definition}
On the other hand, homomorphic equivalence considers the following.\anej{add a definition of a homomorphism}
\begin{definition}{Homomorphism}{homomorphism-equivalence-1}
    Let $\compModel_1$ and $\compModel_2$ be two computational models.
    $\compModel_1$ is \defn{homomorphically equivalent} to $\compModel_2$ if there exists a homomorphisms $\homomorphism\colon \langFun{\compModel_1} \to \langFun{\compModel_2}$ such that
    \begin{equation}
        \homomorphism\left(\langFun{\compModel_1}\right) \defeq \set{\homomorphism\left(\str\right) \mid \str \in \lang\left(\compModel_1\right)} = \lang\left(\compModel_2\right).
    \end{equation}
\end{definition}
\begin{definition}{Homomorphic equivalence}{homomorphism-equivalence}
    Let $\compModel_1$ and $\compModel_2$ be two computational models.
    $\compModel_1$ and $\compModel_2$ are \defn{homomorphically equivalent} if $\compModel_1$ is homomorphically equivalent to $\compModel_2$ and $\compModel_2$ is homomorphically equivalent to $\compModel_1$ as per \cref{def:homomorphism-equivalence-1}.
\end{definition}

\subsubsection{Transformers and the Inability to Recognize Simple Languages} \label{sec:transformers-regular-languages}
We start our exploration of the computational power of transformer models with some negative results, which we will later ``correct'' by using our formalization of a transformer model and different components of the transformer architecture (for example, a different form of attention).
Given their success at modeling human language which is assumed to be at least mildly context-sensitive \citep{huybregts1984van, shieber1985}, it seems surprising that transformers cannot, in fact, recognize some very simple regular languages, such as \textsc{Parity} or \textsc{First} (the FSA shown in \cref{fig:first-fsa}), as well as simple non-regular context-free languages such as \textsc{Dyck} languages:
\begin{align*}
    \textsc{First}  & = \{\str \in \kleene{\alphabet} \mid \alphabet =\{0,1\}, \text{ $\sym_1 = 1$} \}                       \\
    \textsc{Parity} & = \{\str \in \kleene{\alphabet} \mid \alphabet =\{0,1\}, \text{ $\str$ has odd number of 1s} \}        \\
    \textsc{Dyck}   & = \{\str \in \kleene{\alphabet} \mid \alphabet = \{(,)\}, \text{ $\str$ is correctly parenthesized} \}
\end{align*}
This has been formally shown by \citet{hahn-2020-theoretical}, and experimentally verified by \citet{chiang-cholak-2022-overcoming}.
\citet{bhattamishra-etal-2020-ability} found that transformers especially struggle to learn any languages that require counting occurrences in some way, such as the number 0s and 1s in \textsc{Parity} or the number of previous open and closed parentheses in \textsc{Dyck}.
\citet{hahn-2020-theoretical} finds that with \emph{unique hard attention}, these languages cannot be recognized: Recognizing them by a transformer in their formulation would require the number of parameters to increase with the length of the input.
\citet{chiang-cholak-2022-overcoming} consider the setting with soft attention, where the issue is more subtle: In theory, it is possible for a transformer to recognize languages such as \textsc{First} and \textsc{Parity}, however with less and less confidence as the length increases.
This is reflected by the cross-entropy of deciding language membership approaching the worst possible value of 1 bit per symbol.
The reason behind this is quite intuitive: The membership of any of the languages defined above \emph{changes} if a \emph{single} symbol changes.
However, by examining the information flow in a transformer, one can show that the corresponding information gets less and less weight relative to the length of the string due to the attention mechanism averaging over all positions.

\subsubsection{Transformers Can Simulate \ngram{} Models}
\cref{sec:transformers-regular-languages} showed that transformer models can struggle to recognize some of the simplest formal languages.
While we did not discuss those results in detail, intuitively, they stem from the use of unique hard attention and the resulting inability to take into account all values whose keys maximize the attention scoring function.
By relaxing that restriction to \emph{averaging} hard attention, the model becomes more expressive.
To show that, we begin by looking at the very simple \ngram{} language models, as defined in \cref{sec:n-gram-models}.
By constructing, for any \ngram{} model, a transformer representing it, we will show the following theorem.
\begin{theorem}{Transformer language models can simulate \ngram language models}{transformers-ngrams}
    Let $\pLN$ be an \ngram{} language model.
    Then, there exists a transformer $\transformernetwork$ with $\lang\left(\pLN\right) = \lang\left(\transformernetwork\right)$.
\end{theorem}
Alternatively, we could say that transformers can recognize \emph{strictly local languages} (cf. \cref{sec:subregular}).
\begin{proof}
    We prove the theorem by constructing, for $\pLN$, a transformer $\transformernetwork$ with $\lang\left(\pLN\right) = \lang\left(\transformernetwork\right)$.
    Note that we will mostly restrict the proof to the construction of the transformer, i.e., the formal definition of its parameters.
    The (mostly trivial) mathematical details and derivations are left as an exercise to the reader.

    Recall that, by definition, an \ngram{} language model considers a fixed number of previous symbols to define $\pLNSM\left(\sym_\tstep\mid\str_{<\tstep}\right)$---exactly $\ngr - 1$ of them.
    The constructed transformer $\transformernetwork$ will capture this idea  with $\ngr - 1$ \emph{heads}, each of them attending to \emph{exactly one} of those positions in the previous $\ngr - 1$ positions.\footnote{Note that, given an \ngram{} model, the number $\ngr$ is fixed. This means that, for a given \ngram{} we can always fix the number of heads and therefore construct such a transformer.}
    We can then use the symbols the heads attended to (and thus identified) to identify the current \ngram{} and with it define the relevant conditional distribution over the next symbol.
    To be able to attend to the positions of interest---the ones containing the previous $\ngr - 1$ symbols---we have to make use of appropriate positional encodings (cf. \cref{def:positional-encodings}), which will allow the model to attend to them.
    The idea of the construction is abstractly illustrated in \cref{fig:transformer-ngram}.
    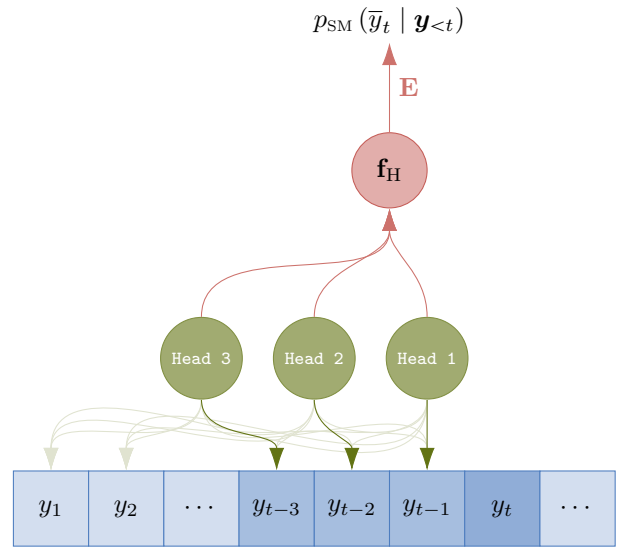
\begin{figure}
        \centering

        \begin{tikzpicture}[
            tape node/.style={draw=ETHBlue!80,minimum size=1cm,fill=ETHBlue!20},
            head node/.style={draw=ETHGreen!80,circle,minimum size=0.75cm,fill=ETHGreen!60,text=white},
            attn arrow/.style={-{Latex[length=3mm,width=2mm]},ETHGreen!100},
            comb arrow/.style={-{Latex[length=3mm,width=2mm]},ETHRed!70},
            comb node/.style={draw=ETHRed!80,circle,minimum size=1cm,fill=ETHRed!40},
            ]

            \foreach \i/\y in {0/$\sym_1$,1/$\sym_2$,2/$\cdots$,3/$\sym_{\tstep-3}$,4/$\sym_{\tstep-2}$,5/$\sym_{\tstep-1}$,6/$\symt$,7/$\cdots$} {
                    \ifnum \i=6
                        \node[tape node,fill=ETHBlue!50] (tape-\i) at (\i,0) {\y};
                    \else
                        \ifnum \i>2
                            \ifnum \i<6
                                \node[tape node,fill=ETHBlue!40] (tape-\i) at (\i,0) {\y};
                            \fi
                        \else
                            \node[tape node,fill=ETHBlue!20] (tape-\i) at (\i,0) {\y};
                        \fi
                        \ifnum \i>6
                            \node[tape node,fill=ETHBlue!20] (tape-\i) at (\i,0) {\y};
                        \fi
                    \fi
                }

            \node[head node] (head-1) at (2,2) {\scriptsize \texttt{Head 3}};
            \node[head node] (head-2) at (3.5,2) {\scriptsize \texttt{Head 2}};
            \node[head node] (head-3) at (5,2) {\scriptsize \texttt{Head 1}};

            \draw[attn arrow, ETHGreen!20] (head-1) to[out=270,in=90] (tape-0.north);
            \draw[attn arrow, ETHGreen!20] (head-1) to[out=270,in=90] (tape-1.north);
            \draw[attn arrow, ETHGreen!20] (head-1) to[out=270,in=90] (tape-4.north);
            \draw[attn arrow, ETHGreen!20] (head-1) to[out=270,in=90] (tape-5.north);
            \draw[attn arrow, ETHGreen!20] (head-2) to[out=270,in=90] (tape-0.north);
            \draw[attn arrow, ETHGreen!20] (head-2) to[out=270,in=90] (tape-1.north);
            \draw[attn arrow, ETHGreen!20] (head-2) to[out=270,in=90] (tape-3.north);
            \draw[attn arrow, ETHGreen!20] (head-2) to[out=270,in=90] (tape-5.north);
            \draw[attn arrow, ETHGreen!20] (head-3) to[out=270,in=90] (tape-0.north);
            \draw[attn arrow, ETHGreen!20] (head-3) to[out=270,in=90] (tape-1.north);
            \draw[attn arrow, ETHGreen!20] (head-3) to[out=270,in=90] (tape-3.north);
            \draw[attn arrow, ETHGreen!20] (head-3) to[out=270,in=90] (tape-4.north);
            \draw[attn arrow] (head-1) to[out=270,in=90] (tape-3.north);
            \draw[attn arrow] (head-2) to[out=270,in=90] (tape-4.north);
            \draw[attn arrow] (head-3) to[out=270,in=90] (tape-5.north);

            \node[comb node] (combiner) at (4.5,4.5) {$\tfheadCombine$};

            \draw[comb arrow] (head-1.north) to[out=90,in=270] (combiner.south);
            \draw[comb arrow] (head-2.north) to[out=90,in=270] (combiner.south);
            \draw[comb arrow] (head-3.north) to[out=90,in=270] (combiner.south);

            \node[fill=none] (out) at (4.5,6.5) {$\pLNSM\left(\eossym_\tstep\mid\str_{<\tstep}\right)$};

            \draw (combiner) edge[comb arrow, right] node{$\outMtx$} (out.south);

        \end{tikzpicture}
        \caption{An abstract depiction of how a transformer can simulate an \ngram{} model using $\ngr - 1$ heads (here, $\ngr = 4$). The stronger arrows from the heads to the symbols in the string show where the heads concentrate their attention. The lighter green arrow is meant to represent that the heads still can consider the entire history of the input so far but are then \emph{configured} such that they only look at the appropriate position.}
        \label{fig:transformer-ngram}
    \end{figure}

    For hopefully a better pedagogical effect, we will present this proof from the ``last'' part of the construction to the ``first''.
    We, therefore, start with the final step: Assuming we have identified the appropriate \ngram{} $\str_{\tstep - \ngr\colon \tstep - 1} \eossym_\tstep$, how can we encode the conditional probability distribution $\pLNSM\left(\eossym_\tstep\mid\str_{\tstep - \ngr\colon \tstep - 1}\right)$?
    The construction we use here directly mirrors the one in Minksy's construction (cf. \cref{lem:minsky-constr}): Knowing what the individual $\pLNSM\left(\eossym_\tstep\mid\str_{\tstep - \ngr\colon \tstep - 1}\right)$ for $\eossym_\tstep \in \eosalphabet$ are (those are, as described in \cref{sec:n-gram-models}, ``hard-coded'', or specified, for each \ngram{} separately in a look-up table), we can simply put their logits ($\log$ probabilities; in case we are using the softmax projection function) or the probabilities directly (if we are using the sparsemax projection function) into a vector and concatenate all the constructed vectors (for all possible \ngram{}s) into a large matrix $\outMtx$.

    If we \emph{one-hot encode} the identified \ngram{} by defining
    \begin{equation}
        \tfencfun\left(\str_{<\tstep}\right) \defeq \onehot{\str_{\tstep - \ngr \colon \tstep - 1}}
    \end{equation}
    we can then, using the formulation of the transformer sequence model from \cref{def:transformer-plnsm}, use the one-hot encoded \ngram{} to lookup the appropriate column containing the conditional probabilities given the identified \ngram{} for all possible $\eossym_\tstep \in \eosalphabet$.\footnote{Note that we are again working over the set of \emph{extended} reals $\Rex = \R \cup \set{-\infty, \infty}$ in case of the softmax activation function.}.
    The formal proof of correctness given that we have identified the correct \ngram{} is therefore analogous to the final part of the Minsky construction.
    \Anej{Exercises:

        How big is the vector $\onehot{\str}$? And the matrix $\outMtx$?

        Define the matrix $\outMtx$ formally.}

    We now consider the preceding step of the simulation: How can we identify the complete \ngram{} given that the $\ngr - 1$ heads of the transformer identified the symbols in the positions they attended to?
    This, it turns out, is a simple instance of the ``\texttt{AND}'' problem investigated in \cref{fact:multi-hot-to-one-hot}: After concatenating the values of the $\ngr - 1$ heads into a common vector $\vv$ (each of which is a $|\eosalphabet|$-dimensional vector), this vector of size $|\eosalphabet|\left(\ngr - 1\right)$ will contain the \defn{multi-hot} representation of the \ngram{} of interest.
    Let $\eossym_1, \ldots, \eossym_{\ngr - 1}$ be the symbols represented by $\vv$.
    This means that $\vv$ is of the form
    \begin{equation}
        \vv = \begin{pmatrix}
            \onehot{\eossym_1} \\
            \vdots             \\
            \onehot{\eossym_{\ngr - 1}}
        \end{pmatrix}
    \end{equation}
    and $\vv_{\idxk |\eosalphabet| + \idxj} = 1$ if and only if $\eossymordering\left(\eossym_\idxk\right) = \idxj$ for an ordering $\eossymordering$ of $\eosalphabet$ determining the one-hot representations of the individual symbols.
    We would then like to transform this vector into a vector $\vu \in \R^{|\alphabet|^{\ngr - 1}}$ such that
    \begin{equation}
        \vu_{\idxi} = 1 \text{ if and only if } \idxi = \anOrdering\left(\eossym_1, \ldots, \eossym_{\ngr - 1}\right)
    \end{equation}
    for some ordering $\anOrdering$ of $\underbrace{\eosalphabet \times \cdots \times \eosalphabet}_{\ngr - 1\text{ times}}$.
    This can be equivalently written as
    \begin{equation}
        \vu_{\idxi} = 1 \text{ if and only if } \vv_{\idxk |\eosalphabet| + \eossymordering\left(\eossym_\idxk\right)} + 1 \text{ for all } \idxk = 1, \ldots, \ngr - 1
    \end{equation}
    where $\idxi = \anOrdering\left(\eossym_1, \ldots, \eossym_{\ngr - 1}\right)$.
    Clearly, this is the same problem as described in \cref{fact:multi-hot-to-one-hot} and can therefore be solved by a linear transformation followed by the application of the thresholded sigmoid nonlinearity, which will together form the transformation $\tfheadCombine$ combining the information obtained from all the heads of the transformer model.
    Note that, to make this more similar to the practice of how transformers are actually implemented, we could also use the $\ReLU$ activation function instead of the saturated sigmoid.

    This brings us to the final part of the proof, which considers the first part of determining the conditional probability of the \ngram{} model by the transformer: Identifying the symbols at the previous $\ngr - 1$ positions by the $\ngr - 1$ heads of the transformer.
    To show how this can be done, let us consider and define the ``degrees of freedom'' we have left when specifying a transformer model in our framework.
    \begin{itemize}
        \item The symbol representations $\tfembfun$.
              We will use simple one-hot encodings of the tokens: $\tfembfun\left(\eossym\right) \defeq \onehot{\eossym}$.
        \item The positional encodings $\posEnc$.
              We will use the following simple positional encoding: $\posEncFun{\tstep} = \begin{pmatrix}
                      \tstep \\ 1
                  \end{pmatrix}$.
              The utility of the constant $1$ will be made clear shortly.
              We will combine positional encodings with symbol representations by \emph{concatenating} them into a vector of size $|\eosalphabet| + 2$.
        \item The number of transformer layers.
              We will use a single transformer layer.
        \item The number of heads $\tfheadnum$: As we mentioned, we will use $\tfheadnum = \ngr -1$ heads to attend to the previous $\ngr - 1$ symbols.
        \item The form of the attention scoring function $\tfscorefun$.
              While not the most typical, we will use the following scoring function:
              \begin{equation} \label{eq:transformer-ngram-function}
                  \tfscorefun\left(\vq, \vk\right) \defeq -|\innerProd{\vq}{\vk}|.
              \end{equation}
              It will, together with the positional encodings, allow us to easily single out the positions in the string that we care about.
        \item The form of attention.
              We will use hard attention (in this case, it can be either unique or averaging).
        \item The parameters of each of the attention heads, that is the transformations $\qTransf$, $\kTransf$, and $\vTransf$.
              Each of those will take the form of a linear transformation of the symbol embedding.
              We describe them and their roles in more detail below.
    \end{itemize}
    As mentioned above, the input symbol $\eossym_\tstep$ is presented to the transformer model together with its positional encoding in the form
    \begin{equation}
        \tfembfun\left(\eossym_\tstep\right) = \begin{pmatrix}
            \onehot{\eossym_\tstep} \\
            \tstep                  \\
            1
        \end{pmatrix} \in \R^{|\eosalphabet| + 2}.
    \end{equation}
    The parameters of all the heads are defined in the same way, with the only difference being a simple parameter that depends on the ``index'' of the head we are considering, $h$.
    Therefore, in the following, we describe the construction of a single head \texttt{Head $h$}.
    At any time step $\tstep$ (i.e., when modeling the conditional distribution $\pLNSM\left(\eossym_\tstep \mid \str_{<\tstep}\right)$), the head $h$ will attend to or be ``responsible for'' recognizing the symbol at position $\tstep - h$, $\sym_{\tstep - h}$.
    This can be seen in \cref{fig:transformer-ngram}, where, for example, \texttt{Head $3$} is responsible for the position $\tstep - 3$, which is denoted by the stronger arrow to that position.
    All we still have to do is describe the individual transformations $\qTransf_h$,  $\kTransf_h$,  $\vTransf_h$ of the head $h$.
    All of them will be \emph{linear} transformations, i.e., matrix multiplication:
    \begin{align}
        \vq \defeq \qTransf\left(\vx\right) \defeq \mQ_h \vx \\
        \vk \defeq \kTransf\left(\vx\right) \defeq \mK_h \vx \\
        \vv \defeq \vTransf\left(\vx\right) \defeq \mV_h \vx
    \end{align}
    We now define the matrices $\mQ_h$, $\mK_h$, and $\mV_h$, specifically in the \emph{first} (in this case, the only) layer of a transformer language model.
    Importantly, since we are talking about only the first layer, we can simply consider as inputs to the layer the original static symbol representations together with their position encodings rather than any contextual representations.
    First, let us consider again what roles the matrices play in computing $\tfencfun\left(\str_{<\tstep}\right)$.
    In the context of language modeling, the matrix $\mQ_h$ takes in the representation of the ``latest'' generated symbol $\sym_{\tstep - 1}$ and produces from it the query vector of $\sym_{\tstep - 1}$.
    It is, therefore, only applied \emph{once} per generation step---only for symbol $\sym_{\tstep - 1}$.
    The matrices $\mK_h$ and $\mV_h$, on the other hand, transform \emph{all} non-masked input symbols to the key and value vectors.
    That is, they take the representations of the input symbols and their positions encodings $\tfencfun\left(\sym_\idxj\right)$ for every $\idxj = 1, \ldots, \tstep - 1$ and transform them into the key and value vectors.
    The keys will then be compared with the query constructed for $\sym_{\tstep - 1}$ with the $\mQ_h$ matrix, while the constructed values will be used to compute the new hidden state $\hiddStatet$.\footnote{Importantly, in a multi-layer transformer, the queries would be constructed for \emph{every} non-masked symbol and its representation (hidden state) would be updated. However, since the updated representations would not be used in the single layer case, we only have to compute the representation of the newest symbol in this case.}

    So, what kind of query, key, and value vectors do we want?
    As mentioned, the head $h$ will be responsible for identifying the symbol at position $\tstep - h$.
    Therefore, we want it to put all its attention to this position.
    In other words, given the query $\vq_{\tstep - 1}$, we want the attention function in \cref{eq:transformer-ngram-function} to be maximized by the key of the symbol at position $\tstep - h$.
    Notice that, therefore, the key does not have to depend on the identity of the symbol at position $\tstep - h$---only the position information matters.
    Let us then consider the following query and key transformations for head $h$:
    \begin{align}
        \qTransf\colon & \begin{pmatrix}
                             \onehot{\eossym_\tstep} \\
                             \tstep                  \\
                             1
                         \end{pmatrix} \mapsto \begin{pmatrix}
                                                   \tstep - h \\
                                                   1
                                               \end{pmatrix} \\
        \kTransf\colon & \begin{pmatrix}
                             \onehot{\eossym_\idxj} \\
                             \idxj                  \\
                             1
                         \end{pmatrix} \mapsto \begin{pmatrix}
                                                   -1 \\
                                                   \idxj
                                               \end{pmatrix}.
    \end{align}
    Given such a query and such keys, the attention scoring function computes
    \begin{equation}
        \tfscorefun\left(\vq_\tstep, \vk_\idxj\right) = -|\innerProd{\begin{pmatrix}
                \tstep - h \\
                1
            \end{pmatrix}}{\begin{pmatrix}
                -1 \\
                \idxj
            \end{pmatrix}}| = -|\tstep - h - \idxj|,
    \end{equation}
    which is maximized exactly when $\idxj = \tstep - h$, that is, at the position that we want the head $h$ to attend to!
    This means that the hard attention we use will put all its probability mass to exactly the position we intended it to.
    Intuitively, both transformations keep only the positional information.
    The query transformation ``injects'' the knowledge of which position should maximize the attention score, while the key transformation (which is, again, applied to all the non-masked positions) simply ``exposes'' the positional information about the symbol.
    The alternating constant $1$ (or $-1$) and the index of the position ensure that the inner product simply computes the \emph{difference} between the position of the symbol and the position of interest---we will use this trick multiple times in later constructions as well.
    It is easy to see that the two transformations are indeed linear.

    This leaves us with the question of how to use this position of the symbol of interest ($\tstep - h$) to extract the one-hot encoding of the symbol at that position.
    Luckily, due to the information contained in the symbol representations $\tfembfun\left(\sym_\idxj\right)$, this is trivial.
    All that the transformation $\vTransf$ has to do is the following:
    \begin{equation}
        \vTransf\colon \begin{pmatrix}
            \onehot{\eossym_\idxj} \\
            \idxj                  \\
            1
        \end{pmatrix} \mapsto \onehot{\eossym_\idxj}.
    \end{equation}
    With this, the identity of the symbol is carried forward through the attention mechanism.
    Again, is easy to see that this is a linear transformation of the symbol representation.
    Notice that the only head-depend transformation is the query transformation---it depends on the index of the head, determining the position of interest, meaning that every head defines a different query transformation, while the keys and values transformations are the same among all heads.

    This concludes the proof.
    \cref{fig:transformer-ngram-full} again shows an illustration of the described model with all the defined components.
    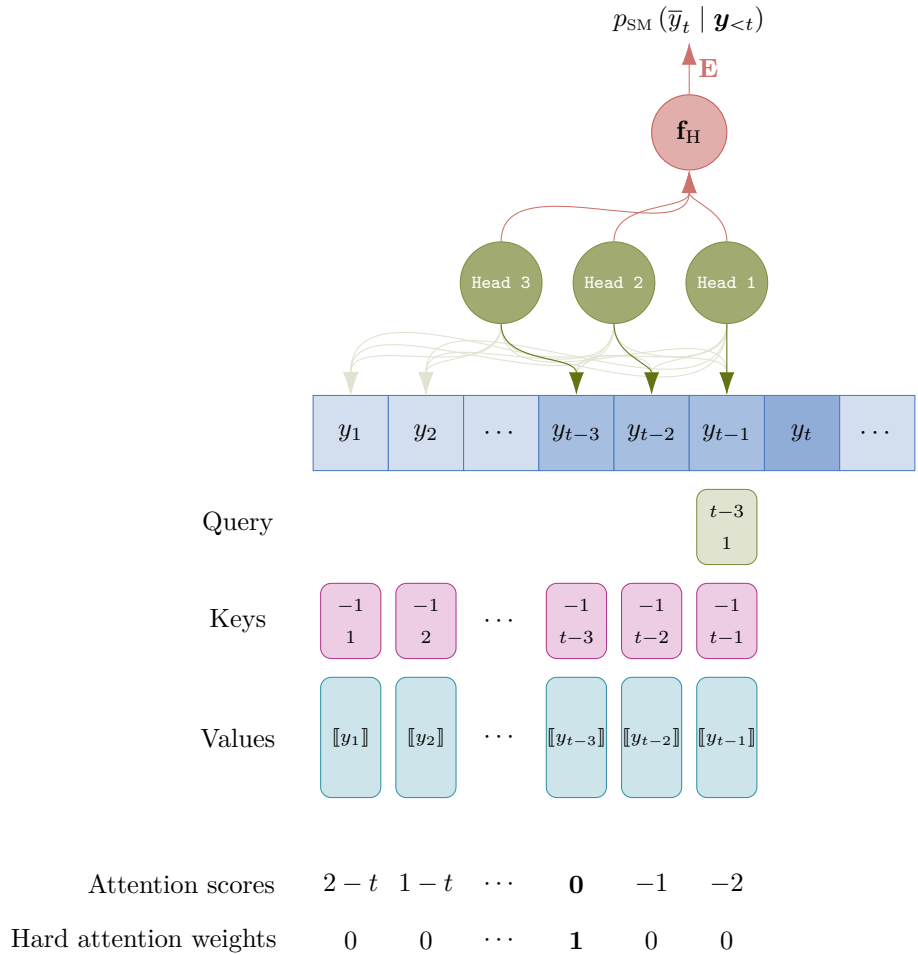
\begin{figure}
        \centering

        \begin{tikzpicture}[
            tape node/.style={draw=ETHBlue!80,minimum size=1cm,fill=ETHBlue!20},
            head node/.style={draw=ETHGreen!80,circle,minimum size=0.75cm,fill=ETHGreen!60,text=white},
            attn arrow/.style={-{Latex[length=3mm,width=2mm]},ETHGreen!100},
            comb arrow/.style={-{Latex[length=3mm,width=2mm]},ETHRed!70},
            comb node/.style={draw=ETHRed!80,circle,minimum size=1cm,fill=ETHRed!40},
            qvect/.style={
                    draw=ETHGreen!80,
                    fill=ETHGreen!20,
                    rounded corners,
                    minimum width=1cm,
                    minimum height=1cm
                },
            kvect/.style={
                    draw=ETHPurple!80,
                    fill=ETHPurple!20,
                    rounded corners,
                    minimum width=1cm,
                    minimum height=1cm
                },
            vvect/.style={
                    draw=ETHPetrol!80,
                    fill=ETHPetrol!20,
                    rounded corners,
                    minimum width=1cm,
                    minimum height=1cm
                },
            ]

            \node[text=black] at (-1.5,0.2) {Values};
            \draw[vvect] (-0.4,-0.6) rectangle (0.4,1) node[midway] {$\scriptstyle \onehot{\sym_1}$};
            \draw[vvect] (0.6,-0.6) rectangle (1.4,1) node[midway] {$\scriptstyle \onehot{\sym_2}$};
            \node[text=black] at (2,0.2) {$\cdots$};
            \draw[vvect] (2.6,-0.6) rectangle (3.4,1) node[midway] {$\scriptstyle \onehot{\sym_{\tstep - 3}}$};
            \draw[vvect] (3.6,-0.6) rectangle (4.4,1) node[midway] {$\scriptstyle \onehot{\sym_{\tstep - 2}}$};
            \draw[vvect] (4.6,-0.6) rectangle (5.4,1) node[midway] {$\scriptstyle \onehot{\sym_{\tstep - 1}}$};

            \node[text=black] at (-1.5,1.75) {Keys};
            \draw[kvect] (-0.4,1.25) rectangle (0.4,2.25) node[midway] {$\begin{array}{c} \scriptstyle -1 \\ \scriptstyle1 \end{array}$};
            \draw[kvect] (0.6,1.25) rectangle (1.4,2.25) node[midway] {$\begin{array}{c} \scriptstyle -1 \\ \scriptstyle 2 \end{array}$};
            \node[text=black] at (2,1.75) {$\cdots$};
            \draw[kvect] (2.6,1.25) rectangle (3.4,2.25) node[midway] {$\begin{array}{c} \scriptstyle -1 \\ \scriptstyle \tstep - 3 \end{array}$};
            \draw[kvect] (3.6,1.25) rectangle (4.4,2.25) node[midway] {$\begin{array}{c} \scriptstyle -1 \\ \scriptstyle \tstep - 2 \end{array}$};
            \draw[kvect] (4.6,1.25) rectangle (5.4,2.25) node[midway] {$\begin{array}{c} \scriptstyle -1 \\ \scriptstyle \tstep - 1 \end{array}$};

            \node[text=black] at (-1.5,3.05) {Query};
            \draw[qvect] (4.6,2.5) rectangle (5.4,3.5) node[midway] {$\begin{array}{c} \scriptstyle \tstep - 3 \\ \scriptstyle 1 \end{array}$};

            \node[text=black] at (-2.25,-1.75) {Attention scores};
            \node[text=black] at (0,-1.75) {$2 - \tstep$};
            \node[text=black] at (1,-1.75) {$1 - \tstep$};
            \node[text=black] at (2,-1.75) {$\cdots$};
            \node[text=black] at (3,-1.75) {$\mathbf{0}$};
            \node[text=black] at (4,-1.75) {$-1$};
            \node[text=black] at (5,-1.75) {$-2$};

            \node[text=black] at (-2.75,-2.5) {Hard attention weights};
            \node[text=black] at (0,-2.5) {$0$};
            \node[text=black] at (1,-2.5) {$0$};
            \node[text=black] at (2,-2.5) {$\cdots$};
            \node[text=black] at (3,-2.5) {$\mathbf{1}$};
            \node[text=black] at (4,-2.5) {$0$};
            \node[text=black] at (5,-2.5) {$0$};

            \foreach \i/\y in {0/$\sym_1$,1/$\sym_2$,2/$\cdots$,3/$\sym_{\tstep-3}$,4/$\sym_{\tstep-2}$,5/$\sym_{\tstep-1}$,6/$\symt$,7/$\cdots$} {
                    \ifnum \i=6
                        \node[tape node,fill=ETHBlue!50] (tape-\i) at (\i,4.25) {\y};
                    \else
                        \ifnum \i>2
                            \ifnum \i<6
                                \node[tape node,fill=ETHBlue!40] (tape-\i) at (\i,4.25) {\y};
                            \fi
                        \else
                            \node[tape node,fill=ETHBlue!20] (tape-\i) at (\i,4.25) {\y};
                        \fi
                        \ifnum \i>6
                            \node[tape node,fill=ETHBlue!20] (tape-\i) at (\i,4.25) {\y};
                        \fi
                    \fi
                }

            \node[head node] (head-1) at (2,6.25) {\scriptsize \texttt{Head 3}};
            \node[head node] (head-2) at (3.5,6.25) {\scriptsize \texttt{Head 2}};
            \node[head node] (head-3) at (5,6.25) {\scriptsize \texttt{Head 1}};

            \draw[attn arrow, ETHGreen!20] (head-1) to[out=270,in=90] (tape-0.north);
            \draw[attn arrow, ETHGreen!20] (head-1) to[out=270,in=90] (tape-1.north);
            \draw[attn arrow, ETHGreen!20] (head-1) to[out=270,in=90] (tape-4.north);
            \draw[attn arrow, ETHGreen!20] (head-1) to[out=270,in=90] (tape-5.north);
            \draw[attn arrow, ETHGreen!20] (head-2) to[out=270,in=90] (tape-0.north);
            \draw[attn arrow, ETHGreen!20] (head-2) to[out=270,in=90] (tape-1.north);
            \draw[attn arrow, ETHGreen!20] (head-2) to[out=270,in=90] (tape-3.north);
            \draw[attn arrow, ETHGreen!20] (head-2) to[out=270,in=90] (tape-5.north);
            \draw[attn arrow, ETHGreen!20] (head-3) to[out=270,in=90] (tape-0.north);
            \draw[attn arrow, ETHGreen!20] (head-3) to[out=270,in=90] (tape-1.north);
            \draw[attn arrow, ETHGreen!20] (head-3) to[out=270,in=90] (tape-3.north);
            \draw[attn arrow, ETHGreen!20] (head-3) to[out=270,in=90] (tape-4.north);
            \draw[attn arrow] (head-1) to[out=270,in=90] (tape-3.north);
            \draw[attn arrow] (head-2) to[out=270,in=90] (tape-4.north);
            \draw[attn arrow] (head-3) to[out=270,in=90] (tape-5.north);

            \node[comb node] (combiner) at (4.5,8.25) {$\tfheadCombine$};

            \draw[comb arrow] (head-1.north) to[out=90,in=270] (combiner.south);
            \draw[comb arrow] (head-2.north) to[out=90,in=270] (combiner.south);
            \draw[comb arrow] (head-3.north) to[out=90,in=270] (combiner.south);

            \node[fill=none] (out) at (4.5,9.75) {$\pLNSM\left(\eossym_\tstep\mid\str_{<\tstep}\right)$};

            \draw (combiner) edge[comb arrow, right] node{$\outMtx$} (out.south);

        \end{tikzpicture}
        \caption{A more complete illustration of the construction described in the proof for the case of the third head, \texttt{Head $3$}, based on \cref{fig:transformer-ngram}.
            Note that, among the three heads, only the \textcolor{ETHGreen}{query vector} (transformation) differs, while the key and value transformations are identical among the heads.}
        \label{fig:transformer-ngram-full}
    \end{figure}

\end{proof}

This proof establishes \emph{the only} ``concrete'' result on the (lower bound of the) expressivity for transformers in the form of \emph{model equivalence} (cf. \cref{def:model-equivalence}) that we know of.
In the next subsections, we discuss how transformer-based language models can simulate more complex formal models.
However, the simulation will not be as ``direct'' as the \ngram{} one, in the sense that we will have to work with modified alphabets which, as we noted above, results in a different notion of equivalence of models than what we have considered so far.
We will thus not model the conditional probabilities $\pLNSM\left(\sym\mid \str_{<\tstep}\right)$, but rather the probabilities over some more complex (but still finitely-many) objects, which will carry in them more information than just the generated symbol.
As we will discuss, this will be required due to the limited abilities of transformers to execute sequential operations compared to RNNs and classical language models, as hinted at in \cref{sec:rnn-parallelization}.

\iftoggle{publish-notes}
{}
{
    \subsubsection{Infinite-precision Transformers and Finite-state Languages}
    Having shown that transformers can perfectly represent at least strictly local subregular languages, we now start our ``standardized'' climb up the hierarchy of regular, context-free, and all computable languages.
    This is also the point where we depart from our stricter notion of equivalence discussed above.
    As we will see shortly, to be able to correctly simulate sequential processing of the classical language models, we will have to augment the alphabet of generated symbols with additional information, meaning that, from now on, we will talk about \emph{homomorphism equivalence} (cf. \cref{def:homomorphism-equivalence}).

    The central result of this subsection can be summarized by the following theorem.
    \begin{theorem}{Transformer language models can simulate probabilistic finite-state automata}{transformers-are-regular}
        Infinite-precision transformers can simulate probabilistic finite-state automata.
    \end{theorem}
    \begin{proof}
        \cref{thm:transformers-are-regular} presents a roughly analogous result to the fact that RNNs can simulate deterministic probabilistic weighted finite-state automata (cf. \cref{lem:minsky-constr}).
        This result, however, is in some sense stronger: it says that a transformer can simulate \emph{any} PFSA---even non-deterministic ones.\anej{We should add this to the RNN section as well}
        This, of course, comes with the caveat of the augmented alphabet and the corresponding homomorphism equivalence, which we discuss shortly.

        As always, we will prove the theorem by constructing, for a given PFSA $\wfsa= \wfsatuple$, a transformer model $\transformernetwork$ with the same weighted language, i.e., $\lang\left(\wfsa\right) = \lang\left(\transformernetwork\right)$.
        While the PFSA can in general have multiple initial states, we assume here, without loss of generality, that it has a single initial state.\anej{Hmmm but do we incur $\eps$ here? how do we handle that?}\footnote{As an exercise, you might think about how one can represent a WFSA with multiple initial states with one that has a single initial state.}

        We start by defining the structure of the hidden states (i.e., contextual representations of symbols) of the transformer, then discuss how those will be used to simulate the PFSA, and lastly discuss the concrete parameter settings which allow the transformer to carry out the actions defined by the PFSA.
        However, for all of that to be possible, we will require an augmented alphabet of symbols.

        \paragraph{An augmented alphabet of symbols.}
        To simulate a PFSA with a transformer, the transformer will work over an extended, but crucially still \emph{finite} alphabet of ``symbols''.\footnote{Some previous work also shows that transformers are Turing complete by showing that they can simulate an RNN by simply encoding the hidden states of the RNN in the generated symbols of the transformer \citep{bhattamishra-etal-2020-computational}. This, together with the Turing completeness of RNNs, would suffice for Turing completeness of transformers. However, notice that by generating RNN hidden states, the transformer model is no longer generating symbols from a \emph{finite} alphabet, meaning that, by our definition (and other standard definitions of a transformer) it is no longer a language model.}
        Let us consider why this might be required.
        As mentioned, transformers have no mechanism to \emph{pass} any notion of an inner state between the subsequent symbols in the string (apart from the notion of passing the state between layers---however, there is always a finite number of those).
        Importantly, the inner state of a machine (e.g., the state in a finite-state automaton, the configuration in a pushdown automaton, or the hidden state in an RNN) is \emph{not} in any direct correspondence to the current symbol, as it depends on all the previously read symbols in the string.
        This is where the modification comes in: What if we encoded the inner state of the machine directly \emph{in the output symbol} itself?
        This would then allow a model to access it simply from the raw context $\str_{<\tstep}$ itself, without having to keep any latent variable in the inner state.
        This is the main idea behind using a modified version of the automaton's alphabet to simulate it using a transformer: we will \emph{expose} the machine's current state $\stateq_\tstep$ into the generated symbol so that the transformer will be able to access it in the next time step with the attention mechanism which can always look at the entire history so far.
        The alphabet whose Kleene closure we will therefore be modeling with the transformer is
        \begin{equation}
            \extAlphabet \defeq \states \times \alphabet.
        \end{equation}
        That is, the transformer will generate (or model conditional distributions of) individual states of and symbols read by the PFSA.
        This will allow it to keep track of both the symbols read or generated by the PFSA as well as its computation steps (even in the non-deterministic case).

        \paragraph{Transformer hidden states.}
        From a very high level, we will encode the possible ``configurations'' of the PFSA in the transformer's hidden states as follows:
        \begin{equation} \label{eq:transf-wfsa-state-abstract}
            \hiddStatet =
            \begin{pmatrix}
                \onehot{\stateq_\tstep} \\
                \onehot{\eossym_\tstep} \\
                \vdots                  \\
                \begin{array}{@{}c@{}}
                    \textbf{Control variables}   \\
                    \vdots                       \\
                    \textbf{Positional encoding} \\
                \end{array}
            \end{pmatrix},
        \end{equation}
        where $\stateq_\tstep$ refers to the state $\wfsa$ is in at time $\tstep$ and $\eossym_\tstep$ the input symbol it is reading.
        $\onehot{\cdot}$ refers to the one-hot encoding function of the appropriate dimensionality, i.e., $\onehot{\stateq} \in \B^{|\states|}$ and $\onehot{\eossym} \in \B^{|\eosalphabet|}$.
        We will again use the orderings of the sets $\states \times \alphabet$ (ordering $\ordering$), $\eosalphabet$ (ordering $\eossymordering$), and $\states$ (ordering $\stateordering$).
        Notice that the first two elements together exactly correspond to the elements of the extended alphabet $\extAlphabet$.

        Lastly, the positional encodings will be of the form
        \begin{equation}
            \posEncFun{\tstep} \defeq
            \begin{pmatrix}
                1                    \\
                \tstep + 1           \\
                \frac{1}{\tstep + 1} \\
                \frac{1}{\left(\tstep + 1\right)^2}
            \end{pmatrix}.
        \end{equation}
        The need for the constant $1$ will become apparent later.

        \paragraph{Modeling sequentiality with a transformer.}
        As mentioned above, transformers, unlike RNNs, do not model their states sequentially, meaning that their transitions are harder to analyze by comparing them to automata.
        However, by considering what the attention mechanism does and looking at transformers as \emph{sequence models}, we can make the connection to sequential machines more apparent.
        Recall from \cref{def:transformer-plnsm} that the transformer computes the probability of the next token $\eossym_\tstep$ given the context $\strlt$ as
        \begin{equation}
            \pLNSM\left(\eossym_\tstep\mid \strlt\right) = \projfuncEosalphabetminusFunc{\outMtx \hiddStatet}_{\eossym_\tstep},
        \end{equation}
        where the hidden state $\hiddStatet$ is the hidden state computed for the position $\tstep$ by the attention block.
        Importantly, during generation, given (all) the symbols $\strlt$, we can always compute $\pLNSM\left(\eossym_\tstep\mid \strlt\right)$, i.e., the distribution over the next symbol, and thus generate it.
        This is done \emph{one symbol at a time}---\emph{this} presents the analogy in the transformer model to how automata work and such sequential generation can never be parallelized, as discussed in \cref{sec:rnn-parallelization}.
        We will, therefore, use the generative steps of a transformer model to simulate the transitions of an automaton one transition at a time.
        Note, however, that the transformer still does not keep any notion of a hidden state that is \emph{passed} through the generation steps, as is the case of an RNN (cf. \cref{eq:rnn-hidden-state-update}).
        Therefore, \emph{all} the information to compute the next state of the automaton (i.e., to simulate the transition) has to be present in the so-far \emph{generated sequence} itself.
        In the case of simulating finite-state automata, this can be done with the augmented alphabet described above quite easily---we will describe the details shortly.
        However, this notion of storing the information about the configuration of the simulated machine in the generated string so far will be especially important once we extend this construction and start working with potentially arbitrarily large amounts of information, for example, encoded by a stack.

        Coming back to the structure of the transformer, our goal in this proof is to construct a model which can sequentially produce hidden states of the form \cref{eq:transf-wfsa-state-abstract}.
        However, before we do that, let us consider why keeping hidden states of the form from \cref{eq:transf-wfsa-state-abstract} is \emph{enough} to capture the distribution defined by the PFSA.
        In other words, we want to show that, if we can design a transformer that manages to generate symbols from the extended alphabet $\extAlphabet$ correctly, we are guaranteed to be able to represent $\wfsa$'s distribution.
        We show this in the following lemma.
        \begin{lemma}{}{}
            Suppose that a transformer $\transformernetwork$ defines the hidden states of the form of  \cref{eq:transf-wfsa-state-abstract} where $\stateq_\tstep$ refers to the state $\wfsa$ is in at time $\tstep$ and $\eossym_\tstep$ the input symbol it is reading.
            Then, we can define a transformation $\vfunc_\textnormal{out}$ of the transformer hidden states and a symbol representation (\emph{output}) matrix $\outMtx \in \Rex^{|\eosalphabet| \times |\states||\eosalphabet|}$ where $\vfunc_\textnormal{out}\left(\hiddStatet\right) \in \R^{|\states||\eosalphabet|}$.
        \end{lemma}
        \begin{proof}
            The idea is to, based on the hidden state $\stateq_\tstep$ and the read symbol $\eossym_\tstep$ encoded in $\hiddStatet$, one-hot encode the pair ``separately'', and construct a lookup matrix $\outMtx$ in a similar way to how we the Minsky construction in \cref{lem:minsky-constr} and the \ngram{} construction in the proof of \cref{thm:transformers-ngrams}.
            Notice that the hidden state always contains the information to do that: We can always extract the one-hot encodings of the state and the symbol and then combine them into a one-hot encoding of the pair $\stateq_\tstep, \eossymt$ by implementing the \texttt{AND} function---this is completely analogous to the construction of the one-hot encoding of an \ngram{} in the proof of \cref{thm:transformers-ngrams}.
            This is done by the transformation $\vfunc_\textnormal{out}$.\footnote{This does represent a slight departure from the standard definition of a transformer language model, in which the \emph{hidden state directly} is used, together with $\outMtx$, to determine $\pLNSMFun{\eossymtplus}{\strlet}$.
                For simplicity, we allow ourselves this departure.
                However, note that this transformation could easily also be done by an application of an additional transformer block, in which the attention mechanism would mostly just copy the values over and the output transformation $\fTransf$ would perform the conjunction.}

            After transforming $\hiddStatet$ into the one-hot encoding of the state-symbol pair, we can look up the logits of the next-state distribution in the representation matrix $\outMtx$.
            This is again analogous to the construction in \cref{thm:transformers-ngrams}.
            The matrix embeds the symbols $\eossym \in \eosalphabet$ and thus enables the computation of the softmax in the definition of a transformer sequence model (cf. \cref{def:transformer-plnsm}).
            More formally, for $\sym \in \alphabet$, we define
            \begin{equation}
                \outMtx_{\eossymordering\left(\sym\right) \ordering\left(\statep, \syma\right)} \defeq
                \begin{cases} \log w  & \mid \textbf{ if } \edge{\statep}{\syma}{w}{\circ} \in \trans \\
              -\infty & \mid \textbf{ otherwise }\end{cases}
            \end{equation}
            and for $\eos$, we define
            \begin{equation}
                \outMtx_{\eossymordering\left(\eos\right) \ordering\left(\statep, \syma\right)} \defeq
                \begin{cases} \log \finalf\left(\stateq\right) & \mid \textbf{ if } \finalf\left(\stateq\right) > 0 \\
              -\infty                          & \mid \textbf{ otherwise }\end{cases}
            \end{equation}
            \Anej{TODO: finish}
            We can show that this ensures that the probabilities assigned to strings by the transformer language model match those assigned by the PFSA by simply multiplying the conditional probabilities of the transformer defined above and checking that they match the string acceptance weights assigned by $\wfsa$.
        \end{proof}

        Having shown that hidden states of the form from \cref{eq:transf-wfsa-state-abstract} suffice, let us now outline the entire architecture of the transformer simulating $\wfsa$.
        As mentioned, by outputting state-symbol pairs, the model can always access the current state and the read symbol of the simulated automaton in the previous generated symbol.
        The application of the transformer layer, therefore, has to ``extract'' this information from the entire generated string (by attending to the previous symbol) and then simulate the transition weighted transition function of the PFSA.
        This will be done in two stages---this corresponds to the transformer having two \emph{layers}.
        The first task is very similar to what we did in the \ngram{} case: the transformer simply needs one head to attend to the previous generated symbol in the string.\anej{consider changing ``generated symbol'' to ``last symbol in the context''}
        This can be done easily by using the same positional encodings and
        The model will have two layers.

    \end{proof}

    \subsubsection{Turing Completeness of Transformers}

    \paragraph{Turing completeness of two-stack pushdown automata.}

    \begin{theorem}{Turing completeness}{}
        Transformer language models are Turing complete.
    \end{theorem}

    \begin{lemma}{}{}
        Transformer language models can simulate probabilistic two-stack pushdown automata.
    \end{lemma}

    Just like in the RNN case in \cref{sec:rnn-turing-completeness}, we will start by proving a weaker claim whose proof is simpler, but conceptually completely the same---the fact that transformers can simulate deterministic single-stack pushdown automata.
    We will then generalize the construction to the two-stack case.

    \paragraph{Transformer hidden states.}
    From a very high level, we will encode the transformer's hidden states as follows:
    \begin{equation}
        \hiddStatet =
        \begin{pmatrix}
            \onehot{\stateq_\tstep}         \\
            \onehot{\stacksym_\tstep}       \\
            \actionToNum{\pdaAction_\tstep} \\
            \vdots                          \\
            \begin{array}{@{}c@{}}
                \textbf{Control variables} \\
                \vdots                     \\
                \textbf{Positional encoding}
            \end{array}
        \end{pmatrix}
    \end{equation}
    More concretely, the positional encodings will be of the form
    \begin{equation}
        \posEncFun{\tstep} \defeq
        \begin{pmatrix}
            1                    \\
            \tstep + 1           \\
            \frac{1}{\tstep + 1} \\
            \frac{1}{\left(\tstep + 1\right)^2}
        \end{pmatrix}.
    \end{equation}
    The need for the constant $1$ will become apparent later.

    \paragraph{Two memory structures.}
    The construction will work with two ``memory structures'': the \defn{sequence} of the generated ``symbols'' so far and the \defn{stack} of the simulated automaton.
    Distinguishing and finding the relationship between them (in the sense of how the sequence of generated symbols encodes the stack) will be the main challenge of the construction.
    The sequence of generated symbols plays an analogous role to the \emph{context} of the transformer sequence model as defined in \cref{def:transformer-plnsm}
    Conceptually, we can imagine the stack being represented on an infinite \emph{tape} (infinite in one direction), whose cells have \emph{indices} through which we can look up the values of the stack.
    See \cref{fig:transformer-stack}\anej{add} for an abstract depiction.
    Note that we will only ever access the \emph{top} of the stack---we will only use the indexing to make the construction and manipulation of the stack through the transformer hidden states easier.

    \begin{lemma}{}{}

    \end{lemma}

    \paragraph{Turing completeness of two-stack pushdown automata.}

    \subsubsection{The Nuances of Transformer Computational Power}

    \begin{itemize}
        \item Importance of positional encodings
        \item Difference between HAT and AHAT
        \item The difficulty to put them onto the Chomsky Hierarchy
              \begin{itemize}
                  \item Hahn's construction
                  \item Satwik counter construction
              \end{itemize}
    \end{itemize}

    This concludes our investigation of the computational capacity of transformers.
    While this does not finish our theoretical investigation of language models, it is the last result on the computational complexity of language models.
    We, therefore, summarize the results covered in the notes in \cref{fig:model-expressivity-plot}, where we pictorially represent the computational capacity of all the different language models we considered.

    \paragraph{Transformers and First Order Logic}
    \Clemente{\url{https://docs.google.com/document/d/1X2M5MnwGlkVsx7ntJ1la4Dab8H7eheFzAOkMqQmqtfQ/edit?usp=sharing}}

    \begin{figure}
        \centering

        \begin{tikzpicture}
            \draw[-{Latex}] (0,0) -- (10,0) node[below] {Time};
            \draw[-{Latex}] (0,0) -- (0,9) node[above, rotate=90] {Expressivity};
            \node[rotate=60] at (-1,1) {Strictly Local};
            \node[rotate=60] at (-1,3) {Regular};
            \node[rotate=60] at (-1,5) {Context-free};
            \node[rotate=60] at (-1,7) {Computable};
            \node[circle, fill=ETHPetrol, inner sep=2pt, label={[rotate=20]right:\ngram{} LMs}] at (2,1) {};
            \node[circle, fill=ETHPetrol, inner sep=2pt, label={[rotate=20]right:PFSA LMs}] at (3,3) {};
            \node[circle, fill=ETHPetrol, inner sep=2pt, label={[rotate=20]right:Context-Free LMs}] at (4,5) {};
            \node[circle, fill=ETHPetrol, inner sep=2pt, label={[rotate=20]right:Pushdown LMs}] at (5,7) {};
            \node[circle, fill=ETHPetrol, inner sep=2pt, label={[rotate=20]right:$\B$-RNN LMs}] at (6,3) {};
            \node[circle, fill=ETHPetrol, inner sep=2pt, label={[rotate=20]right:$\Q$-RNN LMs}] at (6.5,7) {};
            \node[circle, fill=ETHPetrol, inner sep=2pt, label={[rotate=20]right:$\Q$-Transformers}] at (8,1) {};
        \end{tikzpicture}

        \caption{A pictorial representation of the expressive power of various language model architectures.}
        \label{fig:model-expressivity-plot}
    \end{figure}
}




\printindex


\bibliography{reference}
\bibliographystyle{acl_natbib}

\end{document}